\documentclass[10pt]{report}

\usepackage{thesis}

\usepackage{amsmath}
\usepackage{amssymb}
\usepackage[utf8]{inputenc}
\usepackage{url}
\usepackage[pdftex]{graphicx}
\usepackage{rotating}
\usepackage[titletoc,toc,title]{appendix}
\usepackage{titlesec}
\usepackage{titletoc}
\usepackage{listings}
\usepackage{color}

\usepackage{morefloats}
\usepackage{adjustbox}
\usepackage{setspace}

\usepackage{chapterbib}
\usepackage{natbib}

\usepackage{tabularx}
\usepackage{longtable}
\usepackage{multirow}
\usepackage[makeroom]{cancel}
 \usepackage{pdfpages}
\usepackage[normalem]{ulem}

\usepackage{csquotes}

\usepackage{changepage}
\newcommand{\pnastableadjust}{0}
\newcommand{\revtexlatexswitch}[2]{#2}
\newcommand{\etal}{\textit{et al.}}

\allowdisplaybreaks
\DeclareFixedFont{\ttb}{T1}{txtt}{bx}{n}{8} 
\DeclareFixedFont{\ttm}{T1}{txtt}{m}{n}{8}  
\usepackage{color}
\definecolor{deepblue}{rgb}{0,0,0.5}
\definecolor{deepred}{rgb}{0.6,0,0}
\definecolor{deepgreen}{rgb}{0,0.5,0}

\usepackage{listings}
\newcommand\pythonstyle{\lstset{
language=Python,
basicstyle=\ttm,
otherkeywords={self},             keywordstyle=\ttb\color{deepblue},
emph={MyClass,__init__},          emphstyle=\ttb\color{deepred},    stringstyle=\color{deepgreen},
frame=tb,                         numbers=none,
showstringspaces=false            }}
\lstnewenvironment{python}[1][]
{
\pythonstyle
\lstset{#1}
}
{}

\newcommand{\havgfn}{h_{\rm avg}}

\titlecontents{section}
              [3.8em]
              {}
              {\contentslabel{2.3em}}
              {\hspace*{-2.3em}}
              {\titlerule*[1pc]{.}\contentspage}

\titlecontents{subsection}
              [6.0em]
              {}
              {\contentslabel{2.9em}}
              {\hspace*{-2.3em}}
              {\titlerule*[1pc]{.}\contentspage}
              
\title{Towards a science of human stories:
using sentiment analysis and emotional arcs to understand
the building blocks of complex social systems}
\author{Andrew J. Reagan}
\defensedate{February 24, 2017}
\dissertation                             \doctorphilosophy                      \masci                                 \maygrad                            
\advisor{Christopher M. Danforth, Ph.D.}
\readerone{Peter Sheridan Dodds, Ph.D.}
\readertwo{Samuel V. Scarpino, Ph.D.}
\chair{Jacques A. Bailly, Ph.D.}
\dean{Cynthia J. Forehand, Ph.D.}

\begin{document}

\maketitle

\begin{abstract}
  We can leverage data and complex systems science to better understand society and human nature on a population scale through language
  --- utilizing tools that include sentiment analysis,
  machine learning,
  and data visualization.
  Data-driven science and the sociotechnical systems that we use every day are enabling a transformation from
  hypothesis-driven, reductionist methodology
  to complex systems sciences.
  Namely, the emergence and global adoption of social media has rendered possible the real-time estimation of population-scale sentiment,
  with profound implications for our understanding of human behavior.
  Advances in computing power, natural language processing, and digitization of text now make it possible to study a culture's evolution through its texts using a ``big data'' lens.

  Given the growing assortment of sentiment measuring instruments,
it is imperative to understand which aspects of sentiment dictionaries contribute to both their
classification accuracy
and
their ability to provide richer understanding of texts.
Here, we perform detailed, quantitative tests and qualitative assessments of 6 dictionary-based methods
applied to 4 different corpora, and briefly examine a further 20 methods.
We show that while inappropriate for sentences, dictionary-based methods are generally robust in their classification accuracy for longer texts.
Most importantly they can aid understanding of texts
with reliable and meaningful word shift graphs if
(1)
the dictionary covers a sufficiently large enough
portion of a given text's lexicon when weighted by word usage frequency;
and
(2) 
words are scored on a continuous scale.

Our ability to communicate relies in part upon a shared emotional experience, with stories often following distinct emotional trajectories, forming patterns that are meaningful to us.
By classifying the emotional arcs for a filtered subset of 4,803 stories from Project Gutenberg's fiction collection, we find a set of six core trajectories which form the building blocks of complex narratives.
We strengthen our findings by separately applying optimization, linear decomposition, supervised learning, and unsupervised learning.
For each of these six core emotional arcs, we examine the closest characteristic stories in publication today and find that particular emotional arcs enjoy greater success, as measured by downloads.
Within stories lie the core values of social behavior, rich with both strategies and proper protocol, which we can begin to study more broadly and systematically as a true reflection of culture.
Of profound scientific interest will be the degree to which we can eventually understand the full landscape of human stories, and data driven approaches will play a crucial role.

Finally, we utilize web-scale data from Twitter to study the limits of what social data can tell us about public health, mental illness, discourse around the protest movement of \#BlackLivesMatter, discourse around climate change, and hidden networks.
We conclude with a review of published works in complex systems that separately analyze charitable donations, the happiness of words in 10 languages, 100 years of daily temperature data across the United States, and Australian Rules Football games.

\end{abstract}

\begin{citationspage}

\noindent Material from this dissertation has been submitted for publication in \textit{EPJ Data Science} on January 10, 2017, with the preprint available as follows:\\

\noindent Reagan, A. J., Danforth, C. M., Tivnan, B., Williams, J. R., Dodds, P. S.. (2016).
Benchmarking sentiment analysis methods for large-scale texts: A case for using continuum-scored words and word shift graphs.
\textit{Preprint available online at \href{https://arxiv.org/abs/1512.00531}{https://arxiv.org/abs/1512.00531}}.\\

\begin{center}
{\large AND}\\
\end{center}

\noindent Material from this dissertation has been published in \textit{EPJ Data Science} on November 4, 2016, in the following form.
Within a month of publication, our paper was the most shared of all papers published in \textit{EPJ Data Science}, as measured by Altmetric.\\

\noindent Reagan, A. J., Mitchell, L., Kiley, D., Danforth, C. M., Dodds, P. S.. (2016).
 The emotional arcs of stories are dominated by six basic shapes.
 \textit{EPJ Data Science 5}(1), 31.
\end{citationspage}

\tableofcontents

\mainmatter
\sloppy

\doublespacing


\chapter{Introduction and Literature Review}

\begin{quote}
  Epoch watch: Welcome to the Sociotechnocene.\\
  -@peterdodds \href{https://twitter.com/peterdodds/status/156899358573465601}{2012-01-10}
\end{quote}

\section{Introduction}

Individual words encapsulate information and emotion as the building blocks that we use to capture experiences in stories.
Beyond words, multi-word expressions (phrases), conceptual metaphor, and complicated grammar (syntax) coalesce to provide incredible expressive power.
Attempts to quantify semantic content build atop syntactic understand of language with the aim of transforming a model of meaning that has proven useful to our own cognitive machinery into something more readily applicable for another purpose (e.g., summarization by a computer).
One such goal of semantic understanding is to measure the sentiment expressed in written communication, which is broadly known as sentiment analysis.
The next evolution of natural language systems will tackle the harder-yet problems of pragmatics, where narrative understanding and generation can enable common-sense reasoning on par with human intuition.

In our work, we transfer the emotion of single, isolated words into a one-dimensional happiness measure to build the Hedonometer.
Leveraging the Hedonometer technology and modern computational power, we analyze digitized text with the ultimate goal of understanding stories.
This dissertation proceeds as follows: in this chapter we explore the foundations of sentiment analysis and narrative structure.
In Chapter~\ref{chap:benchmarking} we benchmark and compare methods for sentiment analysis.
In Chapter~\ref{chap:emotional-arcs} we apply these methods and extract dominant emotional arcs from digitized text.
In Chapter~\ref{chap:contributions}, we discuss contributions made to published work in the broader science of complex systems.
Finally, in Chapter~\ref{chap:conclusion} we offer some concluding remarks.

Next, we examine prior work in natural language processing, sentiment analysis, and computational narrative understanding.

\section{Sentiment analysis}

The field of Natural Language Processing (NLP) has been around since the advent of computers, but is growing rapidly alongside computational advances.
While major advances have been made, there remain many open problems.
We focus here on a specific NLP problem, namely understanding the emotional content of language.
We refer to the emotional content in a written text broadly as the sentiment.
In addition to the summaries given in recent review articles \citep{giachanou2016like}, the landscape of tools and technologies is expanding quickly and sentiment analysis systems are deployed to tackle important challenges.
As we will see, sentiment analysis is a sub-field of NLP that can benefit from advancement in other realms of NLP as well (e.g., phrase partitioning).

Applications of sentiment analysis span academia, industry, and government.
Just some of the current uses include predicting elections \citep{tumasjan2010predicting}, product sales \citep{liu2007arsa}, stock market movement \citep{bar2011identifying}, and tracking public opinion \citep{cody2015climate}.
NLP and measures of sentiment are used to analyze consumption of information from the media, and societal level decisions are driven by this flow of public opinion online.
Beyond individual and collective decisions, corporate success demands an understanding of the public sentiments directed towards their products.

Advances in Artificial Intelligence (AI) have elucidated the distinction between problems that are hard for computers and those that are hard for humans---a difference that is not obvious at the outset.
Determining sentiment is one such task: understanding the sentiment of our friends and colleagues through informal text is relatively easy for us, but it is hard to codify in a computer algorithm.
As we will see, machine learning (often broadly referred to as AI) is finding uses in all areas of Natural Language Processing (NLP), including advancing the state-of-the-art in sentiment classification and sentiment dictionary creation.
While sentiment analysis benefits from machine learning to create classifiers and sentiment dictionaries, the output of sentiment detection also aids higher level approaches to language understanding.

\subsection{Psychology of emotion}
\label{subsec:psychology}

With few exceptions, current sentiment analysis methods aim to detect sentiment one-dimensionally, giving a score on a range from negative to positive sentiment.
While this pragmatic approach proves useful, \cite{jack2014dynamic} conjectured that there are four basic emotions, \cite{ekman1992argument} names six, and \cite{plutchik1991emotions} identifies two additional basic emotions in humans.
These theories are only the most well known classifications, with at least 90 such classifications being given over the past century, as noted by \cite{plutchik2001nature}.
Through the use of brain imaging and fMRI techniques, researchers in neuroscience have also attempted to distinguish whether basic emotions are best captured as discrete categories (anger, fear) or underlying dimensions (valence, arousal).
Altogether they have found consistent neural locations for basic emotions but no one-to-one mapping, and further research is still needed \citep{harrison2010embodiment,hamann2012mapping}.

\begin{figure}[tbp!]
  \centering
  \includegraphics[width=0.48\textwidth]{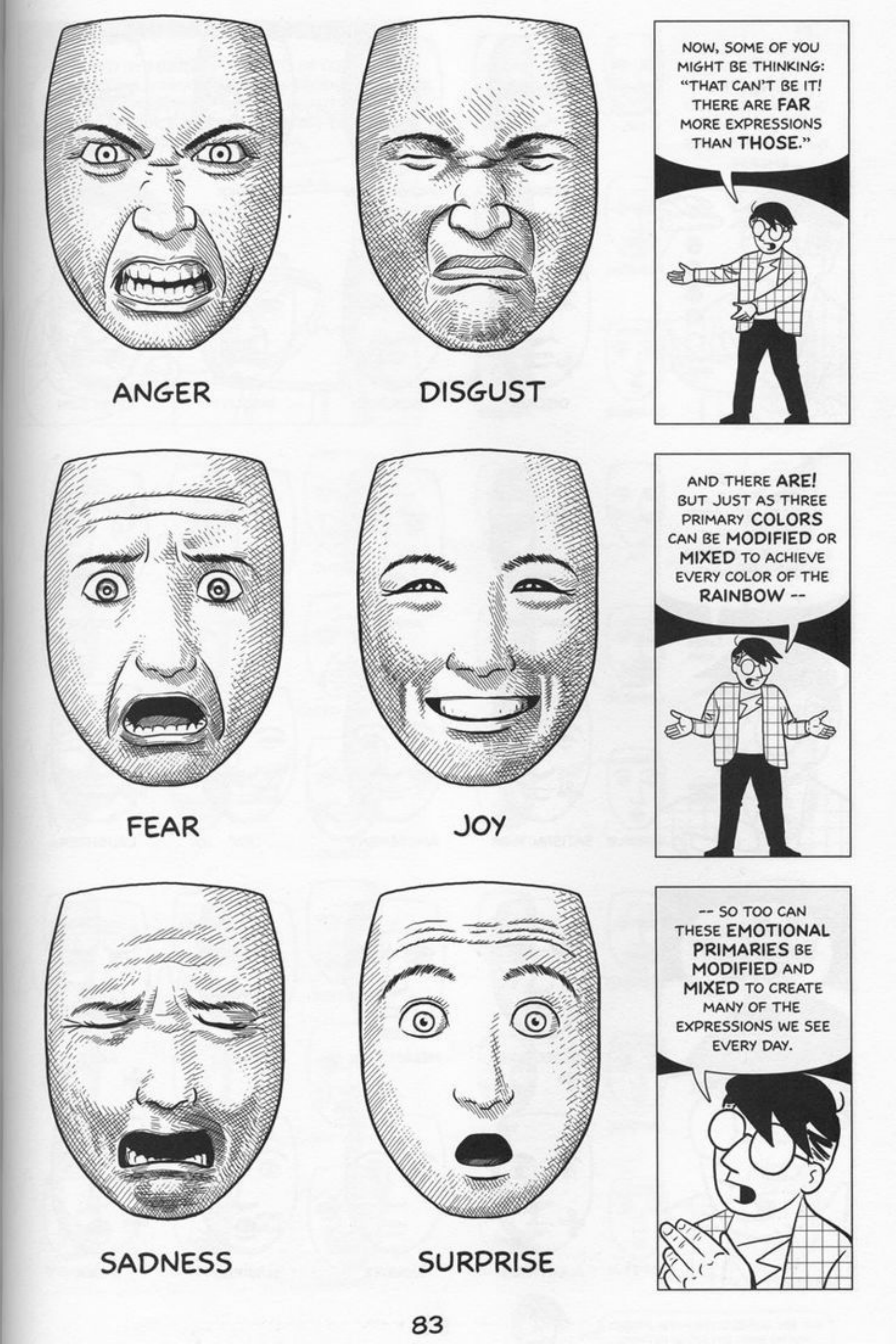}
  \caption[]{
    The six emotions of \cite{ekman1992argument}, illustrated here by \cite{mccloud2006making}.
    In principle, the entire range of human emotions can be constructed from this minimal ``basis'', e.g., the emotion \textit{delight} is the addition of \textit{joy} and \textit{surprise}.
    This theory of basic emotions distinguishes these emotions as being fundamentally distinct, adapted for fundamental life tasks, and universally present through evolution (or, perhaps, universal social learning).
    In particular the distinction between basic emotions is not explained by variation in dimensions of arousal, pleasantness, or activity.
  }
  \label{fig:ekman}
\end{figure}

\begin{figure}[tbp!]
  \centering
  \includegraphics[width=0.48\textwidth]{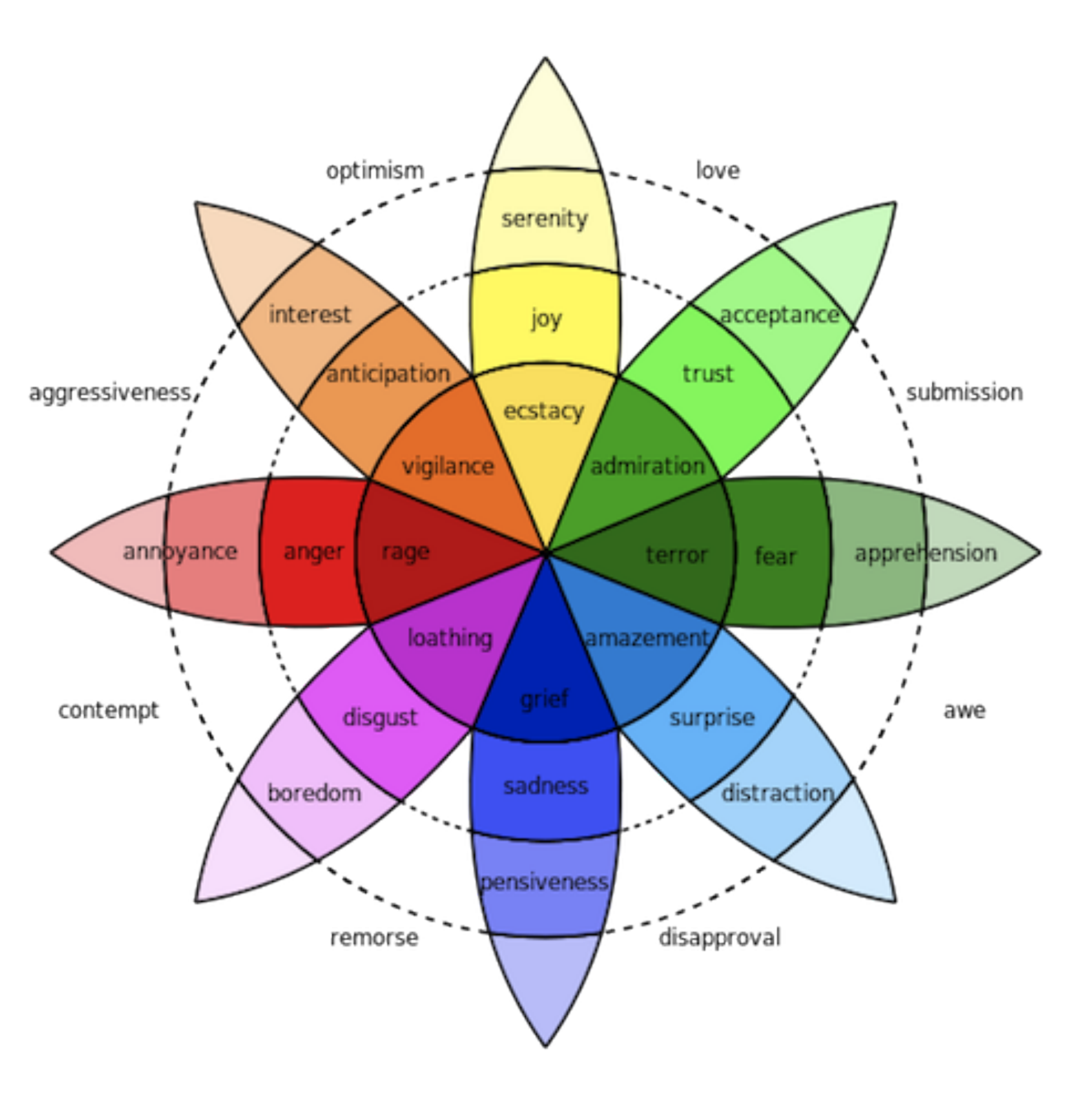}
  \caption[]{
    Schematic of the eight emotions from \cite{plutchik1991emotions}.
    The commonly known eight names (e.g., joy, etc.) are one row out from the center.
    In addition to the six emotions of \cite{ekman1992argument} we find \textit{anticipation} and \textit{trust} on the first level.
  }
  \label{fig:plutchik}
\end{figure}

The widely acknowledged six basic emotions identified by Paul Eckman are:
{
\begin{itemize}
\item \textit{happy},\vspace{-12pt}
  \item \textit{surprised},\vspace{-12pt}
  \item \textit{afraid},\vspace{-12pt}
  \item \textit{disgusted},\vspace{-12pt}
  \item \textit{angry},\vspace{-12pt}
  \item and \textit{sad}.
\end{itemize}
}
In Figure~\ref{fig:ekman}, a visualization of these six basic emotions is shown.
As noted in the caption, these six emotions serve as a basis for more complex emotions.
The eight basic emotions of \cite{plutchik1991emotions} are shown as the variations along four dimensions in Figure~\ref{fig:plutchik}.
While we do not expect that each of the six basic emotions have orthogonal representations in their embodiment in language (as orthogonality may be inferred from the Figures, is found in facial expression, and underlies the theory), a basis of more than a single dimension is likely necessary to represent the full range of emotion.
The basic emotions theory rejects that all emotions can be represented as either positive of negative states, and this should extend to language.
Indeed, attempts to cast the basic emotions as either positive (e.g., happy) or negative (e.g., sad) are subjective, e.g. by \cite{robinson2008brain} classifying \textit{pride} as a negative emotion.
According to \cite{ekman1992argument}, basic emotions are distinguished by nine characteristics:
\begin{enumerate}
  \item Distinctive universal signals.\vspace{-12pt}
  \item Presence in other primates.\vspace{-12pt}
  \item Distinctive physiology.\vspace{-12pt}
  \item Distinctive universals in antecedent events.\vspace{-12pt}
  \item Coherence among emotional response.\vspace{-12pt}
  \item Quick onset.\vspace{-12pt}
  \item Brief duration.\vspace{-12pt}
  \item Automatic appraisal.\vspace{-12pt}
  \item Unbidden occurrence.
\end{enumerate}

\begin{figure}[tbp!]
  \centering
  \includegraphics[width=0.48\textwidth]{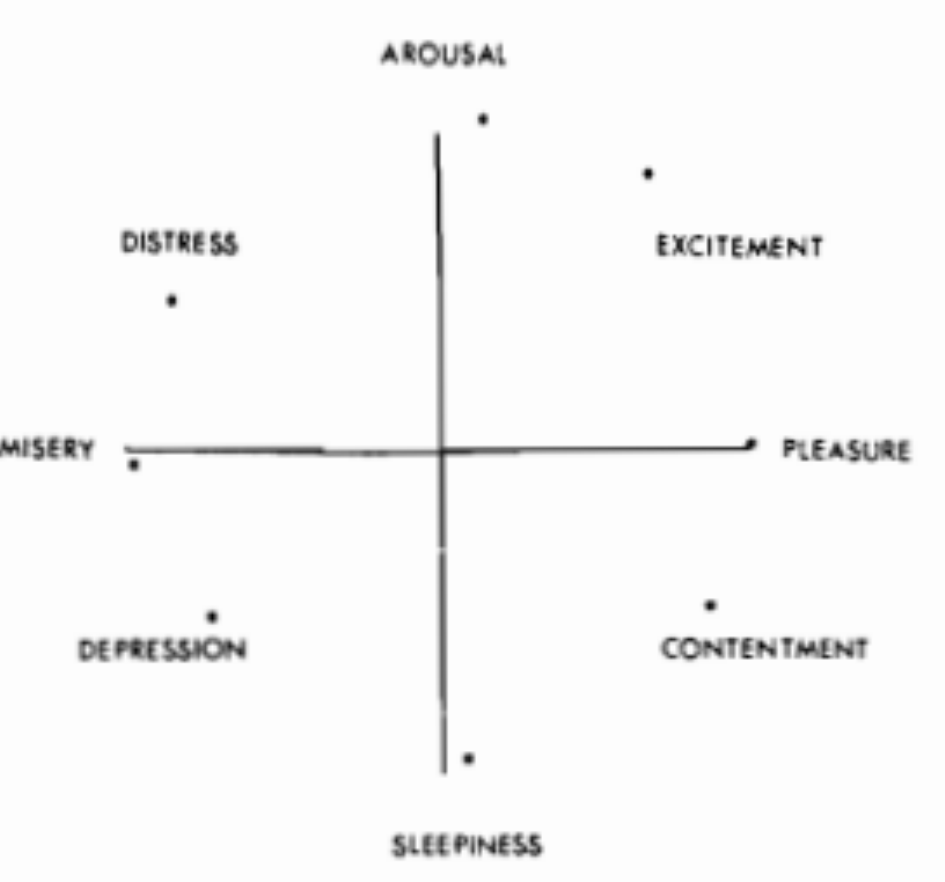}
  \caption[]{
    Eight emotions on the arousal--pleasure axis of \cite{russell1980circumplex}, who finds these axis to be the best representation of emotion.
    To this end, using 28 emotional words manually annotated for the characteristics which they share, Russell finds the two major principal components in a Principal Component Analysis, establishing this ``circular ordering.''
    This circular ordering agrees well with the mental model of emotional states used by psychologists at the time.
  }
  \label{fig:russell}
\end{figure}

To this end, in Figure~\ref{fig:russell} the theory of \cite{russell1980circumplex} attempts to find the core dimensions of emotion using data from emotions manually labelled for 28 adjectives.
The explained variance by the first two principal components would provide an indication of how well we can capture emotion with two abstract dimensions, however this is not provided by \cite{russell1980circumplex}.
Each of these theories expands upon the single dimension considered further in sentiment analysis: positive and negative.
More complex emotions can be constructed from combinations of the basic emotions (e..g., \verb|delight = joy + surprise|), which is not possible from combinations of simply positive and negative states (e.g., it would be nonsensical to find coefficients $a,b$ for the abstract categories positive and negative to satisfy \verb|delight = a*positive + b*negative|).

An alternative to basic, discrete emotions being the building blocks for all emotions is to place all emotions in the dimensions of \textit{valence}, \textit{arousal}, and \textit{dominance}, often referred to as ``norms'' and measured alongside \textit{concreteness} and \textit{age of acquisition} \citep{lindquist2016language}.
In the literature, the term \textit{valence} is used interchangeably to mean the negative/positive emotional dimension.

The positivity bias in language is frequency-independent, as long as the frequency selections are rank ordered (see \cite{dodds2015human} and Chapter~\ref{chap:contributions}).
\cite{schrauf2004preponderance} asked participants to write as many emotion words as they could think of in two minutes, and found that participants were able to recall a larger list of negative emotional words.
At least one theory for this difference, as elaborated in \cite{koch2016general}, posits that this difference is because positive words are more similar than negative words.
In one of six tests, they show that the scores for positive words are more tightly clustered than the scores for negative words from the Warriner \& Kuperman sentiment dictionary.

In addition to the emotion of expression, we note that other work attempts to measure personality traits of individuals based on their expressions (rather than the sentiment of the expressions themselves), specifically \cite{kosinski2013private} and \cite{youyou2015computer}.
As an example, given a person's micro-blog post, the algorithms developed by \cite{kosinski2013private} are trained to measure whether the person is an introvert or extrovert.
These attempts fundamentally differ from sentiment analysis by measuring traits of an individual rather than traits of the expression, though in practice the two goals make use of similar machine learning techniques.

For the remainder of this chapter, we will assume that each emotion is being measured on a scale from -4 $\rightarrow$ 4, with 0 representing no presence of emotion and a score of -4/4 representing the maximum negative/positive emotional priming.
While some dictionaries benefit from considering emotion on a different scale for human evaluation (e.g. ``labMT'' with 1 $\rightarrow$ 9 or ``AFINN'' with $-5\to5$), we make this choice to speak more generally about each sentiment dictionary we test.

\clearpage

\subsection{Goals of sentiment analysis}

It may help to first frame the problem of detecting sentiment in text, and we will utilize the generalization given by Bing Liu in his 2012 book \textit{Sentiment Analysis and Opinion Mining} \citep{liu2012sentiment}.
Here, our goal is to detect and \textit{understand} the average sentiment of a document using the words contained within: \textit{document-level sentiment classification}.
Our definition extends that of \cite{liu2012sentiment} to include the goal of better \textit{understanding} text through sentiment detection, and this goal is complementary (and in some cases outright necessary) to achieve classification.
While document length varies, \cite{liu2012sentiment} subdivides finer-grained classification into two categories: (1) classifying sentence-level sentiment and (2) classifying entity-level sentiment.
Sentence-level sentiment is detecting sentiment in sentences, and entity-level sentiment aims to predict sentiments that are directed at named entities (e.g., products, people, or corporations).
We express caution in pursuing these latter goals using existing methodology, namely in classifying short, informal text.
We will examine in Chapter~\ref{chap:benchmarking} how dictionary based approaches are effective at the document level, but fail at the sentence level (and by extension fail at the entity level as well).
Several examples of different sentences are also given in \cite{liu2012sentiment}, highlighting the difficulty of classifying individual sentences, and we share these examples here.

The accuracy of classifying documents correctly as positive or negative is commonly measured using precision, recall, and F-score statistics, as in \cite{ribeiro2016sentibench}.
These measures assess the classification accuracy, but do not attempt to assess the qualitative goal of achieving better understand of text with sentiment analysis (an area on which our work will build).
Both of these goals can be assessed with ground truth data, and next we review publicly available data sets for sentiment evaluation.

\subsection{Publicly available annotated data}
Review papers such as those by \cite{giachanou2016like} attempt to capture the many advances in the field, including applications of machine learning with training data, although they only identify 3 of the 17 sentiment dictionaries that we list in Chapter~\ref{chap:benchmarking}.
They identify the lack of benchmarks as important issue \citep{giachanou2016like}:
\begin{quote}
  One of the main challenges in evaluating approaches that address Twitter-based sentiment analysis is the absence of benchmark datasets. In the literature, a large number of researchers have used the Twitter API to crawl tweets and create their own datasets, whereas others evaluate their methods on collections that were created by previously reported studies. One major challenge in creating new datasets is how the tweets should be annotated. There are two approaches that have been followed for annotating the tweets according to their polarity: manual annotation and distant supervision.
\end{quote}
To this end, we note the availability of datasets below and attempt to collect each dataset enumerated by \cite{giachanou2016like,saif2013evaluation,ribeiro2016sentibench} in Table~\ref{tbl:twitter-data} and make them accessible in one place online.
In addition to these public datasets, some academic groups choose not to release their tagged data, and there are claims of very large datasets held by private companies in the sentiment analysis space.
Given the time and cost associated with obtaining high quality training data, and the ubiquity of machine learning for sentiment analysis in industry, the training data itself can be viewed as a trade secret.

\begin{table}
  \begin{center}
  {\tiny

\begin{tabular}{l | l | l | l}
  \hline
  Short name & Description & \# Samples & Referenced By\\
  \hline
STS,Tweets\_STF,STS-Test & Stanford Twitter Sentiment & 499 & G, R, S\\
Sanders,Tweets\_SAN,Sanders & Sanders Corpus & 3424 & G, R, S\\
HCR,HCR & Health Care Reform & 4616 & G, S\\
OMD,Tweets\_DBT,OMD & Obama-McCain Debate & 3298 & G, R, S\\
SS-Tweet,Tweets\_RN\_I,SS-Twitter & SentiStrength Twitter Dataset & 4243 & G, R, S\\
SemEval,Tweets\_Semeval,SemEval & SemEval Datasets & 6087 & G, R, S\\
STS-Gold,STS-Gold & STS-Gold & 2036 & G, S\\
DETC,DETC & Dialogue Earth Twitter Corpus & N/A & G, S\\
Tweets\_RND\_IV & aisopos\_ntua & 500 & R\\
Comments\_TED & TED Comments & 839 & R\\
Comments\_BBC & SentiStrength BBC Comments & 1000 & R\\
Comments\_Digg & SentiStrength Digg Comments & 1077 & R\\
Reviews\_I & SentiStrength Myspace Reviews & 1041 & R\\
RW & SentiStrength Runners World Forum & 1046 & R\\
Comments\_YTB & SentiStrength YouTube Comments & 3407 & R\\
Amazon & VADER Amazon Reviews & 3708 & R\\
Reviews\_II & VADER Movie Reviews & 10605 & R\\
Comments\_NYT & VADER NYT Comments & 5190 & R\\
Tweets\_RND\_II & VADER Tweets & 4200 & R\\
Tweets\_RND\_III & DAI-Labor English MT & 3771 & R\\
ORT & Opinion Retrieval Twitter & 5051 & L\\

\end{tabular}  }
  \end{center}
  \caption{Summary of publicly available Twitter datasets tagged with sentiment labels.
    In respect of Twitter's Terms of Service, lists of the Tweet IDs are provided, as well as a script to download the Tweets through Twitter's public API (note some data may not longer be available).
  We shorted the references as follows as G: \cite{giachanou2016like}, S: \cite{saif2013evaluation}, R: \cite{ribeiro2016sentibench}, and L: \cite{luo2012opinion}.}
  \label{tbl:twitter-data}
\end{table}

In addition to the tagged datasets above, we attempt to provide a comprehensive list of sentiment dictionaries in Table~\ref{tbl:summary}.

\subsection{Natural Language Processing techniques}
\label{subsec:nlp}

As itself a tool for NLP,
sentiment analysis leverages approaches that are applied more broadly (e.g., classification),
and can benefit,
if only slightly,
from other such techniques.
In this section, we provide a very brief overview of techniques
for processing raw text,
detecting boundaries of multi-word expressions,
disambiguating word senses,
tagging parts-of-speech,
and dependency parsing.

\subsubsection{Tokenization}

Here,
we consider words as the basis for our computation,
and the process of extracting words from raw text is often referred to as ``tokenization''.
The simplest tokenization procedure is splitting raw text strings on spaces,
with words being any contiguous non-space characters.
For well structured (formal) writing,
this approach presents few false positive matches,
but this approach is often too simple for processing informal text
(e.g., Twitter),
where grammar is not reliable.
To improve upon the aforementioned approach,
we build a list of known ``word characters'' (e.g., the letters a-z, the apostrophe, hyphen, etc.) and extract all contiguous sequences of these characters as words.
An example regular expression implementing this approach is provided in Section~\ref{subsec:regex-parsing}.
The final consideration here are the various uses of individual words;
the representation of a word differs based on,
but not limited to,
the different classes,
inflection,
contractions,
possessive use,
and/or pluralization of the word.
Depending upon the ultimate use case,
a choice can be made for how to process words.
A common choice is to reduce words to their root,
a process called ``stemming'',
which removes the inflection from words,
a popular implementation is provided by \cite{porter2001snowball}.
A widely used source for annotated data based on word stems is the morphology of WordNet \citep{fellbaum1998a}.
In the approach that we adopt for sentiment analysis,
we attempt to retain the most complete representation of words,
without removing the information about usage that may be contained beyond a word's root or stem.
This achieves a very basic
and computationally efficient disambiguation between word senses.

\subsubsection{Multi-word Expressions}
\label{subsec:mwe}

In addition to tokenization,
the meaningful units of language often span multiple words.
These multi-word expressions,
or ``phrases'',
can also be extracted from tokenized words.
Here we summarize two state-of-the-art approaches from \cite{handler2016bag} and \cite{williams2016boundary}.\\

\noindent \textbf{Williams, J.~R. (2016).
\newblock Boundary-based MWE segmentation with text partitioning.
\newblock {\em arXiv preprint arXiv:1608.02025\/}.}

\noindent Williams performs boundary-based MWE segmentation with text partitioning,
building on prior work that introduces random and serial partitioning algorithms,
and showing that phrase frequency follows Zipf's law more closely than words alone. 
Trained models for partitioning rely on (1) phrase likelihood from ``informed random partitioning'', (2) entries the Wiktionary, and (3) annotated corpora.
The model is general purpose for pattern recognition,
and can be run using text data or PoS tags,
combining the output phrases for higher recall.
Altogether, this achieves state-of-the-art performance with flexible application to any text-based corpora.\\

\noindent \textbf{Handler, A., M.~J. Denny, H.~Wallach, and B.~O'Connor (2016).
\newblock Bag of what? simple noun phrase extraction for text analysis.
\newblock {\em NLP+ CSS 2016\/}, 114.}

\noindent Handler and colleagues build upon prior work that defines a grammar of PoS labels for noun phrases.
In essence,
the approach uses patterns to match noun phrases.
The implementation realizes computational feasibility with a Finite State Transducer (FST) compiled to find all matches of their pattern represented by a Finite State Grammar (FSG).
As an example of this general type of approach,
the pattern of word labels Adjective Noun Noun (encoded ANN) would be successfully matched by the grammar
\verb{(A|N)*N(N)*{,
where the \verb|*| represents 0 or more matches of the previous expression
(as in standard regular expression syntax, otherwise known as the Kleene star).
The availability of reliable part-of-speech tags is assumed by this approach,
although this is known to be a harder problem for informal text
(e.g., social media).\\

We conclude that both of these available methods, and even the ``naive'' method described by \cite{mikolov2013distributed} offer an improvement upon unigram models for bag-of-words approaches to sentiment analysis, which includes the methods used in this dissertation.
Sentiment dictionaries only contain ratings for single words, and extending existing dictionary ratings to MWEs is a widely acknowledged area for future research.

\subsubsection{Word Sense Disambiguation (WSD)}
\label{subsec:wsd}

To get a sense of the Word Sense Disambiguation (WSD) problem,
here we examine a scholarly competition:
The English All-Words Task of the SENSEVAL-2 series.
The SENSEVAL competitions began in 1998,
and the second and third instantiations took place in 2001 and 2004.
After 2004, the scope of tasks was broadened and the name switched to SemEval,
being held again in 2007, 2010, and 2012--2017 every year.
First, we summarize the construction of the benchmark by \cite{snyder2004english},
and then we examine the winning entry from \cite{decadt2004gambl}.\\

\noindent \textbf{Snyder, B. and M. Palmer (2004). The english all-words task. In \textit{Senseval-3: Third International Workshop on the Evaluation of Systems for the Semantic Analysis of Text}, pp. 41–43. Association for Computational Linguistics.}

\noindent To develop the training and testing data for Senseval-3, Snyder and Palmer
extracted approximately 5,000 words from two Wall Street Journal articles and one excerpt from the Brown Corpus.
Word sense was annotated by two people using Wordnet senses, and then settled by a third party, for a total of 2,212 words and multi-word-expressions.
They found the inter-annotator agreement at 72.5\%, representing a practical upper bound for the performance of computational methods.\\

\noindent \textbf{Decadt, B., V. Hoste, W. Daelemans, and A. Van den Bosch (2004). Gambl, genetic algorithm optimization of memory-based wsd. In \textit{Senseval-3: Third International Workshop on the Evaluation of Systems for the Semantic Analysis of Text}, pp. 108–112. Association for Computational Linguistics.}

\noindent GAMBL is a ``word expert'' approach to WSD in which a word sense classifier is trained for each individual word.
The parameters of this classifier are optimized using a genetic algorithm, and the method achieves the best precision/recall of .652.\\

\subsubsection{Part-of-Speech tagging}

Part-of-Speech (PoS) tagging aims to disambiguate between the various forms that a word can take: verb, pronoun, preposition, adverb, conjunction, participle, and article are eighth of the most well recognized categories.
This information tells us how a word relates to the neighboring words around it, and finer grained taxonomies of parts of speech in English contain more than 80 types.
To train and test algorithms for this task, large annotated corpora such as the Penn Treebank are available form \cite{marcus1993building} and OntoNotes .\\

\noindent \textbf{Abney, S. (1997). Part-of-speech tagging and partial parsing. In \textit{Corpus-based methods in language and speech processing}, pp. 118–136. Springer.}

\noindent \cite{abney1997part} elaborates upon the work of \cite{church1988stochastic} and \cite{derose1988grammatical} to develop a reasonable, approximate approach to PoS tagging. State-of-the-art approaches can be classified into rule-based and stochastic, the latter making extensive use of Hidden Markov Models (HMMs) to represent state as a latent variable.\\

\noindent \textbf{Toutanova, K., D. Klein, C. D. Manning, and Y. Singer (2003). Feature-rich part-of-speech tagging with a cyclic dependency network. In \textit{Proceedings of the 2003 Conference of the North American Chapter of the Association for Computational Linguistics on Human Language Technology-Volume 1}, pp. 173–180. Association for Computational Linguistics.}

\noindent \cite{toutanova2003feature} develop a PoS tagger with improved accuracy which is competitive in terms of both speed and accuracy with any attempt since.
This is achieved by using a cyclic dependency network to represent the state of the tagger, and achieves 97.24\% accuracy on the Penn Treebank corpus.
The tagger is used by \cite{manning2014stanford} in the most recent release the Stanford CoreNLP natural language processing toolkit.\\

\noindent \textbf{Owoputi, O., B. O’Connor, C. Dyer, K. Gimpel, N. Schneider, and N. A. Smith (2013). Improved part-of-speech tagging for online conversational text with word clusters. Association for Computational Linguistics.}

\noindent Existing PoS taggers excel at the task in well structured language but are not applicable to short, informal text. In \cite{owoputi2013improved}, large-scale unsupervised word clustering and lexical features are used to achieve 93\% accuracy on Twitter.
In addition, guidelines for manually annotating this type of text are provided.\\

The application of PoS tagging in stand-alone tests on tagged corpora has achieved high rates of accuracy on both formal and informal text.
It now stands to reason that this addition of information for individual words and MWEs have applications in an end-to-end system for sentiment analysis.

\subsubsection{Dependency Parsing}

Dependency parsing aims to extract the syntactic relationship between the words used in a sentence.
Also referred to as syntax parsing, dependency parsing is one more NLP tool that aims to solve a disambiguation problem: of all possible dependency parses, choosing the most appropriate.
In many cases, this disambiguation is between two parses that are both grammatically valid, but nonsensical otherwise; consider the different interpretations of ``They ate the pizza with anchovies'' (seen in Figure \ref{fig:displacy}).
In the prior examples, anchovies could either be utensils or a topping or their friends, but this is obvious to us with commonsense knowledge.
Other examples that I found compelling for parsing are garden path sentences---those which confuse the common human parsing by leading our parse down the wrong path---such as ``the old man the boat'' or ``the horse ran past the barn fell''.
Both examples are valid senses, but are easy to read incorrectly on the first pass.
The dependency parsing algorithms that we examine next solve each of the examples we have just given correctly by utilizing neural network approaches that find the most probable parse.

We note that PoS tagging, a shallower form of parsing, is about twenty times faster than parsing, for applications where computational costs of parsing are a bottleneck \citep{handler2016bag}.
State-of-the-art approaches from both \cite{chen2014fast} and \cite{andor2016globally} achieve parse accuracies over 90\%.\\

\noindent \textbf{Chen, D. and C. D. Manning (2014). A fast and accurate dependency parser using neural networks. In \textit{EMNLP}, pp. 740–750.}

\noindent In \cite{chen2014fast}, a dependency parser is built that uses dense features of the surrounding text to improve upon both the accuracy and speed of current parsers.
For performance, they note their ``parser is able to parse more than 1000 sentences per second at 92.2\% unlabeled attachment score on the English Penn Treebank''.\\

\noindent \textbf{Andor, D., C. Alberti, D. Weiss, A. Severyn, A. Presta, K. Ganchev, S. Petrov, and M. Collins (2016). Globally normalized transition-based neural networks. \textit{arXiv preprint arXiv:1603.06042}}.

\noindent \cite{andor2016globally} from Google Inc. (now Alphabet) improve further on the accuracy of neural network parsers and release a pre-trained model for general consumption.
Their pre-trained model is \textit{Parsey McParseface} and they note that ``for dependency parsing on the Wall Street Journal we achieve the best-ever published unlabeled attachment score of 94.61\%''.\\

Much like PoS tagging, dependency parsing algorithms extract meaningful information at the sentence level with high accuracy.
An open challenge for sentiment analysis is the incorporation of this local information while retaining interpretability across large corpora.

\subsubsection{Heuristics}

In our pursuit to understand and evaluate sentiment analysis methods at a human level, it is intuitive yet deceiving to consider individual sentences.
At the level of individual sentences, the bag of words approach is no longer useful.
One attempt to improve these models for short text is to incorporate rules that are manually encoded to fit a given model for language, relying on the grammatical structure of language.
Such a rule might be to consider negation words such as ``not'' to reverse the polarity of the following sentiment word, such that ``not $w_i$'' would be combined and assigned the score of ``$-w_i$''.

Various attempts to incorporate rule-based heuristics and dictionary approaches for sentiment analysis include the work of \cite{thelwall2012sentiment} and \cite{hutto2014vader}.
The systems developed by \cite{kiritchenko2014sentiment}, \cite{wilson2005recognizing}, and \cite{polanyi2006contextual} incorporate a rule for negation.
An analysis of the usefulness of different features for Twitter sentiment analysis is performed by \cite{agarwal2011sentiment}, including PoS and binary lexicon features.
Perhaps unsurprisingly, the polarity of words is the single most useful feature.
The analysis showed that the most useful combination is the one of PoS with the polarity of words.
\cite{hutto2014vader} report an increase on in the F1 score for binary Tweet classification of 2.1\% using negation, extended vowels (``happy'' to ``haaapy''), punctuation, and capitalization as cues.

\subsection{Building corpus-specific sentiment dictionaries}

\subsubsection{Categorization}

Previous work on building sentiment dictionaries using data, as opposed to human evaluation, has taken various forms.
We categorize these approaches by three main categories; (1) the type of data that is used to gain information about how words are similar, (2) how the data is processed, and (3) which methods are used to infer semantic properties.\\

\noindent Types of data include:
\begin{itemize}
\item Thesaurus\vspace{-12pt}
\item Word associations\vspace{-12pt}
\item Unstructured text corpora
\end{itemize}
\noindent Data processing
\begin{itemize}
\item Network from structured data\vspace{-12pt}
\item Network for POS patterns\vspace{-12pt}
\item Word embedding vectors\vspace{-12pt}
\item Vectors similarity (cosine distance, etc) $\rightarrow$ networks ($k$-NN, etc)
\end{itemize}
\noindent Some of the methods employed:
\begin{itemize}
\item Graph clustering\vspace{-12pt}
\item Graph label propagation\vspace{-12pt}
\item Orthogonal subspace projection on embedding
\end{itemize}
We distinguish these approaches from machine learning approaches that estimate emotion of words from tagged training data in that these approaches extend existing scores about words.

Chronologically, the first approach here is by \cite{hatzivassiloglou1997predicting}, and the most recent we have found is the work of \cite{rothe2016ultradense}.
We will proceed by summarizing the main result of each paper, casting the methodology into one of the aforementioned categories.

\subsubsection{Previous approaches}
\label{subsec:prev-propagation}

First, we take a close look at the earliest effort to build a corpus-specific sentiment dictionary to get a deeper sense of the steps involved in this task.\\

\noindent \textbf{Hatzivassiloglou, V. and K.~R. McKeown (1997).
Predicting the semantic orientation of adjectives.
In {\em Proceedings of the eighth conference on European chapter of
  the Association for Computational Linguistics}, pp.\  174--181. Association
  for Computational Linguistics.}

\noindent \cite{hatzivassiloglou1997predicting} use a four-pronged approach: (1) adjectives are extracted from large text corpora that are linked by conjunctions (``and'' or ``but''), (2) a log-linear regression determines whether they are synonyms/antonyms to make a graph of positive/negative connections, (3) a clustering algorithm is run for two clusters, and (4) the cluster with the greatest average frequency is labeled as the positive words.
The 1987 WSJ corpus is used, with PoS tags for adjectives and conjunctions.
They report 82\% accuracy on the binary classification of word pairs as synonym or antonym, and 90\% accuracy on semantic orientation (predicting manual labels on 1336 adjectives).
Their approach does not rely on existing word scores, but nevertheless forms the basis for future work that does incorporate existing sentiment dictionary data.\\

Now that we have seen one approach in more detail, we will look ahead to methodology that more closely informs our own work.
The years following saw an expansion in the methods, processing, and data used to automatically extend affective word scores, including work \citep{turney2002thumbs,turney2003measuring,taboada2004analyzing,kim2004determining,hu2004mining,esuli2006sentiwordnet,das2007yahoo,kaji2007building,blair2008building,bestgen2008building,rao2009semi}.
We start again in more depth with recent work of Velikovich, directly applicable to extending data sets that we are familiar with (e.g., labMT).\\

\noindent \textbf{Velikovich, L., S.~Blair-Goldensohn, K.~Hannan, and R.~McDonald (2010).
\newblock The viability of web-derived polarity lexicons.
\newblock In {\em Human Language Technologies: The 2010 Annual Conference of
  the North American Chapter of the Association for Computational Linguistics},
  pp.\  777--785. Association for Computational Linguistics.}

\noindent In the paper from \cite{velikovich2010viability}, many of the specifics of the approach are left out.
We review this paper because the methodology outlined is very similar in spirit to all of the approaches that follow.
For a domain corpus, they use n-grams up to length 10 scraped from 4 billion web pages, however the details of this corpus are left vague.
They then use the cosine distance between context vectors from these n-grams to build a $k$ nearest neighbor ($k$-NN) network with $k=15$ (the method used to generate context vectors is again left to the reader).
Seed words within the network are labeled with positive and negative sentiment, and scores for all n-grams are determined by shortest paths to the seed set, a using a generic graph propagation algorithm.
For results, \cite{velikovich2010viability} report that their effort compares favorably to the manually constructed lexicon from \cite{wilson2005recognizing} and a lexicon from WordNet used in \cite{blair2008building}.\\

\noindent \textbf{Bestgen, Y. and N.~Vincze (2012).
\newblock Checking and bootstrapping lexical norms by means of word similarity
  indexes.
\newblock {\em Behavior research methods\/}~{\em 44\/}(4), 998--1006.}

\noindent \cite{bestgen2012checking} begin by taking 300-dimensional word embeddings from the Singular Value Decomposition (SVD) of the word co-occurrence matrix of the TASA corpus, comprised of 44K documents.
They use these embeddings to build a $k$-NN network, and then use the DIC-LSA technique of \cite{bestgen2008building} with the ANEW dictionary (using the dictionary scores to measure correlations with words in the network).
This approach extends the ANEW dictionary by adding scores to additional words, directly using the scores in the ANEW dictionary itself.
For different values of $k$, the score for each word in the network is taken to be the average of it's neighbors (the $k$ closest words in the embedding space), and for words with scores from ANEW, the node value itself is held-out.
By using only the most extreme words (those in ANEW with scores closer to 1 and closer to 9), they achieve an correlation coefficient (Cohen's Kappa) of .53--.94 on sets of all--190 of the words from ANEW (the latter .94 correlation achieved with using the 190 most extreme words in ANEW).
In addition, they provide ratings using their method for 17,000 English words.\\

\noindent \textbf{Tang, D., F.~Wei, B.~Qin, M.~Zhou, and T.~Liu (2014).
\newblock Building large-scale twitter-specific sentiment lexicon: A
  representation learning approach.
\newblock In {\em COLING}, pp.\  172--182.}

\noindent \cite{tang2014building} train a neural network (NN) to learn phrase sentiment from phrase embeddings using a graph collected from Urban Dictionary and Tweets with emoticons.
The Tweets with emoticons are used to embed all phrases in a two dimensional space with the loss function as a hybrid between word context (e.g., word2vec) and emoticon label context (happy or sad).
A network of words is extracted from Urban Dictionary and used to apply label propagation for positive (good, \verb|:)|), negative (poor, \verb|:(|), and neutral words (when, he) across the network (which includes phrases).
The word embeddings and scores from label propagation are used as features for a ternary sentiment classifier that is trained to predict scores from label propagation.
Their system outperforms those tested for the SemEval 2013 competition by attaining a performance of macro F1 score .78, and their final dataset, TS-Lex, is composed of 65,685 words with sentiment scores and provided online.\\

\noindent \textbf{Amir, S., R.~Astudillo, W.~Ling, P.~C. Carvalho, and M.~J. Silva (2016).
\newblock Expanding subjective lexicons for social media mining with embedding subspaces.
\newblock {\em arXiv preprint arXiv:1701.00145\/}.}

\noindent Their approach to lexicon expansion ``consists of training models to predict the labels of pre-existing lexicons, leveraging unsupervised word embeddings as features'' \citep{amir2016expanding}.
Correlations between their method and existing continuous datasets had a maximum of 0.68, an improvement over support vector regression.
The resulting lexicon out-performed other methods in Tweet classification, although not all methods were compared.\\

\noindent \textbf{Hamilton, W.~L., K.~Clark, J.~Leskovec, and D.~Jurafsky (2016).
\newblock Inducing domain-specific sentiment lexicons from unlabeled corpora.
\newblock {\em arXiv preprint arXiv:1606.02820\/}.}

\noindent \cite{hamilton2016inducing} utilize the approach set out in \cite{velikovich2010viability} to generate corpus specific word embeddings using SVD and propagating sentiment labels on inferred $k$-NN network.
The most novel part of the approach measures the uncertainty in predicted labels with bootstrapping procedure that holds out fractions of seed set (with a seed set of 10 words, holding out 2).
They claim to measure performance with correlations to existing dataset of \cite{warriner2013norms}, but not found in results.\\

\noindent \textbf{Mandera, P., E.~Keuleers, and M.~Brysbaert (2015).
\newblock How useful are corpus-based methods for extrapolating psycholinguistic variables?
\newblock {\em The Quarterly Journal of Experimental Psychology\/}~{\em
  68\/}(8), 1623--1642.}

\noindent \cite{mandera2015useful} measure sensitivity of the performance of corpus specific sentiment dictionaries to the number of words in the training data.
They split the \cite{warriner2013norms} corpus into training and testing sets at different thresholds (e.g., 70/30 and 80/20).
Networks are built using $k$-NN and Random Forests on four different distances metrics, and the best performance is attained from the SVD of PMI embedding and a $k$-NN with $k=30$.
They show that accuracy for this best method varies from .61--.72 between a 10/90 to 50/50 split into testing and training.
The reported accuracy leads the authors to cast doubts on the efficacy of automated approaches, but their survey is not exhaustive and the next methods we will explore improve upon the accuracy.\\

\noindent \textbf{Van~Rensbergen, B., S.~De~Deyne, and G.~Storms (2016).
\newblock Estimating affective word covariates using word association data.
\newblock {\em Behavior Research Methods\/}~{\em 48\/}(4), 1644--1652.}

\noindent \cite{van2016estimating} estimate word scores using word association data for 14K dutch words, finding the best correlation between this method and human evaluation for $k$-NN algorithm (also tried ``Orientation towards Paradigm Words'').
For $k=10$ they obtained correlations for valence, arousal, and dominance of .91, .84, and .85.
This performance is considerably better than was achieved by \cite{mandera2015useful} for English using corpus derived word similarity.
These results highlight the sensitive differences between word analogy tasks for human readers and the information extracted by vector space embedding methods.\\

\noindent \textbf{Rothe, S., S.~Ebert, and H.~Sch{\"u}tze (2016).
\newblock Ultradense word embeddings by orthogonal transformation.
\newblock {\em arXiv preprint arXiv:1602.07572\/}.}

\noindent \cite{rothe2016ultradense} transform the embedding space of works via optimization of certain dimensions onto known semantic properties.
This amounts to reducing the 300 or so dimensions typically used for vector space embedding into less than three dimensions.
They apply Stochastic Gradient Descent (SGD) to learn a transformation $Q$ that orthogonalizes the embedding matrix $A$ under the constraint of establishing a sentiment dimension.
This approach is more successful than embedding words directly into such a low dimension space, agreeing with previous work that has show vector embedding performs best with more than 100 dimensions, while extracting the relevant semantic information for sentiment analysis.
For lexicon creation, their approach labeled ``Densifier'' achieves the statistically significant best performance on SemEval 2015 Task 10E with Kendall's $\tau$ of .654.\\

Altogether, these approaches provide a roadmap and demonstrate the possibility of constructing a high-quality, general purpose, phrase based sentiment dictionary.

\subsection{Visualization}
\label{subsec:vis}

Lacking from the bulk of research that applies sentiment analysis, but crucial for validation and understanding, is visualization of sentiment analysis.
Despite limited attempts by researchers in sentiment analysis to use visualization to understand their analysis, online tools have been built to allow anyone to build simple visualizations in a straightforward way \citep{viegas2007manyeyes}.
Motivation for our choice of dictionary-based methods along with a straightforward averaging algorithm for generating scores is that the analysis can be visualized to be understood.
The averaging algorithm is linear and this allows for the comparison of the individual word contributions to text sentiment classification, both enabling greater understanding and validating the analysis.

An overview of previous approaches to text visualization can be found in \cite{heer2014text} and \cite{cao2016introduction}. We note the four goals of text visualization as identified by Heer: understanding, comparison, grouping, and correlation.
Here, we focus on the task of understanding.
A selection of recent work that builds on this task is available from \cite{hearst2009search} (Chapter 11), \cite{chuang2012interpretation}, \cite{van2009mapping}, and \cite{chuang2012termite}.

\begin{figure}[tbp!]
  \centering
  \includegraphics[width=0.88\textwidth]{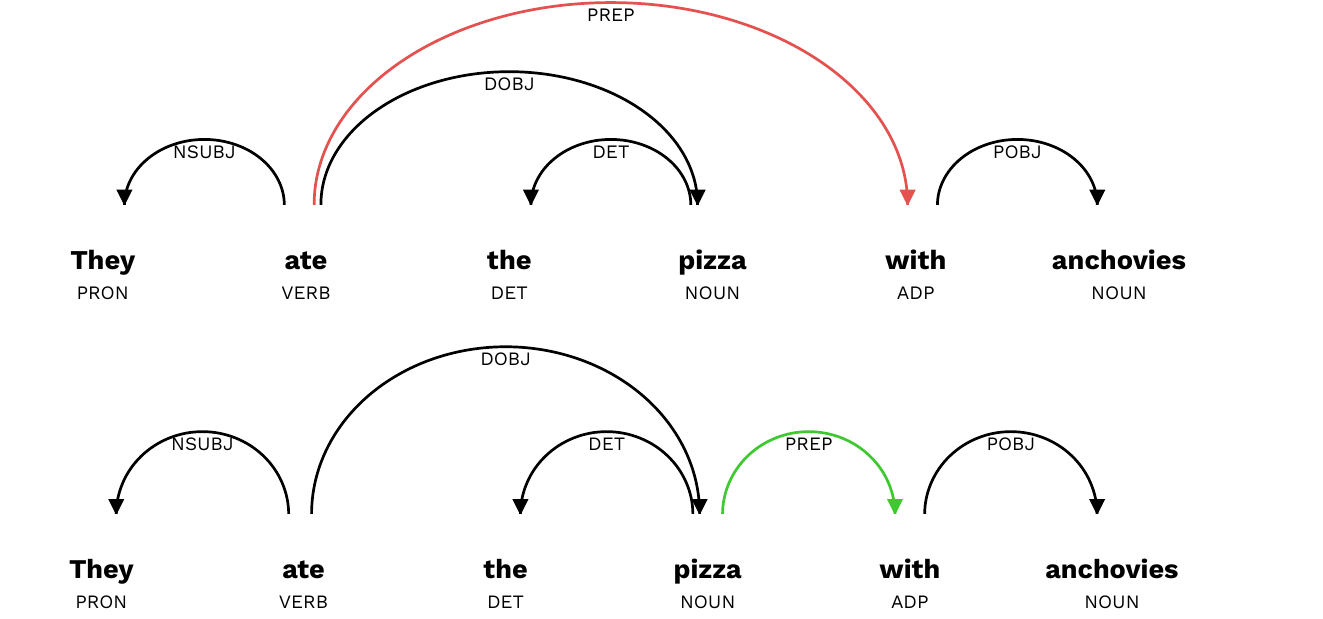}
  \caption[]{
    Visualization of a syntactic dependency parse with the displaCy tool from \cite{honnibal2015displaying}, a companion to the spaCy package for NLP in Python.
    The tool doubles as an annotation tool with key-based input for efficient manual dependency tagging.
  }
  \label{fig:displacy}
\end{figure}

Visualizations of readable portions of text are able to communicate the results of analysis at that level, such as the syntactic parse visualization in \ref{fig:displacy}.
On a sentence level, we can see with or without added visual clues (e.g., colored backgrounds or font size) which individual words have either positive or negative scores, and how their balance contributes to the average-based classification.
When rules become involved, this process is more complicated and it may be necessary to utilize a sentence diagram to understand the classification at even the individual sentence level.
Neither of these approaches scale to visualize more than individual sentences, a fundamental shortcoming in working with big data.

Next, we examine tag clouds as a tool to understand text and the results of text analysis.

\subsubsection{Tag clouds}

Tag clouds are a popular method for displaying the results of text analysis, with the size of text being used to represent one variable from the analysis and the layout of words with random locations, angles, and color, generally positioned to minimize white space.
Various attempts have been made to assess the efficacy of tag clouds compared to more traditional statistical information visualizations such as bar charts with a consensus that they are less effective, though aesthetically pleasing: see \cite{halvey2007assessment}, \cite{rivadeneira2007getting}, and \cite{hearst2008tag}.
One popular package for producing word clouds layouts is ``Wordle'' from \cite{feinberg2009wordle}.

Since tag clouds by wordle have random layouts, improvements that incorporate relevant information into the layout itself have been considered.
In \cite{schrammel2009semantically} they compare the performance and likability of four approaches: alphabetic, random, similarity on Flickr, and distance in WordNet.
From 64 participants, they find that ``semantically clustered tag clouds can provide improvements over random layouts in specific search tasks and that they tend to increase the attention towards tags in small fonts compared to other layouts''.

In \cite{lohmann2009comparison} tag cloud layouts are compared on three tags and results show that there is no single best layout.
The three tasks they test and the best layout for each are:
\begin{itemize}
\item Finding a specific tag: Sequential layout with alphabetical sorting.\vspace{-12pt}
\item Finding the most popular tags: Circular layout with decreasing popularity.\vspace{-12pt}
\item Finding tags that belong to a certain topic: Thematically clustered layout.
\end{itemize}
It is also confirmed using eye tracking that tag clouds are scanned (not read), attention is focused on the center of the tag cloud, and they all perform sub-optimally for looking up specific words.

A study of the social (non-academic) use of Wordle is done by \cite{viegas2009participatory}, finding that the existence of tools for building custom Wordles was crucial to their popularity and that 35/49\% of men/women under the age of 20 did not know that frequency of usage is used for the font size.

Adding a time component to tag clouds with the use of ``sparklines'', \cite{lee2010sparkclouds} find that SparkClouds are able to communicate trends as well.
New layouts attempt to incorporate additional information to tag clouds through layout and color, such as the TAGGLE system of \cite{emerson2015tag}.

Moving beyond tag clouds, we briefly present word shift graphs in the next section.

\subsubsection{Word shift graphs}

\begin{figure}[tbp!]
  \centering
  \includegraphics[width=0.96\textwidth]{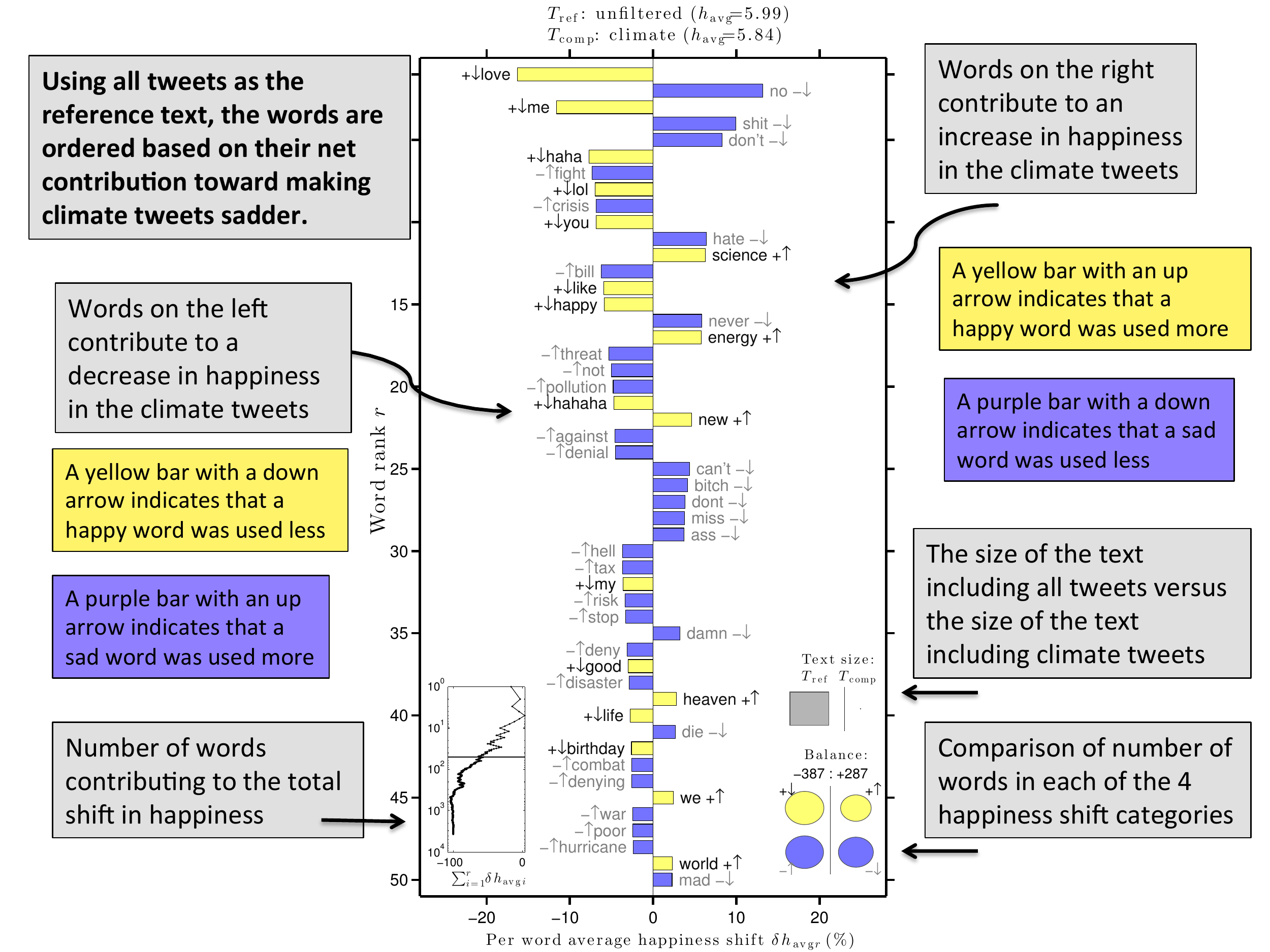}
  \caption[]{
    We quote the following caption and re-use the figure from \cite{cody2015climate}:
    A word shift graph comparing the happiness of tweets containing the word ``climate'' to all unfiltered tweets. The
reference text is roughly 100 billion tweets from September 2008 to July 2014. The comparison text is tweets containing the
word ``climate'' from September 2008 to July 2014. A yellow bar indicates a word with an above average happiness score. A
purple bar indicates a word with below average happiness score. A down arrow indicates that this word is used less within
tweets containing the word ``climate''. An up arrow indicates that this word is used more within tweets containing the word
``climate''. Words on the left side of the graph are contributing to making the comparison text (climate tweets) less happy.
Words on the right side of the graph are contributing to making the comparison text more happy. The small plot in the lower
left corner shows how the individual words contribute to the total shift in happiness. The gray squares in the lower right corner
compare the sizes of the two texts, roughly $10^7$ vs $10^{12}$ words. The circles in the lower right corner indicate how many happy
words were used more or less and how many sad words were used more or less in the comparison text.
  }
  \label{fig:emilyclimatewordshiftexplainer}
\end{figure}

An indispensable, scientific tool for visualizing text analysis is the word shift graph.
The graph was first designed and put to use by \cite{dodds2009b} to understand the result of sentiment analysis.
An online, interactive version of the graphs are used widely at \href{http://hedonometer.org}{hedonometer.org}, and more details on the use of these graphs is available at \href{http://www.uvm.edu/storylab/2014/10/06/hedonometer-2-0-measuring-happiness-and-using-word-shifts/}{compstorylab.org}.
The important difference between the word shift graph and tag cloud is that the word shift graph uses both spatial dimensions meaningfully, encoding the ranking of words in the vertical direction and the relevant statistical value in the horizontal direction, enabling comparison between the values.
We present a closer examination of an example word shift graph in Figure \ref{fig:emilyclimatewordshiftexplainer}.

We elaborate more on the construction, present use cases where the word shift graph helps us understand successes and failures of sentiment analysis, and generally make extensive use of the word shift graph as a tool in Chapter~\ref{chap:emotional-arcs}.
A future effort could aim to assess the efficacy of the word shift graph for text-based research, by performing a task-level user study.

\subsection{Benchmarking literature}

In this section we review recent efforts to benchmark sentiment analysis methods for their performance.\\

\noindent \textbf{Liu, B. (2012, May).
{\em Sentiment analysis and opinion mining}.
Synthesis Lectures on Human Language Technologies. San Rafael, CA:
Morgan \& Claypool Publishers.}

\noindent This book from Bing Liu provides a broad overview of sentiment analysis, and the many different problems that it hopes to address as well as summaries of many common approaches.
Liu provides a framework to understand the aspects of sentiment analysis, with the levels of analysis (aspect, sentence, document level), and goals including classification and opinion summarization.
In Chapter 8, a discussions of the methods for generating sentiment dictionaries is presented, and includes manual, dictionary-based, and corpus based approaches.
Survey methods are not considered (the well-known ANEW dictionary is absent), and there is some confusion between methods that use a dictionary to propagate scores and those that use features of a corpus to propagate scores (\cite{velikovich2010viability} incorrectly classified as the former).
While the references are extensive, no analysis is conducted to understand how the different approaches for generating sentiment dictionaries perform.
Despite these shortcomings, the book is a broad and very useful guide to the landscape of sentiment analysis.\\

\noindent \textbf{Hutto, C.~J. and E.~Gilbert (2014).
Vader: A parsimonious rule-based model for sentiment analysis of
  social media text.
In {\em Eighth International AAAI Conference on Weblogs and Social
  Media}.}

\noindent This paper is focused on the development of a new dictionary-based method for sentiment analysis that incorporates a rule-based system and a dictionary tailored to social media.
While other papers that introduce dictionaries for sentiment analysis have made comparisons between methods (e.g., LIWC correlations between the 2001, 2007, and 2015 dictionaries on their website), we include this as a benchmark because of the uncommon rigor in the comparisons made.
In particular, Hutto and Gilbert compare their new method VADER to 11 other sentiment analysis methodologies.
They compare to seven dictionary-based methods and four ML methods, and find favorable correlations between the classification of Tweets for the dictionary based methods.
In addition they perform tests to measure the performance gains to be had using four rules, and word sense disambiguation, finding mean F1 performance gains of 2 points on individual Tweets.
These rules are a subset of those employed by \cite{thelwall2012sentiment}.
The comparisons between sentiment dictionaries focus on the classification performance, and do not provide any insight into what properties of the dictionaries contributes to their performance.
In addition, no effort is made to use sentiment analysis as more than a binary classifier, a shortcoming that we will address.\\

\noindent \textbf{Giachanou, A. and F.~Crestani (2016, June).
Like it or not: A survey of twitter sentiment analysis methods.
{\em ACM Comput. Surv.\/}~{\em 49\/}(2), 28:1--28:41.}

\noindent This extensive survey from Giachanou \etal provides an overview and categorization of methods used to quantify sentiment on Twitter.
No quantitative comparisons are made between the methods themselves.
The broad categories of the methods they find are based on those from \cite{liu2012sentiment}:
\begin{itemize}
\item Machine Learning.\vspace{-12pt}
\item Lexicon-Based.\vspace{-12pt}
\item Hybrid (Machine Learning \& Lexicon-Based).\vspace{-12pt}
\item Graph-Based.
\end{itemize}
The focus is on ML approaches (as they note: ``The majority of [Twitter Sentiment Analysis] methods use a method from the field of machine learning'').\\

\noindent \textbf{Ribeiro, F.~N., M.~Ara{\'u}jo, P.~Gon{\c{c}}alves,
  M.~Andr{\'e}~Gon{\c{c}}alves, and F.~Benevenuto (2016, jul).
{SentiBench} --- a benchmark comparison of state-of-the-practice
  sentiment analysis methods.
  {\em {EPJ} Data Sci.\/}~{\em 5\/}(1), 23.}

\noindent This recent benchmark from Ribeiro \etal was published while our work was under review, having been submitted after our preprint was released on the arXiv.
The comparisons made by Ribeiro \etal utilize a variety of methods, and provide measures of performance for all methods based on F1 scores.
The methods selected include commercial, ML, and dictionary-based, and they are applied for four corpora.
Beyond metrics of classification performance, no insight is provided into the reasons why certain methods out-perform others, nor is any focus on understanding texts through sentiment (or using visualization), the key tenets of our effort in Chapter~\ref{chap:benchmarking}.\\

\clearpage
\pagebreak

\section{Emotional arcs}

Stories provide a useful framing to condense our experience, and through this they are both ubiquitous and powerful.
In 2011, a DARPA initiative ``Narrative Networks'' \citep{darpa2011narrative} said the following in relation to security:
\begin{quote}
  Narratives exert a powerful influence on human thoughts and behavior.
  They consolidate memory, shape emotions, cue heuristics and biases in judgment, influence in-group/out-group distinctions, and may affect the fundamental contents of personal identity.
  It comes as no surprise that because of these influences stories are important in security contexts: for example, they change the course of insurgencies, frame negotiations, play a role in political radicalization, influence the methods and goals of violent social movements, and likely play a role in clinical conditions important to the military such as post-traumatic stress disorder.
\end{quote}

The ubiquitous nature of stories is summed up well in \cite{dodds2013homo}:
\begin{quote}
We humans are storytelling and story-finding machines: \textit{homo narrativus}, if you will.
In making sense of the world, we look for the shapes of meaningful narratives in everything.
Even in science, we enjoy mathematical equations and algorithms because they are a kind of universal story.
Fluids---the oceans and atmosphere, the blood in your body, honey---all flow according to a single, beautiful set of equations called the Navier-Stokes equations.

In our everyday, human stories, far away from science, we have a limited (if generous) capacity to entertain randomness---we are certainly not \textit{homo probabilisticus}.
Too many coincidences in a movie or book will render it unbelievable and unpalatable.
We would think to ourselves, ``that would never happen in real life!''
This skews our stories. We tend to find or create story threads where there are none.
While it can sometimes be useful to err on the side of causality, the fact remains that our tendency toward teleological explanations often oversteps the evidence.
\end{quote}

\begin{figure}[tbp!]
  \centering
  \includegraphics[width=0.48\textwidth]{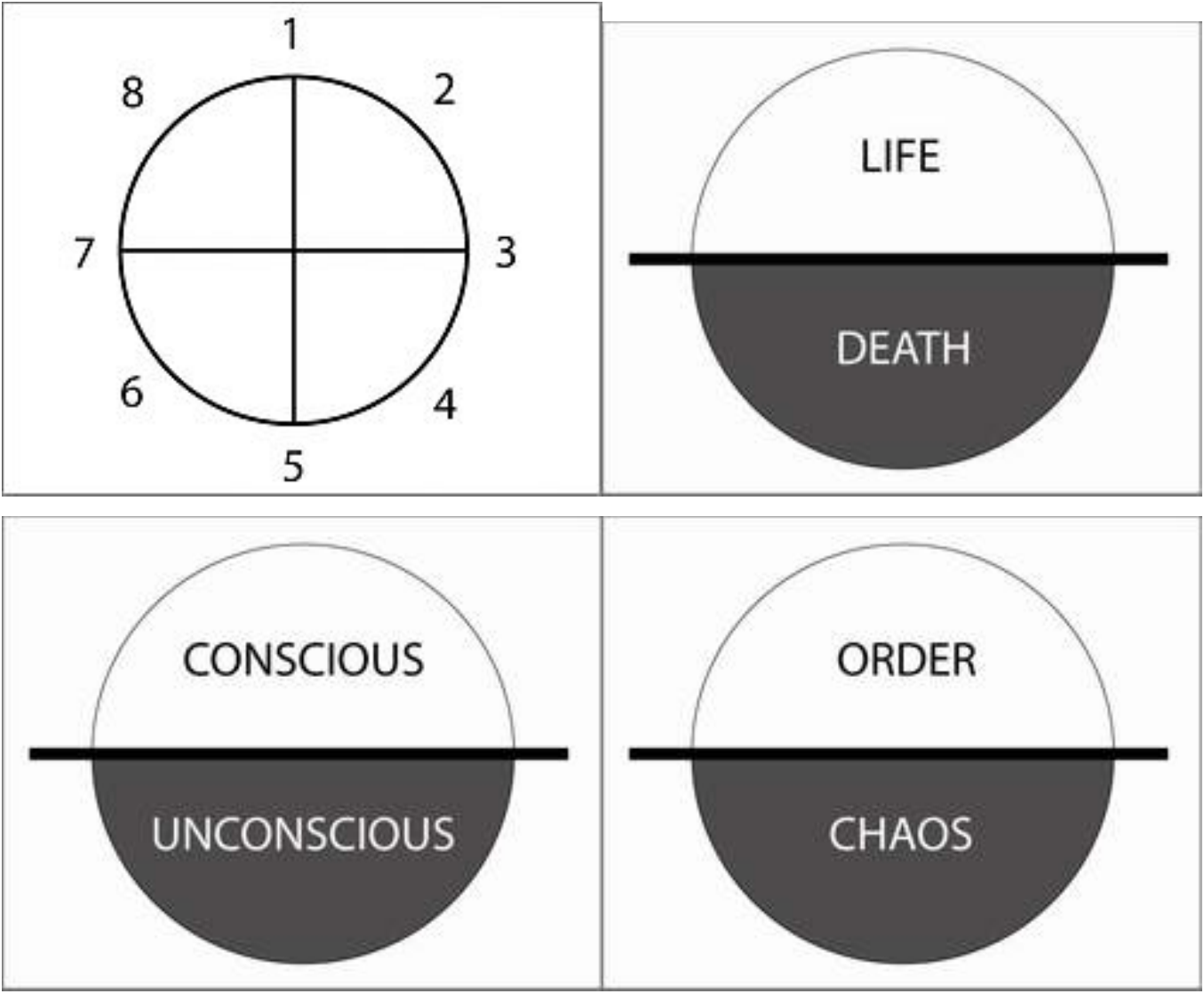}
  \includegraphics[width=0.48\textwidth]{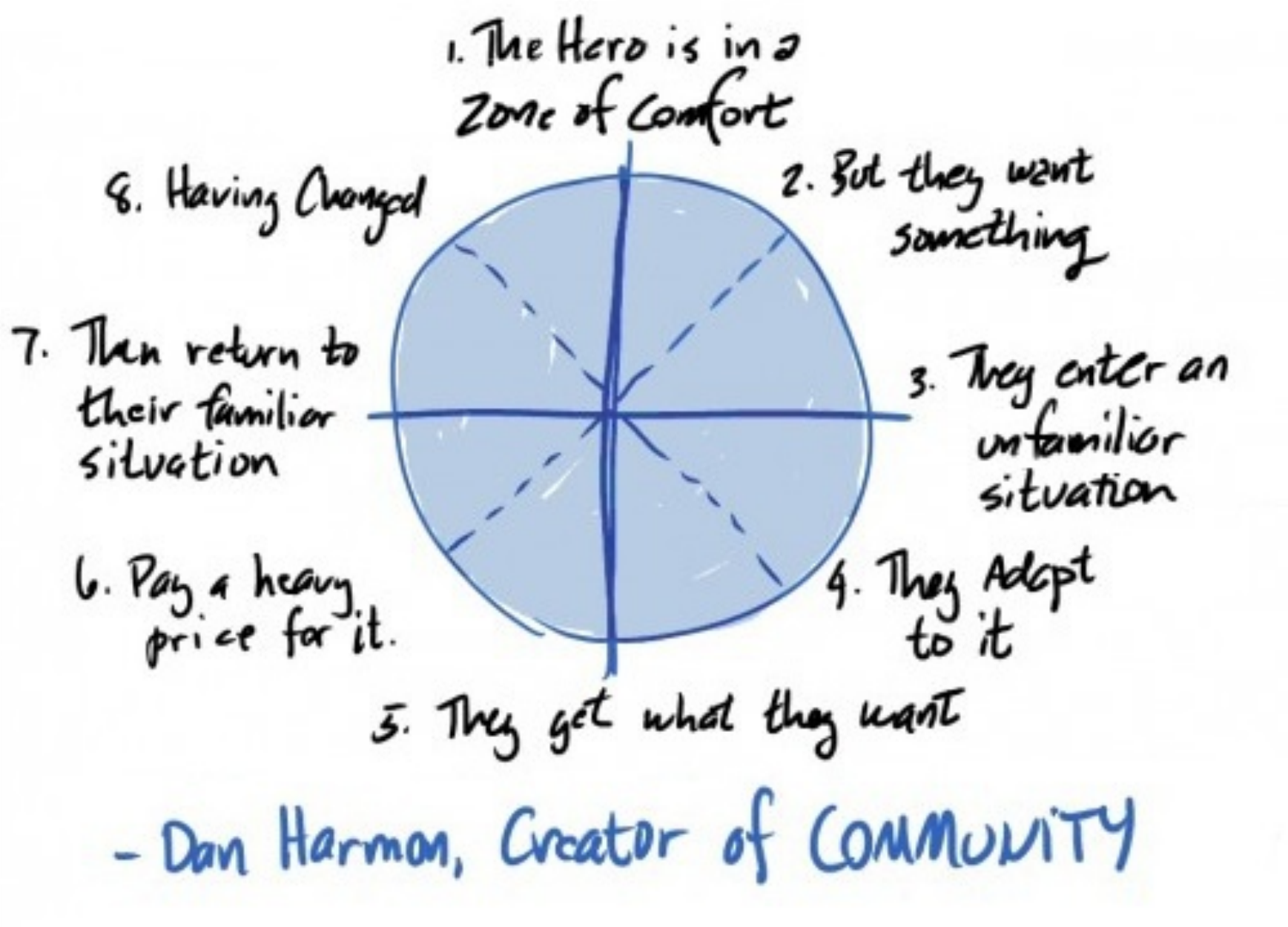}
  \caption[]{
    Harmon cycles with and without labels, as used to develop the show \textit{Community}.
    The cyclical nature of the story has roots in the ``monomyth'' of \cite{campbell1949hero}.
  }
  \label{fig:harmon}
\end{figure}

In Chapter~\ref{chap:emotional-arcs} we consider previous work that finds between one and 36 different plot types: \cite{campbell1949hero,harris1959basic,abbott2008cambridge,booker2006seven,polti1921thirty}.
Of these, the work of \cite{campbell1991a} has gained popular attention as a result of the expositions of Dan Harmon in writing the show \textit{Community} \citep{raftery2011harmon}.
In a series of online posts, Harmon elaborates on the ``monomyth'' and its incorporation into the writing of the \textit{Star Wars} movies \citep{volger1992writer}.
The plot here is cyclical, and therefore represented on a circle, and the argument goes that all well constructed plots can be arranged to fit into this mold.
The basic circle consists of 8 locations, starting and ending in the same place, and show a labeled visualization of these locations in Figure \ref{fig:harmon}.

Lacking from the existing work considering theories of plot is a strong grounding in empirical evidence or stability of the ``universal'' theories across culture.
It is precisely this shortcoming which we hope to address, by using a broad collection of Fiction stories within western culture.

\subsection{Story graphs, plot diagrams, and inferring causality}
\label{subsec:SIG-diagrams}

With the distinction between plot, structure, and emotional trajectory in mind, there have also been attempts to discover plot using data driven methods.
\cite{brewer1980event} makes the distinction between plot and structure is made even clearer.
Through experimentation with different structures, Brewer and Lichtenstein find that the resulting affect in readers is different, with some structures being considered stories and others not (the authors single out ``suspense and resolution'' and ``surprise and resolution'' as indicative of stories).

Plot units were first introduced by \cite{lehnert1981plot}, and form the basis for most all efforts that follow.

Using topic modeling, both \cite{schmidt2015plot} and \cite{jockers2013macroanalysis} find known patterns of plot across many thousand stories.
In \cite{piper2015novel}, computational analysis is applied to realize the potential of distant reading (a term owing to \cite{moretti2013distant}) to find and test scholarly insights.
In \cite{winston2011strong}, a system called ``Genesis'' is developed to compare plot summaries and infer causal connections between events, with the broad aim of the system formalized as the \textit{Strong Story Hypothesis}:
\begin{quote}
  The mechanisms that enable humans to tell, understand, and recombine stories separate human intelligence from that of other primates.
\end{quote}
In his Master's Thesis, \cite{awad2013culturally} extends the Genesis system to model differences in American and Chinese stories by adding commonsense rules that differ between cultures.
With commonsense rules, \textit{Genesis} is able to measure story coherence.

Work by \cite{regneri2010learning} learns event scripts from written descriptions of events that may not always exist in written form (implicit scripts, like shopping), using a graph-based (``temporal script graph'') algorithm and data collected on Amazon's Mechanical Turk. The algorithm is tested to detect similar events with differing descriptions.

The Analogical Story Merging (ASM) system is developed using ``Bayesian model merging'' for story categorization and is applied to 15 Russian folktales \citep{finlayson2011learning}.
The test folktales are annotated for 18 aspects of meaning by 12 annotators using a tool developed for this task.
The folktale categories defined by Vladmir Propp are predicted by ASM and the system achieves a Rand Index of 0.511 (a measure of the similarity between clusters).

In \cite{elson2012detecting} a Story Intention Graph (SIG) is developed to model stories and implemented to measure similarity and analogy.
Elson's propositional similarity metric is used to predict human judgments of story similarity and outperforms human annotation (is better than inter-annotator agreement) on Aesop's fables.

The AESOP system of \cite{goyal2013computational} converts narrative texts into their plot unit model (where plot units are ``conceptual knowledge structure to represent the affect states of and emotional tensions between characters in narrative stories'').
AESOP performs four steps: ``affect state recognition, character identification, affect state projection, and link creation.''
Performance is inspected on a set of Aesop's fables, similar to \cite{elson2012detecting}.

In \textit{Novel Devotions: Conversional Reading, Computational Modeling, and the Modern Novel}, \cite{piper2015novel} applies Multi-Dimensional Scaling (MDS) on representations of novels in a VSM (Vector Space Model --- vectors of word frequencies), and performs hierarchical clustering to understand the differences between novels and autobiographies.

\subsection{Story generation}

In \textit{Plot Induction and Evolutionary Search for Story Generation}, \cite{mcintyre2010plot} build upon their previous work to train a story planner from extracted events, their participants, and preceding relationships from a large corpus. Their system is used to to generate simple, 4 or 5 sentences stories that are mildly coherent.

The Neukom Institute at Dartmouth hosts a competition for algorithms to produce short stories, in a true-fashion Turing test \citep{digilit2016}.
In the 2016 competition, algorithms and writers were given a one-word prompt and tasked to write a 500-word short story.
The stories were then judged by a panel consisting of David Cope, Lynn Neary, and David Krakauer to be either human or machine written.
Each judge received 8 human written stories and 3 machine generated stories, one from each of the 3 entrants into the competition.
To quote their results:
\begin{quote}
  No machine won, but one submission generated by Toksu and Ibrahim on the seed “thesaurus” “fooled” one of the judges!
\end{quote}
With no first place award, the second place award and \$1000 prize was awarded to Judy Malloy whose algorithm rearranged sentences from ``Another Party in Woodside''.

\subsection{Character Identification and Networks}
\label{subsec:characters}

Much work on computational understanding of stories has focused on the extraction and analysis of character networks.
The ideas behind character networks were first examined in the original work of Moretti \citep{moretti2000conjectures,moretti2007a,schulz2011what,moretti2013distant}, and have been used widely in Digital Humanities research.
Below we highlight work that has caught our attention.

\cite{elson2010extracting} use character name chunking, quoted speech attribution and conversation detection to generate character networks from a collection of British novels.
They find a lack of support for characterizations provided by literacy scholars and suggest an alternative explanation.
Namely, the do not find support for the hypothesis that 19th century fiction novels have (1) social networks that differ by the setting of the novel (rural vs. urban) and that (2) novels with more characters have less dialogue (an inverse relationship is suggested by the so-called ``chronotype'' theories).
Instead they find that the point of view of narration (first vs. third person) is strongly correlated with the 
This work applies the distant reading philosophy by first carefully selecting a corpus of books and consulting previous literary research before doing analysis, an approach we aspire to emulate.
Elson later extended this work with models of discourse \citep{elson2012modeling}.

Bamman \etal use Bayesian models, word embedding, and state-of-the-art NLP techniques to learn personas of characters in literature \citep{bamman2014bayesian} and in film \citep{bamman2014learning}.
Their analysis is performed across a large corpora of 15,099 books selected from Hathitrust, 42,306 wikipedia movie plot summaries for film, and is shown to replicate the classification of character roles by a literary scholar.
A similar effort is undertaken by \cite{valls2014toward}, utilizing PoS annotations from syntactic parsing to detect characters in a small set of stories, and using ``action matrices'' in another attempt \citep{valls2014toward2} to encode Propp's narrative theory.
They are able to automatically detect the roles of characters within 10 folktales (developing a system they refer to as ``Voz'').

These methods have also been used to examine popular culture.
In a blog post, \cite{gabasova2015star-blog} finds the most central character in Star Wars.
\cite{xanthos2016visualising} elaborate on the method of constructing and visualizing character networks, an example of their work for Shakespeare is available as a poster: \href{http://www.martingrandjean.ch/network-visualization-shakespeare/}{http://www.martingrandjean.ch/network-visualization-shakespeare/}.
\cite{min2016narrative} perform an in-depth study of Victor Hugo's \textit{Les Misérables}, proposing using the growth of edges in and characters in the network over time to compare different works (with each edge/character curve normalized to sum to 1 at the end of each book).
More recently, \cite{wu2016interactive} has made an interactive exploration of the play Hamilton using discourse and the character network, and Meeks and Averick built an interactive exploration of the dialogue in the show Archer \citep{meeks2017data}

To compare character networks across movies, \cite{ruths2016force} uses network alignment to map characters between the \textit{Stars Wars} movies \textit{The Force Awakens} and \textit{A New Hope} revealing both expected and surprising similarities.
For example R2-D2 maps to BB-8 and Chewbacca maps to Chewbacca, as we might expect.
However, the main characters have more surprising alignments from the interaction networks, with Luke mapping to Poe, Obi-wan mapping to Kylo Ren, and Darth Vader mapping Rey.
A particular problem in using character networks that span an entire movie, TV show, or book is that multiple story lines can intersect in ways that are not accounted for by the method.
\cite{bost2016narrative} examine conversation in TV shows using a smoothing of narration to overcome the multiple narrative problem, finding protagonists more readily than using simpler interaction networks.

\subsection{Frames for NLP}
\label{subsec:frames-NLP}

The seminal work by \cite{schank1977scripts} (and earlier efforts by \cite{rumelhart1975notes}) laid the groundwork for scripts as a framework for cognitive algorithmic computation.
Research programs separately advancing AI capabilities and NLP tasks have made use of this framework.
Although existing knowledge bases such as SUMO (Niles and Pease, 2001), Cyc (Lenat, 1995) or FrameNet (Fillmore et al., 2003) contain such script-like knowledge to a certain extent, their coverage is severely limited.
Increases in computational power have realized the building of systems for script-based event detection, and there have been many efforts made in the past decade to advance such systems.
Schemata such as NarrativeML to annotate narratives are reviewed by \cite{mani2012computational}.
Next, we very briefly highlight some of these approaches, focusing particularly on the research program of Chambers due to the accessibility of the papers and the breadth of research by himself and his students.

In a series of papers \cite{chambers2007classifying,chambers2008unsupervised,chambers2009unsupervised,chambers2010database,chambers2013event} set to classify temporal relations between events, apply unsupervised learning to detect narrative event chains and entities involved, build a database of narrative schemata, and find schemata in large corpora with probability-based models.
A narrative event chain is defined as two events linked by a common actor.
Event chains are identified in text through co-reference between a single entity, ordered by a trained classifier, and all possible event chains are restricted through a clustering approach in \cite{chambers2007classifying,chambers2008unsupervised}.
Both \cite{cheung2013probabilistic} (using the proposed approach of \cite{oconnor2013learning}) and \cite{chambers2013event} utilize generative models for inducing event schemata, with the former utilizing a HMM over latent event variables and the latter using a entity-driven model.
Recent work from \cite{pichotta2015learning} improves on the baseline results of Chambers in detecting scripts using Recurrent Neural Networks (RNNs, particularly a flavor known as Long Short Term Memory (LSTM)) and architectures adapted to this task.

Corpora used by Chambers and by others include the FrameNet from \cite{baker1998berkeley}, Timebank Corpus from \cite{pustejovsky2003timebank}, Opinion Corpus from \cite{mani2006machine}, Narrative Schema Database from \cite{chambers2010database}, the Media Frames Corpus by \cite{card2015media}, and most recently the Story Cloze Dataset from \cite{mostafazadeh2016corpus}.
As an example, \cite{do2011minimally} use a primarily unsupervised approach to specifically learn causality between events in the Penn Discourse Treebank, and \cite{roemmele2017an-rnn-based} use an RNN on the Story Cloze dataset.
The understanding and generation of stories with these data sets and new models may hold promise for major advances in the field of NLP.
\cite{cambria2014jumping} has suggested that the next wave of NLP advances that aim to decode stories (a move from ``bag of words'' approaches to ``bag of narratives'') may very well be a breakthrough in understanding human nature.

Along those lines, stories have been explored as a model to training artificial intelligence systems for commonsense reasoning.
Advanced in this area all recognize and leverage the utility of stories for sense-making \citep{bex2010persuasive,bex2013values,li2012crowdsourcing,riedl2016computational}.

\subsection{Visualization}

\begin{figure}[tbp!]
  \centering
  \includegraphics[width=0.96\textwidth]{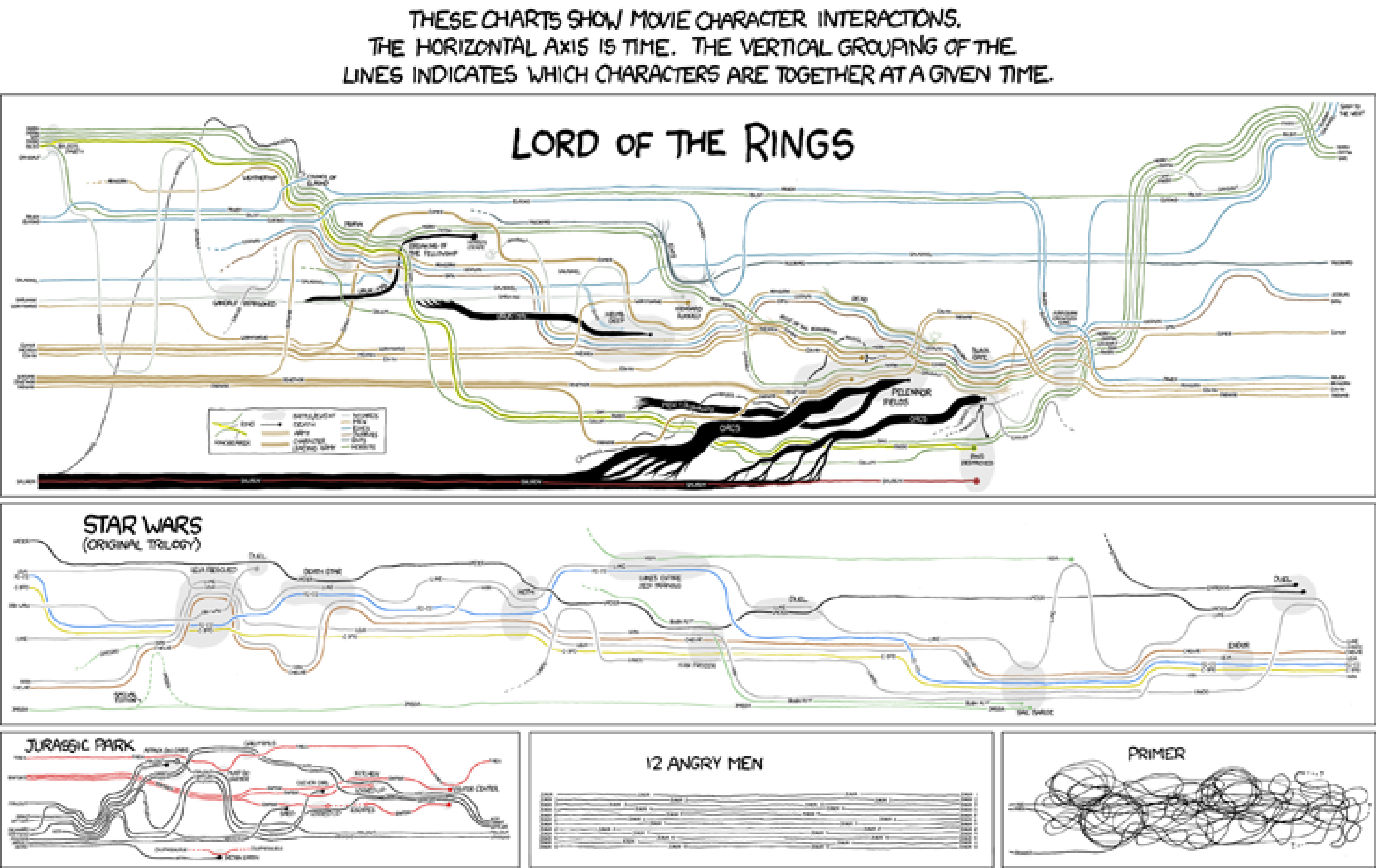}
  \caption[]{
    XKCD number 657 by \cite{munroe657} shows the time evolution of character co-occurrence in \textit{Lord of the Rings}, \textit{Star Wars}, \textit{Jurassic Park}, \textit{12 Angry Men}, and \textit{Primer}.
    Munroe adds: ``in the LotR map, up and down correspond LOOSELY to northwest and southeast respectively.''
    The width of lines correspond to the number of characters in each group, which applies here to the Orcs in \textit{Lord of the Rings}.
  }
  \label{fig:XKCD657}
\end{figure}

Stories as a model for understanding are not readily visualized, as finding a proper encoding for the mental models we use is difficult.
Nevertheless, efforts at capturing the essence of story in a visual form are omnipresent in art and automated attempts to generate such mappings are attempted (recall Figure \ref{fig:harmon}).
The illustrated movie maps of \cite{degraff2015plotted} make representations of movies in the limited space of two pages by using three dimensions to show the movement of time and place.
The web comic XKCD draws inspiration from the well-known visualization of Napoleon's march by Minard and maps the interaction of characters with time as a x-axis and character proximity as distance in the y-axis of a chart, see \cite{munroe657} and Figure \ref{fig:XKCD657}.
\cite{ogawa2010software} attempt to automatically build XKCD-style plots for software development, and an image of their reproduction of the XKCD \textit{Lord of the Rings} visualization is shown by \cite{cao2016introduction}.

\subsection{Emotional arcs}

The emotional arcs drawn by \cite{vonnegut1981palm} are simpler, using time again on the x-axis and representing the fortune of the main character in the vertical direction.
Vonnegut explicitly draws a connection between the New Testament and Cinderella, a story that has incredible popular appeal.
Other story arcs named by Vonnegut are the ``Man in the Hole'' and the ``Boy meets girl'' arcs.

With the same goal of finding commonalities between stories as \cite{vonnegut1981palm}, in a series of blog posts \cite{jockers2014novel} lays out a strategy for generating emotional arcs and eventually finds six story types using hierarchical clustering.
Our work in Chapter~\ref{chap:emotional-arcs} is an continuation of a very different core methodology that we first propose in \cite{dodds2015human}.
Though the core methodology is markedly different, we note that Jocker's first blog post appeared 10 days before the pre-print of our paper
As we note in Chapter~\ref{chap:emotional-arcs} as well, the distinction between plot and emotional arc as well as correct use of using sentiment analysis tools distinguish our contributions from those of \cite{jockers2014novel}.

Attempts to analyze plot more directly than emotional arc have been increasing in the past few years.
\cite{cherny2013bones} applies machine learning over a bag-of-words analysis to predict action and sex scenes using Naive Bayes (NB) and Stochastic Gradient Descent (SGD).
Training data is crowd-sourced from two ratings of 500 word chunks on the survey platform Mechanical Turk (MT), and Cherny develops novel visualizations of the relationships between topics in chapters.
\cite{reiter2014nlp} use an unsupervised method to generate and compare event-based representations of rituals and folktales, but we were unable to obtain their manuscript.
\cite{piper2015novel} analyzes the differences between the first and second half of novels about ``conversion.'' 
We revisit the approach by \cite{schmidt2015plot} here: he uses Latent Semantic Analysis (LSA) and dimensionality reduction to find patterns of plot in a reduced 2-dimensional topic space.
While this is an interesting approach, we would not expect the coefficients of the first two modes in the SVD to hold particular relationships between themselves.
Most recently, the approach of measuring sentiment using sentences and smoothing has been published by \cite{gao2016multiscale}.

The most similar approach to ours (perhaps based on our method from \cite{dodds2015human}, though they cite Vonnegut) was an effort by sentiment analysis startup Indico's Dan Kuster, available at \href{https://indico.io/blog/plotlines/}{https://indico.io/blog/plotlines/} \citep{kuster2015exploring}.
Kuster uses sliding windows and dynamic time warping as a distance metric between emotional arcs, and on single books the method is indeed very similar to ours, yet they don't extend to mine for patterns across a large corpus.

\subsection{Suzyhet and validation}

The work of \cite{jockers2014novel} has been publicly debated in the online sphere.
The back-and-forth between Matt Jockers and Annie Swafford (and others) has happened in blog posts \citep{swafford2015problems}, comments on blogs, and on Twitter.
The extent of this debate is documented in two parts by \cite{clancy2015fabula} (available online: \href{https://storify.com/clancynewyork/contretemps-a-syuzhet}{https://storify.com/clancynewyork/contretemps-a-syuzhet} and \href{https://storify.com/clancynewyork/a-fabula-of-syuzhet-ii}{https://storify.com/clancynewyork/a-fabula-of-syuzhet-ii}).
We attempt to briefly summarize some of the discussion of prominent scholars in digital humanities and how this relates to our own work on emotional arcs, particularly the comments of Bamman, Piper, Schmidt, Enderle, and Underwood.

\begin{figure}[tbp!]
  \centering
  \includegraphics[width=0.96\textwidth]{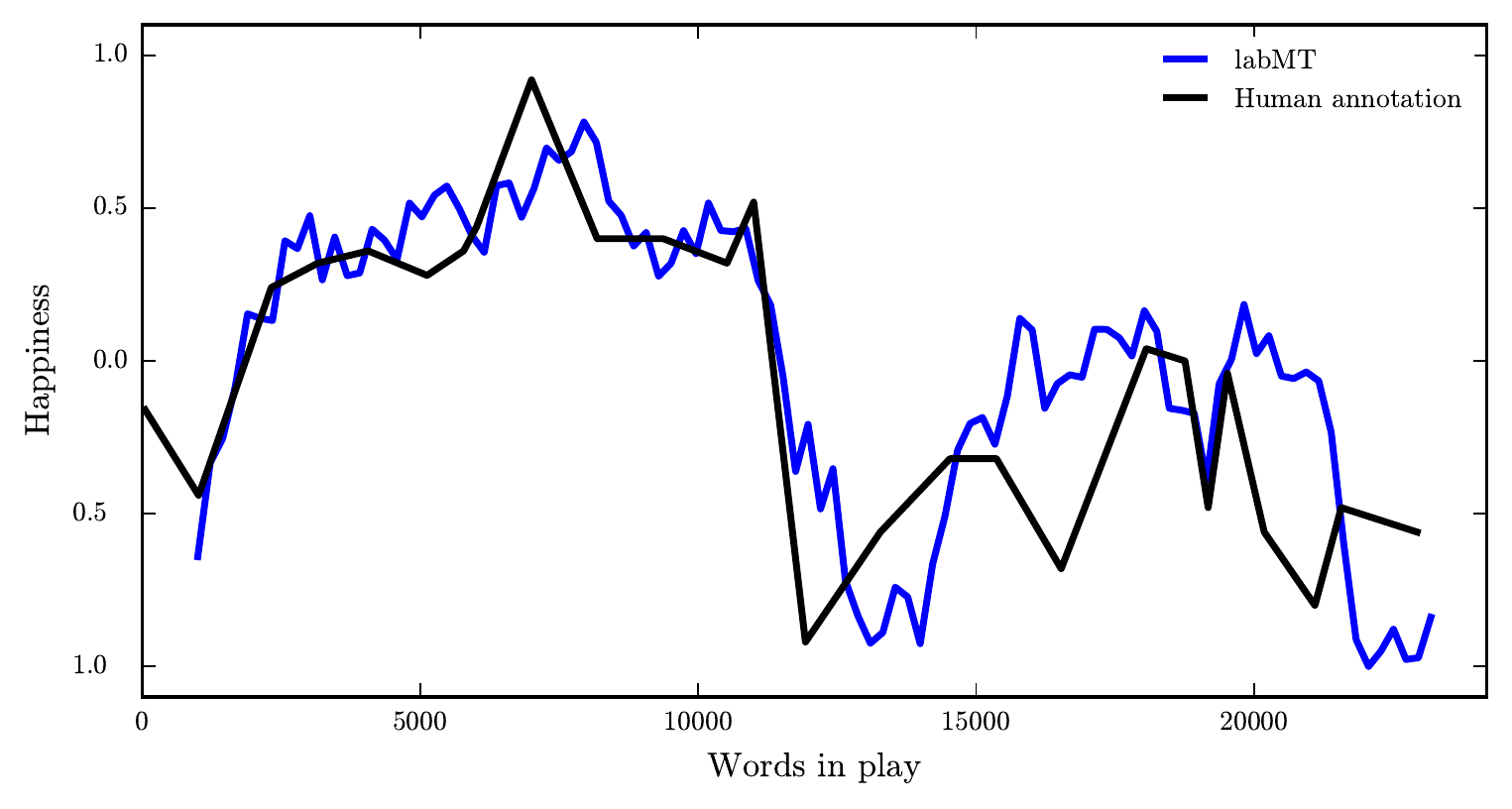}
  \caption[]{
    Emotional arcs of Shakespeare's \textit{Romeo and Juliet}, generated with the labMT sentiment dictionary and the average of 5 human annotations on each scene.
    The labMT approach generated 100 time points, with 2000 rated words at each point shown, $\delta _h = 1$, and ignoring scene boundaries (the same approach used in general).
    The human annotation data is from a survey conducted in \cite{bamman2015validity} with 5 responses for each of the 26 scenes in the play, points are shown on the x-axis in the center of each scene's words.
    The survey collected responses from -5 to 5, which we have re-scaled linearly to -1 to 1 (by dividing by 5), and the labMT data is re-scaled by first mean centering the time series, then multiplying by the inverse of the absolute maximum (such that the time series will touch -1 or 1 in the direction of the absolute maximum).
  }
  \label{fig:bamman-validation}
\end{figure}

\cite{bamman2015validity} elaborates on the discussion around on how to measure validity of emotional arcs
\cite{bamman2015validity} goes on to build a survey to perform the validation proposed by \cite{piper2015validation} and \cite{weingart2015not}.
Bamman's survey for Shakespeare's Romeo and Juliet takes responses from 5 participants on Mechanical Turk for each scene on a -5 to 5 scale along with a free text reasoning for the score.
We plot the mean of these ratings along with our measure of the emotional arc (the happiness of the words in the play for a sliding window of 10000 words and 200 time points) of the play in Figure \ref{fig:bamman-validation}.
This approach could, of course, be extended to provide additional formal validation of the methods and parameters used in our study of emotional arcs.
However, non-expert annotations are not always a proper gold-standard \citep{snow2008cheap}, and there may even be (we might even expect) valid interpretations of a story that produce different emotional responses.
In this case, we would expect that our automated method would find one of these arcs, and the goal of a more advanced system could be to find more than one arc for a given book.

In addition to the problems identified by Swafford, \cite{schmidt2015commodius} builds on \cite{enderle2015sine} and highlights the problem that the low pass filter needs to be circular.
These discussions have provided many interesting future directions for this work and the validation of computational approaches to narratives.

Our own work on emotional arcs (Chapter~\ref{chap:emotional-arcs}) has attracted a great deal of popular attention and has been noticed by those in the digital humanities community, particularly by \cite{schmidt2016plot} and \cite{scott2016brownian}.
We address the concerns raised in both of these critical takes in our work.
Drawing directly from the suggestions from Schmidt, we utilize the Library of Congress classification to produce a better selection of texts from Project Gutenberg in our published manuscript, a notable improvement from the pre-print corpus he analyzes.
In our treatment, we carefully consider the choice of a suitable null hypothesis to test whether there is structure in the emotional arcs of real stories.
Our first pass used the emotional arcs of the same books with randomly shuffled words (``word salad'' books), for a corpora that has no narrative structure but the same emotional words.
The final version of our null model generates stories from a bigram Markov chain trained on the actual text.
These ``nonsense'' narratives have no real structure, but resemble written English.
For more complete details and sample text from each method, see Section \ref{sec:construction}.
Other reasonable attempts could consider shuffling sentences of paragraphs, however Brownian noise and arbitrary random walks are not sensible comparisons.
In particular, the singular value spectrum of Brownian noise is arbitrary.

In the next Chapter, we test sentiment analysis methods for performance in classification and providing understanding of emotional text, methods that form the basis of our study into emotional arcs.

\chapter{
Benchmarking sentiment analysis methods for large-scale texts:
A case for using continuum-scored words and word shift graphs.
}
\chaptermark{Benchmarking}
\label{chap:benchmarking}

\begin{quote}
The emergence and global adoption of social media
has rendered possible
the real-time estimation of population-scale sentiment,
which has profound implications for
our understanding of human behavior.
Given the growing assortment of sentiment-measuring instruments,
it is imperative to understand which aspects of sentiment dictionaries contribute to both their
classification accuracy
and
their ability to provide richer understanding of texts.
Here, we perform detailed, quantitative tests and qualitative assessments of 6 dictionary-based methods
applied to 4 different corpora, and briefly examine a further 20 methods.
We show that while inappropriate for sentences, dictionary-based methods are generally robust in their classification accuracy for longer texts.
Most importantly they can aid understanding of texts
with reliable and meaningful word shift graphs if
(1)
the dictionary covers a sufficiently large
portion of a given text's lexicon when weighted by word usage frequency;
and
(2) 
words are scored on a continuous scale.
\end{quote}

\section{Introduction}

As we move further into what might be called the Sociotechnocene---with
increasingly more interactions, decisions, and impact being made
by globally distributed people and algorithms---the myriad human
social dynamics that have shaped our history have become
far more visible and measurable than ever before.
Driven by the broad implications of being able to characterize social
systems in microscopic detail,
sentiment detection for populations at
all scales has become
a prominent research arena.
Attempts to leverage online expression for
sentiment mining include
prediction of stock markets~\citep{bollen2011twitter,si2013exploiting,chung2011predicting,ruiz2012correlating}, assessing responses to advertising, real-time monitoring of global happiness~\citep{dodds2015human}, and measuring a health-related quality of life~\citep{alajajian2015lexicocalorimeter}.
The diverse set of instruments produced by this work now provide indicators
that help scientists understand collective behavior,
inform public policy makers,
and, in industry,
gauge the sentiment of public response to marketing campaigns.
Given their widespread usage and potential to influence social systems,
understanding how these instruments perform and how they compare with each other has become imperative.
Benchmarking their ability to provide insight into sentiment, and their performance, both focuses future development and provides practical advice to non-experts in selecting a sentiment dictionary.

We identify sentiment detection methods as
belonging to one of three categories, each carrying their own advantages and
disadvantages:
\begin{enumerate}
\item
  Dictionary-based methods~\citep{dodds2015human,bradley1999affective,pennebaker2001linguistic,wilson2005recognizing,liu2010sentiment,warriner2013norms},
\item
  Supervised learning methods~\citep{liu2010sentiment},
  and
\item
  Unsupervised (or deep) learning methods~\citep{socher2013a}.
\end{enumerate}

Here, we focus on dictionary-based methods,
which all center around
the determination of a text $T$'s average happiness (sometimes referred to as \textit{valence})
with sentiment dictionary $D$ through the equation:
\begin{equation}
  h_\textnormal{D}^T =
  \frac{
    \sum_{w\in D}
    h_{\textnormal{D}}(w)
    \cdot
    f ^T (w)
  }
  {
    \sum_{w\in D}
    f^T (w)
  }
  =
  \sum_{w \in D}
  h_\textnormal{D} (w)
  \cdot
  p^T (w),
  \label{eq:havg}
\end{equation}
where we denote each of the words in a given sentiment dictionary $D$ as words $w$,
word sentiment scores as $h_\textnormal{D}(w)$,
word frequency as $f^T(w)$,
and normalized frequency of $w$ in $T$ as
$
p^T (w)
=
f^T (w)
/
\sum _{w\in D}
f^T (w)
$.
In this way, we measure the happiness of a text
in a manner analogous to taking the temperature of a room.
While other simple sentiment metrics may be considered, we will see that
analyzing individual word contributions is important and that this
equation allows for a straightforward, meaningful interpretation.

Dictionary-based methods offer two
distinct advantages which we find necessary: (1) they are in principle
corpus agnostic (applicable to corpora without ground truth data available) and
(2) in contrast to black box (highly non-linear) methods, they offer
the ability to ``look under the hood'' at words contributing to a
particular score through \textit{word shift graphs} (defined fully later; see
also ~\citep{dodds2009b,dodds2011temporal}).
Indeed, if we are at all concerned with understanding why a particular
scoring method
varies---e.g,, our undertaking is scientific---then word shift graphs are
essential tools.
In the absence of word shift graphs, or similar devices, any explanation of sentiment trends
is missing crucial information and rises only to the level of opinion
or guesswork~\citep{golder2011diurnal,garcia2015language,dodds2015reply,wojcik2015conservatives}.

As all methods must, dictionary-based ``bag-of-words'' approaches suffer
from various drawbacks, and three are worth stating up front.
First, they are only applicable to corpora of
sufficient size, well beyond that of a single sentence \citep{ribeiro2016sentibench}
(widespread usage in this misplaced fashion does not suffice as a counterargument).
We directly verify this assertion on individual Tweets, finding that some sentiment dictionaries perform admirably, however the average (median) F1-score on the STS-Gold data set is 0.50 (0.54) from all datasets (Table \ref{tbl:STS}), others having shown similar results for dictionary methods with short text \citep{ribeiro2016sentibench}.
Second, state-of-the-art learning methods with a sufficiently large training set for a specific corpus
will outperform dictionary-based methods on same corpus \citep{liu2012sentiment}.
However, in practice the domains and topics to which sentiment
analysis
are applied are highly varied, such that training to a high degree of
specificity for a single corpus may not be practical and, from
a scientific standpoint, will severely constrain attempts to detect
and understand universal patterns.
Third, words may be evaluated out of context or with the wrong sense.
A simple example is the word ``miss'' occurring frequently
when evaluating articles in the Society section of the New York Times.
This kind of contextual error
is something we can readily identify and correct for through word shift graphs,
but would remain hidden to users of nonlinear learning methods.

We lay out our paper as follows.
We list and describe the dictionary-based
methods we consider in \revtexlatexswitch{Sec.~\ref{sec:dictionariesandcorpora}}{Sec.~Dictionaries,~Corpora,~and~Word~Shift~Graphs},
and outline the corpora we use for tests
in
\revtexlatexswitch{Sec.~\ref{subsec:benchmarks}}{Subsec.~Corpora~Tested}.
We present our results in
\revtexlatexswitch{Sec.~\ref{sec:results}}{Sec.~Results},
comparing all methods in how they perform
for specific analyses of
the New York Times (NYT)
(\revtexlatexswitch{Sec.~\ref{subsec:NYTwordshift}}{Subsec.~New~{Y}ork~{T}imes~Word~Shift~Analysis}),
movie reviews
(\revtexlatexswitch{Sec.~\ref{subsec:moviereviews}}{Subsec.~Movie~{R}eviews~Classification~and~Word~Shift~Analysis}),
Google Books
(\revtexlatexswitch{Sec.~\ref{subsec:googlebooks}}{Subsec.~Google~{B}ooks~Time~Series~and~Word~Shift~Analysis}),
and
Twitter
(\revtexlatexswitch{Sec.~\ref{subsec:twittertimeseries}}{Subsec.~Twitter~Time~Series~Analysis}).
In \revtexlatexswitch{Sec.~\ref{subsec:NB-section}}{Subsec.~Brief~Comparison~to~Machine~Learning~Methods}, we make
some basic comparisons between dictionary-based methods and machine learning approaches.
We provide concluding remarks in \revtexlatexswitch{Sec.~\ref{sec:conclusion}}{Sec.~Conclusion}
and bolster our findings with figures, tables, and additional analysis in the Supporting Information.

\section{Sentiment Dictionaries, Corpora, and Word Shift Graphs}
\label{sec:dictionariesandcorpora}

\begin{sidewaystable*}[tbp!]
  \begin{adjustwidth}{\pnastableadjust in}{0in}
{\scriptsize
  \begin{tabular*}{\linewidth}{ l | l | l | l | l | l}
      \hline
    Dictionary & \# Entries & Range & Construction & License & Ref.\\
    \hline
    \hline
    labMT & 10222 & 1.3 $\to$ 8.5 & Survey: MT, 50 ratings & CC & \citep{dodds2015human}\\
    ANEW & 1034 & 1.2 $\to$ 8.8 & Survey: FSU Psych 101 & Free for research & \citep{bradley1999affective}\\
    LIWC07 & 4483 & [-1,0,1] & Manual & Paid, commercial & \citep{pennebaker2001linguistic}\\
    MPQA & 7192 & [-1,0,1] & Manual + ML & GNU GPL & \citep{wilson2005recognizing}\\
    OL & 6782 & [-1,1] & Dictionary propagation & Free & \citep{liu2010sentiment}\\
    WK & 13915 & 1.3 $\to$ 8.5 & Survey: MT, 14--18 ratings & CC & \citep{warriner2013norms}\\
    \hline
    LIWC01 & 2322 & [-1,0,1] & Manual & Paid, commercial & \citep{pennebaker2001linguistic}\\
    LIWC15 & 6549 & [-1,0,1] & Manual & Paid, commercial & \citep{pennebaker2001linguistic}\\
    PANAS-X & 20 & [-1,1] & Manual & Copyrighted paper & \citep{watson1999panas}\\
    Pattern & 1528 & -1.0 $\to$ 1.0 & Unspecified & BSD & \citep{de2012pattern}\\
    SentiWordNet & 147700 & -1.0 $\to$ 1.0 & Synset synonyms & CC BY-SA 3.0 & \citep{baccianella2010sentiwordnet}\\
    AFINN & 2477 & [-5,-4, $\ldots$,4,5] & Manual & ODbL v1.0 & \citep{nielsen2011new}\\
    GI & 3629 & [-1,1] & Harvard-IV-4 & Unspecified & \citep{stone1966general}\\
    WDAL & 8743 & 0.0 $\to$ 3.0 & Survey: Columbia students & Unspecified & \citep{whissell1986dictionary}\\
    EmoLex & 14182 & [-1,0,1] & Survey: MT & Free for research & \citep{mohammad2013crowdsourcing}\\
    MaxDiff & 1515 & -1.0 $\to$ 1.0 & Survey: MT, MaxDiff & Free for research & \citep{kiritchenko2014sentiment}\\
    HashtagSent & 54129 & -6.9 $\to$ 7.5 & PMI with hashtags & Free for research & \citep{zhu2014nrc}\\
    Sent140Lex & 62468 & -5.0 $\to$ 5.0 & PMI with emoticons & Free for research & \citep{MohammadKZ2013}\\
    SOCAL & 7494 & -30.2 $\to$ 30.7 & Manual & GNU GPL & \citep{taboada2011lexicon}\\
    SenticNet & 30000 & -1.0 $\to$ 1.0 & Label propogation & Citation requested & \citep{cambria2014senticnet}\\
    Emoticons & 132 & [-1,0,1] & Manual & Open source code & \citep{gonccalves2013comparing}\\
    SentiStrength & 2615 & [-5,-4, $\ldots$,4,5] & LIWC+GI & Free for research & \citep{thelwall2010sentiment}\\
    VADER & 7502 & -3.9 $\to$ 3.4 & MT survey, 10 ratings & Freely available & \citep{hutto2014vader}\\
    Umigon & 927 & [-1,1] & Manual & Public Domain & \citep{levallois2013umigon}\\
    USent & 592 & [-1,1] & Manual & CC & \citep{pappas2013distinguishing}\\
    EmoSenticNet & 13188 & [-10,-2,-1,0,1,10] & Bootstrapped extension & Non-commercial & \citep{poria2013enhanced}\\
\end{tabular*}}  \end{adjustwidth}
  \caption{
    Summary of dictionary attributes used in sentiment measurement instruments.
    We provide all acronyms and abbreviations and further information
    regarding sentiment dictionaries in \revtexlatexswitch{Sec.~\ref{subsec:dictionaries}}{Subsec.~Dictionaries}.
    We test the first 6 dictionaries extensively.
                            The range indicates whether scores are continuous or binary (we
    use the term binary for sentiment dictionaries for which words
    are scored as $\pm 1$ and optionally 0).
  }
\label{tbl:summary}
\end{sidewaystable*}

\subsection{Sentiment Dictionaries}
\label{subsec:dictionaries}

The words ``sentiment dictionary,'' ``lexicon,'' and ``corpus'' are often used interchangeably, and for clarity we define our usage as follows.

\begin{description} \itemsep1pt \parskip1pt \parsep0pt
\item[Sentiment Dictionary:] Set of words (possibly including word stems) with ratings.
\item[Corpus:] Collection of texts which we seek to analyze.
\item[Lexicon:] The words contained within a corpus (often said to be ``tokenized'').
\end{description}

We test the following six sentiment dictionaries in depth:

\begin{description} \itemsep1pt \parskip1pt \parsep0pt
\item[labMT] --- language assessment by Mechanical Turk~\citep{dodds2015human}.
\item[ANEW] --- Affective Norms of English Words~\citep{bradley1999affective}.
\item[WK] --- Warriner and Kuperman rated words from SUBTLEX by Mechanical Turk~\citep{warriner2013norms}.
\item[MPQA] --- The Multi-Perspective Question Answering (MPQA) Subjectivity Dictionary~\citep{wilson2005recognizing}.
\item[LIWC\{01,07,15\}] --- Linguistic Inquiry and Word Count, three versions~\citep{pennebaker2001linguistic}.
\item[OL] --- Opinion Lexicon, developed by Bing Liu~\citep{liu2010sentiment}.
\end{description}

We also make note of 18 other sentiment dictionaries:
\begin{description} \itemsep1pt \parskip1pt \parsep0pt
  \item[PANAS-X] --- The Positive and Negative Affect Schedule --- Expanded \citep{watson1999panas}.
    \item[Pattern] --- A web mining module for the Python programming language, version 2.6 \citep{de2012pattern}.
    \item[SentiWordNet] --- WordNet synsets each assigned three sentiment scores: positivity, negativity, and objectivity \citep{baccianella2010sentiwordnet}.
    \item[AFINN] --- Words manually rated -5 to 5 with impact scores by Finn Nielsen \citep{nielsen2011new}.
    \item[GI] --- General Inquirer: database of words and manually created semantic and cognitive categories, including positive and negative connotations \citep{stone1966general}.
    \item[WDAL] --- Whissel's Dictionary of Affective Language: words rated in terms of their Pleasantness, Activation, and Imagery (concreteness) \citep{whissell1986dictionary}.
    \item[EmoLex] --- NRC Word-Emotion Association Lexicon: emotions and sentiment evoked by common words and phrases using Mechanical Turk \citep{mohammad2013crowdsourcing}.
    \item[MaxDiff] --- NRC MaxDiff Twitter Sentiment Lexicon: crowdsourced real-valued scores using the MaxDiff method \citep{kiritchenko2014sentiment}.
    \item[HashtagSent] --- NRC Hashtag Sentiment Lexicon: created from Tweets using Pairwise Mutual Information with sentiment hashtags as positive and negative labels (here we use only the unigrams) \citep{zhu2014nrc}.
    \item[Sent140Lex] --- NRC Sentiment140 Lexicon: created from the ``sentiment140'' corpus of Tweets, using Pairwise Mutual Information with emoticons as positive and negative labels (here we use only the unigrams) \citep{MohammadKZ2013}.
    \item[SOCAL] --- Manually constructed general-purpose sentiment dictionary \citep{taboada2011lexicon}.
    \item[SenticNet] --- Sentiment dataset labeled with semantics and 5 dimensions of emotions by Cambria \etal, version 3 \citep{cambria2014senticnet}.
    \item[Emoticons] --- Commonly used emoticons with their positive, negative, or neutral emotion \citep{gonccalves2013comparing}.
    \item[SentiStrength] --- an API and Java program for general purpose sentiment detection (here we use only the sentiment dictionary) \citep{thelwall2010sentiment}.
    \item[VADER] --- method developed specifically for Twitter and social media analysis \citep{hutto2014vader}.
    \item[Umigon] --- Manually built specifically to analyze Tweets from the sentiment140 corpus \citep{levallois2013umigon}.
    \item[USent] --- set of emoticons and bad words that extend MPQA \citep{pappas2013distinguishing}.
    \item[EmoSenticNet] --- extends SenticNet words with WNA labels \citep{poria2013enhanced}.
\end{description}

All of these sentiment dictionaries were produced by academic groups, and with the exception of LIWC, they are provided free of charge.
In Table~\ref{tbl:summary},
we supply the main aspects---such as word count,
score type (continuum or binary), and license information---for
the sentiment dictionaries listed above.
In the GitHub repository associated with our paper,
\url{https://github.com/andyreagan/sentiment-analysis-comparison},
we include all of the sentiment dictionaries except LIWC.

The labMT, ANEW, and WK sentiment dictionaries have scores ranging on a continuum from 1 (low happiness) to 9 (high happiness) with 5 as neutral, whereas the others we test in detail have scores of $\pm 1$, and either explicitly or implicitly 0 (neutral).
We will refer to the latter sentiment dictionaries as being binary, even if neutral is included.
Other non-binary ranges include a continuous scale from -1 to 1 (SentiWordNet), integers from -5 to 5 (AFINN), continuous from 1 to 3 (GI), and continuous from -5 to 5 (NRC).
For coverage tests, we include all available words, to gain a full sense of the breadth of each sentiment dictionary.
In scoring, we do not include neutral words from any sentiment dictionary.

We test the labMT, ANEW, and WK dictionaries for a range of stop words (starting with the removal of words scoring within $\Delta_{h} = 1$ of the neutral score of 5)~\citep{dodds2011temporal}.
The ability to remove stop words---a common practice for text pre-processing---is one advantage of dictionaries that have a range of scores, allowing us to tune the instrument for maximum performance, while retaining all of the benefits of a dictionary method.
We will show that, in agreement with the original paper introducing labMT and looking at Twitter data, a $\Delta_{h} = 1$ is a pragmatic choice in general~\citep{dodds2011temporal}.

Since we do not apply a part of speech tagger, when using the MPQA
dictionary we are obliged to exclude words with scores of both +1 and -1.
The words and stems with both scores are: blood, boast* (we denote
stems with an asterisk), conscience, deep, destiny, keen, large, and precious.
We choose to match a text's words using
the fixed word set from each sentiment dictionary before stems,
hence words with overlapping matches (a fixed word that also matches a
stem) are first matched by the fixed word.

\subsection{Corpora Tested}
\label{subsec:benchmarks}

For each sentiment dictionary, we test both the coverage and the ability to detect previously observed and/or known patterns within each of the following corpora, noting the pattern we hope to discern:
\begin{enumerate}
  \itemsep1pt \parskip1pt \parsep0pt
\item
  The New York Times (NYT)~\citep{nytimescorpus2008new}: Goal of understanding differences between sections and ranking by sentiment (\revtexlatexswitch{Sec.~\ref{subsec:NYTwordshift}}{Subsec.~New~{Y}ork~{T}imes~Word~Shift~Analysis}).
\item
  Movie reviews~\citep{pang2004sentimental}: Goal of discerning how emotional language differs in positive and negative reviews and how these differences influence classification accuracy (\revtexlatexswitch{Sec.~\ref{subsec:moviereviews}}{Subsec.~Movie~{R}eviews~Classification~and~Word~Shift~Analysis}).
\item
  Google Books~\citep{lin2012syntactic}: Goal of understanding time series (\revtexlatexswitch{Sec.~\ref{subsec:googlebooks}}{Subsec.~Google~{B}ooks~Time~Series~and~Word~Shift~Analysis}).
\item
  Twitter: Goal of understanding time series (\revtexlatexswitch{Sec.~\ref{subsec:twittertimeseries}}{Subsec.~Twitter~Time~Series~Analysis}).
    \end{enumerate}

For the corpora other than the movie reviews and small numbers of tagged Tweets, there is no publicly available ground
truth sentiment, so we instead make comparisons between methods
and examine how words contribute to scores.
We note that measuring how patterns of sentiment compares with societal measures of well being would also
be possible~\citep{mitchell2013happiness}.
We offer greater detail on corpus processing below,
and we also provide the relevant scripts on GitHub at
\url{https://github.com/andyreagan/sentiment-analysis-comparison}.

\subsection{Word Shift Graphs}
\label{subsec:wordshifts}

Sentiment analysis is often applied to classify text as positive or negative.
Indeed if this were the only use case, the value added by sentiment analysis would be severely limited.
Instead we use sentiment analysis as a lens that allow us to see how the emotive words in a text shape the overall content.
This is accomplished by first analyzing each word to find its individual contribution to the difference in sentiment scores between two texts.
Most importantly, the final step is to examine the words themselves, ranked by their individual contribution.
Of the four corpora that we analyze, three rely on this type of qualitative analysis: using the sentiment dictionary as a tool to better understand the sentiment of the corpora rather than as a binary classifier.

To make this possible, we must first find the contribution of each word individually.
Starting with the ANEW sentiment dictionary and two texts which we label reference and comparison, we take the difference of their sentiment scores $h^{\textrm{(comp)}}_{\textrm{ANEW}}$ and $h^{\textrm{(ref)}}_{\textrm{ANEW}}$, rearrange a few things, and arrive at
\[
  h^{\textrm{comp}}_{\textrm{ANEW}} - h^{\textrm{ref}}_{\textrm{ANEW}}
  =
  \sum_{w \in ANEW}
  \underbrace{
  \left[
  h_{\textrm{ANEW}} {(w)} - h^{\textrm{ref}}_{\textrm{ANEW}}
  \right]
  }_{+/-}
  \underbrace{
  \left[
  p^{\textrm{comp}}(w) - p^{\textrm{ref}}(w)
  \right]
    }_{\uparrow/\downarrow}
  \]
  Each word $w$ in the summation contributes to the sentiment difference between the texts according to (1) its sentiment relative
  to the reference text ($+/-$ = more/less emotive),
  and (2) its change in frequency of usage ($\uparrow/\downarrow$ = more/less frequent).
  As a first step, it is possible to visualize this sorted word list in a table, along with the basic indicators of how its contribution is constituted.
  We use word shift graphs to present this information in the most accessible manner to advanced users.
    For further detail, we refer the reader to our instructional post and video at \href{http://www.uvm.edu/storylab/2014/10/06/hedonometer-2-0-measuring-happiness-and-using-word-shifts/}{http://www.uvm.edu/storylab/2014/10/06/}.
\section{Results}
\label{sec:results}

\begin{figure*}[tbp!]
  \centering
\includegraphics[width=0.98\textwidth]{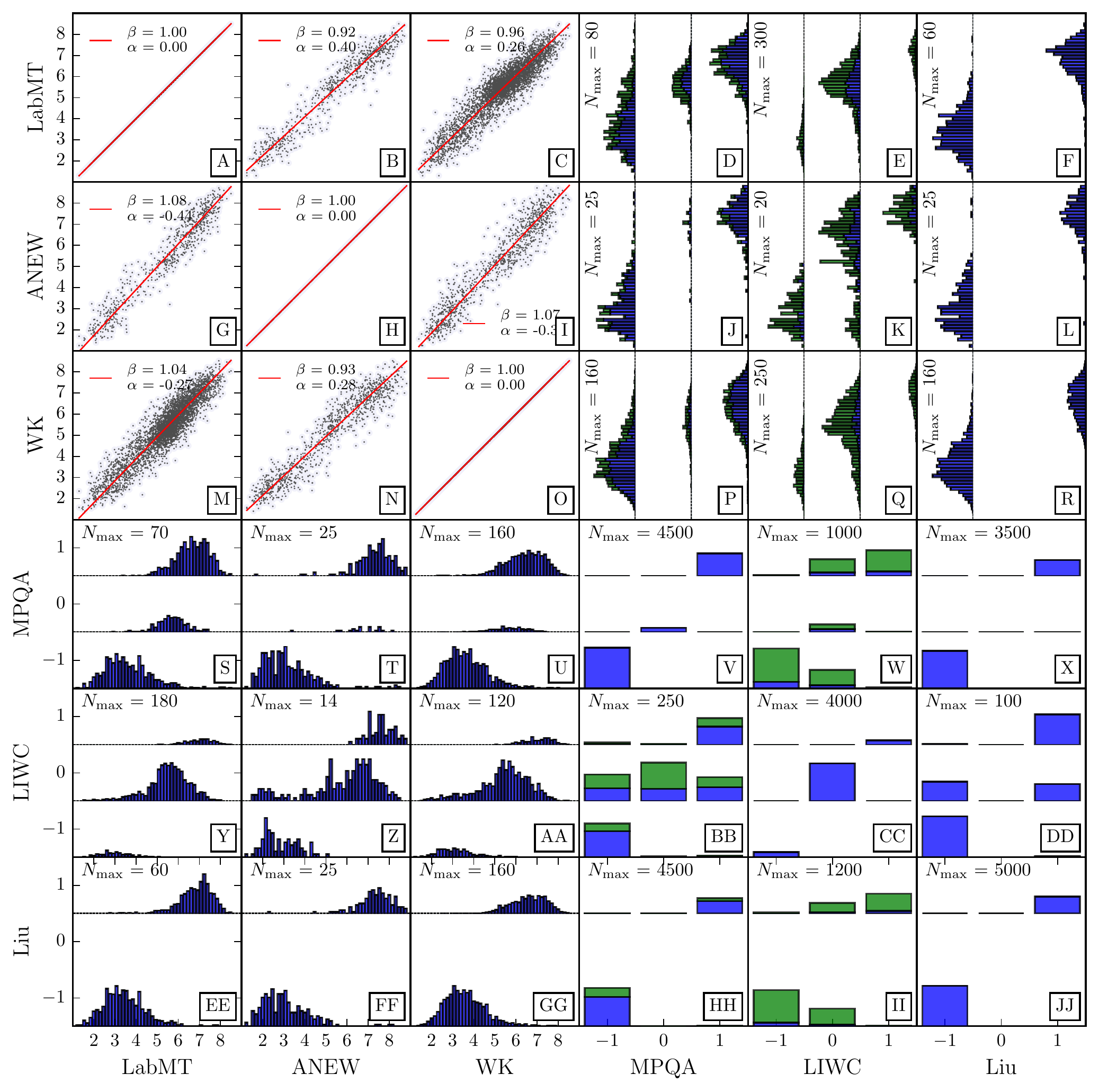}
  \caption{
    Direct comparison of the words in each of the dictionaries tested. For the comparison of two dictionaries, we plot words that are matched by the independent variable ``$x$'' in the dependent variable ``$y$''.
    Because of this, and cross stem matching, the plots are not symmetric across the diagonal of the entire figure.
    Where the scores are continuous in both dictionaries, we compute the RMA linear fit.
    When a sentiment dictionary contains both fixed and stem words, we plot the matches by fixed words in blue and by stem words in green.
    The axes in the bar plots are not of the same height, due to large mismatches in the number of words in the dictionaries, and we note the maximum height of the bar in the upper left of such plots.
    Detailed analysis of Panel C can be found in~\citep{dodds2015reply}.
    We provide a table for each off-diagonal panel in the S2 Appendix with the words whose scores exhibit the greatest mismatch, and a subset of these tables in Fig~\ref{fig:labMT-tables}.
  }
  \label{fig:main}
\end{figure*}

\begin{figure*}[tbp!]
\includegraphics[width=.98\textwidth]{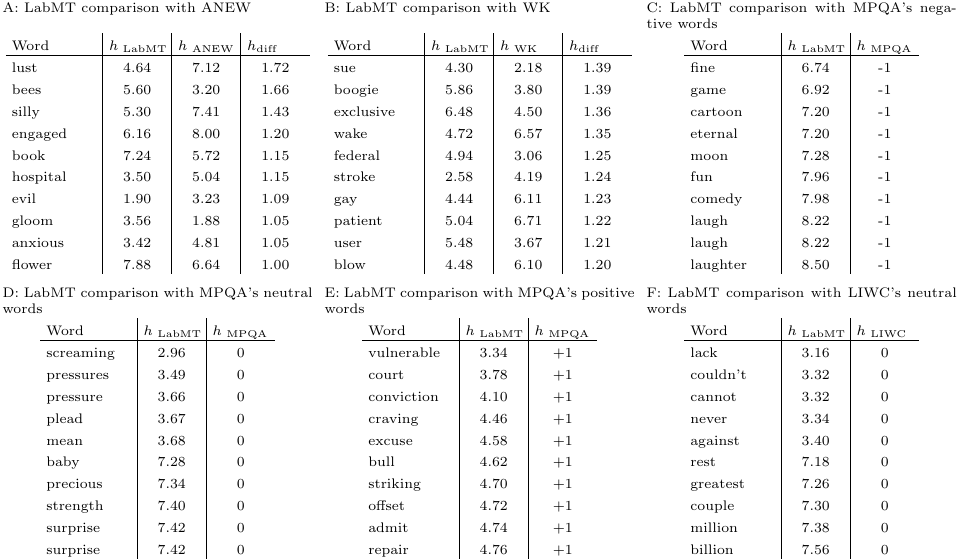}
    \caption{
      We present the specific words from Panels G, M, S and Y of Fig~\ref{fig:main} with the greatest mismatch.
      Only the center histogram from Panel Y of Fig~\ref{fig:main} is included.
      We emphasize that the labMT dictionary scores generally agree well with the other dictionaries, and we are looking at the marginal words with the strongest disagreement.
      Within these words, we detect differences in the creation of these dictionaries that carry through to these edge cases.
      Panel A: The words with most different scores between the labMT and ANEW dictionaries are suggestive of the different meanings that such words entail for the different demographic surveyed to score the words.
      Panel B: Both dictionaries use surveys from the same demographic (Mechanical Turk), where the labMT dictionary required more individual ratings for each word (at least 50, compared to 14) and appears to have dampened the effect of multiple meaning words.
      Panels C--E: The words in labMT matched by MPQA with scores of -1, 0, and +1 in MPQA show that there are at least a few words with negative rating in MPQA that are not negative (including the happiest word in the labMT dictionary: ``laughter''), not all of the MPQA words with score 0 are neutral, and that MPQA's positive words are mostly positive according to the labMT score.
      Panel F: The function words in the expert-curated LIWC dictionary are not emotionally neutral.
  }
  \label{fig:labMT-tables}
\end{figure*}

In Fig~\ref{fig:main}, we show a direct comparison between word
scores for each pair of the 6 dictionaries tested.
Overall, we find strong agreement between all dictionaries
with the exceptions we note below.
As a guide, we will provide more detail on the individual comparison
between the labMT dictionary and the other five dictionaries by
examining the words whose scores disagree across dictionaries shown in Fig~\ref{fig:labMT-tables}.
We refer the reader to the S2 Appendix for the remaining individual comparisons.

To start with, consider the comparison of the labMT and ANEW dictionaries on a word-for-word basis.
Because these dictionaries share the same range of values, a scatterplot is the
natural way to visualize the comparison.
Across the top row of Fig~\ref{fig:main}, which compares labMT to the
other 5 dictionaries, we see in Panel B for the labMT-ANEW comparison
that the RMA best fit~\citep{rayner1985linear} is
\begin{equation*}
  h_{\textrm{labMT}}(w) = 0.92*h_{\textnormal{ANEW}}(w) + 0.40
\end{equation*}
for words $w$ in both labMT and ANEW.
The 10 words with farthest from the line of best fit shown in Panel B of
Fig~\ref{fig:labMT-tables} are (with labMT, ANEW scores in parenthesis):
lust (4.64, 7.12),
bees (5.60, 3.20),
silly (5.30, 7.41),
engaged (6.16, 8.00),
book (7.24, 5.72),
hospital (3.50, 5.04),
evil (1.90, 3.23),
gloom (3.56, 1.88),
anxious (3.42, 4.81),
and flower (7.88, 6.64).
We observe that these words have
high standard deviations in labMT.
While the overall agreement is very good, we should expect some variation
in the emotional associations of words, due to chance, time of survey, and
demographic variability.
Indeed, the Mechanical Turk users who scored the words for the labMT
set in 2011
are evidently different from the University of Florida
students who took the ANEW survey in 2000.

Comparing labMT with WK in Panel C of Fig~\ref{fig:main}, we again
find a fit with slope near 1, and with a smaller positive shift:
$h_{\textnormal{labMT}}(w) = 0.96*h_{\textnormal{WK}}(w)+0.26$.
The 10 words farthest from the best fit line, shown in Panel B of
Fig~\ref{fig:labMT-tables}, are (labMT, WK): sue (4.30, 2.18), boogie
(5.86, 3.80), exclusive (6.48, 4.50), wake (4.72, 6.57), federal
(4.94, 3.06), stroke (2.58, 4.19), gay (4.44, 6.11), patient (5.04,
6.71), user (5.48, 3.67), and blow (4.48,
6.10).
Like labMT, the WK dictionary used a Mechanical Turk online survey to
gather word ratings.
We speculate that the variation is due to differences in the number
of scores required for each word in the surveys, with 14--18 in WK and 50 in labMT.
For an in depth comparison of these sentiment dictionaries, see reference~\citep{dodds2015reply}.

To compare the word scores in a binary sentiment dictionaries (those with $\pm 1$ or $\pm 1,0$) to the word scores in a sentiment dictionary with a 1--9 range, we examine the distribution of the continuous scores for each binary score.
Looking at the labMT-MPQA comparison in Panel D of
Fig~\ref{fig:main}, we see that more of the matches are between words
without stems (blue) than those with stems (orange), and that each score in -1, 0, +1 from MPQA corresponds
to a wider range of scores in labMT.
We examine the shared individual words from labMT with high sentiment scores and MPQA with score -1, both the happiest and the least happy in labMT with MPQA score 0, and the least happy when MPQA is 1 (Fig~\ref{fig:labMT-tables}
Panels C-E).
The 10 happiest words in labMT matched by MPQA words with score -1
are: moonlight (7.50), cutest (7.62), finest (7.66), funniest (7.76),
comedy (7.98), laughs (8.18), laughing (8.20), laugh (8.22), laughed
(8.26), laughter (8.50).
This is an immediately troubling list of evidently positive words
 rated as -1 in MPQA.
We observe the top 5 are matched by
the stem ``laugh*'' in MPQA.
The least happy 5 words and happiest 5 words in labMT matched by words in
MPQA with score 0 are: sorrows (2.69), screaming (2.96), couldn't
(3.32), pressures (3.49), couldnt (3.58), and baby (7.28), precious
(7.34), strength (7.40), surprise (7.42), and song (7.58).
We see that these MPQA word scores are departures from the other dictionaries, warranting concern about their scores.
The least happy words in labMT with score +1 in MPQA that are matched by
MPQA are: vulnerable (3.34), court (3.78), sanctions (3.86), defendant
(3.90), conviction (4.10), backwards (4.22), courts (4.24), defendants
(4.26), court's (4.44), and correction (4.44).
These words have sentiments that appear to vary with context.

While it would be simple to adjust these ratings in the MPQA
dictionary going forward, we are naturally led to be
concerned about existing work using MPQA that does not examine words contributing to overall sentiment.
We note again that the use of word shift graphs of some kind would have exposed these problematic
scores immediately.

For the labMT-LIWC comparison in Panel E of Fig~\ref{fig:main} we
examine the same matched word lists as before.
The 10 happiest words in labMT matched by words in LIWC with score -1
are: trick (5.22), shakin (5.29), number (5.30), geek (5.34), tricks
(5.38), defence (5.39), dwell (5.47), doubtless (5.92), numbers
(6.04), shakespeare (6.88).
From Panel F of Fig~\ref{fig:labMT-tables}, the least happy 5 neutral
words and happiest 5 neutral words in LIWC, matched in LabMT from LIWC words (i.e., using the word stems in LIWC to match across LabMT, directionality matters), are:
negative (2.42), lack (3.16), couldn't (3.32), cannot (3.32), never
(3.34), millions (7.26), couple (7.30), million (7.38), billion
(7.56), millionaire (7.62).
The least happy words in labMT with score +1 in LIWC that are matched by
LIWC are: merrill (4.90), richardson (5.02), dynamite (5.04), careful
(5.10), richard (5.26), silly (5.30), gloria (5.36), securities
(5.38), boldface (5.40), treasury's (5.42).
The +1 and -1 words in LIWC match some neutral words in labMT, which
is not alarming.
However, the problems with the ``neutral'' words in the LIWC set are
immediate: these are not emotionally neutral words.
The range of scores in labMT for these 0-score words in LIWC formed
the basis for Garcia \etal's response to~\citep{dodds2015human}, and we
point out here that the authors must not have looked at the words, an
all-too-common problem in studies using sentiment analysis~\citep{garcia2015language,dodds2015reply}.

For the labMT-OL comparison in Panel E of Fig~\ref{fig:main} we again examine the same matched word lists as before (except the neutral word list because OL has no explicit neutral words).
The 10 happiest words in labMT matched by OL's negative list are: myth (5.90), puppet (5.90), skinny (5.92), jam (6.02), challenging (6.10), fiction (6.16), lemon (6.16), tenderness (7.06), joke (7.62), funny (7.92).
The least happy words in labMT with score +1 in OL that are matched by OL are: defeated (2.74), defeat (3.20), envy (3.33), obsession (3.74), tough (3.96), dominated (4.04), unreal (4.57), striking (4.70), sharp (4.84), sensitive (4.86).
Despite nearly twice as many negative words in OL as positive words
(at odds with the frequency-dependent positivity bias of language~\citep{dodds2015human}), after examining the words which are the most differently scored and seeing how quickly the labMT scores move into the neutral range, we can conclude that these dictionaries generally agree with the exception of only a few bad matches.

Direct comparisons between the word scores in sentiment dictionaries, while evidently tedious, have brought to light many problematic word scores.
In addition, this analysis serves as a template for further comparisons of the words across new sentiment dictionaries.
The six sentiment dictionaries under careful examination in the present study are further analyzed in the Supporting Information.
Next, we examine how each sentiment dictionary can aid in understanding the sentiments contained in articles from the New York Times.

\subsection{New {Y}ork {T}imes Word Shift Analysis}
\label{subsec:NYTwordshift}

The New York Times corpus~\citep{nytimescorpus2008new} is split into
24 sections of the newspaper that are roughly contiguous throughout
the data from 1987--2008.
With each sentiment dictionary, we rate each section and then compute word
shift graphs (described below) against the baseline, and produce a
happiness ranked list of the
sections.

To gain understanding of the sentiment expressed by any given text
relative
to another text,
it is necessary to inspect the words which contribute most
significantly
by their emotional strength and the change in frequency of usage.
We do this through the use of word shift graphs, which plot the
contribution of each word $w$ from the sentiment dictionary (denoted $\delta
h_\textnormal{ANEW} (w)$) to the shift in average happiness between
two texts, sorted by the absolute value of the contribution.
We use word shift graphs to both analyze a single text and to
compare two texts, here focusing on comparing text within corpora.
For a derivation of the algorithm used to make word shift graphs while
separating the frequency and sentiment information, we refer the
reader to Equations 2 and 3 in~\citep{dodds2011temporal}.
We consider both the sentiment difference and frequency difference
components of $\delta h_\textnormal{ANEW} (w)$ by writing each term of
Eq. \ref{eq:havg} as in~\citep{dodds2011temporal}:
\begin{equation}
  \delta h_\textnormal{ANEW} (w)
  =
  100 \frac{
    h_{\textnormal{ANEW}} (w) - h_{\textnormal{ANEW}}^{\textnormal{ref}}
  }{
    h_{\textnormal{ANEW}} ^{\textnormal{comp}} -    h_{\textnormal{ANEW}} ^{\textnormal{ref}}}
  \left [
    p(w) ^{\textnormal{comp}} - p (w)^{\textnormal{ref}}
    \right ].
\end{equation}
An in-depth explanation of how to interpret the word shift graph can
also be found at
\url{http://hedonometer.org/instructions.html#wordshifts}.

\begin{figure*}[tbp!]
\includegraphics[width=0.98\textwidth]{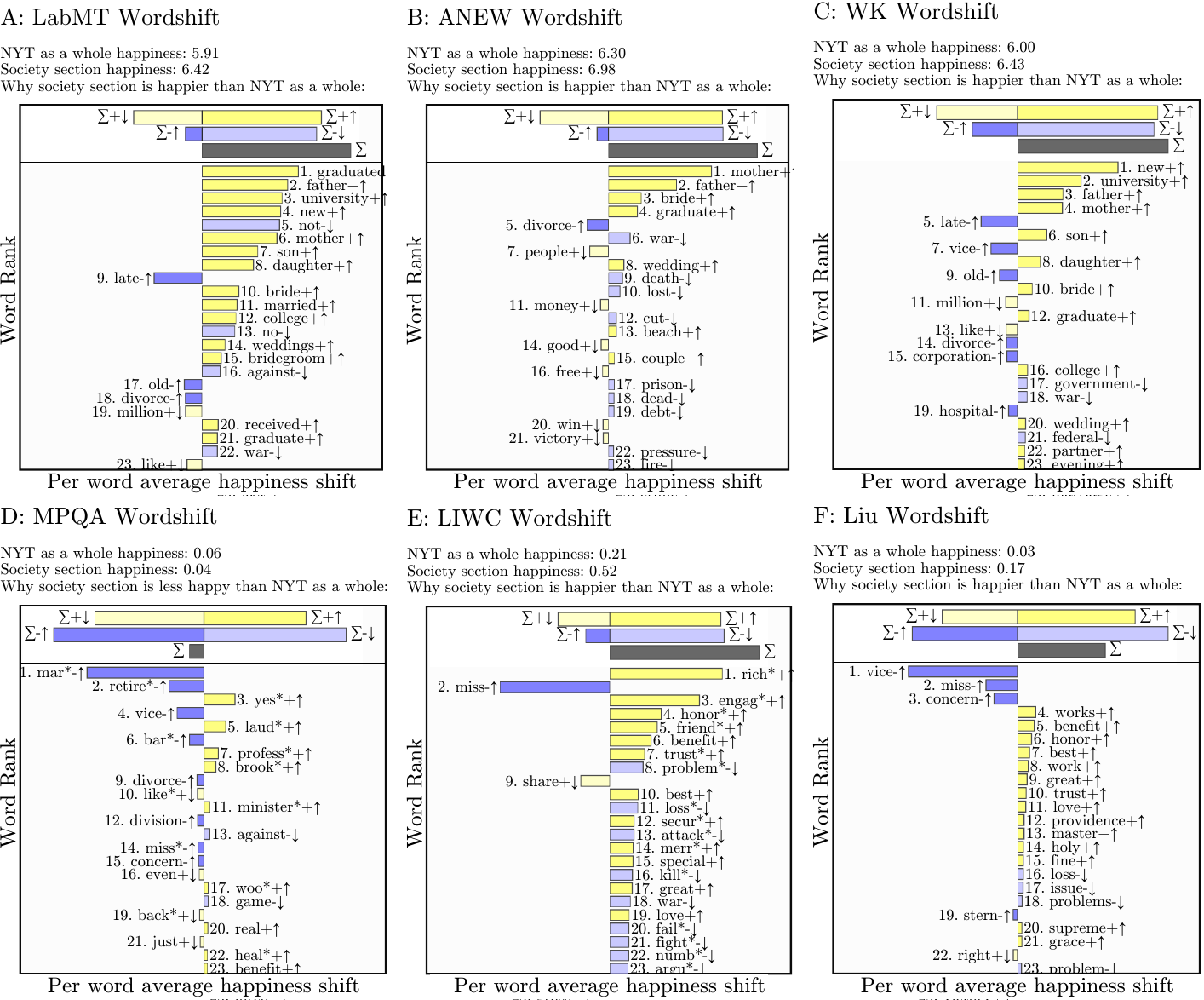}
  \caption{
    New York Times (NYT) ``Society'' section shifted against the entire NYT corpus for each of the six dictionaries listed in tiles A--F.
    We provide a detailed analysis in
    Sec. \ref{subsec:NYTwordshift}.
    Generally, we are able to glean the greatest understanding of the
    sentiment texture associated with this NYT section using the labMT
    dictionary.
    Additionally we note the labMT dictionary has the most coverage
        quantified by word match count (Figure in S3 Appendix), we
    are able to identify and correct problematic words scores in the
    OL dictionary, and we see that the MPQA dictionary disagrees
    entirely with the others because of an overly broad stem match.
  }
  \label{fig:nyt_wordshifts}
\end{figure*}

To both demonstrate the necessity of using word shift graphs in
carrying out sentiment analysis, and to gain understanding about the
ranking of New York Times sections by each sentiment dictionary, we look at word
shift graphs for the ``Society'' section of the newspaper from each
sentiment dictionary in Fig~\ref{fig:nyt_wordshifts}, with the reference text
being the whole of the New York Times.
The ``Society'' section happiness ranks 1, 1, 1, 18, 1, and 11 within the happiness of each of the 24
sections in the dictionaries labMT, ANEW, WK, MPQA, LIWC, and OL,
respectively.
These graphs show only the very top of the distributions which range
in length from 1030 (ANEW) to 13915 words (WK).

First, using the labMT dictionary, we see that the words
``graduated'', ``father'', and ``university'' top the list,
which is dominated by positive words that occur more frequently ($+\uparrow$).
These more frequent positive words paint a clear picture of family life
(relationships, weddings, and divorces), as well as university
accomplishment (graduations and college).
In general, we are able to observe with only these words that the
``Society'' section is where we find the details of these events.

From the ANEW dictionary, we see that a few positive words have increased frequency,
lead by ``mother'', ``father'', and ``bride''.
Looking at this shift in isolation, we see only these words with three
more (``graduate'', ``wedding'', and ``couple'') that would lead us to
suspect these topics are present in the ``Society'' section.

The WK dictionary, with the most individual word scores of any
sentiment dictionary tested, agrees with labMT and ANEW that the ``Society''
section is the happiest section, with somewhat similar set of words at the top:
``new'', ``university'', and ``father''.
Low coverage of the New York Times corpus (see
Fig~\ref{fig:coverage_nyt}) resulted in less specific words describing the ``Society'' section, with more
words that go down in frequency in the shift.
With the words ``bride'' and ``wedding'' up, as well as
``university'', ``graduate'', and ``college'', it is evident that the
``Society'' section covers both graduations and weddings,
in consensus with the other sentiment dictionaries.

The MPQA dictionary ranks the ``Society'' section 18th of the 24 NYT
sections, a departure from the other rankings, with the words
``mar*'', ``retire*'', and ``yes*'' the top three
contributing words (where ``*'' denotes a wildcard ``stem'' match).
Negative words increasing in frequency ($-\uparrow$) are the most common type near
the top, and of these, the words with the biggest contributions are
being scored incorrectly in this context (specifically words
``mar*'', ``retire*'', ``bar*'', ``division'', and ``miss*'').
Looking more in depth at the problems created by the first out of context word match,
we find 1211 unique words match ``mar*''.
The five most frequent, with counts in parenthesis, are married (36750), marriage (5977), marketing (5382), mary (4403),
and mark (2624).
The score for these words in MPQA is -1, in stark contrast to the scores in other sentiment dictionaries (e.g., the labMT scores are 6.76, 6.7, 5.2, 5.88, and 5.48).
These problems plague the MPQA dictionary for scoring the New York
Times corpus, and without using word shift graphs would have gone completely
unseen.
In an attempt to fix contextual issues by fixing corpus-specific
words, we remove ``mar*,retire*,vice,bar*,miss*'' and find that the MPQA dictionary ranks the Society section of the NYT at 15th of the 24 sections

\begin{figure*}[tbp!]
\includegraphics[width=0.96\textwidth]{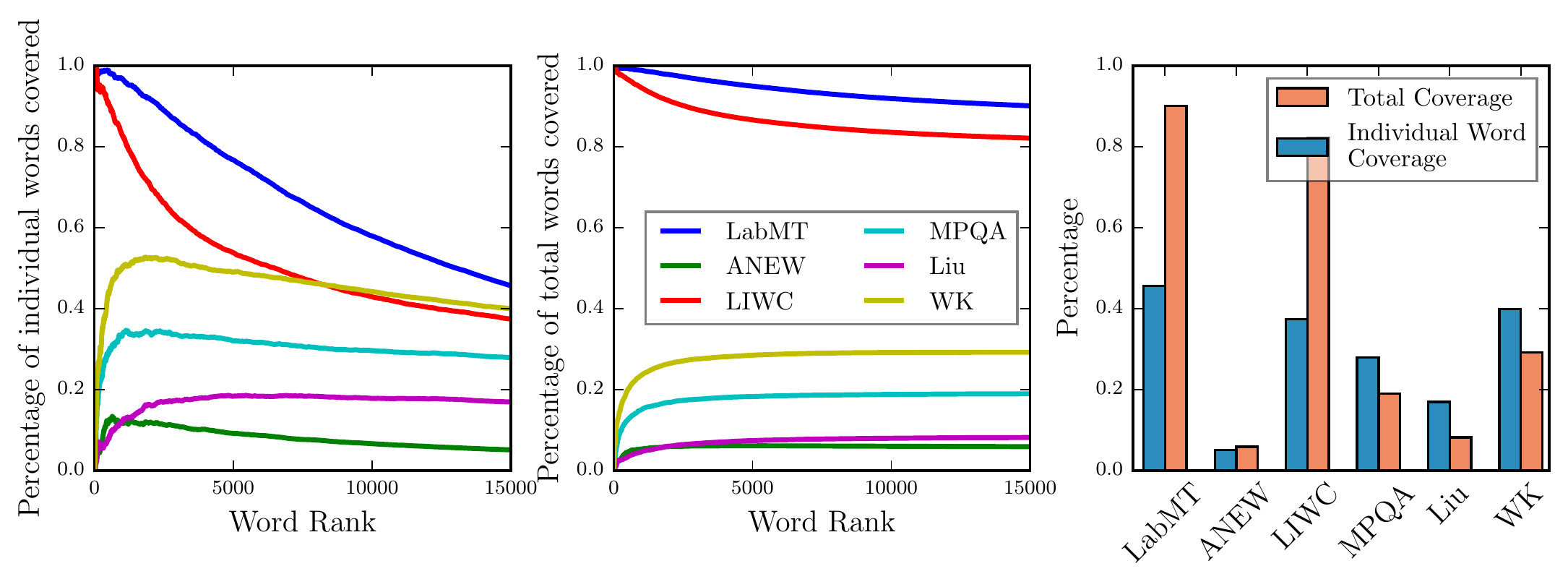}
    \caption{
      Coverage of the words in the movie reviews by each of the dictionaries.
      We observe that the labMT dictionary has the highest coverage of
      words in the movie reviews corpus both across word rank and
      cumulatively.
      The LIWC dictionary has initially high coverage since it
      contains some high-frequency function words, but quickly drops
      off across rank.
      The WK dictionary coverage increases across word rank and
      cumulatively, indicating that it contains a large number of less
      common words in the movie review corpus.
      The OL, ANEW, and MPQA have a cumulative coverage of less than
      20\% of the lexicon.
    }
  \label{fig:coverage_movies}
\end{figure*}

The second binary sentiment dictionary, LIWC, agrees well with the first three
dictionaries and ranks the ``Society'' section at the top with the
words ``rich*'', ``miss'', and ``engage*'' at the top of the list.
We immediately notice that the word ``miss'' is being used
frequently in the ``Society'' section in a different sense than was
coded for in the LIWC dictionary: it is used in the corpus to mean ``the title prefixed to the
name of an unmarried woman'', but is scored as negative in LIWC (with the likely intended
meaning ``to fail to reach an target or to acknowledge loss'').
We would remove this word from LIWC for further analysis of this
corpus (we would also remove the word ``trust'' here).
The words matched by ``miss*'' aside, LIWC finds some positive words
going up ($+\uparrow$), with ``engage*'' hinting at weddings.
Without words that capture the specific behavior happening
in the ``Society'' section, we are unable to see anything about
college, graduations, or marriages, and there is much less to be
gained about the text from the words in LIWC than some of the other
dictionaries we have seen.
Nevertheless, LIWC finds consensus with the
``Society'' section ranked the top section, due in large part to a lack
of negative words ``war'' (rank 18) and ``fight*'' (rank 22).

The OL sentiment dictionary departs from the consensus and ranks the
``Society'' section at 11th out of the 24 sections.
The top three words, ``vice'', ``miss'', and ``concern'',
contribute largely with respect to the rest of distribution, of which
two are clearly being used in the wrong sense.
For a more reasonable analysis we would remove both ``vice'' and
``miss'' from the OL dictionary to score this text.
For a more reasonable analysis we remove both ``vice'' and ``miss'' from the OL dictionary to score this text, and in doing so the happiness goes from 0.168 to 0.297, making the ``Society'' section the second happiest of the 24 sections.
Focusing on the words, we see that the OL dictionary finds many
positive words increasing in frequency ($+\uparrow$) that are mostly generic.
In the word shift graph we do not find the wedding or university events as
in sentiment dictionaries with more coverage, but rather a variety of positive
language surrounding these events, for example 4. ``works'',
``benefit'' (5), ``honor'' (6), ``best'' (7), ``great'' (9),
``trust'' (10), ``love'' (11), etc. While this does not provide insight into the topics, the OL sentiment dictionary
with fixes from the word shift graph analysis does provide details on the emotive words that make the ``Society'' section one of the happiest sections.

In conclusion, we find that 4 of the 6 dictionaries score the
``Society'' section at number 1, and in these cases we use the word
shift graph to uncover the nuances of the language used.
We find, unsurprisingly, that the most matches are found by the labMT
dictionary, which is in part built from the NYT corpus
(see S3 Appendix for coverage plots).
Without as much corpus-specific coverage, we note that while the
nuances of the text remain hidden, the LIWC and OL dictionaries still
highlight the positive language in this section.
Of the two that did not score the ``Society'' section at the top, we are able to assess and repair the MPQA and the OL dictionaries by removing the words ``mar*,retire*,vice*,bar*,miss*'' and ``vice,miss'', respectively.
By identifying words used in the wrong sense/context using the word shift graph, we directly improve the sentiment score for the New York Times corpus from both MPQA and OL dictionaries closer to consensus.
While the OL dictionary, with two corrections, agrees with the other dictionaries, the MPQA dictionary with five corrections still ranks the Society section of the NYT as the 15th happiest of the 24 sections.

In the first Figure in S4 Appendix we show scatterplots for each comparison, and compute the Reduced Major Axes (RMA) regression fit~\citep{rayner1985linear}.
In the second Figure in S4 Appendix we show the sorted bar chart from each sentiment dictionary.

\subsection{Movie {R}eviews Classification and Word Shift Graph Analysis}
\label{subsec:moviereviews}

For the movie reviews, we first provide insight into the language differences and secondly perform binary classification of positive and negative reviews.
The entire dataset consists of 1000 positive and 1000 negative
reviews, as rated with 4 or 5 stars and 1 or 2 stars, respectively.
We show how well each sentiment dictionary covers the review database
in Fig~\ref{fig:coverage_movies}.
The average review length is 650 words, and we plot the distribution
of review lengths in S5 Appendix. We average the sentiment of words in each review individually, using
each sentiment dictionary.
We also combine random samples of $N$ positive or $N$ negative reviews
for $N$ varying from 2 to 900 on a logarithmic scale,
without replacement, and rate the combined text.
With an increase in the size of the text, we expect that the
dictionaries will be better able to distinguish positive from
negative.
The simple statistic we use to describe this ability is the percentage of
distributions that overlap the average.

\begin{figure*}[tbp!]
\includegraphics[width=0.98\textwidth]{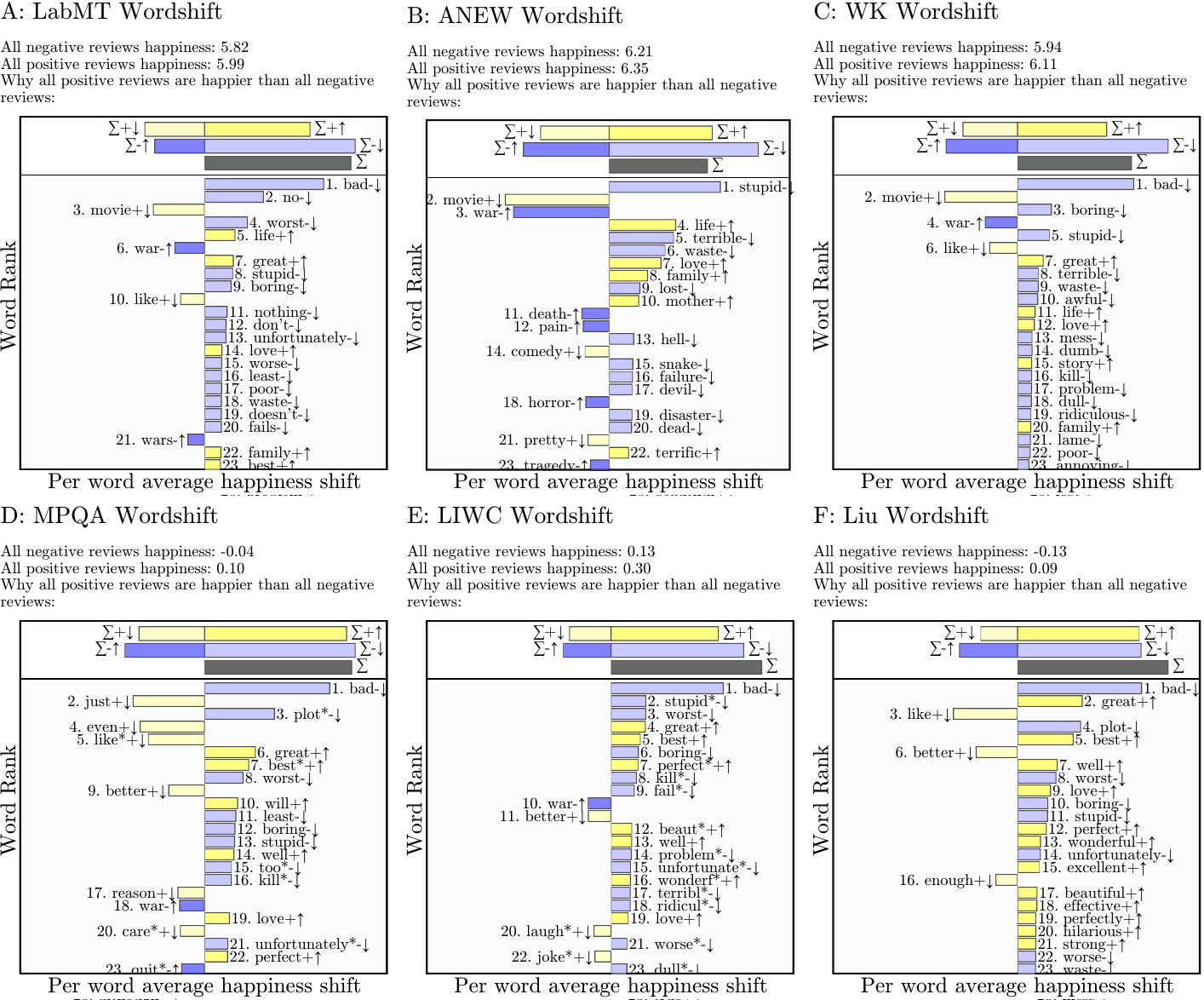}
  \caption{
    Word shift graphs for the movie review corpus.
    By analyzing the words that contribute most significantly to the
    sentiment score produced by each sentiment dictionary we are able to
    identify words that are problematic for each sentiment dictionary at the
    word-level, and generate an understanding of the emotional texture
    of the movie review corpus.
    Again we find that coverage of the lexicon is essential to produce
    meaningful word shift graphs, with the labMT dictionary providing the
    most coverage of this corpus and producing the most detailed word
    shift graphs.
    All words on the left hand side of these word shift graphs are words
    that individually made the positive reviews score more negatively than the
    negative reviews, and the removal of these words would improve the
    accuracy of the ratings given by each sentiment dictionary.
    In particular, across each sentiment dictionary the word shift graphs show that
    domain-specific words such as ``war'' and ``movie'' are used more
    frequently in the positive reviews and are not useful in
    determining the polarity of a single review.
  }
  \label{fig:moviereviews-shifts}
\end{figure*}

To analyze which words are being used by each sentiment dictionary, we compute
word shift graphs of the entire positive corpus versus the entire
negative corpus in Fig~\ref{fig:moviereviews-shifts}.
Across the board, we see that a decrease in negative words is the most
important word type for each sentiment dictionary, with the word ``bad'' being
the top word for every sentiment dictionary in which it is scored (ANEW does not
have it).
Other observations that we can make from the word shift graphs include a few
words that are potentially being used out of context: ``movie'',
``comedy'', ``plot'', ``horror'', ``war'', ``just''.

\begin{figure*}[tbp!]
  \centering
\includegraphics[width=0.98\textwidth]{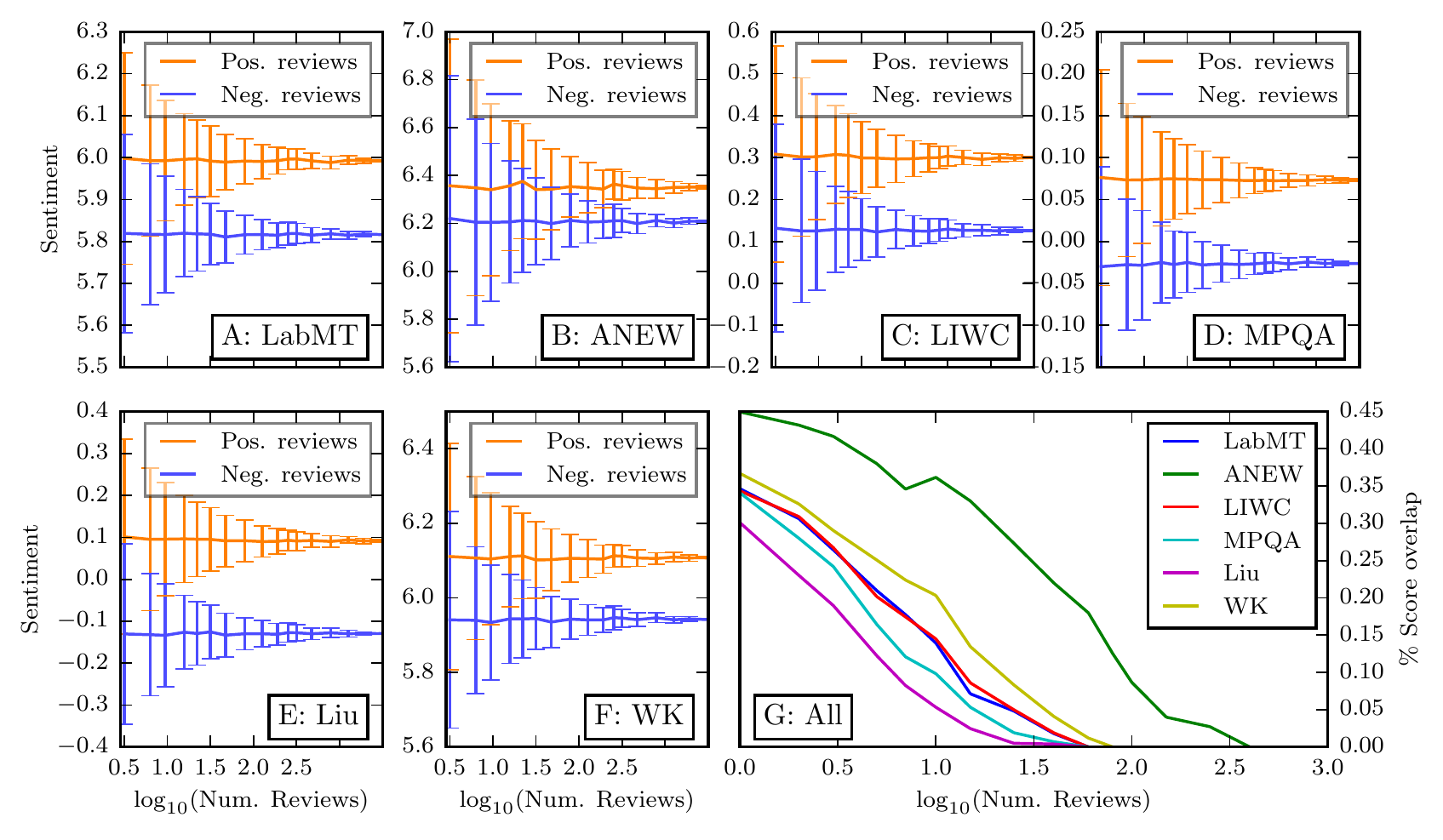}
  \caption{
    The score assigned to increasing numbers of reviews drawn from the tagged positive and negative sets.
    For each sentiment dictionary we show mean sentiment and 1 standard
    deviation over 100 samples for each distribution of reviews in
    Panels A--F.
    For comparison we compute the fraction of the distributions that
    overlap in Panel G.
    At the single review level for each sentiment dictionary this simple
    performance statistic (fraction of distribution overlap) ranks the
    OL dictionary in first place, the MPQA, LIWC, and labMT
    dictionaries in a second place tie, WK in fifth, and ANEW far
    behind.
    All dictionaries require on the order of 1000 words to achieve
    95\% classification accuracy.
  }
  \label{fig:moviereviewtest}
\end{figure*}

In the lower right panel of Fig~\ref{fig:moviereviewtest}, the
percentage overlap of positive and negative review distributions
presents us with a simple summary of sentiment dictionary performance on this
tagged corpus.
The ANEW dictionary stands out as being considerably less capable of
distinguishing positive from negative.
In order, we then see WK is slightly better overall, labMT and LIWC
perform similarly better than WK overall, and then MPQA and OL are
each a degree better again, across the review lengths (see below for
hard numbers at 1 review length).
Two Figures in S5 Appendix
show the distributions for 1 review and for 15 combined reviews.

Classifying single reviews as positive or negative, the F1-scores
are: labMT .63, ANEW .36, LIWC .53, MPQA .66, OL .71, and WK
.34 (see Table \ref{tbl:MR-1}).
We roughly confirm the rule-of-thumb that 10,000 words are enough to
score with a sentiment dictionary confidently, with all dictionaries except MPQA
and ANEW achieving 90\% accuracy with this many words.
We sample the number of reviews evenly in log space, generating sets
of reviews with average word counts of 4550, 6500, 9750, 16250, and
26000 words.
Specifically, the number of reviews necessary to achieve 90\% accuracy
is 15 reviews (9750 words) for labMT, 100 reviews (65000 words) for
ANEW, 10 reviews (6500 words) for LIWC, 10 reviews (6500 words) for
MPQA, 7 reviews (4550 words) for OL, and 25 reviews (16250 words) for
WK.

While we are analyzing the movie review classification, which has ground truth labels, we will take a moment to further support our claims for the inaccuracy of these methods at the sentence level.
The OL dictionary, with the highest performance classifying individual movie reviews of the 6 dictionaries tested in detail, performs worse than guessing at classifying individual sentences in movie reviews.
Specifically, 76.9/74.2\% of sentences in the positive/negative reviews sets have words in the OL dictionary, and of these OL achieves an F1-score of 0.44.
The results for each sentiment dictionary are included in Table \ref{tbl:MR-2}, with an average (median) F1 score of 0.42 (0.45) across all dictionaries.
While these results do cast doubt on the ability to classify positive and negative reviews from single sentences using dictionary based methods, we note that we need not expect the sentiment of individual sentences to be strongly correlated with the overall review polarity.

\subsection{Google {B}ooks Time Series and Word Shift Analysis}
\label{subsec:googlebooks}

We use the Google books 2012 dataset with all English
books~\citep{lin2012syntactic}, from which we remove part of speech
tagging and split into years.
From this, we make time series by year, and word shift graphs of decades
versus the baseline.
In addition, to assess the similarity of each time series, we produce
correlations between each of the time series.

Despite claims from research based on the Google Books corpus
\citep{michel2011quantitative}, we keep in mind that there are several
deep problems with this beguiling data set~\citep{pechenick2015characterizing}.
Leaving aside these issues, the Google Books corpus nevertheless
provides a substantive test of our six dictionaries.

\begin{figure*}[tbp!]
\includegraphics[width=0.98\textwidth]{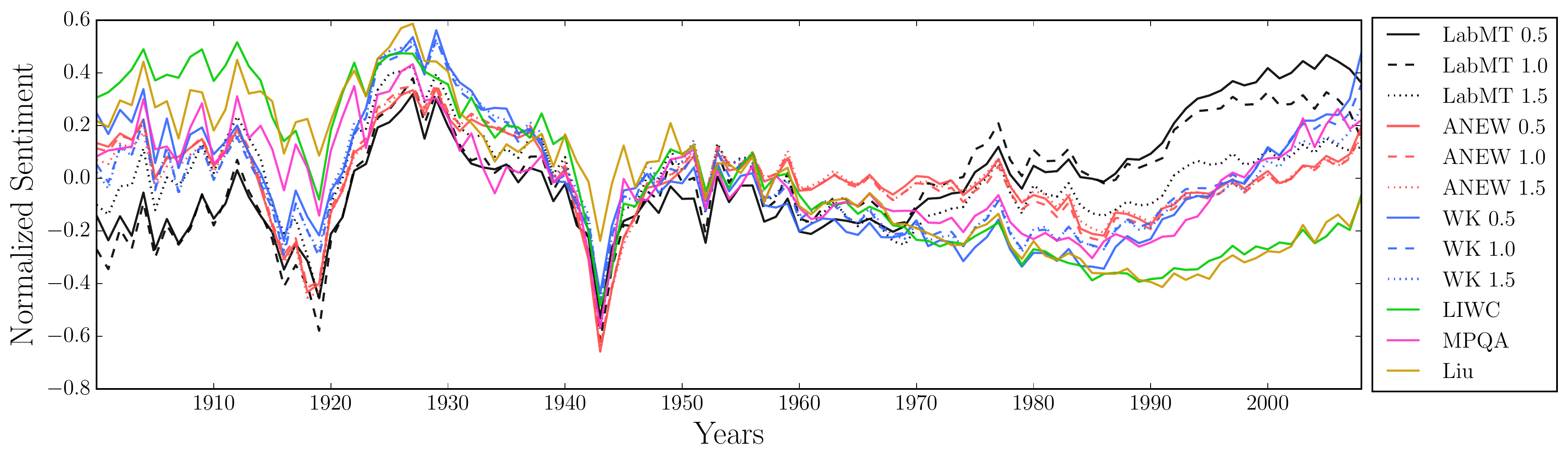}
  \caption{
    Google Books sentiment time series from each sentiment dictionary, with stop values of 0.5, 1.0, and 1.5 from the dictionaries with word scores in the 1--9 range.
    To normalize the sentiment score, we subtract the mean and divide
    by the absolute range.
    We observe that each time series has increased variance, with a few pronounced
    negative time periods, and trending positive towards the end of
    the corpus.
    The score of labMT varies substantially with different stop
    values, although remaining highly correlated, and finds absolute
    lows near the World Wars.
    The LIWC and OL dictionaries trend down towards 1990, dipping as
    low as the war periods.
  }
  \label{fig:gbooks_timeseries}
\end{figure*}

In Fig~\ref{fig:gbooks_timeseries}, we plot the sentiment time series
for Google Books.  Three immediate trends stand out: a dip near the
Great Depression, a dip near World War II, and a general upswing in
the 1990's and 2000's.  From these general trends, a few dictionaries
waver: OL does not dip as much for WW2, OL and LIWC stay lower in
the 90's and 2000's, and labMT with $\Delta_h = 0.5,1.0$ go downward
near the end of the 2000's.  We take a closer look into the 1940's to
see what each sentiment dictionary is picking up in Google Books around World
War 2 in Figure in S6 Appendix.

In each panel of the word shift Figure in S6 Appendix, we see that the top word making the 1940's less positive than the the rest of Google Books is ``war'', which is the top contributor for every sentiment dictionary except OL.
Rounding out the top three contributing words are ``no'' and ``great'', and we infer that the word ``great'' is being seen from mention of ``The Great Depression'' or ``The Great War'', and is possibly being used out of context.
All dictionaries but ANEW have ``great'' in the top 3 words, and each sentiment dictionary could be made more accurate if we remove this word.

In Panel A of the 1940's word shift Figure in S6 Appendix, beyond the top words, increasing words are mostly negative and war-related: ``against'', ``enemy'', ``operation'', which we could expect from this time period.

In Panel B, the ANEW dictionary scores the 1940's of Google Books
lower than the baseline as well, finding ``war'', ``cancer'', and
``cell'' to be the most important three words.
With only 1030 words, there is not enough coverage to see anything
beyond the top word ``war,'' and the shift is dominated by words that
go down in frequency.

In Panel C, the WK dictionary finds the the 1940's with slightly less happy than the baseline, with the top three words being ``war'', ``great'', and ``old''.
We see many of the same war-related words as in labMT, and in addition some positive words like ``good'' and ``be'' are up in frequency.
The word ``first'' could be an artifact of first aid, a claim that could be substantiated with further analysis of the Google Books corpus at the 2-gram level beyond the scope of this manuscript.

In Panel D, the MPQA dictionary rates the 1940's slightly less happy than the baseline, with the top three words being ``war'', ``great'', and ``differ*''.
Beyond the top word ``war'', the score is dominated by words decreasing in frequency, with only a few words up in frequency.
Without specific words increasing in frequency as contextual guides, it is difficult to obtain a good glance at the nature of the text.
For this reason, having a higher coverage of the words in the corpus enables understanding.

In Panel E, the LIWC dictionary rates the 1940's nearly the same as the baseline, with the top three words being ``war'', ``great'', and ``argu*''.
When the scores are nearly the same, although the 1940's are slightly higher happiness here, the word shift is a view into how the words of the reference and comparison text vary.
In addition to a few war related words being up and bringing the score down (``fight'', ``enemy'', ``attack''), we see some positive words up that could also be war related: ``certain'', ``interest'', and ``definite''.
Although LIWC does not manage to find World War II as a low point of the 20th century, the words that contribute to LIWCs score for the 1940's compared to all years are useful in understanding the corpus.

In Panel F, the OL dictionary rates the 1940's as happier than the baseline, with the top three words being ``great'', ``support'', and ``like''.
With 7 positive words up, and 1 negative word up, we see how the OL dictionary misses the war without the word ``war'' itself and with only ``enemy'' contributing from the words surrounding the conflict.
The nature of the positive words that are up is unclear, and could justify a more detailed analysis of why the OL dictionary fails here.

\subsection{Twitter Time Series Analysis}
\label{subsec:twittertimeseries}

For Twitter data, we use the Gardenhose feed, a random 10\% of the full stream. We store data on the Vermont Advanced Computing Core (VACC), and process the text first into hash tables (with approximately 8 million unique English words each day) and then into word vectors for each 15 minutes, for each sentiment dictionary tested.
From this, we build sentiment time series for time resolutions of 15 minutes, 1 hour, 3 hours, 12 hours, and 1 day.
In addition to the raw time series, we compute correlations between each time series to assess the similarity of the ratings between dictionaries.

\begin{figure*}[tbp!]
  \centering
\includegraphics[width=0.98\textwidth]{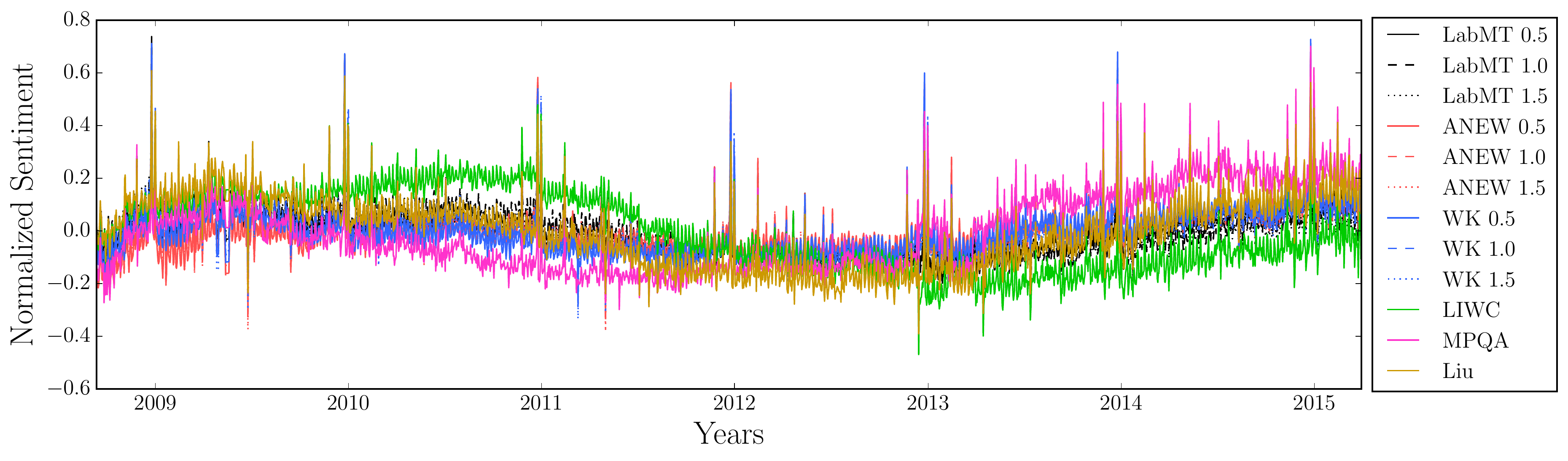}
  \caption{
    Normalized sentiment time series on Twitter using $\Delta _h$ of 1.0 for all dictionaries.
    To normalize the sentiment score, we subtract the mean and divide by the absolute range.
    The resolution is 1 day, and draws on the 10\% gardenhose sample of public Tweets provided by Twitter.
    All of the dictionaries exhibit wide variation for very early Tweets, and from 2012 onward generally track together strongly with the exception of MPQA and LIWC.
    The LIWC and MPQA dictionaries show opposite trends: a rise until 2012 with a decline after 2012 from LIWC, and a decline before 2012 with a rise afterwards from MPQA.
            To analyze the trends we look at the words driving the movement across years using word shift Figures in S7 Appendix.
    }
  \label{fig:twitter_timeseries_4}
\end{figure*}

In Fig~\ref{fig:twitter_timeseries_4}, we present a daily sentiment time series of Twitter processed using each of the dictionaries being tested.
With the exception of LIWC and MPQA we observe that the dictionaries generally track well together across the entire range.
A strong weekly cycle is present in all, although muted for ANEW.

\begin{figure}[tbp!]
  \centering
\includegraphics[width=0.48\textwidth]{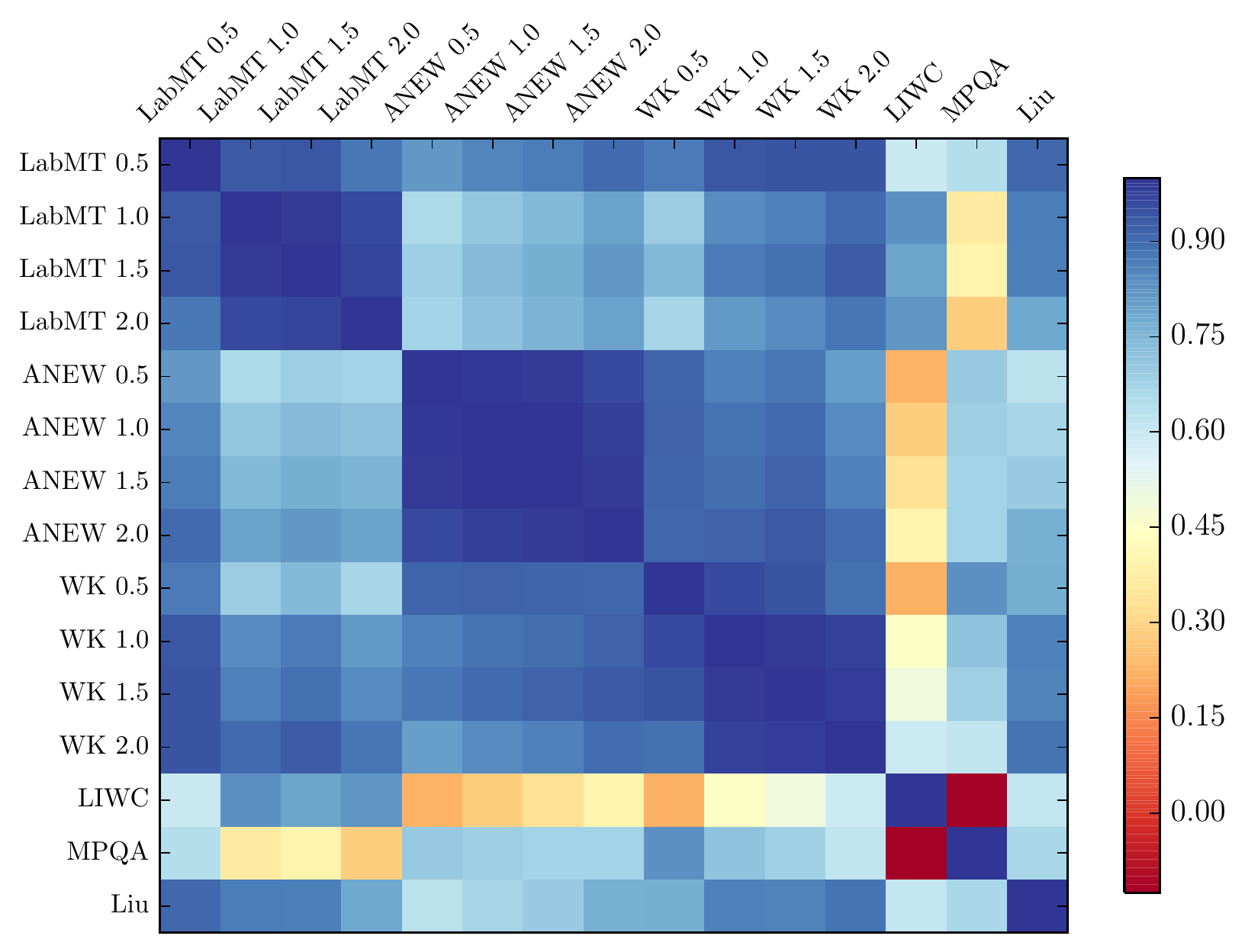}
  \caption{
    Pearson's $r$ correlation between daily resolution Twitter sentiment time series for each sentiment dictionary.
    We see that there is strong agreement within dictionaries, with
    the biggest differences coming from the stop value of $\Delta h =
    0.5$.
    The labMT and OL dictionaries do not strongly disagree with any
    of the others, while LIWC is the least correlated overall with
    other dictionaries.
    labMT and OL correlate strongly with each other, and disagree
    most with the ANEW, LIWC, and MPQA dictionaries.
    The two least correlated dictionaries are the LIWC and MPQA
    dictionaries.
    Again, since there is no publicly accessible ground truth for Twitter sentiment, we
    compare dictionaries against the others, and look at the words.
    With these criteria, we find the labMT dictionary to be the most
    useful.
  }
  \label{fig:twitter_correlation_4}
\end{figure}

We plot the Pearson's correlation between all time series in Fig~\ref{fig:twitter_correlation_4}, and confirm some of the general observations that we can make from the time series.
Namely, the LIWC and MPQA time series disagree the most from the others, and even more so with each other.
Generally, we see strong agreement within dictionaries with varying stop values $\Delta h$.

The time series from each sentiment dictionary exhibits increased variance at the start of the time frame,
when Twitter volume is low in 2008 and into 2009.
As more people join Twitter and the Tweet volume increases through
2010, we see that LIWC rates the text as happier, while the rest start a slow
decline in rating that is led by MPQA in the negative direction.
In 2010, the LIWC dictionary is more positive than the rest with words
like ``haha'', ``lol'' and ``hey'' being used more frequently and
swearing being less frequent than all years of Twitter put
together.
The other dictionaries with more coverage find a decrease in positive
words to balance this increase, with the exception of MPQA which finds
many negative words going up in frequency (see 2010 word shift Figure in Appendix S7).
All of the dictionaries agree most strongly in 2012, all finding a lot
of negative language and swearing that brings scores down
(see 2012 word shift Figure in Appendix S7).
From the bottom at 2012, LIWC continues to go downward while the
others trend back up.
The signal from MPQA jumps to the most positive, and LIWC does start
trending back up eventually.
We analyze the words in 2014 with a word shift against all 7 years of
Tweets for each sentiment dictionary in each panel in the
2014 word shift Figure in Appendix S7:
A. labMT scores 2014 as less happy with
more negative language. B. ANEW finds it happier with a few
positive words up. C. WK finds it happier with more negative words
(like labMT). D. MPQA finds it more positive with less negative
words. E. LIWC finds it less positive with more negative and less
positive words. F. OL finds it to be of the same sentiment as the
background with a balance in positive and negative word usage.
From these word shift graphs, we can analyze which words cause MPQA and LIWC
to disagree with the other dictionaries: the disagreement of MPQA is
again marred by broad stem matches, and the disagreement of LIWC is
due to a lack of coverage.

\subsection{Brief Comparison to Machine Learning Methods}
\label{subsec:NB-section}

We implement a Naive Bayes (NB) classifier (sometimes harshly called idiot Bayes~\citep{hand2001idiot}) on the tagged movie review dataset, using 10\% of the reviews for training and then testing performance on the rest.
Following standard practice, we remove the top 30 ranked words (``stop words'') from the 5000 most frequent words, and use the remaining 4970 words in our classifier for maximum performance (we observe a 0.5\% improvement).
Our implementation is analogous to those found in common Python natural language processing packages (see ``NLTK'' or ``TextBlob'' in \citep{bird2006nltk}).

As we should expect, at the level of single review, NB outperforms the dictionary-based methods with a classification accuracy of 75.7\% averaged over 100 trials.
As the number of reviews is increased, the overlap from NB diminishes, and using our simple ``fraction overlapping'' metric, the error drops to 0 with more than 200 reviews.
Interestingly, NB starts to do worse with more reviews, and with more than 500 of the 1000 reviews concatenated, it rates both the positive and negative reviews as positive (Figure in S8 Appendix).

The rating curves do not touch, and neither do the standard deviation error bars (indicating that the result is not statistically significant), but they both go very slightly above 0 (again, see Figure in S8 Appendix).
Overall, with Naive Bayes we are able to classify a higher percentage of individual reviews correctly, but with more variance.

In the two Tables in S8 Appendix we compute the words which the NB classifier uses to classify all of the positive reviews as positive, and all of the negative reviews as positive.
The Natural Language Toolkit (NLTK~\citep{bird2006nltk}) implements a method to obtain the ``most informative'' words, by taking the ratio of the likelihood of words between all available classes, and looking for the largest ratio:
\begin{equation}
  \max _{\textnormal{all words }w} \frac{P ( w | c_i )}{P ( w | c_j )}
\end{equation}
for all combinations of classes $c_i,c_j$.
This is possible because of the ``naive'' assumption that feature (word) likelihoods are independent, resulting in a classification metric that is linear for each feature.
In S8 Appendix, we provide the derivation of this linearity structure.

We find that the trained NB classifier relies heavily on words that
are very specific to the training set including the names of actors of
the movies themselves, making them useful as classifiers but not in
understanding the nature of the text.
We report the top 10 words for both positive and negative classes
using both the ratio and difference methods in Table in S8 Appendix.
To classify a document using NB, we use the frequency of each word in the
document in conjunction with the probability that that word
occurred in each labeled class $c_i$.
While steps can be taken to avoid this type of over-fitting, it is an ever-present danger that remains hidden without word shift graphs or similar.

We next take the movie-review-trained NB classifier and use it to
classify the New York Times sections, both by ranking them and by looking at
the words (the above ratio and difference weighted by the occurrence of
the words).
We ranked the sections 5 different times, and among those
find the ``Television'' section both by far the happiest,
and by far the least happy in independent tests.
We show these rankings and report the top 10 words used to score the ``Society'' section in Table~\ref{tbl:NB-2}.

We thus see that the NB classifier, a linear learning method, may perform poorly when assessing
sentiment outside of the corpus on which it is trained.
In general, performance will vary depending on
the statistical dissimilarity of the training and novel corpora.
Added to this is the inscrutability of black box methods:
while susceptible to the aforementioned difficulty, nonlinear learning methods (unlike NB) also render detailed examination of how
individual words contribute to a text's score more difficult.

\section{Conclusion}
\label{sec:conclusion}

We have shown that measuring sentiment in various corpora presents
unique challenges, and that sentiment dictionary performance is situation
dependent.
Across the board, the ANEW dictionary performs poorly, and the
continued use of this sentiment dictionary with clearly better alternatives is a
questionable choice.
We have seen that the MPQA dictionary does not agree with the other
five dictionaries on the NYT corpus and Twitter corpus due to a
variety of context, word sense, phrase, and stem matching issues, and we would not
recommend using this sentiment dictionary.
While the OL achieves the highest binary classification accuracy, in comparison to labMT, the WK, LIWC, and OL dictionaries fail to
provide much detail in corpora where their coverage is lower,
including all four corpora tested, the main goal of our analysis.
Sufficient coverage is essential for producing meaningful word shift graphs and
thereby enabling deeper understanding.

In each case, to analyze the output of the dictionary method,
we rely on the use of word shift graphs.
With this tool, we can produce a finer grained analysis of the lexical
content, and we can also detect words that are used out of context and can mask them
directly.
It should be clear that using any of the dictionary-based sentiment detecting method
without looking at how individual words contribute is indefensible,
and analyses that do not use word shift graphs or similar tools cannot be trusted.
The poor word shift performance of binary dictionaries in particular
gravely limits their ability to reveal underlying stories.

In sum, we believe that dictionary-based methods will continue
to play a powerful role---they are fast and well suited for web-scale
data sets---and that the best instruments will be based on dictionaries
with excellent coverage and continuum scores.
To this end, we urge that all dictionaries should be regularly updated
to capture changing lexicons, word usage, and demographics.
Looking further ahead, a move from scoring words to scoring
both phrases and words with senses should realize
considerable improvement for many languages of interest.
With phrase dictionaries, the resulting phrase shift graphs
will allow for a  more nuanced and detailed analysis of a corpus's sentiment
score~\citep{alajajian2015lexicocalorimeter},
ultimately affording clearer stories
for sentiment dynamics.

\singlespacing
\bibliographystyle{chicago}
\bibliography{everything}

\doublespacing

\chapter{
The emotional arcs of stories are dominated by six basic shapes
}
\chaptermark{Emotional Arcs}
\label{chap:emotional-arcs}

\newcommand{\nbooks}{1,327}
\begin{quote}
Advances in computing power, natural language processing, and digitization of text now make it possible to study a culture's evolution through its texts using a ``big data'' lens.
Our ability to communicate relies in part upon a shared emotional experience, with stories often following distinct emotional trajectories and forming patterns that are meaningful to us.
Here, by classifying the emotional arcs for a filtered subset of \nbooks~stories from Project Gutenberg's fiction collection, we find a set of six core emotional arcs which form the essential building blocks of complex emotional trajectories.
We strengthen our findings by separately applying Matrix decomposition, supervised learning, and unsupervised learning.
For each of these six core emotional arcs, we examine the closest characteristic stories in publication today and find that particular emotional arcs enjoy greater success, as measured by downloads.
\end{quote}

\section{Introduction}

The power of stories to transfer information and define our own existence has been shown time and again \citep{pratchett2003science,campbell1949hero,gottschall2013storytelling,cave2013stories}.
We as people are fundamentally driven to find and tell stories, likened to \textit{Pan Narrans} or \textit{Homo Narrativus} \citep{dodds2013homo}.
Stories are encoded in art, language, and even in the mathematics of physics:
We use equations to represent both simple and complicated functions that describe our observations of the real world.
In science, we formalize the ideas that best fit our experience with principles such as Occam's Razor:
The simplest story is the one we should trust.
We tend to prefer stories that fit into the molds which are familiar,
and reject narratives that do not align with our experience \citep{nickerson1998confirmation}.

We seek here to better understand stories that are captured and shared in written form,
a medium that since inception has radically changed how information flows \citep{gleick2011information}.
Without evolved cues from tone, facial expression, or body language,
written stories are forced to capture the entire transfer of experience on a page.
An often integral part of a written story is the emotional experience that is evoked in the reader.
Here, we use a simple, robust sentiment analysis tool to extract the reader-perceived emotional content of written stories as they unfold on the page.

We objectively test aspects of folkloristic theory \citep{propp1968morphology,macdonal1982storytellers},
specifically the commonality of core stories within societal boundaries \citep{cave2013stories,silva2016comparative}.
A major component of folkloristics is the study of society and culture through literary analysis.
This is sometimes referred to as \textit{narratology},
which at its core is ``a series of events, real or fictional, presented to the reader or the listener'' \citep{min2016narrative}.
In our present treatment,
we consider the plot as the ``backbone'' of events that occur in a chronological sequence (more detail on previous theories of plot, and the framing we present next and adopt, are in Appendix~\ref{sec:plots}).
While the plot captures the mechanics of a narrative and the structure encodes their delivery,
in the present work we examine the emotional arc that is invoked through the words used.
The emotional arc of a story does not give us direct information about the plot or the intended meaning of the story,
but rather exists as part of the whole narrative
(e.g., an emotional arc showing a fall in sentiment throughout a story may arise from very different plot and structure combinations).
This distinction between the emotional arc and the plot of a story is one point of misunderstanding in other work that has drawn criticism from the digital humanities community \citep{jockers2014novel}.
Through the identification of motifs,
narrative theories allow us to analyze, interpret, describe, and compare stories across cultures and regions of the world \citep{dundes1997motif,dolby2008literary,uther2011types}.
We show that automated extraction of emotional arcs is not only possibly,
but can test previous theories and provide new insights with the potential to quantify unobserved trends as the field transitions from data-scarce to data-rich \citep{kirschenbaum2007remaking,moretti2013distant}.

The rejected master's thesis of Kurt Vonnegut---which he personally considered his greatest contribution---defines the \textit{emotional arc} of a story on the ``Beginning--End'' and ``Ill Fortune--Great Fortune'' axes \citep{vonnegut1981palm}.
Vonnegut finds a remarkable similarity between Cinderella and the origin story of Christianity in the Old Testament (see Fig.~\ref{fig:vonnegut} in Appendix~\ref{sec:extras}),
leading us to search for all such groupings.
In a recorded lecture available on YouTube \citep{vonnegut1995shapes},
Vonnegut asserted:
\begin{displayquote}
  ``There is no reason why the simple shapes of stories can't be fed into computers, they are beautiful shapes.''
\end{displayquote}

For our analysis,
we apply three independent tools: Matrix decomposition by Singular Value Decomposition (SVD),
supervised learning by agglomerative (hierarchical) clustering with Ward's method,
and unsupervised learning by a Self Organizing Map (SOM, a type of neural network).
Each tool has different strengths: the SVD finds the underlying basis of all of the emotional arcs,
the clustering classifies the emotional arcs into distinct groups,
and the SOM generates arcs from noise which are similar to those in our corpus using a stochastic process.
By considering the results of each tool independently, we are able to confirm our findings of broad support.

We proceed as follows.
We first introduce our methods in Section~\ref{sec:methods},
we then discuss the combined results of each method in Section~\ref{sec:results},
and we present our conclusions in Section~\ref{sec:arcs-conclusion}.
A graphical outline of the methodology and results can be found as Fig.~\ref{fig:infographic} in Appendix~\ref{sec:extras}.

\section{Methods}
\label{sec:methods}

\subsection{Emotional arc construction}
\begin{figure}[tbp!]
  \centering
\includegraphics[width=0.48\textwidth]{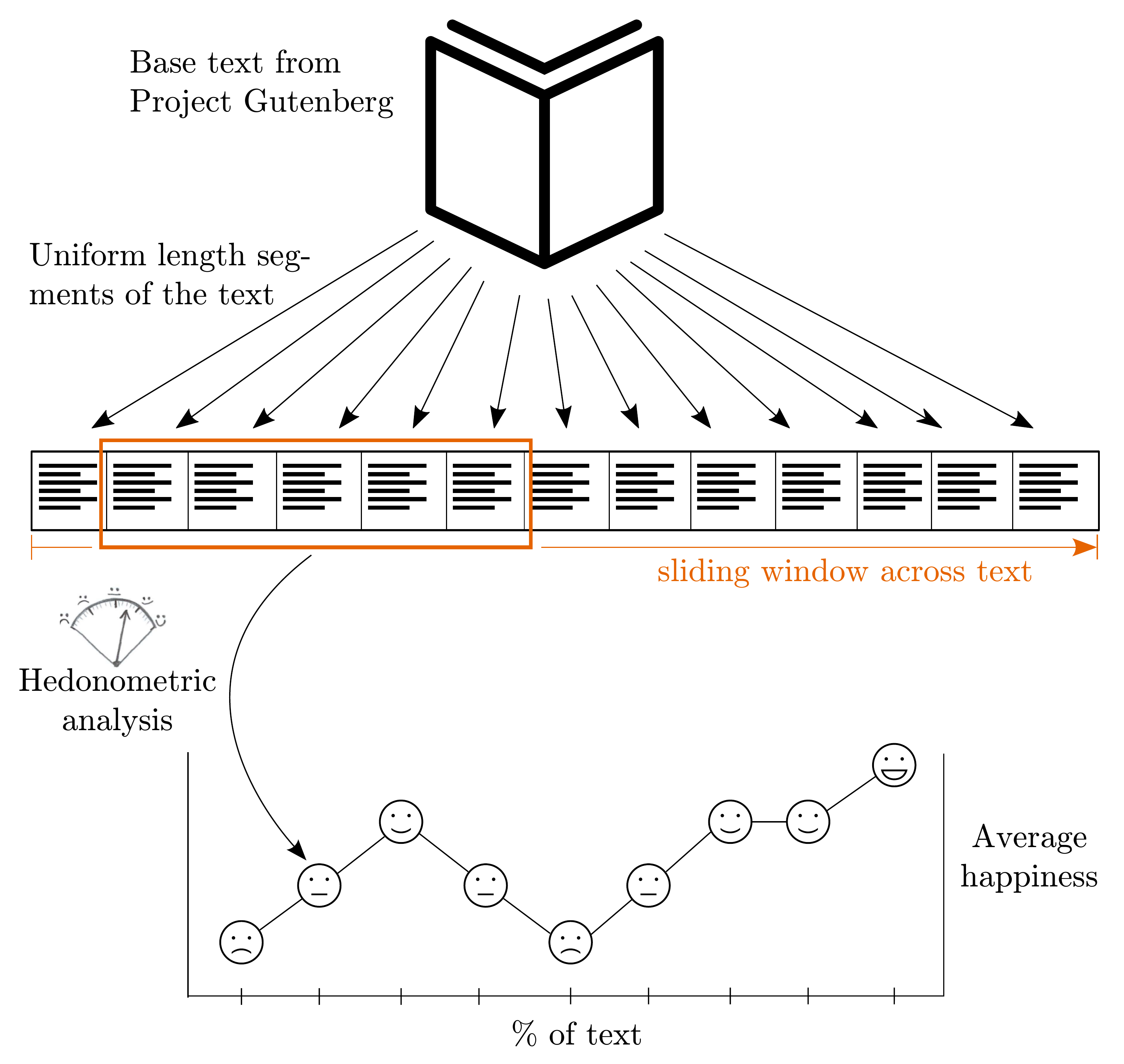}
  \caption[]{
    Schematic of how we compute emotional arcs.
    The indicated uniform length segments (gap between samples) taken from the text form the sample with fixed window size set at $N_w = 10,000$ words.
    The segment length is thus $N_s = (N -(N_w+1))/n$ for $N$ the  length of the book in words,
    and $n$ the number of points in the time series.
    Sliding this fixed size window through the book,
    we generate $n$ sentiment scores with the Hedonometer,
    which comprise the emotional arc \citep{dodds2011temporal}.
  }
  \label{fig:timeseries-schematic}
\end{figure}

\begin{figure*}[tbp!]
  \centering
\includegraphics[width=0.97\textwidth]{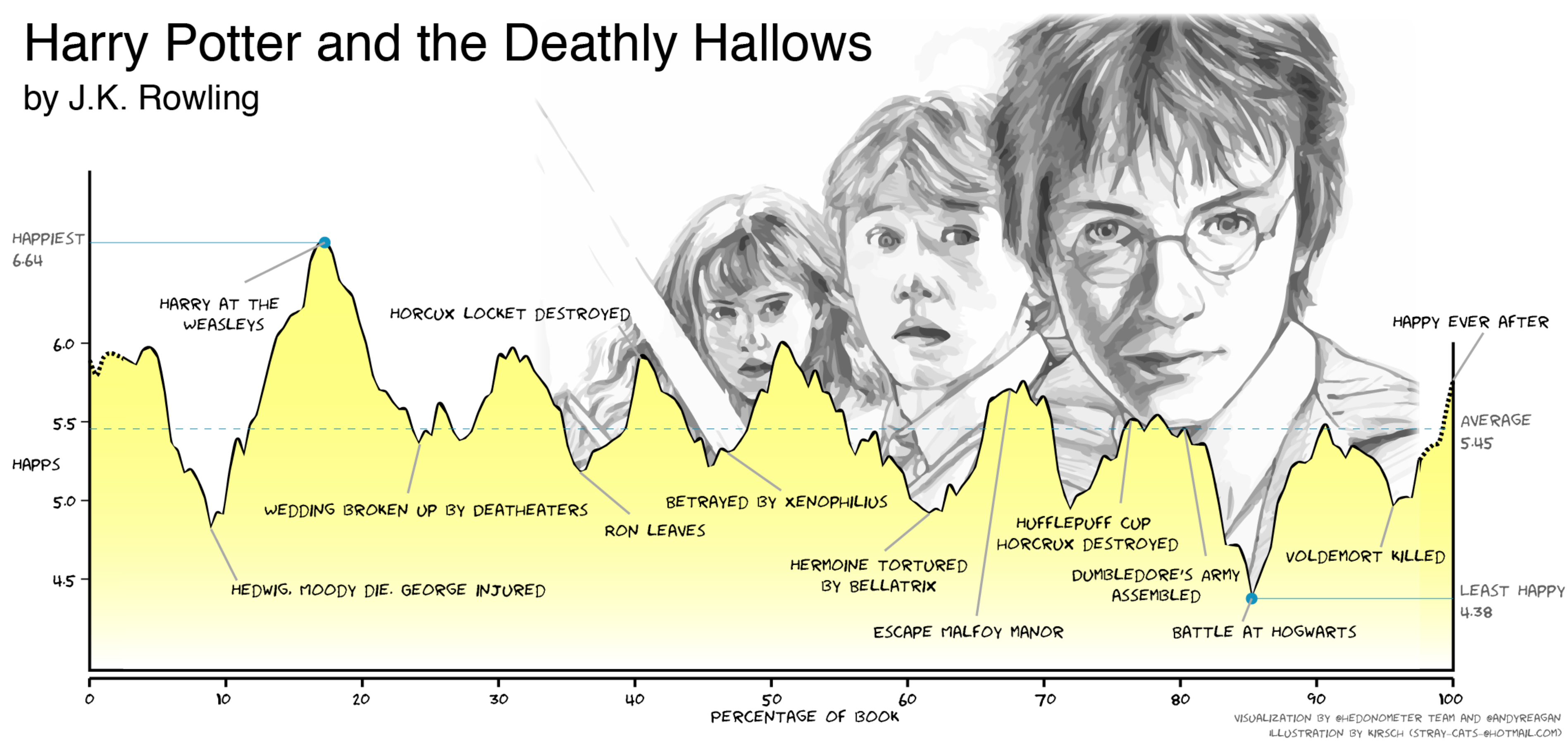}
  \caption[]{
    Annotated emotional arc of \textit{Harry Potter and the Deathly Hallows},
    by J.K. Rowling,
    inspired by the illustration made by Medaris for The Why Files \citep{tenenbaum2015languages}.
    The entire seven book series can be classified as a ``Kill the monster'' plot \citep{booker2006seven},
    while the many sub plots and connections between them complicate the emotional arc of each individual book: this plot could not be readily inferred from the emotional arc alone.
    The emotional arc shown here
    captures the major highs and lows of the story,
    and should be familiar to any reader well acquainted with Harry Potter.
    Our method does not pick up emotional moments discussed briefly,
    perhaps in one paragraph or sentence (e.g., the first kiss of Harry and Ginny).
    We provide interactive visualizations of all Project Gutenberg books at \href{http://hedonometer.org/books/v3/1/}{http://hedonometer.org/books/v3/1/} and a selection of classic and popular books at \href{http://hedonometer.org/books/v1/}{http://hedonometer.org/books/v1/}.
  }
  \label{fig:harry-potter}
\end{figure*}

To generate emotional arcs,
we analyze the sentiment of 10,000 word windows,
which we slide through the text (see Fig.~\ref{fig:timeseries-schematic}).
We rate the emotional content of each window using our Hedonometer with the labMT dataset,
chosen for lexical coverage and its ability to generate meaningful word shift graphs,
specifically using 10,000 words as a minimum necessary to generate meaningful sentiment scores \citep{reagan2016benchmarking,ribeiro2016sentibench}.
We emphasize that dictionary-based methods for sentiment analysis usually perform worse than random on individual sentences \citep{reagan2016benchmarking,ribeiro2016sentibench},
and although this issue can be resolved by using a rolling average of sentences scores,
it betrays a basic misunderstanding of similar efforts \citep{jockers2014novel}.
In Fig.~\ref{fig:harry-potter},
we show the emotional arc of \textit{Harry Potter and the Deathly Hallows},
the final book in the popular Harry Potter series by J.K. Rowling.
While the plot of the book is nested and complicated,
the emotional arc associated with each sub-narrative is clearly visible.
We analyze the emotional arcs corresponding to complete books,
and to limit the conflation of multiple core emotional arcs, we restrict our analysis to shorter books by selecting a maximum number of words when building our filter.
Further details of the emotional arc construction can be found in Appendix~\ref{sec:construction}.

\subsection{Project Gutenberg Corpus}

For a suitable corpus we draw on the open access Project Gutenberg data set \citep{gutenberg}.
We apply rough filters to the collection (roughly 50,000 books) in an attempt to obtain a set of books that represent English works of fiction.
We start by selecting for only English books,
with total words between 20,000 and 100,000,
with more than 40 downloads from the Project Gutenberg website,
and with Library of Congress Class corresponding to English fiction\footnotemark.
\footnotetext{The specific classes have labels PN, PR, PS, and PZ.}
To ensure that the 40-download limit is not influencing the results here,
we repeat the entire analysis for each method with 10, 20, 40, and 80 download thresholds,
in each case confirming the 40 download findings to be qualitatively unchanged.
Next, we remove books with any word in the title from a list of keywords (e.g., ``poems'' and ``collection'',
full list in Appendix~\ref{sec:construction}).
From within this set of books,
we remove the front and back matter of each book using regular expression pattern matches that match on 98.9\% of the books included.
Two slices of the data for download count and the total word count are shown in Appendix~\ref{sec:construction} Fig.~\ref{fig:length-distribution}.\
We provide a list of the book ID's which are included for download in the Online Appendices at \href{http://compstorylab.org/share/papers/reagan2016b/}{http://compstorylab.org/share/papers/reagan2016b/},
the books are listed in Table~\ref{tbl:allbooks} in Appendix~\ref{sec:lists},
and we attempt to provide the Project Gutenberg ID when we mention a book by title herein.
Given the Project Gutenberg ID $n$,
the raw ebook is available online from Project Gutenberg at \href{http://www.gutenberg.org/ebooks/n}{http://www.gutenberg.org/ebooks/n},
e.g., \textit{Alice's Adventures in Wonderland} by Lewis Carroll,
has ID 11 and is available at \href{http://www.gutenberg.org/ebooks/11}{http://www.gutenberg.org/ebooks/11}.
We also provide an online, interactive version of the emotional arc for each book indexed by the ID,
e.g., \textit{Alice's Adventures in Wonderland} is available at \href{http://hedonometer.org/books/v3/11/}{http://hedonometer.org/books/v3/11/}.

\subsection{Principal Component Analysis (SVD)}

We use the standard linear algebra technique Singular Value Decomposition (SVD) to find a decomposition of stories onto an orthogonal basis of emotional arcs.
Starting with the emotional arc (sentiment time series) for each book $b_i$ as row $i$ in the matrix $A$,
we apply the SVD to find
\begin{align}
  A  &= U \Sigma V^{T}
  = W V^{T}
  ,
  \label{eq:SVD}
\end{align}
where $U$ contains the projection of each sentiment time series onto each of the right singular vectors (rows of $V^{T}$,
eigenvectors of $A^TA$),
which have singular values given along the diagonal of $\Sigma$,
with $W = U \Sigma$.
Different intuitive interpretations of the matrices $U, \Sigma,$ and $V^T$ are useful in the various domains in which the SVD is applied; here,
we focus on right singular vectors as an orthonormal basis for the sentiment time series in the rows of $A$,
which we will refer to as the \textit{modes}.
We combine $\Sigma$ and $U$ into the single coefficient matrix $W$ for clarity and convenience,
such that $W$ now represents the mode coefficients.

\subsection{Hierarchical Clustering}
\label{sec:clustering}

We use Ward's method to generate a hierarchical clustering of stories,
which proceeds by minimizing variance between clusters of books \citep{ward1963hierarchical}.
We use the mean-centered books and the distance function
\begin{equation}
  D(b_i, b_j )
  =
  l^{-1}
  \sum ^l _{t=1} |b_i(t) - b_j (t)|
  .
  \label{eq:distance}
\end{equation}
for $t$ indexing the window in books $b_i, b_j$ to generate the distance matrix.

\subsection{Self Organizing Map (SOM)}
\label{sec:SOM}

We implement a Self Organized Map (SOM),
an unsupervised machine learning method (a type of neural network)
to cluster emotional arcs \citep{kohonen1990self}.
The SOM works by finding the most similar emotional arc in a random collection of arcs.
We use an 8x8 SOM (for 64 nodes,
roughly 5\% of the number of books),
connected on a square grid,
training according to the original procedure (with winner take all, and scaling functions across both distance and magnitude).
We take the neighborhood influence function at iteration $i$ as
\begin{equation}
  \text{Nbd}_{k}(i) = \left [ j \in \mathcal{N} ~|~ D(k,j) < \sqrt{N} \, \cdot (i+1)^\alpha \right ]
\end{equation}
for a node $k$ in the set of nodes $\mathcal{N}$,
with distance function $D$ given above and total number of nodes $N$.
For results shown here we take $\alpha = -0.15$.
We implement the learning adaptation function at training iteration $i$ as
$f(i) = (i+1)^{\beta}$,
again with $\beta = -0.15$, a standard value for the training hyper-parameters.

\section{Results}
\label{sec:results}

We obtain a collection of \nbooks~books that are mostly,
but not all,
fictional stories by using metadata from Project Gutenberg to construct a rough filter.
We find broad support for the following six emotional arcs:
\begin{itemize}
\item ``Rags to riches'' (rise).
\item ``Tragedy'', or ``Riches to rags'' (fall).
\item ``Man in a hole'' (fall-rise).
\item ``Icarus'' (rise-fall).
\item ``Cinderella'' (rise-fall-rise).
\item ``Oedipus'' (fall-rise-fall).
\end{itemize}

\noindent Importantly,
we obtain these same six emotional arcs from all possible arcs by observing them as the result of three methods:
As modes from a matrix decomposition by SVD,
as clusters in a hierarchical clustering using Ward's algorithm,
and as clusters using unsupervised machine learning.
We examine each of the results in this section.

\subsection{Principal Component Analysis (SVD)}

\begin{figure*}[tbp!]
  \centering
\includegraphics[width=0.9\textwidth]{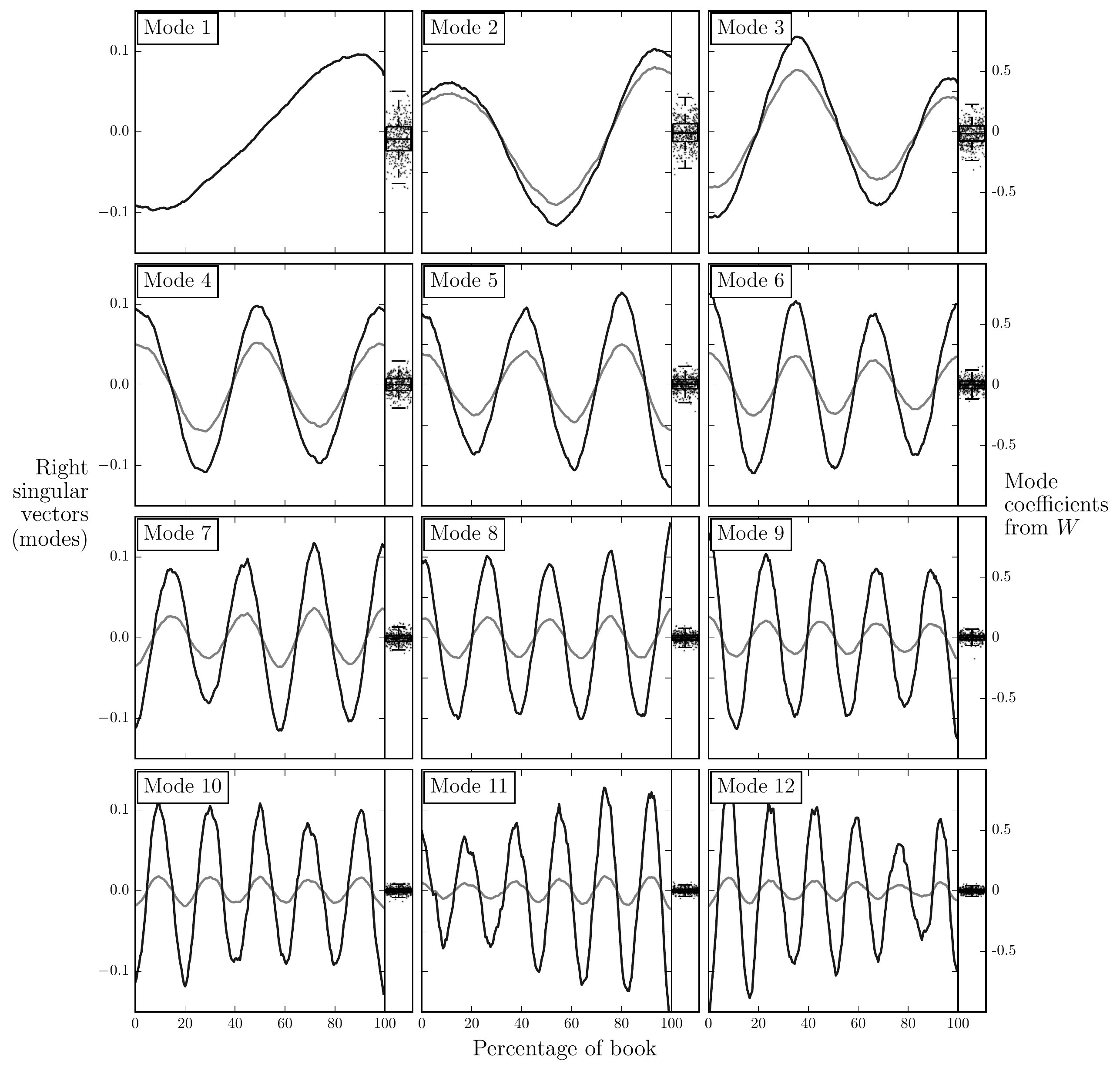}
  \caption[]{
    Top 12 modes from the Singular Value Decomposition of \nbooks~Project Gutenberg books.
    We show in a lighter color modes weighted by their corresponding singular value,
    where we have scaled the matrix $\Sigma$ such that the first entry is 1 for comparison (for reference, the largest singular value is 34.5).
    The mode coefficients normalized for each book are shown in the right panel accompanying each mode, in the range -1 to 1, with the ``Tukey'' box plot.
  }
  \label{fig:SVD-12-comp}
\end{figure*}

In Fig.~\ref{fig:SVD-12-comp} we show the leading 12 modes in both the weighted (dark) and un-weighted (lighter) representation.
In total, the first 12 modes explain 80\% and 94\% of the variance from the mean centered and raw time series, respectively.
The modes are from mean-centered emotional arcs,
such that the first SVD mode need not extract the average from the labMT scores nor the positivity bias present in language \citep{dodds2015human}.
The coefficients for each mode within a single emotional arc are both positive and negative,
so we need to consider both the modes and their negation.
We can immediately recognize the familiar shapes of core emotional arcs in the first four modes,
and compositions of these emotional arcs in modes 5 and 6.
We observe ``Rags to riches'' (mode 1, positive),
``Tragedy'' or ``Riches to rags'' (mode 1, negative),
Vonnegut's ``Man in a hole'' (mode 2, positive),
``Icarus'' (mode 2, negative),
``Cinderella'' (mode 3, positive),
``Oedipus'' (mode 3, negative).
We choose to include modes 7--12 only for completeness,
as these high frequency modes have little contribution to variance and do not align with core emotional arc archetypes from other methods (more below).

We emphasize that by definition of the SVD,
the mode coefficients in $W$ can be either positive and negative,
such that the modes themselves explain variance with both the positive and negative version.
In the right panels of each mode in Fig.~\ref{fig:SVD-12-comp} we project the \nbooks~stories onto each of first six modes and show the resulting coefficients.
While none are far from 0 (as would be expected),
mode 1 has a mean slightly above 0 and both modes 3 and 4 have means slightly below 0.
To sort the books by their coefficient for each mode,
we normalize the coefficients within each book in the rows of $W$ to sum to 1,
accounting for books with higher total energy,
and these are the coefficients shown in the right panels of each mode in Fig.~\ref{fig:SVD-12-comp}.
In Appendix~\ref{sec:SVD-supp},
we provide supporting, intuitive details of the SVD method,
as well as example emotional arc reconstruction using the modes (see Figs.~\ref{fig:SVD-USV}--\ref{fig:SVD-reconstruction}).
As expected, less than 10 modes are enough to reconstruct the emotional arc to a degree of accuracy visible to the eye.

\begin{figure*}[tbp!]
  \centering
\includegraphics[width=.98\textwidth]{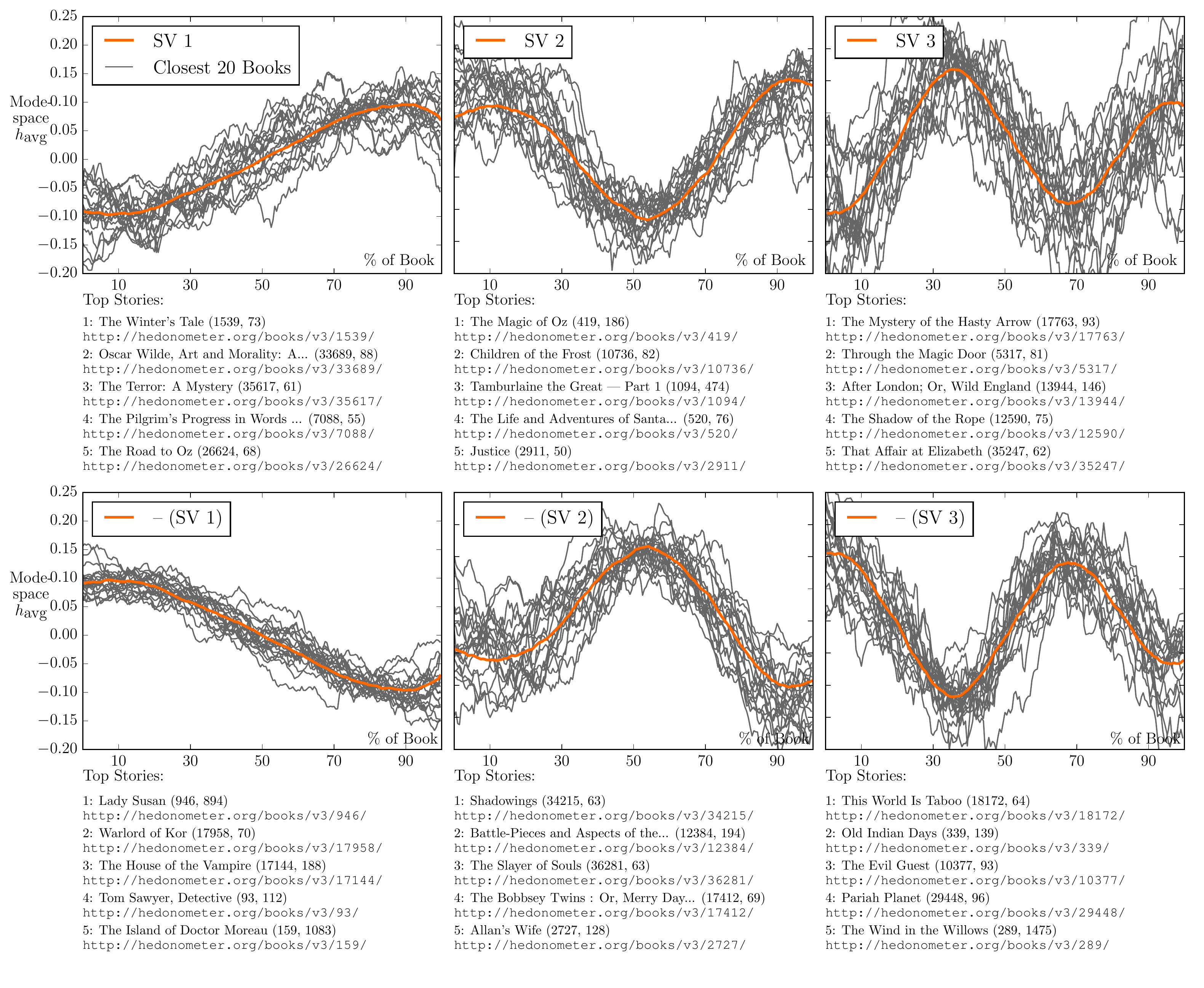}
  \caption[]{
    First 3 SVD modes and their negation with the closest stories to each.
    To locate the emotional arcs on the same scale as the modes,
    we show the modes directly from the rows of $V^T$ and weight the emotional arcs by the inverse of their coefficient in $W$ for the particular mode.
    The closest stories shown for each mode are those stories with emotional arcs which have the greatest coefficient in $W$.
    In parentheses for each story is the Project Gutenberg ID and the number of downloads from the Project Gutenberg website, respectively.
    Links below each story point to an interactive visualization on \href{http://hedonometer.org}{http://hedonometer.org} which enables detailed exploration of the emotional arc for the story.
  }
  \label{fig:SVD-1-3-labelled}
\end{figure*}

We show labeled examples of the emotional arcs closest to the top 6 modes in Figs.~\ref{fig:SVD-1-3-labelled} and~\ref{fig:SVD-4-6-labelled}.

We present both the positive and negative modes,
and the stories closest to each by sorting on the coefficient for that mode.
For the positive stories, we sort in ascending order, and vice versa.
Mode 1, which encompasses both the ``Rags to riches'' and ``Tragedy'' emotional arcs, captures 30\% of the variance of the entire space.
We examine the closest stories to both sides of modes 1--3,
and direct the reader to Fig.~\ref{fig:SVD-4-6-labelled} for more details on the higher order modes.
The two stories that have the most support from the ``Rags to riches'' mode are \textit{The Winter's Tale} (1539) and \textit{Oscar Wilde, Art and Morality: A Defence of ``The Picture of Dorian Gray''} (33689).
Among the most categorical tragedies we find \textit{Lady Susan} (946) and \textit{Warlord of Kor} (17958).
Number 8 in the sorted list of tragedies is perhaps the most famous tragedy: \textit{Romeo and Juliet} by William Shakespeare.
Mode 2 is the ``Man in a hole'' emotional arc,
and we find the stories which most closely follow this path to be \textit{The Magic of Oz} (419) and \textit{Children of the Frost} (10736).
The negation of mode 2 most closely resembles the emotional arc of the ``Icarus'' narrative.
For this emotional arc,
the most characteristic stories are \textit{Shadowings} (34215) and \textit{Battle-Pieces and Aspects of the War} (12384).
Mode 3 is the ``Cinderella'' emotional arc,
and includes \textit{Mystery of the Hasty Arrow} (17763) and \textit{Through the Magic Dorr} (5317).
The negation of Mode 3, which we refer to as ``Oedipus'', is found most characteristically in \textit{This World is Taboo} (18172), \textit{Old Indian Days} (339), and \textit{The Evil Guest} (10377).
We also note that the spread of the stories from their core mode increases strongly for the higher modes.

\subsection{Hierarchical Clustering}
\label{sec:clustering}

\begin{figure*}[tbp!]
  \centering
\includegraphics[width=0.98\textwidth]{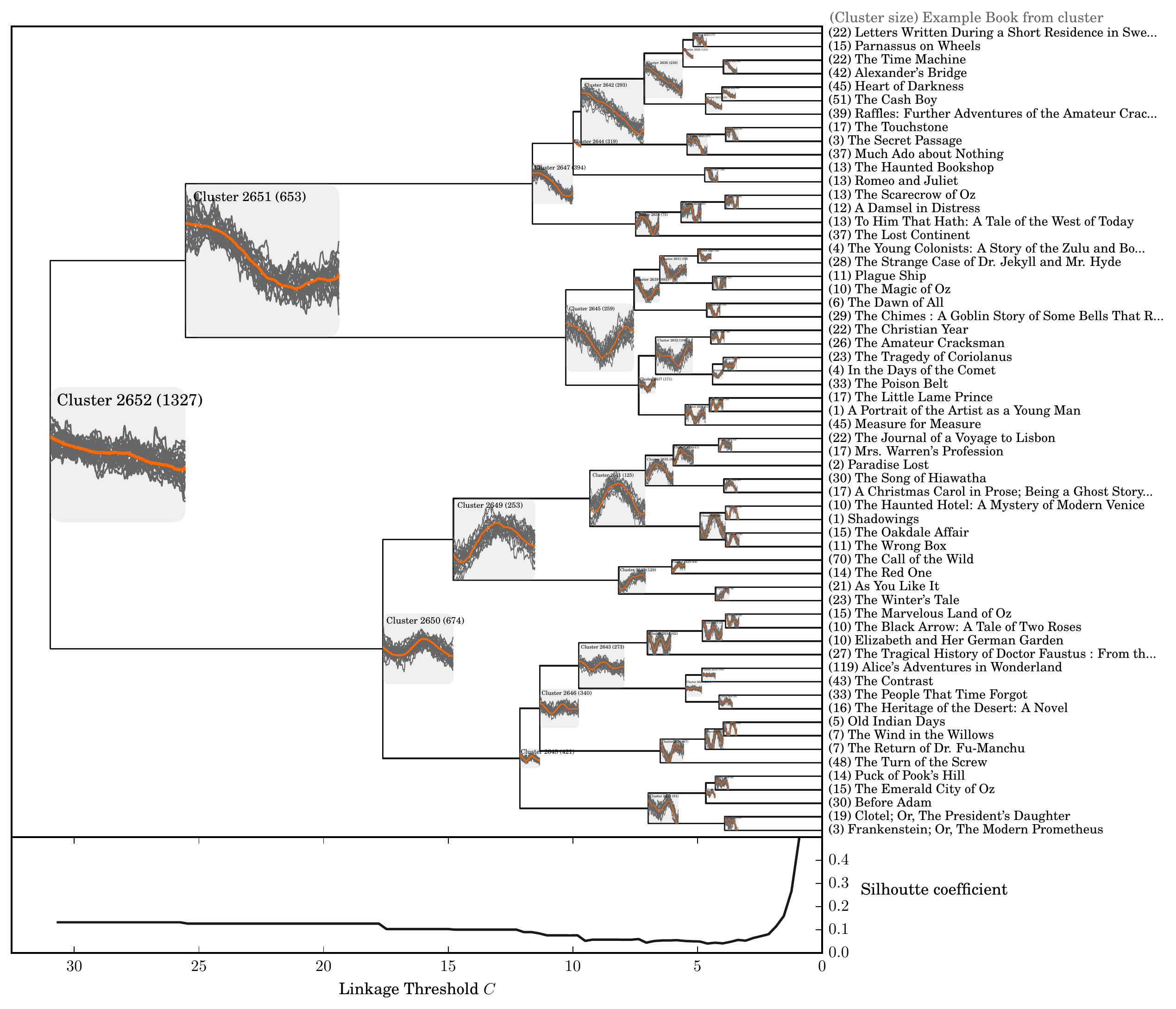}
  \caption[]{
    Dendrogram from the hierarchical clustering procedure using Ward's minimum variance method.
    For each cluster,
    a selection of the 20 most central books to a fully-connected network of books are shown along with the average of the emotional arc for all books in the cluster,
    along with the cluster ID and number of books in each cluster (shown in parenthesis).
    The cluster ID is given by numbering the clusters in order of linkage starting at 0, with each individual book representing a cluster of size 1 such that the final cluster (all books) has the ID $2(N-1)$ for the $N=\nbooks$ books.
    At the bottom,
    we show the average Silhouette value for all books,
    with higher value representing a more appropriate number of clusters.
    For each of the 60 leaf nodes (right side) we show the number of books within the cluster and the most central book to that cluster's book network.
  }
  \label{fig:ward-small}
\end{figure*}

We show a dendrogram of the 60 clusters with highest linkage cost in Fig.~\ref{fig:ward-small}.
The average silhouette coefficient is shown on the bottom of Fig.~\ref{fig:ward-small},
and the distributions of silhouette values within each cluster (see Figs.~\ref{fig:clustering-2-5-clusters}--\ref{fig:clustering-6-9-clusters}) can be used to analyze the appropriate number of clusters \citep{rousseeuw1987silhouettes}.
A characteristic book from each cluster is shown on the leaf nodes by sorting the books within each cluster by the total distance to other books in the cluster (e.g.,
considering each intra-cluster collection as a fully connected weighted network,
we take the most central node),
and in parenthesis the number of books in that cluster.
In other words,
we label each cluster by considering the network centrality of the fully connected cluster with edges weighted by the distance between stories.
By cutting the dendrogram in Fig.~\ref{fig:ward-small} at various linkage costs we are able to extract clusters of the desired granularity.
For the cuts labeled C2, C4, and C8, we show these clusters in Figs.~\ref{fig:ward-cluster-2},~\ref{fig:ward-cluster-4}, and~\ref{fig:ward-cluster-8}.
We find the first four of our final six arcs appearing among the eight most different clusters (Fig.~\ref{fig:ward-cluster-8}).

The clustering method groups stories with a ``Man in a hole'' emotional arc for a range of different variances,
separate from the other arcs.
In total these clusters (Panel A, E, and I of Fig.~\ref{fig:ward-cluster-9}) account for 30\% of the Gutenberg corpus.
The remainder of the stories have emotional arcs that are clustered among the ``Tragedy'' arc (32\%),
``Rags to riches'' arc (5\%),
and the ``Oedipus'' arc (31\%).
A more detailed analysis of the results from hierarchical clustering can be found in Appendix~\ref{sec:clustering-supp},
and this result generally agrees with other attempts that use only hierarchical clustering \citep{jockers2015rest}.a

\subsection{Self Organizing Map (SOM)}
\label{sec:SOM}

Finally, we apply Kohonen's Self-Organizing Map (SOM) and find core arcs from unsupervised machine learning on the emotional arcs.
On the two dimensional component plane, the prescribed network topology, we find seven spatially coherent groups, with five emotional arcs.
These spatial groups are comprised of stories with core emotional arcs of differing variance.

\begin{figure*}[tbp!]
  \centering
\includegraphics[width=0.98\textwidth]{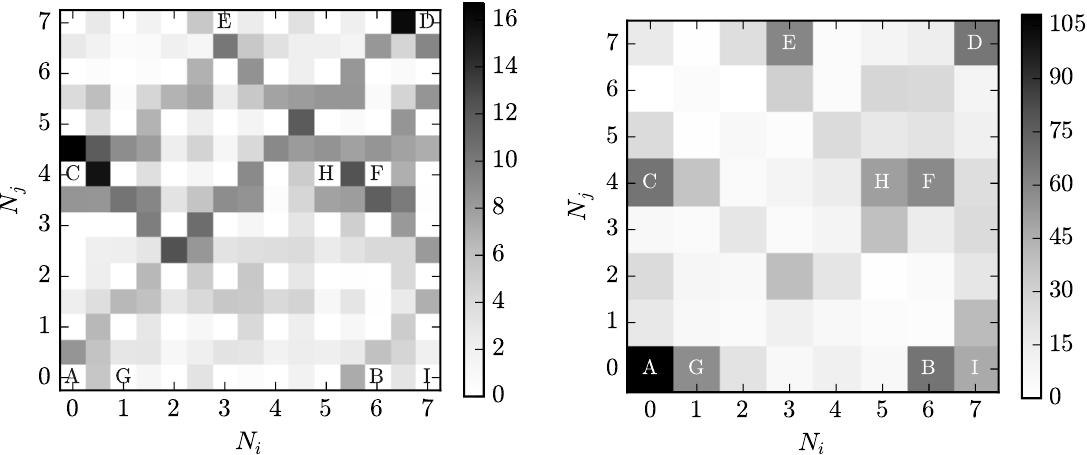}
  \caption[]{
    Results of the SOM applied to Project Gutenberg books.
    Left panel: Nodes on the 2D SOM grid are shaded by the number of stories for which they are the winner.
    Right panel: The B-Matrix shows that there are clear clusters of stories in the 2D space imposed by the SOM network.
  }
  \label{fig:SOM-matrices}
\end{figure*}

In Fig.~\ref{fig:SOM-matrices} we see both the B-Matrix to demonstrate the strength of spatial clustering and a heat-map showing where we find the winning nodes.
The A--I labels refer to the individual nodes shown in Fig.~\ref{fig:SOM-stories},
and we observe seven spatial groups in both panels of Fig.~\ref{fig:SOM-matrices}: (1) A and G, (2) B and I, (3) C, (4) D, (5) E, and (6) H, and (7) F.
These spatial clusters reinforce the visible similarity of the winning node arcs, given that nodes H and F are close spatially but separated by the B-Matrix and contain very distinct arcs.
We show the winning node emotional arcs and the arcs of books for which they are the winners in Fig.~\ref{fig:SOM-stories}.
The legend shows the node ID, numbers the cluster by size, and in parentheses indicates the size of the cluster on that individual node.
In Panels A and G we see varying strengths of the ``Man in a hole'' emotional arc.
In Panels B and I, the second largest individual cluster consists of the ``Rags to riches'' arcs.
In Panel C, and in Panel F, we find the ``Oedipus'' emotional arc,
with a more pronounced positive start and decline in Panel C.
In Panel D we see the ``Icarus'' arc, and in Panel E and Panel H we see the ``Tragedy'' arc.
Each of these top stories are all readily identifiable,
yet again demonstrating the universality of these story types.

\subsection{Null comparison}

There are many possible emotional arcs in the space that we consider.
To demonstrate that these specific arcs are uniquely compelling as stories written by and for \textit{homo narrativus},
we consider the true emotional arcs in relation to their most suitable comparison: the book with randomly shuffled words (``word salad'') and the resulting text from a 2-gram Markov model trained on the individual book itself (``nonsense'').
We chose to compare to ``word salad'' and ``nonsense'' versions as they are more representative of a null model: written stories that are without coherent plot or structure to generate a coherent emotional arc,
which is not true of a stochastic process (e.g., a random walk model or noise).
Examples of the emotional arc and null emotional arcs for a single book are shown in Fig.~\ref{fig:salad-1},
with 10 ``word salad'' and ``nonsense'' versions.
Sampled text using each method is given in Appendix~\ref{sec:construction}.
We re-run each method on the English fiction Gutenberg Corpus with the null versions of each book and verify that the emotional arcs of real stories are not simply an artifact.
The singular value spectrum from the SVD is flatter,
with higher-frequency modes appearing more quickly,
and in total representing 45\% of the total variance present in real stories (see Figs.~\ref{fig:SVD-12-comp-salad} and~\ref{fig:SVD-spectrum-comparison}).
Hierarchical clustering generates less distinct clusters with considerably lower linkage cost (final linkage cost 1400 vs 7000) for the emotional arcs from nonsense books,
and the winning node vectors on a self-organizing map lack coherent structure (see Figs.~\ref{fig:ward-small-salad} and~\ref{fig:SOM-stories-salad} in Appendix~\ref{sec:shuffled}).

\subsection{The Success of Stories}

\begin{figure*}[tbp!]
  \centering
\includegraphics[width=0.9\textwidth]{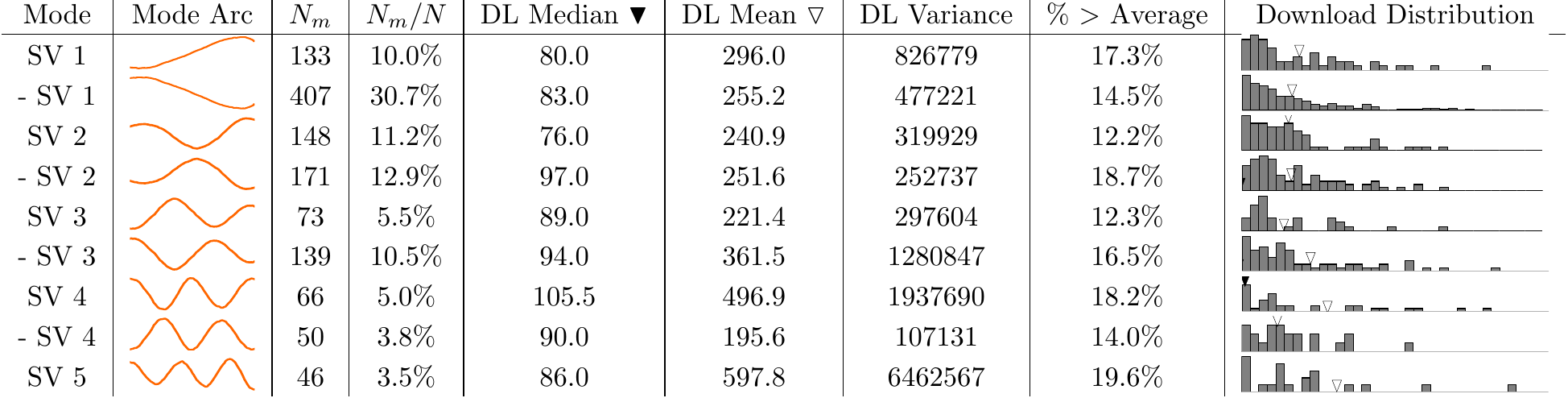}
  \caption[]{
    Download statistics for stories whose SVD Modes comprise more than 2.5\% of books,
    for $N$ the total number of books and $N_m$ the number corresponding to the particular mode.
    Modes \textit{SV 3} through \textit{-SV 4} (both polarities of modes 3 and 4) exhibit a higher average number of downloads and more variance than the others.
    Mode arcs are rows of $V^{T}$ and the download distribution is show in $\log_{10}$ space from 20 to 30,000 downloads.
  }
  \label{fig:download-table-2point5}
\end{figure*}

To examine how the emotional trajectory impacts success,
in Fig.~\ref{fig:download-table-2point5} we examine the downloads for all of the books that are most similar to each SVD mode (for additional modes,
see Fig.~\ref{fig:download-table-point5} in Appendix~\ref{sec:extras}).
We find that the first four modes,
which contain the greatest total number of books,
are not the most popular.
Along with the negative of mode 2, both polarities of modes 3 and 4 have markedly higher median downloads,
while we discount the importance of the mean with the high variance.
The success of the stories underlying these emotional arcs suggests that the emotional experience of readers strongly affects how stories are shared.
We find ``Icarus'' (-SV 2),
``Oedipus'' (-SV 3), and
two sequential ``Man in a hole'' arcs (SV 4),
are the three most successful emotional arcs.
These results are influenced by individual books within each mode which have high numbers of downloads,
and we refer the reader to the download-sorted tables for each mode in Appendix~\ref{sec:SVD-supp}.

\section{Conclusion}
\label{sec:arcs-conclusion}

Using three distinct methods,
we have demonstrated that there is strong support for six core emotional arcs.
Our methodology brings to bear a cross section of data science tools with a knowledge of the potential issues that each present.
We have also shown that consideration of the emotional arc for a given story is important for the success of that story.
Of course, downloads are only a rough proxy for success, and this work may provide an outline for more detailed analysis of the factors that impact meaningful measures of success, i.e., sales or cultural influence.

Our approach could be applied in the opposite direction: namely by beginning with the emotional arc and aiding in the generation of compelling stories \citep{li2013story}.
Understanding the emotional arcs of stories may be useful to aid in constructing arguments \citep{bex2010persuasive} and teaching common sense to artificial intelligence systems \citep{riedl2015using}.

Extensions of our analysis that use a more curated selection of full-text fiction can answer more detailed questions about which stories are the most popular throughout time,
and across regions \citep{silva2016comparative}.
Automatic extraction of character networks would allow a more detailed analysis of plot structure for the Project Gutenberg corpus used here \citep{bost2016narrative,prado2016temporal,min2016narrative}.
Bridging the gap between the full text stories \citep{nenkova2012survey} and systems that analyze plot sequences will allow such systems to undertake studies of this scale \citep{winston2011strong}.
Place could also be used to consider separate character networks through time,
and to help build an analog to Randall Munroe's Movie Narrative Charts \citep{munroe657}.

We are producing data at an ever increasing rate,
including rich sources of stories written to entertain and share knowledge,
from books to television series to news.
Of profound scientific interest will be the degree to which we can eventually understand the full landscape of human stories,
and data driven approaches will play a crucial role.

PSD and CMD acknowledge support from NSF Big Data Grant \#1447634.

\singlespacing
\bibliographystyle{chicago}
\bibliography{everything}

\begin{references}{}

\bibitem[\protect\citeauthoryear{Alajajian, Williams, Reagan, Alajajian, Frank,
  Mitchell, Lahne, Danforth, and Dodds}{Alajajian
  et~al.}{2016}]{alajajian2015lexicocalorimeter}
Alajajian, S.~E., J.~R. Williams, A.~J. Reagan, S.~C. Alajajian, M.~R. Frank,
  L.~Mitchell, J.~Lahne, C.~M. Danforth, and P.~S. Dodds (2016).
\newblock The {L}exicocalorimeter: {G}auging public health through caloric
  input and output on social media.
\newblock Available at
  \href{http://arxiv.org/abs/1507.05098}{http://arxiv.org/abs/1507.05098}.

\bibitem[\protect\citeauthoryear{Baccianella, Esuli, and
  Sebastiani}{Baccianella et~al.}{2010}]{baccianella2010sentiwordnet}
Baccianella, S., A.~Esuli, and F.~Sebastiani (2010).
\newblock Senti{W}ord{N}et 3.0: An enhanced lexical resource for sentiment
  analysis and opinion mining.
\newblock In {\em LREC}, Volume~10, pp.\  2200--2204.

\bibitem[\protect\citeauthoryear{Bird}{Bird}{2006}]{bird2006nltk}
Bird, S. (2006).
\newblock Nltk: the natural language toolkit.
\newblock In {\em Proceedings of the COLING/ACL on Interactive presentation
  sessions}, pp.\  69--72. Association for Computational Linguistics.

\bibitem[\protect\citeauthoryear{Bollen, Mao, and Zeng}{Bollen
  et~al.}{2011}]{bollen2011twitter}
Bollen, J., H.~Mao, and X.~Zeng (2011).
\newblock Twitter mood predicts the stock market.
\newblock {\em Journal of Computational Science\/}~{\em 2\/}(1), 1--8.

\bibitem[\protect\citeauthoryear{Bradley and Lang}{Bradley and
  Lang}{1999}]{bradley1999affective}
Bradley, M.~M. and P.~J. Lang (1999).
\newblock Affective norms for english words ({ANEW}): Stimuli, instruction
  manual and affective ratings.
\newblock Technical report c-1, University of Florida, Gainesville, FL.

\bibitem[\protect\citeauthoryear{Cambria, Olsher, and Rajagopal}{Cambria
  et~al.}{2014}]{cambria2014senticnet}
Cambria, E., D.~Olsher, and D.~Rajagopal (2014).
\newblock Senticnet 3: a common and common-sense knowledge base for
  cognition-driven sentiment analysis.
\newblock In {\em Proceedings of the twenty-eighth AAAI conference on
  artificial intelligence}, pp.\  1515--1521. AAAI Press.

\bibitem[\protect\citeauthoryear{Chung and Liu}{Chung and
  Liu}{2011}]{chung2011predicting}
Chung, S. and S.~Liu (2011).
\newblock Predicting stock market fluctuations from {T}witter.
\newblock {\em Berkeley, California\/}.

\bibitem[\protect\citeauthoryear{De~Smedt and Daelemans}{De~Smedt and
  Daelemans}{2012}]{de2012pattern}
De~Smedt, T. and W.~Daelemans (2012).
\newblock Pattern for {P}ython.
\newblock {\em The Journal of Machine Learning Research\/}~{\em 13\/}(1),
  2063--2067.

\bibitem[\protect\citeauthoryear{Dodds, Clark, Desu, Frank, Reagan, Williams,
  Mitchell, Harris, Kloumann, Bagrow, Megerdoomian, McMahon, Tivnan, and
  Danforth}{Dodds et~al.}{2015a}]{dodds2015human}
Dodds, P.~S., E.~M. Clark, S.~Desu, M.~R. Frank, A.~J. Reagan, J.~R. Williams,
  L.~Mitchell, K.~D. Harris, I.~M. Kloumann, J.~P. Bagrow, K.~Megerdoomian,
  M.~T. McMahon, B.~F. Tivnan, and C.~M. Danforth (2015a).
\newblock Human language reveals a universal positivity bias.
\newblock {\em PNAS\/}~{\em 112\/}(8), 2389--2394.

\bibitem[\protect\citeauthoryear{Dodds, Clark, Desu, Frank, Reagan, Williams,
  Mitchell, Harris, Kloumann, Bagrow, Megerdoomian, McMahon, Tivnan, and
  Danforth}{Dodds et~al.}{2015b}]{dodds2015reply}
Dodds, P.~S., E.~M. Clark, S.~Desu, M.~R. Frank, A.~J. Reagan, J.~R. Williams,
  L.~Mitchell, K.~D. Harris, I.~M. Kloumann, J.~P. Bagrow, K.~Megerdoomian,
  M.~T. McMahon, B.~F. Tivnan, and C.~M. Danforth (2015b).
\newblock Reply to garcia et al.: Common mistakes in measuring
  frequency-dependent word characteristics.
\newblock {\em Proceedings of the National Academy of Sciences\/}~{\em
  112\/}(23), E2984--E2985.

\bibitem[\protect\citeauthoryear{Dodds and Danforth}{Dodds and
  Danforth}{2009}]{dodds2009b}
Dodds, P.~S. and C.~M. Danforth (2009, July).
\newblock Measuring the happiness of large-scale written expression: Songs,
  blogs, and presidents.
\newblock {\em Journal of Happiness Studies\/}~{\em 11\/}(4), 441--456.

\bibitem[\protect\citeauthoryear{Dodds, Harris, Kloumann, Bliss, and
  Danforth}{Dodds et~al.}{2011}]{dodds2011temporal}
Dodds, P.~S., K.~D. Harris, I.~M. Kloumann, C.~A. Bliss, and C.~M. Danforth
  (2011, 12).
\newblock Temporal patterns of happiness and information in a global social
  network: Hedonometrics and {T}witter.
\newblock {\em PLoS ONE\/}~{\em 6\/}(12), e26752.

\bibitem[\protect\citeauthoryear{Garcia, Garas, and Schweitzer}{Garcia
  et~al.}{2015}]{garcia2015language}
Garcia, D., A.~Garas, and F.~Schweitzer (2015).
\newblock The language-dependent relationship between word happiness and
  frequency.
\newblock {\em Proceedings of the National Academy of Sciences\/}~{\em
  112\/}(23), E2983.

\bibitem[\protect\citeauthoryear{Golder and Macy}{Golder and
  Macy}{2011}]{golder2011diurnal}
Golder, S.~A. and M.~W. Macy (2011).
\newblock Diurnal and seasonal mood vary with work, sleep, and daylength across
  diverse cultures.
\newblock {\em Science Magazine\/}~{\em 333}, 1878--1881.

\bibitem[\protect\citeauthoryear{Gon{\c{c}}alves, Ara{\'u}jo, Benevenuto, and
  Cha}{Gon{\c{c}}alves et~al.}{2013}]{gonccalves2013comparing}
Gon{\c{c}}alves, P., M.~Ara{\'u}jo, F.~Benevenuto, and M.~Cha (2013).
\newblock Comparing and combining sentiment analysis methods.
\newblock In {\em Proceedings of the first ACM conference on Online social
  networks}, pp.\  27--38. ACM.

\bibitem[\protect\citeauthoryear{Hand and Yu}{Hand and
  Yu}{2001}]{hand2001idiot}
Hand, D.~J. and K.~Yu (2001).
\newblock Idiot's bayes---not so stupid after all?
\newblock {\em International statistical review\/}~{\em 69\/}(3), 385--398.

\bibitem[\protect\citeauthoryear{Hutto and Gilbert}{Hutto and
  Gilbert}{2014}]{hutto2014vader}
Hutto, C.~J. and E.~Gilbert (2014, May).
\newblock Vader: A parsimonious rule-based model for sentiment analysis of
  social media text.
\newblock In {\em Eighth International AAAI Conference on Weblogs and Social
  Media}. AAAI Publications.

\bibitem[\protect\citeauthoryear{Kiritchenko, Zhu, and Mohammad}{Kiritchenko
  et~al.}{2014}]{kiritchenko2014sentiment}
Kiritchenko, S., X.~Zhu, and S.~M. Mohammad (2014).
\newblock Sentiment analysis of short informal texts.
\newblock {\em Journal of Artificial Intelligence Research\/}~{\em 50},
  723--762.

\bibitem[\protect\citeauthoryear{Levallois}{Levallois}{2013}]{levallois2013umigon}
Levallois, C. (2013).
\newblock Umigon: sentiment analysis for tweets based on terms lists and
  heuristics.
\newblock In {\em Second Joint Conference on Lexical and Computational
  Semantics (* SEM)}, Volume~2, pp.\  414--417.

\bibitem[\protect\citeauthoryear{Lin, Michel, Aiden, Orwant, Brockman, and
  Petrov}{Lin et~al.}{2012}]{lin2012syntactic}
Lin, Y., J.-B. Michel, E.~L. Aiden, J.~Orwant, W.~Brockman, and S.~Petrov
  (2012).
\newblock Syntactic annotations for the google books ngram corpus.
\newblock In {\em Proceedings of the ACL 2012 system demonstrations}, pp.\
  169--174. Association for Computational Linguistics.

\bibitem[\protect\citeauthoryear{Liu}{Liu}{2010}]{liu2010sentiment}
Liu, B. (2010).
\newblock Sentiment analysis and subjectivity.
\newblock {\em Handbook of natural language processing\/}~{\em 2}, 627--666.

\bibitem[\protect\citeauthoryear{Liu}{Liu}{2012}]{liu2012sentiment}
Liu, B. (2012, May).
\newblock {\em Sentiment analysis and opinion mining}.
\newblock Synthesis Lectures on Human Language Technologies. San Rafael, CA:
  Morgan \& Claypool Publishers.

\bibitem[\protect\citeauthoryear{Michel, Shen, Aiden, Veres, Gray, Pickett,
  Hoiberg, Clancy, Norvig, Orwant, et~al.}{Michel
  et~al.}{2011}]{michel2011quantitative}
Michel, J.-B., Y.~K. Shen, A.~P. Aiden, A.~Veres, M.~K. Gray, J.~P. Pickett,
  D.~Hoiberg, D.~Clancy, P.~Norvig, J.~Orwant, et~al. (2011).
\newblock Quantitative analysis of culture using millions of digitized books.
\newblock {\em Science\/}~{\em 331\/}(6014), 176--182.

\bibitem[\protect\citeauthoryear{Mitchell, Frank, Harris, Dodds, and
  Danforth}{Mitchell et~al.}{2013}]{mitchell2013happiness}
Mitchell, L., M.~R. Frank, K.~D. Harris, P.~S. Dodds, and C.~M. Danforth (2013,
  May).
\newblock {The Geography of Happiness: Connecting Twitter Sentiment and
  Expression, Demographics, and Objective Characteristics of Place}.
\newblock {\em PLoS ONE\/}~{\em 8\/}(5), e64417.

\bibitem[\protect\citeauthoryear{Mohammad, Kiritchenko, and Zhu}{Mohammad
  et~al.}{2013}]{MohammadKZ2013}
Mohammad, S.~M., S.~Kiritchenko, and X.~Zhu (2013, June).
\newblock Nrc-canada: Building the state-of-the-art in sentiment analysis of
  tweets.
\newblock In {\em Proceedings of the seventh international workshop on Semantic
  Evaluation Exercises (SemEval-2013)}, Atlanta, Georgia, USA.

\bibitem[\protect\citeauthoryear{Mohammad and Turney}{Mohammad and
  Turney}{2013}]{mohammad2013crowdsourcing}
Mohammad, S.~M. and P.~D. Turney (2013).
\newblock Crowdsourcing a word--emotion association lexicon.
\newblock {\em Computational Intelligence\/}~{\em 29\/}(3), 436--465.

\bibitem[\protect\citeauthoryear{Nielsen}{Nielsen}{2011}]{nielsen2011new}
Nielsen, F.~{\AA}. (2011, May).
\newblock A new {ANEW}: Evaluation of a word list for sentiment analysis in
  microblogs.
\newblock In M.~Rowe, M.~Stankovic, A.-S. Dadzie, and M.~Hardey (Eds.), {\em
  CEUR Workshop Proceedings}, Volume Proceedings of the ESWC2011 Workshop on
  'Making Sense of Microposts': Big things come in small packages 718, pp.\
  93--98.

\bibitem[\protect\citeauthoryear{Pang and Lee}{Pang and
  Lee}{2004}]{pang2004sentimental}
Pang, B. and L.~Lee (2004).
\newblock A sentimental education: Sentiment analysis using subjectivity
  summarization based on minimum cuts.
\newblock In {\em Proceedings of the ACL}.

\bibitem[\protect\citeauthoryear{Pappas, Katsimpras, and Stamatatos}{Pappas
  et~al.}{2013}]{pappas2013distinguishing}
Pappas, N., G.~Katsimpras, and E.~Stamatatos (2013).
\newblock Distinguishing the popularity between topics: A system for up-to-date
  opinion retrieval and mining in the web.
\newblock In {\em International Conference on Intelligent Text Processing and
  Computational Linguistics}, pp.\  197--209. Springer.

\bibitem[\protect\citeauthoryear{Pechenick, Danforth, and Dodds}{Pechenick
  et~al.}{2015}]{pechenick2015characterizing}
Pechenick, E.~A., C.~M. Danforth, and P.~S. Dodds (2015).
\newblock Characterizing the google books corpus: Strong limits to inferences
  of socio-cultural and linguistic evolution.
\newblock {\em arXiv preprint arXiv:1501.00960\/}.

\bibitem[\protect\citeauthoryear{Pennebaker, Francis, and Booth}{Pennebaker
  et~al.}{2001}]{pennebaker2001linguistic}
Pennebaker, J.~W., M.~E. Francis, and R.~J. Booth (2001).
\newblock Linguistic inquiry and word count: {LIWC} 2001.
\newblock {\em Mahway: Lawrence Erlbaum Associates\/}~{\em 71}, 2001.

\bibitem[\protect\citeauthoryear{Poria, Gelbukh, Hussain, Howard, Das, and
  Bandyopadhyay}{Poria et~al.}{2013}]{poria2013enhanced}
Poria, S., A.~Gelbukh, A.~Hussain, N.~Howard, D.~Das, and S.~Bandyopadhyay
  (2013).
\newblock Enhanced senticnet with affective labels for concept-based opinion
  mining.
\newblock {\em IEEE Intelligent Systems\/}~{\em 28\/}(2), 31--38.

\bibitem[\protect\citeauthoryear{Rayner}{Rayner}{1985}]{rayner1985linear}
Rayner, J. M.~V. (1985).
\newblock Linear relations in biomechanics: the statistics of scaling
  functions.
\newblock {\em J. Zool. Lond. (A)\/}~{\em 206}, 415--439.

\bibitem[\protect\citeauthoryear{Ribeiro, Ara{\'u}jo, Gon{\c{c}}alves,
  Andr{\'e}~Gon{\c{c}}alves, and Benevenuto}{Ribeiro
  et~al.}{2016}]{ribeiro2016sentibench}
Ribeiro, F.~N., M.~Ara{\'u}jo, P.~Gon{\c{c}}alves,
  M.~Andr{\'e}~Gon{\c{c}}alves, and F.~Benevenuto (2016, July).
\newblock {SentiBench} --- a benchmark comparison of state-of-the-practice
  sentiment analysis methods.
\newblock {\em {EPJ} Data Sci.\/}~{\em 5\/}(1), 23.

\bibitem[\protect\citeauthoryear{Ruiz, Hristidis, Castillo, Gionis, and
  Jaimes}{Ruiz et~al.}{2012}]{ruiz2012correlating}
Ruiz, E.~J., V.~Hristidis, C.~Castillo, A.~Gionis, and A.~Jaimes (2012).
\newblock Correlating financial time series with micro-blogging activity.
\newblock In {\em Proceedings of the fifth ACM international conference on Web
  search and data mining}, pp.\  513--522. ACM.

\bibitem[\protect\citeauthoryear{Sandhaus}{Sandhaus}{2008}]{nytimescorpus2008new}
Sandhaus, E. (2008).
\newblock The {N}ew {Y}ork {T}imes {A}nnotated {C}orpus.
\newblock {L}inguistic Data Consortium, Philadelphia.

\bibitem[\protect\citeauthoryear{Si, Mukherjee, Liu, Li, Li, and Deng}{Si
  et~al.}{2013}]{si2013exploiting}
Si, J., A.~Mukherjee, B.~Liu, Q.~Li, H.~Li, and X.~Deng (2013).
\newblock Exploiting topic based {T}witter sentiment for stock prediction.
\newblock In {\em ACL (2)}, pp.\  24--29.

\bibitem[\protect\citeauthoryear{Socher, Perelygin, Wu, Chuang, Manning, Ng,
  and Potts}{Socher et~al.}{2013}]{socher2013a}
Socher, R., A.~Perelygin, J.~Y. Wu, J.~Chuang, C.~D. Manning, A.~Y. Ng, and
  C.~Potts (2013).
\newblock Recursive deep models for semantic compositionality over a sentiment
  treebank.
\newblock In {\em Proceedings of the conference on empirical methods in natural
  language processing (EMNLP)}, Volume 1631, pp.\  1642. Citeseer.

\bibitem[\protect\citeauthoryear{Stone, Dunphy, and Smith}{Stone
  et~al.}{1966}]{stone1966general}
Stone, P.~J., D.~C. Dunphy, and M.~S. Smith (1966).
\newblock The general inquirer: A computer approach to content analysis.
\newblock {\em MIT Press\/}.

\bibitem[\protect\citeauthoryear{Taboada, Brooke, Tofiloski, Voll, and
  Stede}{Taboada et~al.}{2011}]{taboada2011lexicon}
Taboada, M., J.~Brooke, M.~Tofiloski, K.~Voll, and M.~Stede (2011).
\newblock Lexicon-based methods for sentiment analysis.
\newblock {\em Computational linguistics\/}~{\em 37\/}(2), 267--307.

\bibitem[\protect\citeauthoryear{Thelwall, Buckley, Paltoglou, Cai, and
  Kappas}{Thelwall et~al.}{2010}]{thelwall2010sentiment}
Thelwall, M., K.~Buckley, G.~Paltoglou, D.~Cai, and A.~Kappas (2010).
\newblock Sentiment strength detection in short informal text.
\newblock {\em Journal of the American Society for Information Science and
  Technology\/}~{\em 61\/}(12), 2544--2558.

\bibitem[\protect\citeauthoryear{Warriner, Kuperman, and Brysbaert}{Warriner
  et~al.}{2013}]{warriner2013norms}
Warriner, A.~B., V.~Kuperman, and M.~Brysbaert (2013).
\newblock Norms of valence, arousal, and dominance for 13,915 english lemmas.
\newblock {\em Behavior research methods\/}~{\em 45\/}(4), 1191--1207.

\bibitem[\protect\citeauthoryear{Watson and Clark}{Watson and
  Clark}{1999}]{watson1999panas}
Watson, D. and L.~A. Clark (1999).
\newblock {\em The {PANAS-X}: Manual for the positive and negative affect
  schedule-expanded form: Manual for the positive and negative affect
  schedule-expanded form}.
\newblock Ph.\ D. thesis, University of Iowa.

\bibitem[\protect\citeauthoryear{Whissell, Fournier, Pelland, Weir, and
  Makarec}{Whissell et~al.}{1986}]{whissell1986dictionary}
Whissell, C., M.~Fournier, R.~Pelland, D.~Weir, and K.~Makarec (1986).
\newblock A dictionary of affect in language: Iv. reliability, validity, and
  applications.
\newblock {\em Perceptual and Motor Skills\/}~{\em 62\/}(3), 875--888.

\bibitem[\protect\citeauthoryear{Wilson, Wiebe, and Hoffmann}{Wilson
  et~al.}{2005}]{wilson2005recognizing}
Wilson, T., J.~Wiebe, and P.~Hoffmann (2005).
\newblock Recognizing contextual polarity in phrase-level sentiment analysis.
\newblock {\em Proceedings of Human Language Technologies Conference/Conference
  on Empirical Methods in Natural Language Processing (HLT/EMNLP 2005)\/}.

\bibitem[\protect\citeauthoryear{Wojcik, Hovasapian, Graham, Motyl, and
  Ditto}{Wojcik et~al.}{2015}]{wojcik2015conservatives}
Wojcik, S.~P., A.~Hovasapian, J.~Graham, M.~Motyl, and P.~H. Ditto (2015).
\newblock Conservatives report, but liberals display, greater happiness.
\newblock {\em Science\/}~{\em 347\/}(6227), 1243--1246.

\bibitem[\protect\citeauthoryear{Zhu, Kiritchenko, and Mohammad}{Zhu
  et~al.}{2014}]{zhu2014nrc}
Zhu, X., S.~Kiritchenko, and S.~M. Mohammad (2014).
\newblock Nrc-canada-2014: Recent improvements in the sentiment analysis of
  tweets.
\newblock In {\em Proceedings of the 8th international workshop on semantic
  evaluation (SemEval 2014)}, pp.\  443--447. Citeseer.

\end{references}


\begin{references}{}

\bibitem[\protect\citeauthoryear{Bex and Bench-Capon}{Bex and
  Bench-Capon}{2010}]{bex2010persuasive}
Bex, F.~J. and T.~J. Bench-Capon (2010).
\newblock Persuasive stories for multi-agent argumentation.
\newblock In {\em AAAI Fall Symposium: Computational Models of Narrative},
  Volume~10, pp.\ ~04.

\bibitem[\protect\citeauthoryear{Booker}{Booker}{2006}]{booker2006seven}
Booker, C. (2006).
\newblock {\em The Seven Basic Plots: Why We Tell Stories}.
\newblock New York: Bloomsbury Academic.

\bibitem[\protect\citeauthoryear{Bost, Labatut, and Linar{\`e}s}{Bost
  et~al.}{2016}]{bost2016narrative}
Bost, X., V.~Labatut, and G.~Linar{\`e}s (2016).
\newblock Narrative smoothing: dynamic conversational network for the analysis
  of tv series plots.

\bibitem[\protect\citeauthoryear{Campbell}{Campbell}{1949}]{campbell1949hero}
Campbell, J. (1949).
\newblock {\em The Hero with a Thousand Faces\/} (third ed.).
\newblock California: New World Library.

\bibitem[\protect\citeauthoryear{Cave}{Cave}{2013}]{cave2013stories}
Cave, S. (2013, Jul).
\newblock The 4 stories we tell ourselves about death.
\newblock
  \url{http://www.ted.com/talks/stephen_cave_the_4_stories_we_tell_ourselves_about_death}.

\bibitem[\protect\citeauthoryear{da~Silva and Tehrani}{da~Silva and
  Tehrani}{2016}]{silva2016comparative}
da~Silva, S.~G. and J.~J. Tehrani (2016).
\newblock Comparative phylogenetic analyses uncover the ancient roots of
  {I}ndo-{E}uropean folktales.
\newblock {\em Royal Society Open Science\/}~{\em 3\/}(1).

\bibitem[\protect\citeauthoryear{Dodds}{Dodds}{2013}]{dodds2013homo}
Dodds, P.~S. (2013).
\newblock Homo {N}arrativus and the trouble with fame.
\newblock Nautilus Magazine.
\newblock
  \href{http://nautil.us/issue/5/fame/homo-narrativus-and-the-trouble-with-fame}{http://nautil.us/issue/5/fame/homo-narrativus-and-the-trouble-with-fame}.

\bibitem[\protect\citeauthoryear{Dodds, Clark, Desu, Frank, Reagan, Williams,
  Mitchell, Harris, Kloumann, Bagrow, Megerdoomian, McMahon, Tivnan, and
  Danforth}{Dodds et~al.}{2015}]{dodds2015human}
Dodds, P.~S., E.~M. Clark, S.~Desu, M.~R. Frank, A.~J. Reagan, J.~R. Williams,
  L.~Mitchell, K.~D. Harris, I.~M. Kloumann, J.~P. Bagrow, K.~Megerdoomian,
  M.~T. McMahon, B.~F. Tivnan, and C.~M. Danforth (2015).
\newblock Human language reveals a universal positivity bias.
\newblock {\em PNAS\/}~{\em 112\/}(8), 2389--2394.

\bibitem[\protect\citeauthoryear{Dodds, Harris, Kloumann, Bliss, and
  Danforth}{Dodds et~al.}{2011}]{dodds2011temporal}
Dodds, P.~S., K.~D. Harris, I.~M. Kloumann, C.~A. Bliss, and C.~M. Danforth
  (2011, 12).
\newblock Temporal patterns of happiness and information in a global social
  network: Hedonometrics and {T}witter.
\newblock {\em PLoS ONE\/}~{\em 6\/}(12), e26752.

\bibitem[\protect\citeauthoryear{Dolby}{Dolby}{2008}]{dolby2008literary}
Dolby, S.~K. (2008).
\newblock {\em Literary {F}olkloristics and the Personal Narrative}.
\newblock Indiana: Trickster Press.

\bibitem[\protect\citeauthoryear{Dundes}{Dundes}{1997}]{dundes1997motif}
Dundes, A. (1997).
\newblock The motif-index and the tale type index: A critique.
\newblock {\em Journal of Folklore Research\/}, 195--202.

\bibitem[\protect\citeauthoryear{Gleick}{Gleick}{2011}]{gleick2011information}
Gleick, J. (2011).
\newblock {\em The Information: A History, A Theory, A Flood}.
\newblock New York: Pantheon.

\bibitem[\protect\citeauthoryear{Gottschall}{Gottschall}{2013}]{gottschall2013storytelling}
Gottschall, J. (2013).
\newblock {\em The Storytelling Animal: How Stories Make Us Human}.
\newblock New York, NY: Mariner Books.

\bibitem[\protect\citeauthoryear{Jockers}{Jockers}{2014}]{jockers2014novel}
Jockers, M. (2014, Jun).
\newblock A novel method for detecting plot.
\newblock
  \url{http://www.matthewjockers.net/2014/06/05/a-novel-method-for-detecting-plot/}.

\bibitem[\protect\citeauthoryear{Jockers}{Jockers}{2015}]{jockers2015rest}
Jockers, M. (2015, Feb).
\newblock The rest of the story.
\newblock
  \url{http://www.matthewjockers.net/2015/02/25/the-rest-of-the-story/}.

\bibitem[\protect\citeauthoryear{Kirschenbaum}{Kirschenbaum}{2007}]{kirschenbaum2007remaking}
Kirschenbaum, M.~G. (2007).
\newblock The remaking of reading: Data mining and the digital humanities.
\newblock In {\em The National Science Foundation Symposium on Next Generation
  of Data Mining and Cyber-Enabled Discovery for Innovation, Maryland}.

\bibitem[\protect\citeauthoryear{Kohonen}{Kohonen}{1990}]{kohonen1990self}
Kohonen, T. (1990).
\newblock The self-organizing map.
\newblock {\em Proceedings of the IEEE\/}~{\em 78\/}(9), 1464--1480.

\bibitem[\protect\citeauthoryear{Li, Lee-Urban, Johnston, and Riedl}{Li
  et~al.}{2013}]{li2013story}
Li, B., S.~Lee-Urban, G.~Johnston, and M.~Riedl (2013).
\newblock Story generation with crowdsourced plot graphs.
\newblock In {\em AAAI}.

\bibitem[\protect\citeauthoryear{MacDonald}{MacDonald}{1982}]{macdonal1982storytellers}
MacDonald, M.~R. (1982).
\newblock {\em Storytellers Sourcebook: A Subject, Title, and Motif Index to
  Folklore Collections for Children}.
\newblock Michigan: Gale Group.

\bibitem[\protect\citeauthoryear{Min and Park}{Min and
  Park}{2016}]{min2016narrative}
Min, S. and J.~Park (2016).
\newblock Narrative as a complex network: A study of {V}ictor {H}ugo's les
  mis{\'e}rables.
\newblock In {\em Proceedings of HCI Korea}.

\bibitem[\protect\citeauthoryear{Moretti}{Moretti}{2013}]{moretti2013distant}
Moretti, F. (2013).
\newblock {\em Distant Reading}.
\newblock New York: Verso.

\bibitem[\protect\citeauthoryear{Munroe}{Munroe}{2009}]{munroe657}
Munroe, R. (2009, 11).
\newblock Movie narrative charts.
\newblock http://xkcd.com/657/.

\bibitem[\protect\citeauthoryear{Nenkova and McKeown}{Nenkova and
  McKeown}{2012}]{nenkova2012survey}
Nenkova, A. and K.~McKeown (2012).
\newblock A survey of text summarization techniques.
\newblock In {\em Mining text data}, pp.\  43--76. Berlin, Germany: Springer.

\bibitem[\protect\citeauthoryear{Nickerson}{Nickerson}{1998}]{nickerson1998confirmation}
Nickerson, R.~S. (1998).
\newblock Confirmation {B}ias; {A} ubiquitous phenomenon in many guises.
\newblock {\em Review of General Psychology\/}~{\em 2}, 175--220.

\bibitem[\protect\citeauthoryear{Prado, Dahmen, Bazzan, Carron, and
  Kenna}{Prado et~al.}{2016}]{prado2016temporal}
Prado, S.~D., S.~R. Dahmen, A.~L.~C. Bazzan, P.~M. Carron, and R.~Kenna (2016).
\newblock Temporal network analysis of literary texts.

\bibitem[\protect\citeauthoryear{Pratchett, Stewart, and Cohen}{Pratchett
  et~al.}{2003}]{pratchett2003science}
Pratchett, T., I.~Stewart, and J.~Cohen (2003).
\newblock {\em The Science of Discworld II: The Globe}.
\newblock London, UK: Ebury Press.

\bibitem[\protect\citeauthoryear{Propp}{Propp}{1968}]{propp1968morphology}
Propp, V. (1968).
\newblock {\em Morphology of the Folktale. 1928}.
\newblock Texas: Texas University Press.

\bibitem[\protect\citeauthoryear{Reagan, Tivnan, Williams, Danforth, and
  Dodds}{Reagan et~al.}{2015}]{reagan2016benchmarking}
Reagan, A., B.~Tivnan, J.~R. Williams, C.~M. Danforth, and P.~S. Dodds (2015).
\newblock Benchmarking sentiment analysis methods for large-scale texts: A case
  for using continuum-scored words and word shift graphs.
\newblock Preprint available at
  \href{https://arxiv.org/abs/1512.00531}{https://arxiv.org/abs/1512.00531}.

\bibitem[\protect\citeauthoryear{Ribeiro, Ara{\'u}jo, Gon{\c{c}}alves,
  Andr{\'e}~Gon{\c{c}}alves, and Benevenuto}{Ribeiro
  et~al.}{2016}]{ribeiro2016sentibench}
Ribeiro, F.~N., M.~Ara{\'u}jo, P.~Gon{\c{c}}alves,
  M.~Andr{\'e}~Gon{\c{c}}alves, and F.~Benevenuto (2016, July).
\newblock {SentiBench} --- a benchmark comparison of state-of-the-practice
  sentiment analysis methods.
\newblock {\em {EPJ} Data Sci.\/}~{\em 5\/}(1), 23.

\bibitem[\protect\citeauthoryear{Riedl and Harrison}{Riedl and
  Harrison}{2015}]{riedl2015using}
Riedl, M.~O. and B.~Harrison (2015).
\newblock Using stories to teach human values to artificial agents.

\bibitem[\protect\citeauthoryear{Rousseeuw}{Rousseeuw}{1987}]{rousseeuw1987silhouettes}
Rousseeuw, P.~J. (1987).
\newblock Silhouettes: a graphical aid to the interpretation and validation of
  cluster analysis.
\newblock {\em Journal of computational and applied mathematics\/}~{\em 20},
  53--65.

\bibitem[\protect\citeauthoryear{Tenenbaum, Barrett, Medaris, and
  Devitt}{Tenenbaum et~al.}{2015}]{tenenbaum2015languages}
Tenenbaum, D.~J., K.~Barrett, S.~Medaris, and T.~Devitt (2015, February).
\newblock In 10 languages, happy words beat sad ones.
\newblock http://whyfiles.org/2015/in-10-languages-happy-words-beat-sad-ones/.

\bibitem[\protect\citeauthoryear{Uther}{Uther}{2011}]{uther2011types}
Uther, H.-J. (2011).
\newblock {\em The Types of International Folktales. A Classification and
  Bibliography. Based on the System of Antti Aarne and Stith Thompson. Part I.
  Animal Tales, Tales of Magic, Religious Tales, and Realistic Tales, with an
  Introduction (FF Communications, 284)}.
\newblock Helsinki, Finland: Finnish Academy of Science and Letters.

\bibitem[\protect\citeauthoryear{Various}{Various}{}]{gutenberg}
Various.
\newblock Project {G}utenberg.

\bibitem[\protect\citeauthoryear{Vonnegut}{Vonnegut}{1981}]{vonnegut1981palm}
Vonnegut, K. (1981).
\newblock {\em Palm Sunday}.
\newblock New York: RosettaBooks LLC.

\bibitem[\protect\citeauthoryear{Vonnegut}{Vonnegut}{1995}]{vonnegut1995shapes}
Vonnegut, K. (1995).
\newblock Shapes of stories.
\newblock https://www.youtube.com/watch?v=oP3c1h8v2ZQ.

\bibitem[\protect\citeauthoryear{Ward~Jr}{Ward~Jr}{1963}]{ward1963hierarchical}
Ward~Jr, J.~H. (1963).
\newblock Hierarchical grouping to optimize an objective function.
\newblock {\em Journal of the American statistical association\/}~{\em
  58\/}(301), 236--244.

\bibitem[\protect\citeauthoryear{Winston}{Winston}{2011}]{winston2011strong}
Winston, P.~H. (2011).
\newblock The strong story hypothesis and the directed perception hypothesis.

\end{references}


\begin{thebibliography}{}

\bibitem[\protect\citeauthoryear{Abbott}{Abbott}{2008}]{abbott2008cambridge}
Abbott, H.~P. (2008).
\newblock {\em The Cambridge introduction to narrative}.
\newblock Massachusetts: Cambridge University Press.

\bibitem[\protect\citeauthoryear{Abney}{Abney}{1997}]{abney1997part}
Abney, S. (1997).
\newblock Part-of-speech tagging and partial parsing.
\newblock In {\em Corpus-based methods in language and speech processing}, pp.\
   118--136. Springer.

\bibitem[\protect\citeauthoryear{Agarwal, Xie, Vovsha, Rambow, and
  Passonneau}{Agarwal et~al.}{2011}]{agarwal2011sentiment}
Agarwal, A., B.~Xie, I.~Vovsha, O.~Rambow, and R.~Passonneau (2011).
\newblock Sentiment analysis of twitter data.
\newblock In {\em Proceedings of the workshop on languages in social media},
  pp.\  30--38. Association for Computational Linguistics.

\bibitem[\protect\citeauthoryear{Alajajian, Williams, Reagan, Alajajian, Frank,
  Mitchell, Lahne, Danforth, and Dodds}{Alajajian
  et~al.}{2016}]{alajajian2015lexicocalorimeter}
Alajajian, S.~E., J.~R. Williams, A.~J. Reagan, S.~C. Alajajian, M.~R. Frank,
  L.~Mitchell, J.~Lahne, C.~M. Danforth, and P.~S. Dodds (2016).
\newblock The {L}exicocalorimeter: {G}auging public health through caloric
  input and output on social media.
\newblock Available at
  \href{http://arxiv.org/abs/1507.05098}{http://arxiv.org/abs/1507.05098}.

\bibitem[\protect\citeauthoryear{Alexander, Dallachy, Piao, Baron, and
  Rayson}{Alexander et~al.}{2015}]{alexander2015metaphor}
Alexander, M., F.~Dallachy, S.~Piao, A.~Baron, and P.~Rayson (2015).
\newblock Metaphor, popular science, and semantic tagging: Distant reading with
  the historical thesaurus of english.
\newblock {\em Digital Scholarship in the Humanities\/}~{\em 30\/}(suppl 1),
  i16--i27.

\bibitem[\protect\citeauthoryear{Amendola, Marra, and Quartin}{Amendola
  et~al.}{2015}]{amendola2015evolving}
Amendola, L., V.~Marra, and M.~Quartin (2015, Jul).
\newblock The evolving perception of controversial movies.
\newblock {\em Palgrave Communication\/}~(1), 15038.

\bibitem[\protect\citeauthoryear{Amir, Astudillo, Ling, Carvalho, and
  Silva}{Amir et~al.}{2016}]{amir2016expanding}
Amir, S., R.~Astudillo, W.~Ling, P.~C. Carvalho, and M.~J. Silva (2016).
\newblock Expanding subjective lexicons for social media mining with embedding
  subspaces.
\newblock {\em arXiv preprint arXiv:1701.00145\/}.

\bibitem[\protect\citeauthoryear{Anderson}{Anderson}{1972}]{anderson1972a}
Anderson, P.~W. (1972).
\newblock More is different.
\newblock {\em Science\/}~{\em 177\/}(4047), 393--396.

\bibitem[\protect\citeauthoryear{Andor, Alberti, Weiss, Severyn, Presta,
  Ganchev, Petrov, and Collins}{Andor et~al.}{2016}]{andor2016globally}
Andor, D., C.~Alberti, D.~Weiss, A.~Severyn, A.~Presta, K.~Ganchev, S.~Petrov,
  and M.~Collins (2016).
\newblock Globally normalized transition-based neural networks.
\newblock {\em arXiv preprint arXiv:1603.06042\/}.

\bibitem[\protect\citeauthoryear{Awad}{Awad}{2013}]{awad2013culturally}
Awad, H. (2013).
\newblock {\em Culturally based story understanding}.
\newblock Ph.\ D. thesis, Citeseer.

\bibitem[\protect\citeauthoryear{Baccianella, Esuli, and
  Sebastiani}{Baccianella et~al.}{2010}]{baccianella2010sentiwordnet}
Baccianella, S., A.~Esuli, and F.~Sebastiani (2010).
\newblock Senti{W}ord{N}et 3.0: An enhanced lexical resource for sentiment
  analysis and opinion mining.
\newblock In {\em LREC}, Volume~10, pp.\  2200--2204.

\bibitem[\protect\citeauthoryear{Bagrow, Desu, Frank, Manukyan, Mitchell,
  Reagan, Bloedorn, Booker, Smith, Tivnan, Danforth, Dodds, and Bongard}{Bagrow
  et~al.}{2014}]{bagrow2014shadow}
Bagrow, J., S.~Desu, M.~R. Frank, N.~Manukyan, L.~Mitchell, A.~Reagan,
  E.~Bloedorn, L.~K. Booker, L. B.~Branting, M.~J. Smith, B.~F. Tivnan, C.~M.
  Danforth, P.~S. Dodds, and J.~C. Bongard (2014).
\newblock Shadow networks: {D}iscovering hidden nodes with models of
  information flow.
\newblock Preprint available at
  \href{http://arxiv.org/abs/1312.6122}{http://arxiv.org/abs/1312.6122}.

\bibitem[\protect\citeauthoryear{Baker, Fillmore, and Lowe}{Baker
  et~al.}{1998}]{baker1998berkeley}
Baker, C.~F., C.~J. Fillmore, and J.~B. Lowe (1998).
\newblock The berkeley framenet project.
\newblock In {\em Proceedings of the 36th Annual Meeting of the Association for
  Computational Linguistics and 17th International Conference on Computational
  Linguistics-Volume 1}, pp.\  86--90. Association for Computational
  Linguistics.

\bibitem[\protect\citeauthoryear{Bamman}{Bamman}{2015}]{bamman2015validity}
Bamman, D. (2015, Apr).
\newblock Validity.
\newblock
  \href{http://www.davidbamman.com/?p=52}{http://www.davidbamman.com/?p=52}.

\bibitem[\protect\citeauthoryear{Bamman, O'Connor, and Smith}{Bamman
  et~al.}{2014}]{bamman2014learning}
Bamman, D., B.~O'Connor, and N.~A. Smith (2014).
\newblock Learning latent personas of film characters.
\newblock In {\em Proceedings of the Annual Meeting of the Association for
  Computational Linguistics (ACL)}, pp.\  352.

\bibitem[\protect\citeauthoryear{Bamman, Underwood, and Smith}{Bamman
  et~al.}{2014}]{bamman2014bayesian}
Bamman, D., T.~Underwood, and N.~A. Smith (2014).
\newblock A bayesian mixed effects model of literary character.
\newblock In {\em ACL (1)}, pp.\  370--379.

\bibitem[\protect\citeauthoryear{Bar-Haim, Dinur, Feldman, Fresko, and
  Goldstein}{Bar-Haim et~al.}{2011}]{bar2011identifying}
Bar-Haim, R., E.~Dinur, R.~Feldman, M.~Fresko, and G.~Goldstein (2011).
\newblock Identifying and following expert investors in stock microblogs.
\newblock In {\em Proceedings of the Conference on Empirical Methods in Natural
  Language Processing}, pp.\  1310--1319. Association for Computational
  Linguistics.

\bibitem[\protect\citeauthoryear{Bestgen et~al.}{Bestgen
  et~al.}{2008}]{bestgen2008building}
Bestgen, Y. et~al. (2008).
\newblock Building affective lexicons from specific corpora for automatic
  sentiment analysis.
\newblock In {\em LREC}. Citeseer.

\bibitem[\protect\citeauthoryear{Bestgen and Vincze}{Bestgen and
  Vincze}{2012}]{bestgen2012checking}
Bestgen, Y. and N.~Vincze (2012).
\newblock Checking and bootstrapping lexical norms by means of word similarity
  indexes.
\newblock {\em Behavior research methods\/}~{\em 44\/}(4), 998--1006.

\bibitem[\protect\citeauthoryear{Bex}{Bex}{2013}]{bex2013values}
Bex, F. (2013).
\newblock Values as the point of a story.

\bibitem[\protect\citeauthoryear{Bex and Bench-Capon}{Bex and
  Bench-Capon}{2010}]{bex2010persuasive}
Bex, F.~J. and T.~J. Bench-Capon (2010).
\newblock Persuasive stories for multi-agent argumentation.
\newblock In {\em AAAI Fall Symposium: Computational Models of Narrative},
  Volume~10, pp.\ ~04.

\bibitem[\protect\citeauthoryear{Bird}{Bird}{2006}]{bird2006nltk}
Bird, S. (2006).
\newblock Nltk: the natural language toolkit.
\newblock In {\em Proceedings of the COLING/ACL on Interactive presentation
  sessions}, pp.\  69--72. Association for Computational Linguistics.

\bibitem[\protect\citeauthoryear{Blair-Goldensohn, Hannan, McDonald, Neylon,
  Reis, and Reynar}{Blair-Goldensohn et~al.}{2008}]{blair2008building}
Blair-Goldensohn, S., K.~Hannan, R.~McDonald, T.~Neylon, G.~A. Reis, and
  J.~Reynar (2008).
\newblock Building a sentiment summarizer for local service reviews.
\newblock In {\em WWW workshop on NLP in the information explosion era},
  Volume~14, pp.\  339--348.

\bibitem[\protect\citeauthoryear{Bollen, Mao, and Zeng}{Bollen
  et~al.}{2011}]{bollen2011twitter}
Bollen, J., H.~Mao, and X.~Zeng (2011).
\newblock Twitter mood predicts the stock market.
\newblock {\em Journal of Computational Science\/}~{\em 2\/}(1), 1--8.

\bibitem[\protect\citeauthoryear{Booker}{Booker}{2006}]{booker2006seven}
Booker, C. (2006).
\newblock {\em The Seven Basic Plots: Why We Tell Stories}.
\newblock New York: Bloomsbury Academic.

\bibitem[\protect\citeauthoryear{Bost, Labatut, and Linar{\`e}s}{Bost
  et~al.}{2016}]{bost2016narrative}
Bost, X., V.~Labatut, and G.~Linar{\`e}s (2016).
\newblock Narrative smoothing: dynamic conversational network for the analysis
  of tv series plots.

\bibitem[\protect\citeauthoryear{Bostock, Ogievetsky, and Heer}{Bostock
  et~al.}{2011}]{bostock2011d3}
Bostock, M., V.~Ogievetsky, and J.~Heer (2011).
\newblock D3: Data-driven documents.
\newblock {\em IEEE Trans. Visualization \& Comp. Graphics (Proc. InfoVis)\/}.

\bibitem[\protect\citeauthoryear{Bradley and Lang}{Bradley and
  Lang}{1999}]{bradley1999affective}
Bradley, M.~M. and P.~J. Lang (1999).
\newblock Affective norms for english words ({ANEW}): Stimuli, instruction
  manual and affective ratings.
\newblock Technical report c-1, University of Florida, Gainesville, FL.

\bibitem[\protect\citeauthoryear{Brewer and Lichtenstein}{Brewer and
  Lichtenstein}{1980}]{brewer1980event}
Brewer, W.~F. and E.~H. Lichtenstein (1980).
\newblock Event schemas, story schemas, and story grammars.
\newblock {\em Center for the Study of Reading Technical Report; no. 197\/}.

\bibitem[\protect\citeauthoryear{Cambria, Olsher, and Rajagopal}{Cambria
  et~al.}{2014}]{cambria2014senticnet}
Cambria, E., D.~Olsher, and D.~Rajagopal (2014).
\newblock Senticnet 3: a common and common-sense knowledge base for
  cognition-driven sentiment analysis.
\newblock In {\em Proceedings of the twenty-eighth AAAI conference on
  artificial intelligence}, pp.\  1515--1521. AAAI Press.

\bibitem[\protect\citeauthoryear{Cambria and White}{Cambria and
  White}{2014}]{cambria2014jumping}
Cambria, E. and B.~White (2014).
\newblock Jumping nlp curves: a review of natural language processing research
  [review article].
\newblock {\em IEEE Computational Intelligence Magazine\/}~{\em 9\/}(2),
  48--57.

\bibitem[\protect\citeauthoryear{Campbell}{Campbell}{1949}]{campbell1949hero}
Campbell, J. (1949).
\newblock {\em The Hero with a Thousand Faces\/} (third ed.).
\newblock California: New World Library.

\bibitem[\protect\citeauthoryear{Campbell and Moyers}{Campbell and
  Moyers}{1991}]{campbell1991a}
Campbell, J. and B.~Moyers (1991).
\newblock {\em The Power of Myth}.
\newblock Anchor.

\bibitem[\protect\citeauthoryear{Cao and Cui}{Cao and
  Cui}{2016}]{cao2016introduction}
Cao, N. and W.~Cui (2016).
\newblock Introduction to text visualization.
\newblock {\em Atlantis briefs in artificial intelligence (\/}~{\em 1}.

\bibitem[\protect\citeauthoryear{Card, Boydstun, Gross, Resnik, and Smith}{Card
  et~al.}{2015}]{card2015media}
Card, D., A.~E. Boydstun, J.~H. Gross, P.~Resnik, and N.~A. Smith (2015).
\newblock The media frames corpus: Annotations of frames across issues.
\newblock In {\em Proceedings of the 53rd Annual Meeting of the Association for
  Computational Linguistics and the 7th International Joint Conference on
  Natural Language Processing}, Volume~2, pp.\  438--444.

\bibitem[\protect\citeauthoryear{Cave}{Cave}{2013}]{cave2013stories}
Cave, S. (2013, Jul).
\newblock The 4 stories we tell ourselves about death.
\newblock
  \url{http://www.ted.com/talks/stephen_cave_the_4_stories_we_tell_ourselves_about_death}.

\bibitem[\protect\citeauthoryear{Chambers}{Chambers}{2013}]{chambers2013event}
Chambers, N. (2013).
\newblock Event schema induction with a probabilistic entity-driven model.
\newblock In {\em EMNLP}, Volume~13, pp.\  1797--1807.

\bibitem[\protect\citeauthoryear{Chambers and Jurafsky}{Chambers and
  Jurafsky}{2008}]{chambers2008unsupervised}
Chambers, N. and D.~Jurafsky (2008).
\newblock Unsupervised learning of narrative event chains.
\newblock In {\em ACL}, Volume 94305, pp.\  789--797. Citeseer.

\bibitem[\protect\citeauthoryear{Chambers and Jurafsky}{Chambers and
  Jurafsky}{2009}]{chambers2009unsupervised}
Chambers, N. and D.~Jurafsky (2009).
\newblock Unsupervised learning of narrative schemas and their participants.
\newblock In {\em Proceedings of the Joint Conference of the 47th Annual
  Meeting of the ACL and the 4th International Joint Conference on Natural
  Language Processing of the AFNLP: Volume 2-Volume 2}, pp.\  602--610.
  Association for Computational Linguistics.

\bibitem[\protect\citeauthoryear{Chambers and Jurafsky}{Chambers and
  Jurafsky}{2010}]{chambers2010database}
Chambers, N. and D.~Jurafsky (2010).
\newblock A database of narrative schemas.
\newblock In {\em LREC}.

\bibitem[\protect\citeauthoryear{Chambers, Wang, and Jurafsky}{Chambers
  et~al.}{2007}]{chambers2007classifying}
Chambers, N., S.~Wang, and D.~Jurafsky (2007).
\newblock Classifying temporal relations between events.
\newblock In {\em Proceedings of the 45th Annual Meeting of the ACL on
  Interactive Poster and Demonstration Sessions}, pp.\  173--176. Association
  for Computational Linguistics.

\bibitem[\protect\citeauthoryear{Chen and Manning}{Chen and
  Manning}{2014}]{chen2014fast}
Chen, D. and C.~D. Manning (2014).
\newblock A fast and accurate dependency parser using neural networks.
\newblock In {\em EMNLP}, pp.\  740--750.

\bibitem[\protect\citeauthoryear{Cherny}{Cherny}{2016}]{cherny2013bones}
Cherny, L. (2016, Jun).
\newblock The bones of a bestseller: Visualizing fiction.
\newblock
  \href{http://blogger.ghostweather.com/2013/06/analysis-of-fiction-my-openvisconf-talk.html}{http://blogger.ghostweather.com/2013/06/analysis-of-fiction-my-openvisconf-talk.html}.

\bibitem[\protect\citeauthoryear{Cheung, Poon, and Vanderwende}{Cheung
  et~al.}{2013}]{cheung2013probabilistic}
Cheung, J. C.~K., H.~Poon, and L.~Vanderwende (2013).
\newblock Probabilistic frame induction.
\newblock {\em arXiv preprint arXiv:1302.4813\/}.

\bibitem[\protect\citeauthoryear{Chuang, Manning, and Heer}{Chuang
  et~al.}{2012}]{chuang2012termite}
Chuang, J., C.~D. Manning, and J.~Heer (2012).
\newblock Termite: Visualization techniques for assessing textual topic models.
\newblock In {\em Proceedings of the International Working Conference on
  Advanced Visual Interfaces}, pp.\  74--77. ACM.

\bibitem[\protect\citeauthoryear{Chuang, Ramage, Manning, and Heer}{Chuang
  et~al.}{2012}]{chuang2012interpretation}
Chuang, J., D.~Ramage, C.~Manning, and J.~Heer (2012).
\newblock Interpretation and trust: Designing model-driven visualizations for
  text analysis.
\newblock In {\em Proceedings of the SIGCHI Conference on Human Factors in
  Computing Systems}, pp.\  443--452. ACM.

\bibitem[\protect\citeauthoryear{Chung and Liu}{Chung and
  Liu}{2011}]{chung2011predicting}
Chung, S. and S.~Liu (2011).
\newblock Predicting stock market fluctuations from {T}witter.
\newblock {\em Berkeley, California\/}.

\bibitem[\protect\citeauthoryear{Church}{Church}{1988}]{church1988stochastic}
Church, K.~W. (1988).
\newblock A stochastic parts program and noun phrase parser for unrestricted
  text.
\newblock In {\em Proceedings of the second conference on Applied natural
  language processing}, pp.\  136--143. Association for Computational
  Linguistics.

\bibitem[\protect\citeauthoryear{Clancy}{Clancy}{2015}]{clancy2015fabula}
Clancy, E. (2015).
\newblock A fabula of syuzhet: A contretemps of digital humanities and
  sentiment analysis.
\newblock
  \href{https://storify.com/clancynewyork/contretemps-a-syuzhet}{https://storify.com/clancynewyork/contretemps-a-syuzhet}
  and
  \href{https://storify.com/clancynewyork/a-fabula-of-syuzhet-ii}{https://storify.com/clancynewyork/a-fabula-of-syuzhet-ii}.

\bibitem[\protect\citeauthoryear{Clauset, Shalizi, and Newman}{Clauset
  et~al.}{2009}]{clauset2009b}
Clauset, A., C.~R. Shalizi, and M.~E.~J. Newman (2009).
\newblock Power-law distributions in empirical data.
\newblock {\em SIAM Review\/}~{\em 51}, 661--703.

\bibitem[\protect\citeauthoryear{Cobley}{Cobley}{2005}]{cobley2005narratology}
Cobley, P. (2005).
\newblock Narratology.
\newblock {\em The Johns Hopkins Guide to Literary Theory and Criticism, 2nd
  ed. John Hopkins University Press, London\/}.

\bibitem[\protect\citeauthoryear{Cody, Reagan, Mitchell, Dodds, and
  Danforth}{Cody et~al.}{2015}]{cody2015climate}
Cody, E.~M., A.~J. Reagan, L.~Mitchell, P.~S. Dodds, and C.~M. Danforth (2015).
\newblock Climate change sentiment on twitter: An unsolicited public opinion
  poll.
\newblock {\em PLOS ONE\/}.

\bibitem[\protect\citeauthoryear{Collier}{Collier}{2011}]{collier2011understanding}
Collier, D. (2011).
\newblock Understanding process tracing.
\newblock {\em PS: Political Science \& Politics\/}~{\em 44\/}(04), 823--830.

\bibitem[\protect\citeauthoryear{da~Silva and Tehrani}{da~Silva and
  Tehrani}{2016}]{silva2016comparative}
da~Silva, S.~G. and J.~J. Tehrani (2016).
\newblock Comparative phylogenetic analyses uncover the ancient roots of
  {I}ndo-{E}uropean folktales.
\newblock {\em Royal Society Open Science\/}~{\em 3\/}(1).

\bibitem[\protect\citeauthoryear{DARPA}{DARPA}{2011}]{darpa2011narrative}
DARPA (2011, 12).
\newblock Broad agency announcement: Narrative networks.
\newblock available at \url{http://www.fedbizopps.gov/}, accessed June 20,
  2016.

\bibitem[\protect\citeauthoryear{Das and Chen}{Das and
  Chen}{2007}]{das2007yahoo}
Das, S.~R. and M.~Y. Chen (2007).
\newblock Yahoo! for amazon: Sentiment extraction from small talk on the web.
\newblock {\em Management Science\/}~{\em 53\/}(9), 1375--1388.

\bibitem[\protect\citeauthoryear{De~Smedt and Daelemans}{De~Smedt and
  Daelemans}{2012}]{de2012pattern}
De~Smedt, T. and W.~Daelemans (2012).
\newblock Pattern for {P}ython.
\newblock {\em The Journal of Machine Learning Research\/}~{\em 13\/}(1),
  2063--2067.

\bibitem[\protect\citeauthoryear{Decadt, Hoste, Daelemans, and Van~den
  Bosch}{Decadt et~al.}{2004}]{decadt2004gambl}
Decadt, B., V.~Hoste, W.~Daelemans, and A.~Van~den Bosch (2004).
\newblock Gambl, genetic algorithm optimization of memory-based wsd.
\newblock In {\em Senseval-3: Third International Workshop on the Evaluation of
  Systems for the Semantic Analysis of Text}, pp.\  108--112. Association for
  Computational Linguistics.

\bibitem[\protect\citeauthoryear{DeGraff and Harmon}{DeGraff and
  Harmon}{2015}]{degraff2015plotted}
DeGraff, A. and D.~Harmon (2015).
\newblock {\em Plotted: A Literary Atlas}.
\newblock Houghton Mifflin Harcourt.

\bibitem[\protect\citeauthoryear{DeRose}{DeRose}{1988}]{derose1988grammatical}
DeRose, S.~J. (1988).
\newblock Grammatical category disambiguation by statistical optimization.
\newblock {\em Computational linguistics\/}~{\em 14\/}(1), 31--39.

\bibitem[\protect\citeauthoryear{Do, Chan, and Roth}{Do
  et~al.}{2011}]{do2011minimally}
Do, Q.~X., Y.~S. Chan, and D.~Roth (2011).
\newblock Minimally supervised event causality identification.
\newblock In {\em Proceedings of the Conference on Empirical Methods in Natural
  Language Processing}, EMNLP '11, Stroudsburg, PA, USA, pp.\  294--303.
  Association for Computational Linguistics.

\bibitem[\protect\citeauthoryear{Dodds}{Dodds}{2013}]{dodds2013homo}
Dodds, P.~S. (2013).
\newblock Homo {N}arrativus and the trouble with fame.
\newblock Nautilus Magazine.
\newblock
  \href{http://nautil.us/issue/5/fame/homo-narrativus-and-the-trouble-with-fame}{http://nautil.us/issue/5/fame/homo-narrativus-and-the-trouble-with-fame}.

\bibitem[\protect\citeauthoryear{Dodds, Clark, Desu, Frank, Reagan, Williams,
  Mitchell, Harris, Kloumann, Bagrow, Megerdoomian, McMahon, Tivnan, and
  Danforth}{Dodds et~al.}{2015a}]{dodds2015human}
Dodds, P.~S., E.~M. Clark, S.~Desu, M.~R. Frank, A.~J. Reagan, J.~R. Williams,
  L.~Mitchell, K.~D. Harris, I.~M. Kloumann, J.~P. Bagrow, K.~Megerdoomian,
  M.~T. McMahon, B.~F. Tivnan, and C.~M. Danforth (2015a).
\newblock Human language reveals a universal positivity bias.
\newblock {\em PNAS\/}~{\em 112\/}(8), 2389--2394.

\bibitem[\protect\citeauthoryear{Dodds, Clark, Desu, Frank, Reagan, Williams,
  Mitchell, Harris, Kloumann, Bagrow, Megerdoomian, McMahon, Tivnan, and
  Danforth}{Dodds et~al.}{2015b}]{dodds2015reply}
Dodds, P.~S., E.~M. Clark, S.~Desu, M.~R. Frank, A.~J. Reagan, J.~R. Williams,
  L.~Mitchell, K.~D. Harris, I.~M. Kloumann, J.~P. Bagrow, K.~Megerdoomian,
  M.~T. McMahon, B.~F. Tivnan, and C.~M. Danforth (2015b).
\newblock Reply to garcia et al.: Common mistakes in measuring
  frequency-dependent word characteristics.
\newblock {\em Proceedings of the National Academy of Sciences\/}~{\em
  112\/}(23), E2984--E2985.

\bibitem[\protect\citeauthoryear{Dodds and Danforth}{Dodds and
  Danforth}{2009}]{dodds2009b}
Dodds, P.~S. and C.~M. Danforth (2009, July).
\newblock Measuring the happiness of large-scale written expression: Songs,
  blogs, and presidents.
\newblock {\em Journal of Happiness Studies\/}~{\em 11\/}(4), 441--456.

\bibitem[\protect\citeauthoryear{Dodds, Harris, Kloumann, Bliss, and
  Danforth}{Dodds et~al.}{2011}]{dodds2011temporal}
Dodds, P.~S., K.~D. Harris, I.~M. Kloumann, C.~A. Bliss, and C.~M. Danforth
  (2011, 12).
\newblock Temporal patterns of happiness and information in a global social
  network: Hedonometrics and {T}witter.
\newblock {\em PLoS ONE\/}~{\em 6\/}(12), e26752.

\bibitem[\protect\citeauthoryear{Dodds, Mitchell, Reagan, and Danforth}{Dodds
  et~al.}{2016}]{dodds2016tracking}
Dodds, P.~S., L.~Mitchell, A.~J. Reagan, and C.~M. Danforth (2016, may).
\newblock Tracking climate change through the spatiotemporal dynamics of the
  teletherms, the statistically hottest and coldest days of the year.
\newblock {\em {PLOS} {ONE}\/}~{\em 11\/}(5), e0154184.

\bibitem[\protect\citeauthoryear{Dolby}{Dolby}{2008}]{dolby2008literary}
Dolby, S.~K. (2008).
\newblock {\em Literary {F}olkloristics and the Personal Narrative}.
\newblock Indiana: Trickster Press.

\bibitem[\protect\citeauthoryear{Dundes}{Dundes}{1997}]{dundes1997motif}
Dundes, A. (1997).
\newblock The motif-index and the tale type index: A critique.
\newblock {\em Journal of Folklore Research\/}, 195--202.

\bibitem[\protect\citeauthoryear{Ekman}{Ekman}{1992}]{ekman1992argument}
Ekman, P. (1992).
\newblock An argument for basic emotions.
\newblock {\em Cognition \& emotion\/}~{\em 6\/}(3-4), 169--200.

\bibitem[\protect\citeauthoryear{Elson}{Elson}{2012a}]{elson2012detecting}
Elson, D.~K. (2012a).
\newblock Detecting story analogies from annotations of time, action and
  agency.
\newblock In {\em Proceedings of the LREC 2012 Workshop on Computational Models
  of Narrative, Istanbul, Turkey}.

\bibitem[\protect\citeauthoryear{Elson}{Elson}{2012b}]{elson2012modeling}
Elson, D.~K. (2012b).
\newblock {\em Modeling narrative discourse}.
\newblock Ph.\ D. thesis, Citeseer.

\bibitem[\protect\citeauthoryear{Elson, Dames, and McKeown}{Elson
  et~al.}{2010}]{elson2010extracting}
Elson, D.~K., N.~Dames, and K.~R. McKeown (2010).
\newblock Extracting social networks from literary fiction.
\newblock In {\em Proceedings of the 48th annual meeting of the association for
  computational linguistics}, pp.\  138--147. Association for Computational
  Linguistics.

\bibitem[\protect\citeauthoryear{Emerson, Churcher, and Cockburn}{Emerson
  et~al.}{2015}]{emerson2015tag}
Emerson, J., N.~Churcher, and A.~Cockburn (2015).
\newblock Tag clouds for software and information visualisation.
\newblock In {\em Proceedings of the 14th Annual ACM SIGCHI\_NZ conference on
  Computer-Human Interaction}, pp.\ ~1. ACM.

\bibitem[\protect\citeauthoryear{Enderle}{Enderle}{2015}]{enderle2015sine}
Enderle, S. (2015, April).
\newblock What's a sine wave of sentiment?
\newblock
  \href{http://www.lagado.name/blog/sine-of-the-times/}{http://www.lagado.name/blog/sine-of-the-times/}.

\bibitem[\protect\citeauthoryear{Enderle}{Enderle}{2016}]{scott2016brownian}
Enderle, S. (2016, Sept).
\newblock Brownian noise and plot arcs.
\newblock
  \href{http://www.lagado.name/blog/brownian-noise-and-plot-arcs/}{http://www.lagado.name/blog/brownian-noise-and-plot-arcs/}.

\bibitem[\protect\citeauthoryear{Esuli and Sebastiani}{Esuli and
  Sebastiani}{2006}]{esuli2006sentiwordnet}
Esuli, A. and F.~Sebastiani (2006).
\newblock Sentiwordnet: A publicly available lexical resource for opinion
  mining.
\newblock In {\em Proceedings of LREC}, Volume~6, pp.\  417--422. Citeseer.

\bibitem[\protect\citeauthoryear{Feinberg}{Feinberg}{2009}]{feinberg2009wordle}
Feinberg, J. (2009).
\newblock Wordle-beautiful word clouds.

\bibitem[\protect\citeauthoryear{Fellbaum}{Fellbaum}{1998}]{fellbaum1998a}
Fellbaum, C. (Ed.) (1998).
\newblock {\em {WordNet}: {A}n Electronic Lexical Database}.
\newblock Cambridge, MA: MIT Press.

\bibitem[\protect\citeauthoryear{Finlayson}{Finlayson}{2011}]{finlayson2011learning}
Finlayson, M.~A. (2011).
\newblock {\em Learning narrative structure from annotated folktales}.
\newblock Ph.\ D. thesis, Massachusetts Institute of Technology.

\bibitem[\protect\citeauthoryear{Gabasova}{Gabasova}{2015}]{gabasova2015star-blog}
Gabasova, E. (2015, Dec).
\newblock The star wars social network.
\newblock
  \href{http://evelinag.com/blog/2015/12-15-star-wars-social-network/}{http://evelinag.com/blog/2015/12-15-star-wars-social-network/index.html}.

\bibitem[\protect\citeauthoryear{Gallagher, Reagan, Danforth, and
  Dodds}{Gallagher et~al.}{2016}]{gallagher2016divergent}
Gallagher, R.~J., A.~J. Reagan, C.~M. Danforth, and P.~S. Dodds (2016).
\newblock Divergent discourse between protests and counter-protests:
  {\#}blacklivesmatter and {\#}alllivesmatter.
\newblock {\em CoRR\/}~{\em abs/1606.06820}.

\bibitem[\protect\citeauthoryear{Gao, Jockers, Laudun, and Tangherlini}{Gao
  et~al.}{2016}]{gao2016multiscale}
Gao, J., M.~L. Jockers, J.~Laudun, and T.~Tangherlini (2016).
\newblock A multiscale theory for the dynamical evolution of sentiment in
  novels.
\newblock In {\em Behavioral, Economic and Socio-cultural Computing (BESC),
  2016 International Conference on}, pp.\  1--4. IEEE.

\bibitem[\protect\citeauthoryear{Garcia, Garas, and Schweitzer}{Garcia
  et~al.}{2015}]{garcia2015language}
Garcia, D., A.~Garas, and F.~Schweitzer (2015).
\newblock The language-dependent relationship between word happiness and
  frequency.
\newblock {\em Proceedings of the National Academy of Sciences\/}~{\em
  112\/}(23), E2983.

\bibitem[\protect\citeauthoryear{Gelman and Basb{\o}ll}{Gelman and
  Basb{\o}ll}{2014}]{gelman2014stories}
Gelman, A. and T.~Basb{\o}ll (2014).
\newblock When do stories work? evidence and illustration in the social
  sciences.
\newblock {\em Sociological Methods \& Research\/}, 0049124114526377.

\bibitem[\protect\citeauthoryear{Giachanou and Crestani}{Giachanou and
  Crestani}{2016}]{giachanou2016like}
Giachanou, A. and F.~Crestani (2016, June).
\newblock Like it or not: A survey of twitter sentiment analysis methods.
\newblock {\em ACM Comput. Surv.\/}~{\em 49\/}(2), 28:1--28:41.

\bibitem[\protect\citeauthoryear{Gleick}{Gleick}{2011}]{gleick2011information}
Gleick, J. (2011).
\newblock {\em The Information: A History, A Theory, A Flood}.
\newblock New York: Pantheon.

\bibitem[\protect\citeauthoryear{Golder and Macy}{Golder and
  Macy}{2011}]{golder2011diurnal}
Golder, S.~A. and M.~W. Macy (2011).
\newblock Diurnal and seasonal mood vary with work, sleep, and daylength across
  diverse cultures.
\newblock {\em Science Magazine\/}~{\em 333}, 1878--1881.

\bibitem[\protect\citeauthoryear{Gon{\c{c}}alves, Ara{\'u}jo, Benevenuto, and
  Cha}{Gon{\c{c}}alves et~al.}{2013}]{gonccalves2013comparing}
Gon{\c{c}}alves, P., M.~Ara{\'u}jo, F.~Benevenuto, and M.~Cha (2013).
\newblock Comparing and combining sentiment analysis methods.
\newblock In {\em Proceedings of the first ACM conference on Online social
  networks}, pp.\  27--38. ACM.

\bibitem[\protect\citeauthoryear{Gottesman, Reagan, and Dodds}{Gottesman
  et~al.}{2014}]{gottesman2014collective}
Gottesman, W.~G., A.~J. Reagan, and P.~S. Dodds (2014).
\newblock Collective philanthropy: {D}escribing and modeling the ecology of
  giving.
\newblock {\em PLoS ONE\/}~{\em 9}, e98876.

\bibitem[\protect\citeauthoryear{Gottschall}{Gottschall}{2013}]{gottschall2013storytelling}
Gottschall, J. (2013).
\newblock {\em The Storytelling Animal: How Stories Make Us Human}.
\newblock New York, NY: Mariner Books.

\bibitem[\protect\citeauthoryear{Goyal, Riloff, et~al.}{Goyal
  et~al.}{2013}]{goyal2013computational}
Goyal, A., E.~Riloff, et~al. (2013).
\newblock A computational model for plot units.
\newblock {\em Computational Intelligence\/}~{\em 29\/}(3), 466--488.

\bibitem[\protect\citeauthoryear{Halvey and Keane}{Halvey and
  Keane}{2007}]{halvey2007assessment}
Halvey, M.~J. and M.~T. Keane (2007).
\newblock An assessment of tag presentation techniques.
\newblock In {\em Proceedings of the 16th international conference on World
  Wide Web}, pp.\  1313--1314. ACM.

\bibitem[\protect\citeauthoryear{Hamann}{Hamann}{2012}]{hamann2012mapping}
Hamann, S. (2012).
\newblock Mapping discrete and dimensional emotions onto the brain:
  controversies and consensus.
\newblock {\em Trends in cognitive sciences\/}~{\em 16\/}(9), 458--466.

\bibitem[\protect\citeauthoryear{Hamilton, Clark, Leskovec, and
  Jurafsky}{Hamilton et~al.}{2016}]{hamilton2016inducing}
Hamilton, W.~L., K.~Clark, J.~Leskovec, and D.~Jurafsky (2016).
\newblock Inducing domain-specific sentiment lexicons from unlabeled corpora.
\newblock {\em arXiv preprint arXiv:1606.02820\/}.

\bibitem[\protect\citeauthoryear{Hand and Yu}{Hand and
  Yu}{2001}]{hand2001idiot}
Hand, D.~J. and K.~Yu (2001).
\newblock Idiot's bayes---not so stupid after all?
\newblock {\em International statistical review\/}~{\em 69\/}(3), 385--398.

\bibitem[\protect\citeauthoryear{Handler, Denny, Wallach, and O'Connor}{Handler
  et~al.}{2016}]{handler2016bag}
Handler, A., M.~J. Denny, H.~Wallach, and B.~O'Connor (2016).
\newblock Bag of what? simple noun phrase extraction for text analysis.
\newblock {\em NLP+ CSS 2016\/}, 114.

\bibitem[\protect\citeauthoryear{Harris}{Harris}{1959}]{harris1959basic}
Harris, W.~F. (1959).
\newblock {\em The basic patterns of plot}.
\newblock Oklahoma: University of Oklahoma Press.

\bibitem[\protect\citeauthoryear{Harrison, Gray, Gianaros, and
  Critchley}{Harrison et~al.}{2010}]{harrison2010embodiment}
Harrison, N.~A., M.~A. Gray, P.~J. Gianaros, and H.~D. Critchley (2010).
\newblock The embodiment of emotional feelings in the brain.
\newblock {\em Journal of Neuroscience\/}~{\em 30\/}(38), 12878--12884.

\bibitem[\protect\citeauthoryear{Hatzivassiloglou and McKeown}{Hatzivassiloglou
  and McKeown}{1997}]{hatzivassiloglou1997predicting}
Hatzivassiloglou, V. and K.~R. McKeown (1997).
\newblock Predicting the semantic orientation of adjectives.
\newblock In {\em Proceedings of the eighth conference on European chapter of
  the Association for Computational Linguistics}, pp.\  174--181. Association
  for Computational Linguistics.

\bibitem[\protect\citeauthoryear{Hearst}{Hearst}{2009}]{hearst2009search}
Hearst, M. (2009).
\newblock {\em Search user interfaces}.
\newblock Cambridge University Press.

\bibitem[\protect\citeauthoryear{Hearst and Rosner}{Hearst and
  Rosner}{2008}]{hearst2008tag}
Hearst, M.~A. and D.~Rosner (2008).
\newblock Tag clouds: Data analysis tool or social signaller?
\newblock In {\em Hawaii International Conference on System Sciences,
  Proceedings of the 41st Annual}, pp.\  160--160. IEEE.

\bibitem[\protect\citeauthoryear{Heer}{Heer}{2014}]{heer2014text}
Heer, J. (2014).
\newblock Text visualizatoin.
\newblock CSE 512 Lecture available at
  \href{http://courses.cs.washington.edu/courses/cse512/14wi/}{http://courses.cs.washington.edu/courses/cse512/14wi/}.

\bibitem[\protect\citeauthoryear{Honnibal}{Honnibal}{2015}]{honnibal2015displaying}
Honnibal, M. (2015, Aug).
\newblock Displaying linguistic structure with css.
\newblock \url{https://spacy.io/blog/displacy-dependency-visualizer}.

\bibitem[\protect\citeauthoryear{Hu and Liu}{Hu and Liu}{2004}]{hu2004mining}
Hu, M. and B.~Liu (2004).
\newblock Mining and summarizing customer reviews.
\newblock In {\em Proceedings of the tenth ACM SIGKDD international conference
  on Knowledge discovery and data mining}, pp.\  168--177. ACM.

\bibitem[\protect\citeauthoryear{Hutto and Gilbert}{Hutto and
  Gilbert}{2014}]{hutto2014vader}
Hutto, C.~J. and E.~Gilbert (2014, May).
\newblock Vader: A parsimonious rule-based model for sentiment analysis of
  social media text.
\newblock In {\em Eighth International AAAI Conference on Weblogs and Social
  Media}. AAAI Publications.

\bibitem[\protect\citeauthoryear{Jack, Garrod, and Schyns}{Jack
  et~al.}{2014}]{jack2014dynamic}
Jack, R.~E., O.~G. Garrod, and P.~G. Schyns (2014).
\newblock Dynamic facial expressions of emotion transmit an evolving hierarchy
  of signals over time.
\newblock {\em Current biology\/}~{\em 24\/}(2), 187--192.

\bibitem[\protect\citeauthoryear{Jockers}{Jockers}{2014}]{jockers2014novel}
Jockers, M. (2014, Jun).
\newblock A novel method for detecting plot.
\newblock
  \url{http://www.matthewjockers.net/2014/06/05/a-novel-method-for-detecting-plot/}.

\bibitem[\protect\citeauthoryear{Jockers}{Jockers}{2015}]{jockers2015rest}
Jockers, M. (2015, Feb).
\newblock The rest of the story.
\newblock
  \url{http://www.matthewjockers.net/2015/02/25/the-rest-of-the-story/}.

\bibitem[\protect\citeauthoryear{Jockers}{Jockers}{2013}]{jockers2013macroanalysis}
Jockers, M.~L. (2013).
\newblock {\em Macroanalysis: Digital methods and literary history}.
\newblock University of Illinois Press.

\bibitem[\protect\citeauthoryear{Kaji and Kitsuregawa}{Kaji and
  Kitsuregawa}{2007}]{kaji2007building}
Kaji, N. and M.~Kitsuregawa (2007).
\newblock Building lexicon for sentiment analysis from massive collection of
  {HTML} documents.
\newblock In {\em EMNLP-CoNLL}, pp.\  1075--1083.

\bibitem[\protect\citeauthoryear{Kay, Roberts, Samuels, and Wotherspoon}{Kay
  et~al.}{2009}]{kay2009historical}
Kay, C., J.~Roberts, M.~Samuels, and I.~Wotherspoon (2009).
\newblock {\em Historical Thesaurus of the Oxford English Dictionary}.
\newblock Oxford University Press.

\bibitem[\protect\citeauthoryear{Kiley, Reagan, Mitchell, Danforth, and
  Dodds}{Kiley et~al.}{2016}]{kiley2016game}
Kiley, D.~P., A.~J. Reagan, L.~Mitchell, C.~M. Danforth, and P.~S. Dodds (2016,
  May).
\newblock Game story space of professional sports: Australian rules football.
\newblock {\em Phys. Rev. E\/}~{\em 93}, 052314.

\bibitem[\protect\citeauthoryear{Kim and Hovy}{Kim and
  Hovy}{2004}]{kim2004determining}
Kim, S.-M. and E.~Hovy (2004).
\newblock Determining the sentiment of opinions.
\newblock In {\em Proceedings of the 20th international conference on
  Computational Linguistics}, pp.\  1367. Association for Computational
  Linguistics.

\bibitem[\protect\citeauthoryear{King, Pan, and Roberts}{King
  et~al.}{2016}]{king2016chinese}
King, G., J.~Pan, and M.~E. Roberts (2016).
\newblock How the chinese government fabricates social media posts for
  strategic distraction, not engaged argument.
\newblock {\em Harvard University\/}.

\bibitem[\protect\citeauthoryear{Kiritchenko, Zhu, and Mohammad}{Kiritchenko
  et~al.}{2014}]{kiritchenko2014sentiment}
Kiritchenko, S., X.~Zhu, and S.~M. Mohammad (2014).
\newblock Sentiment analysis of short informal texts.
\newblock {\em Journal of Artificial Intelligence Research\/}~{\em 50},
  723--762.

\bibitem[\protect\citeauthoryear{Kirschenbaum}{Kirschenbaum}{2007}]{kirschenbaum2007remaking}
Kirschenbaum, M.~G. (2007).
\newblock The remaking of reading: Data mining and the digital humanities.
\newblock In {\em The National Science Foundation Symposium on Next Generation
  of Data Mining and Cyber-Enabled Discovery for Innovation, Maryland}.

\bibitem[\protect\citeauthoryear{Koch, Alves, Kr{\"u}ger, and Unkelbach}{Koch
  et~al.}{2016}]{koch2016general}
Koch, A., H.~Alves, T.~Kr{\"u}ger, and C.~Unkelbach (2016).
\newblock A general valence asymmetry in similarity: Good is more alike than
  bad.
\newblock {\em Journal of Experimental Psychology: Learning, Memory, and
  Cognition\/}~{\em 42\/}(8), 1171.

\bibitem[\protect\citeauthoryear{Kohonen}{Kohonen}{1990}]{kohonen1990self}
Kohonen, T. (1990).
\newblock The self-organizing map.
\newblock {\em Proceedings of the IEEE\/}~{\em 78\/}(9), 1464--1480.

\bibitem[\protect\citeauthoryear{Kosinski, Stillwell, and Graepel}{Kosinski
  et~al.}{2013}]{kosinski2013private}
Kosinski, M., D.~Stillwell, and T.~Graepel (2013).
\newblock Private traits and attributes are predictable from digital records of
  human behavior.
\newblock {\em Proceedings of the National Academy of Sciences\/}~{\em
  110\/}(15), 5802--5805.

\bibitem[\protect\citeauthoryear{Kuster}{Kuster}{2015}]{kuster2015exploring}
Kuster, D. (2015, Jul).
\newblock Exploring the shapes of stories using python and sentiment apis.
\newblock
  \href{https://indico.io/blog/plotlines/}{https://indico.io/blog/plotlines/}.

\bibitem[\protect\citeauthoryear{Lee, Riche, Karlson, and Carpendale}{Lee
  et~al.}{2010}]{lee2010sparkclouds}
Lee, B., N.~H. Riche, A.~K. Karlson, and S.~Carpendale (2010).
\newblock Sparkclouds: Visualizing trends in tag clouds.
\newblock {\em IEEE transactions on visualization and computer graphics\/}~{\em
  16\/}(6), 1182--1189.

\bibitem[\protect\citeauthoryear{Lehnert}{Lehnert}{1981}]{lehnert1981plot}
Lehnert, W.~G. (1981).
\newblock Plot units and narrative summarization.
\newblock {\em Cognitive Science\/}~{\em 5\/}(4), 293--331.

\bibitem[\protect\citeauthoryear{Levallois}{Levallois}{2013}]{levallois2013umigon}
Levallois, C. (2013).
\newblock Umigon: sentiment analysis for tweets based on terms lists and
  heuristics.
\newblock In {\em Second Joint Conference on Lexical and Computational
  Semantics (* SEM)}, Volume~2, pp.\  414--417.

\bibitem[\protect\citeauthoryear{Levy}{Levy}{2008}]{levy2008case}
Levy, J.~S. (2008).
\newblock Case studies: Types, designs, and logics of inference.
\newblock {\em Conflict Management and Peace Science\/}~{\em 25\/}(1), 1--18.

\bibitem[\protect\citeauthoryear{Li, Lee-Urban, Appling, and Riedl}{Li
  et~al.}{2012}]{li2012crowdsourcing}
Li, B., S.~Lee-Urban, D.~S. Appling, and M.~O. Riedl (2012).
\newblock Crowdsourcing narrative intelligence.
\newblock {\em Advances in Cognitive Systems\/}~{\em 2}, 25--42.

\bibitem[\protect\citeauthoryear{Li, Lee-Urban, Johnston, and Riedl}{Li
  et~al.}{2013}]{li2013story}
Li, B., S.~Lee-Urban, G.~Johnston, and M.~Riedl (2013).
\newblock Story generation with crowdsourced plot graphs.
\newblock In {\em AAAI}.

\bibitem[\protect\citeauthoryear{Lin, Michel, Aiden, Orwant, Brockman, and
  Petrov}{Lin et~al.}{2012}]{lin2012syntactic}
Lin, Y., J.-B. Michel, E.~L. Aiden, J.~Orwant, W.~Brockman, and S.~Petrov
  (2012).
\newblock Syntactic annotations for the google books ngram corpus.
\newblock In {\em Proceedings of the ACL 2012 system demonstrations}, pp.\
  169--174. Association for Computational Linguistics.

\bibitem[\protect\citeauthoryear{Lindquist, Gendron, Satpute, Barrett, Lewis,
  and Haviland-Jones}{Lindquist et~al.}{2016}]{lindquist2016language}
Lindquist, K., M.~Gendron, A.~Satpute, L.~Barrett, M.~Lewis, and
  J.~Haviland-Jones (2016).
\newblock Language and emotion: Putting words into feelings and feelings into
  words.
\newblock {\em Handbook of emotions\/}.

\bibitem[\protect\citeauthoryear{Liu}{Liu}{2010}]{liu2010sentiment}
Liu, B. (2010).
\newblock Sentiment analysis and subjectivity.
\newblock {\em Handbook of natural language processing\/}~{\em 2}, 627--666.

\bibitem[\protect\citeauthoryear{Liu}{Liu}{2012}]{liu2012sentiment}
Liu, B. (2012, May).
\newblock {\em Sentiment analysis and opinion mining}.
\newblock Synthesis Lectures on Human Language Technologies. San Rafael, CA:
  Morgan \& Claypool Publishers.

\bibitem[\protect\citeauthoryear{Liu, Huang, An, and Yu}{Liu
  et~al.}{2007}]{liu2007arsa}
Liu, Y., X.~Huang, A.~An, and X.~Yu (2007).
\newblock Arsa: a sentiment-aware model for predicting sales performance using
  blogs.
\newblock In {\em Proceedings of the 30th annual international ACM SIGIR
  conference on Research and development in information retrieval}, pp.\
  607--614. ACM.

\bibitem[\protect\citeauthoryear{Lohmann, Ziegler, and Tetzlaff}{Lohmann
  et~al.}{2009}]{lohmann2009comparison}
Lohmann, S., J.~Ziegler, and L.~Tetzlaff (2009).
\newblock Comparison of tag cloud layouts: Task-related performance and visual
  exploration.
\newblock In {\em IFIP Conference on Human-Computer Interaction}, pp.\
  392--404. Springer.

\bibitem[\protect\citeauthoryear{Luo, Osborne, and Wang}{Luo
  et~al.}{2012}]{luo2012opinion}
Luo, Z., M.~Osborne, and T.~Wang (2012).
\newblock Opinion retrieval in twitter.
\newblock In {\em ICWSM}.

\bibitem[\protect\citeauthoryear{MacDonald}{MacDonald}{1982}]{macdonal1982storytellers}
MacDonald, M.~R. (1982).
\newblock {\em Storytellers Sourcebook: A Subject, Title, and Motif Index to
  Folklore Collections for Children}.
\newblock Michigan: Gale Group.

\bibitem[\protect\citeauthoryear{Mahoney and Goertz}{Mahoney and
  Goertz}{2006}]{mahoney2006tale}
Mahoney, J. and G.~Goertz (2006).
\newblock A tale of two cultures: Contrasting quantitative and qualitative
  research.
\newblock {\em Political analysis\/}~{\em 14\/}(3), 227--249.

\bibitem[\protect\citeauthoryear{Mandera, Keuleers, and Brysbaert}{Mandera
  et~al.}{2015}]{mandera2015useful}
Mandera, P., E.~Keuleers, and M.~Brysbaert (2015).
\newblock How useful are corpus-based methods for extrapolating
  psycholinguistic variables?
\newblock {\em The Quarterly Journal of Experimental Psychology\/}~{\em
  68\/}(8), 1623--1642.

\bibitem[\protect\citeauthoryear{Mani}{Mani}{2012}]{mani2012computational}
Mani, I. (2012).
\newblock Computational modeling of narrative.
\newblock {\em Synthesis Lectures on Human Language Technologies\/}~{\em
  5\/}(3), 1--142.

\bibitem[\protect\citeauthoryear{Mani, Verhagen, Wellner, Lee, and
  Pustejovsky}{Mani et~al.}{2006}]{mani2006machine}
Mani, I., M.~Verhagen, B.~Wellner, C.~M. Lee, and J.~Pustejovsky (2006).
\newblock Machine learning of temporal relations.
\newblock In {\em Proceedings of the 21st International Conference on
  Computational Linguistics and the 44th annual meeting of the Association for
  Computational Linguistics}, pp.\  753--760. Association for Computational
  Linguistics.

\bibitem[\protect\citeauthoryear{Manning, Surdeanu, Bauer, Finkel, Bethard, and
  McClosky}{Manning et~al.}{2014}]{manning2014stanford}
Manning, C.~D., M.~Surdeanu, J.~Bauer, J.~R. Finkel, S.~Bethard, and
  D.~McClosky (2014).
\newblock The stanford corenlp natural language processing toolkit.
\newblock In {\em ACL (System Demonstrations)}, pp.\  55--60.

\bibitem[\protect\citeauthoryear{Marcus, Marcinkiewicz, and Santorini}{Marcus
  et~al.}{1993}]{marcus1993building}
Marcus, M.~P., M.~A. Marcinkiewicz, and B.~Santorini (1993).
\newblock Building a large annotated corpus of english: The penn treebank.
\newblock {\em Computational linguistics\/}~{\em 19\/}(2), 313--330.

\bibitem[\protect\citeauthoryear{McCloud}{McCloud}{2006}]{mccloud2006making}
McCloud, S. (2006).
\newblock {\em Making comics: storytelling secrets of comics, manga and graphic
  novels}.
\newblock New York: Harper.

\bibitem[\protect\citeauthoryear{McIntyre and Lapata}{McIntyre and
  Lapata}{2010}]{mcintyre2010plot}
McIntyre, N. and M.~Lapata (2010).
\newblock Plot induction and evolutionary search for story generation.
\newblock In {\em Proceedings of the 48th Annual Meeting of the Association for
  Computational Linguistics}, ACL '10, Stroudsburg, PA, USA, pp.\  1562--1572.
  Association for Computational Linguistics.

\bibitem[\protect\citeauthoryear{Meeks and Averick}{Meeks and
  Averick}{}]{meeks2017data}
Meeks, E. and M.~Averick.
\newblock A data-driven exploration of archer.
\newblock
  \href{https://archervisualization.herokuapp.com/}{https://archervisualization.herokuapp.com/}.

\bibitem[\protect\citeauthoryear{Michel, Shen, Aiden, Veres, Gray, Pickett,
  Hoiberg, Clancy, Norvig, Orwant, et~al.}{Michel
  et~al.}{2011}]{michel2011quantitative}
Michel, J.-B., Y.~K. Shen, A.~P. Aiden, A.~Veres, M.~K. Gray, J.~P. Pickett,
  D.~Hoiberg, D.~Clancy, P.~Norvig, J.~Orwant, et~al. (2011).
\newblock Quantitative analysis of culture using millions of digitized books.
\newblock {\em Science\/}~{\em 331\/}(6014), 176--182.

\bibitem[\protect\citeauthoryear{Mikolov and Dean}{Mikolov and
  Dean}{2013}]{mikolov2013distributed}
Mikolov, T. and J.~Dean (2013).
\newblock Distributed representations of words and phrases and their
  compositionality.
\newblock {\em Advances in neural information processing systems\/}.

\bibitem[\protect\citeauthoryear{Min and Park}{Min and
  Park}{2016}]{min2016narrative}
Min, S. and J.~Park (2016).
\newblock Narrative as a complex network: A study of {V}ictor {H}ugo's les
  mis{\'e}rables.
\newblock In {\em Proceedings of HCI Korea}.

\bibitem[\protect\citeauthoryear{Mitchell, Frank, Harris, Dodds, and
  Danforth}{Mitchell et~al.}{2013}]{mitchell2013happiness}
Mitchell, L., M.~R. Frank, K.~D. Harris, P.~S. Dodds, and C.~M. Danforth (2013,
  May).
\newblock {The Geography of Happiness: Connecting Twitter Sentiment and
  Expression, Demographics, and Objective Characteristics of Place}.
\newblock {\em PLoS ONE\/}~{\em 8\/}(5), e64417.

\bibitem[\protect\citeauthoryear{Mohammad, Kiritchenko, and Zhu}{Mohammad
  et~al.}{2013}]{MohammadKZ2013}
Mohammad, S.~M., S.~Kiritchenko, and X.~Zhu (2013, June).
\newblock Nrc-canada: Building the state-of-the-art in sentiment analysis of
  tweets.
\newblock In {\em Proceedings of the seventh international workshop on Semantic
  Evaluation Exercises (SemEval-2013)}, Atlanta, Georgia, USA.

\bibitem[\protect\citeauthoryear{Mohammad and Turney}{Mohammad and
  Turney}{2013}]{mohammad2013crowdsourcing}
Mohammad, S.~M. and P.~D. Turney (2013).
\newblock Crowdsourcing a word--emotion association lexicon.
\newblock {\em Computational Intelligence\/}~{\em 29\/}(3), 436--465.

\bibitem[\protect\citeauthoryear{Moretti}{Moretti}{2000}]{moretti2000conjectures}
Moretti, F. (2000).
\newblock Conjectures on world literature.
\newblock {\em New Left Review\/}~{\em 1}, 54.

\bibitem[\protect\citeauthoryear{Moretti}{Moretti}{2007}]{moretti2007a}
Moretti, F. (2007).
\newblock {\em Graphs, Maps, Trees: Abstract Models for a Literary History}.
\newblock New York: Verso.

\bibitem[\protect\citeauthoryear{Moretti}{Moretti}{2013}]{moretti2013distant}
Moretti, F. (2013).
\newblock {\em Distant Reading}.
\newblock New York: Verso.

\bibitem[\protect\citeauthoryear{Mostafazadeh, Chambers, He, Parikh, Batra,
  Vanderwende, Kohli, and Allen}{Mostafazadeh
  et~al.}{2016}]{mostafazadeh2016corpus}
Mostafazadeh, N., N.~Chambers, X.~He, D.~Parikh, D.~Batra, L.~Vanderwende,
  P.~Kohli, and J.~Allen (2016, June).
\newblock A corpus and cloze evaluation for deeper understanding of commonsense
  stories.
\newblock In {\em Proceedings of the 2016 Conference of the North American
  Chapter of the Association for Computational Linguistics: Human Language
  Technologies}, San Diego, California, pp.\  839--849w. Association for
  Computational Linguistics.

\bibitem[\protect\citeauthoryear{Munroe}{Munroe}{2009}]{munroe657}
Munroe, R. (2009, 11).
\newblock Movie narrative charts.
\newblock http://xkcd.com/657/.

\bibitem[\protect\citeauthoryear{Munzner}{Munzner}{2014}]{munzner2014visualization}
Munzner, T. (2014).
\newblock {\em Visualization analysis and design}.
\newblock CRC Press.

\bibitem[\protect\citeauthoryear{Nenkova and McKeown}{Nenkova and
  McKeown}{2012}]{nenkova2012survey}
Nenkova, A. and K.~McKeown (2012).
\newblock A survey of text summarization techniques.
\newblock In {\em Mining text data}, pp.\  43--76. Berlin, Germany: Springer.

\bibitem[\protect\citeauthoryear{Neukom~Institute}{Neukom~Institute}{2016}]{digilit2016}
Neukom~Institute, D. (2016).
\newblock Turing tests in creative arts: Digilit 2016.
\newblock \url{https://math.dartmouth.edu/~turingtests/DigiLit.php}.

\bibitem[\protect\citeauthoryear{Nickerson}{Nickerson}{1998}]{nickerson1998confirmation}
Nickerson, R.~S. (1998).
\newblock Confirmation {B}ias; {A} ubiquitous phenomenon in many guises.
\newblock {\em Review of General Psychology\/}~{\em 2}, 175--220.

\bibitem[\protect\citeauthoryear{Nielsen}{Nielsen}{2011}]{nielsen2011new}
Nielsen, F.~{\AA}. (2011, May).
\newblock A new {ANEW}: Evaluation of a word list for sentiment analysis in
  microblogs.
\newblock In M.~Rowe, M.~Stankovic, A.-S. Dadzie, and M.~Hardey (Eds.), {\em
  CEUR Workshop Proceedings}, Volume Proceedings of the ESWC2011 Workshop on
  'Making Sense of Microposts': Big things come in small packages 718, pp.\
  93--98.

\bibitem[\protect\citeauthoryear{O'Connor}{O'Connor}{2013}]{oconnor2013learning}
O'Connor, B. (2013).
\newblock Learning frames from text with an unsupervised latent variable model.
\newblock {\em arXiv preprint arXiv:1307.7382\/}.

\bibitem[\protect\citeauthoryear{Ogawa and Ma}{Ogawa and
  Ma}{2010}]{ogawa2010software}
Ogawa, M. and K.-L. Ma (2010).
\newblock Software evolution storylines.
\newblock In {\em Proceedings of the 5th international symposium on Software
  visualization}, pp.\  35--42. ACM.

\bibitem[\protect\citeauthoryear{Orman}{Orman}{2015}]{orman2015information}
Orman, L.~V. (2015).
\newblock Information paradox: Drowning in information, starving for knowledge.

\bibitem[\protect\citeauthoryear{Owoputi, O'Connor, Dyer, Gimpel, Schneider,
  and Smith}{Owoputi et~al.}{2013}]{owoputi2013improved}
Owoputi, O., B.~O'Connor, C.~Dyer, K.~Gimpel, N.~Schneider, and N.~A. Smith
  (2013).
\newblock Improved part-of-speech tagging for online conversational text with
  word clusters.
\newblock Association for Computational Linguistics.

\bibitem[\protect\citeauthoryear{Pang and Lee}{Pang and
  Lee}{2004}]{pang2004sentimental}
Pang, B. and L.~Lee (2004).
\newblock A sentimental education: Sentiment analysis using subjectivity
  summarization based on minimum cuts.
\newblock In {\em Proceedings of the ACL}.

\bibitem[\protect\citeauthoryear{Pappas, Katsimpras, and Stamatatos}{Pappas
  et~al.}{2013}]{pappas2013distinguishing}
Pappas, N., G.~Katsimpras, and E.~Stamatatos (2013).
\newblock Distinguishing the popularity between topics: A system for up-to-date
  opinion retrieval and mining in the web.
\newblock In {\em International Conference on Intelligent Text Processing and
  Computational Linguistics}, pp.\  197--209. Springer.

\bibitem[\protect\citeauthoryear{Pechenick, Danforth, and Dodds}{Pechenick
  et~al.}{2015}]{pechenick2015characterizing}
Pechenick, E.~A., C.~M. Danforth, and P.~S. Dodds (2015).
\newblock Characterizing the google books corpus: Strong limits to inferences
  of socio-cultural and linguistic evolution.
\newblock {\em arXiv preprint arXiv:1501.00960\/}.

\bibitem[\protect\citeauthoryear{Pennebaker, Francis, and Booth}{Pennebaker
  et~al.}{2001}]{pennebaker2001linguistic}
Pennebaker, J.~W., M.~E. Francis, and R.~J. Booth (2001).
\newblock Linguistic inquiry and word count: {LIWC} 2001.
\newblock {\em Mahway: Lawrence Erlbaum Associates\/}~{\em 71}, 2001.

\bibitem[\protect\citeauthoryear{Pichotta and Mooney}{Pichotta and
  Mooney}{2015}]{pichotta2015learning}
Pichotta, K. and R.~J. Mooney (2015).
\newblock Learning statistical scripts with lstm recurrent neural networks.
\newblock In {\em Proceedings of the 30th AAAI Conference on Artificial
  Intelligence}.

\bibitem[\protect\citeauthoryear{Piper}{Piper}{2015a}]{piper2015novel}
Piper, A. (2015a).
\newblock Novel devotions: Conversional reading, computational modeling, and
  the modern novel.
\newblock {\em New Literary History\/}~{\em 46\/}(1), 63--98.

\bibitem[\protect\citeauthoryear{Piper}{Piper}{2015b}]{piper2015validation}
Piper, A. (2015b, Mar).
\newblock Validation and subjective computing.
\newblock \href{http://txtlab.org/?p=470}{http://txtlab.org/?p=470}.

\bibitem[\protect\citeauthoryear{Plutchik}{Plutchik}{1991}]{plutchik1991emotions}
Plutchik, R. (1991).
\newblock {\em The emotions}.
\newblock University Press of America.

\bibitem[\protect\citeauthoryear{Plutchik}{Plutchik}{2001}]{plutchik2001nature}
Plutchik, R. (2001).
\newblock The nature of emotions human emotions have deep evolutionary roots, a
  fact that may explain their complexity and provide tools for clinical
  practice.
\newblock {\em American scientist\/}~{\em 89\/}(4), 344--350.

\bibitem[\protect\citeauthoryear{Polanyi and Zaenen}{Polanyi and
  Zaenen}{2006}]{polanyi2006contextual}
Polanyi, L. and A.~Zaenen (2006).
\newblock Contextual valence shifters.
\newblock In {\em Computing attitude and affect in text: Theory and
  applications}, pp.\  1--10. Springer.

\bibitem[\protect\citeauthoryear{Polti}{Polti}{1921}]{polti1921thirty}
Polti, G. (1921).
\newblock {\em The Thirty-Six Dramatic Situations}.
\newblock Ohio: James Knapp Reeve.

\bibitem[\protect\citeauthoryear{Poria, Gelbukh, Hussain, Howard, Das, and
  Bandyopadhyay}{Poria et~al.}{2013}]{poria2013enhanced}
Poria, S., A.~Gelbukh, A.~Hussain, N.~Howard, D.~Das, and S.~Bandyopadhyay
  (2013).
\newblock Enhanced senticnet with affective labels for concept-based opinion
  mining.
\newblock {\em IEEE Intelligent Systems\/}~{\em 28\/}(2), 31--38.

\bibitem[\protect\citeauthoryear{Porter}{Porter}{2001}]{porter2001snowball}
Porter, M.~F. (2001).
\newblock Snowball: A language for stemming algorithms.

\bibitem[\protect\citeauthoryear{Prado, Dahmen, Bazzan, Carron, and
  Kenna}{Prado et~al.}{2016}]{prado2016temporal}
Prado, S.~D., S.~R. Dahmen, A.~L.~C. Bazzan, P.~M. Carron, and R.~Kenna (2016).
\newblock Temporal network analysis of literary texts.

\bibitem[\protect\citeauthoryear{Pratchett, Stewart, and Cohen}{Pratchett
  et~al.}{2003}]{pratchett2003science}
Pratchett, T., I.~Stewart, and J.~Cohen (2003).
\newblock {\em The Science of Discworld II: The Globe}.
\newblock London, UK: Ebury Press.

\bibitem[\protect\citeauthoryear{Propp}{Propp}{1968}]{propp1968morphology}
Propp, V. (1968).
\newblock {\em Morphology of the Folktale. 1928}.
\newblock Texas: Texas University Press.

\bibitem[\protect\citeauthoryear{Pustejovsky, Hanks, Sauri, See, Gaizauskas,
  Setzer, Radev, Sundheim, Day, Ferro, et~al.}{Pustejovsky
  et~al.}{2003}]{pustejovsky2003timebank}
Pustejovsky, J., P.~Hanks, R.~Sauri, A.~See, R.~Gaizauskas, A.~Setzer,
  D.~Radev, B.~Sundheim, D.~Day, L.~Ferro, et~al. (2003).
\newblock The timebank corpus.
\newblock In {\em Corpus linguistics}, Volume 2003, pp.\ ~40.

\bibitem[\protect\citeauthoryear{Radford, Jozefowicz, and Sutskever}{Radford
  et~al.}{2017}]{1704.01444}
Radford, A., R.~Jozefowicz, and I.~Sutskever (2017).
\newblock Learning to generate reviews and discovering sentiment.

\bibitem[\protect\citeauthoryear{Raftery}{Raftery}{2011}]{raftery2011harmon}
Raftery, B. (2011, Sep).
\newblock How {D}an {H}armon drives himself crazy making {C}ommunity.
\newblock at \url{http://www.wired.com/2011/09/mf_harmon/}, accessed June 20,
  2016.

\bibitem[\protect\citeauthoryear{Rao and Ravichandran}{Rao and
  Ravichandran}{2009}]{rao2009semi}
Rao, D. and D.~Ravichandran (2009).
\newblock Semi-supervised polarity lexicon induction.
\newblock In {\em Proceedings of the 12th Conference of the European Chapter of
  the Association for Computational Linguistics}, pp.\  675--682. Association
  for Computational Linguistics.

\bibitem[\protect\citeauthoryear{Rayner}{Rayner}{1985}]{rayner1985linear}
Rayner, J. M.~V. (1985).
\newblock Linear relations in biomechanics: the statistics of scaling
  functions.
\newblock {\em J. Zool. Lond. (A)\/}~{\em 206}, 415--439.

\bibitem[\protect\citeauthoryear{Reagan, Tivnan, Williams, Danforth, and
  Dodds}{Reagan et~al.}{2015}]{reagan2016benchmarking}
Reagan, A., B.~Tivnan, J.~R. Williams, C.~M. Danforth, and P.~S. Dodds (2015).
\newblock Benchmarking sentiment analysis methods for large-scale texts: A case
  for using continuum-scored words and word shift graphs.
\newblock Preprint available at
  \href{https://arxiv.org/abs/1512.00531}{https://arxiv.org/abs/1512.00531}.

\bibitem[\protect\citeauthoryear{Regneri, Koller, and Pinkal}{Regneri
  et~al.}{2010}]{regneri2010learning}
Regneri, M., A.~Koller, and M.~Pinkal (2010).
\newblock Learning script knowledge with web experiments.
\newblock In {\em Proceedings of the 48th Annual Meeting of the Association for
  Computational Linguistics}, ACL '10, Stroudsburg, PA, USA, pp.\  979--988.
  Association for Computational Linguistics.

\bibitem[\protect\citeauthoryear{Reiter, Frank, and Hellwig}{Reiter
  et~al.}{2014}]{reiter2014nlp}
Reiter, N., A.~Frank, and O.~Hellwig (2014).
\newblock An nlp-based cross-document approach to narrative structure
  discovery.
\newblock {\em Literary and Linguistic Computing\/}~{\em 29\/}(4), 583--605.

\bibitem[\protect\citeauthoryear{Ribeiro, Ara{\'u}jo, Gon{\c{c}}alves,
  Andr{\'e}~Gon{\c{c}}alves, and Benevenuto}{Ribeiro
  et~al.}{2016}]{ribeiro2016sentibench}
Ribeiro, F.~N., M.~Ara{\'u}jo, P.~Gon{\c{c}}alves,
  M.~Andr{\'e}~Gon{\c{c}}alves, and F.~Benevenuto (2016, July).
\newblock {SentiBench} --- a benchmark comparison of state-of-the-practice
  sentiment analysis methods.
\newblock {\em {EPJ} Data Sci.\/}~{\em 5\/}(1), 23.

\bibitem[\protect\citeauthoryear{Ribeiro, Singh, and Guestrin}{Ribeiro
  et~al.}{2016}]{ribeiro2016should}
Ribeiro, M.~T., S.~Singh, and C.~Guestrin (2016).
\newblock Why should i trust you?: Explaining the predictions of any
  classifier.
\newblock In {\em Proceedings of the 22nd ACM SIGKDD International Conference
  on Knowledge Discovery and Data Mining}, pp.\  1135--1144. ACM.

\bibitem[\protect\citeauthoryear{Riedl}{Riedl}{2016}]{riedl2016computational}
Riedl, M.~O. (2016).
\newblock Computational narrative intelligence: A human-centered goal for
  artificial intelligence.
\newblock {\em arXiv preprint arXiv:1602.06484\/}.

\bibitem[\protect\citeauthoryear{Riedl and Harrison}{Riedl and
  Harrison}{2015}]{riedl2015using}
Riedl, M.~O. and B.~Harrison (2015).
\newblock Using stories to teach human values to artificial agents.

\bibitem[\protect\citeauthoryear{Rivadeneira, Gruen, Muller, and
  Millen}{Rivadeneira et~al.}{2007}]{rivadeneira2007getting}
Rivadeneira, A.~W., D.~M. Gruen, M.~J. Muller, and D.~R. Millen (2007).
\newblock Getting our head in the clouds: toward evaluation studies of
  tagclouds.
\newblock In {\em Proceedings of the SIGCHI conference on Human factors in
  computing systems}, pp.\  995--998. ACM.

\bibitem[\protect\citeauthoryear{Robinson}{Robinson}{2008}]{robinson2008brain}
Robinson, D.~L. (2008).
\newblock Brain function, emotional experience and personality.
\newblock {\em Netherlands Journal of Psychology\/}~{\em 64\/}(4), 152--168.

\bibitem[\protect\citeauthoryear{Roemmele, Kobayashi, Inoue, and
  Gordon}{Roemmele et~al.}{2017}]{roemmele2017an-rnn-based}
Roemmele, M., S.~Kobayashi, N.~Inoue, and A.~Gordon (2017, April).
\newblock An rnn-based binary classifier for the story cloze test.
\newblock Proceedings of Linking Models of Lexical, Sentential and
  Discourse-level Semantics, workshop at European Association for Computational
  Linguistics.

\bibitem[\protect\citeauthoryear{Rothe, Ebert, and Sch{\"u}tze}{Rothe
  et~al.}{2016}]{rothe2016ultradense}
Rothe, S., S.~Ebert, and H.~Sch{\"u}tze (2016).
\newblock Ultradense word embeddings by orthogonal transformation.
\newblock {\em arXiv preprint arXiv:1602.07572\/}.

\bibitem[\protect\citeauthoryear{Rousseeuw}{Rousseeuw}{1987}]{rousseeuw1987silhouettes}
Rousseeuw, P.~J. (1987).
\newblock Silhouettes: a graphical aid to the interpretation and validation of
  cluster analysis.
\newblock {\em Journal of computational and applied mathematics\/}~{\em 20},
  53--65.

\bibitem[\protect\citeauthoryear{Ruiz, Hristidis, Castillo, Gionis, and
  Jaimes}{Ruiz et~al.}{2012}]{ruiz2012correlating}
Ruiz, E.~J., V.~Hristidis, C.~Castillo, A.~Gionis, and A.~Jaimes (2012).
\newblock Correlating financial time series with micro-blogging activity.
\newblock In {\em Proceedings of the fifth ACM international conference on Web
  search and data mining}, pp.\  513--522. ACM.

\bibitem[\protect\citeauthoryear{Rumelhart}{Rumelhart}{1975}]{rumelhart1975notes}
Rumelhart, D.~E. (1975).
\newblock Notes on a schema for stories.
\newblock {\em Representation and understanding: Studies in cognitive
  science\/}~{\em 211\/}(236), 45.

\bibitem[\protect\citeauthoryear{Russell}{Russell}{1980}]{russell1980circumplex}
Russell, J.~A. (1980).
\newblock A circumplex model of affect.
\newblock {\em Journal of Personality and Social Psychology\/}~{\em 39\/}(6),
  1161--1178.

\bibitem[\protect\citeauthoryear{Ruths}{Ruths}{2016}]{ruths2016force}
Ruths, D. (2016, Mar).
\newblock Why the force awakens is not just a remake of a new hope.
\newblock
  \href{http://www.derekruths.com/2016/03/05/why-the-force-awakens-is-not-just-a-remake-of-a-new-hope/}{http://www.derekruths.com/2016/03/05/why-the-force-awakens-is-not-just-a-remake-of-a-new-hope/}.

\bibitem[\protect\citeauthoryear{Saif, Fernandez, He, and Alani}{Saif
  et~al.}{2013}]{saif2013evaluation}
Saif, H., M.~Fernandez, Y.~He, and H.~Alani (2013).
\newblock Evaluation datasets for twitter sentiment analysis: A survey and a
  new dataset, the sts-gold.

\bibitem[\protect\citeauthoryear{Sandhaus}{Sandhaus}{2008}]{nytimescorpus2008new}
Sandhaus, E. (2008).
\newblock The {N}ew {Y}ork {T}imes {A}nnotated {C}orpus.
\newblock {L}inguistic Data Consortium, Philadelphia.

\bibitem[\protect\citeauthoryear{Schank and Abelson}{Schank and
  Abelson}{1977}]{schank1977scripts}
Schank, R.~C. and R.~P. Abelson (1977).
\newblock {\em Scripts, plans, goals, and understanding: An inquiry into human
  knowledge structures}.
\newblock Psychology Press.

\bibitem[\protect\citeauthoryear{Schmidt}{Schmidt}{2015a}]{schmidt2015commodius}
Schmidt, B. (2015a, Apr).
\newblock Commodius vici of recirculation: The real problem with syuzhet.
\newblock
  \href{http://benschmidt.org/2015/04/03/commodius-vici-of-recirculation-the-real-problem-with-syuzhet/}{http://benschmidt.org/2015/04/03/commodius-vici-of-recirculation-the-real-problem-with-syuzhet/}.

\bibitem[\protect\citeauthoryear{Schmidt}{Schmidt}{2016}]{schmidt2016plot}
Schmidt, B. (2016, Jul).
\newblock Plot arceology 2016: emotion and tension.
\newblock
  \href{http://sappingattention.blogspot.com/2016/07/plot-arceology-emotion-and-tension.html}{http://sappingattention.blogspot.com/2016/07/plot-arceology-emotion-and-tension.html}.

\bibitem[\protect\citeauthoryear{Schmidt}{Schmidt}{2015b}]{schmidt2015plot}
Schmidt, B.~M. (2015b).
\newblock Plot arceology: A vector-space model of narrative structure.
\newblock In {\em Big Data (Big Data), 2015 IEEE International Conference on},
  pp.\  1667--1672. IEEE.

\bibitem[\protect\citeauthoryear{Schrammel, Leitner, and Tscheligi}{Schrammel
  et~al.}{2009}]{schrammel2009semantically}
Schrammel, J., M.~Leitner, and M.~Tscheligi (2009).
\newblock Semantically structured tag clouds: an empirical evaluation of
  clustered presentation approaches.
\newblock In {\em Proceedings of the SIGCHI Conference on Human Factors in
  Computing Systems}, pp.\  2037--2040. ACM.

\bibitem[\protect\citeauthoryear{Schrauf and Sanchez}{Schrauf and
  Sanchez}{2004}]{schrauf2004preponderance}
Schrauf, R.~W. and J.~Sanchez (2004).
\newblock The preponderance of negative emotion words across generations and
  across cultures.
\newblock {\em Journal of Multilingual and Multicultural Development\/}~{\em
  25}, 266--284.

\bibitem[\protect\citeauthoryear{Schulz}{Schulz}{2011}]{schulz2011what}
Schulz, K. (2011, Jun).
\newblock What is distant reading?
\newblock
  \href{http://www.nytimes.com/2011/06/26/books/review/the-mechanic-muse-what-is-distant-reading.html}{http://www.nytimes.com/2011/06/26/books/review/the-mechanic-muse-what-is-distant-reading.html}.

\bibitem[\protect\citeauthoryear{Shriller}{Shriller}{2017}]{shriller2017narrative}
Shriller, R.~J. (2017, Jan).
\newblock Narrative economics.
\newblock In C.~F. for Research In~Economics (Ed.), {\em 129th annual meeting
  of the American Economic Association}, Number 2069.

\bibitem[\protect\citeauthoryear{Si, Mukherjee, Liu, Li, Li, and Deng}{Si
  et~al.}{2013}]{si2013exploiting}
Si, J., A.~Mukherjee, B.~Liu, Q.~Li, H.~Li, and X.~Deng (2013).
\newblock Exploiting topic based {T}witter sentiment for stock prediction.
\newblock In {\em ACL (2)}, pp.\  24--29.

\bibitem[\protect\citeauthoryear{Snow, O'Connor, Jurafsky, and Ng}{Snow
  et~al.}{2008}]{snow2008cheap}
Snow, R., B.~O'Connor, D.~Jurafsky, and A.~Y. Ng (2008).
\newblock Cheap and fast---but is it good?: evaluating non-expert annotations
  for natural language tasks.
\newblock In {\em Proceedings of the conference on empirical methods in natural
  language processing}, pp.\  254--263. Association for Computational
  Linguistics.

\bibitem[\protect\citeauthoryear{Snyder and Palmer}{Snyder and
  Palmer}{2004}]{snyder2004english}
Snyder, B. and M.~Palmer (2004).
\newblock The english all-words task.
\newblock In {\em Senseval-3: Third International Workshop on the Evaluation of
  Systems for the Semantic Analysis of Text}, pp.\  41--43. Association for
  Computational Linguistics.

\bibitem[\protect\citeauthoryear{Socher, Perelygin, Wu, Chuang, Manning, Ng,
  and Potts}{Socher et~al.}{2013}]{socher2013a}
Socher, R., A.~Perelygin, J.~Y. Wu, J.~Chuang, C.~D. Manning, A.~Y. Ng, and
  C.~Potts (2013).
\newblock Recursive deep models for semantic compositionality over a sentiment
  treebank.
\newblock In {\em Proceedings of the conference on empirical methods in natural
  language processing (EMNLP)}, Volume 1631, pp.\  1642. Citeseer.

\bibitem[\protect\citeauthoryear{Stone, Dunphy, and Smith}{Stone
  et~al.}{1966}]{stone1966general}
Stone, P.~J., D.~C. Dunphy, and M.~S. Smith (1966).
\newblock The general inquirer: A computer approach to content analysis.
\newblock {\em MIT Press\/}.

\bibitem[\protect\citeauthoryear{Storr}{Storr}{2014}]{storr2014unpersuadables}
Storr, W. (2014).
\newblock {\em The unpersuadables: Adventures with the enemies of science}.
\newblock The Overlook Press.

\bibitem[\protect\citeauthoryear{Swafford}{Swafford}{2015}]{swafford2015problems}
Swafford, A. (2015, Mar).
\newblock Problems with the syuzhet package.
\newblock
  \href{https://annieswafford.wordpress.com/2015/03/02/syuzhet/}{https://annieswafford.wordpress.com/2015/03/02/syuzhet/}.

\bibitem[\protect\citeauthoryear{Taboada, Brooke, Tofiloski, Voll, and
  Stede}{Taboada et~al.}{2011}]{taboada2011lexicon}
Taboada, M., J.~Brooke, M.~Tofiloski, K.~Voll, and M.~Stede (2011).
\newblock Lexicon-based methods for sentiment analysis.
\newblock {\em Computational linguistics\/}~{\em 37\/}(2), 267--307.

\bibitem[\protect\citeauthoryear{Taboada and Grieve}{Taboada and
  Grieve}{2004}]{taboada2004analyzing}
Taboada, M. and J.~Grieve (2004).
\newblock Analyzing appraisal automatically.
\newblock In {\em Proceedings of AAAI Spring Symposium on Exploring Attitude
  and Affect in Text (AAAI Technical Re\# port SS\# 04\# 07), Stanford
  University, CA, pp. 158q161. AAAI Press}.

\bibitem[\protect\citeauthoryear{Tang, Wei, Qin, Zhou, and Liu}{Tang
  et~al.}{2014}]{tang2014building}
Tang, D., F.~Wei, B.~Qin, M.~Zhou, and T.~Liu (2014).
\newblock Building large-scale twitter-specific sentiment lexicon: A
  representation learning approach.
\newblock In {\em COLING}, pp.\  172--182.

\bibitem[\protect\citeauthoryear{Tenenbaum, Barrett, Medaris, and
  Devitt}{Tenenbaum et~al.}{2015}]{tenenbaum2015languages}
Tenenbaum, D.~J., K.~Barrett, S.~Medaris, and T.~Devitt (2015, February).
\newblock In 10 languages, happy words beat sad ones.
\newblock http://whyfiles.org/2015/in-10-languages-happy-words-beat-sad-ones/.

\bibitem[\protect\citeauthoryear{Thelwall, Buckley, and Paltoglou}{Thelwall
  et~al.}{2012}]{thelwall2012sentiment}
Thelwall, M., K.~Buckley, and G.~Paltoglou (2012).
\newblock Sentiment strength detection for the social web.
\newblock {\em Journal of the American Society for Information Science and
  Technology\/}~{\em 63\/}(1), 163--173.

\bibitem[\protect\citeauthoryear{Thelwall, Buckley, Paltoglou, Cai, and
  Kappas}{Thelwall et~al.}{2010}]{thelwall2010sentiment}
Thelwall, M., K.~Buckley, G.~Paltoglou, D.~Cai, and A.~Kappas (2010).
\newblock Sentiment strength detection in short informal text.
\newblock {\em Journal of the American Society for Information Science and
  Technology\/}~{\em 61\/}(12), 2544--2558.

\bibitem[\protect\citeauthoryear{Tobias}{Tobias}{1993}]{tobias1993master}
Tobias, R.~B. (1993).
\newblock {\em 20 Master Plots: And How to Build Them}.
\newblock Ohio: Writer's Digest Books.

\bibitem[\protect\citeauthoryear{Toutanova, Klein, Manning, and
  Singer}{Toutanova et~al.}{2003}]{toutanova2003feature}
Toutanova, K., D.~Klein, C.~D. Manning, and Y.~Singer (2003).
\newblock Feature-rich part-of-speech tagging with a cyclic dependency network.
\newblock In {\em Proceedings of the 2003 Conference of the North American
  Chapter of the Association for Computational Linguistics on Human Language
  Technology-Volume 1}, pp.\  173--180. Association for Computational
  Linguistics.

\bibitem[\protect\citeauthoryear{Tumasjan, Sprenger, Sandner, and
  Welpe}{Tumasjan et~al.}{2010}]{tumasjan2010predicting}
Tumasjan, A., T.~O. Sprenger, P.~G. Sandner, and I.~M. Welpe (2010).
\newblock Predicting elections with twitter: What 140 characters reveal about
  political sentiment.
\newblock {\em ICWSM\/}~{\em 10}, 178--185.

\bibitem[\protect\citeauthoryear{Turney}{Turney}{2002}]{turney2002thumbs}
Turney, P.~D. (2002).
\newblock Thumbs up or thumbs down?: semantic orientation applied to
  unsupervised classification of reviews.
\newblock In {\em Proceedings of the 40th annual meeting on association for
  computational linguistics}, pp.\  417--424. Association for Computational
  Linguistics.

\bibitem[\protect\citeauthoryear{Turney and Littman}{Turney and
  Littman}{2003}]{turney2003measuring}
Turney, P.~D. and M.~L. Littman (2003).
\newblock Measuring praise and criticism: Inference of semantic orientation
  from association.
\newblock {\em ACM Transactions on Information Systems (TOIS)\/}~{\em 21\/}(4),
  315--346.

\bibitem[\protect\citeauthoryear{Uther}{Uther}{2011}]{uther2011types}
Uther, H.-J. (2011).
\newblock {\em The Types of International Folktales. A Classification and
  Bibliography. Based on the System of Antti Aarne and Stith Thompson. Part I.
  Animal Tales, Tales of Magic, Religious Tales, and Realistic Tales, with an
  Introduction (FF Communications, 284)}.
\newblock Helsinki, Finland: Finnish Academy of Science and Letters.

\bibitem[\protect\citeauthoryear{Valls-Vargas, Ontan{\'o}n, and
  Zhu}{Valls-Vargas et~al.}{2014}]{valls2014toward}
Valls-Vargas, J., S.~Ontan{\'o}n, and J.~Zhu (2014).
\newblock Toward automatic character identification in unannotated narrative
  text.
\newblock In {\em Seventh Intelligent Narrative Technologies Workshop}.

\bibitem[\protect\citeauthoryear{Valls-Vargas, Zhu, and
  Onta{\~n}{\'o}n}{Valls-Vargas et~al.}{2014}]{valls2014toward2}
Valls-Vargas, J., J.~Zhu, and S.~Onta{\~n}{\'o}n (2014).
\newblock Toward automatic role identification in unannotated folk tales.
\newblock In {\em Tenth Artificial Intelligence and Interactive Digital
  Entertainment Conference}.

\bibitem[\protect\citeauthoryear{Van~Ham, Wattenberg, and Vi{\'e}gas}{Van~Ham
  et~al.}{2009}]{van2009mapping}
Van~Ham, F., M.~Wattenberg, and F.~B. Vi{\'e}gas (2009).
\newblock Mapping text with phrase nets.
\newblock {\em IEEE transactions on visualization and computer graphics\/}~{\em
  15\/}(6).

\bibitem[\protect\citeauthoryear{Van~Rensbergen, De~Deyne, and
  Storms}{Van~Rensbergen et~al.}{2016}]{van2016estimating}
Van~Rensbergen, B., S.~De~Deyne, and G.~Storms (2016).
\newblock Estimating affective word covariates using word association data.
\newblock {\em Behavior Research Methods\/}~{\em 48\/}(4), 1644--1652.

\bibitem[\protect\citeauthoryear{Various}{Various}{}]{gutenberg}
Various.
\newblock Project {G}utenberg.

\bibitem[\protect\citeauthoryear{Velikovich, Blair-Goldensohn, Hannan, and
  McDonald}{Velikovich et~al.}{2010}]{velikovich2010viability}
Velikovich, L., S.~Blair-Goldensohn, K.~Hannan, and R.~McDonald (2010).
\newblock The viability of web-derived polarity lexicons.
\newblock In {\em Human Language Technologies: The 2010 Annual Conference of
  the North American Chapter of the Association for Computational Linguistics},
  pp.\  777--785. Association for Computational Linguistics.

\bibitem[\protect\citeauthoryear{Viegas, Wattenberg, and Feinberg}{Viegas
  et~al.}{2009}]{viegas2009participatory}
Viegas, F.~B., M.~Wattenberg, and J.~Feinberg (2009).
\newblock Participatory visualization with wordle.
\newblock {\em IEEE transactions on visualization and computer graphics\/}~{\em
  15\/}(6).

\bibitem[\protect\citeauthoryear{Viegas, Wattenberg, Van~Ham, Kriss, and
  McKeon}{Viegas et~al.}{2007}]{viegas2007manyeyes}
Viegas, F.~B., M.~Wattenberg, F.~Van~Ham, J.~Kriss, and M.~McKeon (2007).
\newblock Manyeyes: a site for visualization at internet scale.
\newblock {\em IEEE transactions on visualization and computer graphics\/}~{\em
  13\/}(6).

\bibitem[\protect\citeauthoryear{Volger}{Volger}{1992}]{volger1992writer}
Volger, C. (1992).
\newblock The writer's journey. mythic structure for storytellers and
  screenwriters.

\bibitem[\protect\citeauthoryear{Vonnegut}{Vonnegut}{1981}]{vonnegut1981palm}
Vonnegut, K. (1981).
\newblock {\em Palm Sunday}.
\newblock New York: RosettaBooks LLC.

\bibitem[\protect\citeauthoryear{Vonnegut}{Vonnegut}{1995}]{vonnegut1995shapes}
Vonnegut, K. (1995).
\newblock Shapes of stories.
\newblock https://www.youtube.com/watch?v=oP3c1h8v2ZQ.

\bibitem[\protect\citeauthoryear{Ward~Jr}{Ward~Jr}{1963}]{ward1963hierarchical}
Ward~Jr, J.~H. (1963).
\newblock Hierarchical grouping to optimize an objective function.
\newblock {\em Journal of the American statistical association\/}~{\em
  58\/}(301), 236--244.

\bibitem[\protect\citeauthoryear{Warriner, Kuperman, and Brysbaert}{Warriner
  et~al.}{2013}]{warriner2013norms}
Warriner, A.~B., V.~Kuperman, and M.~Brysbaert (2013).
\newblock Norms of valence, arousal, and dominance for 13,915 english lemmas.
\newblock {\em Behavior research methods\/}~{\em 45\/}(4), 1191--1207.

\bibitem[\protect\citeauthoryear{Watson and Clark}{Watson and
  Clark}{1999}]{watson1999panas}
Watson, D. and L.~A. Clark (1999).
\newblock {\em The {PANAS-X}: Manual for the positive and negative affect
  schedule-expanded form: Manual for the positive and negative affect
  schedule-expanded form}.
\newblock Ph.\ D. thesis, University of Iowa.

\bibitem[\protect\citeauthoryear{Weingart}{Weingart}{}]{weingart2015not}
Weingart, S.
\newblock Not enough perspectives, pt. 1.
\newblock
  \href{http://scottbot.net/not-enough-perspectives-pt-1/}{http://scottbot.net/not-enough-perspectives-pt-1/}.

\bibitem[\protect\citeauthoryear{Whissell, Fournier, Pelland, Weir, and
  Makarec}{Whissell et~al.}{1986}]{whissell1986dictionary}
Whissell, C., M.~Fournier, R.~Pelland, D.~Weir, and K.~Makarec (1986).
\newblock A dictionary of affect in language: Iv. reliability, validity, and
  applications.
\newblock {\em Perceptual and Motor Skills\/}~{\em 62\/}(3), 875--888.

\bibitem[\protect\citeauthoryear{Williams}{Williams}{2016}]{williams2016boundary}
Williams, J.~R. (2016).
\newblock Boundary-based mwe segmentation with text partitioning.
\newblock {\em arXiv preprint arXiv:1608.02025\/}.

\bibitem[\protect\citeauthoryear{Wilson, Wiebe, and Hoffmann}{Wilson
  et~al.}{2005}]{wilson2005recognizing}
Wilson, T., J.~Wiebe, and P.~Hoffmann (2005).
\newblock Recognizing contextual polarity in phrase-level sentiment analysis.
\newblock {\em Proceedings of Human Language Technologies Conference/Conference
  on Empirical Methods in Natural Language Processing (HLT/EMNLP 2005)\/}.

\bibitem[\protect\citeauthoryear{Winston}{Winston}{2011}]{winston2011strong}
Winston, P.~H. (2011).
\newblock The strong story hypothesis and the directed perception hypothesis.

\bibitem[\protect\citeauthoryear{Wojcik, Hovasapian, Graham, Motyl, and
  Ditto}{Wojcik et~al.}{2015}]{wojcik2015conservatives}
Wojcik, S.~P., A.~Hovasapian, J.~Graham, M.~Motyl, and P.~H. Ditto (2015).
\newblock Conservatives report, but liberals display, greater happiness.
\newblock {\em Science\/}~{\em 347\/}(6227), 1243--1246.

\bibitem[\protect\citeauthoryear{Wu}{Wu}{2016}]{wu2016interactive}
Wu, S. (2016, Dec).
\newblock An interactive visualization of every line in hamilton.
\newblock
  \href{http://polygraph.cool/hamilton/}{http://polygraph.cool/hamilton/}.

\bibitem[\protect\citeauthoryear{Xanthos, Pante, Rochat, and Grandjean}{Xanthos
  et~al.}{2016}]{xanthos2016visualising}
Xanthos, A., I.~Pante, Y.~Rochat, and M.~Grandjean (2016).
\newblock Visualising the dynamics of character networks.
\newblock {\em Digital Humanities 2016: Conference Abstracts\/}, 417--419.

\bibitem[\protect\citeauthoryear{Youyou, Kosinski, and Stillwell}{Youyou
  et~al.}{2015}]{youyou2015computer}
Youyou, W., M.~Kosinski, and D.~Stillwell (2015).
\newblock Computer-based personality judgments are more accurate than those
  made by humans.
\newblock {\em Proceedings of the National Academy of Sciences\/}~{\em
  112\/}(4), 1036--1040.

\bibitem[\protect\citeauthoryear{Zhu, Kiritchenko, and Mohammad}{Zhu
  et~al.}{2014}]{zhu2014nrc}
Zhu, X., S.~Kiritchenko, and S.~M. Mohammad (2014).
\newblock Nrc-canada-2014: Recent improvements in the sentiment analysis of
  tweets.
\newblock In {\em Proceedings of the 8th international workshop on semantic
  evaluation (SemEval 2014)}, pp.\  443--447. Citeseer.

\end{thebibliography}

\doublespacing

\chapter{Selected contributions to published work}
\chaptermark{Contributions} 
\label{chap:contributions}

Throughout the course of my studies at the University of Vermont,
I have enjoyed the collaborative research environment
afforded by the Computational Story Lab \footnote{In this Chapter I use the singular first person noun in place of the plural pronoun to discuss my individual contributions}.
As a result of these collaborations, I have assisted in the preparation of 10 other research papers.
I have variously done data visualization work, curated data from our Twitter database, built interactive online appendices, and assisted in performing mathematical analysis.
In this Chapter, I detail my contributions to each of these 10 papers, beginning with the paper abstract and then discussing my personal contribution.

\pagebreak

\section{Collective Philanthropy: Describing and Modeling the Ecology of Giving}

The first paper is \textit{Collective Philanthropy: Describing and Modeling the Ecology of Giving} by William L. Gottesman, Andrew James Reagan, and Peter Sheridan Dodds, cited as \cite{gottesman2014collective}.

\subsection{Abstract}
\begin{quote}
  Reflective of income and wealth distributions,
  philanthropic gifting appears to follow an approximate power-law size distribution as measured by the size of gifts received by individual institutions.
We explore the ecology of gifting by analyzing data sets of individual gifts for a diverse group of institutions dedicated to education, medicine, art, public support, and religion.
We find that the detailed forms of gift-size distributions differ across but are relatively constant within charity categories.
We construct a model for how a donor's income affects their giving preferences in different charity categories, offering a mechanistic explanation for variations in institutional gift-size distributions.
We discuss how knowledge of gift-sized distributions may be used to assess an institution's gift-giving profile, to help set fund-raising goals, and to design an institution-specific giving pyramid.
\end{quote}

\subsection{Contribution}

In this paper I prepared final versions of each visualization in the paper, working from the initial designs from both Professor Dodds and Bill Gottesman, and working closely with Professor Dodds in their preparation.
Additionally and at the request of the reviewers, I performed the statistical tests for support of power law distributions discussed in the paper, and included in the Appendix.
In addition to testing for support of power law distributions using the MLE estimator \cite{clauset2009b}, I ran likelihood comparison tests across many distributions, which we argue in the manuscript are potentially more applicable here to determine the most appropriate distribution.
In Figure 
The parameters for the various distributions mentioned in the paper are written using LaTeX variables, written in a .tex file by the MATLAB and Python scripts that perform the statistical procedures.
To the extend possible, all figures and analysis can be reproduced by running a single script.
In this Section we include a reprint of Figure 1, Figure S1, and the power law fit tables from the paper.
The codebase for creating the figures and performing the statistical procedures is available at \href{https://github.com/andyreagan/philanthropy-distributions-code}{https://github.com/andyreagan/philanthropy-distributions-code}.

\begin{figure}[tbp!]
  \centering
  \includegraphics[width=0.96\textwidth]{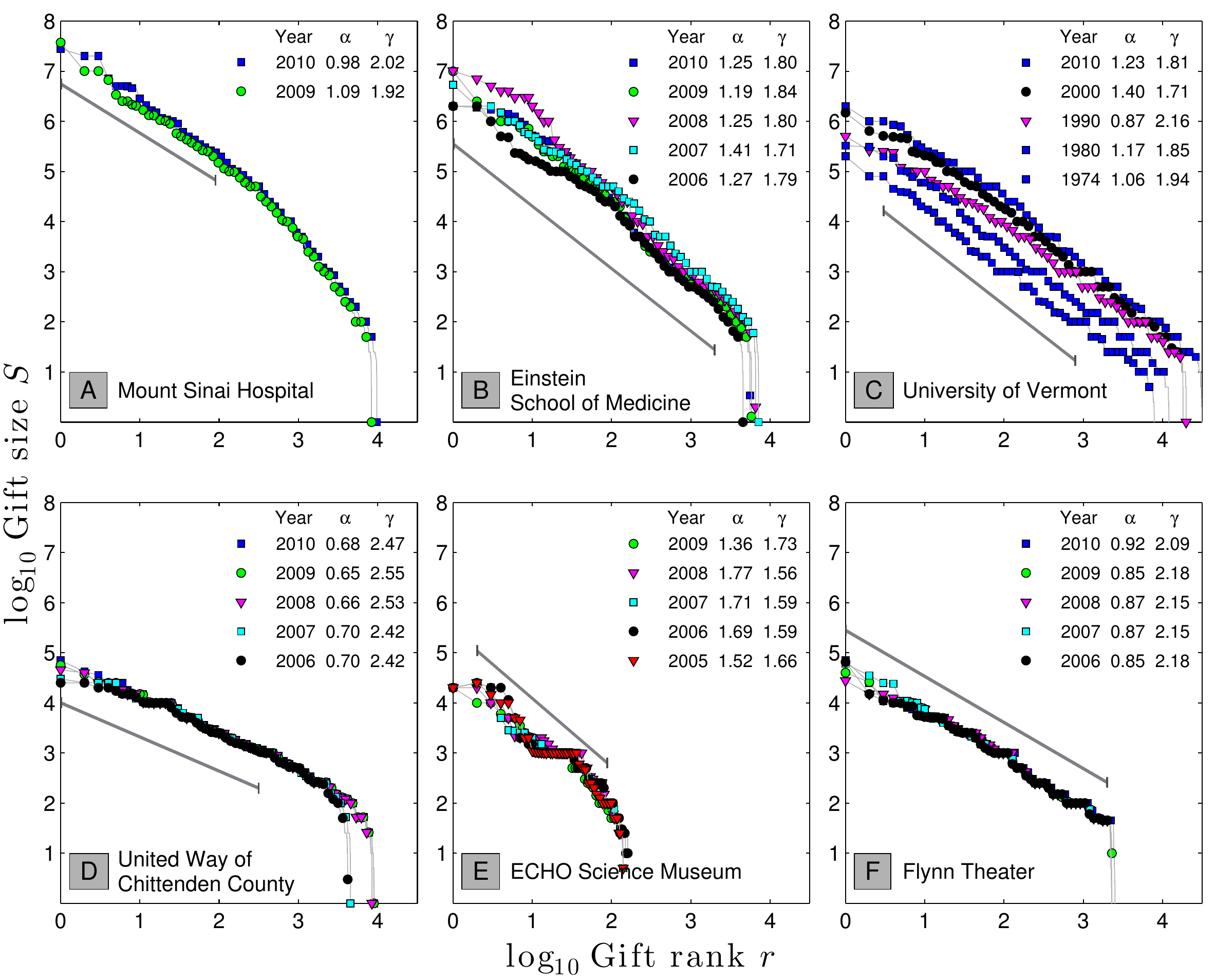}
  \caption[]{
    A reprint of Figure 1 from \cite{gottesman2014collective}, part of the caption is as follows:
    ``Gift size distributions for a range of institutions.
    The reported $\alpha$ and $\gamma$ were fitted to the region
indicated by solid gray line, and the 95\% CI of this fit, as well as
year for which the fit is plotted, are included for each organization.
    The ranges over which the data were fit was chosen empirically; other approaches were found to be inconsistent (see Supplementary).''
      }
  \label{fig:philanthropy-001}
\end{figure}

\begin{figure}[tbp!]
  \centering
  \includegraphics[width=0.96\textwidth]{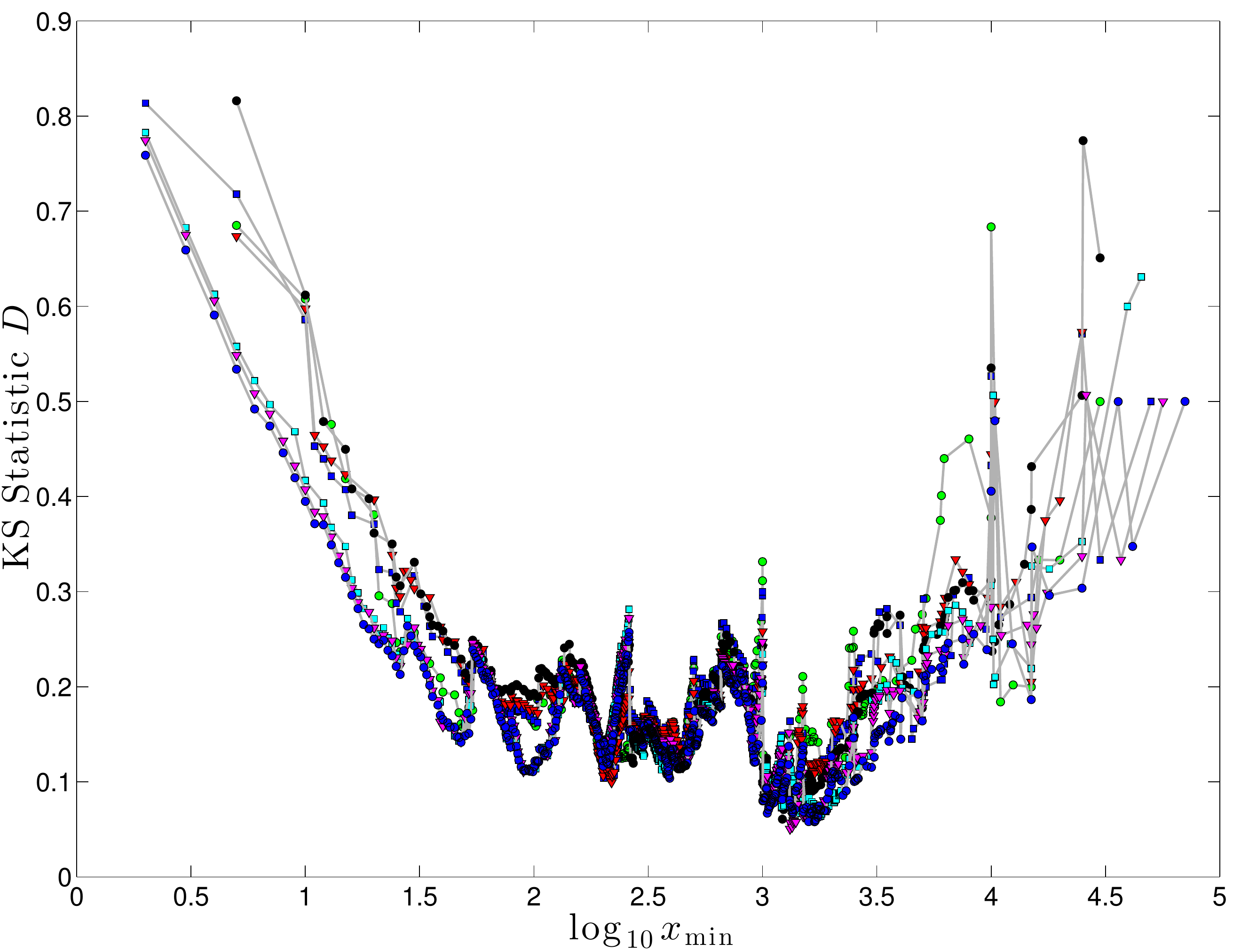}
  \caption[]{
    A reprint of Figure S1 from \cite{gottesman2014collective}, part of the caption is as follows:
    ``The Kolmogorov-Smirnoff statistic $D$ plotted over the log of $x_{\min}$, the minimum value fit for power law behavior, for the United Way of Chittenden County over the years 2006-2010.
    $D$ is generated from the ML estimate.
    Existence of multiple minima in our data indicate that there are multiple possible fitting regions for which the KS statistic details a good fit.
    The variability of this value over each year plotted produced widely varying scaling parameters $\gamma$, and thus cannot be used without actually looking at the data.''
      }
  \label{fig:philanthropy-S001}
\end{figure}

{\scriptsize
\begin{sidewaystable*}
\begin{tabular}{|lc|rrr|cl|cc|}
\hline
{\bf Institution} & {\bf Year} & $~~~~~~~~~~~\mathbf{ \langle x \rangle}$ & $~~~~~~~~~~~~~\mathbf{ \sigma}$ & $~~~~~~~~~\mathbf{x_\text{max}}$ &  $~~~~~~~~~\mathbf{\gamma}~~~~~~~~~$ & $\mathbf{\text{Range} }~~~~~$ & ~~{\bf D}~~ & ~~~{\bf p} ~~~\\ 
\hline
Mount Sinai Hospital & 2009 & 17618.40 & 450408.65 & 37259947 & 1.92 $\pm$ 0.08 & 1 to 90 & 0.12 & 0.00\\
 & 2010 & 19348.18 & 429587.88 & 27885708 & 2.02 $\pm$ 0.10 & 1 to 90 & 0.10 & 0.00\\
\hline
Einstein School of Medicine & 2006 & 3247.30 & 46940.29 & 2000000 & 1.79 $\pm$ 0.02 & 1 to 2000 & 0.11 & 0.00\\
 & 2007 & 4768.09 & 78762.48 & 5350000 & 1.71 $\pm$ 0.01 & 1 to 2000 & 0.15 & 0.00\\
 & 2008 & 10385.80 & 199751.68 & 10200000 & 1.80 $\pm$ 0.01 & 1 to 2000 & 0.21 & 0.00\\
 & 2009 & 5212.92 & 139468.89 & 10000000 & 1.84 $\pm$ 0.01 & 1 to 2000 & 0.15 & 0.00\\
 & 2010 & 4917.94 & 61893.49 & 2000000 & 1.80 $\pm$ 0.06 & 1 to 2000 & 0.15 & 0.00\\
\hline
Univeristy of Vermont & 1974 & 155.76 & 2811.94 & 200000 & 1.94 $\pm$ 0.01 & 3 to 794 & 0.18 & 0.00\\
 & 1980 & 284.31 & 5284.36 & 326000 & 1.85 $\pm$ 0.03 & 3 to 794 & 0.11 & 0.00\\
 & 1990 & 350.23 & 5382.45 & 500000 & 2.16 $\pm$ 0.01 & 3 to 794 & 0.38 & 0.00\\
 & 2000 & 805.33 & 15120.53 & 1488000 & 1.71 $\pm$ 0.03 & 3 to 794 & 0.09 & 0.00\\
 & 2010 & 741.40 & 17029.10 & 2000000 & 1.81 $\pm$ 0.05 & 3 to 794 & 0.13 & 0.00\\
\hline
United Way, Chittendon County & 2004 & 441.71 & 1133.02 & 30000 & 2.77 $\pm$ 0.04 & 1 to 316 & 0.21 & 0.00\\
 & 2005 & 464.47 & 1444.26 & 50000 & 2.58 $\pm$ 0.22 & 1 to 316 & 0.13 & 0.00\\
 & 2006 & 456.86 & 1199.92 & 25000 & 2.42 $\pm$ 0.05 & 1 to 316 & 0.07 & 0.00\\
 & 2007 & 456.16 & 1279.14 & 30000 & 2.42 $\pm$ 0.14 & 1 to 316 & 0.07 & 0.00\\
 & 2008 & 287.53 & 1089.92 & 45460 & 2.53 $\pm$ 0.00 & 1 to 316 & 0.14 & 0.00\\
 & 2009 & 278.93 & 1122.44 & 56500 & 2.55 $\pm$ 0.08 & 1 to 316 & 0.12 & 0.00\\
 & 2010 & 287.58 & 1271.10 & 70518 & 2.47 $\pm$ 0.09 & 1 to 316 & 0.08 & 0.00\\
\hline
ECHO Science Museum & 2005 & 977.77 & 3153.41 & 25000 & 1.66 $\pm$ 0.03 & 2 to 88 & 0.20 & 0.00\\
 & 2006 & 951.16 & 3415.22 & 25000 & 1.59 $\pm$ 0.02 & 2 to 88 & 0.28 & 0.00\\
 & 2007 & 941.61 & 3161.08 & 25000 & 1.59 $\pm$ 0.07 & 2 to 88 & 0.31 & 0.00\\
 & 2008 & 956.88 & 2688.31 & 20000 & 1.56 $\pm$ 0.01 & 2 to 88 & 0.26 & 0.00\\
 & 2009 & 676.84 & 2098.96 & 20000 & 1.73 $\pm$ 0.15 & 2 to 88 & 0.17 & 0.00\\
\hline
Flynn Theater & 2006 & 241.87 & 1528.82 & 65065 & 2.18 $\pm$ 0.04 & 1 to 2000 & 0.26 & 0.00\\
 & 2007 & 268.54 & 1732.33 & 60000 & 2.15 $\pm$ 0.05 & 1 to 2000 & 0.25 & 0.00\\
 & 2008 & 248.00 & 1015.39 & 27500 & 2.15 $\pm$ 0.00 & 1 to 2000 & 0.22 & 0.00\\
 & 2009 & 242.90 & 1212.42 & 40000 & 2.18 $\pm$ 0.04 & 1 to 2000 & 0.23 & 0.00\\
 & 2010 & 246.13 & 1606.43 & 70000 & 2.09 $\pm$ 0.05 & 1 to 2000 & 0.22 & 0.00\\
\hline
\end{tabular}
\label{table:fitstats}
\caption{Summary statistics of all of the donation data is presented. The reported $\gamma$ and range are fit with the MLE method, and the $x_\text{min}$ which was found to minimize the Kolmogorov-Smirnoff statistc {\bf D} is reported along with {\bf D} itself. In this case, lower values of {\bf D} indicate a better fit.}
\end{sidewaystable*}
\begin{sidewaystable*}
\begin{tabular}{|lc|c|cc|cc|cc|cc|}
\hline
 &  & & \multicolumn{2}{c}{{\bf Log-Normal }} & \multicolumn{2}{|c|}{{\bf Exponential }} & \multicolumn{2}{|c|}{{\bf Stretched Exp. }} & \multicolumn{2}{|c|}{{\bf Cutoff Power Law}} \\ 
{\bf Institution} & {\bf Year} & {\bf ~~p~~} & {\bf ~~LR~~} & {\bf ~~p~~}  &  {\bf ~~LR~~} & {\bf ~~p~~} & {\bf ~~LR~~} & {\bf ~~p~~} & {\bf ~~LR~~} & {\bf ~~p~~}\\ 
\hline
Mount Sinai Hospital & 2009 & 0.00& -0.21 & 0.67& \textcolor{blue}{31.80} &  \textcolor{blue}{$\mathbf{0.01}$}& -0.19 & 0.82& -0.53 & 0.30\\
 & 2010 & 0.00& -0.00 & 0.99& \textcolor{blue}{47.31} &  \textcolor{blue}{$\mathbf{0.00}$}& 0.46 & 0.60& -0.23 & 0.50\\
\hline
Einstein School of Medicine & 2006 & 0.00& \textcolor{red}{-6.22} &  \textcolor{red}{$\mathbf{0.03}$}& \textcolor{blue}{378.82} &  \textcolor{blue}{$\mathbf{0.00}$}& \textcolor{red}{-7.06} &  \textcolor{red}{$\mathbf{0.03}$}& \textcolor{red}{-8.31} &  \textcolor{red}{$\mathbf{0.00}$}\\
 & 2007 & 0.00& -0.30 & 0.59& \textcolor{blue}{17.65} &  \textcolor{blue}{$\mathbf{0.01}$}& -0.35 & 0.61& -0.67 & 0.25\\
 & 2008 & 0.00& -1.03 & 0.37& \textcolor{blue}{1235.22} &  \textcolor{blue}{$\mathbf{0.00}$}& 0.71 & 0.81& \textcolor{red}{-2.85} &  \textcolor{red}{$\mathbf{0.02}$}\\
 & 2009 & 0.00& -2.48 & 0.13& \textcolor{blue}{578.27} &  \textcolor{blue}{$\mathbf{0.00}$}& -2.75 & 0.22& \textcolor{red}{-5.82} &  \textcolor{red}{$\mathbf{0.00}$}\\
 & 2010 & 0.00& -1.52 & 0.22& \textcolor{blue}{842.87} &  \textcolor{blue}{$\mathbf{0.00}$}& -0.64 & 0.80& \textcolor{red}{-5.19} &  \textcolor{red}{$\mathbf{0.00}$}\\
\hline
Univeristy of Vermont & 1974 & 0.00& -0.39 & 0.54& \textcolor{blue}{20.93} &  \textcolor{blue}{$\mathbf{0.00}$}& -0.49 & 0.54& -1.17 & 0.13\\
 & 1980 & 0.00& -0.72 & 0.41& \textcolor{blue}{82.27} &  \textcolor{blue}{$\mathbf{0.00}$}& -0.81 & 0.47& \textcolor{red}{-1.82} &  \textcolor{red}{$\mathbf{0.06}$}\\
 & 1990 & 0.00& -0.94 & 0.36& \textcolor{blue}{23.05} &  \textcolor{blue}{$\mathbf{0.01}$}& -1.11 & 0.34& \textcolor{red}{-1.79} &  \textcolor{red}{$\mathbf{0.06}$}\\
 & 2000 & 0.00& -0.65 & 0.45& \textcolor{blue}{30.59} &  \textcolor{blue}{$\mathbf{0.00}$}& -0.78 & 0.44& \textcolor{red}{-1.52} &  \textcolor{red}{$\mathbf{0.08}$}\\
 & 2010 & 0.00& -inf & nan& \textcolor{blue}{7.75} &  \textcolor{blue}{$\mathbf{0.02}$}& 0.39 & 0.34& -0.00 & 0.94\\
\hline
United Way, Chittendon County & 2004 & 0.00& -0.46 & 0.47& \textcolor{blue}{28.75} &  \textcolor{blue}{$\mathbf{0.00}$}& -0.53 & 0.55& -1.29 & 0.11\\
 & 2005 & 0.00& -0.08 & 0.77& \textcolor{blue}{54.69} &  \textcolor{blue}{$\mathbf{0.00}$}& 0.36 & 0.74& -0.69 & 0.24\\
 & 2006 & 0.00& -0.12 & 0.71& \textcolor{blue}{68.71} &  \textcolor{blue}{$\mathbf{0.00}$}& 0.44 & 0.71& -0.85 & 0.19\\
 & 2007 & 0.00& -0.61 & 0.43& \textcolor{blue}{48.21} &  \textcolor{blue}{$\mathbf{0.00}$}& -0.65 & 0.57& \textcolor{red}{-1.64} &  \textcolor{red}{$\mathbf{0.07}$}\\
 & 2008 & 0.00& -0.13 & 0.72& \textcolor{blue}{46.52} &  \textcolor{blue}{$\mathbf{0.00}$}& 0.14 & 0.90& -0.71 & 0.23\\
 & 2009 & 0.00& -0.35 & 0.55& \textcolor{blue}{48.39} &  \textcolor{blue}{$\mathbf{0.00}$}& -0.28 & 0.80& -1.15 & 0.13\\
 & 2010 & 0.00& -0.32 & 0.58& \textcolor{blue}{35.25} &  \textcolor{blue}{$\mathbf{0.00}$}& -0.30 & 0.77& -0.90 & 0.18\\
\hline
ECHO Science Museum & 2005 & 0.00& -2.47 & 0.25& \textcolor{blue}{31.43} &  \textcolor{blue}{$\mathbf{0.04}$}& -3.04 & 0.21& \textcolor{red}{-3.56} &  \textcolor{red}{$\mathbf{0.01}$}\\
 & 2006 & 0.00& -0.20 & 0.69& 1.42 & 0.57& -0.28 & 0.68& -0.53 & 0.30\\
 & 2007 & 0.00& -inf & nan& \textcolor{blue}{4.56} &  \textcolor{blue}{$\mathbf{0.03}$}& 0.20 & 0.35& 0.00 & 1.00\\
 & 2008 & 0.00& -inf & nan& \textcolor{blue}{4.28} &  \textcolor{blue}{$\mathbf{0.00}$}& 0.29 & 0.19& 0.00 & 1.00\\
 & 2009 & 0.00& -0.87 & 0.47& \textcolor{blue}{31.48} &  \textcolor{blue}{$\mathbf{0.01}$}& -1.23 & 0.44& \textcolor{red}{-2.51} &  \textcolor{red}{$\mathbf{0.03}$}\\
\hline
Flynn Theater & 2006 & 0.00& -0.52 & 0.46& \textcolor{blue}{272.93} &  \textcolor{blue}{$\mathbf{0.00}$}& 0.32 & 0.87& \textcolor{red}{-2.80} &  \textcolor{red}{$\mathbf{0.02}$}\\
 & 2007 & 0.00& -0.06 & 0.80& 4.53 & 0.14& -0.08 & 0.86& -0.26 & 0.47\\
 & 2008 & 0.00& -0.56 & 0.45& \textcolor{blue}{303.73} &  \textcolor{blue}{$\mathbf{0.00}$}& 0.38 & 0.86& \textcolor{red}{-3.35} &  \textcolor{red}{$\mathbf{0.01}$}\\
 & 2009 & 0.00& -0.25 & 0.63& \textcolor{blue}{281.34} &  \textcolor{blue}{$\mathbf{0.00}$}& 1.11 & 0.59& \textcolor{red}{-2.16} &  \textcolor{red}{$\mathbf{0.04}$}\\
 & 2010 & 0.00& \textcolor{red}{-3.96} &  \textcolor{red}{$\mathbf{0.07}$}& \textcolor{blue}{129.19} &  \textcolor{blue}{$\mathbf{0.00}$}& \textcolor{red}{-4.61} &  \textcolor{red}{$\mathbf{0.06}$}& \textcolor{red}{-6.78} &  \textcolor{red}{$\mathbf{0.00}$}\\
\hline
\end{tabular}
\label{table:fittest}
\caption{The results of the Likelihood-Ratio and its associated {\bf p}-value are reported for different distributions. Here, positive values lend support to the Power Law and negative values to the other stated distribution. The significance of the {\bf LR} is {\bf p}, where low values of {\bf p} indicate a trustworthy {\bf LR}. Values for which $\mathbf{ p} < 0.05$ are bolded.}
\end{sidewaystable*}
}

\pagebreak
\clearpage

\section{Shadow networks: Discovering hidden nodes with models of information flow}

Paper number two is \textit{Shadow networks: Discovering hidden nodes with models of information flow} by James P. Bagrow, Suma Desu, Morgan R. Frank, Narine Manukyan, Lewis Mitchell, Andrew Reagan, Eric E. Bloedorn, Lashon B. Booker, Luther K. Branting, Michael J. Smith, Brian F. Tivnan, Christopher M. Danforth, Peter S. Dodds, and Joshua C. Bongard, cited as \cite{bagrow2014shadow}.

\subsection{Abstract}
\begin{quote}
    Complex, dynamic networks underlie many systems, and understanding these networks is the concern of a great span of important scientific and engineering problems.
  Quantitative description is crucial for this understanding yet, due to a range of measurement problems, many real network datasets are incomplete.
Here we explore how accidentally missing or deliberately hidden nodes may be detected in networks by the effect of their absence on predictions of the speed with which information flows through the network.
We use Symbolic Regression (SR) to learn models relating information flow to network topology.
  These models show localized, systematic, and non-random discrepancies when applied to test networks with intentionally masked nodes, demonstrating the ability to detect the presence of missing nodes and where in the network those nodes are likely to reside.
\end{quote}

\subsection{Contribution}

This paper is the result of a multi-day intensive collaboration called a Flash Mob Research Event.
The format is one or two days of everyone in the same room, brain storming how to tackle an important open question.
An outline of the paper is written, and after the event each member works to complete their part in carrying out the research idea.
My responsibility was to build reciprocal reply networks from Twitter data, in an effort to measure information flow over the network.
The network construction proceeded in three steps: (1) build a network using replies, (2) measure information flow over this reciprocal reply network, and (3) collect edges in the network for the actual information flow.
Each step of the construction would be carried out over a number of days, and using a single note on the VACC, we were able to build networks in memory for a total of 9 days.
These 9 days were considered for combinations 3/3/3 or 4/4/1 days, respectively.
These data were used in a real world test, to accompany testing of simulated data.

\section{Human language reveals a universal positivity bias}

Paper number three is \textit{Human language reveals a universal positivity bias} by Peter Sheridan Dodds, Eric M. Clark, Suma Desu, Morgan R. Frank, Andrew J. Reagan, Jake Ryland Williams, Lewis Mitchell, Kameron Decker Harris, Isabel M. Kloumann, James P. Bagrow, Karine Megerdoomian, Matthew T. McMahon, Brian F. Tivnan, and Christopher M. Danforth, cited as \cite{dodds2015human}.

\subsection{Abstract}
\begin{quote}
    Using human evaluation of 100,000 words spread across 24 corpora in 10 languages diverse in origin and culture, we present evidence of a deep imprint of human sociality in language, observing that
  (1) the words of natural human language possess a universal positivity bias;
  (2) the estimated emotional content of words is consistent between languages under translation;
  and (3) this positivity bias is strongly independent of frequency of word usage.
  Alongside these general regularities, we describe inter-language variations in the emotional spectrum of languages which allow us to rank corpora.
We also show how our word evaluations can be used to construct physical-like instruments for both real-time and offline measurement of the emotional content of large-scale texts.
\end{quote}

\subsection{Contribution}

In this paper I built the online appendices and performed additional tests of our method for building the sentiment timeseries for books (measuring their emotional arcs).
This included building a fully interactive version of an application of this dataset to analyze the emotional arcs of stories, which was done for a selection of the Western Canon and Project Gutenberg books.
In particular, we analyzed the emotional arc for these books in their original language, providing translations of the word shifts graphs into English.
The translations relied upon the translations of Google Translate, as curated by Eric Clark.
The additional statistical tests amounted to randomly shuffling the words in each book which we showcased, to demonstrate that the emotional arcs were meaningful.

\section{Climate change sentiment on Twitter: An unsolicited public opinion poll}

Paper number four is \textit{Climate change sentiment on Twitter: An unsolicited public opinion poll} by Emily M. Cody, Andrew J. Reagan, Lewis Mitchell, Peter Sheridan Dodds, and Christopher M. Danforth, cited as \cite{cody2015climate}.

\subsection{Abstract}
\begin{quote}
    The consequences of anthropogenic climate change are extensively debated through scientific papers, newspaper articles, and blogs.
Newspaper articles may lack accuracy, while the severity of findings in scientific papers may be too opaque for the public to understand.
Social media, however, is a forum where individuals of diverse backgrounds can share their thoughts and opinions.
As consumption shifts from old media to new, Twitter has become a valuable resource for analyzing current events and headline news.
In this research, we analyze tweets containing the word "climate" collected between September 2008 and July 2014.
Through use of a previously developed sentiment measurement tool called the Hedonometer, we determine how collective sentiment varies in response to climate change news, events, and natural disasters.
We find that natural disasters, climate bills, and oil-drilling can contribute to a decrease in happiness while climate rallies, a book release, and a green ideas contest can contribute to an increase in happiness.
Words uncovered by our analysis suggest that responses to climate change news are predominantly from climate change activists rather than climate change deniers, indicating that Twitter is a valuable resource for the spread of climate change awareness.
\end{quote}

\subsection{Contribution}

In this paper I was responsible for the data curation.
This amounted to searching the Twitter database on the VACC for a variety of keywords, storing those results, and processing them into useful formats for analysis.
Weighing at approximately 37TB of compressed JSON files, the Twitter database is difficult to search quickly over the GPFS architecture of the VACC, and only possible through the use of many short runtime (less than 2 hour) jobs.
Given all of this, a single search of the database takes approximately 2 days if everything is running smoothly.

\section{Reply to Garcia et al.: Common mistakes in measuring frequency dependent word characteristics}

The fifth paper is \textit{Reply to Garcia et al.: Common mistakes in measuring frequency dependent word characteristics} by P. S. Dodds, E. M. Clark, S. Desu, M. R. Frank, A. J. Reagan, J. R. Williams, L. Mitchell, K. D. Harris, I. M. Kloumann, J. P. Bagrow, K. Megerdoomian, M. T. McMahon, B. F. Tivnan, and C. M. Danforth, cited as \cite{dodds2015reply}.

\subsection{Abstract}
\begin{quote}
    We demonstrate that the concerns expressed by Garcia et al. are misplaced, due to
  (1) a misreading of our findings in \cite{dodds2015human};
  (2) a widespread failure to examine and present words in support of asserted summary quantities based on word usage frequencies;
  and (3) a range of misconceptions about word usage frequency, word rank, and expert-constructed word lists.
In particular, we show that the English component of our study compares well statistically with two related surveys, that no survey design influence is apparent, and that estimates of measurement error do not explain the positivity biases reported in our work and that of others.
We further demonstrate that for the frequency dependence of positivity
---of which we explored the nuances in great detail in \cite{dodds2015human}
---Garcia et al did not perform a reanalysis of our data---
they instead carried out an analysis of a
statistically improper data set and introduced a nonlinearity before performing linear regression.
\end{quote}

\subsection{Contribution}

For this paper I built a new online appendix, performed tests of the claims made by Garcia \etal (including re-making their visualizations), and built visualizations for the extended version of the reply (e.g. Table I and Figure 1 in the arXiv version).
Below, we include a reprint of the aforementioned Figure 1 and reproduction of the Figure from Garcia \etal:

\begin{figure}[ht]
  \centering
  \includegraphics[width=0.96\textwidth]{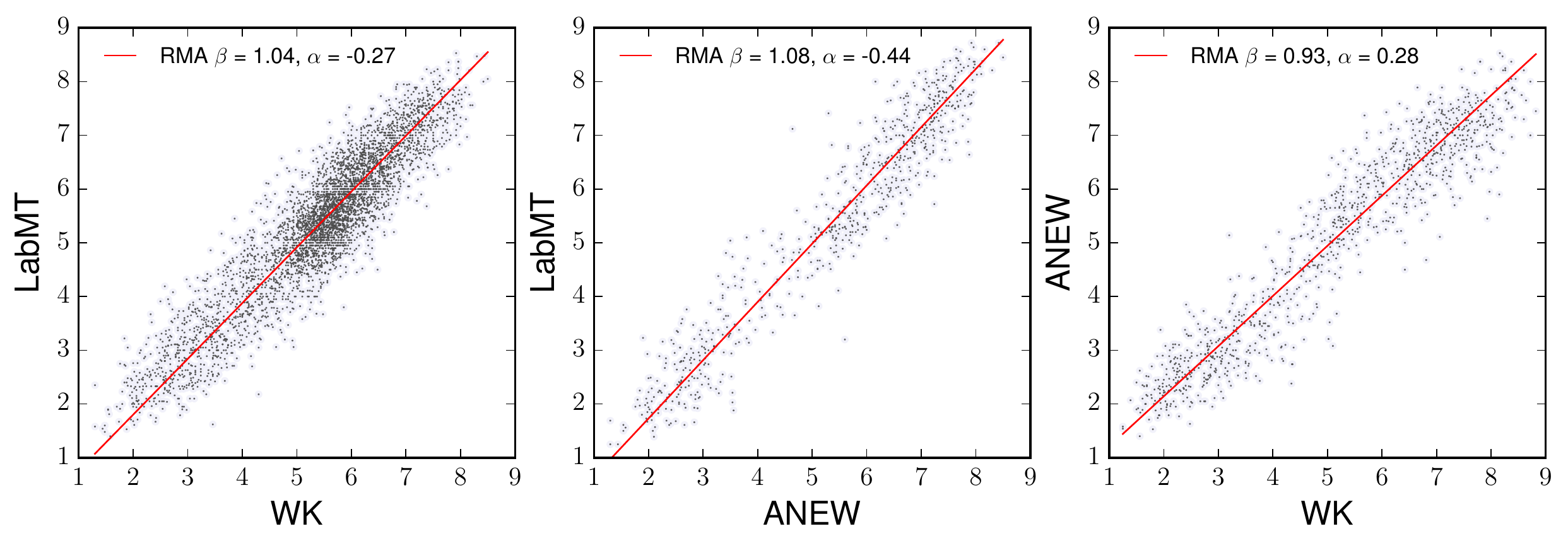}
  \caption[]{
    Reprint of Figure 1 from \cite{dodds2015reply}, with the caption as follows:
       ``Comparison of word ratings for three studies for
    overlapping words:
    labMT~\citep{dodds2011temporal},
    ANEW~\citep{bradley1999affective},
    and Warriner and Kuperman~\citep{warriner2013norms}
    Reduced major axis regression~\citep{rayner1985linear} yield
    the fits $\havgfn' = \beta \havgfn + \alpha$.''
  }
  \label{fig:reply-001}
\end{figure}

\begin{figure}[ht]
  \centering
  \includegraphics[width=0.45\textwidth]{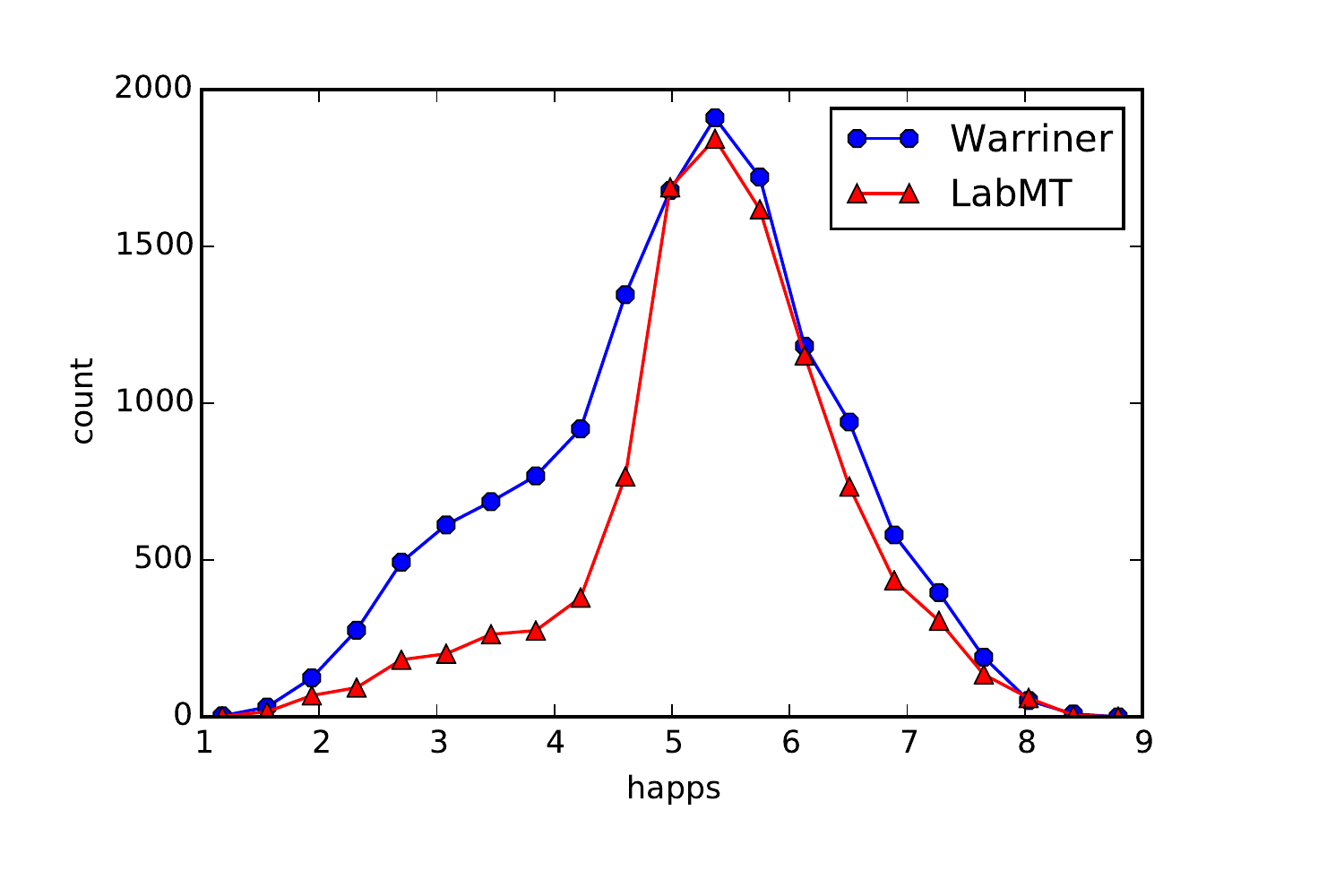}
  \includegraphics[width=0.45\textwidth]{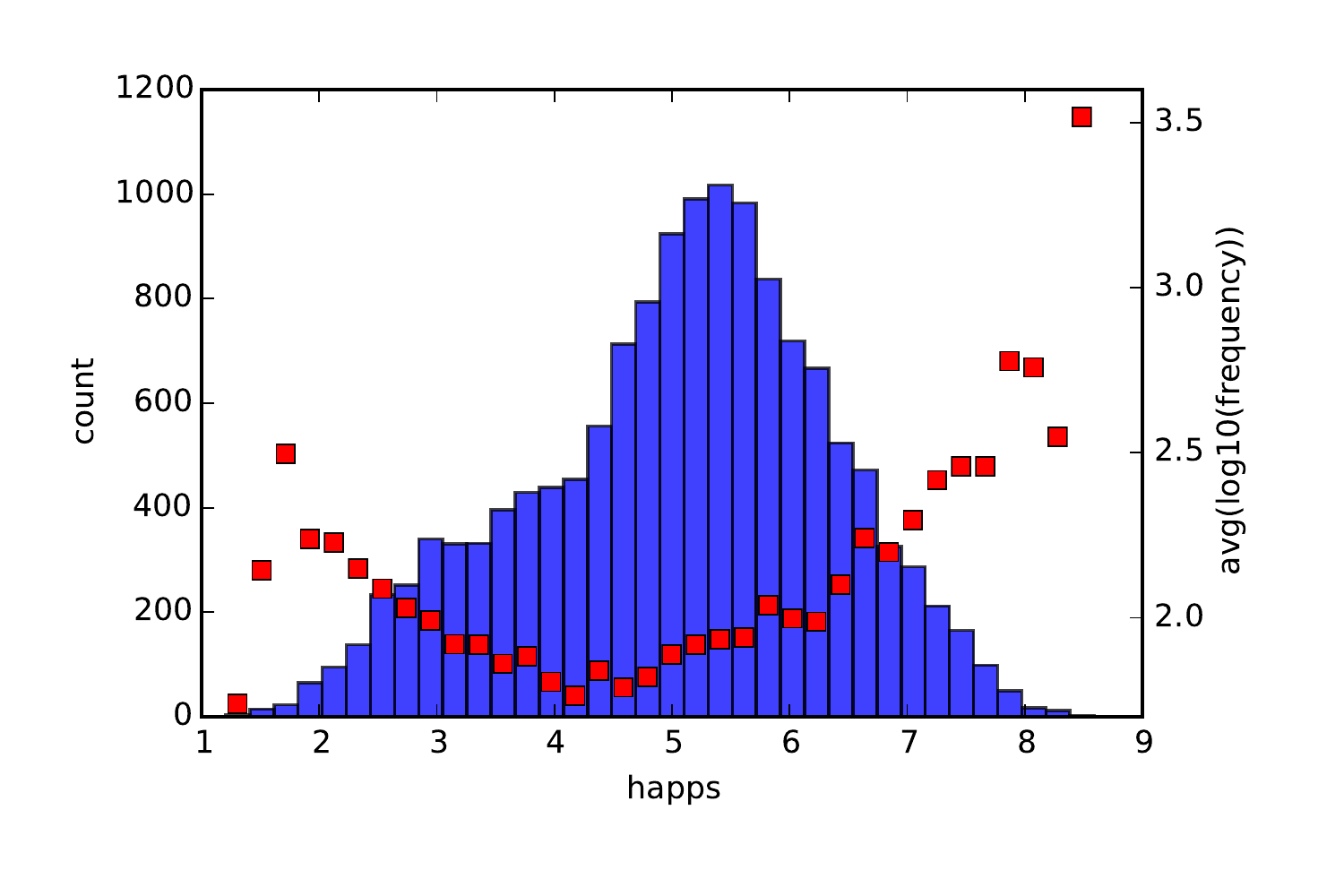}
  \caption[]{
    A reproduction of the Figure 1A and 1B from \cite{garcia2015language}.
      }
  \label{fig:reply-002}
\end{figure}

\clearpage
\pagebreak

\section{The game story space of professional sports: Australian Rules Football}

Paper number six is \textit{The game story space of professional sports: Australian Rules Football} by D. P. Kiley, A. J. Reagan, L. Mitchell, C. M. Danforth, and P. S. Dodds, cited as \cite{kiley2016game}.

\subsection{Abstract}
\begin{quote}
  Sports are spontaneous generators of stories.
Through skill and chance, the script of each game is dynamically written in real time by players acting out possible trajectories allowed by a sport's rules.
By properly characterizing a given sport's ecology of `game stories',
we are able to capture the sport's capacity for unfolding interesting narratives,
in part by contrasting them with random walks.
Here, we explore the game story space afforded by a data set of 1,310 Australian Football League (AFL) score lines.
We find that AFL games exhibit a continuous spectrum of stories rather than distinct clusters.
We show how coarse-graining reveals identifiable motifs ranging from last minute comeback wins to one-sided blowouts.
Through an extensive comparison with biased random walks,
we show that real AFL games deliver a broader array of motifs than null models,
and we provide consequent insights into the narrative appeal of real games.
\end{quote}

\subsection{Contribution}

For this paper I consulted with lead author Dilan Kiley on the statistical methods used, and assisted in performing the statistical analysis by leveraging the computational resources of the VACC.

\section{The Lexicocalorimeter: Gauging public health through caloric input and output on social media}

Paper number seven is \textit{The Lexicocalorimeter: Gauging public health through caloric input and output on social media} by S. E. Alajajian, J. R. Williams, A. J. Reagan, S. C. Alajajian, M. R. Frank, L. Mitchell, J. Lahne, C. M. Danforth, and P. S. Dodds, cited as \cite{alajajian2015lexicocalorimeter}.

\subsection{Abstract}
\begin{quote}
  We propose and develop a Lexicocalorimeter: an online, interactive instrument for measuring the ``caloric content'' of social media and other large-scale texts.
We do so by constructing extensive yet improvable tables of food and activity related phrases, and respectively assigning them with sourced estimates of caloric intake and expenditure.
We show that for Twitter, our naive measures of ``caloric input'', ``caloric output'', and the ratio of these measures are all strong correlates with health and well-being measures for the contiguous United States.
Our caloric balance measure in many cases outperforms both its constituent quantities,
is tunable to specific health and well-being measures such as diabetes rates,
has the capability of providing a real-time signal reflecting a population's health,
and has the potential to be used alongside traditional survey data in the development of public policy and collective self-awareness.
Because our Lexicocalorimeter is a linear superposition of principled phrase scores,
we also show we can move beyond correlations to explore what people talk about in collective detail,
and assist in the understanding and explanation of how population-scale conditions vary,
a capacity unavailable to black-box type methods.
\end{quote}

\subsection{Contribution}

\begin{figure}[ht]
  \centering
  \includegraphics[width=0.6\textwidth]{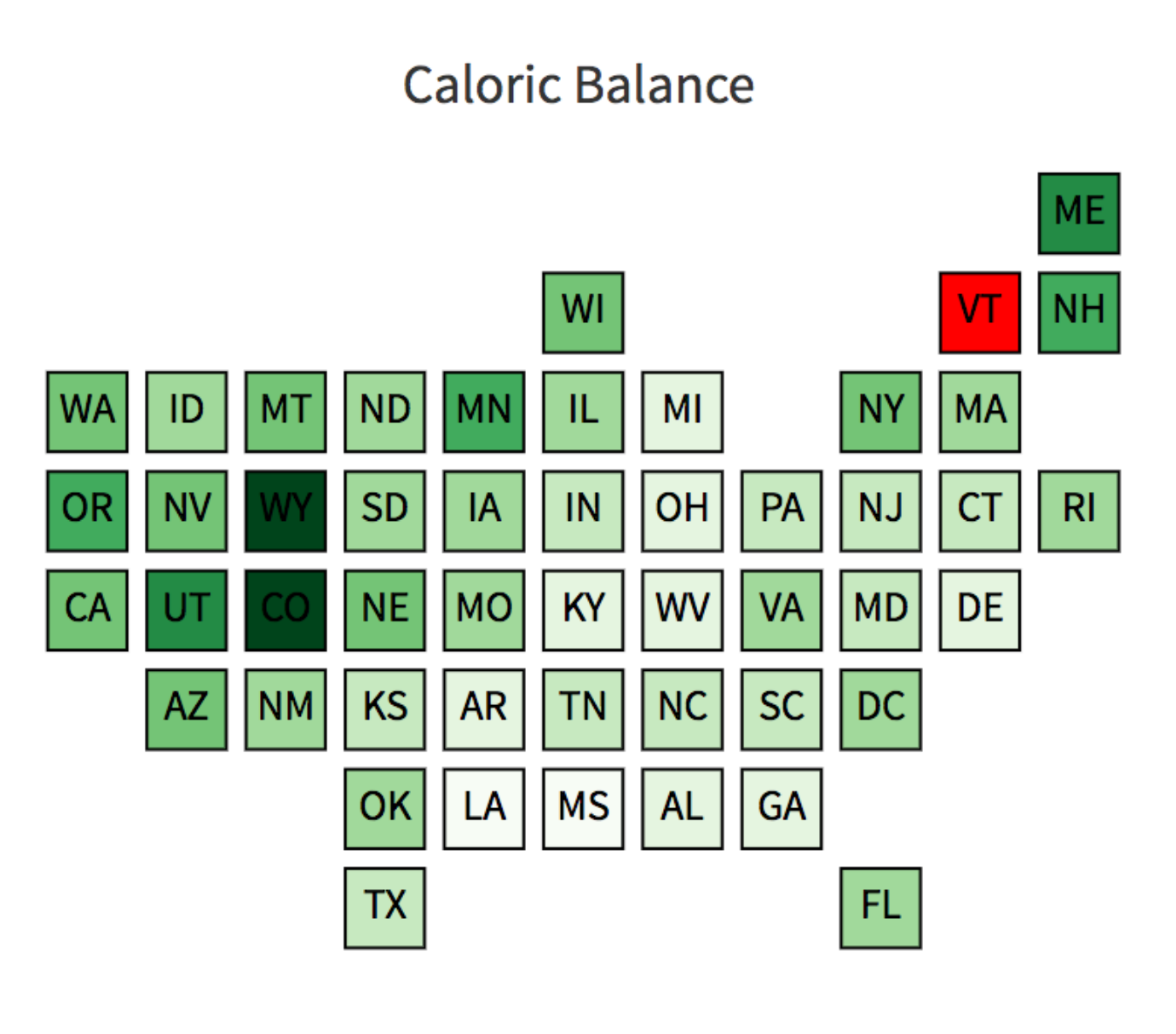}
  \caption[]{
    Lexicocalorimeter map, using square states to control for the disproportionate area and population of US States.
    Here, Vermont is highlighted by a hover.
      }
  \label{fig:panometer-001}
\end{figure}

\begin{figure}[ht]
  \centering
  \includegraphics[width=0.9\textwidth]{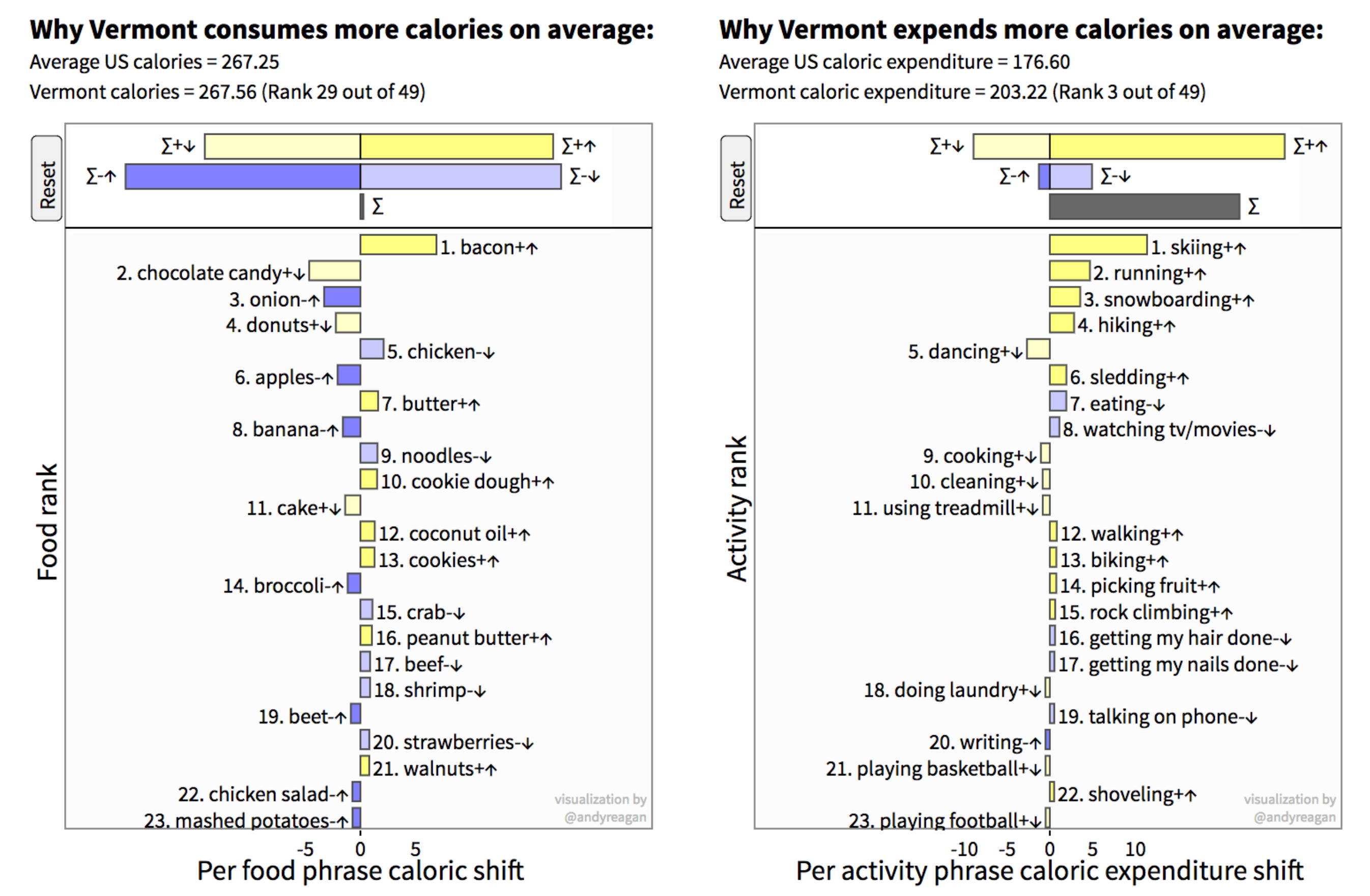}
  \caption[]{
    Lexicocalorimeter food and activity shifts.
    Here we see which foods and which activities contribute to Vermont's difference in caloric intake and expenditure from the US as a whole.
    We see that Bacon contributes most to caloric intake in Vermont relative to the average US intake, and overall Vermont is a middle-of-the-pack state (29th out of 49).
    On the right, Tweets from Vermont expend more calories than the US average with activities such as skiing, running, snowboarding, hiking, and sledding, giving the outdoorsy Vermont Twitter population the 3rd highest expenditure.
      }
  \label{fig:panometer-002}
\end{figure}

\begin{figure}[ht]
  \centering
  \includegraphics[width=0.9\textwidth]{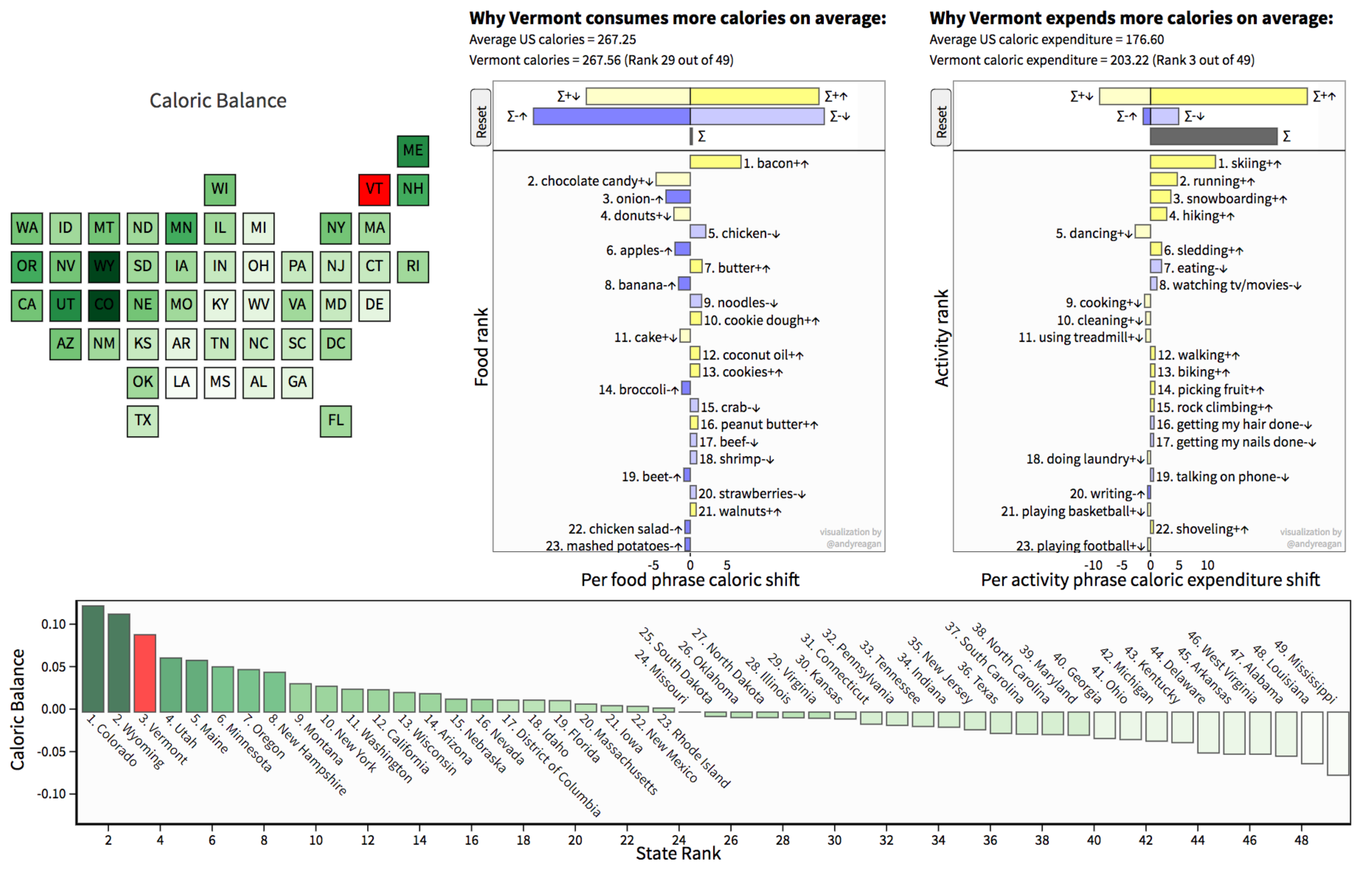}
  \caption[]{
    Overview of the Lexicocalorimeter dashboard.
    Each view is linked by hovering, and we can explore details of the caloric difference balances between states.
      }
  \label{fig:panometer-003}
\end{figure}

\begin{figure}[ht]
  \centering
  \includegraphics[width=0.9\textwidth]{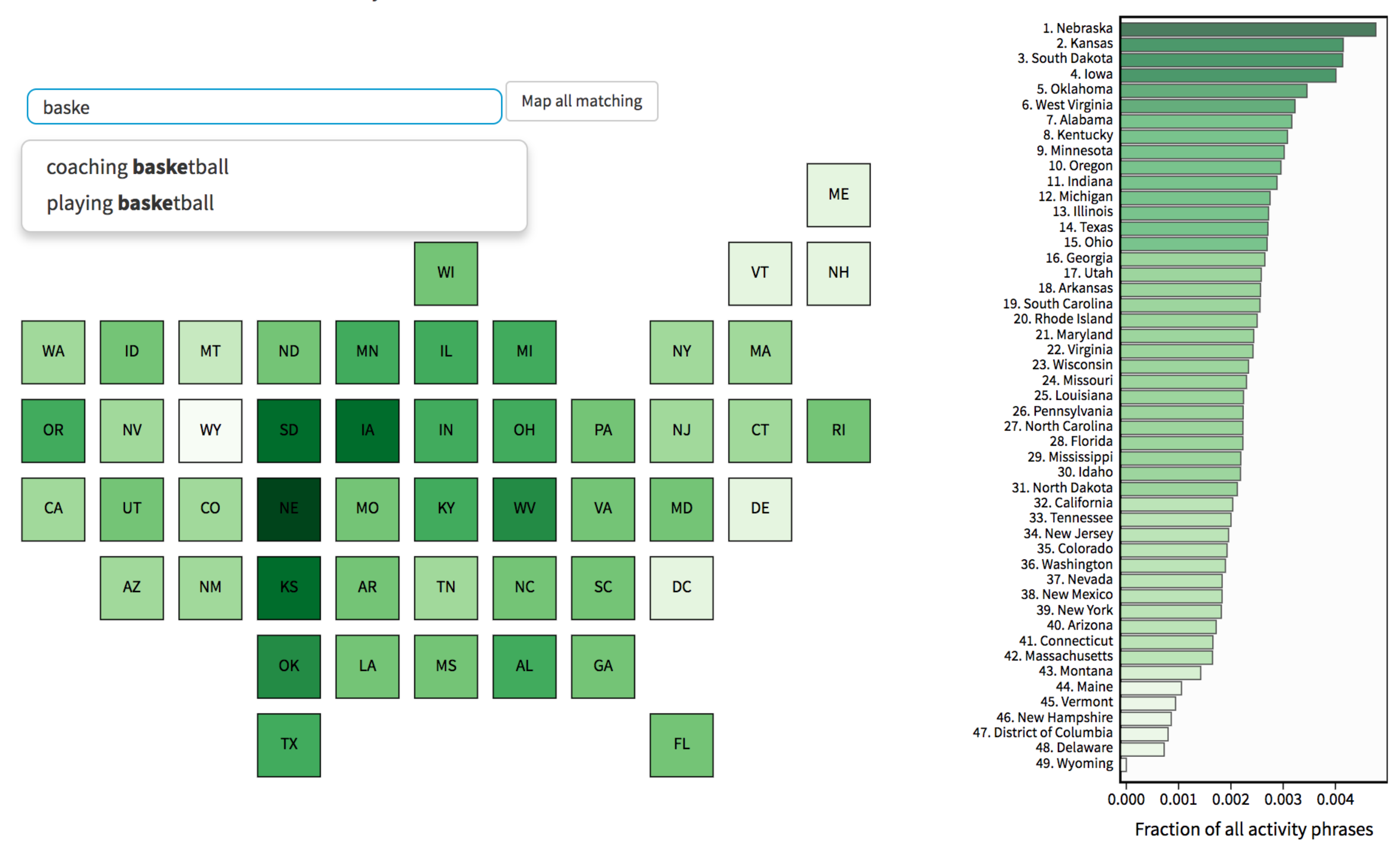}
  \caption[]{
    Snapshot of the Lexicocalorimeter activity search page.
    A similar page exists for foods.
    Here, we submit the query for ``basketball'', seeing that Nebraskans Tweet more about basketball relative to other activities than other US States.
      }
  \label{fig:panometer-004}
\end{figure}

For this paper I built an extensive online appendix and the accompanying website.
The online appendix at
\href{http://www.uvm.edu/storylab/share/papers/alajajian2015a/}{http://www.uvm.edu/storylab/share/papers/alajajian2015a/}
features an interactive dashboard provided at \href{http://panometer.org}{http://panometer.org}.
In addition to this tool, we provide searchable maps for all food and activity words used in the study.
Next, we show snapshots of the various visualizations available on the website, in Figures \ref{fig:panometer-001}--\ref{fig:panometer-004}.

\clearpage
\pagebreak

\section{Tracking the Teletherms: The spatiotemporal dynamics of the hottest and coldest days of the year}

Paper number eight is
\textit{Tracking the Teletherms: The spatiotemporal dynamics of the hottest and coldest days of the year}
by Peter Sheridan Dodds, Lewis Mitchell, Andrew J. Reagan, and Christopher M. Danforth,
cited as \cite{dodds2016tracking}.
\subsection{Abstract}
\begin{quote}
  Instabilities and long term shifts in seasons,
  whether induced by natural drivers or human activities,
  pose great disruptive threats to ecological, agricultural, and social systems.
  Here, we propose, measure, and explore two fundamental markers of location-sensitive seasonal variations:
  the Summer and Winter Teletherms
  ---
  the on-average annual dates of the hottest and coldest days of the year.
  We analyze daily temperature extremes recorded at 1218 stations across the contiguous United States
  from 1853--2012,
  and observe large regional variation with the Summer Teletherm falling up to 90 days after the Summer Solstice,
  and 50 days for the Winter Teletherm after the Winter Solstice.
  We show that Teletherm temporal dynamics are substantive
  with clear and in some cases dramatic shifts reflective of system bifurcations.
  We also compare recorded daily temperature extremes
  with output from two regional climate models finding considerable though relatively unbiased error.
  Our work demonstrates that Teletherms are an intuitive, powerful, and statistically sound measure
  of local climate change,
  and that they pose detailed, stringent challenges for future theoretical and computational models.
\end{quote}

\subsection{Contribution}

\begin{figure}[tbp!]
  \centering
  \includegraphics[width=0.96\textwidth]{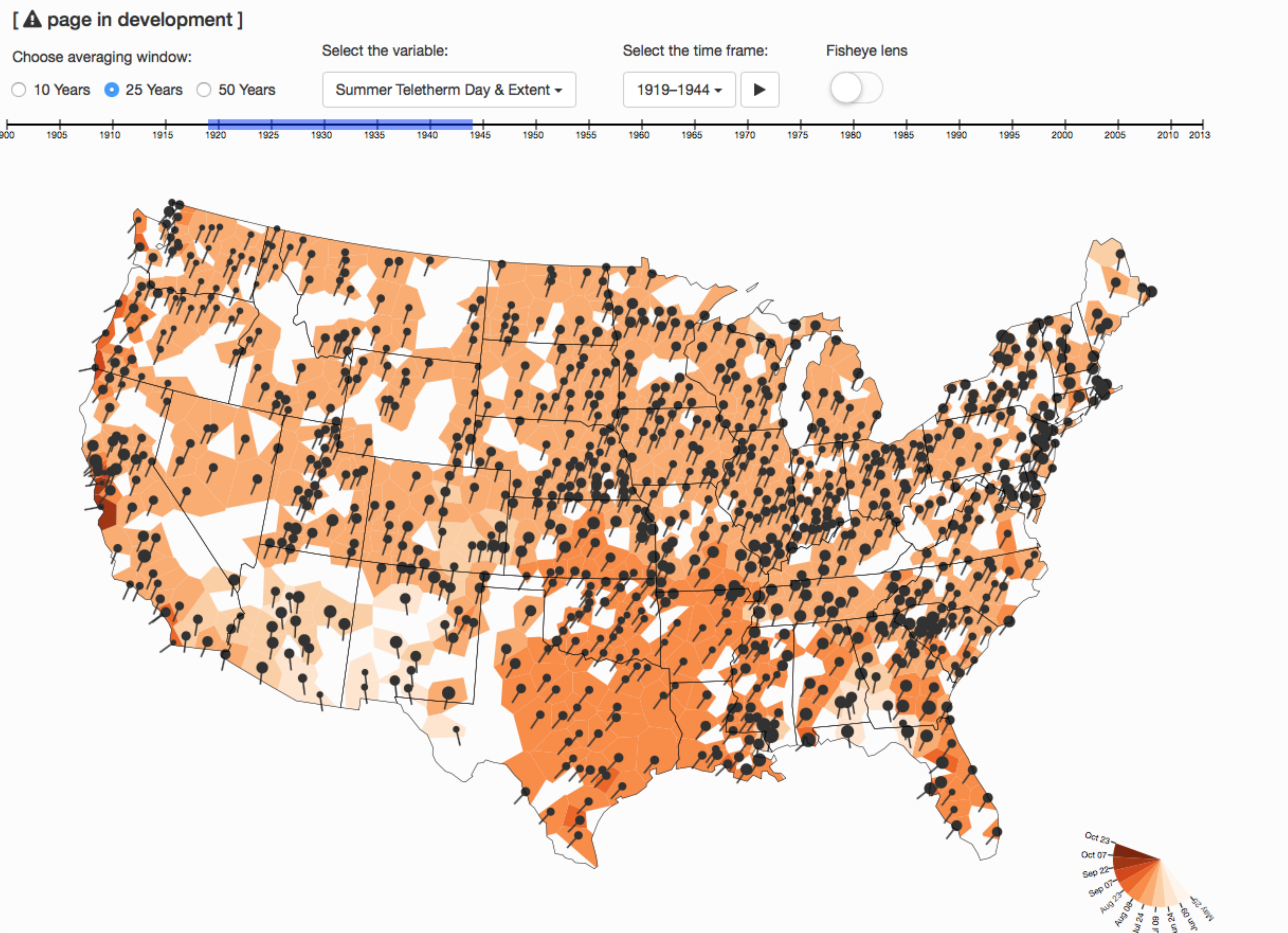}
  \caption[]{
    Interactive teletherm map with time and variable controls. Select between the teletherm day \& extent and teletherm temperature, the averaging window to compute the teletherms, and the time to show on the map.
    A linear color scale, ``oranges'', is shown for teletherm day and extent.
    A diverging color scale is shown for temperatures, inspired by \href{https://darksky.net}{https://darksky.net}.
    For each weather station, a tooltip hover shows details on demand.
  }
  \label{fig:teletherm-map}
\end{figure}

For this paper,
I built the online appendices and transformed the visualizations into online, interactive versions
at \href{http://teletherm.org/}{http://teletherm.org/} using D3 Javascript \citep{bostock2011d3}.
The online appendices are available at \href{http://compstorylab.org/share/papers/dodds2015c/index.html}{http://compstorylab.org/share/papers/dodds2015c/index.html}.
Maps of the United States are shown in Figure \ref{fig:teletherm-map}, with Voronoi cells for each
station colored in addition to the direction and color of the arrows used in
the static maps.
Other features of these online maps include the ability to animate through time,
select a fisheye lens for inspecting the map, and toggle between the various indicators
(Summer/Winter Teletherm day and temperature).

\begin{figure}[tbp!]
  \centering
  \includegraphics[width=0.96\textwidth]{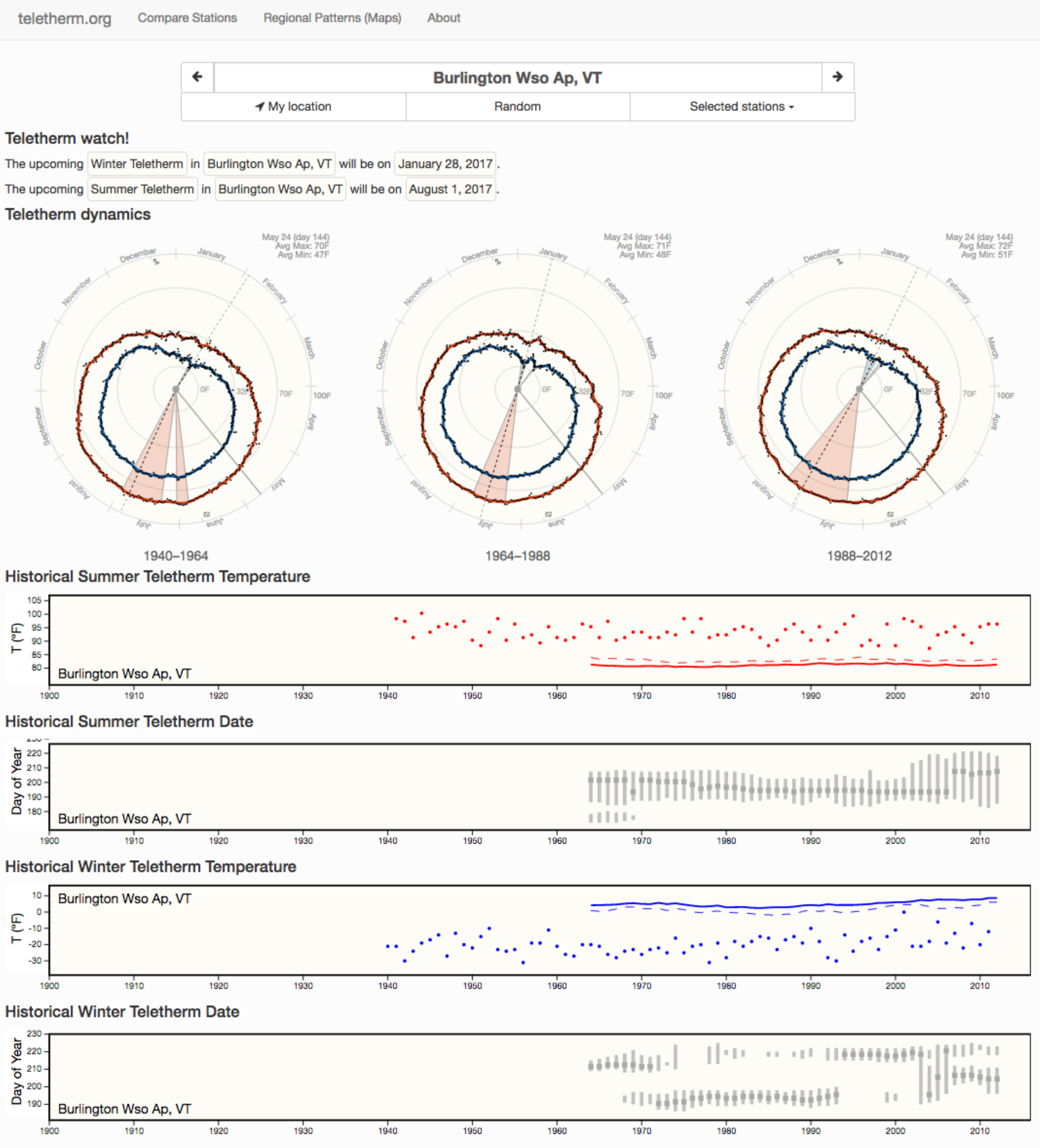}
  \caption[]{
    Teletherm dials shows the yearly temperature dynamics for a single location over a period of time, and time series below show the trends for both temperature extremes and teletherm dates.
    The min and max temperature for each day of the year are smoothed over three 25 year windows, one for each dial, and show in blue and red, respectively.
    As in the paper, the smoothed temperature is computed with a Gaussian kernel smoothing over the average min/max over days of the year.
    To avoid issues with the boundary, to compute the Gaussian kernel the temperature is wrapped on both ends of the year (with the same data).
    Summer and winter solstice are shown with icons, and the details of the day of year are shown in the upper right of each dial (over which the hover is linked between each dial---they all move together).
  }
  \label{fig:teletherm-splat}
\end{figure}

To realize the goals of this research, the website is designed to communicate
the patterns of Teletherm dynamics at both a local and a regional level.
In addition to building interactive versions of the US maps, I worked with
Professor Dodds to design novel visualizations for the individual station teletherm dynamics.
These plots are shown in Figure \ref{fig:teletherm-splat}, and accompany visualizations of
the time dynamics of Teletherm days, extends, and temperatures.
The online source code repository is publicly available at \href{https://github.com/andyreagan/teletherm.org}{https://github.com/andyreagan/teletherm.org}.

\clearpage

\section{Divergent Discourse Between Protests and Counter-Protests: \#BlackLivesMatter and \#AllLivesMatter}

Paper number 10 is \textit{Divergent Discourse Between Protests and Counter-Protests: \#BlackLivesMatter and \#AllLivesMatter} by Ryan J. Gallagher, Andrew J. Reagan, Christopher M. Danforth, and Peter Sheridan Dodds, cited as \cite{gallagher2016divergent}.

\subsection{Abstract}

\begin{quote}
  Since the shooting of Black teenager Michael Brown by White police officer Darren Wilson in Ferguson, Missouri,
the protest hashtag \#BlackLivesMatter has amplified critiques of extrajudicial killings of Black Americans.
In response to \#BlackLivesMatter,
other Twitter users have adopted \#AllLivesMatter,
a counter-protest hashtag whose content argues that equal attention should be given to all lives regardless of race.
Through a multi-level analysis,
we study how these protests and counter-protests diverge by quantifying aspects of their discourse.
In particular,
we introduce methodology that not only quantifies these divergences,
but also reveals whether they are from widespread discussion or a few popular retweets within these groups.
We find that \#BlackLivesMatter exhibits many information rich conversations,
while those within \#AllLivesMatter are more muted and susceptible to hijacking.
We also show that the discussion within \#BlackLivesMatter is more likely to center around the deaths of Black Americans,
while that of \#AllLivesMatter is more likely to sympathize with the lives of police officers and express politically conservative views.
\end{quote}

\subsection{Contribution}

My main contribution to this paper was working closely with lead author Ryan Gallagher to collect the data from our Twitter database on the VACC.
We collected data for a number of hashtags, specifically all of the following:

\begin{python}
keywords = [{"re": re.compile(r"#blacklivesmatter\b",flags=re.IGNORECASE)}},
            {"re": re.compile(r"#alllivesmatter\b",flags=re.IGNORECASE)}},
            {"re": re.compile(r"#bluelivesmatter\b",flags=re.IGNORECASE)},
            {"re": re.compile(r"#policelivesmatter\b",flags=re.IGNORECASE)},
            {"re": re.compile(r"#michaelbrown\b",flags=re.IGNORECASE)},
            {"re": re.compile(r"#ferguson\b",flags=re.IGNORECASE)},
            {"re": re.compile(r"#freddiegray\b",flags=re.IGNORECASE)},
            {"re": re.compile(r"#ericgarner\b",flags=re.IGNORECASE)},
            {"re": re.compile(r"#icantbreathe\b",flags=re.IGNORECASE)},
            {"re": re.compile(r"#sarahbland\b",flags=re.IGNORECASE)},
            {"re": re.compile(r"#templeton\b",flags=re.IGNORECASE)},]
\end{python}

After collect the Tweets for these hashtags, they were reorganized by user, and then collected into a \verb|sqlite| database using Django, a Python web framework.
This web framework was then used to go back and collect the most recent 3,200 Tweets from each public Twitter account that we had found in our initial search.
The collection ended on Nov 25th, 2015, so these Tweets were the 3,200 most recent as of that date.
From this data, we were able to construct the social networks for analysis of the dynamics of these online communities.

\chapter{Conclusion}
\chaptermark{Conclusion}
\label{chap:conclusion}

\section{Future directions}

First we take a look to the future research around sentiment analysis, emotional arcs, and the related projects we covered in Chapter \ref{chap:contributions}.

\subsection{Sentiment analysis}

Our work looked in detail at dictionary-based sentiment analysis methodology, focusing on the use of these methods in qualitative and quantitative analysis.
Immediate directions for the extension of dictionary based methods can examine the creation and use of dictionaries that offer (1) many emotions (Section \ref{subsec:psychology}), (2) MWEs (Section~\ref{subsec:mwe}), (3) multiple word senses (Section~\ref{subsec:wsd}), and (4) corpus-specific tuning.
We reviewed automated methods to build corpus-specific dictionaries in Section \ref{subsec:prev-propagation}, and while most approaches are low precision, we identified directions for that provide the highest precision and recall.
Combining automated (machine learning, propagation-based) approaches with MWEs, word senses, and many emotions will provide many opportunities for the study of the sentiment properties of language and the improvement of sentiment analysis.

In addition to the improvement of the dictionaries, many unanswered questions remain around the visualization of sentiment analysis measures.
We reviewed some approaches in Section \ref{subsec:vis} and reiterate that future work can (1) incorporate task-specific usability testing \citep{munzner2014visualization}, (2) visualize non-linear features \citep{ribeiro2016should}, and (3) continue to build more tools that enable other researchers to make use of visualization.

\subsection{Emotional arcs}

Here we enumerate some directions for research on emotional arcs in addition those mentioned at the end of Chapter 3 (see Section \ref{sec:arcs-conclusion}).

The emotional arcs of movies could be considered as a feature driving once controversial movies towards normalization over time, a closer examination of the trend presented by \cite{amendola2015evolving}.
Various studies have examined the changes in the valence of language over time, and in a similar fashion this will be possible to see how the emotional trajectories of stories has changed.

The emotional arc of a book can be used to predict the Library of Congress classification, using fiction and non-fiction separately to demonstrate the applicability of emotional arcs.
In particular, one could feed the coefficient vector from the SVD projection for the first $n$ modes into a predictor and see how much predictive power is contained in each mode, and exploring $n$ can provide additional testing of how explanatory the first 6 modes are.
Clustering on the emotional arc embedding vector would show whether these groups can be separated in a purely unsupervised manner.

Extending the approach of \cite{bamman2015validity} and the validation shown in Figure \ref{fig:bamman-validation}, it will remain important to keep people in the loop of the analysis of emotional arcs, since it is our reaction to stories that is being measured.
A follow-up project to our work on emotional arcs could build a more complete user study to examine the human aspect of emotion in narrative more directly.

We broadly examined the the forefront of NLP research (Section \ref{subsec:nlp}), and can use the advancing methods to answer such questions as ``is a character good or bad?''.
The analysis of character networks (Section \ref{subsec:characters}) will continue to improve with identification of the nature of relationships, and the events for particular characters (e.g., birth, marriage, death, and the associated sentiments).

Connecting the scripts, frames, and SIG-like approaches (see Section~\ref{subsec:SIG-diagrams} and Section~\ref{subsec:frames-NLP}) to narrative more directly to the emotional arcs will be provide a finer-grained emotional arc representation, connected to the events in a narrative.
This approach will in-part realize the jump from a bag-of-words to a bag-of-stories approach to natural language.
As neural network approaches pust the state-of-the-art in NLP, there may be utility to considering architectures that have an explicit representations of abstraction levels.
This approach is analagous to the Convolution Neural Network (CNN) architecture that has proven successful in image recognition tasks.
An example structure to build upon is the Historical Thesaurus of English \citep{kay2009historical}, as is done by \cite{alexander2015metaphor}.
In contrast to this proposed approach, the ``automatic'' feature selection (magic) of neural networks remains powerful \citep{1704.01444}.

\subsection{Other projects}

We have shown that it is possible to build population scale measures of well-being and public health.
The Hedonometer and the Lexicocalorimeter can be utilized as only two of many broad measures that extend our dashboard of societal indicators; such additional  ``meters'' of general interest that the Computational StoryLab has considered include such tools as an ``insomniometer''.
Considering the Lexicocalorimeter, taking these lexical meters from snapshot-in-time analysis to real-time feeds remains a difficult challenge that has been accomplished with \href{http://hedonometer.org/}{http://hedonometer.org/} and can be extended to additional meters.

There are many improvements possible for the visualizations hosted online at \href{http://teletherm.org/}{http://teletherm.org/}.
The teletherm animations can be improved through the use of the \verb|d3.timer| module for smoother animation.
Voronoi cells on the map are clipped at the boundary of the contiguous United States using a clipping mask that contains all 50 states as individual paths, and this does not work reliably in Google Chrome.
More issues for improvement are noted in the ``issues'' tab of the online source code repository at \href{https://github.com/andyreagan/teletherm.org}{https://github.com/andyreagan/teletherm.org}.
In addition, it will be possible to extend the teletherm project to incorporate temperature data from across the world.

\section{Parting thoughts}

Narratives are not unique in their explanation of causal links between events, and often the ``adjacent narratives'' are in direct competition.
We saw in Section \ref{subsec:frames-NLP} that the the disambiguation of competing event chains is an active area of NLP research.
This is identified as one factor contributing to information overload on the Internet \citep{orman2015information}, and participating in a collective cognitive denial of service attack \citep{king2016chinese}.
We are biased to seeing the world through narratives that have the most support from our existing experiences.
Embodied in the principle of Occam's Razor, we often prefer stories that are the simplest.
This premise is explored anecdotally \citep{storr2014unpersuadables}, and the competition between competing narratives is a new avenue for computation study.

The use of narratives in science belies an understanding of natural phenomena through metaphor, the consequences of stories in science has been examined by \cite{mahoney2006tale,levy2008case,collier2011understanding,gelman2014stories}.
Narrative itself has been in the spotlight, being put forth to frame the decisions of economists in times of crisis and related to the political functions of democratic elections \citep{shriller2017narrative}.

Every-day causality and personal narrative build upon a fundamental assumption of personal agency and free will.
Post-hoc rationalization is only useful to explain behavior that was intentional.
Deterministic laws of physics are at odds with this worldview, but the science of complex systems has shown us that systems at different levels can exhibit emergent behavior that cannot be predicted from lower level interactions \citep{anderson1972a}.
Applying computational thinking to the human concepts of metaphor and narrative can force us to further elucidate these distinctions and provide us with a deeper understanding of the world around us as we see it.

\pagebreak
\singlespacing
\addcontentsline{toc}{chapter}{Bibliography}
\bibliographystyle{chicago}

\doublespacing
\appendix
\addappheadtotoc

\titleformat{\chapter}[hang]
{\normalfont\huge\bfseries}{\chaptertitlename\ \thechapter:}{1em}{}

\chapter{Supplementary Material for Sentiment Dictionary Comparisons}

\section{S1 Appendix: Computational methods} \label{supp:methods}

All of the code to perform these tests is available and document on GitHub.
The repository can be found here: \url{https://github.com/andyreagan/sentiment-analysis-comparison}.

\subsection{Stem matching} \label{supp:stems}

Of the dictionaries tested, both LIWC and MPQA use ``word stems''.
Here we quickly note some of the technical difficulties with using word stems, and how we processed them, for future research to build upon and improve.

An example is \verb|abandon*|, which is intended to the match words of the standard RE form \verb|abandon[a-z]*|.
A naive approach is to check each word against the regular expression, but this is prohibitively slow.
We store each of the dictionaries in a ``trie'' data structure with a record.
We use the easily available ``marisa-trie'' Python library, which wraps the C++ counterpart.
The speed of these libraries made the comparison possible over large corpora, in particular for the dictionaries with stemmed words, where the \verb|prefix| search is necessary.
Specifically, the ``trie'' structure is 70 times faster than a regular expression based search for stem words.
In particular, we construct two tries for each dictionary: a fixed and stemmed trie.
We first attempt to match words against the fixed list, and then turn to the prefix match on the stemmed list.

\subsection{Regular expression parsing}
\label{subsec:regex-parsing}

The first step in processing the text of each corpora is extracting the words from the raw text.
Here we rely on a regular expression search, after first removing some punctuation.
We choose to include a set of all characters that are found within the words in each of the six dictionaries tested in detail, such that it respects the parse used to create these dictionaries by retaining such characters.
This takes the following form in Python, for \verb|raw_text| as a string:
\begin{verbatim}
punctuation_to_replace = ["---","--","''"]
for punctuation in punctuation_to_replace:
    raw_text = raw_text.replace(punctuation," ")
words = [x.lower() for x in re.findall(\
    r"""(?:[0-9][0-9,\.]*[0-9])|
(?:http://[\w\./\-\?\&\#]+)|
(?:[\w\@\#\'\&\]\[]+)|
(?:[b}/3D;p)|’\-@x#^_0\\P(o:O{X$[=<>\]*B]+)""",
    raw_text,flags=re.UNICODE)]

\end{verbatim}

\clearpage
\pagebreak

\section{S2 Appendix: Continued individual comparisons} \label{supp:comparisons}

Picking up right where we left off in Section \ref{sec:results}, we next compare ANEW with the other dictionaries.
The ANEW-WK comparison in Panel I of Fig. \ref{fig:main} contains all 1030 words of ANEW, with a fit of $h_{\text{ANEW}}(w) = 1.07*h_{\text{WK}} (w)-0.30$, making ANEW more positive and with increasing positivity for more positive words.
The 20 most different scores are (ANEW,WK): fame (7.93,5.45), god (8.15,5.90), aggressive (5.10,3.08), casino (6.81,4.68), rancid (4.34,2.38), bees (3.20,5.14), teacher (5.68,7.37), priest (6.42,4.50), aroused (7.97,5.95), skijump (7.06,5.11), noisy (5.02,3.21), heroin (4.36,2.74), insolent (4.35,2.74), rain (5.08,6.58), patient (5.29,6.71), pancakes (6.08,7.43), hospital (5.04,3.52), valentine (8.11,6.40), and book (5.72,7.05).
We again see some of the same words from the LabMT comparisons with these dictionaries, and again can attribute some differences to small sample sizes and differing demographics.

For the ANEW-MPQA comparison in Panel J of Fig. \ref{fig:main} we show the same matched word lists as before.
The happiest 10 words in ANEW matched by MPQA are: clouds (6.18), bar (6.42), mind (6.68), game (6.98), sapphire (7.00), silly (7.41), flirt (7.52), rollercoaster (8.02), comedy (8.37), laughter (8.45).
The least happy 5 neutral words and happiest 5 neutral words in MPQA, matched with MPQA, are: pressure (3.38), needle (3.82), quiet (5.58), key (5.68), alert (6.20), surprised (7.47), memories (7.48), knowledge (7.58), nature (7.65), engaged (8.00), baby (8.22).
The least happy words in ANEW with score +1 in MPQA that are matched by MPQA are: terrified (1.72), meek (3.87), plain (4.39), obey (4.52), contents (4.89), patient (5.29), reverent (5.35), basket (5.45), repentant (5.53), trumpet (5.75).
Again we see some very questionable matches by the MPQA dictionary, with broad stems capturing words with both positive and negative scores.

For the ANEW-LIWC comparison in Panel K of Fig. \ref{fig:main} we show the same matched word lists as before.
The happiest 10 words in ANEW matched by LIWC are: lazy (4.38), neurotic (4.45), startled (4.50), obsession (4.52), skeptical (4.52), shy (4.64), anxious (4.81), tease (4.84), serious (5.08), aggressive (5.10).
There are only 5 words in ANEW that are matched by LIWC with LIWC score of 0: part (5.11), item (5.26), quick (6.64), couple (7.41), millionaire (8.03).
The least happy words in ANEW with score +1 in LIWC that are matched by LIWC are: heroin (4.36), virtue (6.22), save (6.45), favor (6.46), innocent (6.51), nice (6.55), trust (6.68), radiant (6.73), glamour (6.76), charm (6.77).

For the ANEW-Liu comparison in Panel L of Fig. \ref{fig:main} we show the same matched word lists as before, except the neutral word list because Liu has no explicit neutral words.
The happiest 10 words in ANEW matched by Liu are: pig (5.07), aggressive (5.10), tank (5.16), busybody (5.17), hard (5.22), mischief (5.57), silly (7.41), flirt (7.52), rollercoaster (8.02), joke (8.10).
The least happy words in ANEW with score +1 in Liu that are matched by Liu are: defeated (2.34), obsession (4.52), patient (5.29), reverent (5.35), quiet (5.58), trumpet (5.75), modest (5.76), humble (5.86), salute (5.92), idol (6.12).

For the WK-MPQA comparison in Panel P of Fig. \ref{fig:main} we show the same matched word lists as before.
The happiest 10 words in WK matched by MPQA are: cutie (7.43), pancakes (7.43), panda (7.55), laugh (7.56), marriage (7.56), lullaby (7.57), fudge (7.62), pancake (7.71), comedy (8.05), laughter (8.05).
The least happy 5 neutral words and happiest 5 neutral words in MPQA, matched with MPQA, are: sociopath (2.44), infectious (2.63), sob (2.65), soulless (2.71), infertility (3.00), thinker (7.26), knowledge (7.28), legacy (7.38), surprise (7.44), song (7.59).
The least happy words in WK with score +1 in MPQA that are matched by MPQA are: kidnapper (1.77), kidnapping (2.05), kidnap (2.19), discriminating (2.33), terrified (2.51), terrifying (2.63), terrify (2.84), courtroom (2.84), backfire (3.00), indebted (3.21).

For the WK-LIWC comparison in Panel Q of Fig. \ref{fig:main} we show the same matched word lists as before.
The happiest 10 words in WK matched by LIWC are: geek (5.56), number (5.59), fiery (5.70), trivia (5.70), screwdriver (5.76), foolproof (5.82), serious (5.88), yearn (5.95), dumpling (6.48), weeping willow (6.53).
The least happy 5 neutral words and happiest 5 neutral words in LIWC, matched with LIWC, are: negative (2.52), negativity (2.74), quicksand (3.62), lack (3.68), wont (4.09), unique (7.32), millionaire (7.32), first (7.33), million (7.55), rest (7.86).
The least happy words in WK with score +1 in LIWC that are matched by LIWC are: heroin (2.74), friendless (3.15), promiscuous (3.32), supremacy (3.48), faithless (3.57), laughingstock (3.77), promiscuity (3.95), tenderfoot (4.26), succession (4.52), dynamite (4.79).

For the WK-Liu comparison in Panel R of Fig. \ref{fig:main} we show the same matched word lists as before, except the neutral word list because Liu has no explicit neutral words.
The happiest 10 words in WK matched by Liu are: goofy (6.71), silly (6.72), flirt (6.73), rollercoaster (6.75), tenderness (6.89), shimmer (6.95), comical (6.95), fanciful (7.05), funny (7.59), fudge (7.62), joke (7.88).
The least happy words in WK with score +1 in Liu that are matched by Liu are: defeated (2.59), envy (3.05), indebted (3.21), supremacy (3.48), defeat (3.74), overtake (3.95), trump (4.18), obsession (4.38), dominate (4.40), tough (4.45).

Now we'll focus our attention on the MPQA row, and first we see comparisons against the three full range dictionaries.
For the first match against LabMT in Panel D of Fig. \ref{fig:main}, the MPQA match catches 431 words with MPQA score 0, while LabMT (without stems) matches 268 words in MPQA  in Panel S (1039/809 and 886/766 for the positive and negative words of MPQA).
Since we've already highlighted most of these words, we move on and focus our attention on comparing the $\pm 1$ dictionaries.

In Panels V--X, BB--DD, and HH--JJ of Fig. \ref{fig:main} there are a total of 6 bins off of the diagonal, and we focus out attention on the bins that represent words that have opposite scores in each of the dictionaries.
For example, consider the matches made my MPQA in Panel BB: the words in the top left corner and bottom right corner with are scored in a opposite manner in LIWC, and are of particular concern.
Looking at the words from Panel W with a +1 in MPQA and a -1 in LIWC (matched by LIWC) we see: stunned, fiery, terrified, terrifying, yearn, defense, doubtless, foolproof, risk-free, exhaustively, exhaustive, blameless, low-risk, low-cost, lower-priced, guiltless, vulnerable, yearningly, and yearning.
The words with a -1 in MPQA that are +1 in LIWC (matched by LIWC) are: silly, madly, flirt, laugh, keen, superiority, supremacy, sillily, dearth, comedy, challenge, challenging, cheerless, faithless, laughable, laughably, laughingstock, laughter, laugh, grating, opportunistic, joker, challenge, flirty.

In Panel W of \ref{fig:main}, the words with a +1 in MPQA and a -1 in Liu (matched by Liu) are: solicitude, flair, funny, resurgent, untouched, tenderness, giddy, vulnerable, and joke.
The words with a -1 in MPQA that are +1 in Liu, matched by Liu, are: superiority, supremacy, sharp, defeat, dumbfounded, affectation, charisma, formidable, envy, empathy, trivially, obsessions, and obsession.

In Panel BB of \ref{fig:main}, the words with a +1 in LIWC and a -1 in MQPA (matched by MPQA) are: silly, madly, flirt, laugh, keen, determined, determina, funn, fearless, painl, cute, cutie, and gratef.
The words with a -1 in LIWC and a +1 in MQPA, that are matched by MPQA, are: stunned, terrified, terrifying, fiery, yearn, terrify, aversi, pressur, careless, helpless, and hopeless.

In Panel DD of \ref{fig:main}, the words with a -1 in LIWC and a +1 in Liu, that are matched by Liu, are: silly, and madly.
The words with a +1 in LIWC and a -1 in Liu, that are matched by Liu, are: stunned, and fiery.

In Panel HH of \ref{fig:main}, the words with a -1 in Liu and a +1 in MPQA, that are matched by MPQA, are: superiority, supremacy, sharp, defeat, dumbfounded, charisma, affectation, formidable, envy, empathy, trivially, obsessions, obsession, stabilize, defeated, defeating, defeats, dominated, dominates, dominate, dumbfounding, cajole, cuteness, faultless, flashy, fine-looking, finer, finest, panoramic, pain-free, retractable, believeable, blockbuster, empathize, err-free, mind-blowing, marvelled, marveled, trouble-free, thumb-up, thumbs-up, long-lasting, and viewable.
The words with a +1 in Liu and a -1 in MPQA, that are matched by MPQA, are: solicitude, flair, funny, resurgent, untouched, tenderness, giddy, vulnerable, joke, shimmer, spurn, craven, aweful, backwoods, backwood, back-woods, back-wood, back-logged, backaches, backache, backaching, backbite, tingled, glower, and gainsay.

In Panel II of \ref{fig:main}, the words with a +1 in Liu and a -1 in LIWC, that are matched by LIWC, are: stunned, fiery, defeated, defeating, defeats, defeat, doubtless, dominated, dominates, dominate, dumbfounded, dumbfounding, faultless, foolproof, problem-free, problem-solver, risk-free, blameless, envy, trivially, trouble-free, tougher, toughest, tough, low-priced, low-price, low-risk, low-cost, lower-priced, geekier, geeky, guiltless, obsessions, and obsession.
The words with a -1 in Liu and a +1 in LIWC, that are matched by LIWC, are: silly, madly, sillily, dearth, challenging, cheerless, faithless, flirty, flirt, funnily, funny, tenderness, laughable, laughably, laughingstock, grating, opportunistic, joker, and joke.

In the off-diagonal bins for all of the $\pm 1$ dictionaries, we see many of the same words.
Again MPQA stem matches are disparagingly broad.
We also find matches by LIWC that are concerning, and should in all likelihood be removed from the dictionary.

\clearpage
\pagebreak

\section{S3 Appendix: Coverage for all corpuses} \label{supp:coverage}
We provide coverage plots for the other three corpuses.

\begin{figure*}[!htb]
\includegraphics[width=0.98\textwidth]{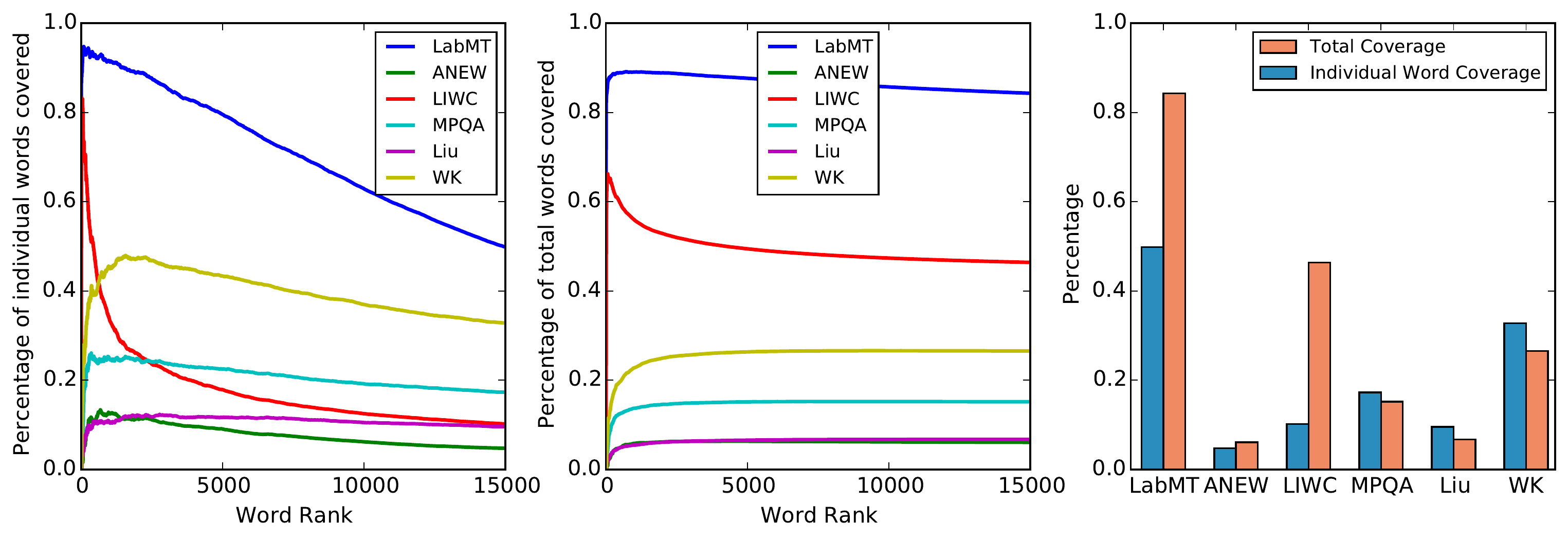}
    \caption[]{
      Coverage of the words on twitter by each of the dictionaries.
  }
  \label{fig:coverage_twitter}
\end{figure*}

\begin{figure*}[!htb]
\includegraphics[width=0.96\textwidth]{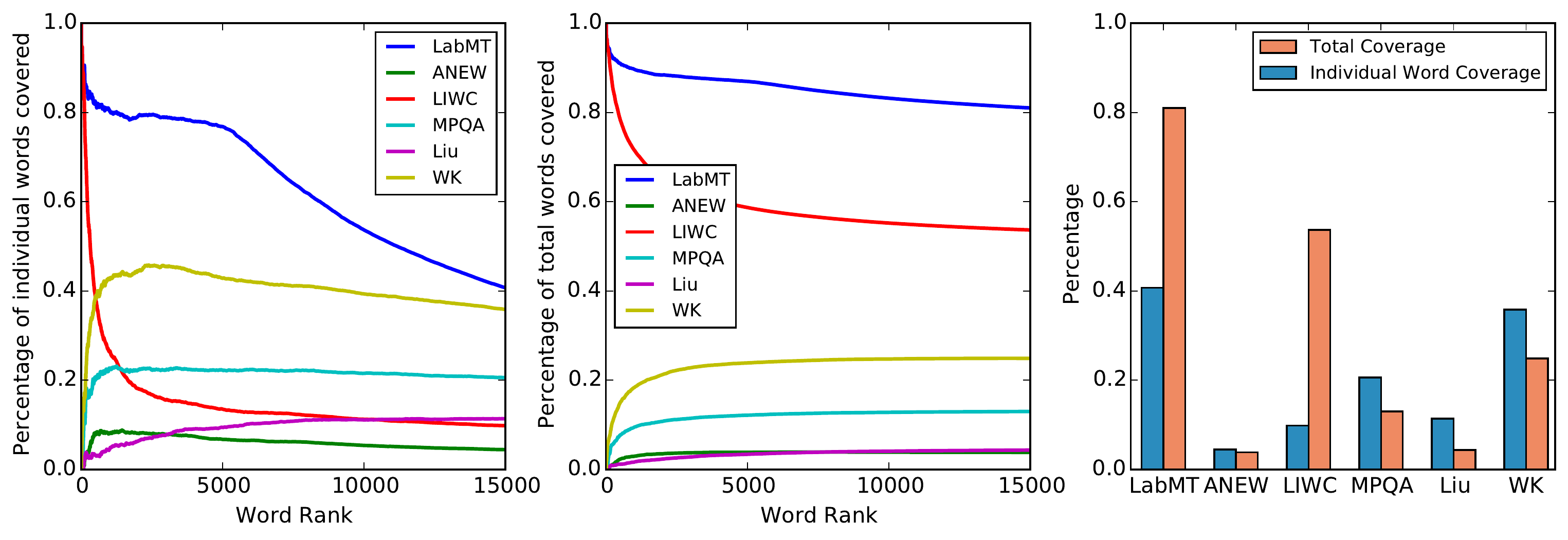}
    \caption[]{
    Coverage of the words in Google books by each of the dictionaries.
  }
  \label{fig:coverage_gbooks}
\end{figure*}

\begin{figure*}[!htb]
\includegraphics[width=0.96\textwidth]{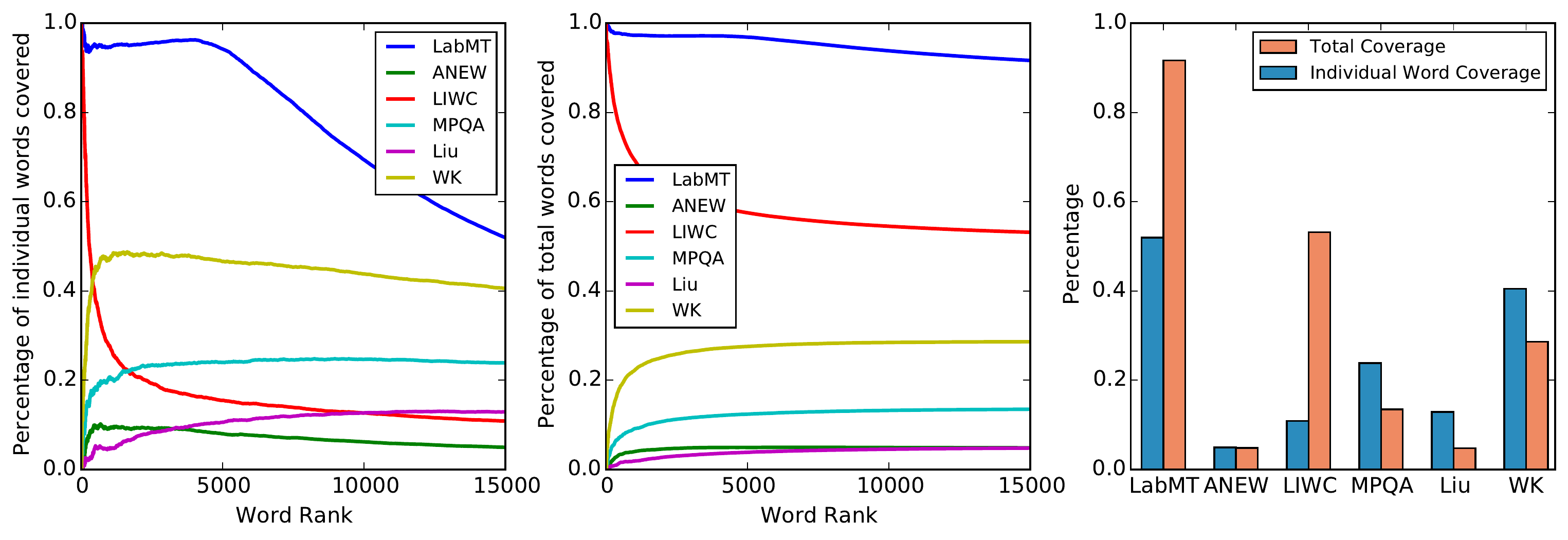}
    \caption[]{
    Coverage of the words in the New York Times by each of the dictionaries.
  }
  \label{fig:coverage_nyt}
\end{figure*}

\clearpage
\pagebreak

\section{S4 Appendix: Sorted New York Times rankings} \label{supp:nyt}

\begin{figure*}[!htb]
\includegraphics[width=0.98\textwidth]{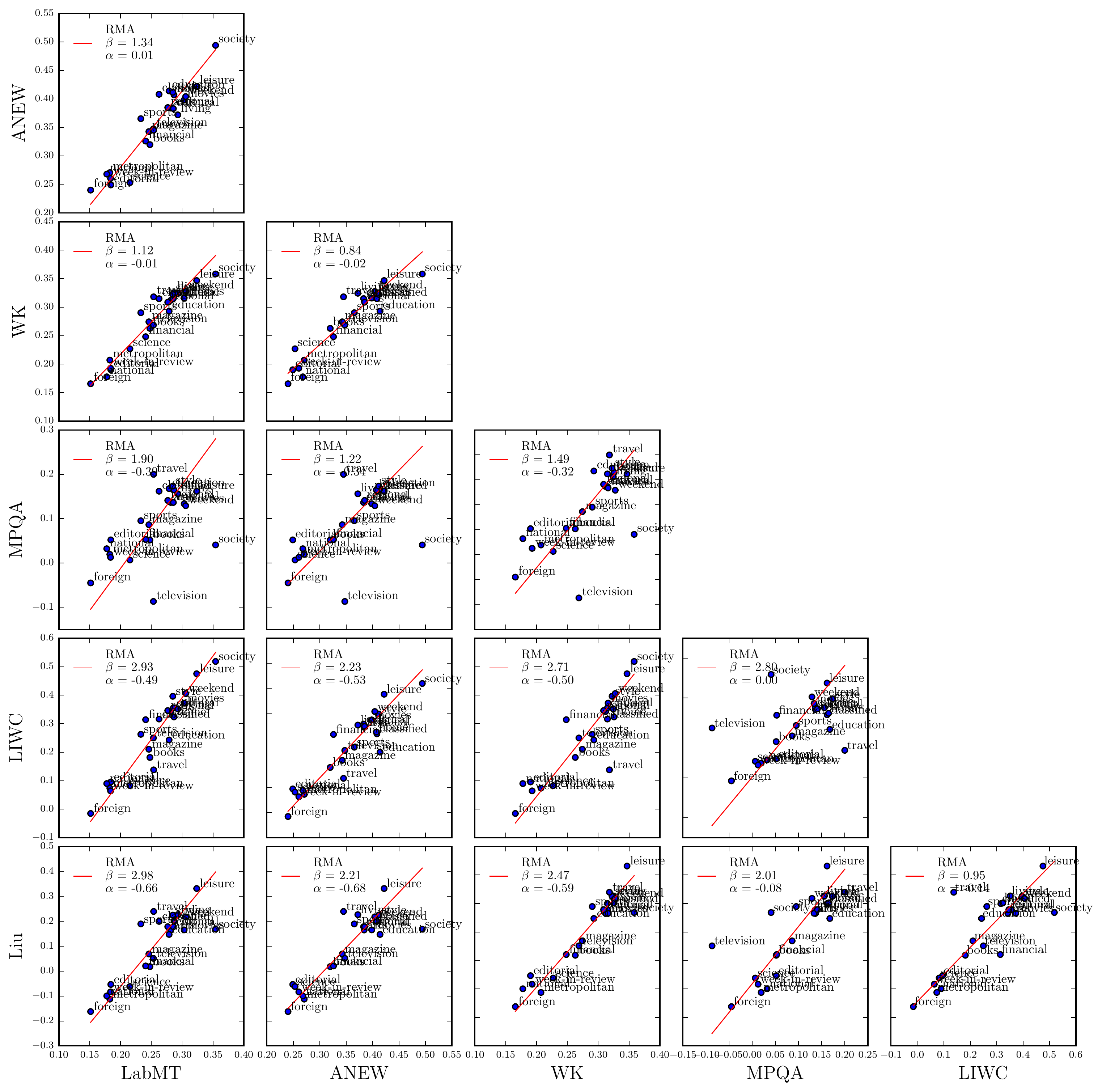}
    \caption[]{
      NYT Sections scatterplot.
      The RMA fit $\alpha$ and $\beta$ for the formula $y = \alpha + \beta x$.
      For the sake of comparison, we normalized each dictionary to the range [-.5,.5] by subtracting the mean score (5 or 0) and dividing by the range (8 or 2).
  }
  \label{fig:nyt_scatter_all}
\end{figure*}

\begin{figure*}[!htb]
\includegraphics[width=0.96\textwidth]{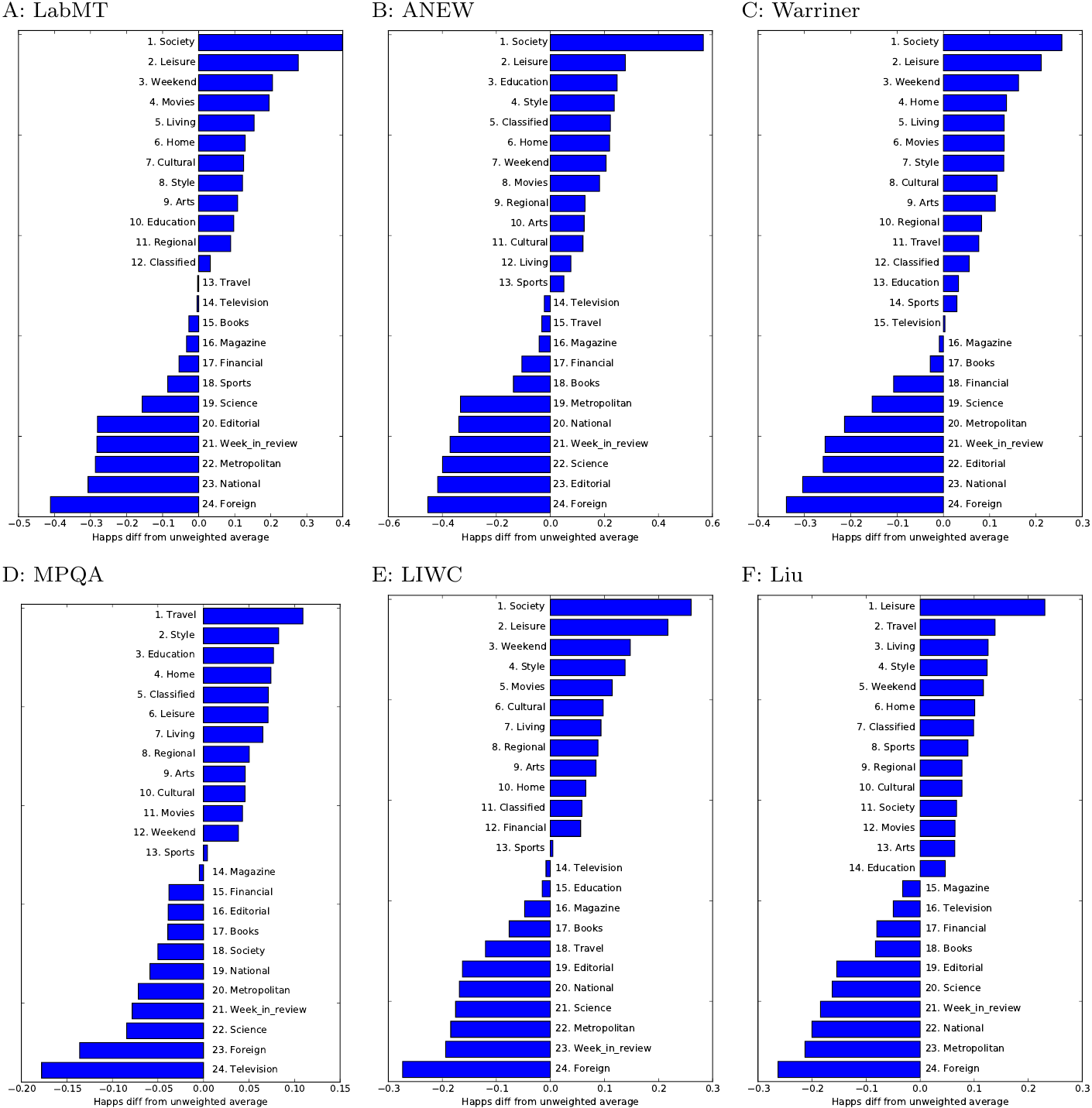}
    \caption[]{
    Sorted bar charts ranking each of the 24 New York Times Sections for each dictionary tested.
  }
  \label{fig:nyt_barcharts_all}
\end{figure*}

\clearpage
\pagebreak

\section{S5 Appendix: Movie Review Distributions} \label{supp:movies}

\begin{figure*}[!htb]
  \centering
\includegraphics[width=0.96\textwidth]{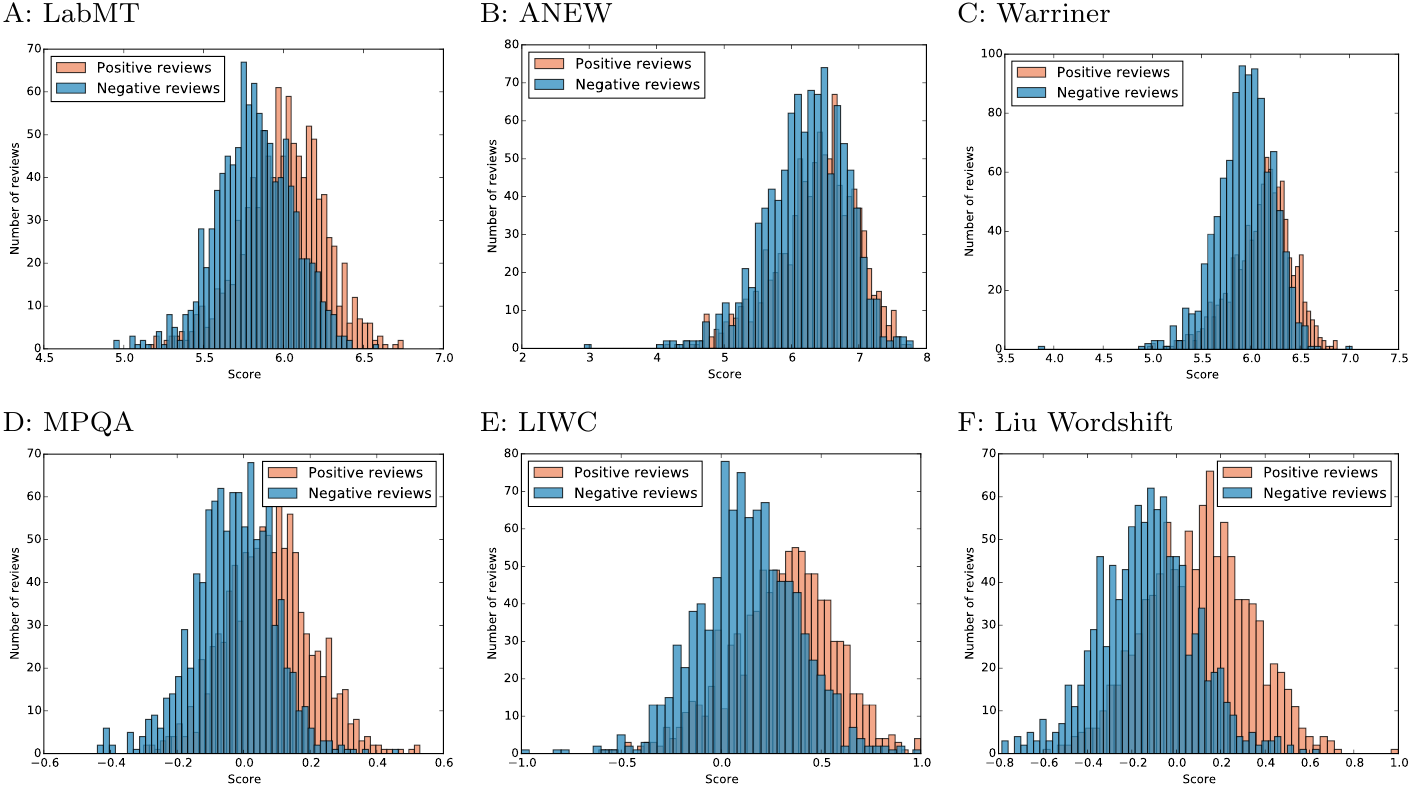}
  \caption[]{
    Binned scores for each review by each corpus with a stop value of $\Delta _h = 1.0$.
  }
  \label{fig:moviereview-dist-1}
\end{figure*}

Here we examine the distributions of movie review scores.
These distributions are each summarized by their mean and standard deviation in panels of Figure 2 for each dictionary.
For example, the left most error bar of each panel in Figure 2 shows the standard deviation around the mean for the distribution of individual review scores (Figure \ref{fig:moviereview-dist-1}).

\begin{figure*}[!htb]
  \centering
\includegraphics[width=0.96\textwidth]{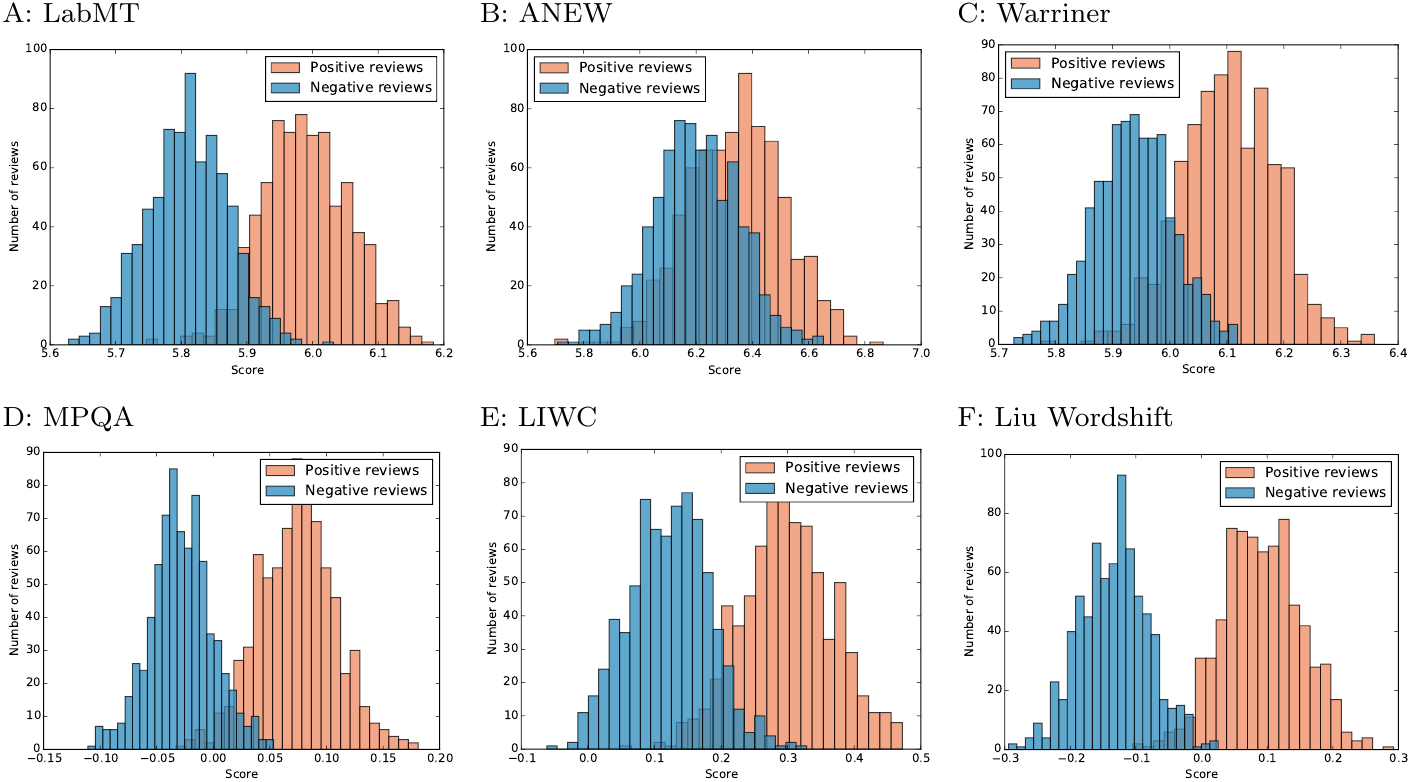}
  \caption[]{
    Binned scores for samples of 15 concatenated random reviews.
    Each dictionary uses stop value of $\Delta _h = 1.0$.
  }
  \label{fig:moviereview-dist-15}
\end{figure*}

\begin{figure*}[!htb]
  \centering
\includegraphics[width=0.45\textwidth]{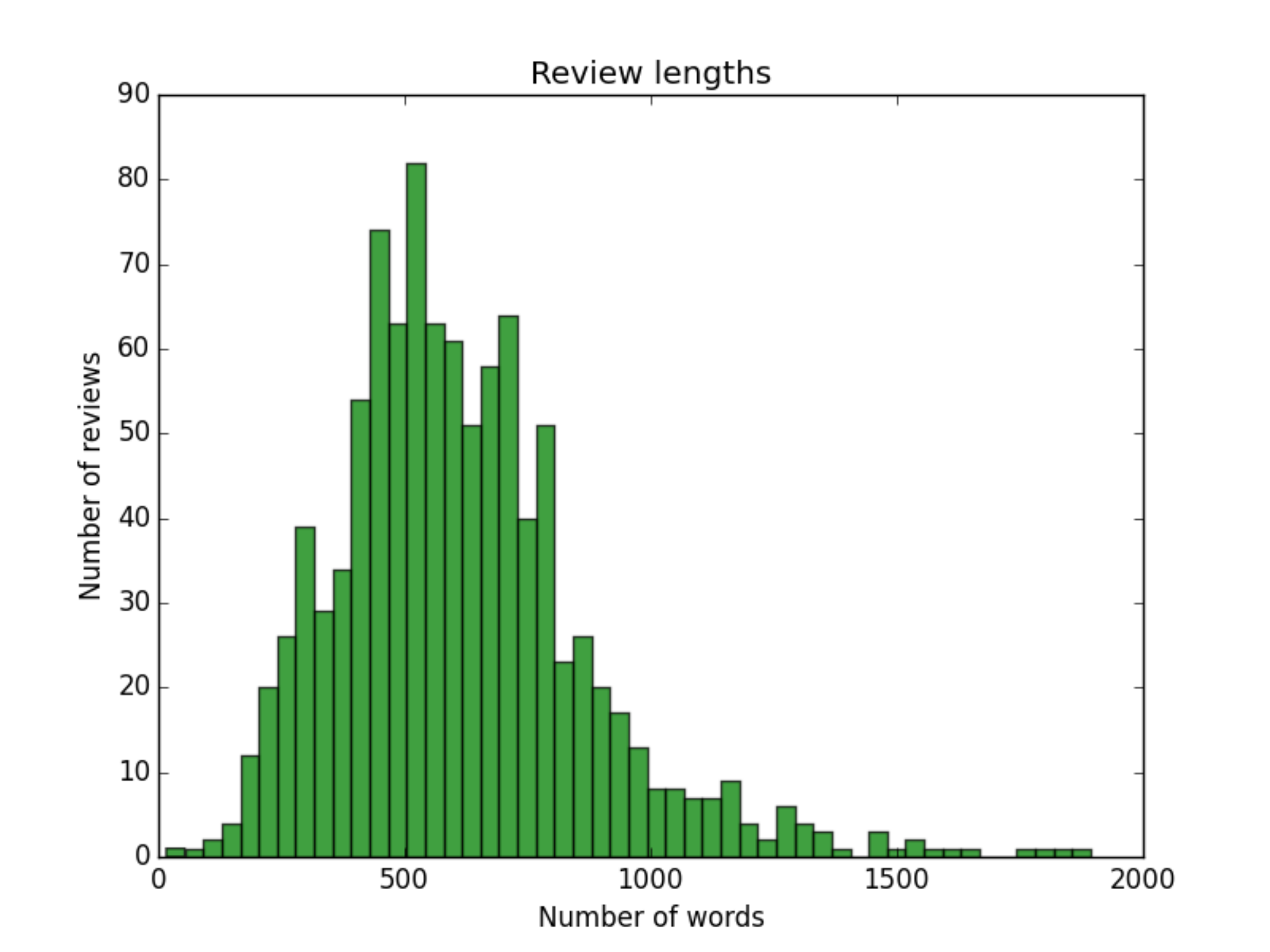}
  \caption[]{
    Binned length of positive reviews, in words.
  }
  \label{fig:moviereview-lengths}
\end{figure*}

\clearpage
\pagebreak

\section{S6 Appendix: Google Books correlations and word shifts} \label{supp:gbooks}

\begin{figure*}[!htb]
\includegraphics[width=0.48\textwidth]{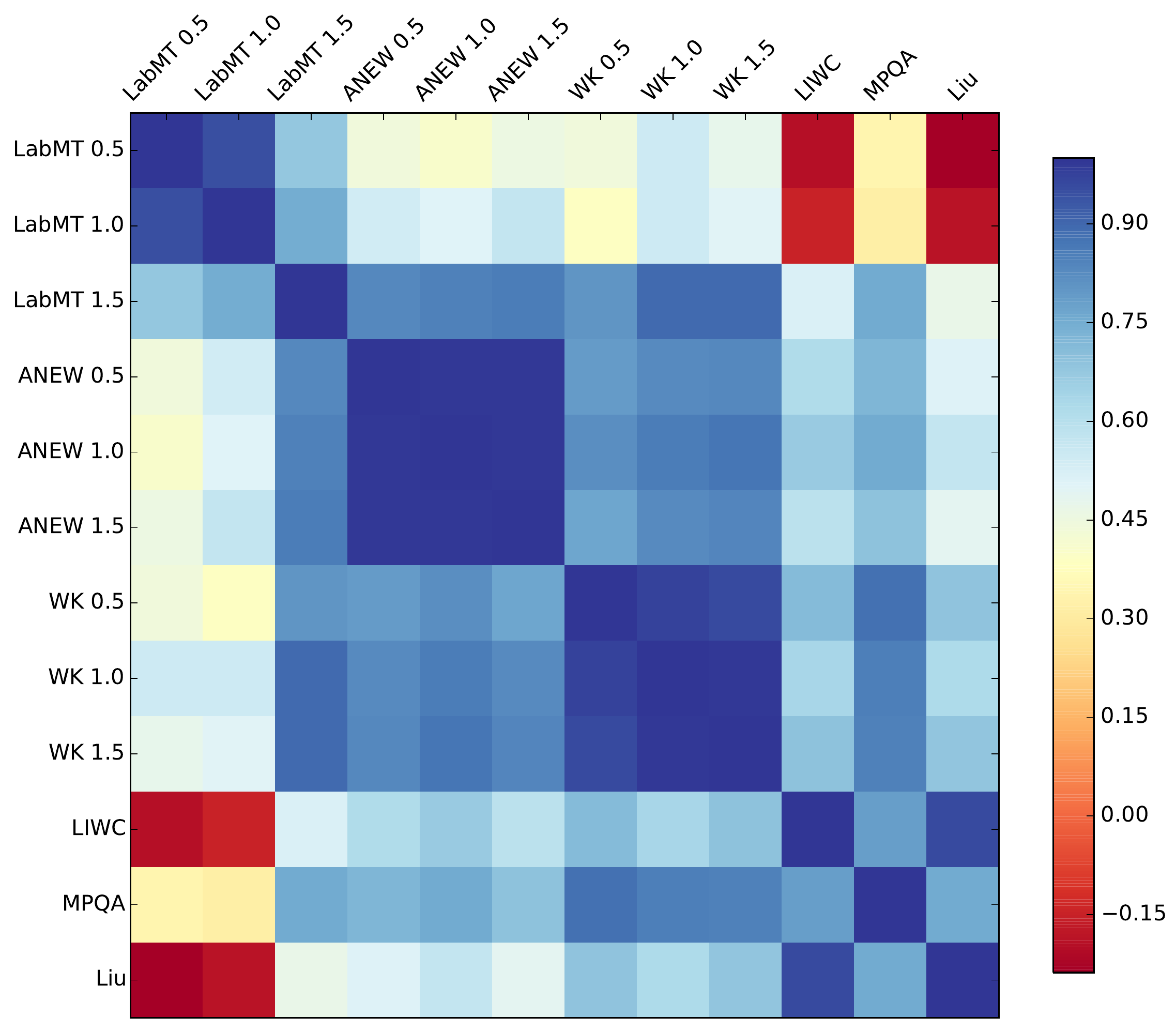}
    \caption[]{
    Google Books correlations.
    Here we include correlations for the google books time series, and word shifts for selected decades (1920's,1940's,1990's,2000's).

  }
  \label{fig:gbooks_correlations}
\end{figure*}

\begin{figure*}[!htb]
\includegraphics[width=0.98\textwidth]{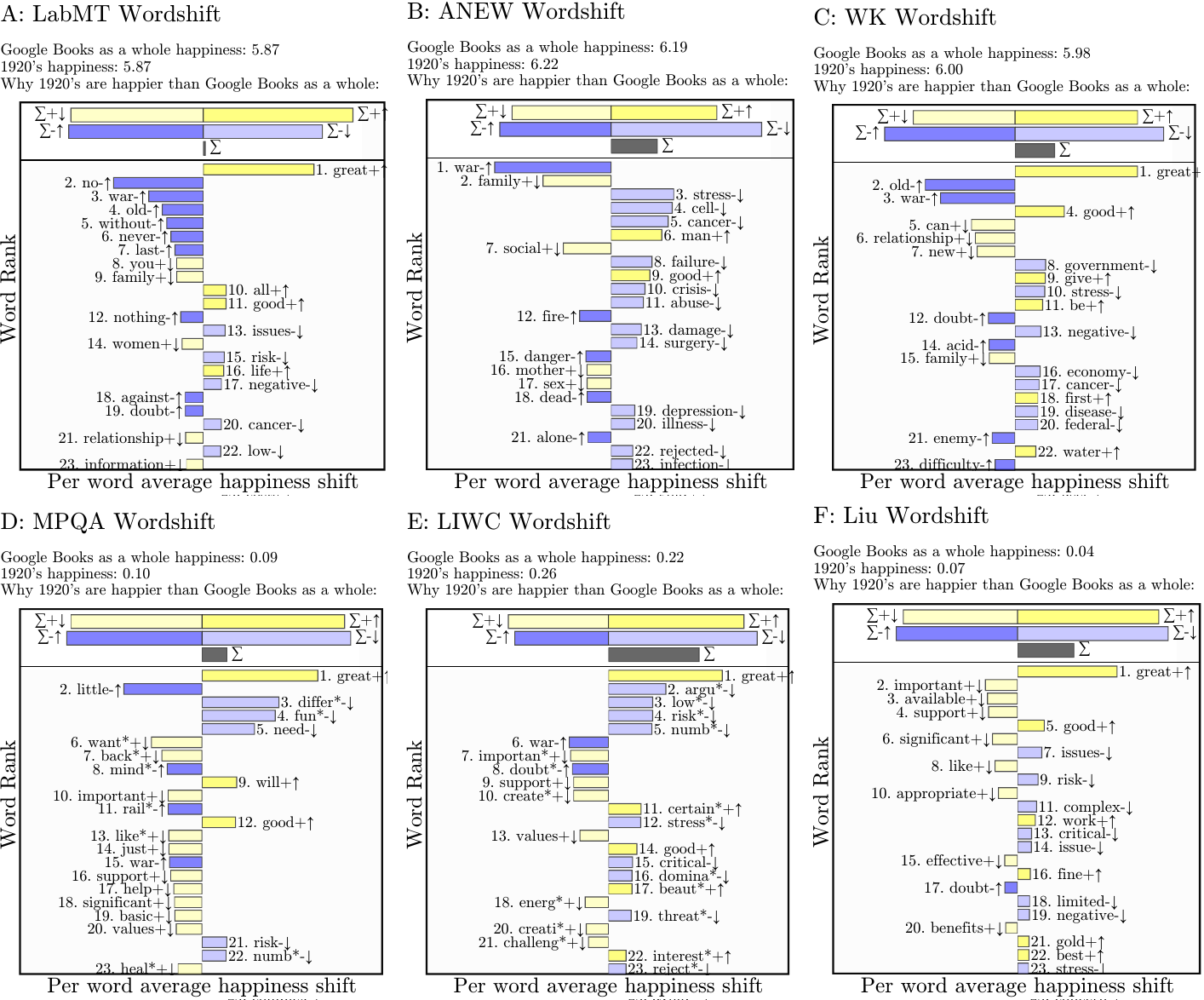}
  \caption[]{
    Google Books shifts in the 1920's against the baseline of Google Books.
  }
  \label{fig:gbooks-shifts-1920}
\end{figure*}

\begin{figure*}[!htb]
\includegraphics[width=0.98\textwidth]{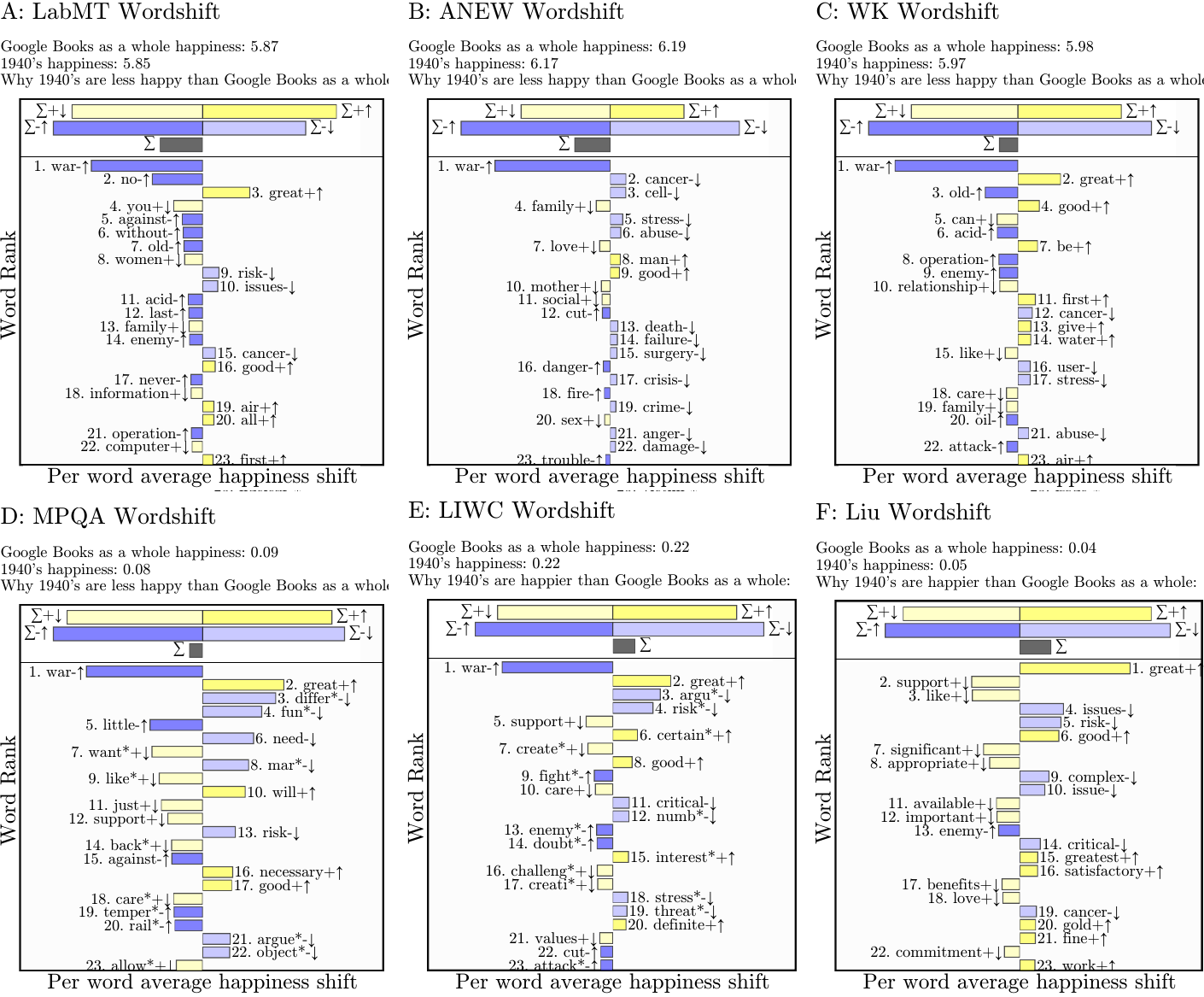}
  \caption[]{
    Google Books shifts in the 1940's against the baseline of Google Books.
  }
  \label{fig:gbooks-shifts-1940}
\end{figure*}

\begin{figure*}[!htb]
\includegraphics[width=0.98\textwidth]{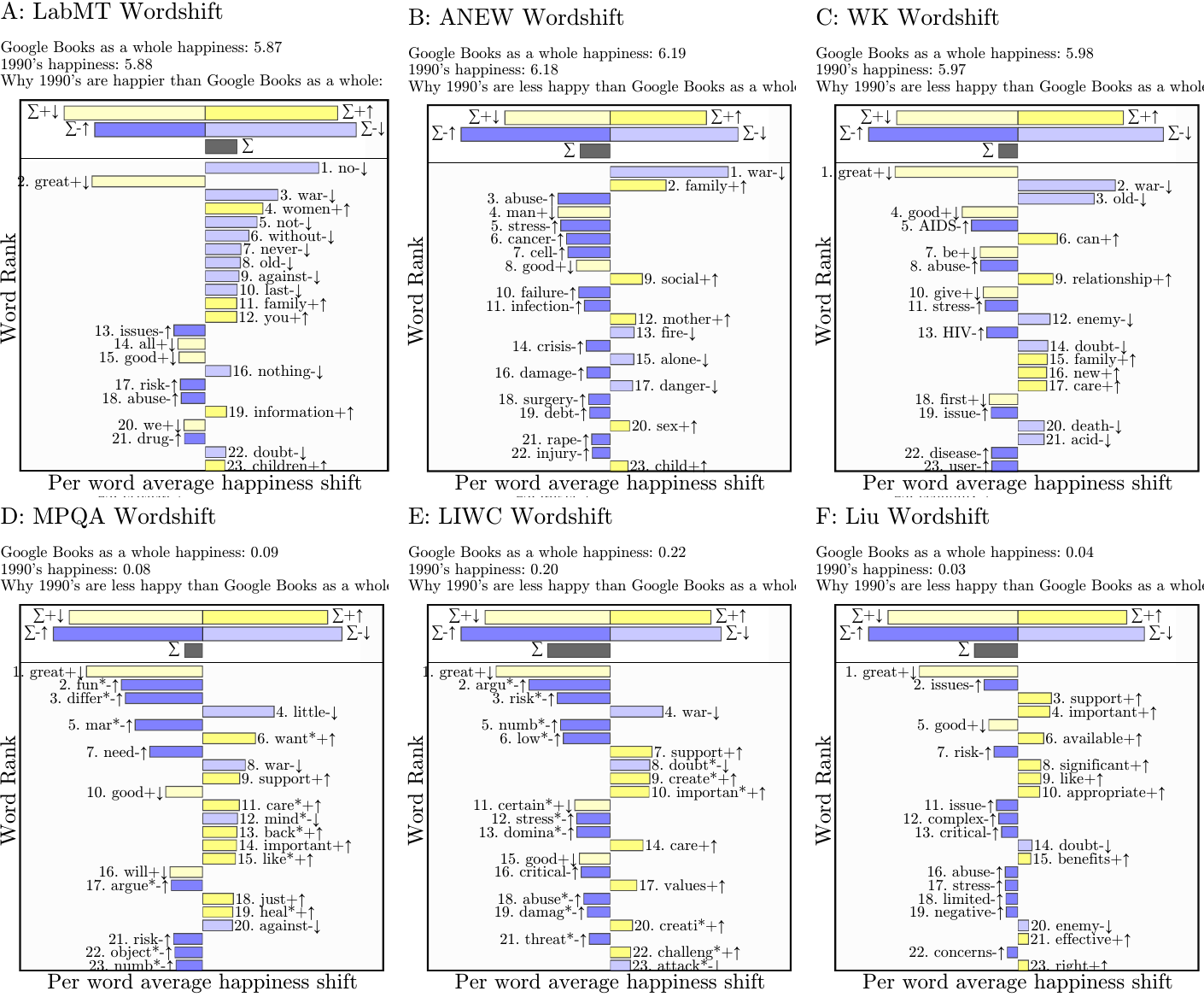}
  \caption[]{
    Google Books shifts in the 1990's against the baseline of Google Books.
  }
  \label{fig:gbooks-shifts-1990}
\end{figure*}

\begin{figure*}[!htb]
\includegraphics[width=0.98\textwidth]{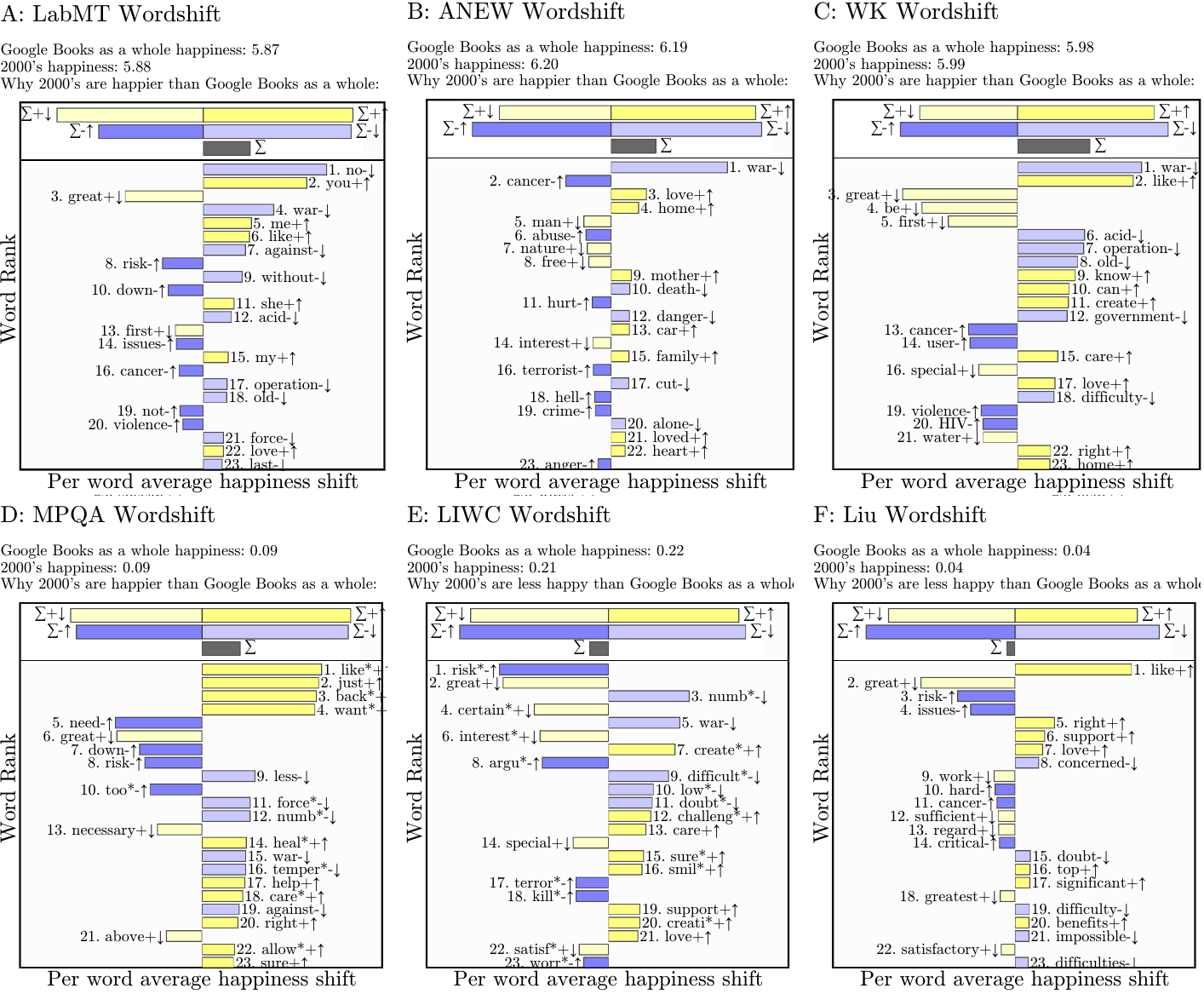}
  \caption[]{
    Google Books shifts in the 2000's against the baseline of Google Books.
  }
  \label{fig:gbooks-shifts-2000}
\end{figure*}

\clearpage
\pagebreak

\section{S7 Appendix: Additional Twitter time series, correlations, and shifts} \label{supp:twitter}

First, we present additional Twitter time series:

\begin{figure*}[!htb]
  \centering
\includegraphics[width=0.98\textwidth]{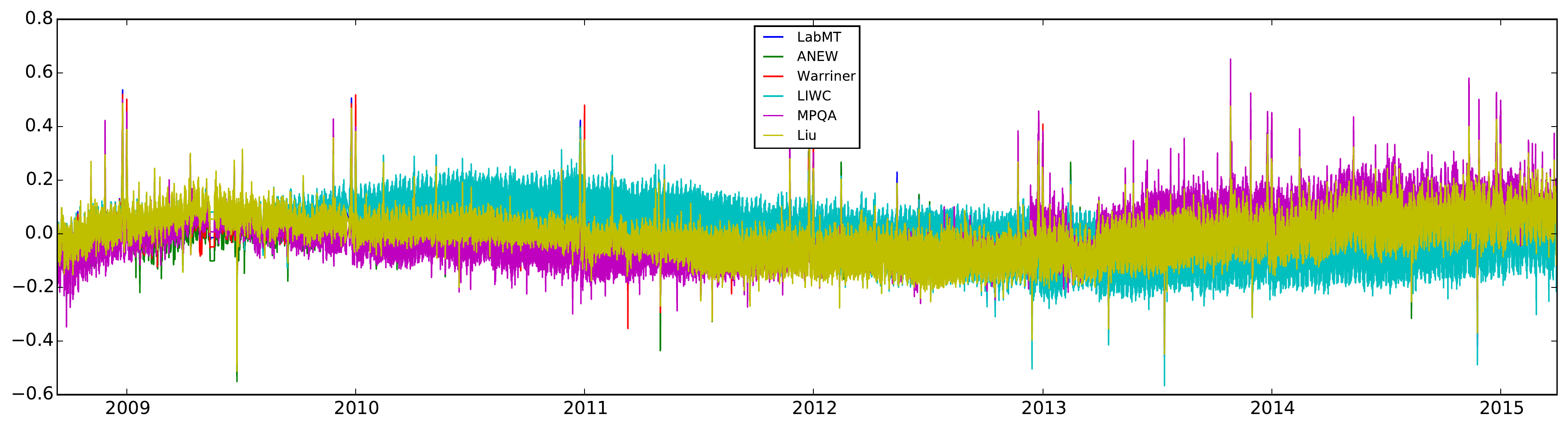}
  \caption[]{
    Normalized time series on Twitter using $\Delta _h$ of 1.0 for all.
    For resolution of 3 hours.
    We do not include any of the time series with resolution below 3 hours here because there are too many data points to see.
  }
  \label{fig:twitter_timeseries_2}
\end{figure*}

\begin{figure*}[!htb]
  \centering
\includegraphics[width=0.98\textwidth]{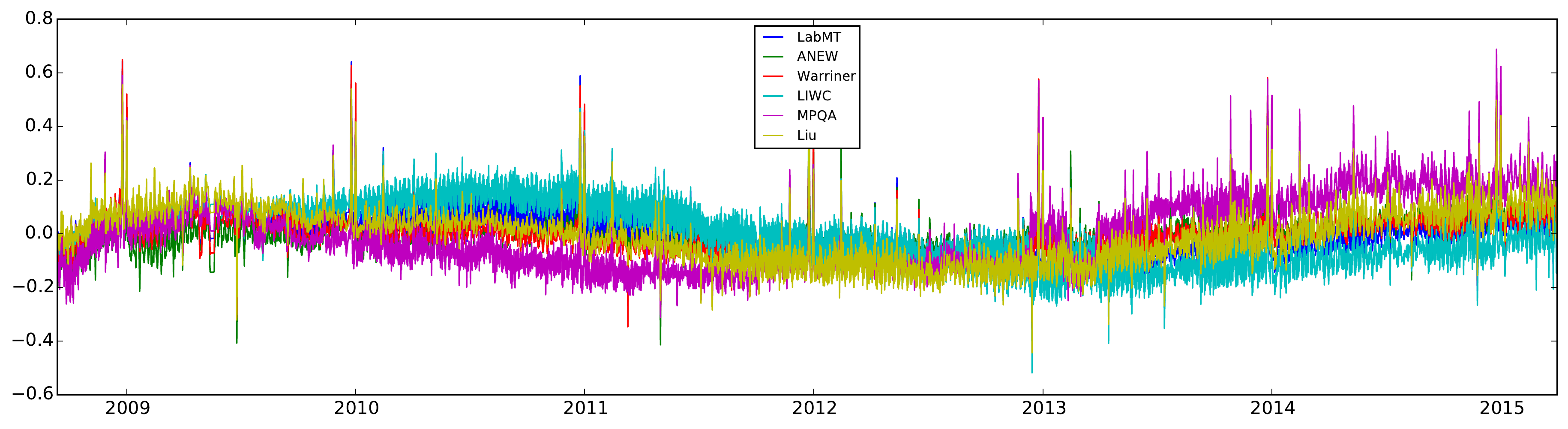}
  \caption[]{
    Normalized time series on Twitter using $\Delta _h$ of 1.0 for all.
    For resolution of 12 hours.
  }
  \label{fig:twitter_timeseries_3}
\end{figure*}

Next, we take a look at more correlations:

\begin{figure*}[ht]
\includegraphics[width=0.96\textwidth]{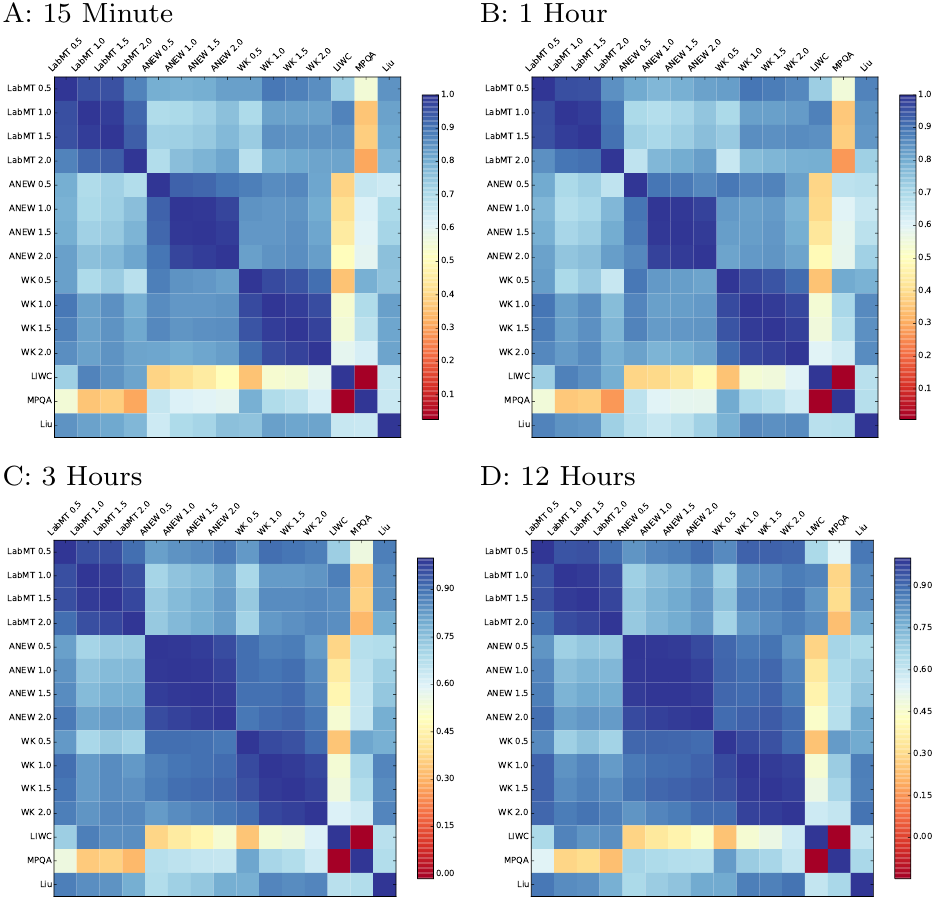}
    \caption[]{
    Pearson's $r$ correlation between Twitter time series for all resolutions below 1 day.
  }
  \label{fig:twitter_correlation_others}
\end{figure*}

\clearpage
\pagebreak

Now we include word shift graphs that are absent from the manuscript itself.

\begin{figure*}[ht]
  \centering
\includegraphics[width=0.98\textwidth]{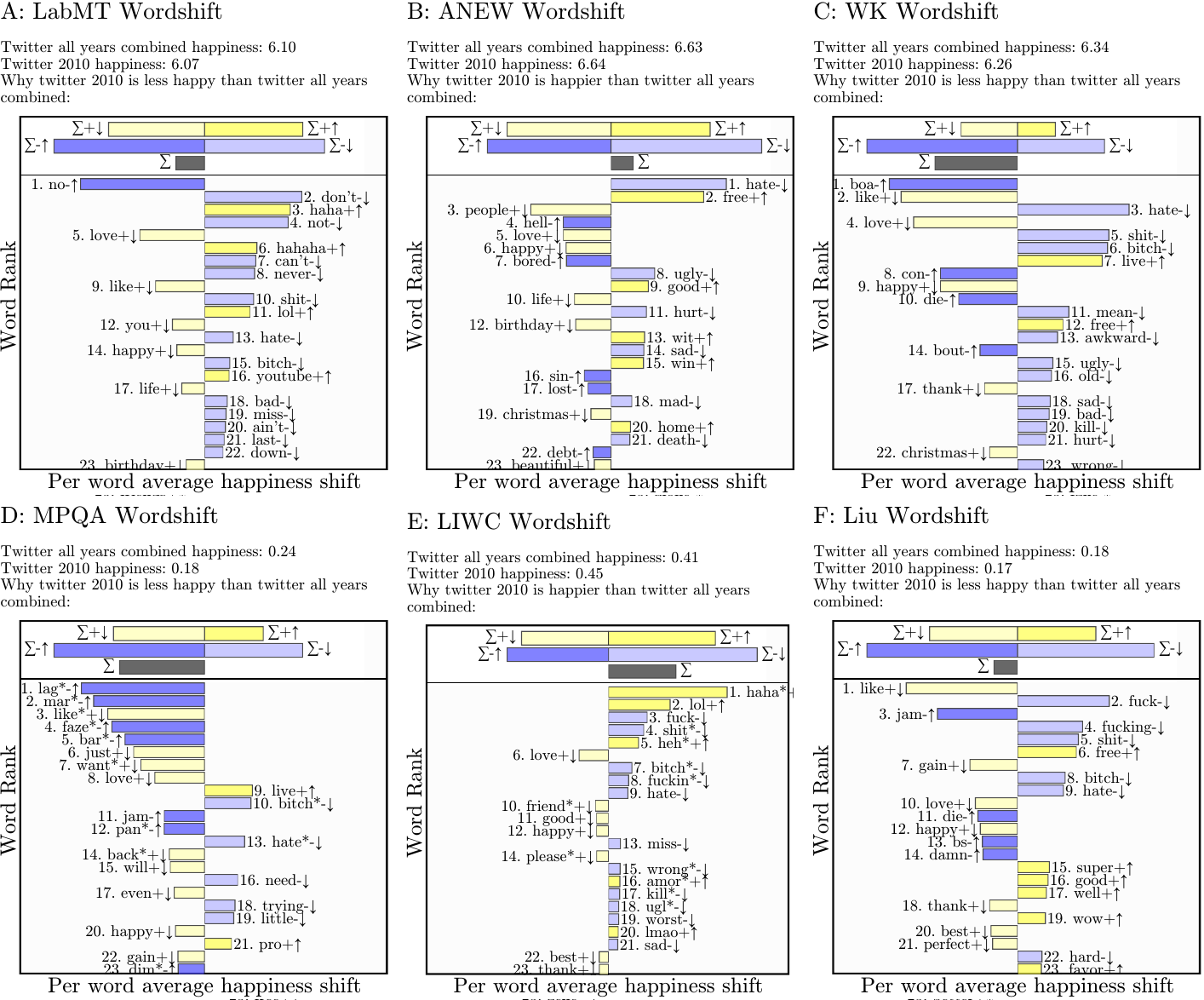}
  \caption[]{
    Word Shifts for Twitter in 2010.
    The reference word usage is all of Twitter (the 10\% Gardenhose feed) from September 2008 through April 2015, with the word usage normalized by year.
  }
  \label{fig:twitter-shift-2010}
\end{figure*}

\begin{figure*}[ht]
  \centering
\includegraphics[width=0.98\textwidth]{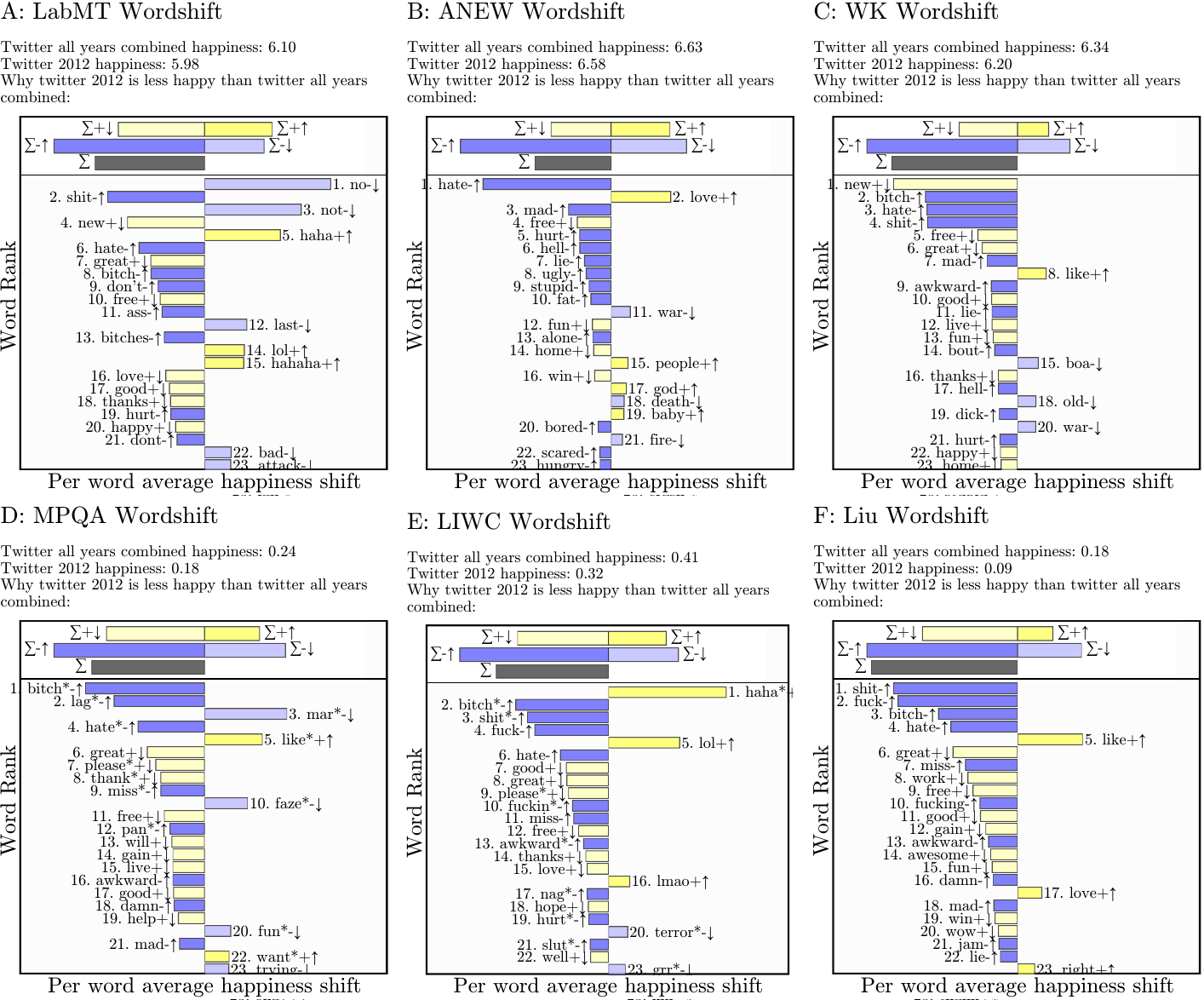}
  \caption[]{
    Word Shifts for Twitter in 2012.
    The reference word usage is all of Twitter (the 10\% Gardenhose feed) from September 2008 through April 2015, with the word usage normalized by year.
  }
  \label{fig:twitter-shift-2012}
\end{figure*}

\begin{figure*}[ht]
  \centering
\includegraphics[width=0.98\textwidth]{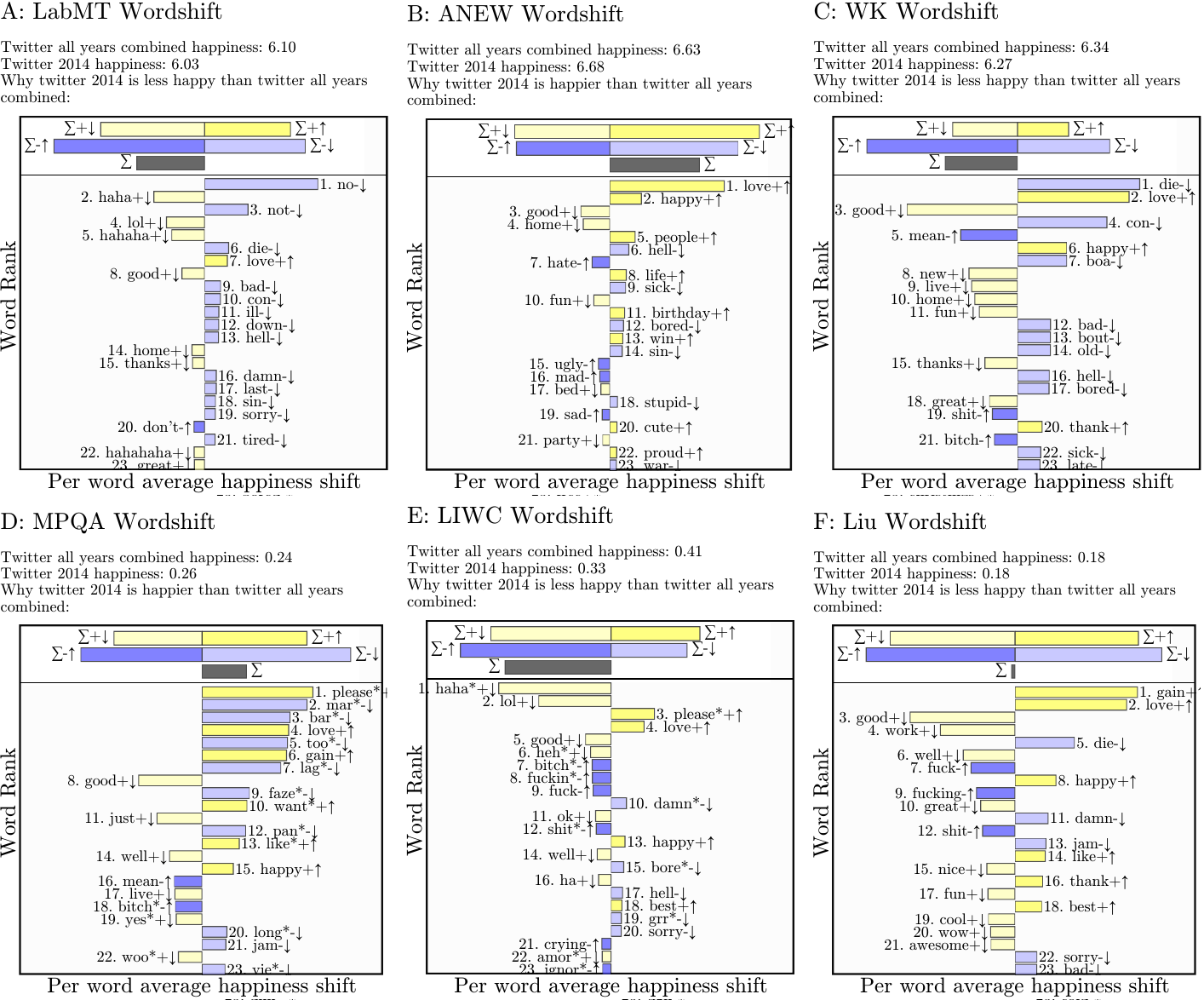}
  \caption[]{
    Word Shifts for Twitter in 2014.
    The reference word usage is all of Twitter (the 10\% Gardenhose feed) from September 2008 through April 2015, with the word usage normalized by year.
  }
  \label{fig:twitter-shift-2014}
\end{figure*}

\clearpage
\pagebreak

Finally, we include the results of each dictionary applied to a set of annotated Twitter data.
We apply sentiment dictionaries to rate individual Tweets and classify a Tweet as positive (negative) if the Tweet rating is greater (less) than the average of all scores in dictionary.

\begin{table*}[ht]
  \begin{adjustwidth}{\pnastableadjust in}{0in}
    \centering
\begin{tabular}{l | l | c | c | c | c}
Rank & Dictionary & \% Tweets scored & F1 of Tweets scored & Calibrated F1 & Overall F1\\
\hline
1. & Sent140Lex & 100.0 & 0.89 & 0.88 & 0.89\\
2. & labMT & 100.0 & 0.69 & 0.78 & 0.69\\
3. & HashtagSent & 100.0 & 0.67 & 0.64 & 0.67\\
4. & SentiWordNet & 98.6 & 0.67 & 0.68 & 0.67\\
5. & VADER & 81.3 & 0.75 & 0.81 & 0.61\\
6. & SentiStrength & 73.9 & 0.83 & 0.81 & 0.61\\
7. & SenticNet & 97.3 & 0.61 & 0.64 & 0.59\\
8. & Umigon & 67.1 & 0.87 & 0.85 & 0.58\\
9. & SOCAL & 82.2 & 0.71 & 0.75 & 0.58\\
10. & WDAL & 99.9 & 0.58 & 0.64 & 0.58\\
11. & AFINN & 73.6 & 0.78 & 0.80 & 0.57\\
12. & OL & 66.7 & 0.83 & 0.82 & 0.55\\
13. & MaxDiff & 94.1 & 0.58 & 0.70 & 0.54\\
14. & EmoSenticNet & 96.0 & 0.56 & 0.59 & 0.54\\
15. & MPQA & 73.2 & 0.73 & 0.72 & 0.53\\
16. & WK & 96.5 & 0.53 & 0.72 & 0.51\\
17. & LIWC15 & 61.8 & 0.81 & 0.78 & 0.50\\
18. & Pattern & 69.0 & 0.71 & 0.75 & 0.49\\
19. & GI & 67.6 & 0.72 & 0.70 & 0.49\\
20. & LIWC07 & 60.3 & 0.80 & 0.75 & 0.48\\
21. & LIWC01 & 54.3 & 0.83 & 0.75 & 0.45\\
22. & EmoLex & 59.4 & 0.73 & 0.69 & 0.43\\
23. & ANEW & 64.1 & 0.65 & 0.68 & 0.42\\
24. & USent & 4.5 & 0.74 & 0.73 & 0.03\\
25. & PANAS-X & 1.7 & 0.88 & -- & 0.01\\
26. & Emoticons & 1.4 & 0.72 & 0.77 & 0.01\\
\end{tabular}
    \end{adjustwidth}
  \caption{Ranked results of sentiment dictionary performance on individual Tweets from STS-Gold dataset (Saif, 2013).
    We report the percentage of Tweets for which each dictionary contains at least 1 entry, the F1 score on those Tweets, and the overall classification F1 score.
    The calibrated F1 score tunes the decision threshold between positive and negative Tweets with a random 10\% training sample.}
\label{tbl:STS}
\end{table*}

\clearpage
\pagebreak

\section{S8 Appendix: Naive Bayes results and derivation} \label{supp:bayes}

We now provide more details on the implementation of Naive Bayes, a derivation of the linearity structure, and more results from the classification of Movie Reviews.

First, to implement a binary Naive Bayes classifier for a collection of documents, we denote each of the $N$ words in the given document $T$ as $w_i$, thus the normalized word frequency is $f_i(T) = w_i/N$, and finally we denote the class labels $c_1,c_2$.
The probability of a document $T$ belonging to class $c_1$ can be written as
$$ P(c_1 | T) = \frac{P(c_1)P(T|c_1)}{P(T)}. $$
Since we do not know $P(T|c_1)$ explicitly, we make the \textit{naive} assumption that each word appears independently, and thus write
$$ P(c_1 | T) = \frac{P(c_1)\cdot \left [ P(f_1(T)|c_1) \cdot P(f_2(T)|c_1) \cdots P(f_N(T)|c_1)\right] }{P(T)}. $$
Since we are only interested in comparing $P(c_1 | T)$ and $P(c_2 | T)$, we disregard the shared denominator and have
$$ P(c_1 | T) \propto P(c_1)\cdot \left [ P(f_1(T)|c_1) \cdot P(f_2(T)|c_1) \cdots P(f_N(T)|c_1)\right] . $$
Finally we say that document $T$ belongs to class $c_1$ if $P(c_1 | T) > P(c_2 | T)$.
Given that the probabilities of individual words are small, to avoid machine truncation error we compute these probabilities in log space, such that the product of individual word likelihoods becomes a sum
$$ \log P(c_1 | T) \propto \log P(c_1) + \sum_{i=1}^N \log P(f_i(T)|c_1)  . $$
Assigning a classification of class $c_1$ if $P(c_1 | T) > P(c_2 | T)$ is the same as saying that the difference between the two is positive, i.e. $P(c_1 | T) - P(c_2 | T) > 0$ and since the logarithm is monotonic, $\log P(c_1 | T) - \log P(c_2 | T) > 0$.
To examine how individual words contribute to this difference, we can write
\begin{align*} 0 &< \log P(c_1 | T) - \log P(c_2 | T) \\
  &\propto \log P(c_1) + \sum_{i=1}^N \log P(f_i(T)|c_1) -\log P(c_2) - \sum_{i=1}^N \log P(f_i(T)|c_2)\\
  &\propto \log P(c_1) -\log P(c_2) + \sum_{i=1}^N \left [\log P(f_i(T)|c_1) - \log P(f_i(T)|c_2) \right ]\\
  &\propto \log \frac{P(c_1)}{P(c_2)} + \sum_{i=1}^N \log \frac{P(f_i(T)|c_1)}{P(f_i(T)|c_2)}.\end{align*}
We can see from the above that the contribution of each word $w_i$ (or more accurately, the likelihood of the frequency in document $T$ being predictive of class $c$ as $P(f_i(T)|c_1)$) is a linear constituent of the classification.

Next, we include the detailed results of the Naive Bayes classifier on the Movie Review corpus.

\begin{figure*}[!htb]
\includegraphics[width=0.96\textwidth]{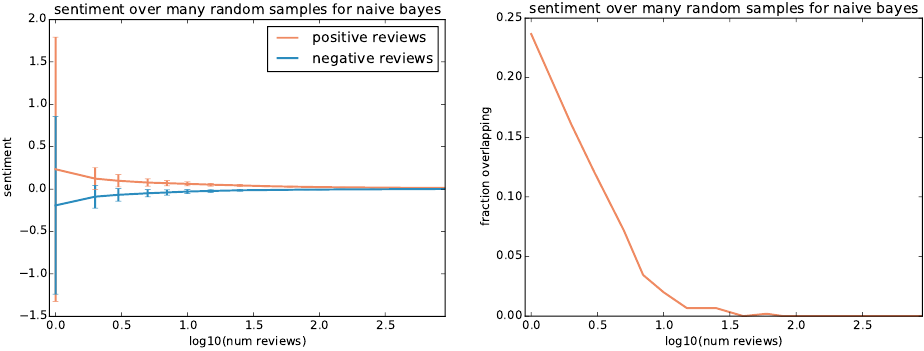}
    \caption[]{
    Results of the NB classifier on the Movie Reviews corpus.
  }
  \label{fig:NBresult}
\end{figure*}

\begin{figure*}[!htb]
\includegraphics[width=0.96\textwidth]{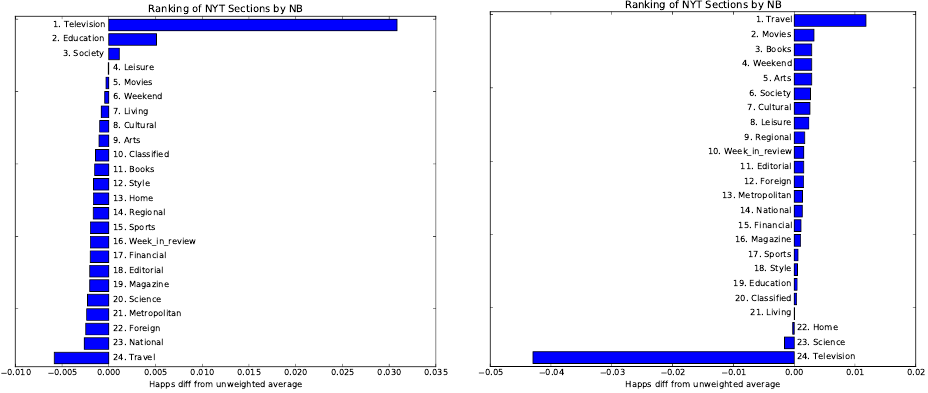}
    \caption[]{
    NYT Sections ranked by Naive Bayes in two of the five trials.
  }
  \label{fig:NBbars}
\end{figure*}

\begin{table*}[!htb]
  \begin{adjustwidth}{\pnastableadjust in}{0in}
    \centering
    \begin{tabular}{ | p{2cm} p{2cm} | p{2cm} p{2cm} |  }
  \hline
  \multicolumn{4}{|c|}{Most informative}  \\
  \hline
  \multicolumn{2}{|c}{Positive} & \multicolumn{2}{|c}{Negative} \\
  \hline
  Word & Value & Word & Value  \\
  \hline
  27.27 & flynt & 20.21 & godzilla  \\
26.33 & truman & 15.95 & werewolf \\
20.68 & charles & 13.83 & gorilla \\
15.04 & event & 13.83 & spice \\
14.10 & shrek & 13.83 & memphis \\
13.16 & cusack & 13.83 & sgt \\
13.16 & bulworth & 12.76 & jennifer \\
13.16 & robocop & 12.76 & hill \\
12.22 & jedi & 11.70 & max \\
12.22 & gangster & 11.70 & 200 \\
\hline
\end{tabular}
\\[3ex]
\begin{tabular}{ | p{2cm} p{2cm} | p{2cm} p{2cm} |  }
\hline
  \multicolumn{4}{|c|}{NYT Society} \\
  \hline
  \multicolumn{2}{|c}{Positive} & \multicolumn{2}{|c}{Negative}  \\
  \hline  
  Word & Value & Word & Value \\
  \hline
  26.08 & truman & 20.40 & godzilla \\
  20.49 & charles & 12.88 & hill  \\
  12.11 & gangster & 12.88 & jennifer  \\
  10.25 & speech & 10.73 & fatal  \\
  9.32 & melvin & 8.59 & freddie  \\
  8.85 & wars & 8.59 & = \\
  7.45 & agents & 8.59 & mess \\
  6.52 & dance & 8.59 & gene \\
  6.52 & bleak & 8.59 & apparent \\
  6.52 & pitt & 7.51 & travolta \\
  \hline
\end{tabular}
    \end{adjustwidth}
  \caption{Trial 1 of Naive Bayes trained on a random 10\% of the movie review corpus, and applied to the New York Times Society section.
    We show the words which are used by the trained classifier to classify individual reviews (in corpus), and on the New York Times (out of corpus).
    In addition, we report a second trial in Table \ref{tbl:NB-2}, since Naive Bayes is trained on a random subset of data, to show the variation in individual words between trials (while performance is consistent).}
\label{tbl:NB-1}
\end{table*}

\begin{table*}[!htb]
  \begin{adjustwidth}{\pnastableadjust in}{0in}
    \centering
    \begin{tabular}{ | p{2cm} p{2cm} | p{2cm} p{2cm} |  }
  \hline
  \multicolumn{4}{|c|}{Most informative}  \\
  \hline
  \multicolumn{2}{|c}{Positive} & \multicolumn{2}{|c}{Negative} \\
  \hline
  Word & Value & Word & Value  \\
  \hline
  18.11 & shrek & 34.63 & west\\
  17.15 & poker & 24.14 & webb\\
  15.25 & shark & 18.89 & jackal\\
  14.29 & maggie & 17.84 & travolta\\
  13.34 & guido & 17.84 & woo\\
  13.34 & outstanding & 17.84 & coach\\
  13.34 & political & 16.79 & awful \\
  13.34 & journey & 16.79 & brenner \\
  13.34 & bulworth & 15.74 & gabriel \\
  12.39 & bacon & 15.74 & general's \\
\hline
\end{tabular}
\\[3ex]
\begin{tabular}{ | p{2cm} p{2cm} | p{2cm} p{2cm} |  }
\hline
  \multicolumn{4}{|c|}{NYT Society} \\
  \hline
  \multicolumn{2}{|c}{Positive} & \multicolumn{2}{|c}{Negative}  \\
  \hline  
  Word & Value & Word & Value \\
  \hline
  17.79 & poker & 33.39 & west \\
  13.84 & journey & 17.20 & coach \\
  13.84 & political & 17.20 & travolta \\
  8.90 & tribe & 15.18 & gabriel \\
  7.91 & tony & 12.14 & pointless \\
  7.91 & price & 9.44 & stupid \\
  7.91 & threat & 8.09 & screaming \\
  7.12 & titanic & 7.59 & mess \\
  6.92 & dicaprio & 7.42 & boring \\
  6.92 & kate & 7.08 & = \\
  \hline
\end{tabular}
    \end{adjustwidth}
  \caption{Trial 2 of Naive Bayes trained on a random 10\% of the movie review corpus, and applied to the New York Times Society section.
    We show the words which are used by the trained classifier to classify individual reviews (in corpus), and on the New York Times (out of corpus).
    This second trial is in addition to the first trial in Table \ref{tbl:NB-1}, since Naive Bayes is trained on a random subset of data, to show the variation in individual words between trials (while performance is consistent).}
\label{tbl:NB-2}
\end{table*}

\clearpage
\pagebreak

\section{S9 Appendix: Movie review benchmark of additional dictionaries} \label{supp:additional}

Here, we present the accuracy of each dictionary applied to binary classification of Movie Reviews.

\begin{table*}[!htb]
  \begin{adjustwidth}{\pnastableadjust in}{0in}
          \centering
\begin{tabular}{l | l | c | c | c |}
Rank & Title & \% Scored & F1 Trained & F1 Untrained\\
\hline
1. & OL & 100 & 0.70 & 0.71\\
2. & HashtagSent & 100 & 0.67 & 0.66\\
3. & MPQA & 100 & 0.67 & 0.66\\
4. & SentiWordNet & 100 & 0.65 & 0.65\\
5. & labMT & 100 & 0.64 & 0.63\\
6. & AFINN & 100 & 0.67 & 0.63\\
7. & Umigon & 100 & 0.65 & 0.62\\
8. & GI & 100 & 0.65 & 0.61\\
9. & SOCAL & 100 & 0.71 & 0.60\\
10. & VADER & 100 & 0.67 & 0.60\\
11. & WDAL & 100 & 0.60 & 0.59\\
12. & SentiStrength & 100 & 0.63 & 0.58\\
13. & EmoLex & 100 & 0.65 & 0.56\\
14. & LIWC15 & 100 & 0.64 & 0.55\\
15. & LIWC01 & 100 & 0.65 & 0.54\\
16. & LIWC07 & 100 & 0.64 & 0.53\\
17. & Pattern & 100 & 0.73 & 0.52\\
18. & PANAS-X & 33 & 0.51 & 0.51\\
19. & Sent140Lex & 100 & 0.68 & 0.47\\
20. & SenticNet & 100 & 0.62 & 0.45\\
21. & ANEW & 100 & 0.57 & 0.36\\
22. & MaxDiff & 100 & 0.66 & 0.36\\
23. & EmoSenticNet & 100 & 0.58 & 0.34\\
24. & WK & 100 & 0.63 & 0.34\\
25. & Emoticons & 0 & -- & --\\
26. & USent & 40 & -- & --\\
\end{tabular}    \end{adjustwidth}
    \caption{Ranked performance of dictionaries on the Movie Review corpus.
                }
\label{tbl:MR-1}
\end{table*}

\begin{sidewaystable*}[!htb]
    \begin{adjustwidth}{\pnastableadjust in}{0in}
            \centering
\begin{tabular}{l | l | c | c | c | c }
Rank & Title & \% Scored & F1 Trained of Scored & F1 Untrained of Scored & F1 Untrained, All\\
\hline
1. & HashtagSent & 100 & 0.55 & 0.55 & 0.55\\
2. & LIWC15 & 99 & 0.53 & 0.55 & 0.55\\
3. & LIWC07 & 99 & 0.53 & 0.55 & 0.54\\
4. & LIWC01 & 99 & 0.52 & 0.55 & 0.54\\
5. & labMT & 99 & 0.54 & 0.54 & 0.54\\
6. & Sent140Lex & 100 & 0.55 & 0.54 & 0.54\\
7. & SentiWordNet & 99 & 0.54 & 0.53 & 0.53\\
8. & WDAL & 99 & 0.53 & 0.53 & 0.52\\
9. & EmoLex & 95 & 0.54 & 0.55 & 0.52\\
10. & MPQA & 93 & 0.54 & 0.55 & 0.52\\
11. & SenticNet & 97 & 0.53 & 0.52 & 0.50\\
12. & SOCAL & 88 & 0.56 & 0.55 & 0.49\\
13. & EmoSenticNet & 98 & 0.52 & 0.46 & 0.45\\
14. & Pattern & 81 & 0.55 & 0.55 & 0.45\\
15. & GI & 80 & 0.55 & 0.55 & 0.44\\
16. & WK & 97 & 0.54 & 0.45 & 0.44\\
17. & OL & 76 & 0.56 & 0.57 & 0.44\\
18. & VADER & 79 & 0.56 & 0.55 & 0.43\\
19. & SentiStrength & 77 & 0.54 & 0.54 & 0.41\\
20. & MaxDiff & 83 & 0.54 & 0.49 & 0.41\\
21. & AFINN & 70 & 0.56 & 0.56 & 0.39\\
22. & ANEW & 63 & 0.52 & 0.48 & 0.30\\
23. & Umigon & 53 & 0.56 & 0.56 & 0.30\\
24. & PANAS-X & 1 & 0.53 & 0.53 & 0.01\\
25. & Emoticons & 0 & -- & -- & --\\
26. & USent & 2 & -- & -- & --\\
\end{tabular}    \end{adjustwidth}
    \caption{Ranked performance of dictionaries on the Movie Review corpus, broken into sentences.
                }
\label{tbl:MR-2}
\end{sidewaystable*}

\begin{figure*}[!htb]
\includegraphics[width=0.96\textwidth]{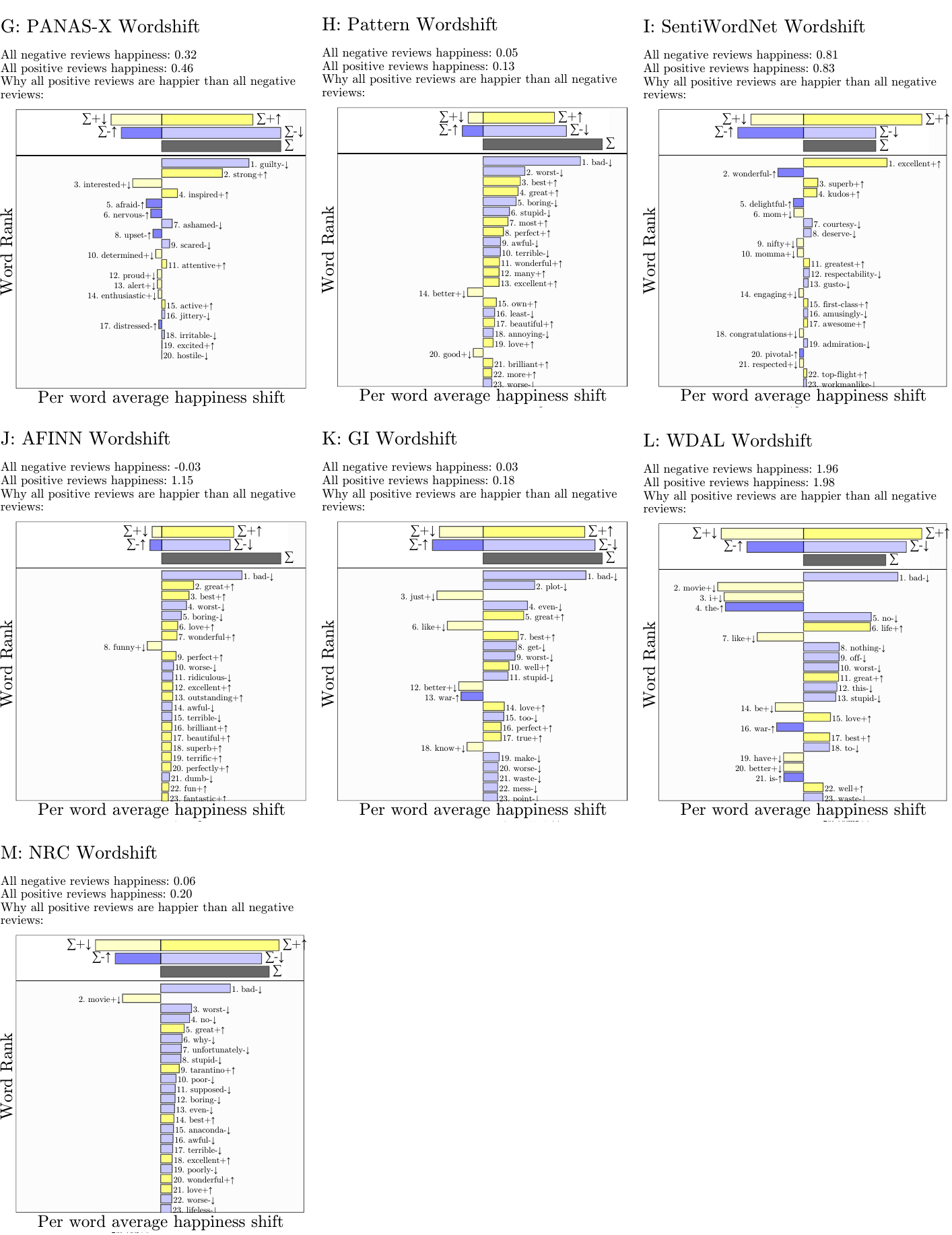}
    \caption[]{
      Word shifts for the movie review corpus, with panel letters continuing from Fig. \ref{fig:moviereviews-shifts}.
      We again see many of the same patterns, and refer the reader to Fig. \ref{fig:moviereviews-shifts} for a more in depth analysis.
  }
  \label{fig:moviereviews-shifts-extras}
\end{figure*}

\clearpage
\pagebreak

\section{S10 Appendix: Coverage removal and binarization tests of labMT dictionary } \label{supp:labMT-test}

    Here, we perform a detailed analysis of the labMT dictionary to further isolate the effects of dictionary coverage and scoring type.
    This analysis is motivated by ensuring that the our results are not confounded entirely by the quality of the word scores across dictionaries, such that the effect of coverage and scoring type are isolated.
    We focus on the Movie Review corpus for this analysis and analyzing the different between positive and negative reviews using word shift graphs.
    While our attention is focused on a qualitative understanding of the differences in these two sets of documents, we also report the accuracy of the labMT dictionary with the aforementioned modifications using the F1 score.

\subsection{Binarization}
    First, we gradually reduce the range of scores in the labMT dictionary from a centered -4 $\to$ 4, down to just the integer scores $-1$ and $1$.
    This process is accomplished by first using a $\Delta_h = 1.00$, leaving words with scores from 1--4 and 6--9, and then applying a linear transformation to these sets of words.
    We subtract the center value of 5.0 from the words, leaving words with ranges from -4-- -1 and 1--4, and then linearly map these sets to scores with a reduced range.
    For a binarization of 25\%, we map -4-- -1 to -3.25 -- -1 and 1--4 to 1--3.25, reducing the range in direction from 3 to 2.25 (a 25\% reduction).
    For a binarization of 50\%, this becomes a map of -4-- -1 to -2.5 -- -1 and 1--4 to 1--2.5, leaving only half of the original range of values.
    Finally, a binarization of 100\% sets the score for all words -4-- -1 to -1, and words 1--4 to 1.

\begin{figure*}[!htb]
\includegraphics[width=0.96\textwidth]{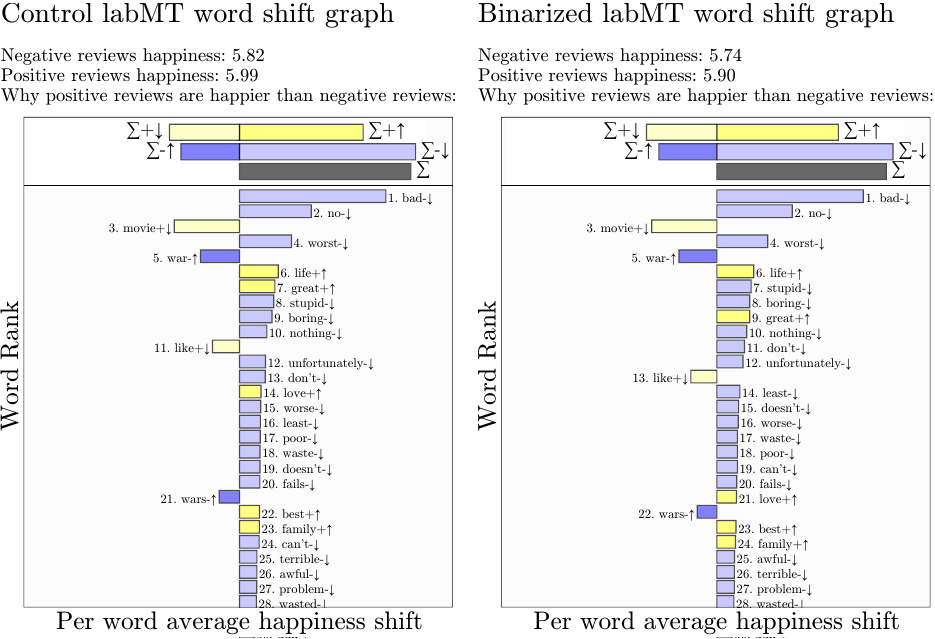}
  \caption[]{
    Word shift graph resulting from the 25\% binarization of the labMT dictionary.
  }
  \label{fig:labMT-binary-shift-1}
\end{figure*}

    In Figs. \ref{fig:labMT-binary-shift-1}--\ref{fig:labMT-binary-shift-4} we observe that the binarization of the labMT dictionary results in observably different word shift graphs by changing which words contribute to the sentiment differences as well as reducing the difference in sentiment scores between the two corpora.
    Looking specifically at Fig. \ref{fig:labMT-binary-shift-4}, the top 5 words in the control word shift graph are bad, no, movie, worst, and war.
    In the binarized version, the top 5 are bad, no, movie, nothing, and worst.
    The top 5 from the continuous dictionary move into places 1, 2, 3, 5, and 10.
    Examining only the positive words that increased in frequency (not all shown in the Figure), we have ``3. movie (3)'', ``11. like (24)'', ``32. funny (102)'', ``33. better (46)'', and ``43. jokes (133)'' in the control version, with these words' positions in the binarized version in parenthesis.
    In the binarized version, these top words are ``3. movie (3)'', ``24. like (11)'', ``30. you (84)'', ``36. up (126)'', ``37. all (98)'', where the first number is the place in the overall list for the given labMT score list, with the place for that word in the control word shift graph in parenthesis.

    In Figure \ref{fig:labMT-binary}, the F1 score is show across this gradual, linear change to a binary dictionary.
    We observe that the full binarization of the labMT dictionary results in a degradation of performance, although the differences are not statistically significant.

\begin{figure*}[!htb]
\includegraphics[width=0.96\textwidth]{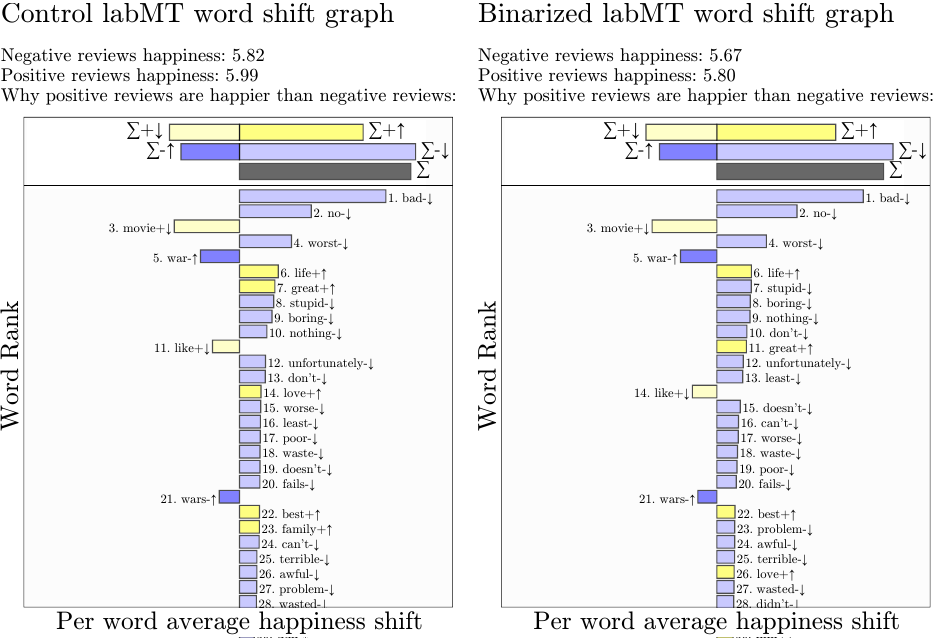}
  \caption[]{
    Word shift graph resulting from the 50\% binarization of the labMT dictionary.
  }
  \label{fig:labMT-binary-shift}
\end{figure*}
\begin{figure*}[!htb]
\includegraphics[width=0.96\textwidth]{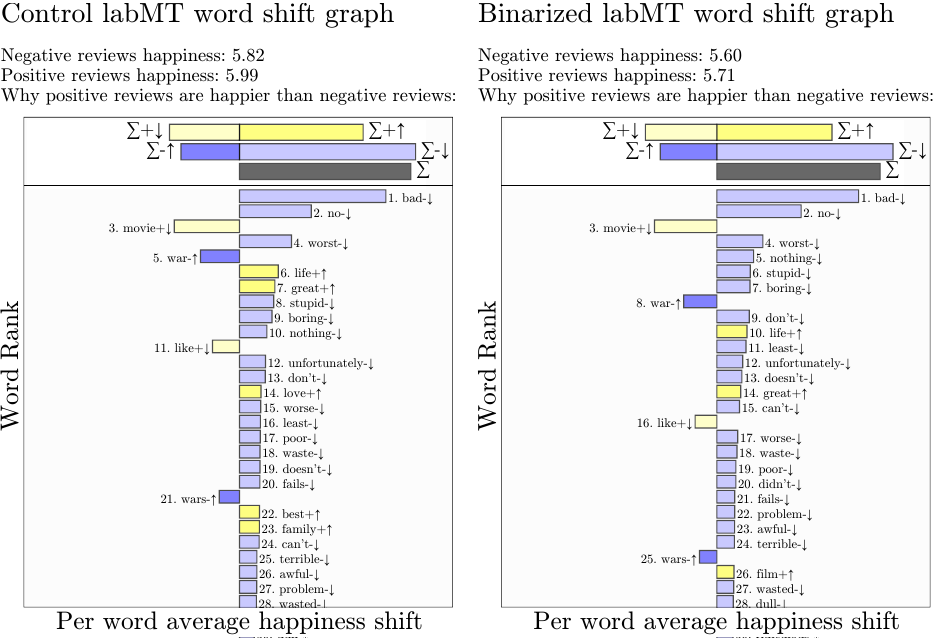}
  \caption[]{
    Word shift graph resulting from the 75\% binarization of the labMT dictionary.
  }
  \label{fig:labMT-binary-shift}
\end{figure*}
\begin{figure*}[!htb]
\includegraphics[width=0.96\textwidth]{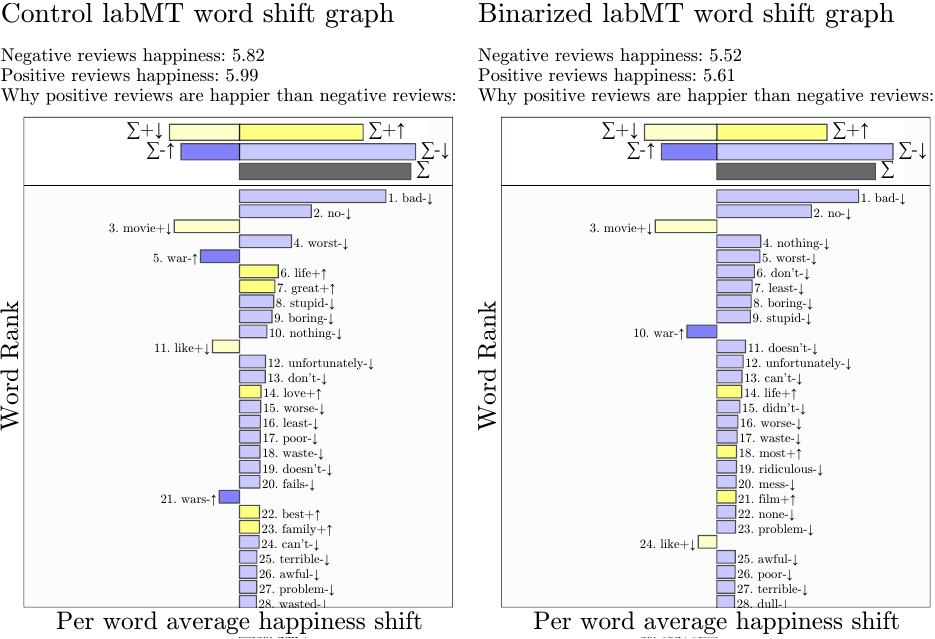}
  \caption[]{
    Word shift graph resulting from the full binarization of the labMT dictionary.
  }
  \label{fig:labMT-binary-shift-4}
\end{figure*}

\begin{figure*}[!htb]
  \begin{center}
\includegraphics[width=0.48\textwidth]{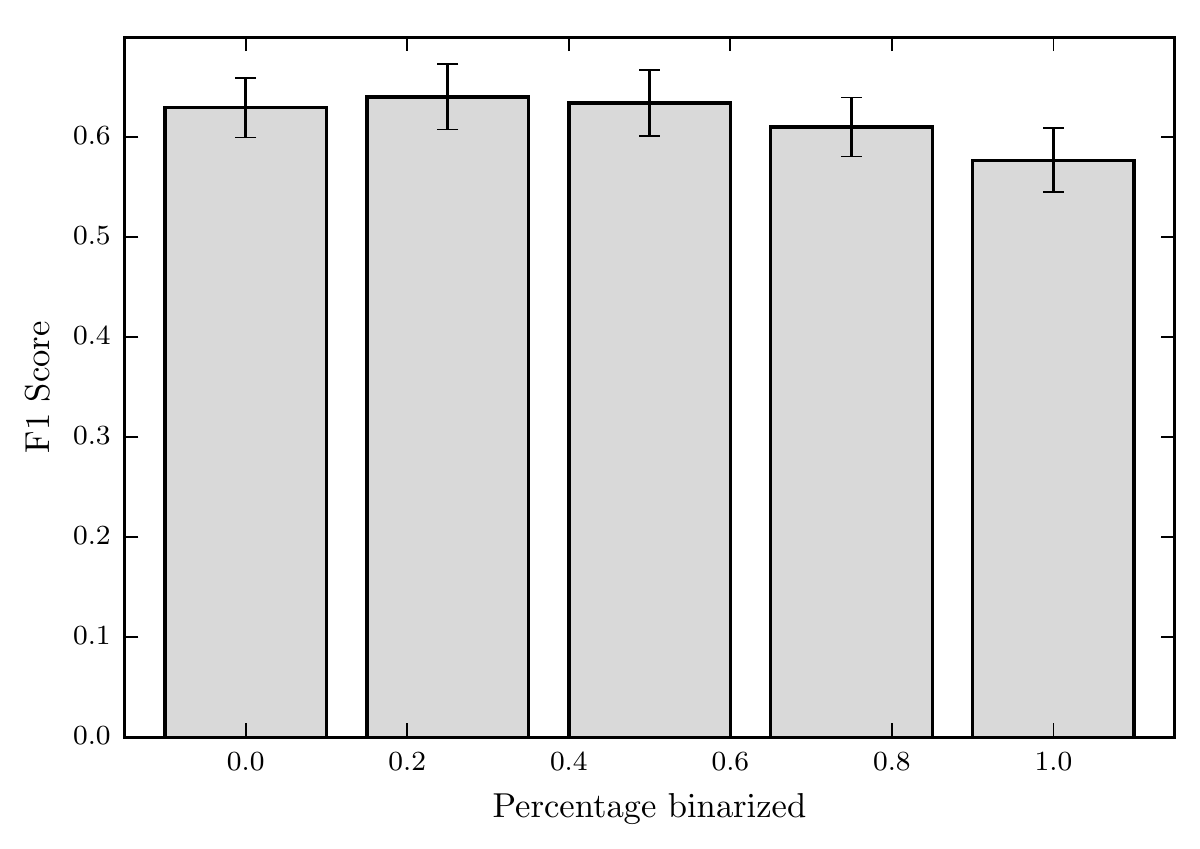}
  \end{center}
  \caption[]{
    The direct binarization of the labMT dictionary results in a degradation of performance.
    The binarization is accomplished by linearly reducing the range of scores in the labMT dictionary from a centered -4 $\to$ 4 to the integer scores $-1$ and $1$.
  }
  \label{fig:labMT-binary}
\end{figure*}

\clearpage

    \subsection{Reduced coverage}
    Second, to test the effect of coverage alone, we systematically reduce the coverage of the labMT dictionary and again attempt the binary classification task of identifying Movie Review polarity.
    Three possible strategies to reduce the coverage are (1) removing the most frequent words, (2) removing the least frequent words, and (3) removing words randomly (irrespective of their frequency of usage).

\begin{figure*}[!htb]
\includegraphics[width=0.96\textwidth]{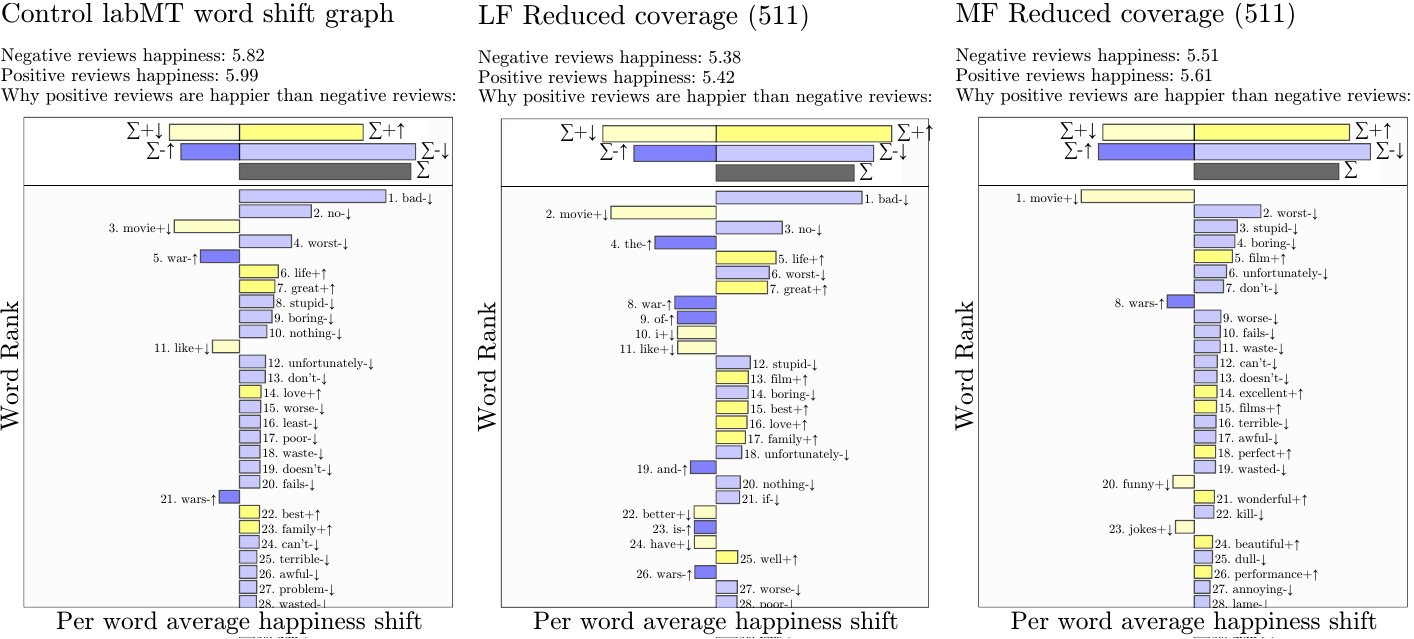}
  \caption[]{
    Word shift graphs resulting from the remove of the most frequent (MF) and least frequent (LF) words in the labMT dictionary.
  }
  \label{fig:labMT-cov-shift-first}
\end{figure*}

    In Figs. \ref{fig:labMT-cov-shift-first}--\ref{fig:labMT-cov-shift-last}, we show the resulting word shift graphs with the control (all words included) alongside word shift graphs using the labMT dictionary with the least frequent (LF) and most frequent (MF) words removed.
    Each word shift graph with reduced coverage shows the number of words removed in parenthesis in the title,
    e.g.,
    in Fig. \ref{fig:labMT-cov-shift-first} we see the titles ``LF Reduced coverage (511)'' and ``MF Reduced coverage (511)''
    which indicate that 511 words were removed in the indicated fashion.
    We first observe that the difference in sentiment scores between the positive and negative movie reviews is decreased from 0.17 to 0.02--0.05 and 0.09--0.15 for the LF and MF strategies, respectively, while noting that these differences do not result in predictive accuracy (i.e., classification accuracy is not statistically significant worsened).
    Examining the words in Fig. \ref{fig:labMT-cov-shift-first} more closely, where only 5\% of the words have been removed, we already observe departures in individual word contributions.
    Of the top 5 words in the control graph (``bad'', ``no'', ``movie'', ``worst'', and ``war''), we see only 3 of these in the top 5 for LF (all in the top 8) and only 1 in the top for MF (with 2 of the 5 showing on the graph at all).
    In the LF graph we lose words like ``don't'', ``least'', ``doesn't'', ``terrible'', ``awful'', ``problem'', and instead see the words ``the'', ``of'', ``i'', ``is'', ``have'' contribute more strongly.
    In the MF graph we lose common words like ``best'', ``family'', ``love'', ``life'', ``like'' and instead see the less common words ``excellent'', ``perfect'', ``funny'', ``wonderful'', ``kill'', ``jokes'', ``beautiful'', ``dull'', ``performance'', ``annoying'', and ``lame''.
    As one might expect, these trends of common/uncommon words varying across the word shifts graphs continue for increasingly reduced coverage.

    With approximately half of the words from the labMT dictionary removed, in Fig. \ref{fig:labMT-cov-shift-5110} we observe high overlap between the words in the control and LF, and only a single word in common between the control and MF word shift graphs.
    In addition to this, the sentiment score difference between the positive and negative reviews is 0.17 for the control, 0.04 for LF, and 0.14 for MF.
        In Fig. \label{fig:labMT-cov-shift-9198}, only 1,024 (of 10,222) words remain in the LF and MF reduced coverage dictionaries, and again we see similar trends.
    Higher overlap exists between the LF and control, with only two words (``don't'', ``can't'') in common between MF and control.
    While coverage remains above 50\% for the LF strategy, the word shift graph shows more words that are weighting the classification incorrectly: ``the'', ``i'', ``war'', ``like'', etc.
    The MF word shift graph shows interesting words but also has many words that detracting from the classification: ``i'm'', ``spice'', ``they're'', ``drunken'', etc.
    We can conclude again, with these observations, that sentiment classification and sentiment understanding using word shifts graphs relies on broad coverage of the words used in the text being analyzed.

    In Figures \ref{fig:labMT-coverage-removal-result} and \ref{fig:labMT-coverage-removal}, we show the resulting F1 score of classification performance for each of these three strategies and the total coverage from each removal strategy.
    We observe that while certain strategies are more effective at retaining performance, lower coverage scores are all lower despite substantial variation, and the overall pattern for each strategy is a decrease in performance for decreasing coverage.
    In both cases these results are consistent with those seen across dictionaries: integer scores and low coverage strongly reduce the performance of the 2-class movie review classification task, as measured by the F1-score.
    We note that this trend is not statistically significant, as can be observed with the standard deviation error bars.

\begin{figure*}[!htb]
\includegraphics[width=0.96\textwidth]{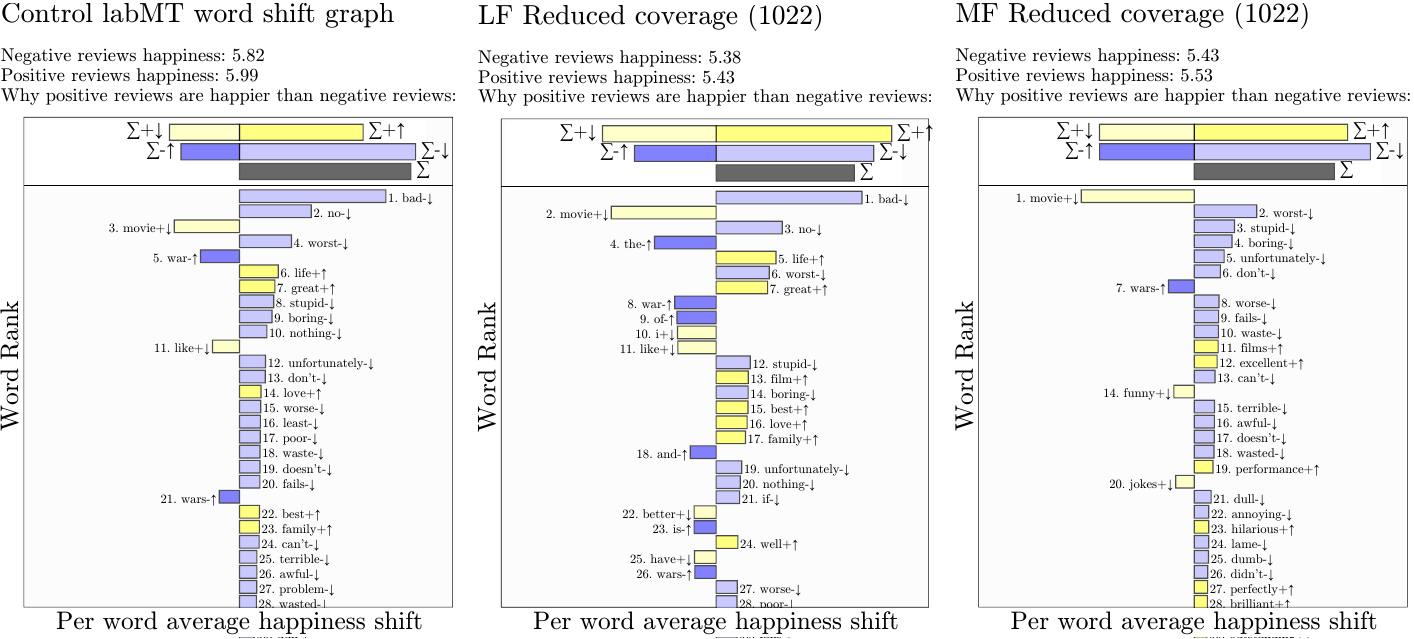}
  \caption[]{
    Word shift graphs resulting from the remove of the most frequent (MF) and least frequent (LF) words in the labMT dictionary.
  }
  \label{fig:labMT-cov-shift}
\end{figure*}
\begin{figure*}[!htb]
\includegraphics[width=0.96\textwidth]{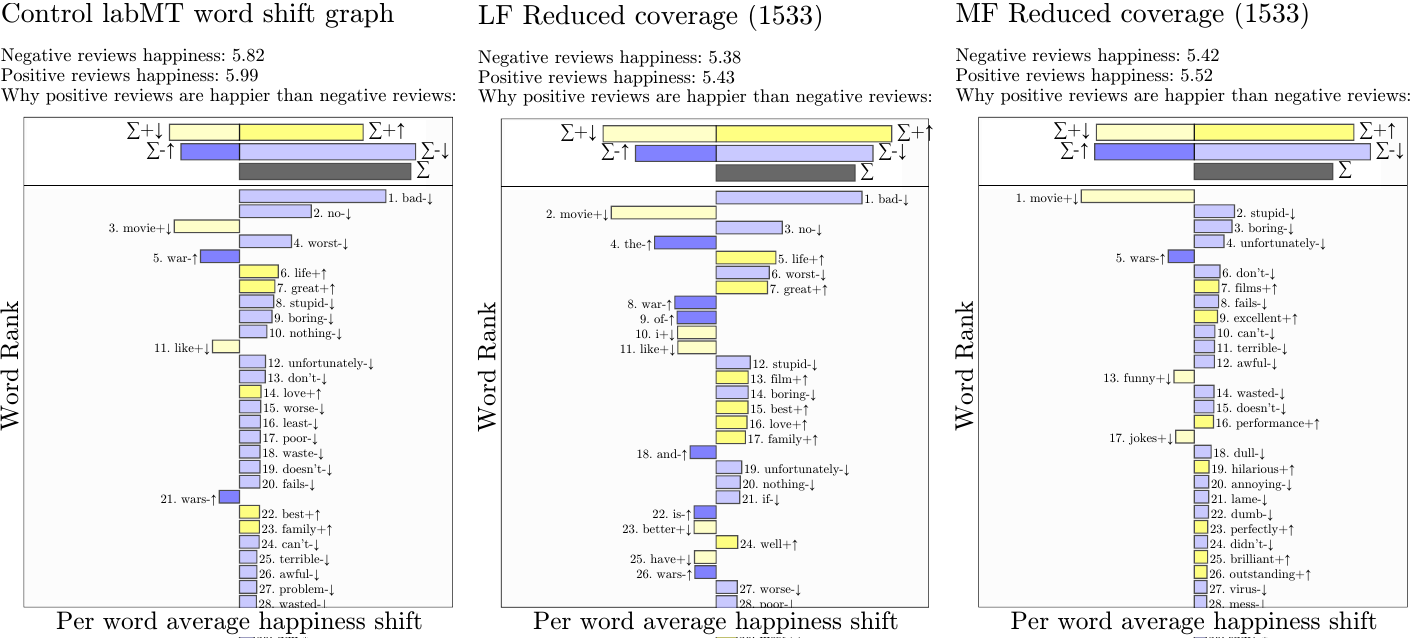}
  \caption[]{
    Word shift graphs resulting from the remove of the most frequent (MF) and least frequent (LF) words in the labMT dictionary.
  }
  \label{fig:labMT-cov-shift}
\end{figure*}
\begin{figure*}[!htb]
\includegraphics[width=0.96\textwidth]{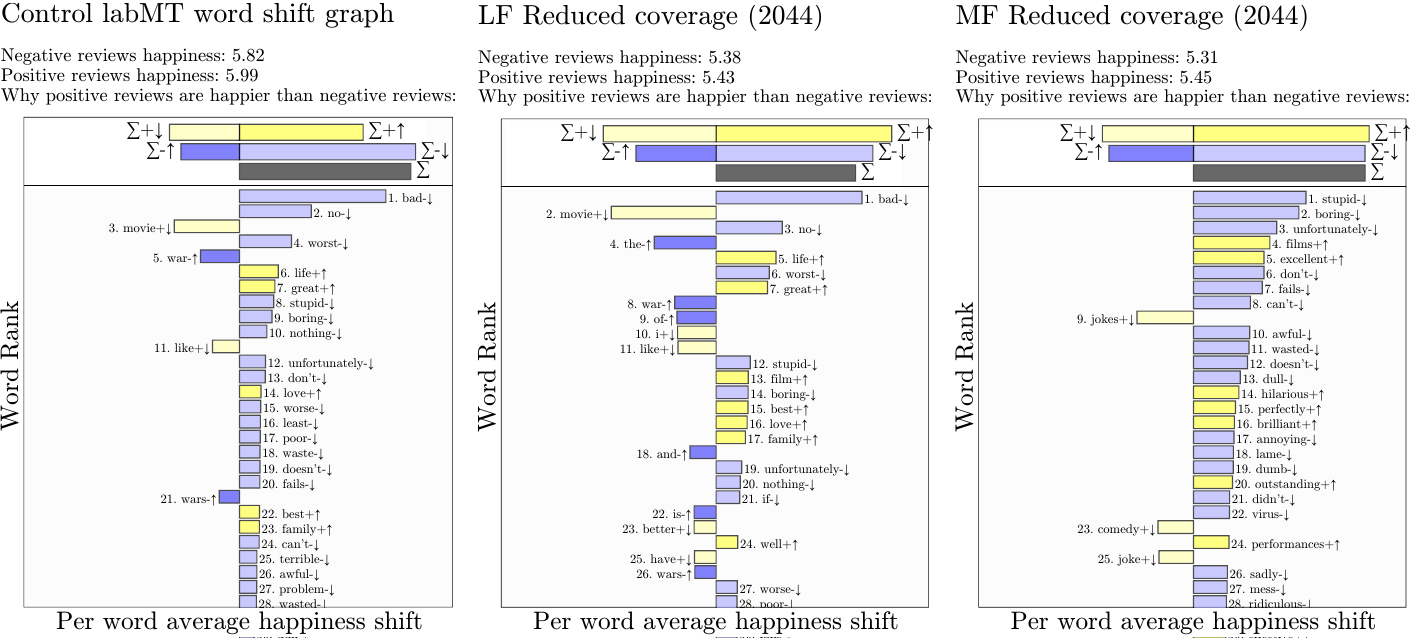}
  \caption[]{
    Word shift graphs resulting from the remove of the most frequent (MF) and least frequent (LF) words in the labMT dictionary.
  }
  \label{fig:labMT-cov-shift}
\end{figure*}
\begin{figure*}[!htb]
\includegraphics[width=0.96\textwidth]{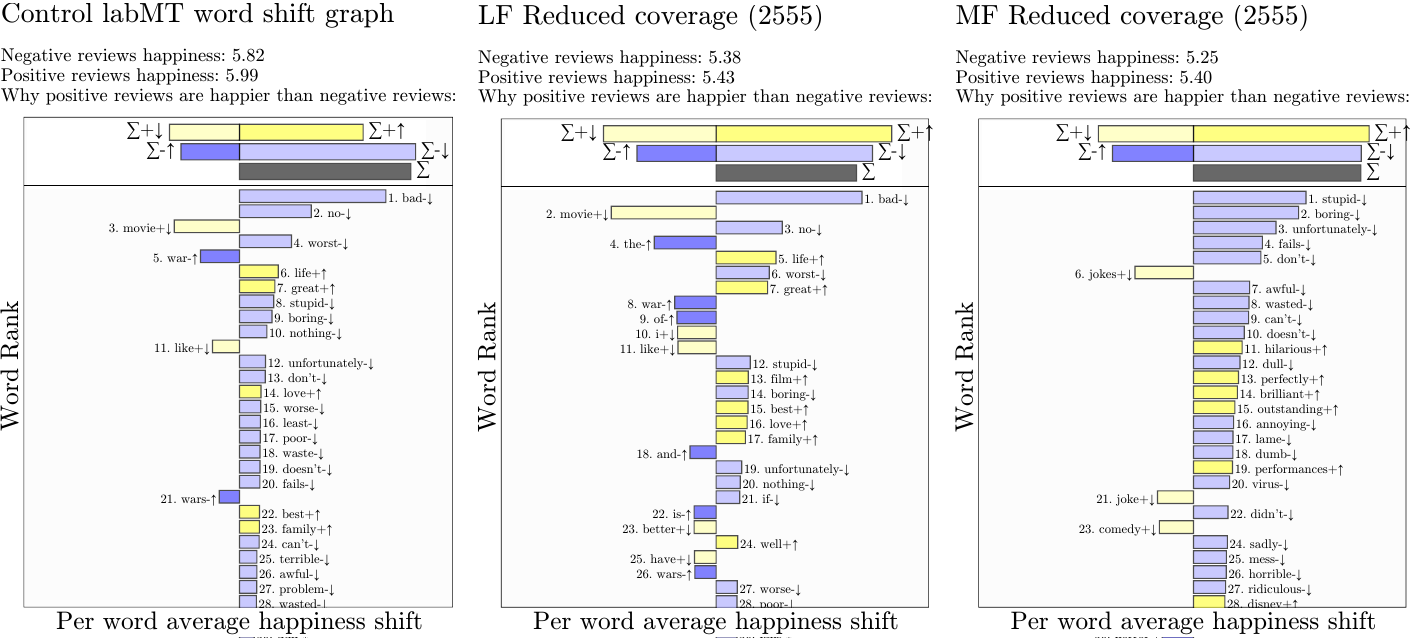}
  \caption[]{
    Word shift graphs resulting from the remove of the most frequent (MF) and least frequent (LF) words in the labMT dictionary.
  }
  \label{fig:labMT-cov-shift}
\end{figure*}
\begin{figure*}[!htb]
\includegraphics[width=0.96\textwidth]{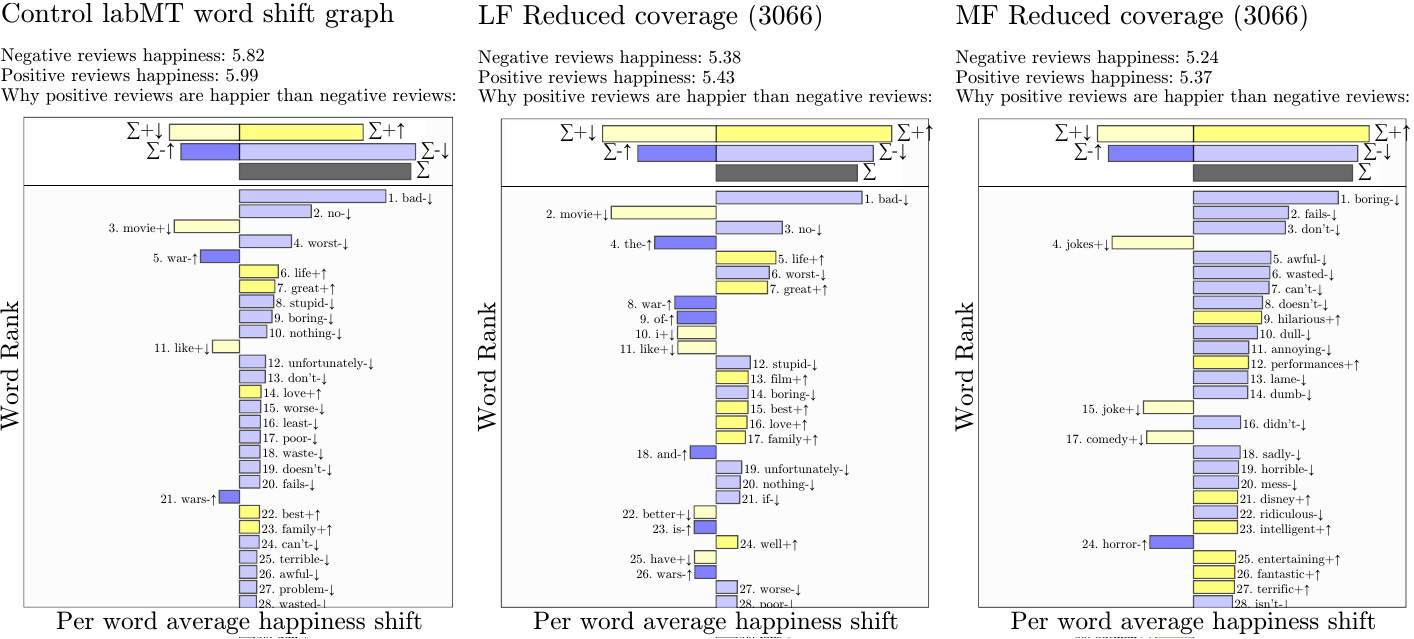}
  \caption[]{
    Word shift graphs resulting from the remove of the most frequent (MF) and least frequent (LF) words in the labMT dictionary.
  }
  \label{fig:labMT-cov-shift}
\end{figure*}
\begin{figure*}[!htb]
\includegraphics[width=0.96\textwidth]{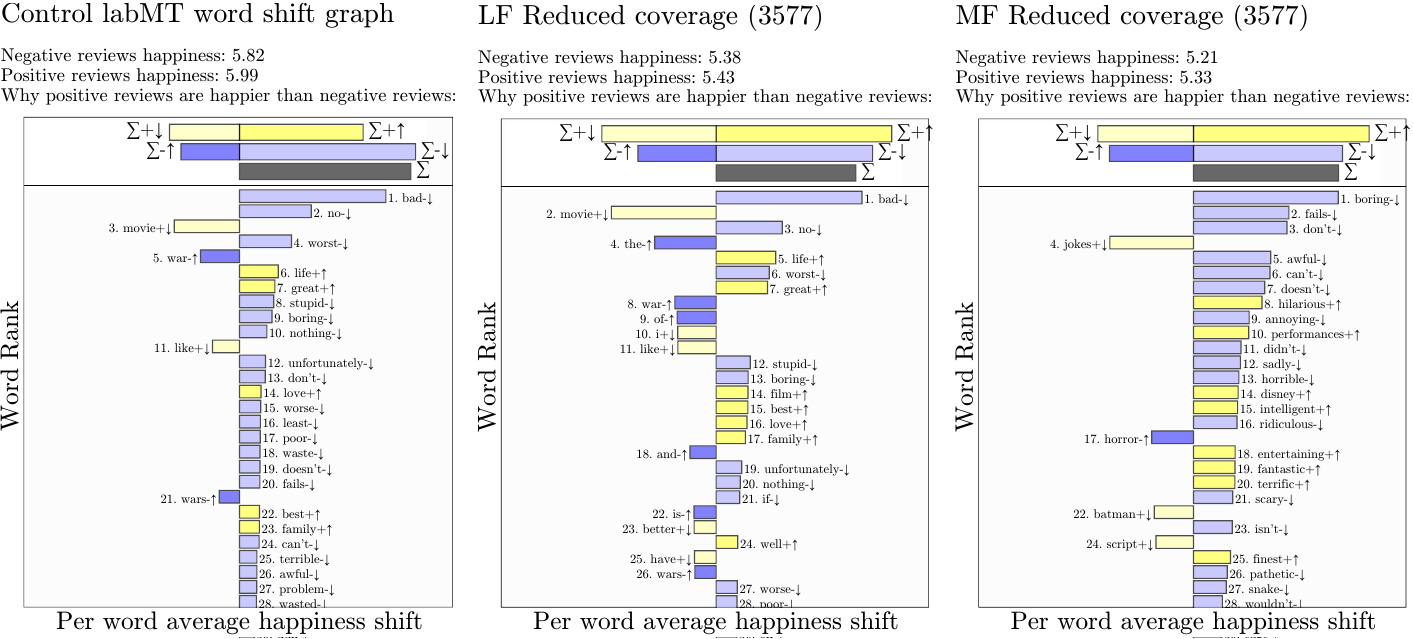}
  \caption[]{
    Word shift graphs resulting from the remove of the most frequent (MF) and least frequent (LF) words in the labMT dictionary.
  }
  \label{fig:labMT-cov-shift}
\end{figure*}
\begin{figure*}[!htb]
\includegraphics[width=0.96\textwidth]{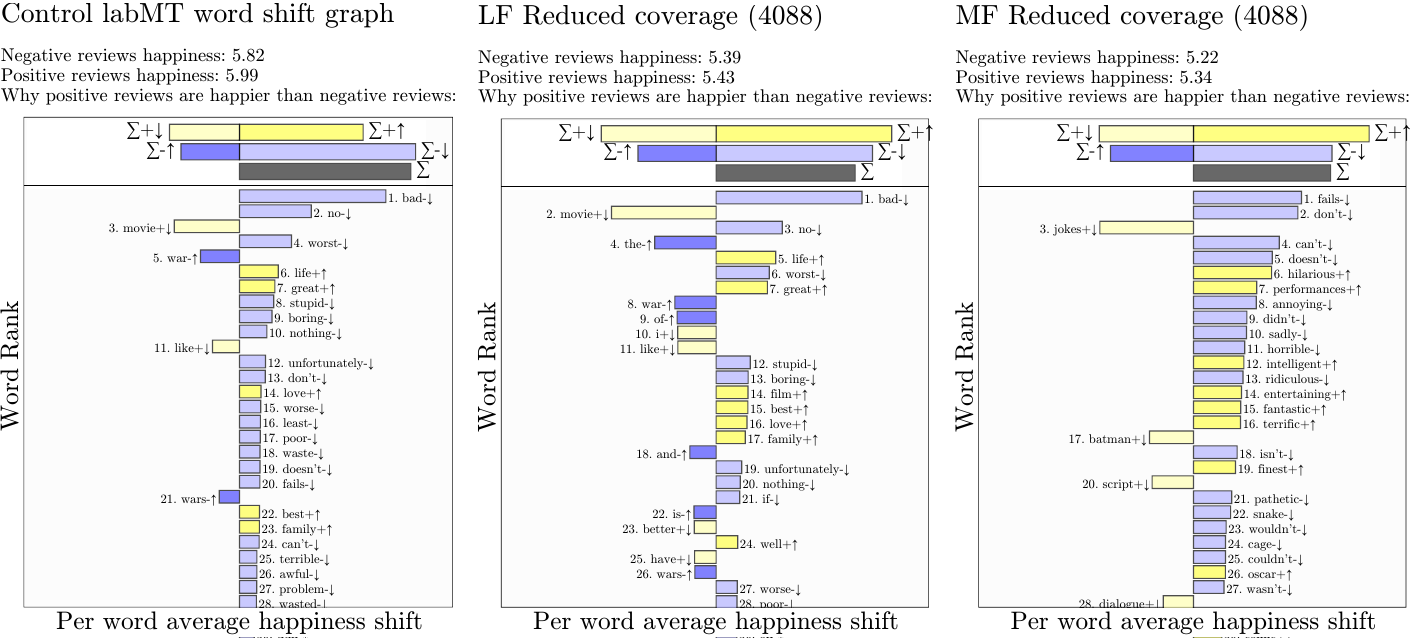}
  \caption[]{
    Word shift graphs resulting from the remove of the most frequent (MF) and least frequent (LF) words in the labMT dictionary.
  }
  \label{fig:labMT-cov-shift}
\end{figure*}
\begin{figure*}[!htb]
\includegraphics[width=0.96\textwidth]{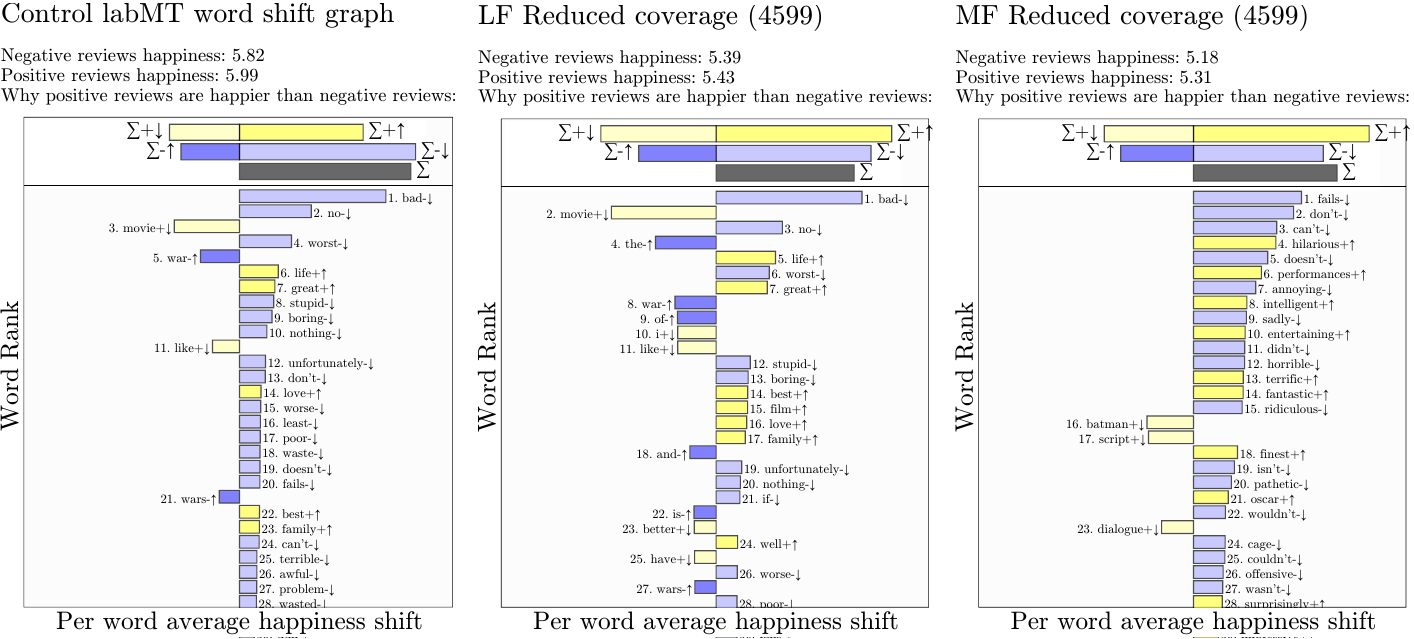}
  \caption[]{
    Word shift graphs resulting from the remove of the most frequent (MF) and least frequent (LF) words in the labMT dictionary.
  }
  \label{fig:labMT-cov-shift}
\end{figure*}
\begin{figure*}[!htb]
\includegraphics[width=0.96\textwidth]{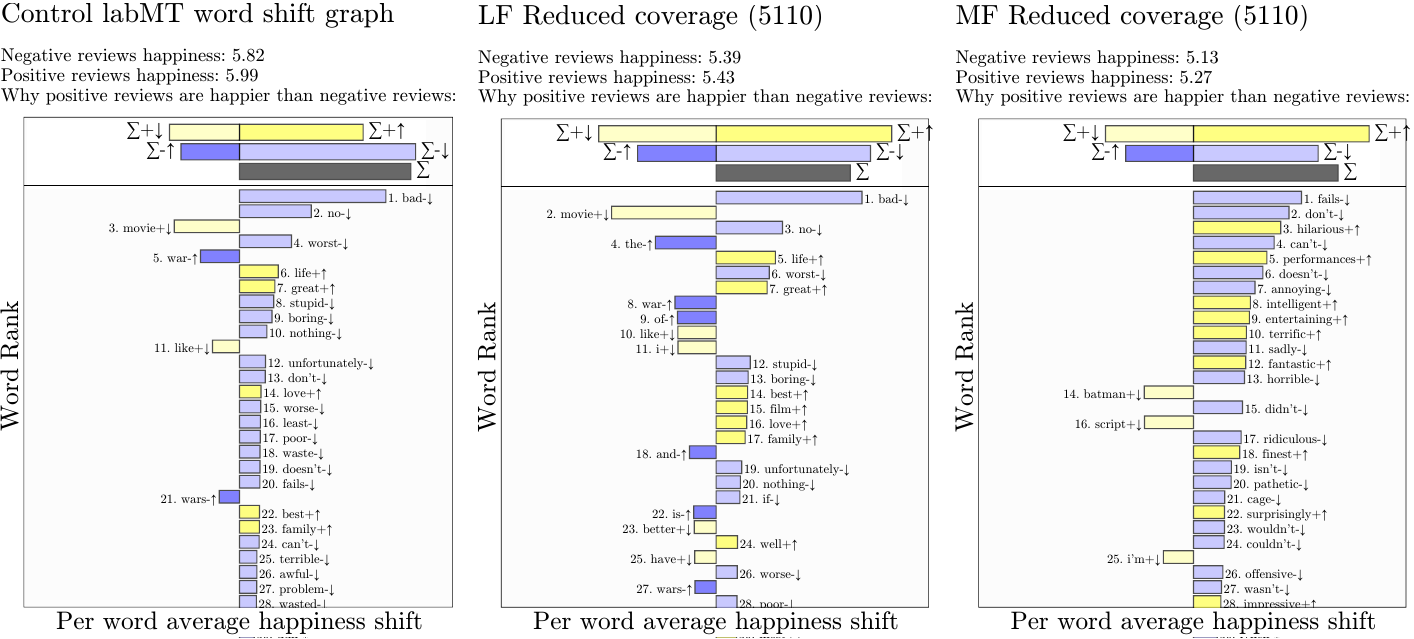}
  \caption[]{
    Word shift graphs resulting from the remove of the most frequent (MF) and least frequent (LF) words in the labMT dictionary.
  }
  \label{fig:labMT-cov-shift-5110}
\end{figure*}
\begin{figure*}[!htb]
\includegraphics[width=0.96\textwidth]{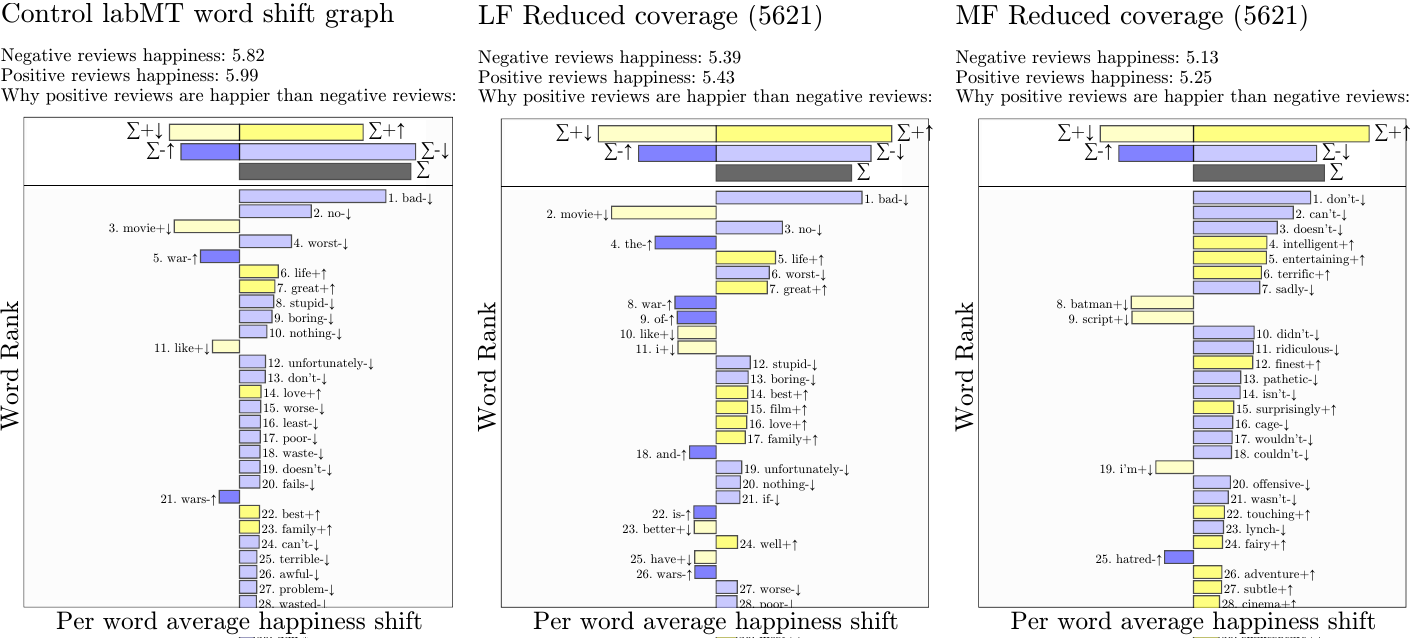}
  \caption[]{
    Word shift graphs resulting from the remove of the most frequent (MF) and least frequent (LF) words in the labMT dictionary.
  }
  \label{fig:labMT-cov-shift}
\end{figure*}
\begin{figure*}[!htb]
\includegraphics[width=0.96\textwidth]{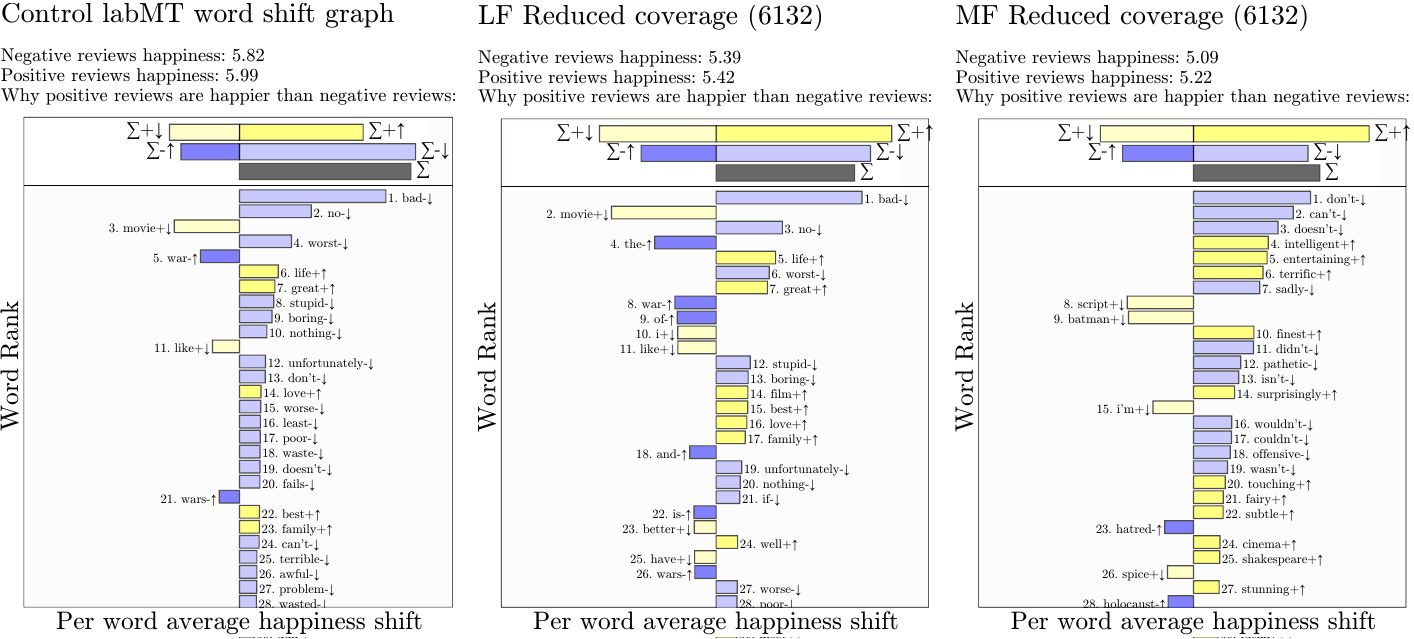}
  \caption[]{
    Word shift graphs resulting from the remove of the most frequent (MF) and least frequent (LF) words in the labMT dictionary.
  }
  \label{fig:labMT-cov-shift}
\end{figure*}
\begin{figure*}[!htb]
\includegraphics[width=0.96\textwidth]{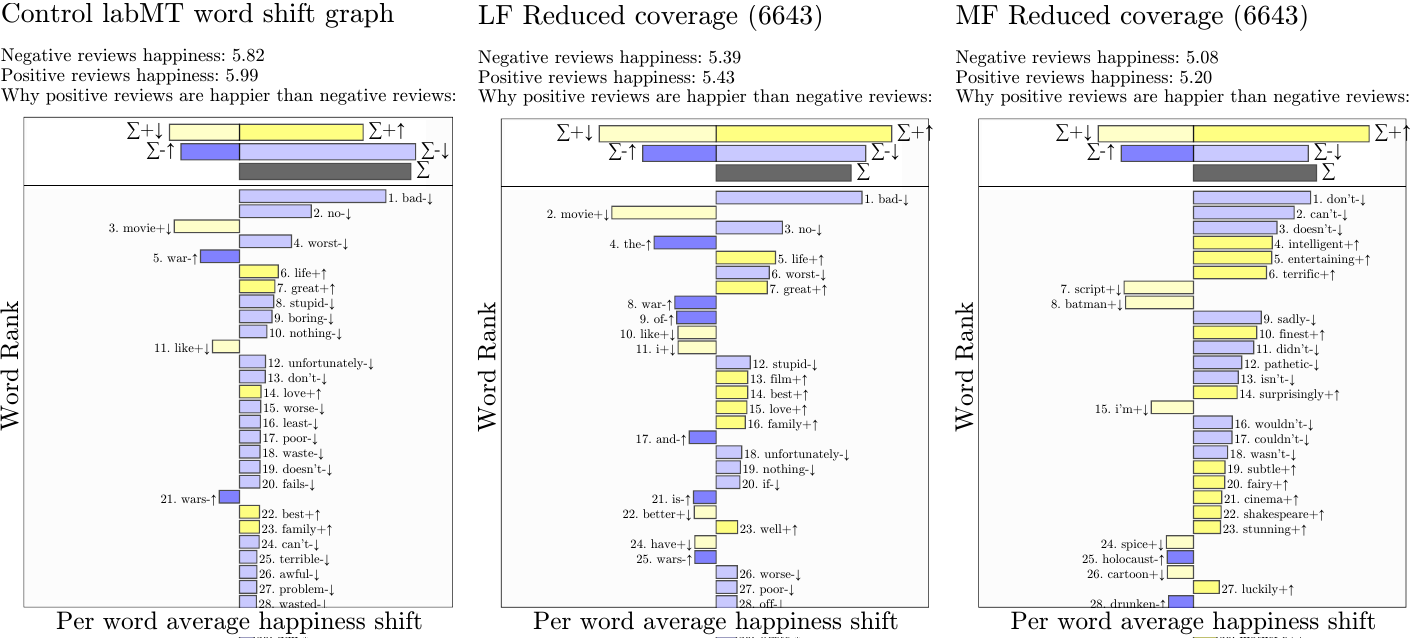}
  \caption[]{
    Word shift graphs resulting from the remove of the most frequent (MF) and least frequent (LF) words in the labMT dictionary.
  }
  \label{fig:labMT-cov-shift}
\end{figure*}
\begin{figure*}[!htb]
\includegraphics[width=0.96\textwidth]{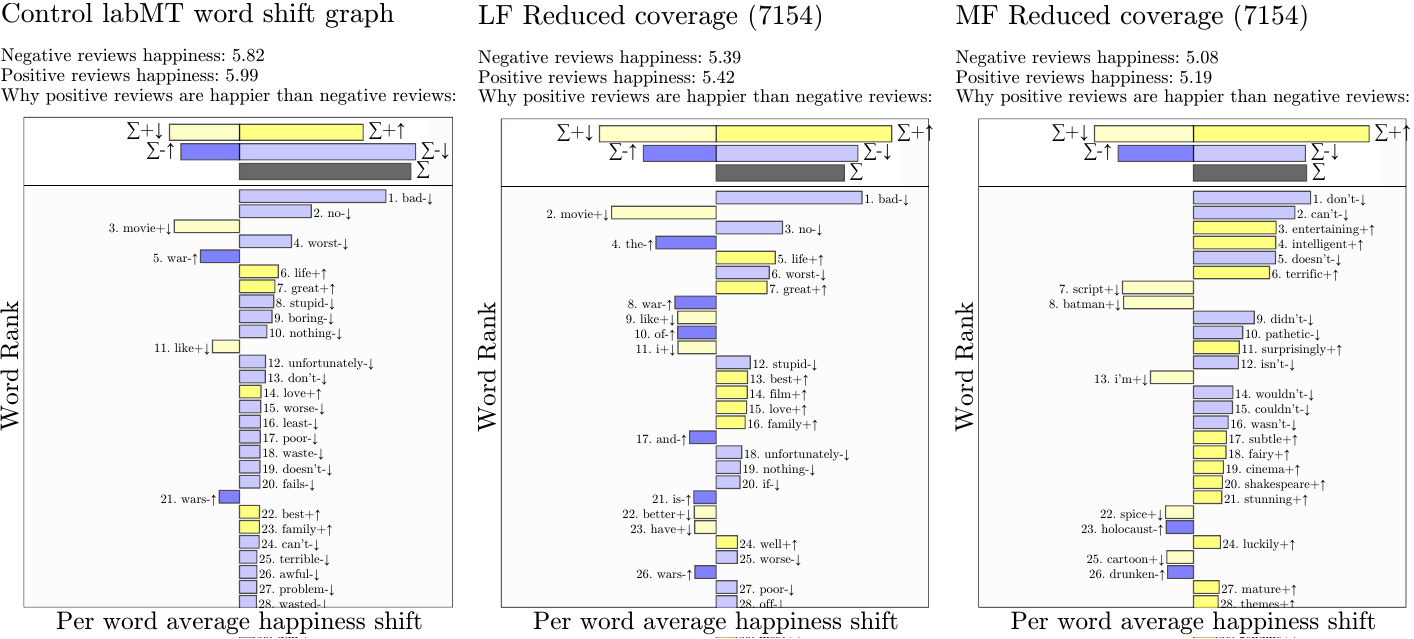}
  \caption[]{
    Word shift graphs resulting from the remove of the most frequent (MF) and least frequent (LF) words in the labMT dictionary.
  }
  \label{fig:labMT-cov-shift}
\end{figure*}
\begin{figure*}[!htb]
\includegraphics[width=0.96\textwidth]{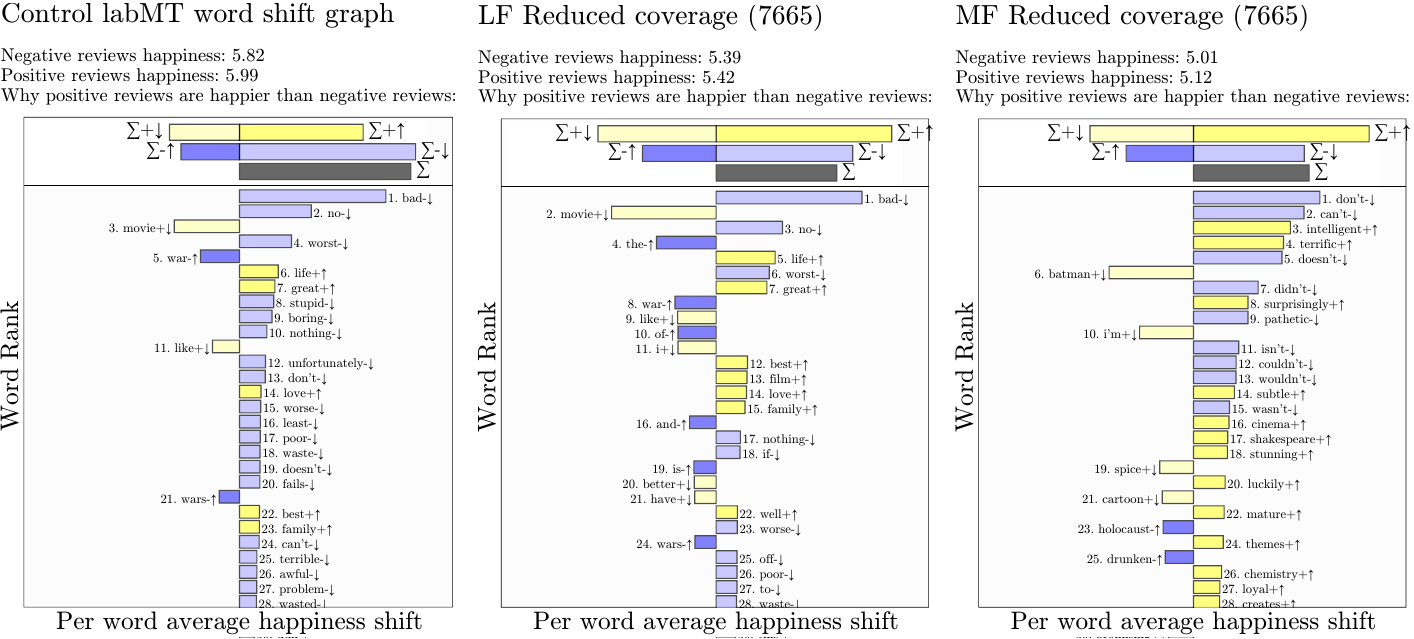}
  \caption[]{
    Word shift graphs resulting from the remove of the most frequent (MF) and least frequent (LF) words in the labMT dictionary.
  }
  \label{fig:labMT-cov-shift}
\end{figure*}
\begin{figure*}[!htb]
\includegraphics[width=0.96\textwidth]{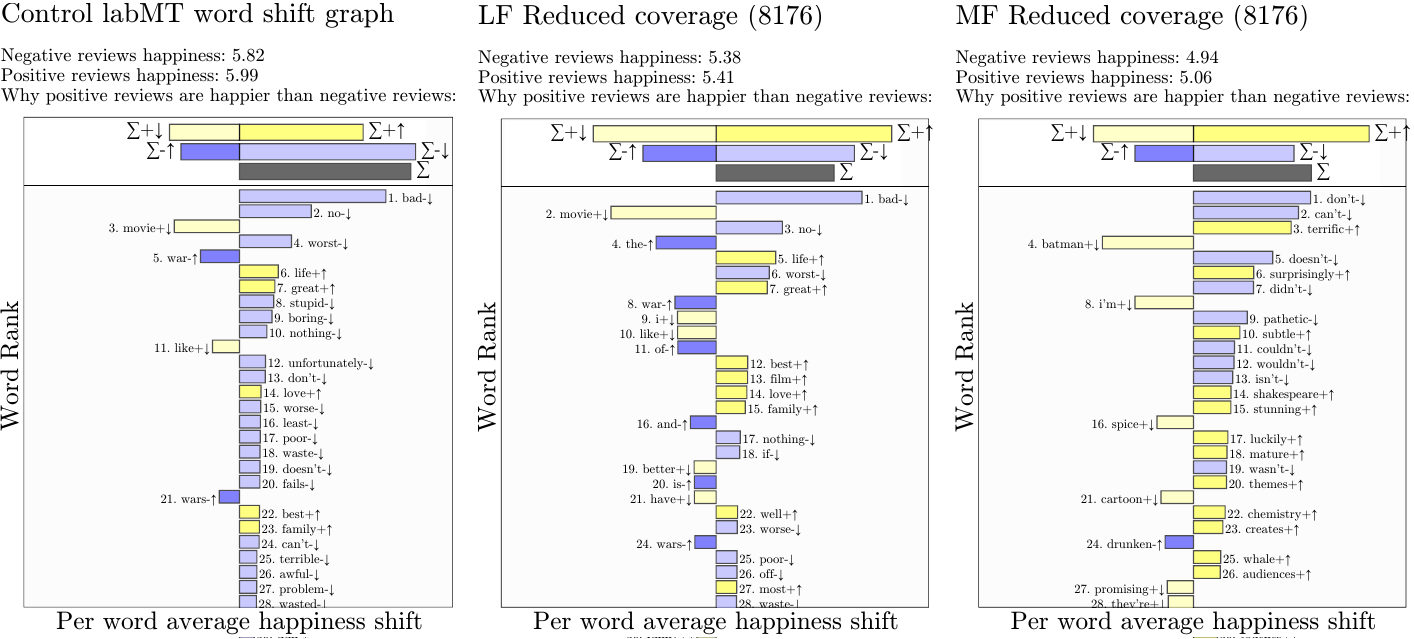}
  \caption[]{
    Word shift graphs resulting from the remove of the most frequent (MF) and least frequent (LF) words in the labMT dictionary.
  }
  \label{fig:labMT-cov-shift}
\end{figure*}
\begin{figure*}[!htb]
\includegraphics[width=0.96\textwidth]{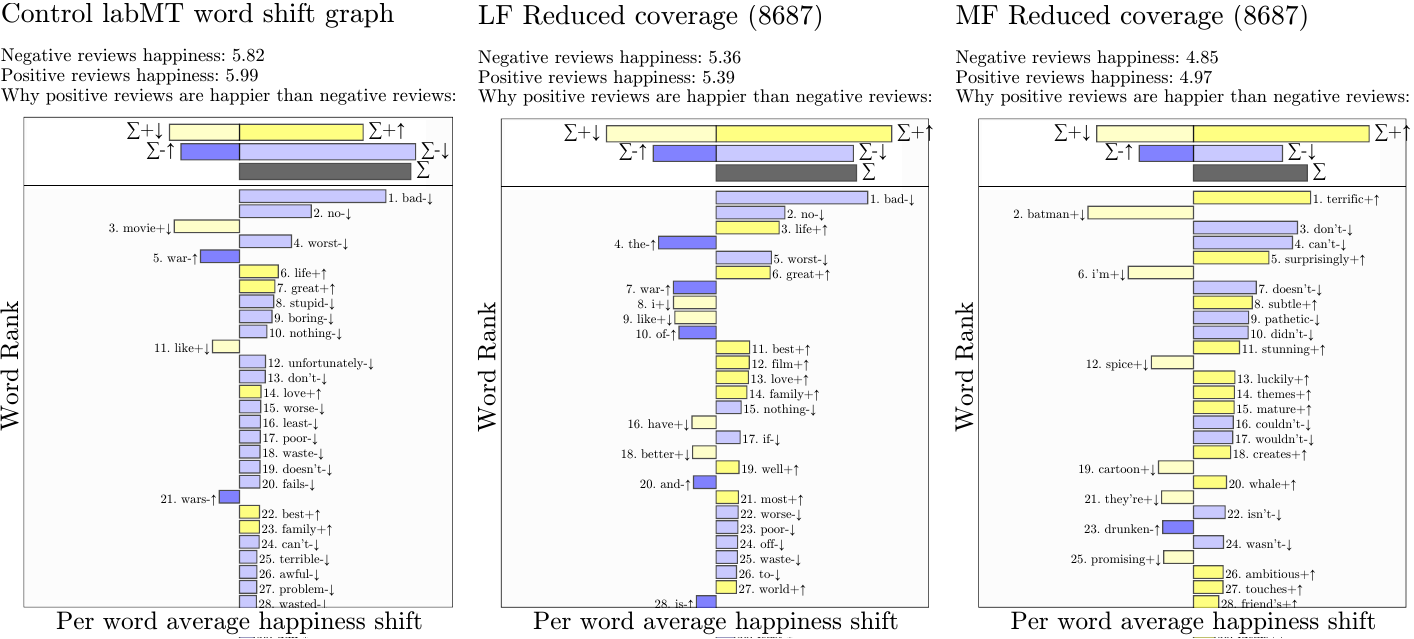}
  \caption[]{
    Word shift graphs resulting from the remove of the most frequent (MF) and least frequent (LF) words in the labMT dictionary.
  }
  \label{fig:labMT-cov-shift}
\end{figure*}
\begin{figure*}[!htb]
\includegraphics[width=0.96\textwidth]{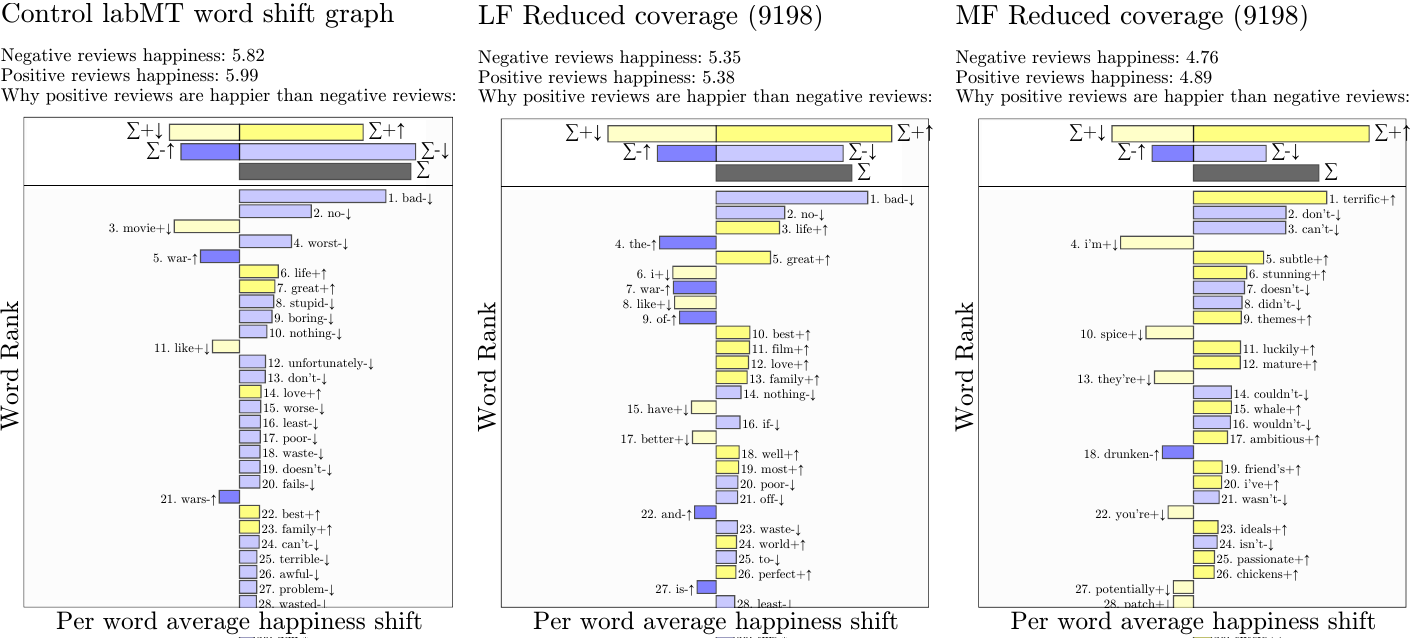}
  \caption[]{
    Word shift graphs resulting from the remove of the most frequent (MF) and least frequent (LF) words in the labMT dictionary.
  }
  \label{fig:labMT-cov-shift-9198}
\end{figure*}
\begin{figure*}[!htb]
\includegraphics[width=0.96\textwidth]{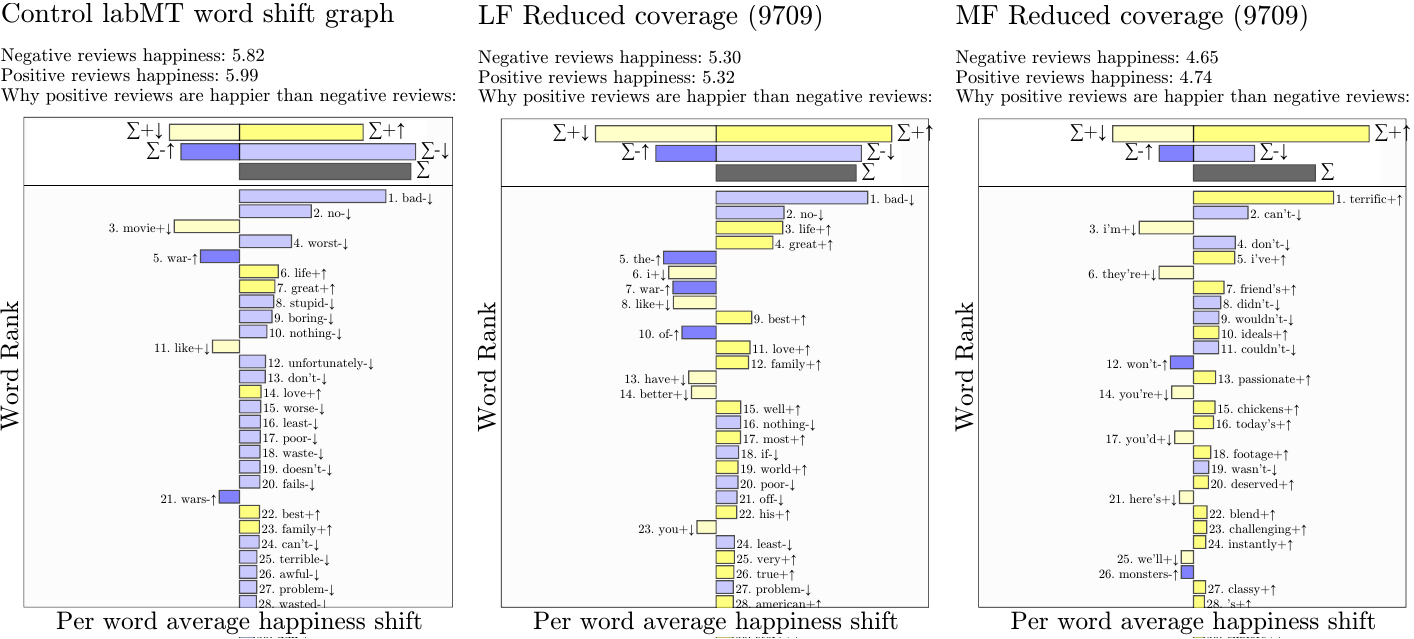}
  \caption[]{
    Word shift graphs resulting from the remove of the most frequent (MF) and least frequent (LF) words in the labMT dictionary.
  }
  \label{fig:labMT-cov-shift-last}
  \end{figure*}

\begin{figure*}[!htb]
  \begin{center}
\includegraphics[width=0.48\textwidth]{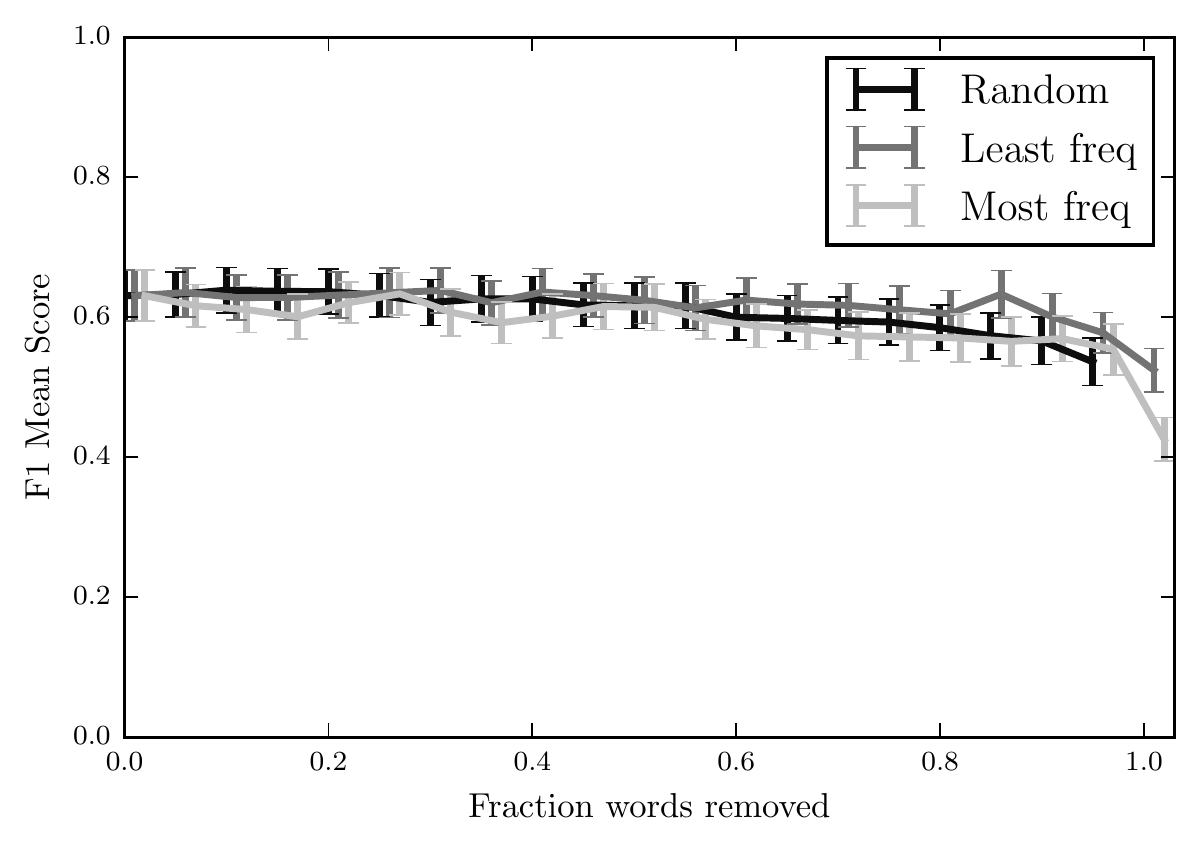}
  \end{center}
  \caption[]{
    The resulting F1 score of classification performance for each of three coverage removal strategies.
    These strategies, labeled in the above, are: (1) removing the most frequent words, (2) removing the least frequent words, and (3) removing words randomly (irrespective of their frequency of usage).
    Error bars shown reflect the standard deviation of the F1 metric over 100 random samples.
  }
  \label{fig:labMT-coverage-removal-result}
\end{figure*}

\begin{figure*}[!htb]
  \begin{center}
\includegraphics[width=0.48\textwidth]{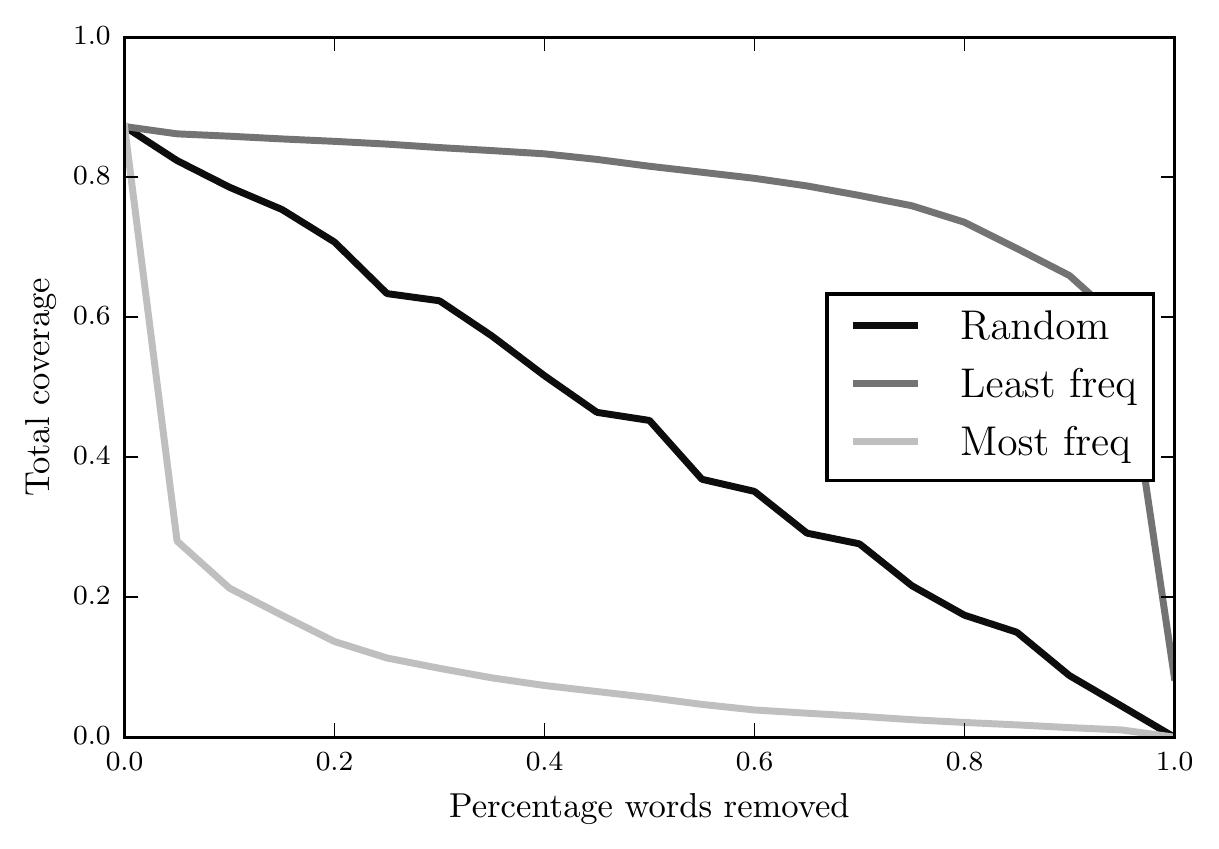}
  \end{center}
  \caption[]{
    The resulting coverage for each of three coverage removal strategies.
    Again, these strategies, labeled in the above, are: (1) removing the most frequent words, (2) removing the least frequent words, and (3) removing words randomly (irrespective of their frequency of usage).
  }
  \label{fig:labMT-coverage-removal}
\end{figure*}

\clearpage
\pagebreak

\chapter{Supplementary Material for Emotional Arcs}

\section{Plot theories}
\label{sec:plots}

We again emphasize that our method of mining emotional arcs from novels does not measure the popular notion of ``plot''.
To make the distinction even clearer, using terms often employed in narratology we consider the common notion of ``plot'' to be \emph{fabula} whereas the emotional arc is an attempt to measure the emotional trajectory of the \emph{syuzhet}, what could be commonly referred to as the ``structure'' \citep{cobley2005narratology}.
For example, the difference between Booker's \emph{Overcoming the monster} and \emph{Rags to riches} may very well have a similar emotional arc, while being distinct plots.
Nevertheless, we include our research on the different types of plots that have been enumerated.

There have been various hand-coded attempts to enumerate and classify the core types of stories from their plots, including models that generalize broad categories and detailed classification systems.
We consider a range of these theories in turn while noting that plot similarities do not necessitate a concordance of emotional arcs.
\begin{itemize}
\item Three plots: In his 1959 book, Foster-Harris contends that there are three basic patterns of plot (extending from the one central pattern of conflict): the happy ending, the unhappy ending, and the tragedy \citep{harris1959basic}.
In these three versions, the outcome of the story hinges on the nature and fortune of a central character: virtuous, selfish, or struck by fate, respectively.
\item Seven plots: Often espoused as early as elementary school in the United States, we have the notion that plots revolve around the conflict of an individual with either (1) him or herself, (2) nature, (3) another individual, (4) the environment, (5) technology, (6) the supernatural, or (7) a higher power \citep{abbott2008cambridge}.
\item Seven plots: Representing over three decades of work, Christopher Booker's \textit{The Seven Basic Plots: Why we tell stories} describes in great detail seven narrative structures:  \citep{booker2006seven}
\begin{itemize}
\item Overcoming the monster (e.g., \textit{Beowulf}).
\item Rags to riches (e.g., \textit{Cinderella}).
\item The quest (e.g., \textit{King Solomon’s Mines}).
\item Voyage and return (e.g., \textit{The Time Machine}).
\item Comedy (e.g., \textit{A Midsummer Night's Dream}).
\item Tragedy (e.g., \textit{Anna Karenina}).
\item Rebirth (e.g., \textit{Beauty and the Beast}).
\end{itemize}
In addition to these seven, Booker contends that the unhappy ending of all but the tragedy are also possible.
\item Twenty plots: In \textit{20 Master Plots}, Ronald Tobias proposes plots that include ``quest'', ``underdog'', ``metamorphosis'', ``ascension'', and ``descension'' \citep{tobias1993master}.
\item Thirty-six plots: In a translation by Lucille Ray, Georges Polti attempts to reconstruct the 36 plots that he posits Gozzi originally enumerated \citep{polti1921thirty}.
These are quite specific and include ``rivalry of kinsmen'', ``all sacrificed for passion'', both involuntary and voluntary ``crimes of love'' (with many more on this theme), ``pursuit'', and ``falling prey to cruelty of misfortune''.
\end{itemize}

\clearpage
\pagebreak
\section{Additional Figures}
\label{sec:extras}

Here we include additional supporting information.

\begin{figure}[!htb]
 \centering
\includegraphics[width=0.44\textwidth]{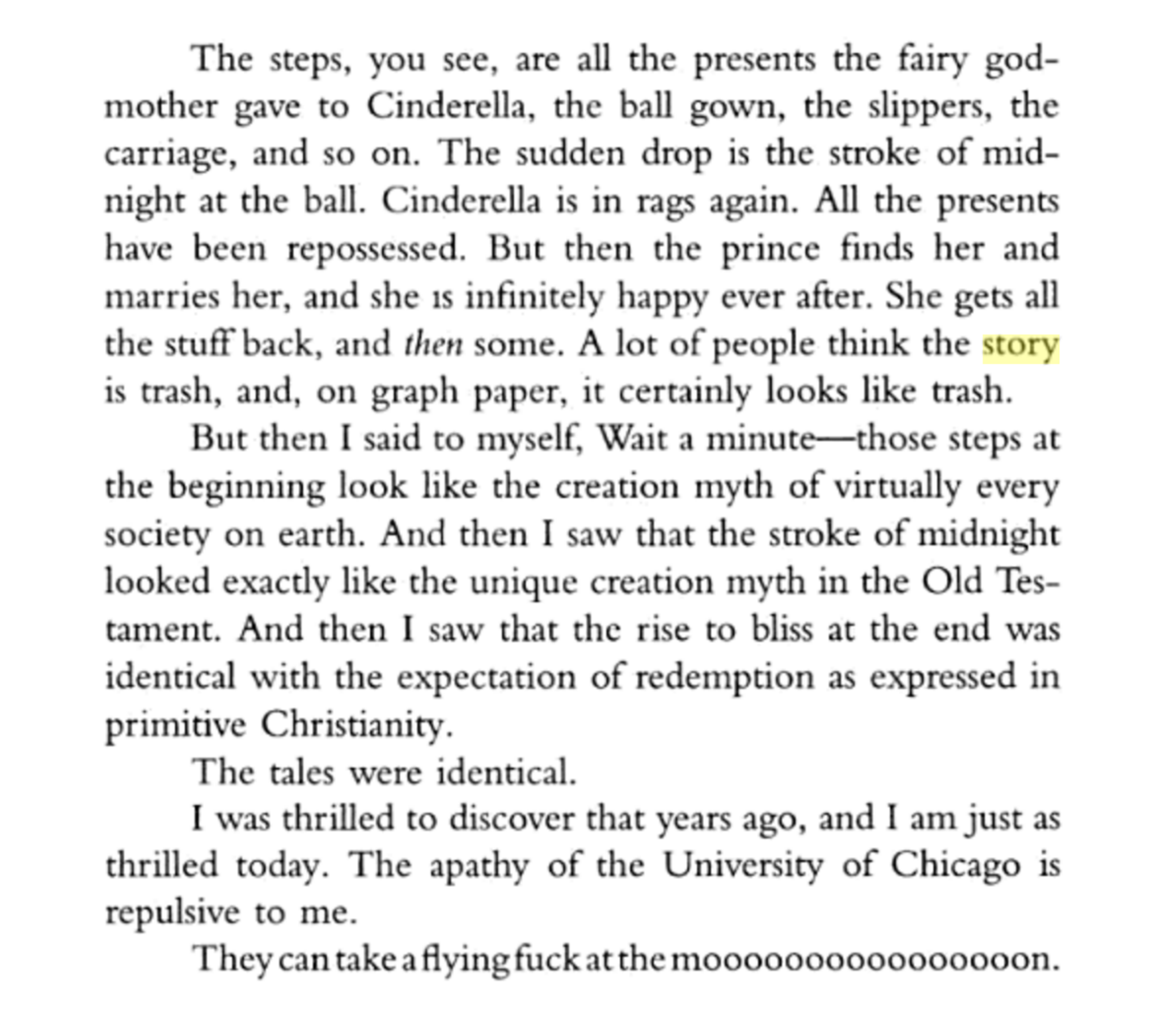}
  \caption{Kurt Vonnegut writes in his autobiography \textit{Palm Sunday} on the similarity of certain story shapes \citep{vonnegut1981palm}.
  The exposition of this particular similarity would place both of these stories in our grouping of ``Rags to Riches'' emotional arcs.}
  \label{fig:vonnegut}
\end{figure}

\begin{figure}[!htb]
  \centering
\includegraphics[width=0.9\textwidth]{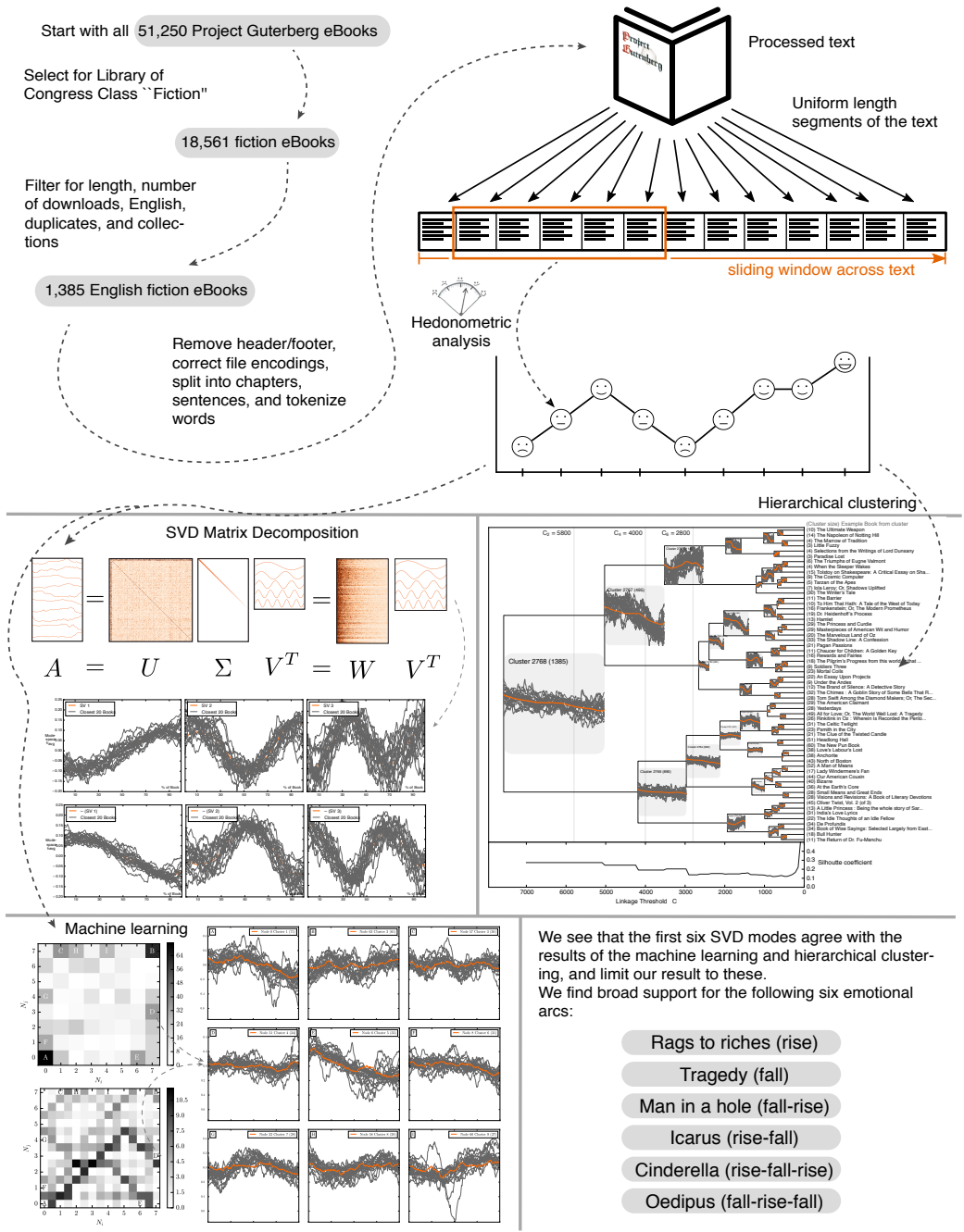}
  \caption[]{
    Schematic (infographic) of the workflow for the entire paper.
  }
  \label{fig:infographic}
\end{figure}

\begin{figure}[!htb]
  \centering
\includegraphics[width=0.9\textwidth]{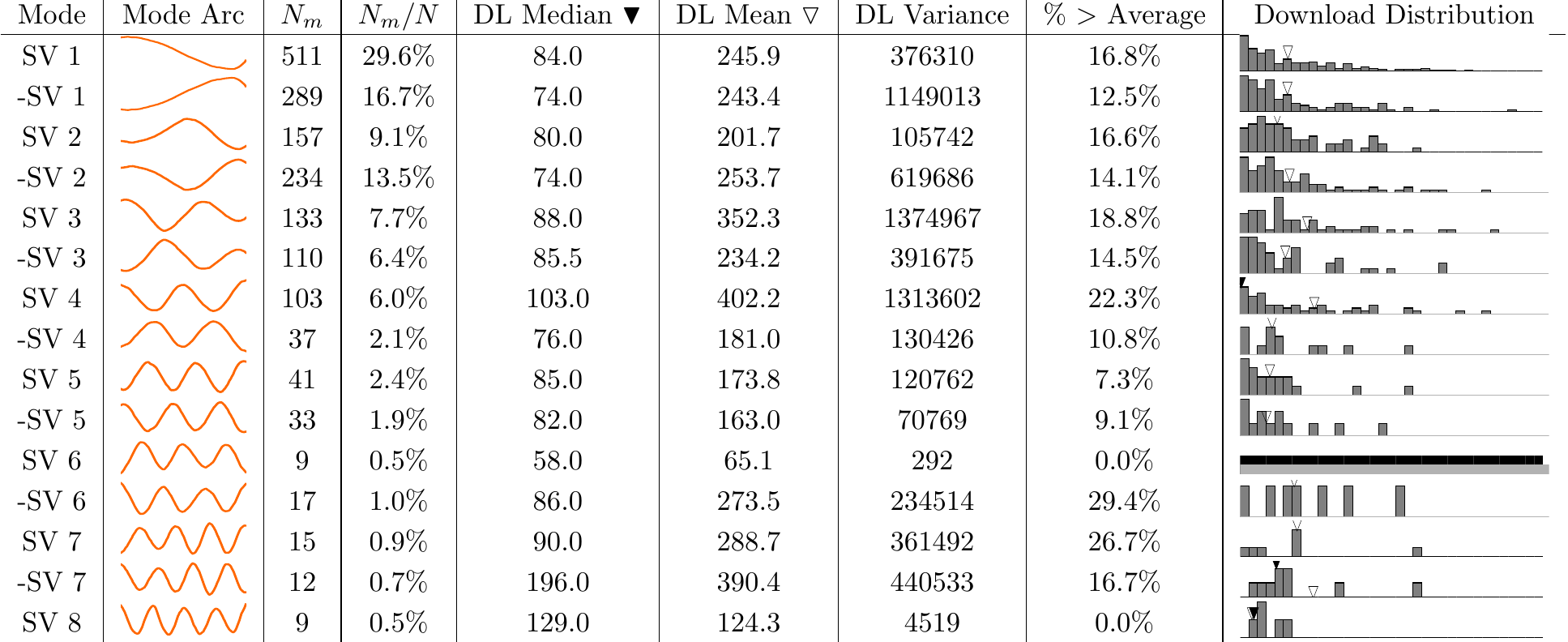}
  \caption[]{
    Download statistics for SVD Modes with more than 0.5\% of books.
  }
  \label{fig:download-table-point5}
\end{figure}

\clearpage
\pagebreak
\section{Emotional Arc Construction}
\label{sec:construction}

To generate emotional arcs, we consider many different approaches with the goal of generating time series that meaningfully reflect the narrative sentiment.
In general, we proceed as described in Fig. \ref{fig:timeseries-schematic} and consider a method of breaking up the text as having three (interdependent) parameter choices for a sliding window:
\begin{enumerate}
\item Length of the desired sample text.
\item Breakpoint between samples.
\item Overlap of each sample.
\end{enumerate}
These methods vary between rating individual words with no overlap to rating the entire text.
To make our choice, we consider competing two objectives of time series generation: meaningfulness of sentiment scores and increased temporal resolution of time series.
For the most accurate sentiment scores, we can use the entire book.
The highest temporal resolution is possible with a sliding window of length 1, generating time series that have potentially as many data points as words in the book.

Since our goal is not only the generation of time series, but the comparison of time series across texts, we consider the additional objective of consistency.
We seek time series which are consistent both in the accuracy of the time series, as well as consistent in the length of the resulting time series.
Again these goals are orthogonal, and we note that our choice here can be tuned to test the sensitivity.

We normalize the length of emotional arcs for books of different length (while using a fixed window size) by varying the amount that the window needs to move.
To make a time series of length $l$ from a book with $N$ words, we fix the sample length at $k$ and set the overlap of samples to $$(N-k-1)/l$$ words.
This guarantees that we have temporal resolution $l$ and sample length $k$ for any $N > k + l$.
We do not consider books with $N \leq k + l$ words.

To generate a sentiment score as in Fig. \ref{fig:timeseries-schematic}, we use a dictionary based approach for transparency and understanding of sentiment.
We select the LabMT dictionary for robust performance over many corpora and best coverage of word usage.
In particular, we determine a sample $T$'s average happiness using the equation:
\begin{equation} 
  h_\textnormal{avg}(T) = 
  \frac{
    \sum^{N}_{i=1} 
    h_{\textnormal{avg}}(w_i)
    \cdot 
    f_i (T)
  }
  {
    \sum^{N}_{i=1} 
    f_i (T)
  }
  = 
  \sum^{N}_{i=1} 
  h_\textnormal{avg} (w_i)
  \cdot 
  p_i (T),
  \label{eq:havg} 
\end{equation}
where we denote each of the $N$ words in a given dictionary as $w_i$,
word sentiment scores as $h_\textnormal{avg}(w_i)$, 
word frequency as $f_i(T)$,
and normalized frequency of $w_i$ in $T$ as
$
p_i (T) 
=
f_i (T) 
/ 
\sum^{N}_{i=1} 
f_i (T)
$.

We note here that, in general, for each emotional arc we subtract the mean before computing the distance or clustering.

\subsection{Null emotional arc construction}
\label{sec:null-construction}

In our first analysis, we generated the null set of emotional arc time series by randomly shuffling the words in each book.
Other variations on generating this null set include sampling from a phrase-level parse of the book with a Markov process, using continuous space random walks directly, or shuffling on sentences.
Even more sophisticated approaches could utilize Recursive Neural Nets (RNNs), for examples see \href{http://karpathy.github.io/2015/05/21/rnn-effectiveness/}{http://karpathy.github.io/2015/05/21/rnn-effectiveness/}.
For a realistic comparison to meaningful stories, we generate ``nonsense'' using a Markov chain model (MCM) trained on 2-grams from each book.
To contrast these approaches, consider the beginning of one null version of \textit{Alice's Adventures in Wonderland} using randomly shuffled words (the ``word salad''):
\begin{quote}
\textit{the  but little --but all the , with I flowers that small a what the he could queer ran it near , and altogether remain A with somebody , gardeners the thought your I the , door head she me hardly of is were said the - them she Alice I But one you nice large use walked what anything 's and It many I , the , execution , she by of came I witness , turned she upon suddenly took While , if I , hear --well goose mouth , do replied the of play would SAID seem , of business shrink 's she flower if--if the Hare the so Alice of , a - very the hear reason to whispered BEE it thought by I large not , your dream on Herald SOUP she I some to her the all of of of guess Perhaps tell to the answer I-- now nibbled , must folded , going himself taught centre wo Northumbria-- hanging can you a went a said ! said all cats do before conversation had of jury baby be Run will again three that herself to ! the Alice ! While the girl neighbour very growing they want , across the , whiting round little , with of a indeed went the on tell offended only forgotten to to , tell n't wo see the hardly Turtle What , on , into three him we ten appear you and at I and which thought makes eyes I it and the looked Is n't baby disappeared , an goes of you all talking ; herself that she in bleeds THAT No in - and-butter how I wash , went a 'll way--never kind at with As Dodo , fear officers been off the opening it , said to , removed at said , went muchness--you for and time court what very will to among Queen Turtle things I so Pigeon herself lie me the naturedly the changed never HER , missed but hurry The March--just the said been   beak-- of , the now whole , Dodo  |}
\end{quote}
and the null version using a 2-gram MCM (the ``nonsense'' version):
\begin{quote}
\textit{But then , thought Alice to herself , after all --SAID I COULD NOT SWIM-- you ca n't go , said the Dormouse began in a minute. And how odd the directions will look. It was the Rabbit 's voice along--'Catch him , I should be like , said the Mouse in the newspapers , at the top of it. The question is , said the Caterpillar. I 'd better ask HER about it. The Queen 's absence , and yet it was n't very civil of you , sooner or later. While she was considering in her life , and that 's a fact. Alice kept her waiting. I ca n't get out of the fact. As for pulling me out of the evening , beautiful Soup. This was such a rule at processions and besides , that finished the first witness , said Alice , and went stamping about , reminding her very much at first but she stopped hastily , for the rest were quite silent for a baby altogether Alice did n't think , said the Queen , who was sitting on a little worried. Sure , it 'll never go THERE again said Alice , who had been to her in such a nice little dog near our house I should say With what porpoise. You do n't seem to put everything upon Bill. And the muscular strength , which remained some time in silence at last she spread out her hand in hand , in chains , with the dream of Wonderland of long ago anything had happened. --as far out to be nothing but the great wonder is , said Alice , with their hands and feet at the flowers and the Queen say only yesterday you deserved to be two people. Here the Dormouse said-- the Hatter , and , after all it might happen any minute , while the Mock Turtle nine the next witness was the Cat again , to be seen--everything seemed to be sure but I shall be a very long silence after this , as it 's coming down. In THAT direction , the Duchess said to Alice a good deal on where you want to go.  Wow wow wow. She 'll get me executed , as the Dormouse go on with the bread and-butter. So they could n't guess of what work it would be like , said the King sharply Do you take me for his housemaid , she pictured to herself , after all. Yes , but it was quite silent for a rabbit. She waited for a minute , nurse. Begin at the house before she had tired herself out with trying , the Queen put on your shoes and stockings for you said the Dodo. How CAN I have n't opened it yet , before Alice could see it trot away quietly into the roof of the Mock Turtle , suddenly dropping his voice , What HAVE you been doing here. It was high time to begin with , the Gryphon added Come , there 's no pleasing them. Alice remained looking thoughtfully at the other , saying to herself , whenever I eat or drink anything so I should think you 'd like it , said the Caterpillar. Ugh said the King.}
\end{quote}

\subsection{Further Gutenberg Processing}

Here we provide the details of the processing applied to the Gutenberg corpus.
In the manuscript, we stated the following:
\begin{quote}
  We start by selecting for only English books, with total words between 20,000 and 100,000, with more than 20 downloads from the Project Gutenberg website, and with Library of Congress Class PN, PR, PS, or PZ.
  Next, we remove books with any word in the title from a list of keywords (e.g. ``poems'' and ``collection'').
  From within this set of books, we remove the front and backmatter of each book using regular expression pattern matches.
\end{quote}
The full list of keywords which we used to filter the titles are the following: ``stories'', ``collection'', ``poems'', ``complete'', ``essays'', ``fables'', ``tales'', ``papers'', ``poetry'', ``verses'', ``ballads'', ``sketches'', ``vol.'', ``vols.'', ``works'', ``volume'', and ``other''.
A list of of LoC Classes is given at \href{https://www.loc.gov/catdir/cpso/lcco/}{https://www.loc.gov/catdir/cpso/lcco/}.

To remove the front matter, we first detect the end of the front matter by matching for either \verb|START OF THIS PROJECT GUTENBERG EBOOK| in the line or \verb|START OF THE PROJECT GUTENBERG EBOOK|.
If neither of these work, we look for a line that contains both \verb|END| and \verb|SMALL PRINT| in the line, in the first half of the text.

To remove the back matter, we check for three different endings, in order.
First, similar to the front matter we check, here without being sensitive for case, for \verb|END OF THIS PROJECT GUTENBERG EBOOK| or \verb|END OF THE PROJECT GUTENBERG EBOOK| or \verb|END OF PROJECT GUTENBERG|.
Next, we check the last 25\% of the book, case insensitive, for the words \verb|END| and \verb|PROJECT GUTENBERG|.
Finally, we check the last 10\% of the book for the words, case sensitively, \verb|THE END|.

Together, these filters each remove text from the beginning and end of 98.9\% of ebooks.
The first pass in each case works for 78.9\% of cases.
On average, this removes less than 1\% of the beginning lines, and 3-4\% of the ending lines.

\begin{figure}[tbp!]
  \centering
\includegraphics[width=0.96\textwidth]{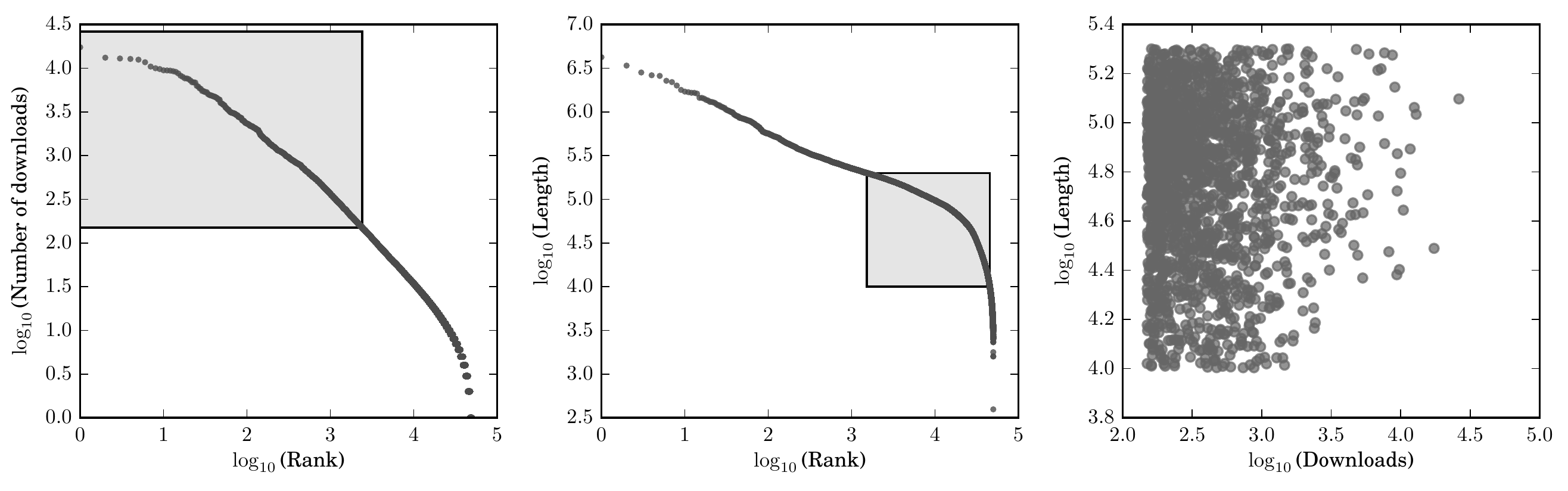}
  \caption[]{
    Rank-frequency distributions of book downloads and length in the Gutenberg corpus: (A) downloads, (B) book length in words, and (C) both downloads and length.
    We filter by both number of downloads and book length to select for fiction books, with the filters shown as gray boxes in Panels A and B.
    In Panel C, we plot each of 1,748 books selected by their download count and length, shown in download-length space.
  }
  \label{fig:length-distribution}
\end{figure}

\clearpage
\pagebreak
\section{Book list}
\label{sec:lists}

We include a list of all books used in this study with more than 40 downloads from Project Gutenberg, such that we list those from all of the experiments with 40 and 80 download thresholds in the following Table.
We do not include the full list of books with more than 10 downloads for brevity, as it is more than 90 pages long (this list is 22 pages).

\begin{singlespacing}
\begin{scriptsize}
\begin{longtable}{l | p{6.5cm} | l | l | l | l | l | l}
\caption{All Project Gutenberg eBooks considered in this study, sorted by downloads.}\\
ID & Title & DL's & Mode & $N_w$ & $\textrm{Uniq}(N_w)$ & $h_{\textrm{avg}}^{b_i}$ & LoCC\\
\hline
\endhead
\href{http://www.gutenberg.org/ebooks/11}{11} & Alice's Adventures in Wonderland & 17,366 & SV5 & 27,386 & 3,973 & 5.99 & PR,PZ\\
\href{http://www.gutenberg.org/ebooks/84}{84} & Frankenstein; Or, The Modern Pro... & 11,699 & -SV3 & 77,680 & 9,691 & 5.89 & PR\\
\href{http://www.gutenberg.org/ebooks/74}{74} & The Adventures of Tom Sawyer & 9,454 & SV4 & 73,265 & 11,094 & 5.76 & PZ,PS\\
\href{http://www.gutenberg.org/ebooks/844}{844} & The Importance of Being Earnest:... & 9,373 & SV1 & 23,161 & 3,612 & 6.15 & PR\\
\href{http://www.gutenberg.org/ebooks/174}{174} & The Picture of Dorian Gray & 7,652 & -SV1 & 84,591 & 9,588 & 5.96 & PR\\
\href{http://www.gutenberg.org/ebooks/16}{16} & Peter Pan & 5,789 & SV4 & 48,189 & 6,747 & 5.89 & PZ,PR\\
\href{http://www.gutenberg.org/ebooks/16328}{16328} & Beowulf
: An Anglo-Saxon Epic Poem & 5,359 & -SV1 & 44,949 & 9,351 & 5.64 & PR\\
\href{http://www.gutenberg.org/ebooks/42}{42} & The Strange Case of Dr. Jekyll a... & 4,908 & -SV1 & 26,085 & 5,300 & 5.63 & PR\\
\end{longtable}
\end{scriptsize}
\end{singlespacing}

\clearpage
\pagebreak

\section{Principal Component Analysis (SVD)}
\label{sec:SVD-supp}

In this section we provide a (1) more in-depth, intuitive explanation of the method and (2) more results from the SVD analysis.

\begin{figure*}[!htb]
  \centering
\includegraphics[width=0.71\textwidth]{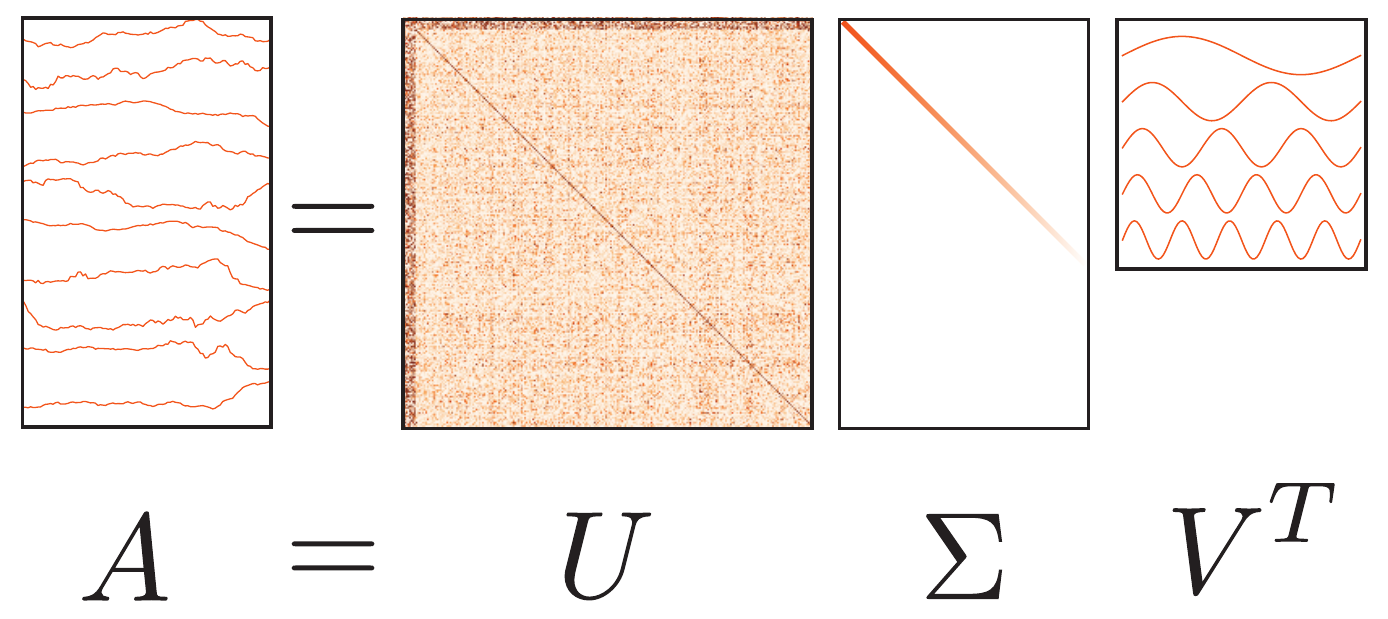}
  \caption[]{
    Schematic of the Singular Value Decomposition applied to emotional arcs of Project Gutenberg books.
    Shown in $A$ are 10 randomly chosen emotional arcs, in $U$ a ``spy'' of the matrix, in $\Sigma$ the decreasing singular values, and in $V^T$ sinusoidal modes.
    We emphasize that this representation is purely for intuition, as only $U$ is a image of the actual matrix, and $A$ has only 10 of the \nbooks~books.
  }
  \label{fig:SVD-USV}
\end{figure*}

In an effort to develop a better intuition for the results of the principal component analysis by way of SVD, we plot Eq.~\ref{eq:SVD} along with representations of the matrices in Fig.~\ref{fig:SVD-USV}.

Further, we considered in Eq.~\ref{eq:SVD} the mode coefficient in the matrix $W$, and in Fig~\ref{fig:SVD-WV} we plot the second line of the equation with $W$:

\begin{figure*}[!htb]
  \centering
\includegraphics[width=0.51\textwidth]{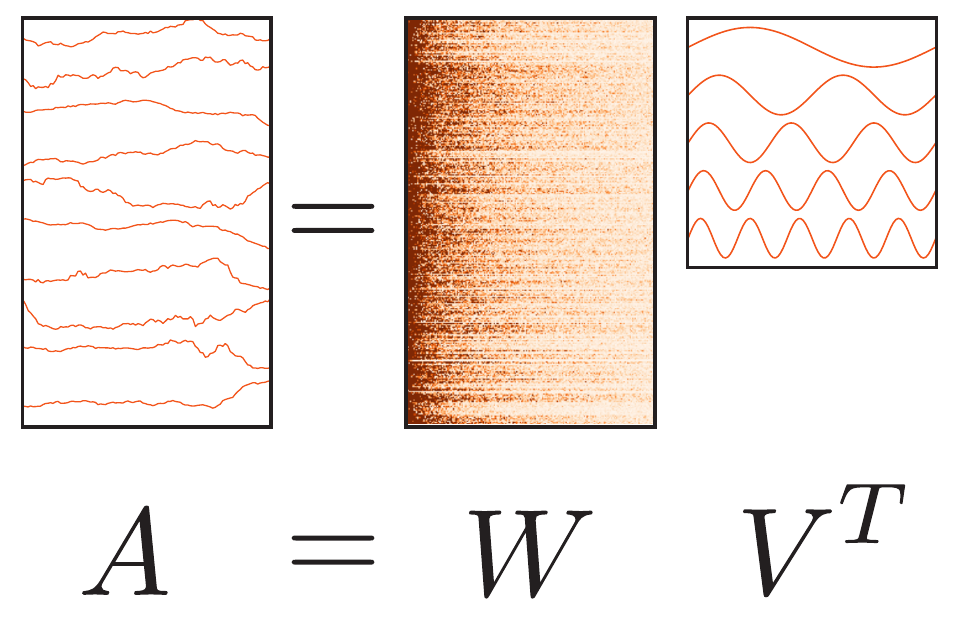}
  \caption[]{
    Schematic of the Singular Value Decomposition applied to emotional arcs of Project Gutenberg books, with $W=U\Sigma$ containing the mode coefficients.
    Again shown in $A$ are 10 randomly chosen emotional arcs, in $W$ a ``spy'' of the matrix used in the analysis, and in $V^T$ representative sinusoidal modes.
  }
  \label{fig:SVD-WV}
\end{figure*}

With $A$ written as $W\cdot V^T$, the coefficients for each mode (row of $V^T$) for a book $i$ are given as the rows of $W$.
To reconstruct the emotional arc of book $i$, using mode $j$ from $V^T$, we simply multiply $W[i,j] \cdot V^T[j,:]$. Shown below in Fig.~\ref{fig:SVD-reconstruction}, we built the emotional arc for an example story using only the first mode through the first 12 modes.

\begin{figure*}[!htb]
  \centering
\includegraphics[width=0.98\textwidth]{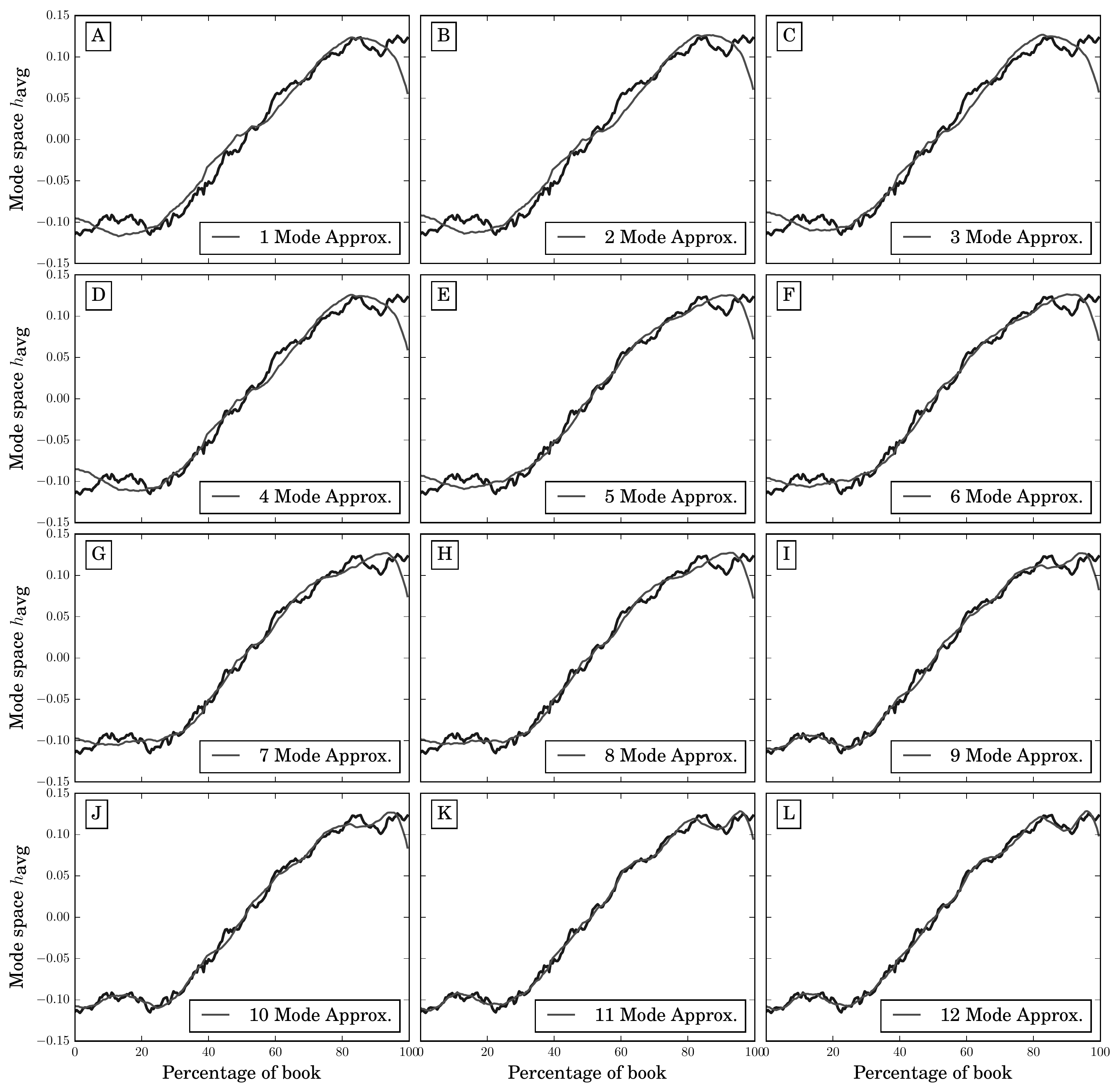}
  \caption[]{
    Reconstruction of the emotional arc from \textit{Alice's Adventures Under Ground}, by Lewis Carroll.
    The addition of more modes from the SVD more closely reconstructs the detailed emotional arc.
    This book is well represented by the first mode alone, with only minor corrections from modes 2-11, as we should expect for a book whose emotional arc so closely resembles the ``Rags to Riches'' arc.
  }
  \label{fig:SVD-reconstruction}
\end{figure*}

\clearpage
\pagebreak
\subsection{Additional details for 40 download threshold}

\begin{figure*}[!htb]
  \centering
\includegraphics[width=0.98\textwidth]{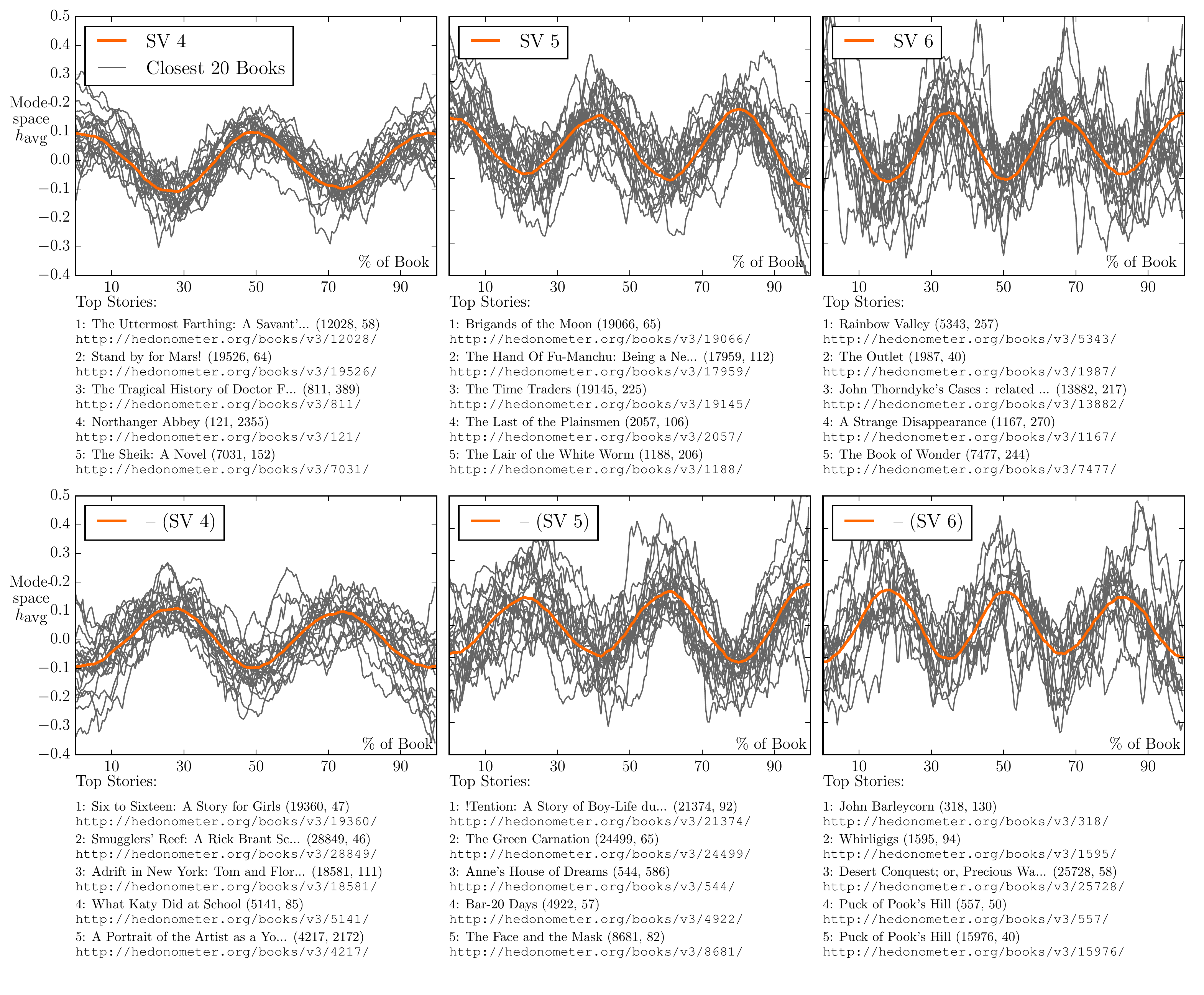}
  \caption[]{
    SVD modes 4--6 (and their negation) with closest stories.
    Again, to show the emotional arcs on the same scale as the modes, we show the modes directly from the rows of $V^T$ and weight the emotional arcs by the inverse of their coefficient in $W$ for the particular mode.
    Shown in parenthesis for each story is the Project Gutenberg ID and the number of downloads from the Project Gutenberg website, respectively.
    Links below each story point to an interactive visualization on \href{http://hedonometer.org}{http://hedonometer.org} which enables detailed exploration of the emotional arc for the story.
  }
  \label{fig:SVD-4-6-labelled}
\end{figure*}

First, we consider modes 4--6 and their closest stories in Fig. \ref{fig:SVD-4-6-labelled}.

Next, we provide a full list of the books closest to each mode in the analysis, both sorted by downloads and support from the mode.

\begin{singlespacing}
\begin{scriptsize}
\begin{longtable}{l | l | l | l | c}
\caption{Stories which are are represented by core emotional arc 1, sorted by the variance explained in their emotional arc by core emotional arc 1.}\\
ID & Title & DL's & max($W$[i,:]) & Arc\\
\hline
\endhead
\href{http://www.gutenberg.org/ebooks/1539}{1539} & The Winter's Tale & 73 & 0.5217 & \adjustimage{height=12px,width=45px,valign=m}{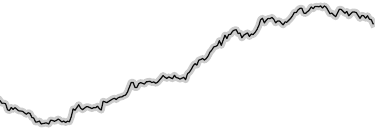} \\
\href{http://www.gutenberg.org/ebooks/33689}{33689} & Oscar Wilde, Art and Morality: A... & 88 & 0.3804 & \adjustimage{height=12px,width=45px,valign=m}{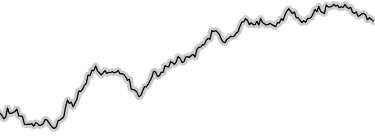} \\
\href{http://www.gutenberg.org/ebooks/35617}{35617} & The Terror: A Mystery & 61 & 0.3734 & \adjustimage{height=12px,width=45px,valign=m}{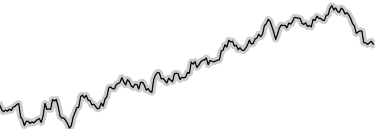} \\
\href{http://www.gutenberg.org/ebooks/7088}{7088} & The Pilgrim's Progress in Words ... & 55 & 0.3522 & \adjustimage{height=12px,width=45px,valign=m}{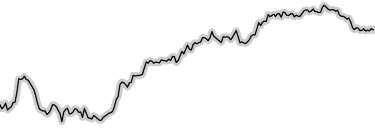} \\
\href{http://www.gutenberg.org/ebooks/26624}{26624} & The Road to Oz & 68 & 0.3412 & \adjustimage{height=12px,width=45px,valign=m}{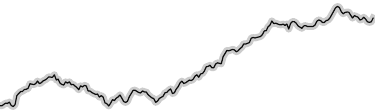} \\
\href{http://www.gutenberg.org/ebooks/21130}{21130} & Book of Wise Sayings: Selected La... & 86 & 0.3361 & \adjustimage{height=12px,width=45px,valign=m}{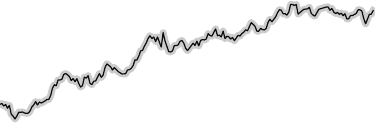} \\
\href{http://www.gutenberg.org/ebooks/2248}{2248} & The Winter's Tale & 137 & 0.3355 & \adjustimage{height=12px,width=45px,valign=m}{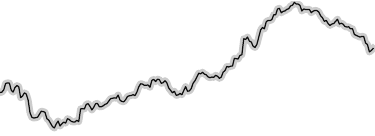} \\
\href{http://www.gutenberg.org/ebooks/485}{485} & The Road to Oz & 178 & 0.3320 & \adjustimage{height=12px,width=45px,valign=m}{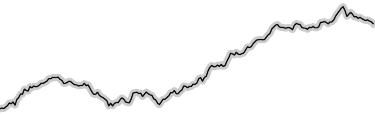} \\
\end{longtable}
\begin{longtable}{l | l | l | l  | c}
\caption{Top 10 stories which are are represented by core emotional arc 1, sorted by downloads.}\\
ID & Title & DL's & max(W[i,:]) & Arc\\
\hline
\endhead
\href{http://www.gutenberg.org/ebooks/844}{844} & The Importance of Being Earnest:... & 9,373 & 0.0963 & \adjustimage{height=12px,width=45px,valign=m}{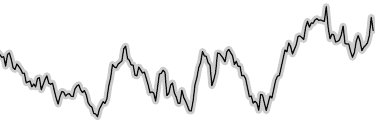} \\
\href{http://www.gutenberg.org/ebooks/236}{236} & The Jungle Book & 3,478 & 0.1728 & \adjustimage{height=12px,width=45px,valign=m}{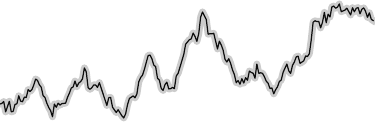} \\
\href{http://www.gutenberg.org/ebooks/215}{215} & The Call of the Wild & 2,439 & 0.2046 & \adjustimage{height=12px,width=45px,valign=m}{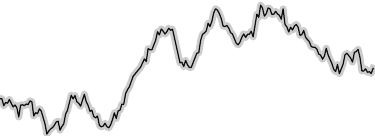} \\
\href{http://www.gutenberg.org/ebooks/779}{779} & The Tragical History of Doctor F... & 2,133 & 0.1133 & \adjustimage{height=12px,width=45px,valign=m}{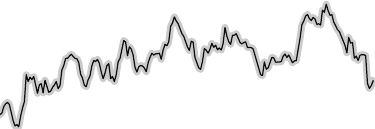} \\
\href{http://www.gutenberg.org/ebooks/35997}{35997} & The Jungle Book & 1,370 & 0.1623 & \adjustimage{height=12px,width=45px,valign=m}{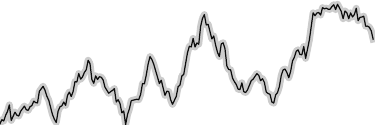} \\
\href{http://www.gutenberg.org/ebooks/885}{885} & An Ideal Husband & 1,303 & 0.2557 & \adjustimage{height=12px,width=45px,valign=m}{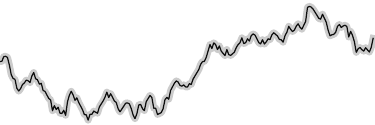} \\
\href{http://www.gutenberg.org/ebooks/78}{78} & Tarzan of the Apes & 1,272 & 0.1341 & \adjustimage{height=12px,width=45px,valign=m}{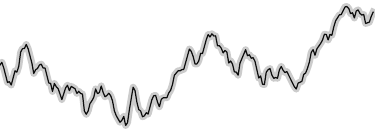} \\
\href{http://www.gutenberg.org/ebooks/500}{500} & The Adventures of Pinocchio & 863 & 0.2108 & \adjustimage{height=12px,width=45px,valign=m}{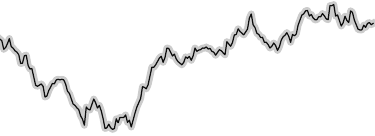} \\
\href{http://www.gutenberg.org/ebooks/269}{269} & Beasts and Super-Beasts & 804 & 0.2271 & \adjustimage{height=12px,width=45px,valign=m}{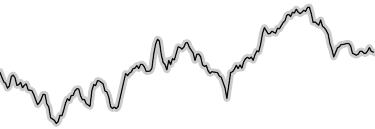} \\
\end{longtable}
\begin{longtable}{l | l | l | l | c}
\caption{Stories which are are represented by core emotional arc 2, sorted by the variance explained in their emotional arc by core emotional arc 2.}\\
ID & Title & DL's & max($W$[i,:]) & Arc\\
\hline
\endhead
\href{http://www.gutenberg.org/ebooks/946}{946} & Lady Susan & 894 & 0.5006 & \adjustimage{height=12px,width=45px,valign=m}{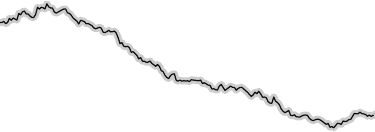} \\
\href{http://www.gutenberg.org/ebooks/17958}{17958} & Warlord of Kor & 70 & 0.4664 & \adjustimage{height=12px,width=45px,valign=m}{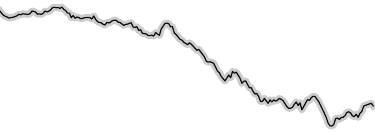} \\
\href{http://www.gutenberg.org/ebooks/17144}{17144} & The House of the Vampire & 188 & 0.4657 & \adjustimage{height=12px,width=45px,valign=m}{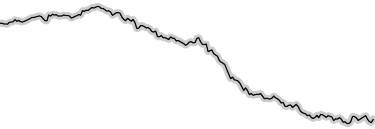} \\
\href{http://www.gutenberg.org/ebooks/93}{93} & Tom Sawyer, Detective & 112 & 0.4566 & \adjustimage{height=12px,width=45px,valign=m}{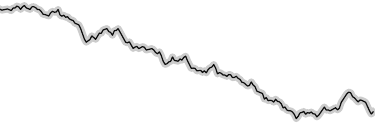} \\
\href{http://www.gutenberg.org/ebooks/159}{159} & The Island of Doctor Moreau & 1,083 & 0.4538 & \adjustimage{height=12px,width=45px,valign=m}{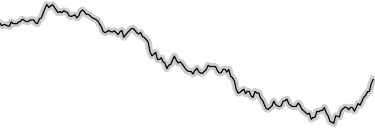} \\
\href{http://www.gutenberg.org/ebooks/32208}{32208} & The Star Lord & 58 & 0.4504 & \adjustimage{height=12px,width=45px,valign=m}{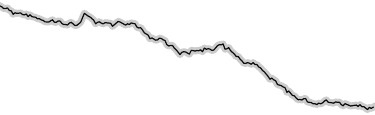} \\
\href{http://www.gutenberg.org/ebooks/790}{790} & Lady Windermere's Fan & 485 & 0.4257 & \adjustimage{height=12px,width=45px,valign=m}{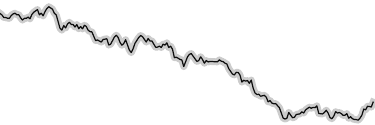} \\
\href{http://www.gutenberg.org/ebooks/1777}{1777} & Romeo and Juliet & 186 & 0.4013 & \adjustimage{height=12px,width=45px,valign=m}{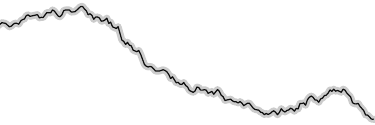} \\
\end{longtable}
\begin{longtable}{l | l | l | l | c}
\caption{Top 10 stories which are are represented by core emotional arc 2, sorted by downloads.}\\
ID & Title & DL's & max($W$[i,:]) & Arc\\
\hline
\endhead
\href{http://www.gutenberg.org/ebooks/174}{174} & The Picture of Dorian Gray & 7,652 & 0.1893 & \adjustimage{height=12px,width=45px,valign=m}{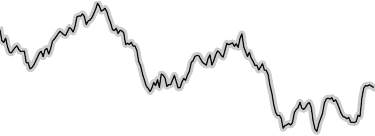} \\
\href{http://www.gutenberg.org/ebooks/16328}{16328} & Beowulf
: An Anglo-Saxon Epic Poem & 5,359 & 0.1584 & \adjustimage{height=12px,width=45px,valign=m}{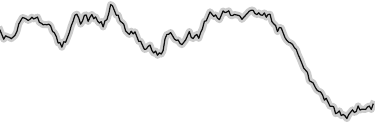} \\
\href{http://www.gutenberg.org/ebooks/42}{42} & The Strange Case of Dr. Jekyll a... & 4,908 & 0.2379 & \adjustimage{height=12px,width=45px,valign=m}{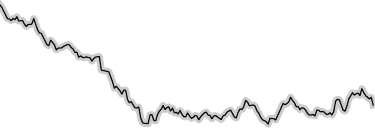} \\
\href{http://www.gutenberg.org/ebooks/526}{526} & Heart of Darkness & 4,362 & 0.2597 & \adjustimage{height=12px,width=45px,valign=m}{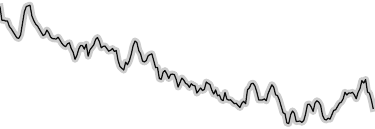} \\
\href{http://www.gutenberg.org/ebooks/35}{35} & The Time Machine & 3,732 & 0.3199 & \adjustimage{height=12px,width=45px,valign=m}{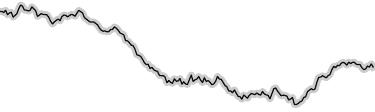} \\
\href{http://www.gutenberg.org/ebooks/3825}{3825} & Pygmalion & 3,580 & 0.2446 & \adjustimage{height=12px,width=45px,valign=m}{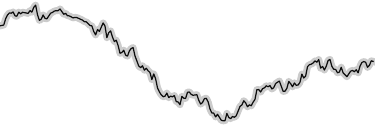} \\
\href{http://www.gutenberg.org/ebooks/2852}{2852} & The Hound of the Baskervilles & 3,358 & 0.1951 & \adjustimage{height=12px,width=45px,valign=m}{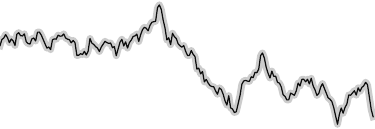} \\
\href{http://www.gutenberg.org/ebooks/219}{219} & Heart of Darkness & 3,243 & 0.2184 & \adjustimage{height=12px,width=45px,valign=m}{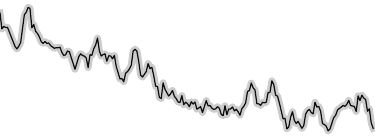} \\
\href{http://www.gutenberg.org/ebooks/28520}{28520} & Forbidden Fruit: Luscious and exc... & 2,716 & 0.2028 & \adjustimage{height=12px,width=45px,valign=m}{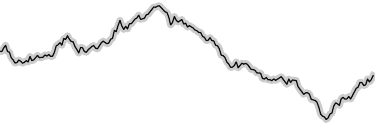} \\
\end{longtable}
\begin{longtable}{l | l | l | l | c}
\caption{Stories which are are represented by core emotional arc 3, sorted by the variance explained in their emotional arc by core emotional arc 3.}\\
ID & Title & DL's & max($W$[i,:]) & Arc\\
\hline
\endhead
\href{http://www.gutenberg.org/ebooks/419}{419} & The Magic of Oz & 186 & 0.3150 & \adjustimage{height=12px,width=45px,valign=m}{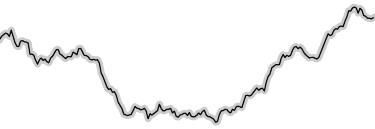} \\
\href{http://www.gutenberg.org/ebooks/10736}{10736} & Children of the Frost & 82 & 0.3083 & \adjustimage{height=12px,width=45px,valign=m}{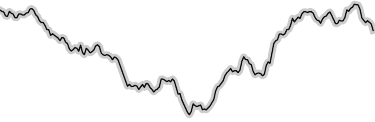} \\
\href{http://www.gutenberg.org/ebooks/1094}{1094} & Tamburlaine the Great — Part 1 & 474 & 0.3081 & \adjustimage{height=12px,width=45px,valign=m}{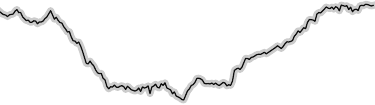} \\
\href{http://www.gutenberg.org/ebooks/520}{520} & The Life and Adventures of Santa... & 76 & 0.3041 & \adjustimage{height=12px,width=45px,valign=m}{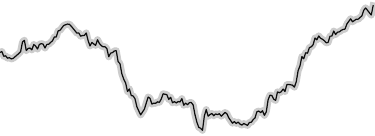} \\
\href{http://www.gutenberg.org/ebooks/2911}{2911} & Justice & 50 & 0.2989 & \adjustimage{height=12px,width=45px,valign=m}{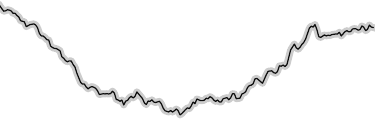} \\
\href{http://www.gutenberg.org/ebooks/18768}{18768} & The Sky Is Falling & 113 & 0.2978 & \adjustimage{height=12px,width=45px,valign=m}{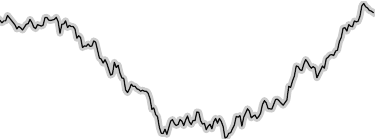} \\
\href{http://www.gutenberg.org/ebooks/29774}{29774} & A Yankee Flier Over Berlin & 42 & 0.2864 & \adjustimage{height=12px,width=45px,valign=m}{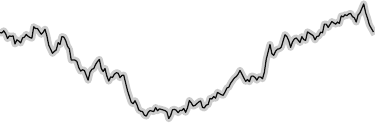} \\
\href{http://www.gutenberg.org/ebooks/19726}{19726} & The Door Through Space & 201 & 0.2826 & \adjustimage{height=12px,width=45px,valign=m}{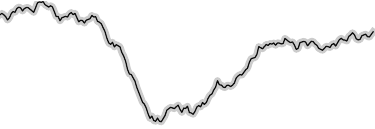} \\
\end{longtable}
\begin{longtable}{l | l | l | l  | c}
\caption{Top 10 stories which are are represented by core emotional arc 3, sorted by downloads.}\\
ID & Title & DL's & max(W[i,:]) & Arc\\
\hline
\endhead
\href{http://www.gutenberg.org/ebooks/2814}{2814} & Dubliners & 4,742 & 0.2583 & \adjustimage{height=12px,width=45px,valign=m}{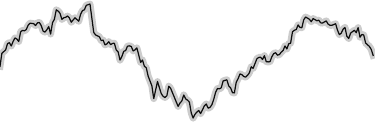} \\
\href{http://www.gutenberg.org/ebooks/55}{55} & The Wonderful Wizard of Oz & 3,035 & 0.1770 & \adjustimage{height=12px,width=45px,valign=m}{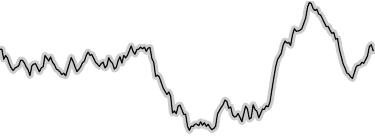} \\
\href{http://www.gutenberg.org/ebooks/12}{12} & Through the Looking-Glass & 2,892 & 0.0742 & \adjustimage{height=12px,width=45px,valign=m}{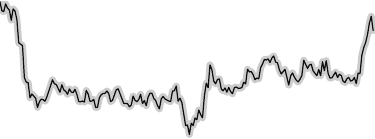} \\
\href{http://www.gutenberg.org/ebooks/834}{834} & The Memoirs of Sherlock Holmes & 2,164 & 0.1476 & \adjustimage{height=12px,width=45px,valign=m}{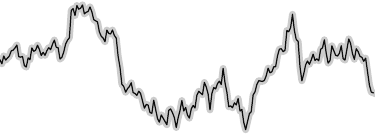} \\
\href{http://www.gutenberg.org/ebooks/1837}{1837} & The Prince and the Pauper & 1,389 & 0.1604 & \adjustimage{height=12px,width=45px,valign=m}{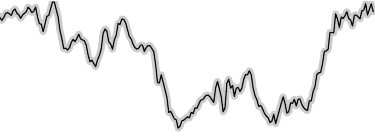} \\
\href{http://www.gutenberg.org/ebooks/447}{447} & Maggie: A Girl of the Streets & 1,295 & 0.1616 & \adjustimage{height=12px,width=45px,valign=m}{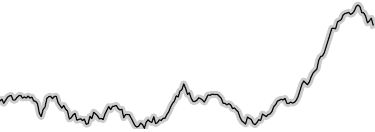} \\
\href{http://www.gutenberg.org/ebooks/113}{113} & The Secret Garden & 1,153 & 0.1416 & \adjustimage{height=12px,width=45px,valign=m}{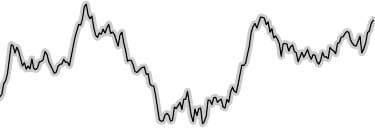} \\
\href{http://www.gutenberg.org/ebooks/421}{421} & Kidnapped & 1,132 & 0.1274 & \adjustimage{height=12px,width=45px,valign=m}{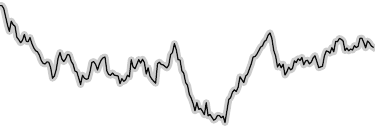} \\
\href{http://www.gutenberg.org/ebooks/5230}{5230} & The Invisible Man: A Grotesque R... & 1,011 & 0.1250 & \adjustimage{height=12px,width=45px,valign=m}{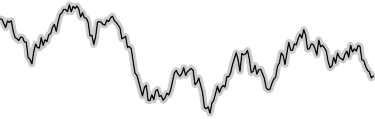} \\
\end{longtable}
\begin{longtable}{l | l | l | l | c}
\caption{Stories which are are represented by core emotional arc 4, sorted by the variance explained in their emotional arc by core emotional arc 4.}\\
ID & Title & DL's & max($W$[i,:]) & Arc\\
\hline
\endhead
\href{http://www.gutenberg.org/ebooks/34215}{34215} & Shadowings & 63 & 0.4141 & \adjustimage{height=12px,width=45px,valign=m}{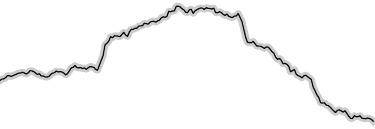} \\
\href{http://www.gutenberg.org/ebooks/12384}{12384} & Battle-Pieces and Aspects of the... & 194 & 0.3546 & \adjustimage{height=12px,width=45px,valign=m}{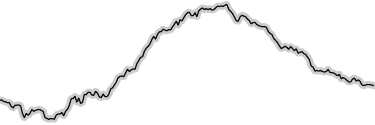} \\
\href{http://www.gutenberg.org/ebooks/36281}{36281} & The Slayer of Souls & 63 & 0.3283 & \adjustimage{height=12px,width=45px,valign=m}{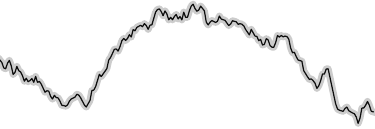} \\
\href{http://www.gutenberg.org/ebooks/17412}{17412} & The Bobbsey Twins
: Or, Merry Day... & 69 & 0.3259 & \adjustimage{height=12px,width=45px,valign=m}{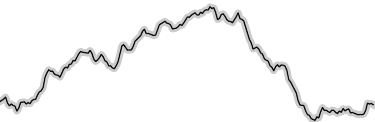} \\
\href{http://www.gutenberg.org/ebooks/2727}{2727} & Allan's Wife & 128 & 0.3124 & \adjustimage{height=12px,width=45px,valign=m}{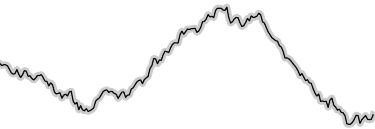} \\
\href{http://www.gutenberg.org/ebooks/27761}{27761} & Hamlet, Prince of Denmark & 301 & 0.3057 & \adjustimage{height=12px,width=45px,valign=m}{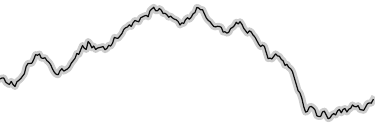} \\
\href{http://www.gutenberg.org/ebooks/363}{363} & The Oakdale Affair & 91 & 0.2968 & \adjustimage{height=12px,width=45px,valign=m}{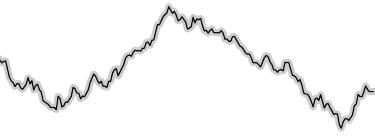} \\
\href{http://www.gutenberg.org/ebooks/17854}{17854} & The Sport of the Gods & 61 & 0.2855 & \adjustimage{height=12px,width=45px,valign=m}{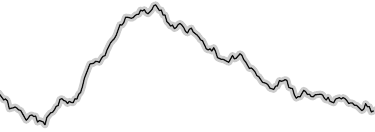} \\
\end{longtable}
\begin{longtable}{l | l | l | l | c}
\caption{Top 10 stories which are are represented by core emotional arc 4, sorted by downloads.}\\
ID & Title & DL's & max($W$[i,:]) & Arc\\
\hline
\endhead
\href{http://www.gutenberg.org/ebooks/46}{46} & A Christmas Carol in Prose; Bein... & 4,602 & 0.1917 & \adjustimage{height=12px,width=45px,valign=m}{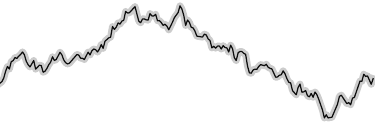} \\
\href{http://www.gutenberg.org/ebooks/20}{20} & Paradise Lost & 2,522 & 0.1475 & \adjustimage{height=12px,width=45px,valign=m}{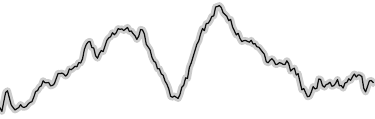} \\
\href{http://www.gutenberg.org/ebooks/62}{62} & A Princess of Mars & 2,515 & 0.1032 & \adjustimage{height=12px,width=45px,valign=m}{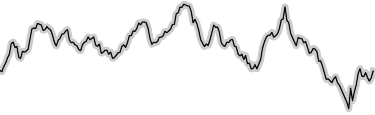} \\
\href{http://www.gutenberg.org/ebooks/308}{308} & Three Men in a Boat & 2,059 & 0.0815 & \adjustimage{height=12px,width=45px,valign=m}{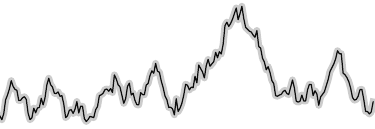} \\
\href{http://www.gutenberg.org/ebooks/5131}{5131} & Childe Harold's Pilgrimage & 1,481 & 0.1987 & \adjustimage{height=12px,width=45px,valign=m}{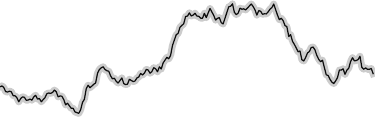} \\
\href{http://www.gutenberg.org/ebooks/972}{972} & The Devil's Dictionary & 1,257 & 0.0936 & \adjustimage{height=12px,width=45px,valign=m}{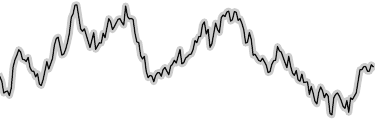} \\
\href{http://www.gutenberg.org/ebooks/3289}{3289} & The Valley of Fear & 1,228 & 0.1845 & \adjustimage{height=12px,width=45px,valign=m}{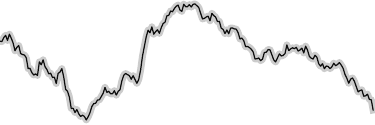} \\
\href{http://www.gutenberg.org/ebooks/131}{131} & The Pilgrim's Progress from this... & 1,126 & 0.1888 & \adjustimage{height=12px,width=45px,valign=m}{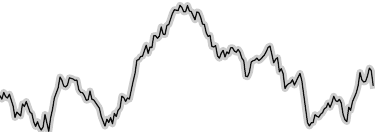} \\
\href{http://www.gutenberg.org/ebooks/28522}{28522} & Laura Middleton; Her Brother and... & 1,097 & 0.1851 & \adjustimage{height=12px,width=45px,valign=m}{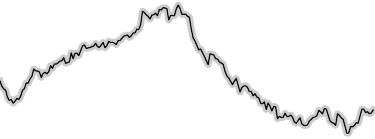} \\
\end{longtable}
\begin{longtable}{l | l | l | l | c}
\caption{Stories which are are represented by core emotional arc 5, sorted by the variance explained in their emotional arc by core emotional arc 5.}\\
ID & Title & DL's & max($W$[i,:]) & Arc\\
\hline
\endhead
\href{http://www.gutenberg.org/ebooks/17763}{17763} & The Mystery of the Hasty Arrow & 93 & 0.2301 & \adjustimage{height=12px,width=45px,valign=m}{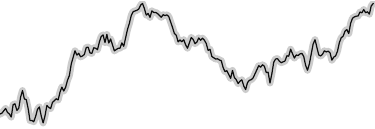} \\
\href{http://www.gutenberg.org/ebooks/5317}{5317} & Through the Magic Door & 81 & 0.2299 & \adjustimage{height=12px,width=45px,valign=m}{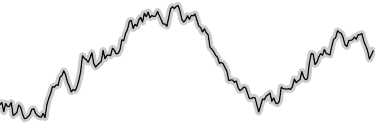} \\
\href{http://www.gutenberg.org/ebooks/13944}{13944} & After London; Or, Wild England & 146 & 0.2259 & \adjustimage{height=12px,width=45px,valign=m}{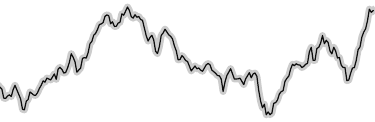} \\
\href{http://www.gutenberg.org/ebooks/12590}{12590} & The Shadow of the Rope & 75 & 0.2202 & \adjustimage{height=12px,width=45px,valign=m}{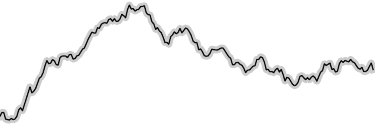} \\
\href{http://www.gutenberg.org/ebooks/35247}{35247} & That Affair at Elizabeth & 62 & 0.2179 & \adjustimage{height=12px,width=45px,valign=m}{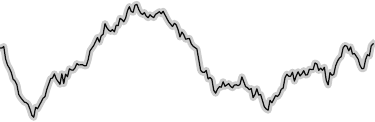} \\
\href{http://www.gutenberg.org/ebooks/39827}{39827} & The Conduct of Life & 176 & 0.2041 & \adjustimage{height=12px,width=45px,valign=m}{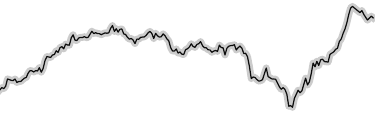} \\
\href{http://www.gutenberg.org/ebooks/295}{295} & The Early Short Fiction of Edith... & 49 & 0.2012 & \adjustimage{height=12px,width=45px,valign=m}{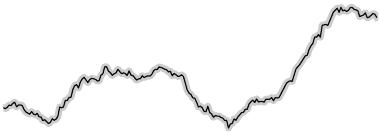} \\
\href{http://www.gutenberg.org/ebooks/10459}{10459} & The Celtic Twilight & 146 & 0.1974 & \adjustimage{height=12px,width=45px,valign=m}{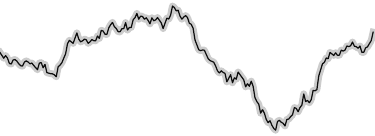} \\
\end{longtable}
\begin{longtable}{l | l | l | l  | c}
\caption{Top 10 stories which are are represented by core emotional arc 5, sorted by downloads.}\\
ID & Title & DL's & max(W[i,:]) & Arc\\
\hline
\endhead
\href{http://www.gutenberg.org/ebooks/120}{120} & Treasure Island & 4,402 & 0.1620 & \adjustimage{height=12px,width=45px,valign=m}{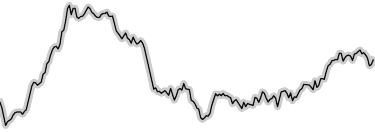} \\
\href{http://www.gutenberg.org/ebooks/696}{696} & The Castle of Otranto & 1,663 & 0.1465 & \adjustimage{height=12px,width=45px,valign=m}{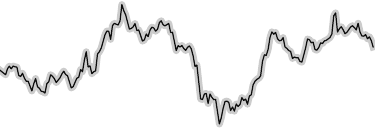} \\
\href{http://www.gutenberg.org/ebooks/2166}{2166} & King Solomon's Mines & 788 & 0.1652 & \adjustimage{height=12px,width=45px,valign=m}{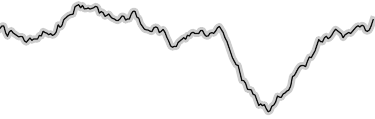} \\
\href{http://www.gutenberg.org/ebooks/1280}{1280} & Spoon River Anthology & 671 & 0.1297 & \adjustimage{height=12px,width=45px,valign=m}{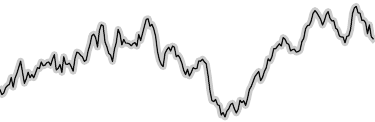} \\
\href{http://www.gutenberg.org/ebooks/1212}{1212} & Love and Freindship [sic] & 611 & 0.1657 & \adjustimage{height=12px,width=45px,valign=m}{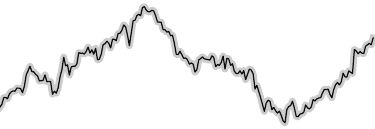} \\
\href{http://www.gutenberg.org/ebooks/708}{708} & The Princess and the Goblin & 579 & 0.0996 & \adjustimage{height=12px,width=45px,valign=m}{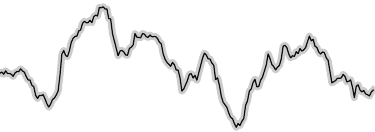} \\
\href{http://www.gutenberg.org/ebooks/1164}{1164} & The Iron Heel & 506 & 0.1395 & \adjustimage{height=12px,width=45px,valign=m}{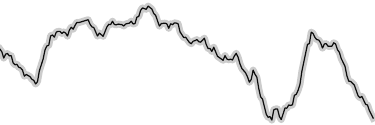} \\
\href{http://www.gutenberg.org/ebooks/2243}{2243} & The Merchant of Venice & 506 & 0.0787 & \adjustimage{height=12px,width=45px,valign=m}{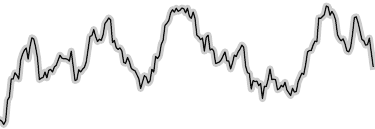} \\
\href{http://www.gutenberg.org/ebooks/394}{394} & Cranford & 285 & 0.1029 & \adjustimage{height=12px,width=45px,valign=m}{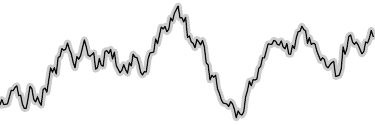} \\
\end{longtable}
\begin{longtable}{l | l | l | l | c}
\caption{Stories which are are represented by core emotional arc 6, sorted by the variance explained in their emotional arc by core emotional arc 6.}\\
ID & Title & DL's & max($W$[i,:]) & Arc\\
\hline
\endhead
\href{http://www.gutenberg.org/ebooks/18172}{18172} & This World Is Taboo & 64 & 0.3281 & \adjustimage{height=12px,width=45px,valign=m}{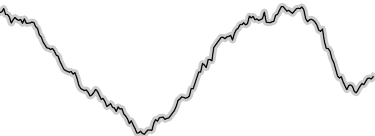} \\
\href{http://www.gutenberg.org/ebooks/339}{339} & Old Indian Days & 139 & 0.3148 & \adjustimage{height=12px,width=45px,valign=m}{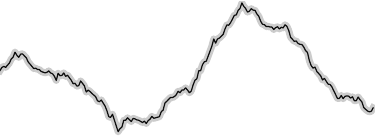} \\
\href{http://www.gutenberg.org/ebooks/10377}{10377} & The Evil Guest & 93 & 0.2868 & \adjustimage{height=12px,width=45px,valign=m}{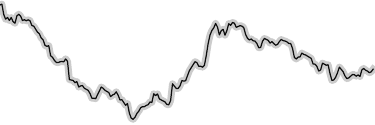} \\
\href{http://www.gutenberg.org/ebooks/29448}{29448} & Pariah Planet & 96 & 0.2864 & \adjustimage{height=12px,width=45px,valign=m}{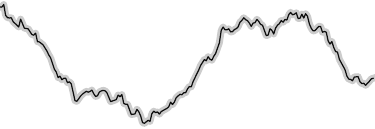} \\
\href{http://www.gutenberg.org/ebooks/289}{289} & The Wind in the Willows & 1,475 & 0.2750 & \adjustimage{height=12px,width=45px,valign=m}{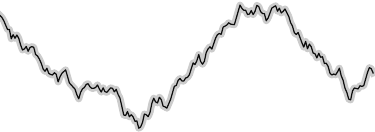} \\
\href{http://www.gutenberg.org/ebooks/20727}{20727} & The Cosmic Computer & 221 & 0.2716 & \adjustimage{height=12px,width=45px,valign=m}{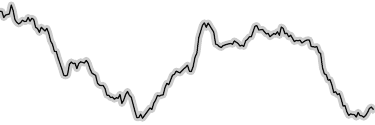} \\
\href{http://www.gutenberg.org/ebooks/27805}{27805} & The Wind in the Willows & 543 & 0.2702 & \adjustimage{height=12px,width=45px,valign=m}{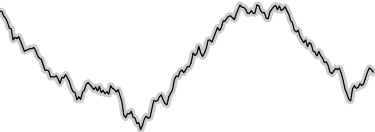} \\
\href{http://www.gutenberg.org/ebooks/27991}{27991} & The Blue Bird for Children: The W... & 111 & 0.2677 & \adjustimage{height=12px,width=45px,valign=m}{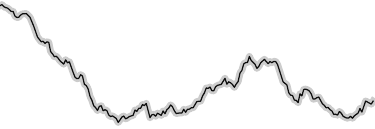} \\
\end{longtable}
\begin{longtable}{l | l | l | l | c}
\caption{Top 10 stories which are are represented by core emotional arc 6, sorted by downloads.}\\
ID & Title & DL's & max($W$[i,:]) & Arc\\
\hline
\endhead
\href{http://www.gutenberg.org/ebooks/84}{84} & Frankenstein; Or, The Modern Pro... & 11,699 & 0.1602 & \adjustimage{height=12px,width=45px,valign=m}{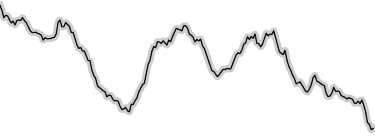} \\
\href{http://www.gutenberg.org/ebooks/244}{244} & A Study in Scarlet & 4,535 & 0.1955 & \adjustimage{height=12px,width=45px,valign=m}{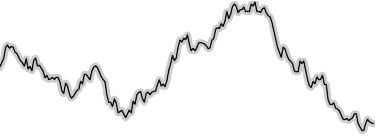} \\
\href{http://www.gutenberg.org/ebooks/863}{863} & The Mysterious Affair at Styles & 3,112 & 0.1937 & \adjustimage{height=12px,width=45px,valign=m}{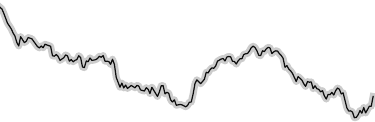} \\
\href{http://www.gutenberg.org/ebooks/36}{36} & The War of the Worlds & 2,496 & 0.1060 & \adjustimage{height=12px,width=45px,valign=m}{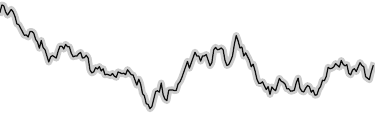} \\
\href{http://www.gutenberg.org/ebooks/2097}{2097} & The Sign of the Four & 2,283 & 0.1199 & \adjustimage{height=12px,width=45px,valign=m}{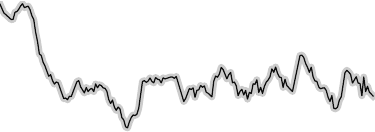} \\
\href{http://www.gutenberg.org/ebooks/209}{209} & The Turn of the Screw & 2,175 & 0.2251 & \adjustimage{height=12px,width=45px,valign=m}{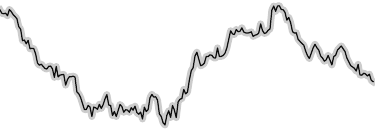} \\
\href{http://www.gutenberg.org/ebooks/289}{289} & The Wind in the Willows & 1,475 & 0.2750 & \adjustimage{height=12px,width=45px,valign=m}{all-timeseries/289.pdf} \\
\href{http://www.gutenberg.org/ebooks/10007}{10007} & Carmilla & 1,416 & 0.1333 & \adjustimage{height=12px,width=45px,valign=m}{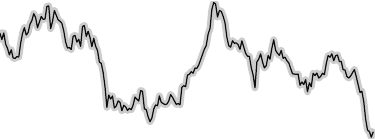} \\
\href{http://www.gutenberg.org/ebooks/73}{73} & The Red Badge of Courage: An Epi... & 1,163 & 0.1494 & \adjustimage{height=12px,width=45px,valign=m}{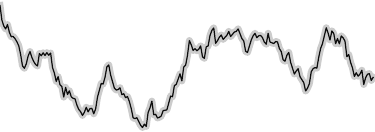} \\
\end{longtable}
\begin{longtable}{l | l | l | l | c}
\caption{Stories which are are represented by core emotional arc 7, sorted by the variance explained in their emotional arc by core emotional arc 7.}\\
ID & Title & DL's & max($W$[i,:]) & Arc\\
\hline
\endhead
\href{http://www.gutenberg.org/ebooks/12028}{12028} & The Uttermost Farthing: A Savant'... & 58 & 0.2612 & \adjustimage{height=12px,width=45px,valign=m}{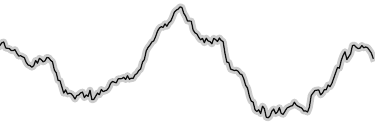} \\
\href{http://www.gutenberg.org/ebooks/19526}{19526} & Stand by for Mars! & 64 & 0.2425 & \adjustimage{height=12px,width=45px,valign=m}{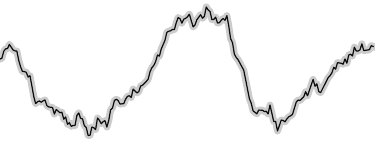} \\
\href{http://www.gutenberg.org/ebooks/811}{811} & The Tragical History of Doctor F... & 389 & 0.2245 & \adjustimage{height=12px,width=45px,valign=m}{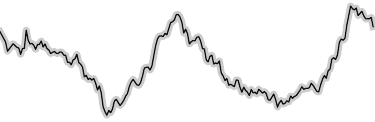} \\
\href{http://www.gutenberg.org/ebooks/121}{121} & Northanger Abbey & 2,355 & 0.2196 & \adjustimage{height=12px,width=45px,valign=m}{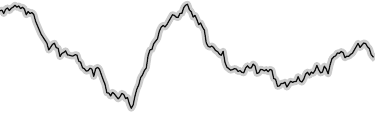} \\
\href{http://www.gutenberg.org/ebooks/7031}{7031} & The Sheik: A Novel & 152 & 0.2071 & \adjustimage{height=12px,width=45px,valign=m}{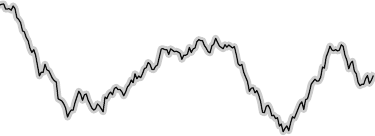} \\
\href{http://www.gutenberg.org/ebooks/24035}{24035} & The Pirates of Ersatz & 113 & 0.2017 & \adjustimage{height=12px,width=45px,valign=m}{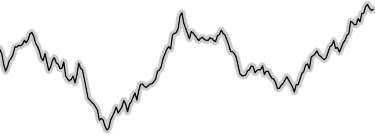} \\
\href{http://www.gutenberg.org/ebooks/2429}{2429} & Lost Face & 190 & 0.1975 & \adjustimage{height=12px,width=45px,valign=m}{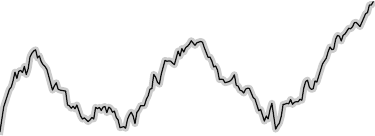} \\
\href{http://www.gutenberg.org/ebooks/74}{74} & The Adventures of Tom Sawyer & 9,454 & 0.1906 & \adjustimage{height=12px,width=45px,valign=m}{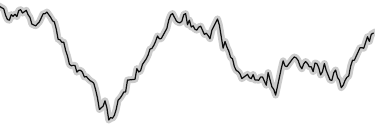} \\
\end{longtable}
\begin{longtable}{l | l | l | l  | c}
\caption{Top 10 stories which are are represented by core emotional arc 7, sorted by downloads.}\\
ID & Title & DL's & max(W[i,:]) & Arc\\
\hline
\endhead
\href{http://www.gutenberg.org/ebooks/74}{74} & The Adventures of Tom Sawyer & 9,454 & 0.1906 & \adjustimage{height=12px,width=45px,valign=m}{all-timeseries/74.pdf} \\
\href{http://www.gutenberg.org/ebooks/16}{16} & Peter Pan & 5,789 & 0.1197 & \adjustimage{height=12px,width=45px,valign=m}{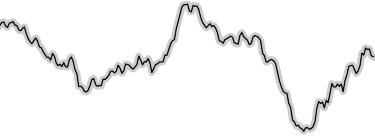} \\
\href{http://www.gutenberg.org/ebooks/4517}{4517} & Ethan Frome & 2,895 & 0.1220 & \adjustimage{height=12px,width=45px,valign=m}{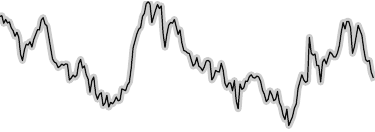} \\
\href{http://www.gutenberg.org/ebooks/121}{121} & Northanger Abbey & 2,355 & 0.2196 & \adjustimage{height=12px,width=45px,valign=m}{all-timeseries/121.pdf} \\
\href{http://www.gutenberg.org/ebooks/8492}{8492} & The King in Yellow & 1,504 & 0.1354 & \adjustimage{height=12px,width=45px,valign=m}{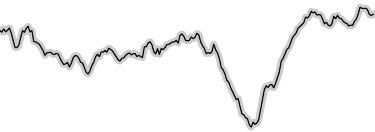} \\
\href{http://www.gutenberg.org/ebooks/910}{910} & White Fang & 1,178 & 0.1223 & \adjustimage{height=12px,width=45px,valign=m}{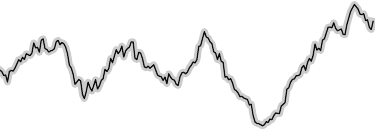} \\
\href{http://www.gutenberg.org/ebooks/242}{242} & My Antonia & 847 & 0.1024 & \adjustimage{height=12px,width=45px,valign=m}{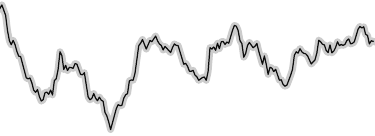} \\
\href{http://www.gutenberg.org/ebooks/146}{146} & A Little Princess
: Being the who... & 825 & 0.1189 & \adjustimage{height=12px,width=45px,valign=m}{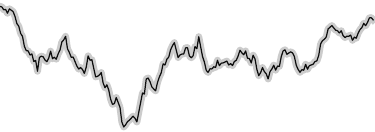} \\
\href{http://www.gutenberg.org/ebooks/47}{47} & Anne of Avonlea & 803 & 0.1000 & \adjustimage{height=12px,width=45px,valign=m}{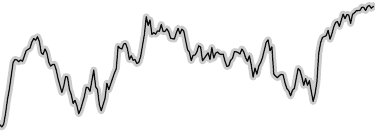} \\
\end{longtable}
\begin{longtable}{l | l | l | l | c}
\caption{Stories which are are represented by core emotional arc 8, sorted by the variance explained in their emotional arc by core emotional arc 8.}\\
ID & Title & DL's & max($W$[i,:]) & Arc\\
\hline
\endhead
\href{http://www.gutenberg.org/ebooks/19360}{19360} & Six to Sixteen: A Story for Girls & 47 & 0.2415 & \adjustimage{height=12px,width=45px,valign=m}{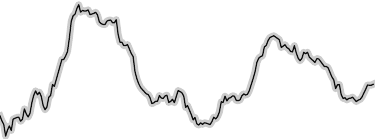} \\
\href{http://www.gutenberg.org/ebooks/28849}{28849} & Smugglers' Reef: A Rick Brant Sc... & 46 & 0.2320 & \adjustimage{height=12px,width=45px,valign=m}{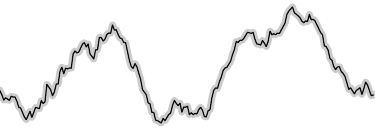} \\
\href{http://www.gutenberg.org/ebooks/18581}{18581} & Adrift in New York: Tom and Flor... & 111 & 0.2182 & \adjustimage{height=12px,width=45px,valign=m}{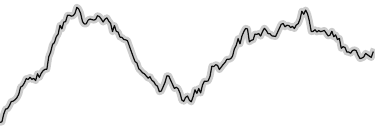} \\
\href{http://www.gutenberg.org/ebooks/5141}{5141} & What Katy Did at School & 85 & 0.2065 & \adjustimage{height=12px,width=45px,valign=m}{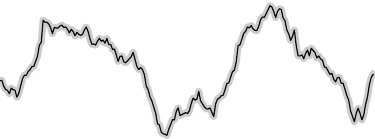} \\
\href{http://www.gutenberg.org/ebooks/4217}{4217} & A Portrait of the Artist as a Yo... & 2,172 & 0.2039 & \adjustimage{height=12px,width=45px,valign=m}{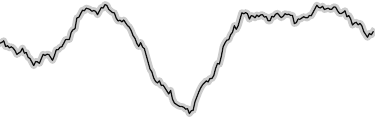} \\
\href{http://www.gutenberg.org/ebooks/31308}{31308} & Orientations & 76 & 0.1924 & \adjustimage{height=12px,width=45px,valign=m}{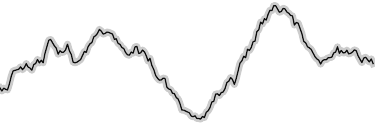} \\
\href{http://www.gutenberg.org/ebooks/35117}{35117} & Lord Tony's Wife: An Adventure o... & 66 & 0.1748 & \adjustimage{height=12px,width=45px,valign=m}{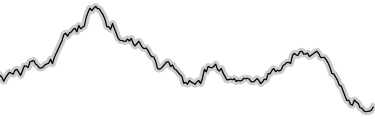} \\
\href{http://www.gutenberg.org/ebooks/4715}{4715} & An African Millionaire: Episodes... & 163 & 0.1714 & \adjustimage{height=12px,width=45px,valign=m}{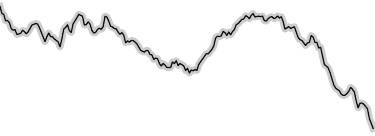} \\
\end{longtable}
\begin{longtable}{l | l | l | l | c}
\caption{Top 10 stories which are are represented by core emotional arc 8, sorted by downloads.}\\
ID & Title & DL's & max($W$[i,:]) & Arc\\
\hline
\endhead
\href{http://www.gutenberg.org/ebooks/4217}{4217} & A Portrait of the Artist as a Yo... & 2,172 & 0.2039 & \adjustimage{height=12px,width=45px,valign=m}{all-timeseries/4217.pdf} \\
\href{http://www.gutenberg.org/ebooks/204}{204} & The Innocence of Father Brown & 800 & 0.1357 & \adjustimage{height=12px,width=45px,valign=m}{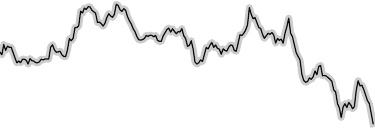} \\
\href{http://www.gutenberg.org/ebooks/26}{26} & Paradise Lost & 730 & 0.1565 & \adjustimage{height=12px,width=45px,valign=m}{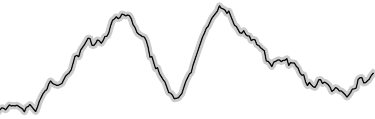} \\
\href{http://www.gutenberg.org/ebooks/64}{64} & The Gods of Mars & 628 & 0.1481 & \adjustimage{height=12px,width=45px,valign=m}{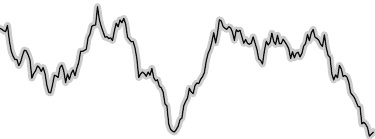} \\
\href{http://www.gutenberg.org/ebooks/5670}{5670} & Jacob's Room & 403 & 0.1013 & \adjustimage{height=12px,width=45px,valign=m}{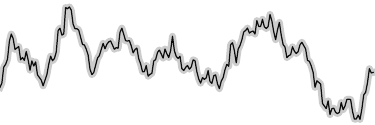} \\
\href{http://www.gutenberg.org/ebooks/4352}{4352} & Laughter: An Essay on the Meanin... & 365 & 0.1251 & \adjustimage{height=12px,width=45px,valign=m}{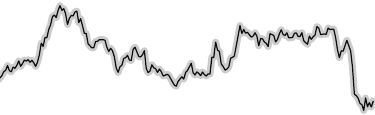} \\
\href{http://www.gutenberg.org/ebooks/225}{225} & At the Back of the North Wind & 288 & 0.0889 & \adjustimage{height=12px,width=45px,valign=m}{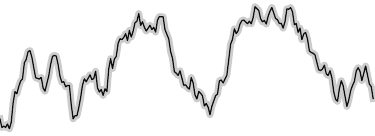} \\
\href{http://www.gutenberg.org/ebooks/517}{517} & The Emerald City of Oz & 266 & 0.1097 & \adjustimage{height=12px,width=45px,valign=m}{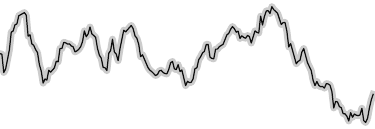} \\
\href{http://www.gutenberg.org/ebooks/2688}{2688} & The Clue of the Twisted Candle & 237 & 0.1402 & \adjustimage{height=12px,width=45px,valign=m}{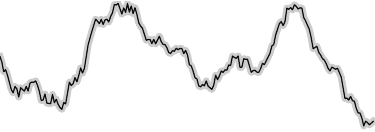} \\
\end{longtable}
\begin{longtable}{l | l | l | l | c}
\caption{Stories which are are represented by core emotional arc 9, sorted by the variance explained in their emotional arc by core emotional arc 9.}\\
ID & Title & DL's & max($W$[i,:]) & Arc\\
\hline
\endhead
\href{http://www.gutenberg.org/ebooks/19066}{19066} & Brigands of the Moon & 65 & 0.2113 & \adjustimage{height=12px,width=45px,valign=m}{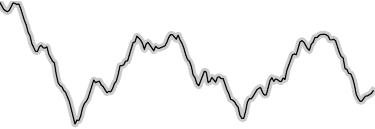} \\
\href{http://www.gutenberg.org/ebooks/17959}{17959} & The Hand Of Fu-Manchu: Being a Ne... & 112 & 0.1861 & \adjustimage{height=12px,width=45px,valign=m}{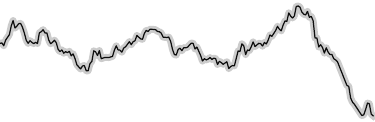} \\
\href{http://www.gutenberg.org/ebooks/19145}{19145} & The Time Traders & 225 & 0.1801 & \adjustimage{height=12px,width=45px,valign=m}{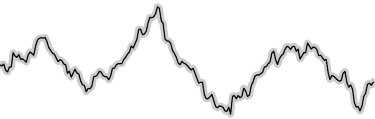} \\
\href{http://www.gutenberg.org/ebooks/2057}{2057} & The Last of the Plainsmen & 106 & 0.1681 & \adjustimage{height=12px,width=45px,valign=m}{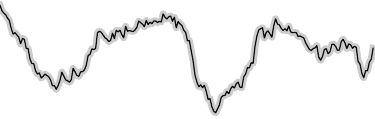} \\
\href{http://www.gutenberg.org/ebooks/1188}{1188} & The Lair of the White Worm & 206 & 0.1632 & \adjustimage{height=12px,width=45px,valign=m}{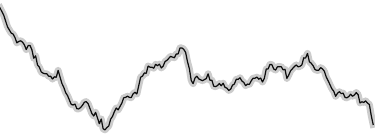} \\
\href{http://www.gutenberg.org/ebooks/546}{546} & Under the Andes & 87 & 0.1544 & \adjustimage{height=12px,width=45px,valign=m}{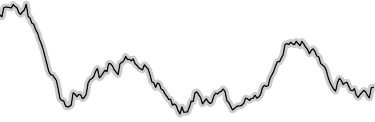} \\
\href{http://www.gutenberg.org/ebooks/18668}{18668} & In Search of the Unknown & 159 & 0.1489 & \adjustimage{height=12px,width=45px,valign=m}{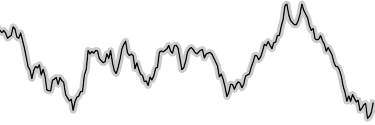} \\
\href{http://www.gutenberg.org/ebooks/6955}{6955} & The Prince and Betty & 85 & 0.1488 & \adjustimage{height=12px,width=45px,valign=m}{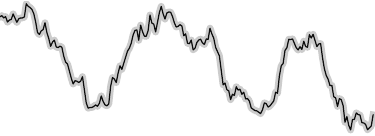} \\
\end{longtable}
\begin{longtable}{l | l | l | l  | c}
\caption{Top 10 stories which are are represented by core emotional arc 9, sorted by downloads.}\\
ID & Title & DL's & max(W[i,:]) & Arc\\
\hline
\endhead
\href{http://www.gutenberg.org/ebooks/11}{11} & Alice's Adventures in Wonderland & 17,366 & 0.0988 & \adjustimage{height=12px,width=45px,valign=m}{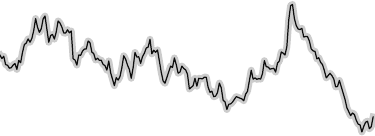} \\
\href{http://www.gutenberg.org/ebooks/33}{33} & The Scarlet Letter & 3,045 & 0.1195 & \adjustimage{height=12px,width=45px,valign=m}{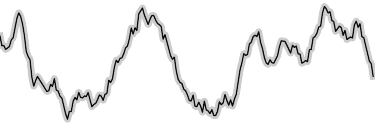} \\
\href{http://www.gutenberg.org/ebooks/28885}{28885} & Alice's Adventures in Wonderland... & 1,051 & 0.0862 & \adjustimage{height=12px,width=45px,valign=m}{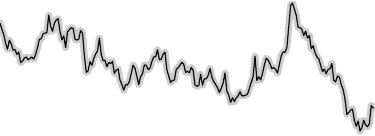} \\
\href{http://www.gutenberg.org/ebooks/271}{271} & Black Beauty & 780 & 0.0870 & \adjustimage{height=12px,width=45px,valign=m}{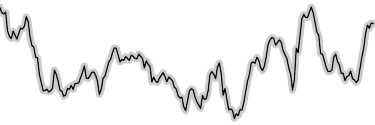} \\
\href{http://www.gutenberg.org/ebooks/2775}{2775} & The Good Soldier & 426 & 0.1144 & \adjustimage{height=12px,width=45px,valign=m}{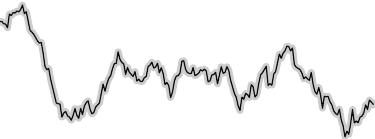} \\
\href{http://www.gutenberg.org/ebooks/1872}{1872} & The Red House Mystery & 422 & 0.1421 & \adjustimage{height=12px,width=45px,valign=m}{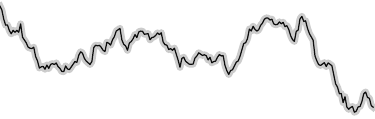} \\
\href{http://www.gutenberg.org/ebooks/25344}{25344} & The Scarlet Letter & 386 & 0.1266 & \adjustimage{height=12px,width=45px,valign=m}{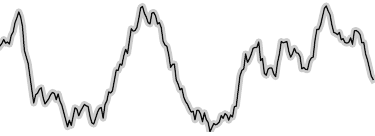} \\
\href{http://www.gutenberg.org/ebooks/1051}{1051} & Sartor Resartus: The Life and Op... & 347 & 0.1074 & \adjustimage{height=12px,width=45px,valign=m}{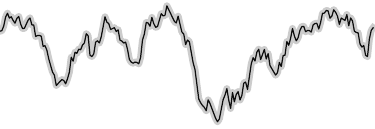} \\
\href{http://www.gutenberg.org/ebooks/3328}{3328} & Man and Superman: A Comedy and a... & 312 & 0.1181 & \adjustimage{height=12px,width=45px,valign=m}{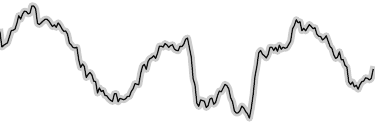} \\
\end{longtable}
\begin{longtable}{l | l | l | l | c}
\caption{Stories which are are represented by core emotional arc 10, sorted by the variance explained in their emotional arc by core emotional arc 10.}\\
ID & Title & DL's & max($W$[i,:]) & Arc\\
\hline
\endhead
\href{http://www.gutenberg.org/ebooks/21374}{21374} & !Tention: A Story of Boy-Life du... & 92 & 0.2195 & \adjustimage{height=12px,width=45px,valign=m}{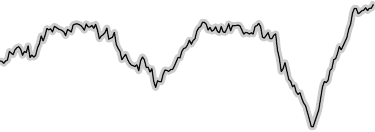} \\
\href{http://www.gutenberg.org/ebooks/24499}{24499} & The Green Carnation & 65 & 0.1986 & \adjustimage{height=12px,width=45px,valign=m}{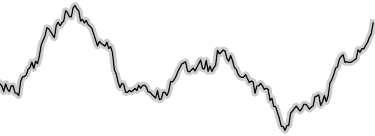} \\
\href{http://www.gutenberg.org/ebooks/544}{544} & Anne's House of Dreams & 586 & 0.1676 & \adjustimage{height=12px,width=45px,valign=m}{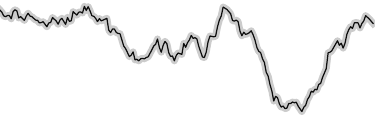} \\
\href{http://www.gutenberg.org/ebooks/4922}{4922} & Bar-20 Days & 57 & 0.1648 & \adjustimage{height=12px,width=45px,valign=m}{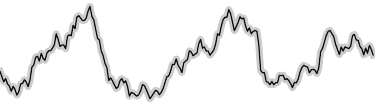} \\
\href{http://www.gutenberg.org/ebooks/8681}{8681} & The Face and the Mask & 82 & 0.1528 & \adjustimage{height=12px,width=45px,valign=m}{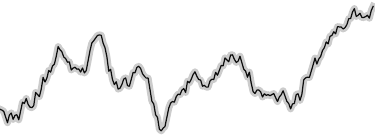} \\
\href{http://www.gutenberg.org/ebooks/1805}{1805} & The Gentle Grafter & 84 & 0.1505 & \adjustimage{height=12px,width=45px,valign=m}{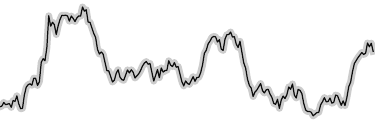} \\
\href{http://www.gutenberg.org/ebooks/10443}{10443} & The Rayner-Slade Amalgamation & 61 & 0.1486 & \adjustimage{height=12px,width=45px,valign=m}{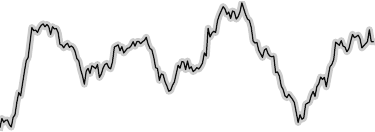} \\
\href{http://www.gutenberg.org/ebooks/16339}{16339} & The Passenger from Calais & 125 & 0.1454 & \adjustimage{height=12px,width=45px,valign=m}{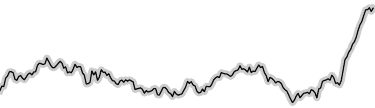} \\
\end{longtable}
\begin{longtable}{l | l | l | l | c}
\caption{Top 10 stories which are are represented by core emotional arc 10, sorted by downloads.}\\
ID & Title & DL's & max($W$[i,:]) & Arc\\
\hline
\endhead
\href{http://www.gutenberg.org/ebooks/25305}{25305} & Memoirs Of Fanny Hill: A New and ... & 2,222 & 0.1296 & \adjustimage{height=12px,width=45px,valign=m}{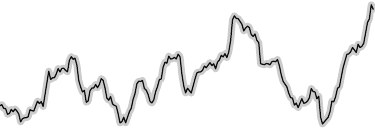} \\
\href{http://www.gutenberg.org/ebooks/544}{544} & Anne's House of Dreams & 586 & 0.1676 & \adjustimage{height=12px,width=45px,valign=m}{all-timeseries/544.pdf} \\
\href{http://www.gutenberg.org/ebooks/4540}{4540} & In His Steps & 230 & 0.0902 & \adjustimage{height=12px,width=45px,valign=m}{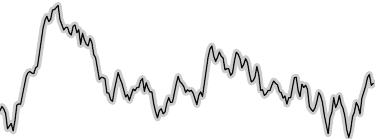} \\
\href{http://www.gutenberg.org/ebooks/38070}{38070} & The Norwegian Fairy Book & 149 & 0.1044 & \adjustimage{height=12px,width=45px,valign=m}{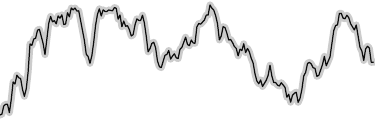} \\
\href{http://www.gutenberg.org/ebooks/16339}{16339} & The Passenger from Calais & 125 & 0.1454 & \adjustimage{height=12px,width=45px,valign=m}{all-timeseries/16339.pdf} \\
\href{http://www.gutenberg.org/ebooks/1535}{1535} & The Tragedy of Coriolanus & 116 & 0.1137 & \adjustimage{height=12px,width=45px,valign=m}{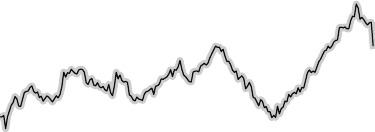} \\
\href{http://www.gutenberg.org/ebooks/35517}{35517} & The Three Impostors; or, The Tra... & 111 & 0.1148 & \adjustimage{height=12px,width=45px,valign=m}{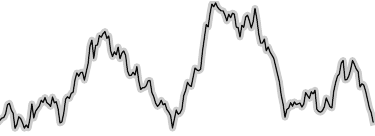} \\
\href{http://www.gutenberg.org/ebooks/21374}{21374} & !Tention: A Story of Boy-Life du... & 92 & 0.2195 & \adjustimage{height=12px,width=45px,valign=m}{all-timeseries/21374.pdf} \\
\href{http://www.gutenberg.org/ebooks/24775}{24775} & Up the River; or, Yachting on th... & 88 & 0.1449 & \adjustimage{height=12px,width=45px,valign=m}{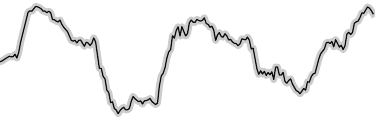} \\
\end{longtable}
\begin{longtable}{l | l | l | l | c}
\caption{Stories which are are represented by core emotional arc 11, sorted by the variance explained in their emotional arc by core emotional arc 11.}\\
ID & Title & DL's & max($W$[i,:]) & Arc\\
\hline
\endhead
\href{http://www.gutenberg.org/ebooks/5343}{5343} & Rainbow Valley & 257 & 0.1533 & \adjustimage{height=12px,width=45px,valign=m}{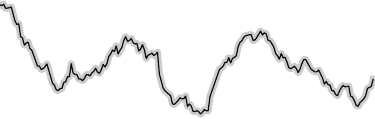} \\
\href{http://www.gutenberg.org/ebooks/1987}{1987} & The Outlet & 40 & 0.1444 & \adjustimage{height=12px,width=45px,valign=m}{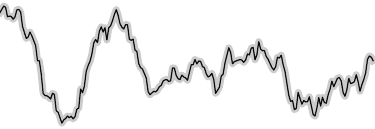} \\
\href{http://www.gutenberg.org/ebooks/13882}{13882} & John Thorndyke's Cases
: related ... & 217 & 0.1343 & \adjustimage{height=12px,width=45px,valign=m}{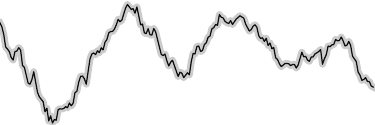} \\
\href{http://www.gutenberg.org/ebooks/1167}{1167} & A Strange Disappearance & 270 & 0.1282 & \adjustimage{height=12px,width=45px,valign=m}{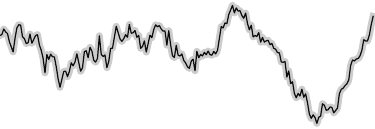} \\
\href{http://www.gutenberg.org/ebooks/7477}{7477} & The Book of Wonder & 244 & 0.1276 & \adjustimage{height=12px,width=45px,valign=m}{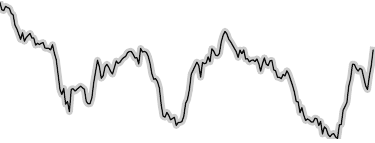} \\
\href{http://www.gutenberg.org/ebooks/6753}{6753} & Psmith in the City & 199 & 0.1245 & \adjustimage{height=12px,width=45px,valign=m}{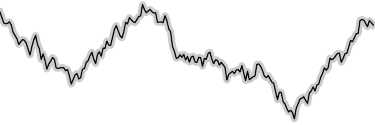} \\
\href{http://www.gutenberg.org/ebooks/5265}{5265} & The Ball and the Cross & 118 & 0.1229 & \adjustimage{height=12px,width=45px,valign=m}{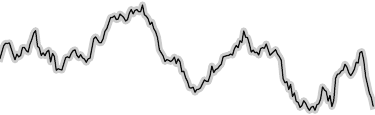} \\
\href{http://www.gutenberg.org/ebooks/974}{974} & The Secret Agent: A Simple Tale & 1,142 & 0.1224 & \adjustimage{height=12px,width=45px,valign=m}{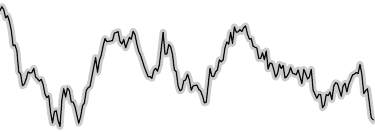} \\
\end{longtable}
\begin{longtable}{l | l | l | l  | c}
\caption{Top 10 stories which are are represented by core emotional arc 11, sorted by downloads.}\\
ID & Title & DL's & max(W[i,:]) & Arc\\
\hline
\endhead
\href{http://www.gutenberg.org/ebooks/2641}{2641} & A Room with a View & 1,354 & 0.0681 & \adjustimage{height=12px,width=45px,valign=m}{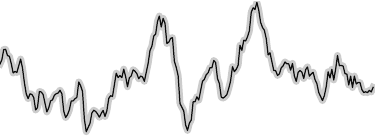} \\
\href{http://www.gutenberg.org/ebooks/974}{974} & The Secret Agent: A Simple Tale & 1,142 & 0.1224 & \adjustimage{height=12px,width=45px,valign=m}{all-timeseries/974.pdf} \\
\href{http://www.gutenberg.org/ebooks/767}{767} & Agnes Grey & 287 & 0.0856 & \adjustimage{height=12px,width=45px,valign=m}{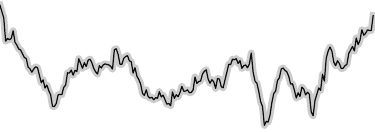} \\
\href{http://www.gutenberg.org/ebooks/1167}{1167} & A Strange Disappearance & 270 & 0.1282 & \adjustimage{height=12px,width=45px,valign=m}{all-timeseries/1167.pdf} \\
\href{http://www.gutenberg.org/ebooks/5343}{5343} & Rainbow Valley & 257 & 0.1533 & \adjustimage{height=12px,width=45px,valign=m}{all-timeseries/5343.pdf} \\
\href{http://www.gutenberg.org/ebooks/7477}{7477} & The Book of Wonder & 244 & 0.1276 & \adjustimage{height=12px,width=45px,valign=m}{all-timeseries/7477.pdf} \\
\href{http://www.gutenberg.org/ebooks/13882}{13882} & John Thorndyke's Cases
: related ... & 217 & 0.1343 & \adjustimage{height=12px,width=45px,valign=m}{all-timeseries/13882.pdf} \\
\href{http://www.gutenberg.org/ebooks/6753}{6753} & Psmith in the City & 199 & 0.1245 & \adjustimage{height=12px,width=45px,valign=m}{all-timeseries/6753.pdf} \\
\href{http://www.gutenberg.org/ebooks/10832}{10832} & Carnacki, the Ghost Finder & 135 & 0.1032 & \adjustimage{height=12px,width=45px,valign=m}{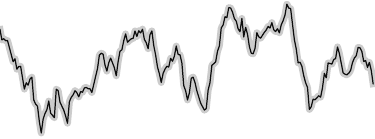} \\
\end{longtable}
\begin{longtable}{l | l | l | l | c}
\caption{Stories which are are represented by core emotional arc 12, sorted by the variance explained in their emotional arc by core emotional arc 12.}\\
ID & Title & DL's & max($W$[i,:]) & Arc\\
\hline
\endhead
\href{http://www.gutenberg.org/ebooks/318}{318} & John Barleycorn & 130 & 0.1909 & \adjustimage{height=12px,width=45px,valign=m}{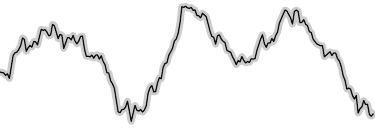} \\
\href{http://www.gutenberg.org/ebooks/1595}{1595} & Whirligigs & 94 & 0.1563 & \adjustimage{height=12px,width=45px,valign=m}{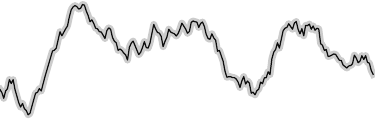} \\
\href{http://www.gutenberg.org/ebooks/25728}{25728} & Desert Conquest; or, Precious Wa... & 58 & 0.1543 & \adjustimage{height=12px,width=45px,valign=m}{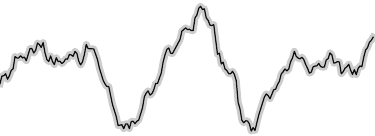} \\
\href{http://www.gutenberg.org/ebooks/557}{557} & Puck of Pook's Hill & 50 & 0.1482 & \adjustimage{height=12px,width=45px,valign=m}{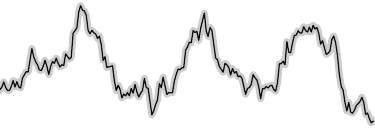} \\
\href{http://www.gutenberg.org/ebooks/15976}{15976} & Puck of Pook's Hill & 40 & 0.1376 & \adjustimage{height=12px,width=45px,valign=m}{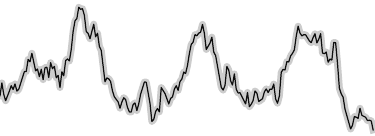} \\
\href{http://www.gutenberg.org/ebooks/26027}{26027} & Puck of Pook's Hill & 78 & 0.1357 & \adjustimage{height=12px,width=45px,valign=m}{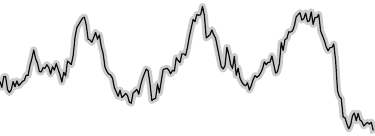} \\
\href{http://www.gutenberg.org/ebooks/3326}{3326} & The Well-Beloved: A Sketch of a ... & 76 & 0.1226 & \adjustimage{height=12px,width=45px,valign=m}{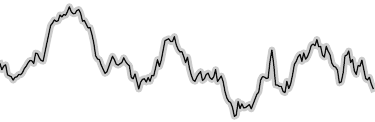} \\
\href{http://www.gutenberg.org/ebooks/1611}{1611} & Seventeen
: A Tale of Youth and S... & 56 & 0.1145 & \adjustimage{height=12px,width=45px,valign=m}{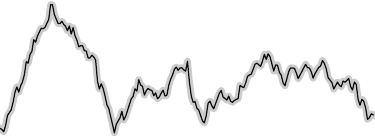} \\
\end{longtable}
\begin{longtable}{l | l | l | l | c}
\caption{Top 10 stories which are are represented by core emotional arc 12, sorted by downloads.}\\
ID & Title & DL's & max($W$[i,:]) & Arc\\
\hline
\endhead
\href{http://www.gutenberg.org/ebooks/624}{624} & Looking Backward, 2000 to 1887 & 679 & 0.1060 & \adjustimage{height=12px,width=45px,valign=m}{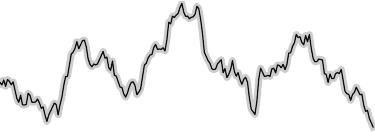} \\
\href{http://www.gutenberg.org/ebooks/610}{610} & Idylls of the King & 494 & 0.1075 & \adjustimage{height=12px,width=45px,valign=m}{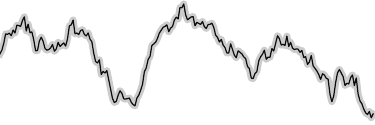} \\
\href{http://www.gutenberg.org/ebooks/316}{316} & The Golden Road & 151 & 0.0864 & \adjustimage{height=12px,width=45px,valign=m}{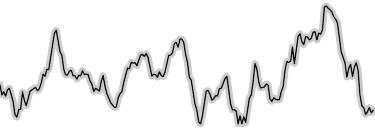} \\
\href{http://www.gutenberg.org/ebooks/318}{318} & John Barleycorn & 130 & 0.1909 & \adjustimage{height=12px,width=45px,valign=m}{all-timeseries/318.pdf} \\
\href{http://www.gutenberg.org/ebooks/12803}{12803} & Headlong Hall & 124 & 0.1113 & \adjustimage{height=12px,width=45px,valign=m}{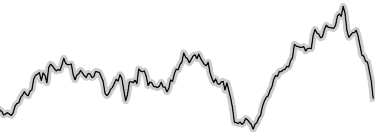} \\
\href{http://www.gutenberg.org/ebooks/37532}{37532} & The Scottish Fairy Book & 98 & 0.1036 & \adjustimage{height=12px,width=45px,valign=m}{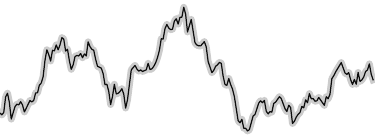} \\
\href{http://www.gutenberg.org/ebooks/1595}{1595} & Whirligigs & 94 & 0.1563 & \adjustimage{height=12px,width=45px,valign=m}{all-timeseries/1595.pdf} \\
\href{http://www.gutenberg.org/ebooks/6840}{6840} & Queen Lucia & 81 & 0.0801 & \adjustimage{height=12px,width=45px,valign=m}{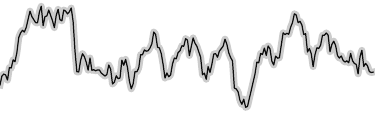} \\
\href{http://www.gutenberg.org/ebooks/26027}{26027} & Puck of Pook's Hill & 78 & 0.1357 & \adjustimage{height=12px,width=45px,valign=m}{all-timeseries/26027.pdf} \\
\end{longtable}
\begin{longtable}{l | l | l | l | c}
\caption{Stories which are are represented by core emotional arc 13, sorted by the variance explained in their emotional arc by core emotional arc 13.}\\
ID & Title & DL's & max($W$[i,:]) & Arc\\
\hline
\endhead
\href{http://www.gutenberg.org/ebooks/3815}{3815} & Rolling Stones & 107 & 0.1377 & \adjustimage{height=12px,width=45px,valign=m}{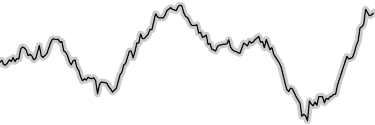} \\
\href{http://www.gutenberg.org/ebooks/26732}{26732} & Free Air & 42 & 0.1295 & \adjustimage{height=12px,width=45px,valign=m}{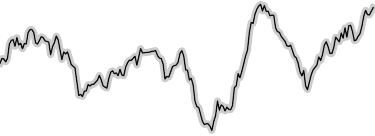} \\
\href{http://www.gutenberg.org/ebooks/30836}{30836} & Seven Keys to Baldpate & 57 & 0.1245 & \adjustimage{height=12px,width=45px,valign=m}{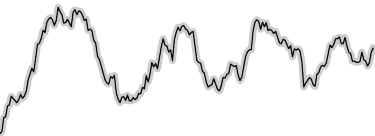} \\
\href{http://www.gutenberg.org/ebooks/22342}{22342} & Supermind & 61 & 0.1185 & \adjustimage{height=12px,width=45px,valign=m}{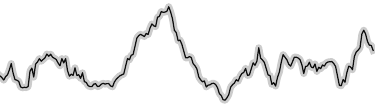} \\
\href{http://www.gutenberg.org/ebooks/9925}{9925} & Black Jack & 111 & 0.1174 & \adjustimage{height=12px,width=45px,valign=m}{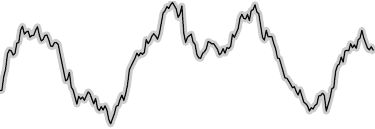} \\
\href{http://www.gutenberg.org/ebooks/21687}{21687} & The Youngest Girl in the Fifth: ... & 59 & 0.1097 & \adjustimage{height=12px,width=45px,valign=m}{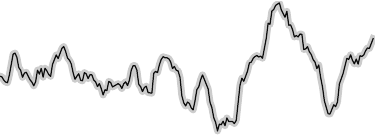} \\
\href{http://www.gutenberg.org/ebooks/19527}{19527} & The Yukon Trail: A Tale of the N... & 44 & 0.1085 & \adjustimage{height=12px,width=45px,valign=m}{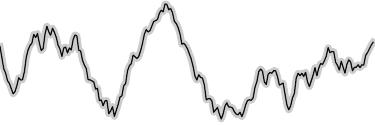} \\
\href{http://www.gutenberg.org/ebooks/6312}{6312} & Representative Men: Seven Lectures & 170 & 0.0986 & \adjustimage{height=12px,width=45px,valign=m}{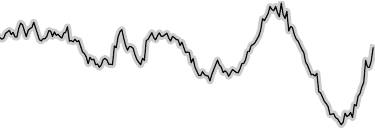} \\
\href{http://www.gutenberg.org/ebooks/33391}{33391} & Bill Nye's Cordwood & 248 & 0.0888 & \adjustimage{height=12px,width=45px,valign=m}{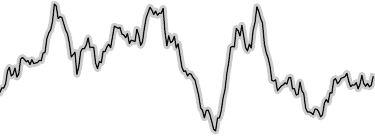} \\
\end{longtable}
\begin{longtable}{l | l | l | l  | c}
\caption{Top 10 stories which are are represented by core emotional arc 13, sorted by downloads.}\\
ID & Title & DL's & max(W[i,:]) & Arc\\
\hline
\endhead
\href{http://www.gutenberg.org/ebooks/33391}{33391} & Bill Nye's Cordwood & 248 & 0.0888 & \adjustimage{height=12px,width=45px,valign=m}{all-timeseries/33391.pdf} \\
\href{http://www.gutenberg.org/ebooks/6312}{6312} & Representative Men: Seven Lectures & 170 & 0.0986 & \adjustimage{height=12px,width=45px,valign=m}{all-timeseries/6312.pdf} \\
\href{http://www.gutenberg.org/ebooks/9925}{9925} & Black Jack & 111 & 0.1174 & \adjustimage{height=12px,width=45px,valign=m}{all-timeseries/9925.pdf} \\
\href{http://www.gutenberg.org/ebooks/3815}{3815} & Rolling Stones & 107 & 0.1377 & \adjustimage{height=12px,width=45px,valign=m}{all-timeseries/3815.pdf} \\
\href{http://www.gutenberg.org/ebooks/22342}{22342} & Supermind & 61 & 0.1185 & \adjustimage{height=12px,width=45px,valign=m}{all-timeseries/22342.pdf} \\
\href{http://www.gutenberg.org/ebooks/21687}{21687} & The Youngest Girl in the Fifth: ... & 59 & 0.1097 & \adjustimage{height=12px,width=45px,valign=m}{all-timeseries/21687.pdf} \\
\href{http://www.gutenberg.org/ebooks/30836}{30836} & Seven Keys to Baldpate & 57 & 0.1245 & \adjustimage{height=12px,width=45px,valign=m}{all-timeseries/30836.pdf} \\
\href{http://www.gutenberg.org/ebooks/19527}{19527} & The Yukon Trail: A Tale of the N... & 44 & 0.1085 & \adjustimage{height=12px,width=45px,valign=m}{all-timeseries/19527.pdf} \\
\href{http://www.gutenberg.org/ebooks/26732}{26732} & Free Air & 42 & 0.1295 & \adjustimage{height=12px,width=45px,valign=m}{all-timeseries/26732.pdf} \\
\end{longtable}
\begin{longtable}{l | l | l | l | c}
\caption{Stories which are are represented by core emotional arc 14, sorted by the variance explained in their emotional arc by core emotional arc 14.}\\
ID & Title & DL's & max($W$[i,:]) & Arc\\
\hline
\endhead
\href{http://www.gutenberg.org/ebooks/5077}{5077} & Marmion: A Tale of Flodden Field... & 125 & 0.1476 & \adjustimage{height=12px,width=45px,valign=m}{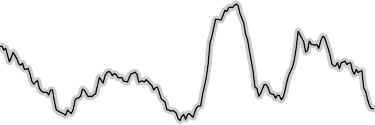} \\
\href{http://www.gutenberg.org/ebooks/20869}{20869} & The Skylark of Space & 246 & 0.1457 & \adjustimage{height=12px,width=45px,valign=m}{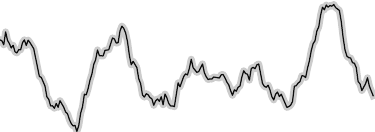} \\
\href{http://www.gutenberg.org/ebooks/15272}{15272} & Spenser's The Faerie Queene, Book I & 978 & 0.1302 & \adjustimage{height=12px,width=45px,valign=m}{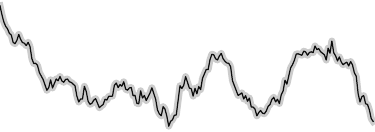} \\
\href{http://www.gutenberg.org/ebooks/1027}{1027} & The Lone Star Ranger: A Romance ... & 253 & 0.1277 & \adjustimage{height=12px,width=45px,valign=m}{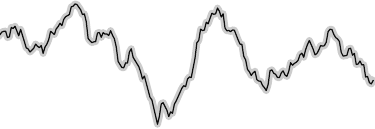} \\
\href{http://www.gutenberg.org/ebooks/2233}{2233} & A Damsel in Distress & 201 & 0.1256 & \adjustimage{height=12px,width=45px,valign=m}{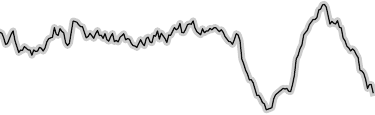} \\
\href{http://www.gutenberg.org/ebooks/27690}{27690} & Nobody's Girl: (En Famille) & 70 & 0.1219 & \adjustimage{height=12px,width=45px,valign=m}{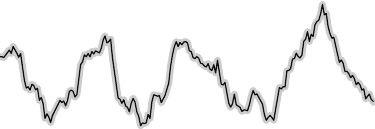} \\
\href{http://www.gutenberg.org/ebooks/32759}{32759} & Red Nails & 151 & 0.1158 & \adjustimage{height=12px,width=45px,valign=m}{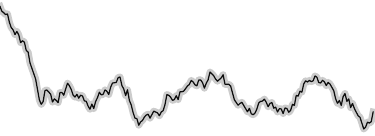} \\
\href{http://www.gutenberg.org/ebooks/27063}{27063} & The Hero & 60 & 0.1155 & \adjustimage{height=12px,width=45px,valign=m}{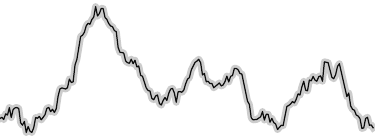} \\
\end{longtable}
\begin{longtable}{l | l | l | l | c}
\caption{Top 10 stories which are are represented by core emotional arc 14, sorted by downloads.}\\
ID & Title & DL's & max($W$[i,:]) & Arc\\
\hline
\endhead
\href{http://www.gutenberg.org/ebooks/105}{105} & Persuasion & 2,535 & 0.0980 & \adjustimage{height=12px,width=45px,valign=m}{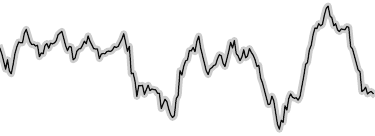} \\
\href{http://www.gutenberg.org/ebooks/15272}{15272} & Spenser's The Faerie Queene, Book I & 978 & 0.1302 & \adjustimage{height=12px,width=45px,valign=m}{all-timeseries/15272.pdf} \\
\href{http://www.gutenberg.org/ebooks/1027}{1027} & The Lone Star Ranger: A Romance ... & 253 & 0.1277 & \adjustimage{height=12px,width=45px,valign=m}{all-timeseries/1027.pdf} \\
\href{http://www.gutenberg.org/ebooks/20869}{20869} & The Skylark of Space & 246 & 0.1457 & \adjustimage{height=12px,width=45px,valign=m}{all-timeseries/20869.pdf} \\
\href{http://www.gutenberg.org/ebooks/619}{619} & The Warden & 215 & 0.0946 & \adjustimage{height=12px,width=45px,valign=m}{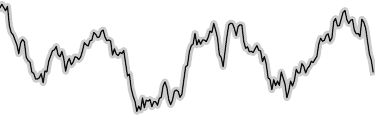} \\
\href{http://www.gutenberg.org/ebooks/2233}{2233} & A Damsel in Distress & 201 & 0.1256 & \adjustimage{height=12px,width=45px,valign=m}{all-timeseries/2233.pdf} \\
\href{http://www.gutenberg.org/ebooks/2804}{2804} & Rose in Bloom
: A Sequel to "Eigh... & 168 & 0.1083 & \adjustimage{height=12px,width=45px,valign=m}{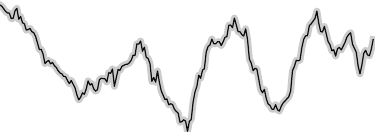} \\
\href{http://www.gutenberg.org/ebooks/32759}{32759} & Red Nails & 151 & 0.1158 & \adjustimage{height=12px,width=45px,valign=m}{all-timeseries/32759.pdf} \\
\href{http://www.gutenberg.org/ebooks/5077}{5077} & Marmion: A Tale of Flodden Field... & 125 & 0.1476 & \adjustimage{height=12px,width=45px,valign=m}{all-timeseries/5077.pdf} \\
\end{longtable}
\begin{longtable}{l | l | l | l | c}
\caption{Stories which are are represented by core emotional arc 15, sorted by the variance explained in their emotional arc by core emotional arc 15.}\\
ID & Title & DL's & max($W$[i,:]) & Arc\\
\hline
\endhead
\href{http://www.gutenberg.org/ebooks/33066}{33066} & The Garden of Eden & 72 & 0.1165 & \adjustimage{height=12px,width=45px,valign=m}{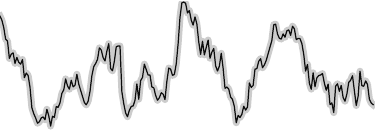} \\
\href{http://www.gutenberg.org/ebooks/40852}{40852} & Instigations: Together with An Es... & 79 & 0.0806 & \adjustimage{height=12px,width=45px,valign=m}{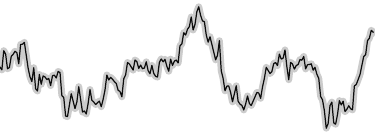} \\
\end{longtable}
\begin{longtable}{l | l | l | l  | c}
\caption{Top 10 stories which are are represented by core emotional arc 15, sorted by downloads.}\\
ID & Title & DL's & max(W[i,:]) & Arc\\
\hline
\endhead
\href{http://www.gutenberg.org/ebooks/40852}{40852} & Instigations: Together with An Es... & 79 & 0.0806 & \adjustimage{height=12px,width=45px,valign=m}{all-timeseries/40852.pdf} \\
\href{http://www.gutenberg.org/ebooks/33066}{33066} & The Garden of Eden & 72 & 0.1165 & \adjustimage{height=12px,width=45px,valign=m}{all-timeseries/33066.pdf} \\
\end{longtable}
\begin{longtable}{l | l | l | l | c}
\caption{Stories which are are represented by core emotional arc 16, sorted by the variance explained in their emotional arc by core emotional arc 16.}\\
ID & Title & DL's & max($W$[i,:]) & Arc\\
\hline
\endhead
\href{http://www.gutenberg.org/ebooks/872}{872} & Reprinted Pieces & 49 & 0.1103 & \adjustimage{height=12px,width=45px,valign=m}{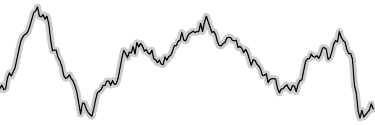} \\
\href{http://www.gutenberg.org/ebooks/8673}{8673} & A Columbus of Space & 56 & 0.1077 & \adjustimage{height=12px,width=45px,valign=m}{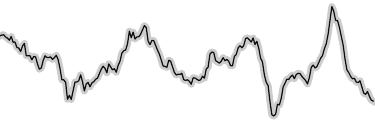} \\
\href{http://www.gutenberg.org/ebooks/434}{434} & The Circular Staircase & 189 & 0.0978 & \adjustimage{height=12px,width=45px,valign=m}{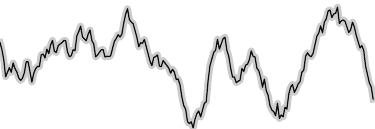} \\
\href{http://www.gutenberg.org/ebooks/1263}{1263} & The Glimpses of the Moon & 58 & 0.0864 & \adjustimage{height=12px,width=45px,valign=m}{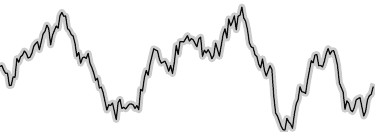} \\
\href{http://www.gutenberg.org/ebooks/3075}{3075} & The Return & 73 & 0.0759 & \adjustimage{height=12px,width=45px,valign=m}{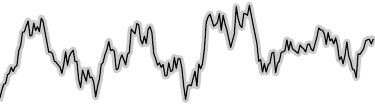} \\
\end{longtable}%
\end{scriptsize}
\end{singlespacing}

\clearpage
\pagebreak
\section{Additional Hierarchical Clustering Figures}
\label{sec:clustering-supp}

In the section, we include additional results from the hierarchical clustering analysis.
The distance function between clusters is defined in the \verb|scipy| package using the incremental algorithm, starting with all arcs as separate clusters and iteratively merging them:
$$ d(u,v) = \sqrt{\frac{|v|+|s|}
                    {T}d(v,s)^2
             + \frac{|v|+|t|}
                    {T}d(v,t)^2
             - \frac{|v|}
                    {T}d(s,t)^2}$$
where $|v|$ denotes the cardinality of set $v$ (single arcs have cardinality 1), $u$ is the merged cluster of $s,t$, the denominator $T$ is the sum of the sizes, and $v$ is an unused cluster.
Similar to the MATLAB implementation, this relies on a nearest-neighbor chain to be computed efficiently.

\begin{figure*}[!htb]
  \centering
\includegraphics[width=0.98\textwidth]{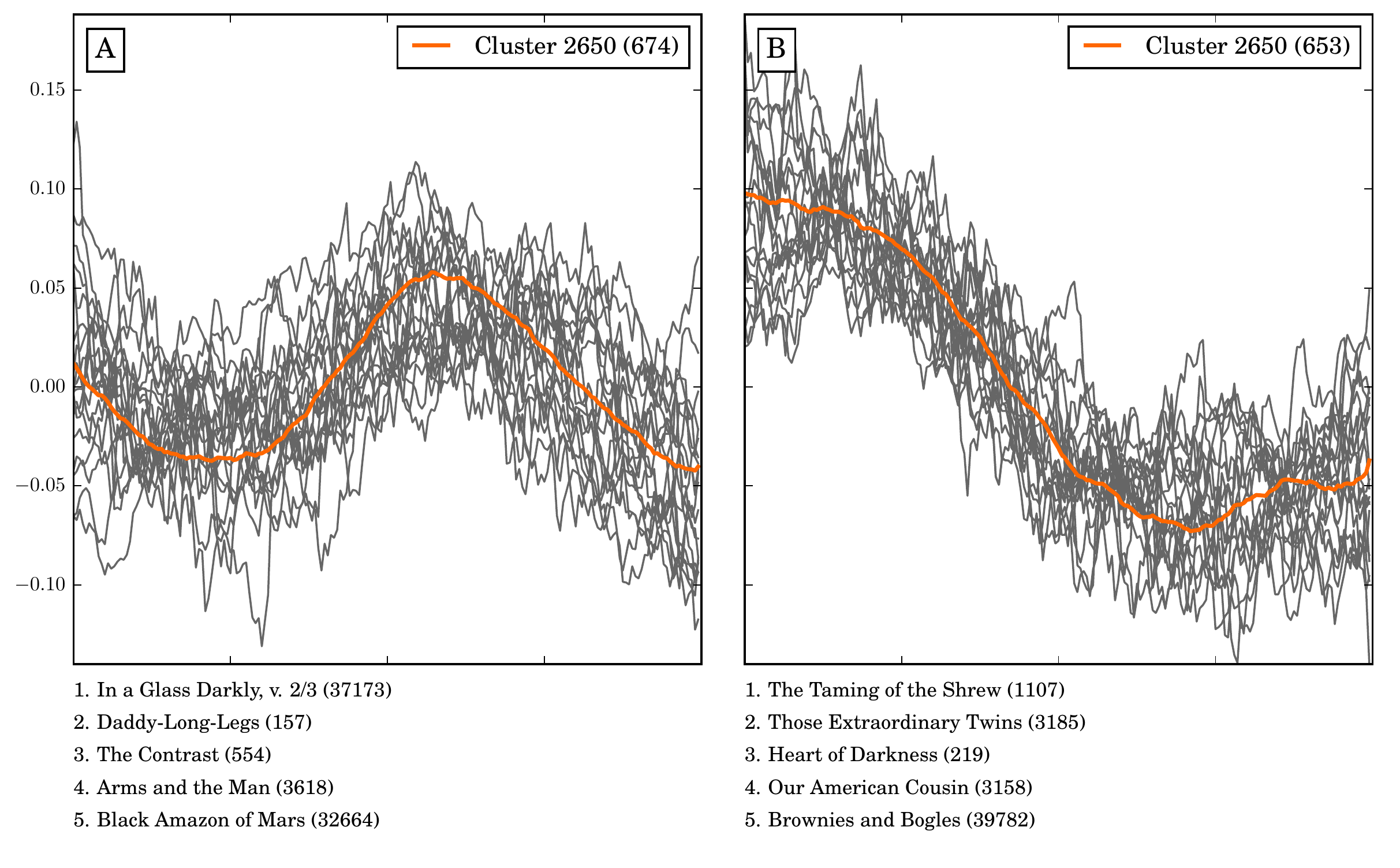}
  \caption[]{
    The 2 clusters identified by Agglomerative Clustering using Ward's method.
  }
  \label{fig:ward-cluster-2}
\end{figure*}

\begin{figure*}[!htb]
  \centering
\includegraphics[width=0.98\textwidth]{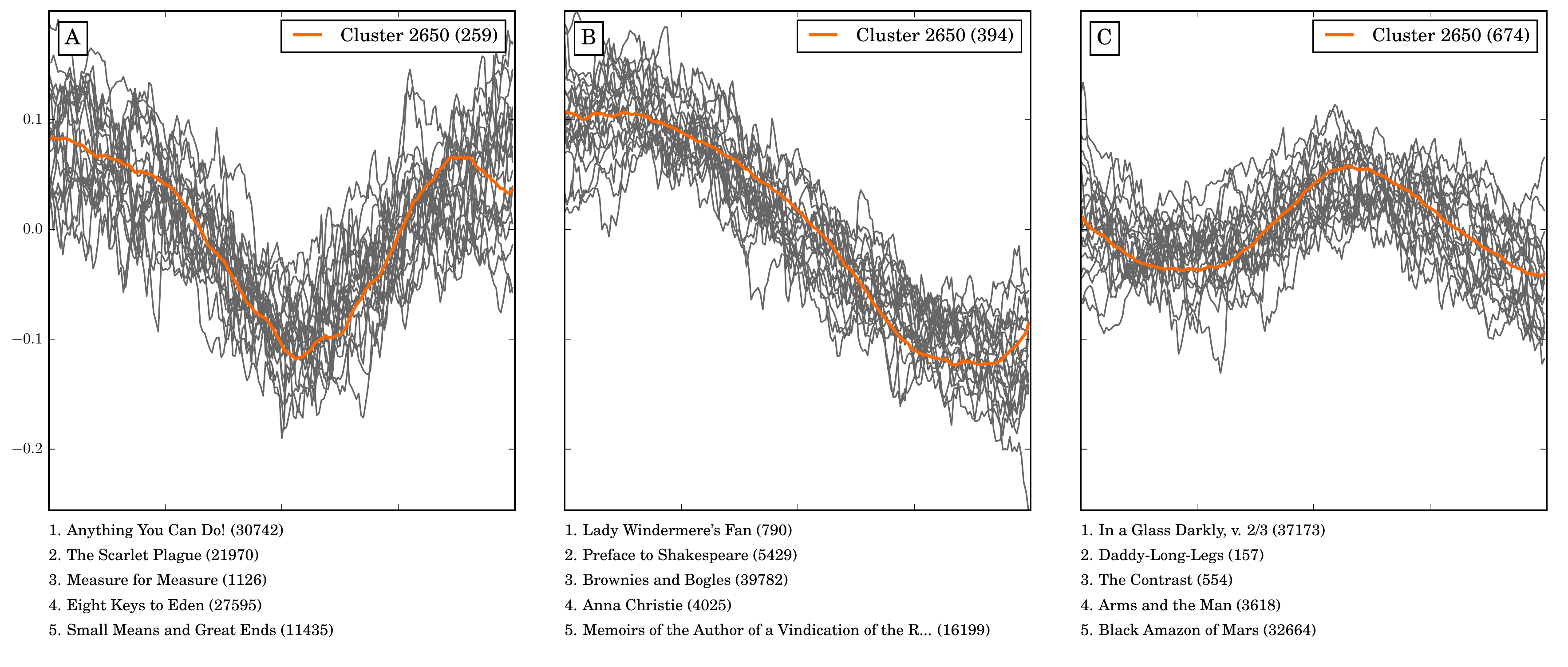}
  \caption[]{
    The 3 clusters identified by Agglomerative Clustering using Ward's method.
  }
  \label{fig:ward-cluster-3}
\end{figure*}

\begin{figure*}[!htb]
  \centering
\includegraphics[width=0.98\textwidth]{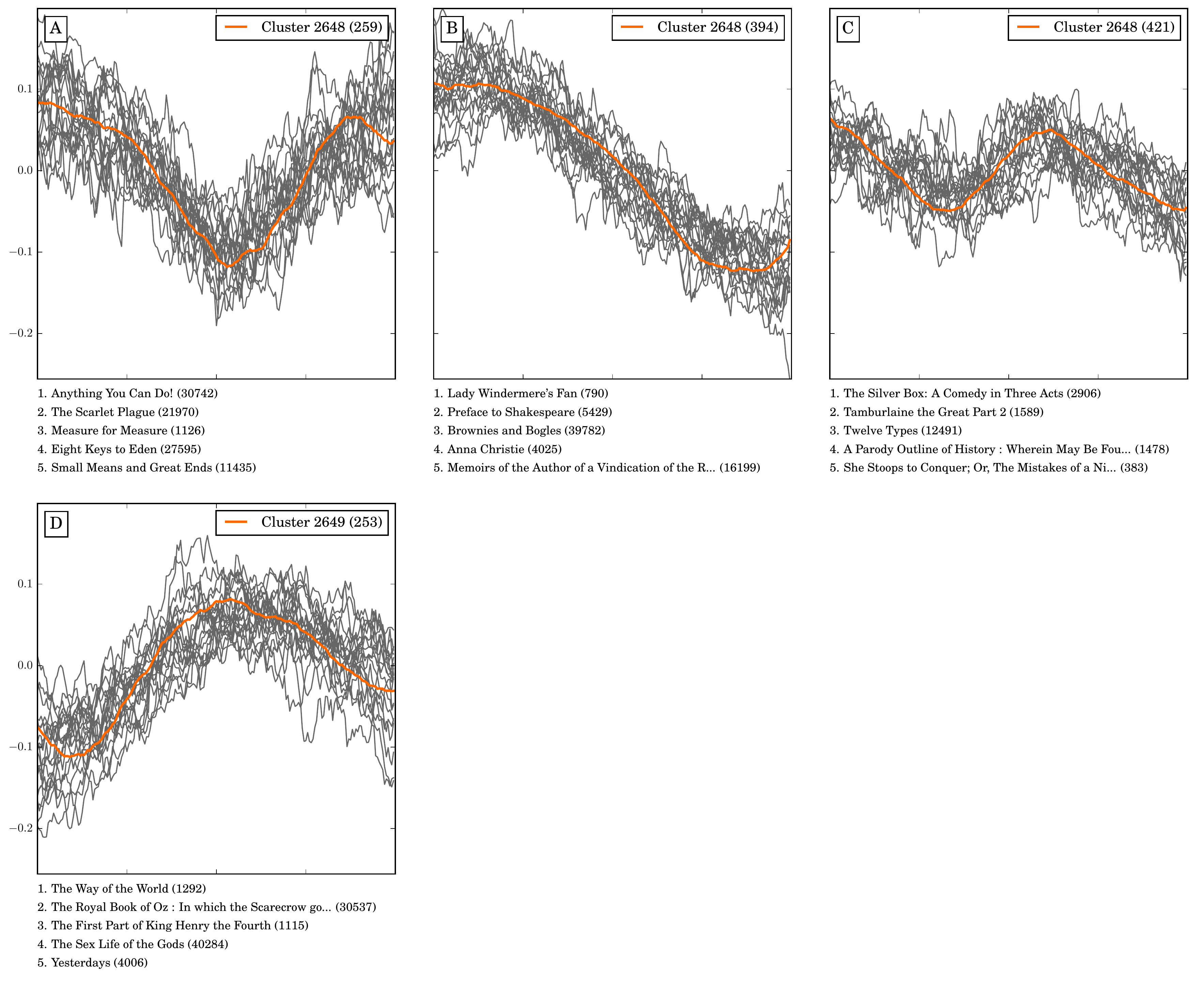}
  \caption[]{
    The 4 clusters identified by Agglomerative Clustering using Ward's method.
  }
    \label{fig:ward-cluster-4}
\end{figure*}

\begin{figure*}[!htb]
  \centering
\includegraphics[width=0.98\textwidth]{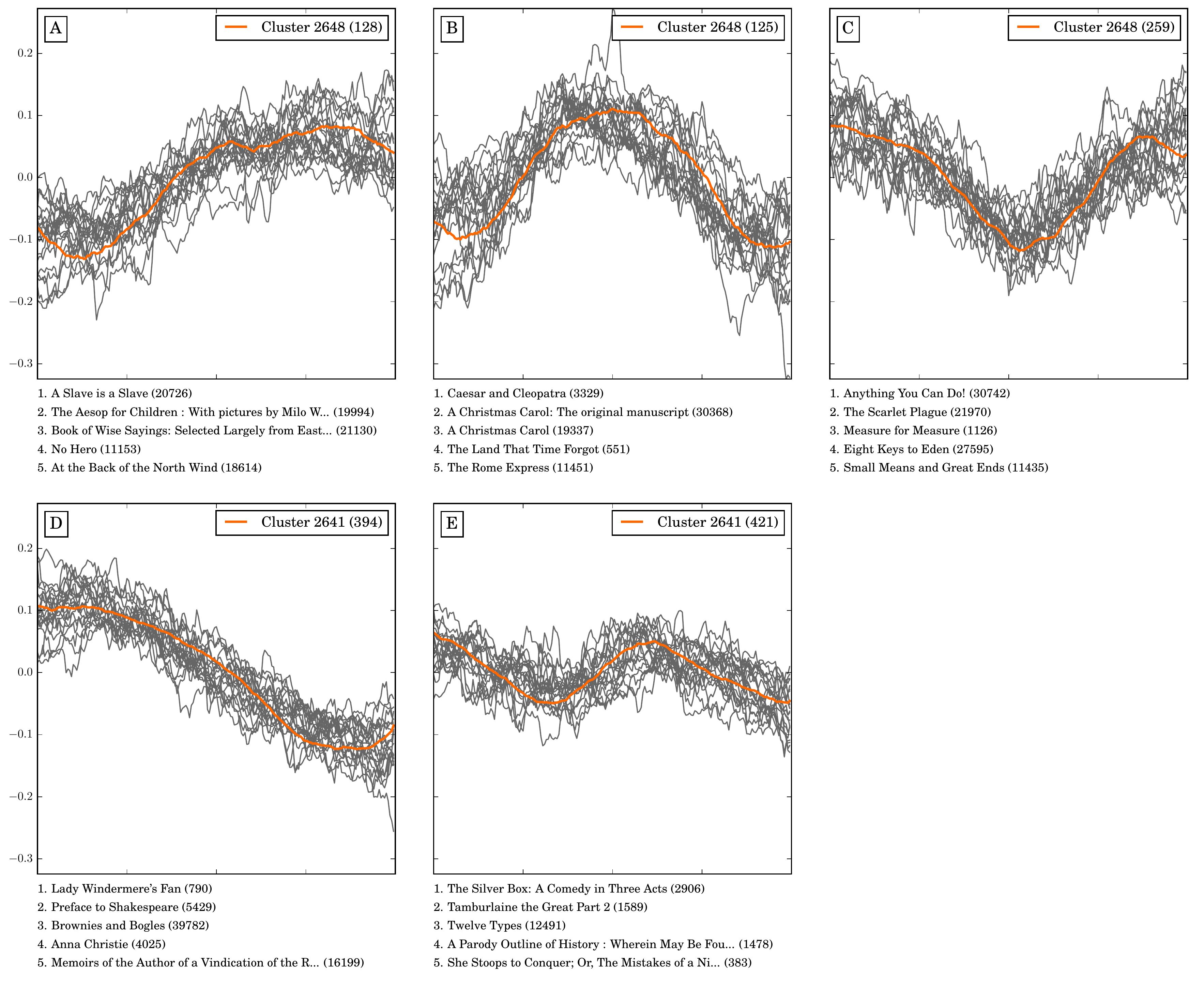}
  \caption[]{
    The 5 clusters identified by Agglomerative Clustering using Ward's method.
  }
    \label{fig:ward-cluster-5}
\end{figure*}

\begin{figure*}[!htb]
  \centering
\includegraphics[width=0.98\textwidth]{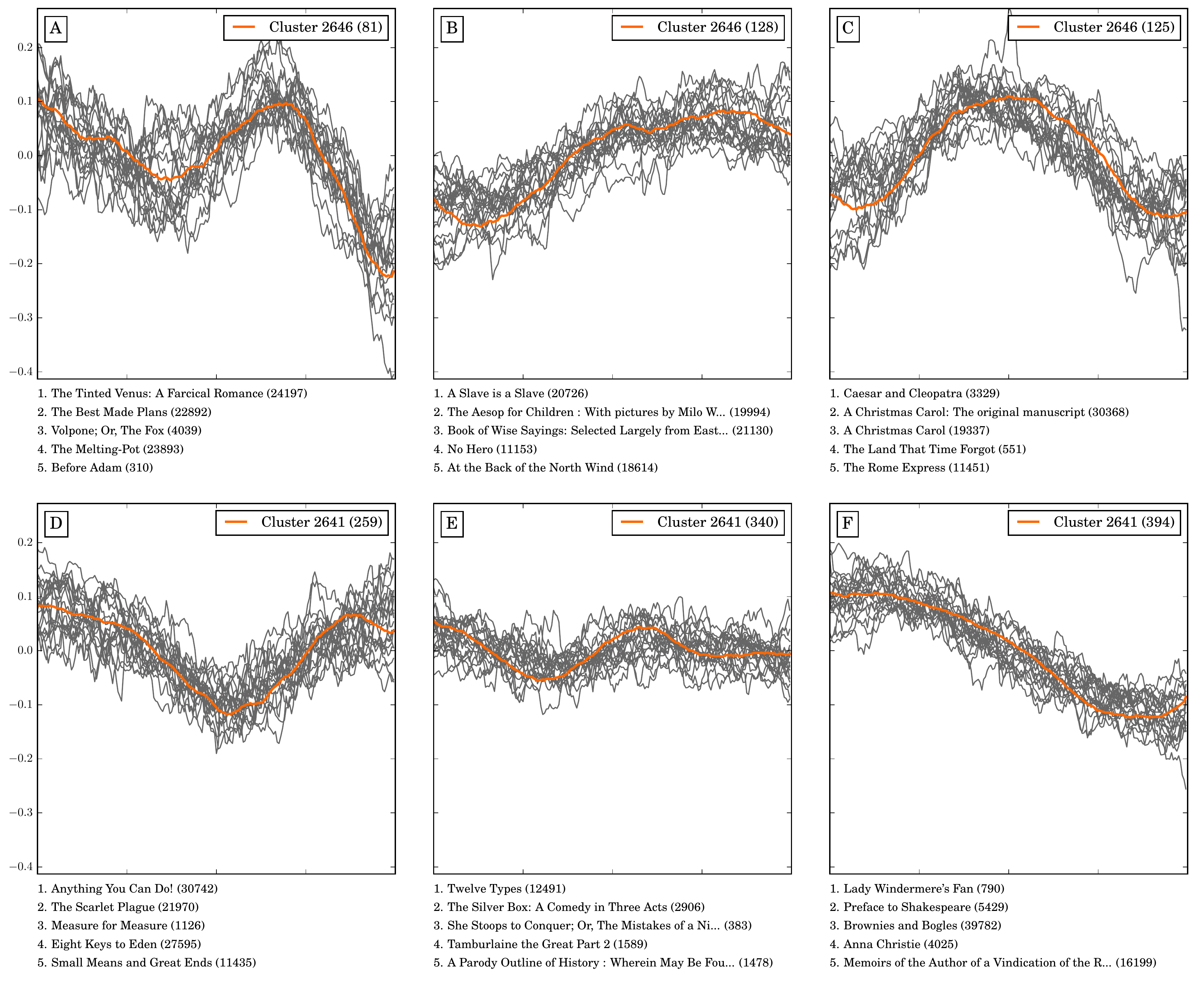}
  \caption[]{
    The 6 clusters identified by Agglomerative Clustering using Ward's method.
  }
    \label{fig:ward-cluster-6}
\end{figure*}

\begin{figure*}[!htb]
  \centering
\includegraphics[width=0.98\textwidth]{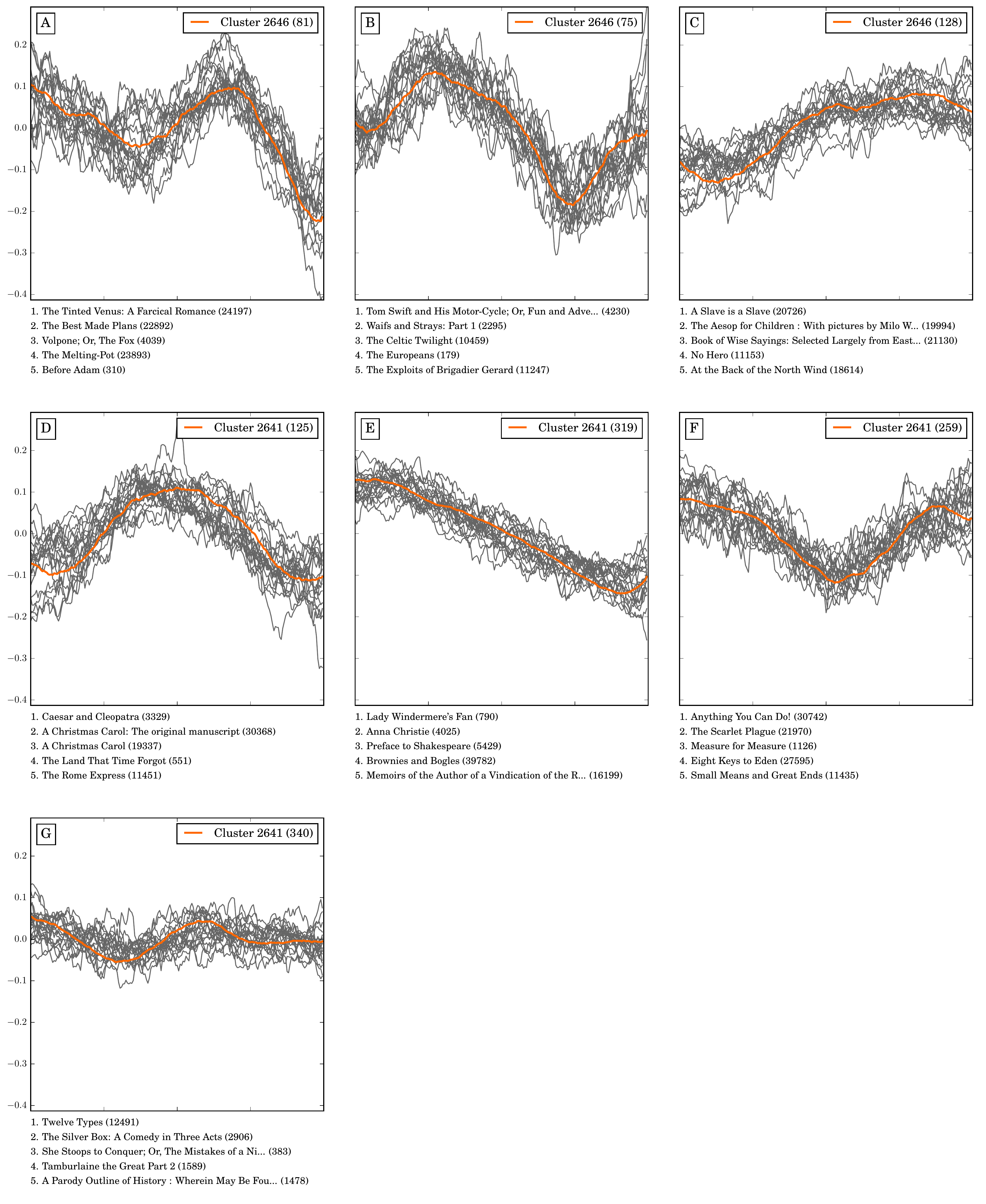}
  \caption[]{
    The 7 clusters identified by Agglomerative Clustering using Ward's method.
  }
    \label{fig:ward-cluster-7}
\end{figure*}

\begin{figure*}[!htb]
  \centering
\includegraphics[width=0.98\textwidth]{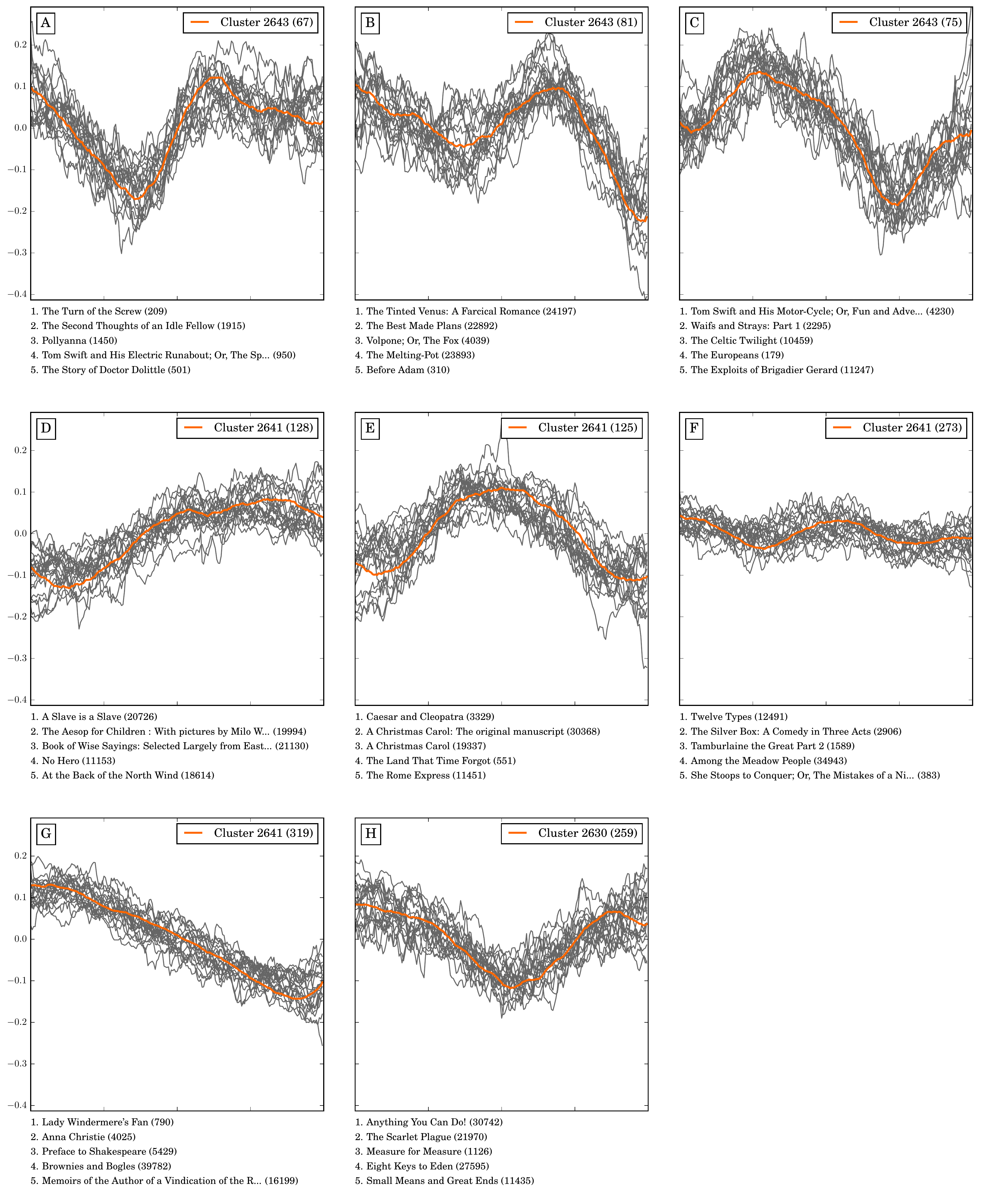}
  \caption[]{
    The 8 clusters identified by Agglomerative Clustering using Ward's method.
  }
    \label{fig:ward-cluster-8}
\end{figure*}

\begin{figure*}[!htb]
  \centering
\includegraphics[width=0.98\textwidth]{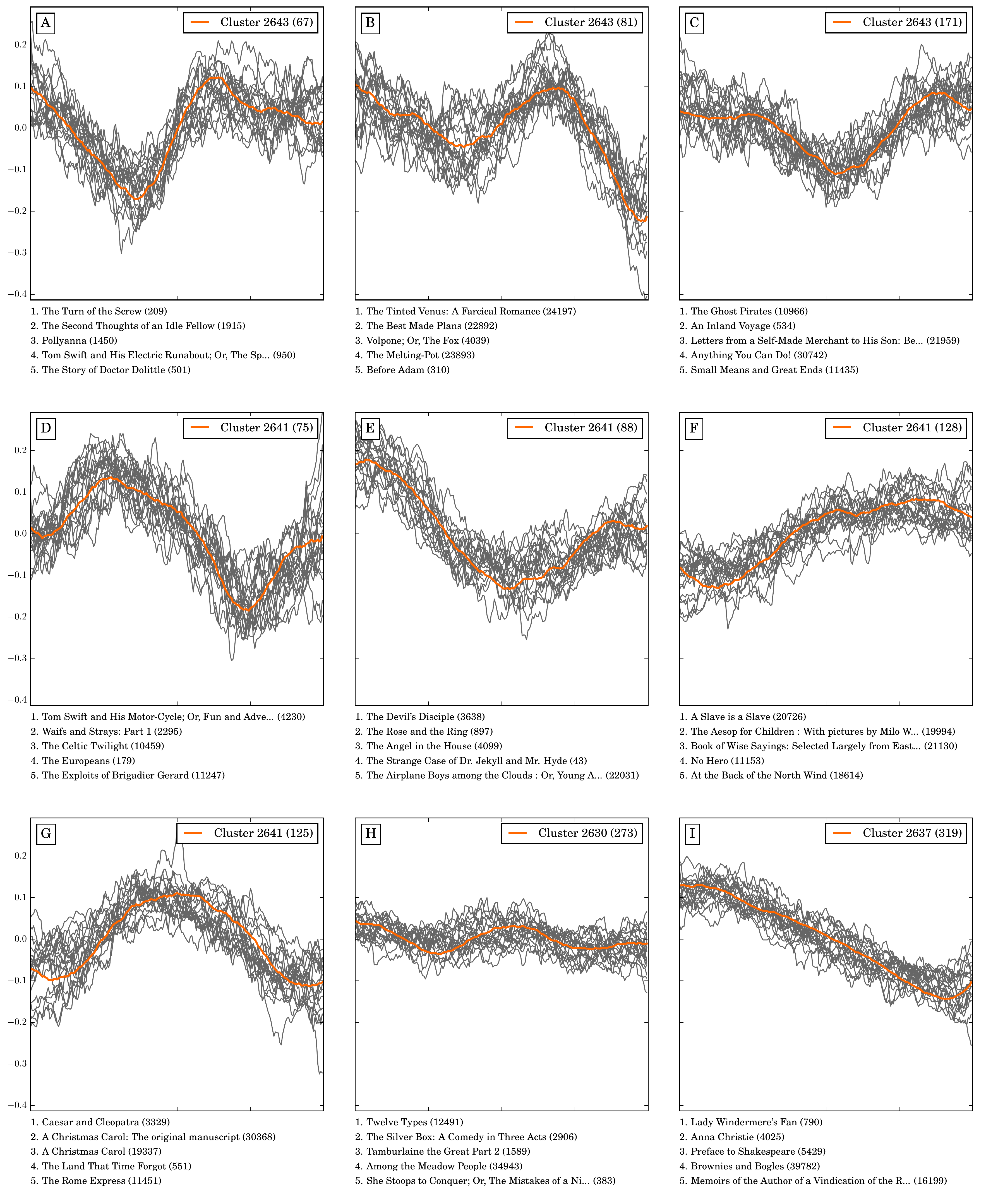}
  \caption[]{
    The 9 clusters identified by Agglomerative Clustering using Ward's method.
  }
    \label{fig:ward-cluster-9}
\end{figure*}

\begin{figure*}[!htb]
  \centering
\includegraphics[width=0.9\textwidth]{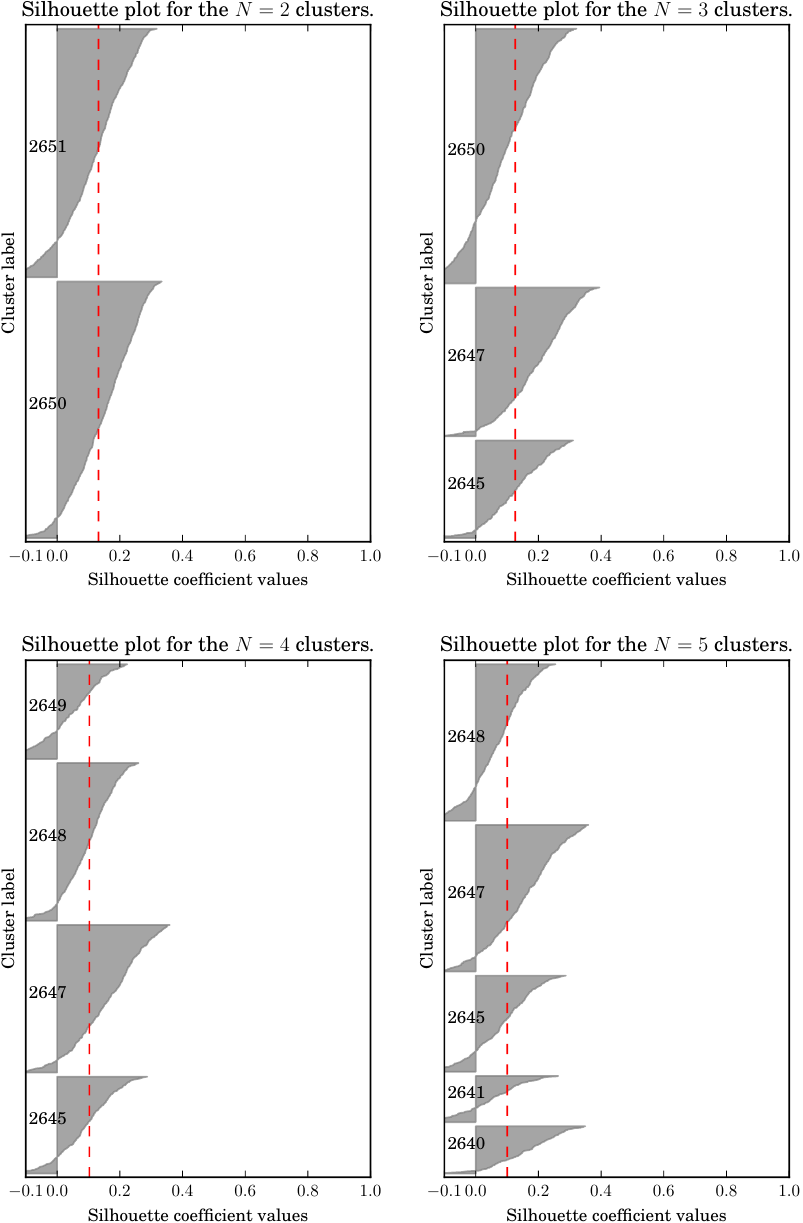}
  \caption[]{
    The silhouette plots for 2--5 clusters identified by Agglomerative Clustering using Ward's method.
  }
  \label{fig:clustering-2-5-clusters}
\end{figure*}

\begin{figure*}[!htb]
  \centering
\includegraphics[width=0.9\textwidth]{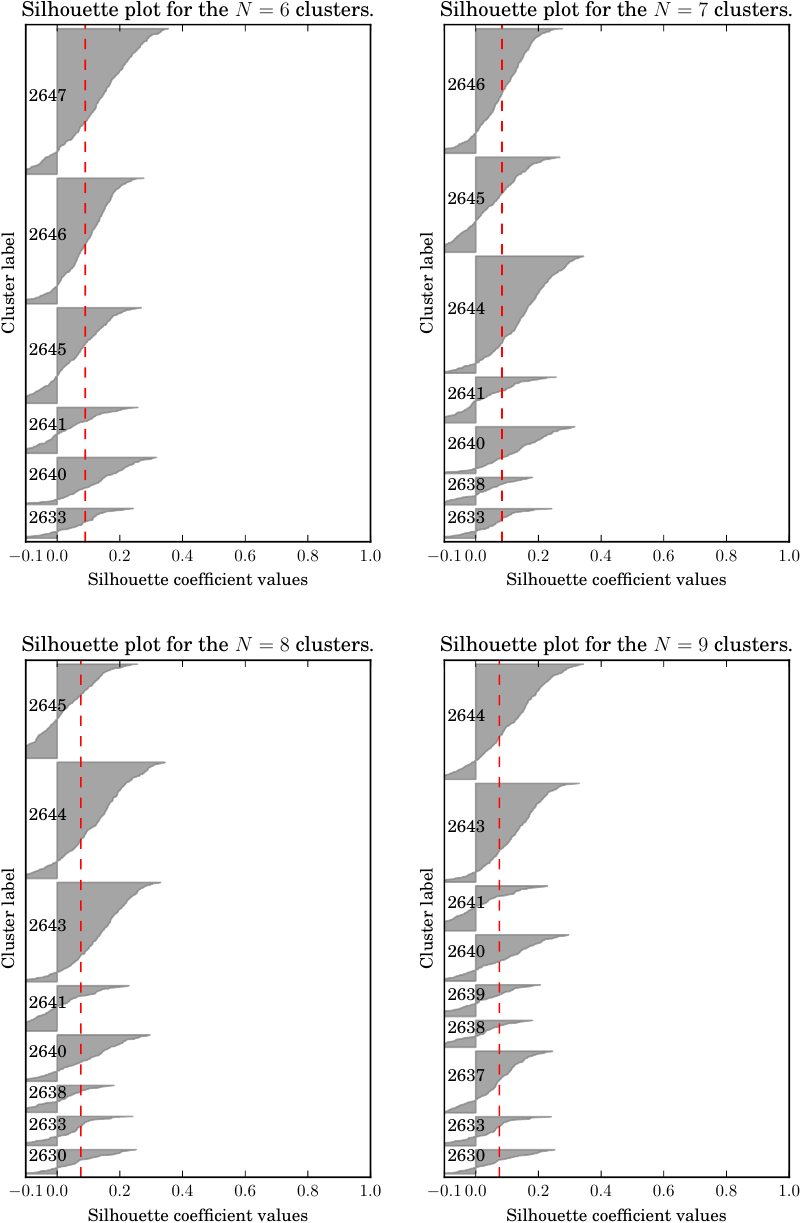}
  \caption[]{
    The silhouette plots for 6--9 clusters identified by Agglomerative Clustering using Ward's method.
  }
  \label{fig:clustering-6-9-clusters}
\end{figure*}

\clearpage
\pagebreak
\section{Additional SOM Figures}
\label{sec:SOM-supp}

\begin{figure}[ht]
  \centering
\includegraphics[width=0.78\textwidth]{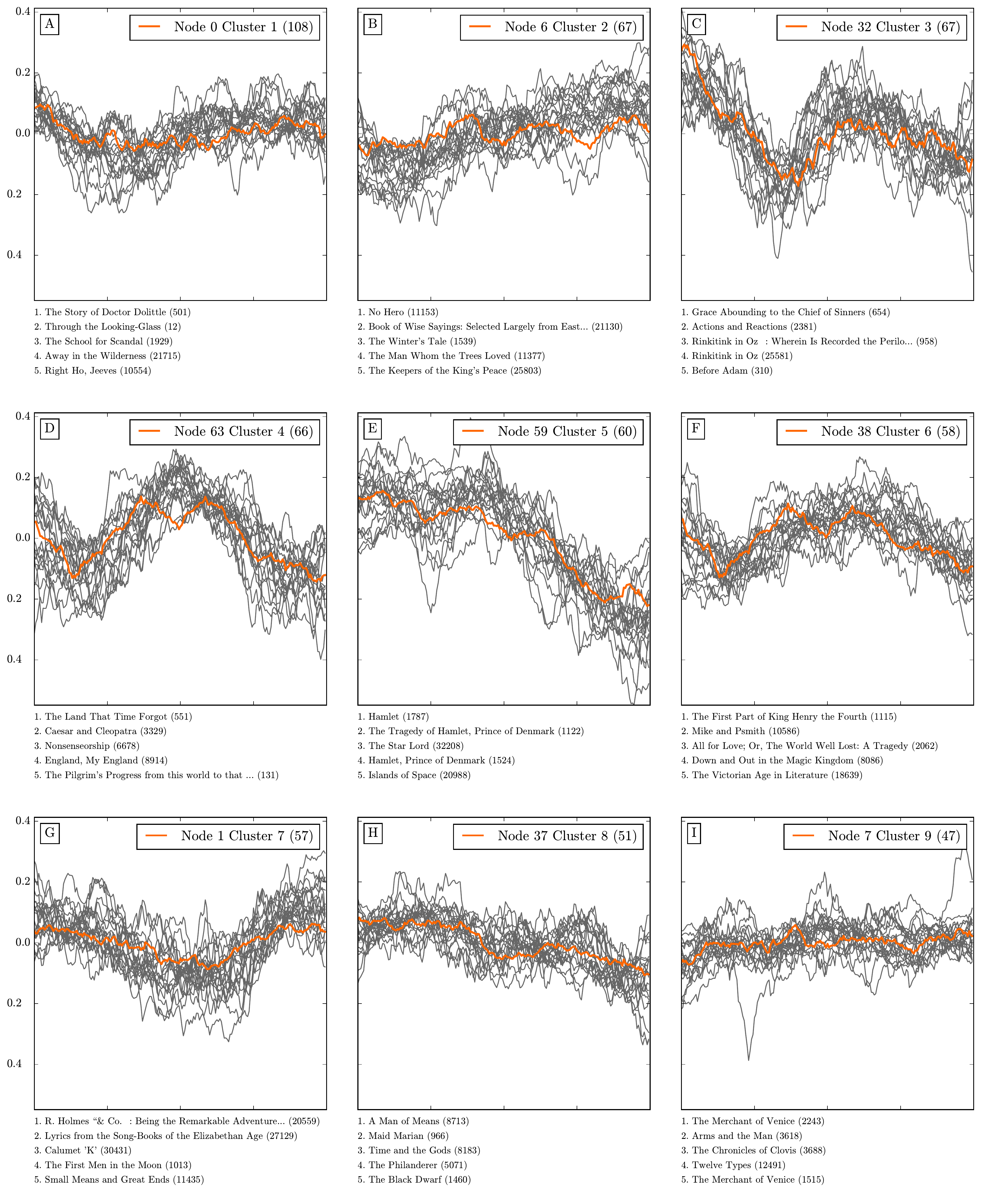}
  \caption[]{
    The vector for each of the top 9 SOM nodes, accompanied with those sentiment time series which are closest to that node.
    The core stories which we have found with other methods are readily visible.
  }
  \label{fig:SOM-stories}
\end{figure}

In Fig.~\ref{fig:SOM-stories} we show the emotional arcs that are closest to each of 9 most frequently winning nodes in the winner-take-all implementation the Self Organizing Map.

\clearpage
\pagebreak
\section{Null comparison details}
\label{sec:shuffled}

\begin{figure}[!htb]
  \centering
\includegraphics[width=0.58\textwidth]{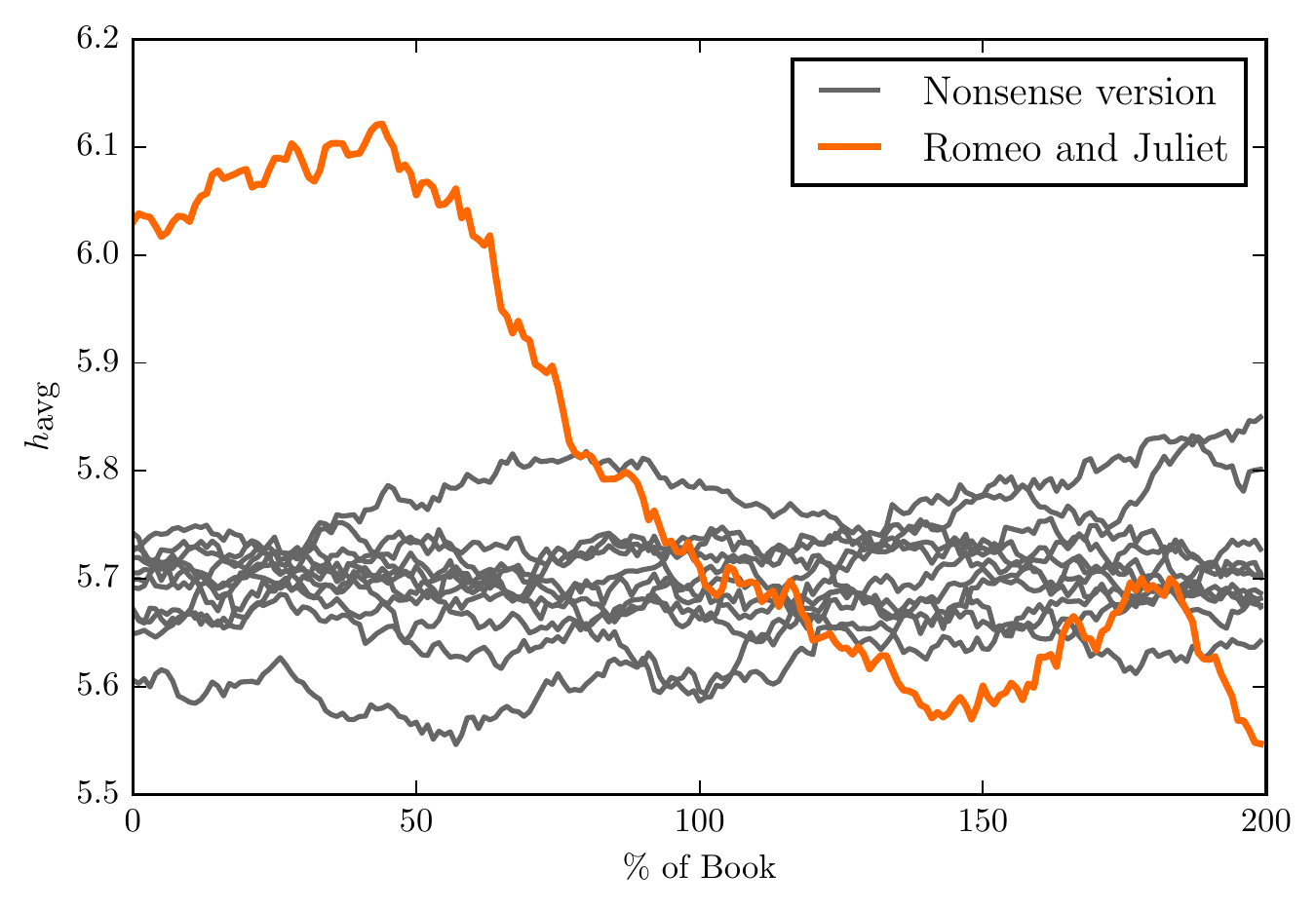}
  \caption{The emotional arc of \textit{Romeo And Juliet} by William Shakespeare (Gutenberg ID 1777), along with 11 ``nonsense'' versions, as produced by a 2-gram Markov model.
    We see that the emotional arc from the true version has more structure than the nonsense versions.}
  \label{fig:salad-1}
\end{figure}

\begin{figure}[!htb]
  \centering
\includegraphics[width=0.58\textwidth]{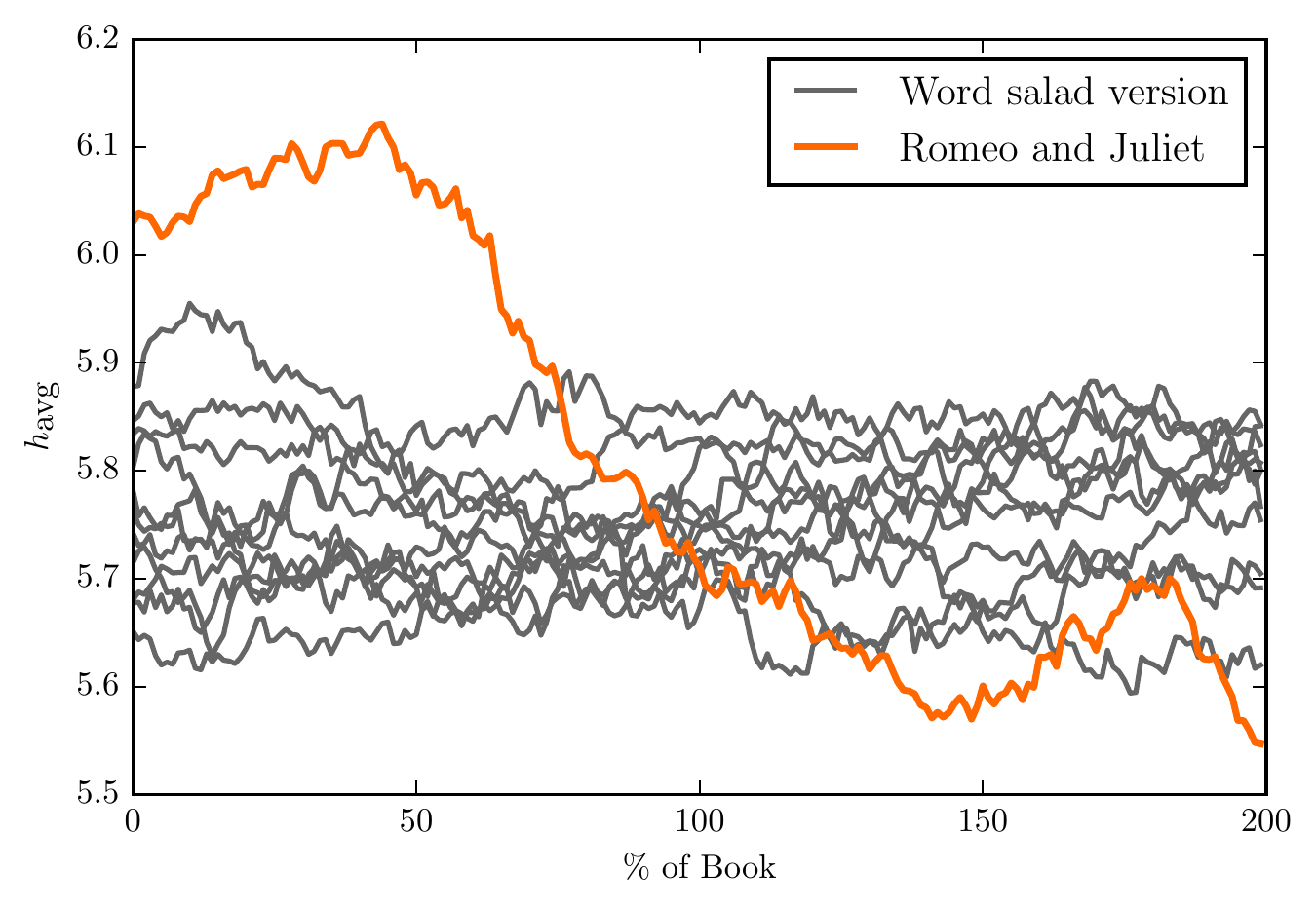}
  \caption{The emotional arc of \textit{Romeo And Juliet} by William Shakespeare, along with 11 ``word salad'' versions, as produced by randomly shuffling the words in the book.
    We see that the emotional arc from the true version has more structure than the word salad versions as well.}
  \label{fig:salad-2}
\end{figure}

An example of the ``nonsense'' and ``word salad'' text is presented first in Appendix \ref{sec:construction}.
First, we examine the resulting timeseries for an example book in Figs. \ref{fig:salad-1} and \ref{fig:salad-2}.
We then go on to present the full result of the SVD, agglomerative clustering, and SOM to ``nonsense'' English fiction books with more than 40 downloads.

\clearpage
\pagebreak
\subsection{Null SVD}

SVD modes from the emotional arcs of word salad books.
We observe higher frequency modes appearing more quickly, and a more even spread of mode coefficients.

\begin{figure*}[!htb]
  \centering
\includegraphics[width=0.9\textwidth]{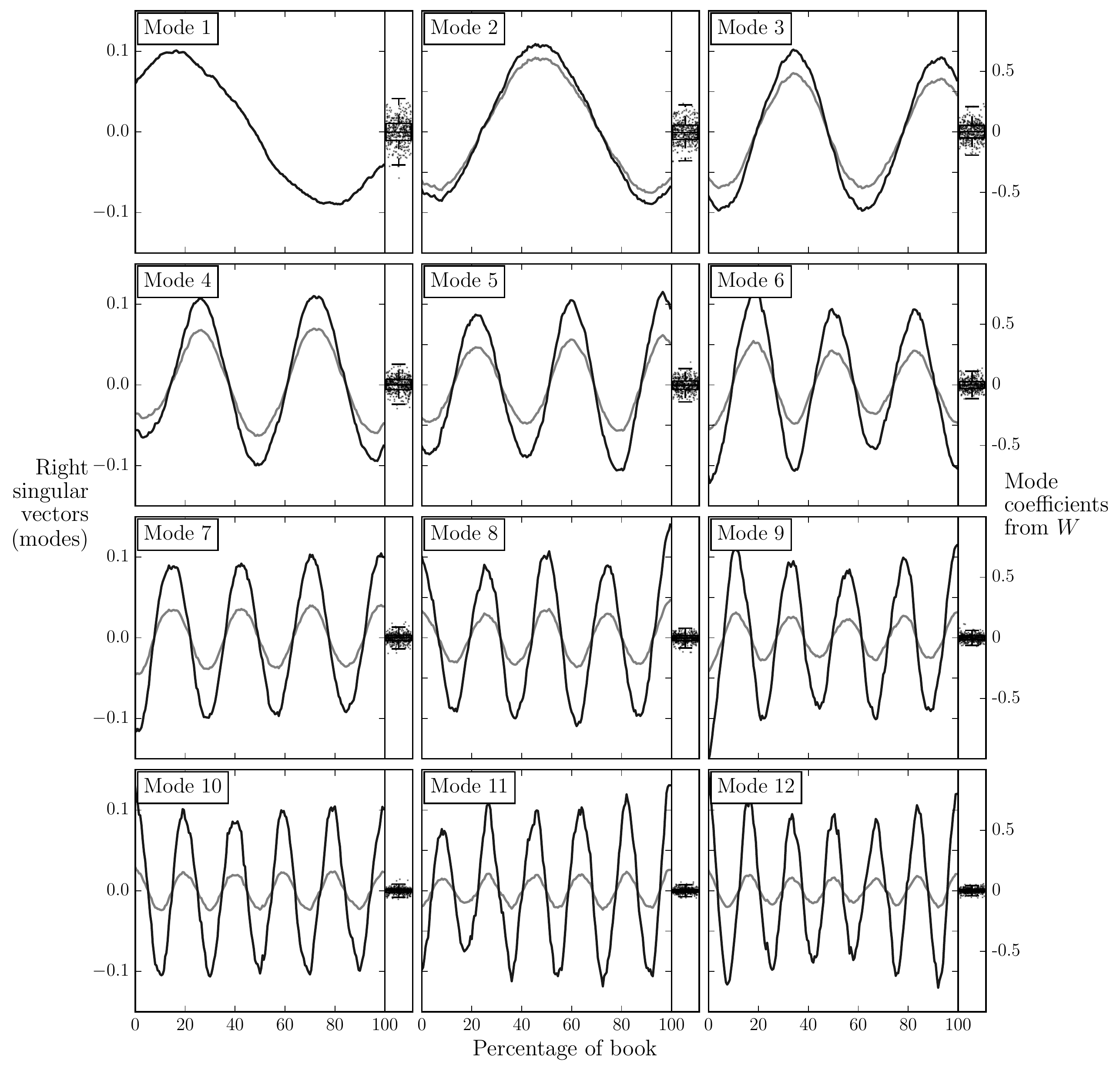}
  \caption[]{
    Top 12 modes from the Singular Value Decomposition of \nbooks~nonsense Project Gutenberg books.
    We show in a lighter color modes weighted by their corresponding singular value, where we have scaled the matrix $\Sigma$ such that the first entry is 1 for comparison.
    The mode coefficients normalized for each book are shown in the right panel accompanying each mode, in the range -1 to 1, with the ``Tukey'' box plot.
  }
  \label{fig:SVD-12-comp-salad}
\end{figure*}

\begin{figure*}[!htb]
  \centering
\includegraphics[width=.98\textwidth]{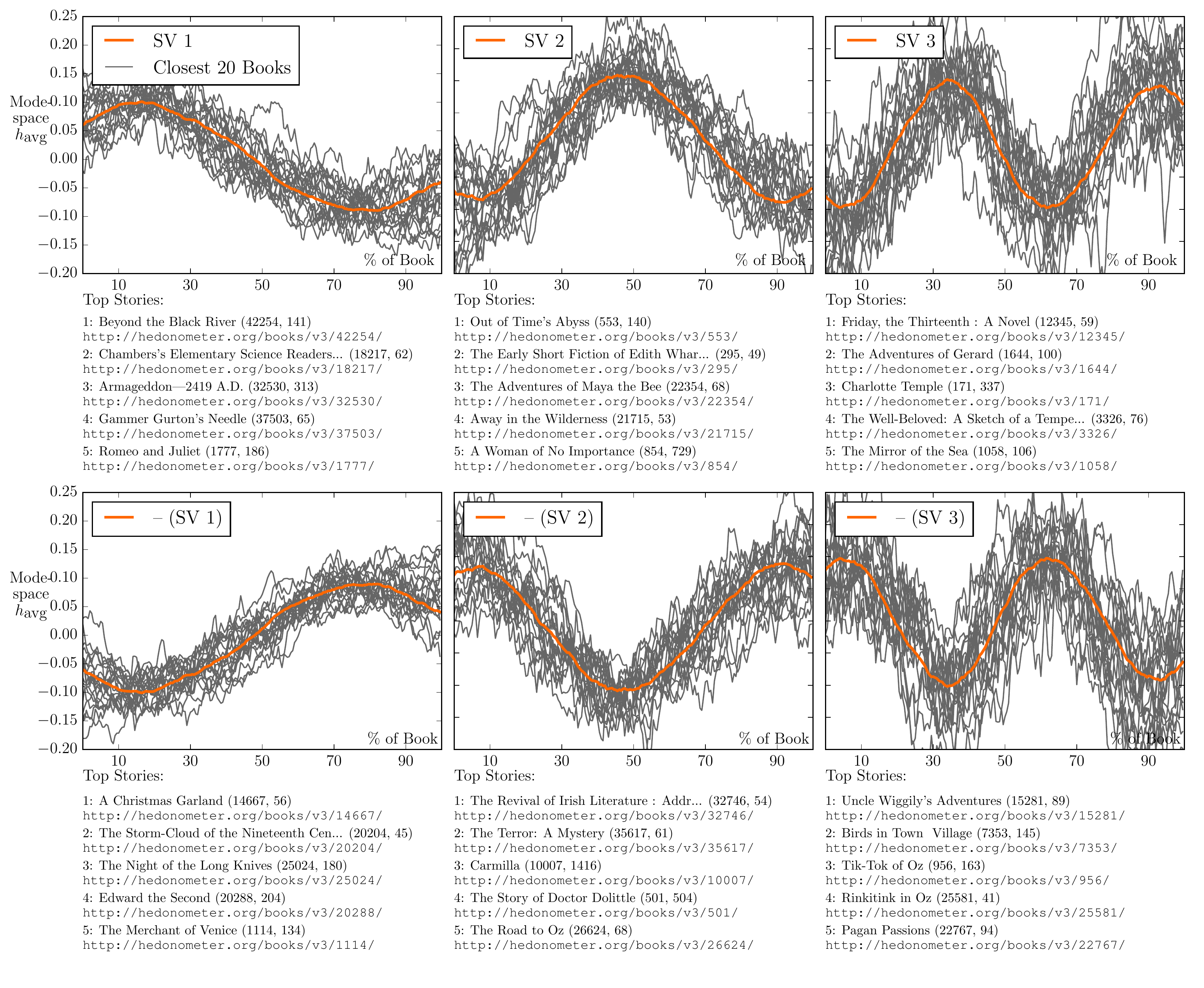}
  \caption[]{
    First 3 SVD modes from nonsense books and their negation with the closest stories to each.
    Links below each story point to an interactive visualization on \href{http://hedonometer.org}{http://hedonometer.org} which enables detailed exploration of the emotional arc for the story.
  }
  \label{fig:SVD-1-3-labelled-salad}
\end{figure*}

\begin{figure*}[!htb]
  \centering
\includegraphics[width=.98\textwidth]{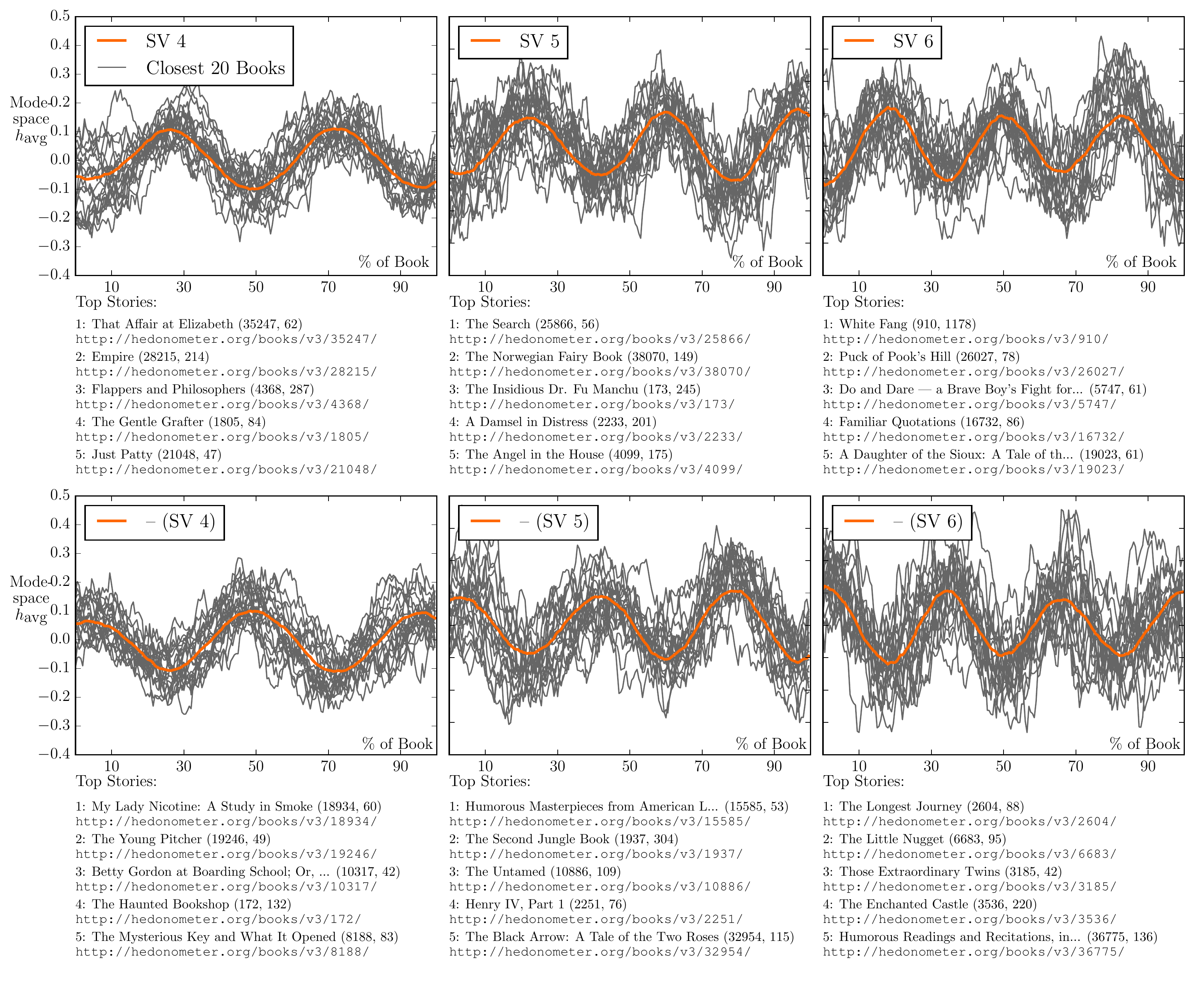}
  \caption[]{
    Modes 4--6 from the SVD analysis of nonsense books and their negation with the closest stories to each.
    Links below each story point to an interactive visualization on \href{http://hedonometer.org}{http://hedonometer.org} which enables detailed exploration of the emotional arc for the story.
  }
  \label{fig:SVD-4-6-labelled-salad}
\end{figure*}

\begin{figure*}[!htb]
  \centering
\includegraphics[width=.6\textwidth]{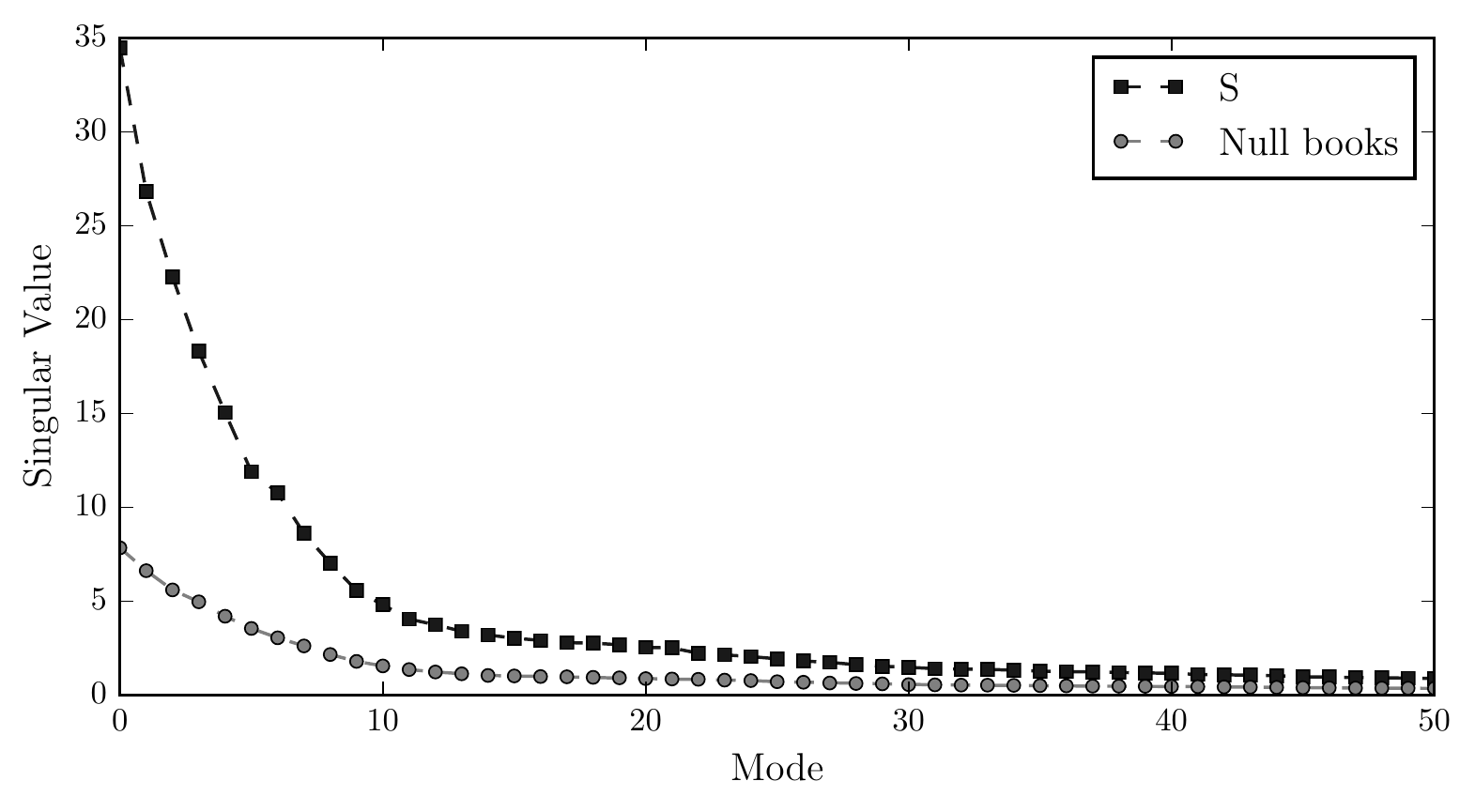}
  \caption[]{
    Comparison of the singular value spectra from the emotional arcs of nonsense books and the emotional arcs of individual Project Gutenberg books.
    The spectra from the nonsense books is muted, indicating both lower total variance explained and less important ordering of the singular vectors.}
  \label{fig:SVD-spectrum-comparison}
\end{figure*}

\clearpage
\pagebreak
\subsection{Null Hierarchical Clustering}

Dendrogram of clustering using Ward's method on the emotional arcs of word salad books.
We observe comparatively low linkage cost for these emotional arcs, indicating the absence of distinct clusters.

\begin{figure*}[!htb]
  \centering
\includegraphics[width=0.98\textwidth]{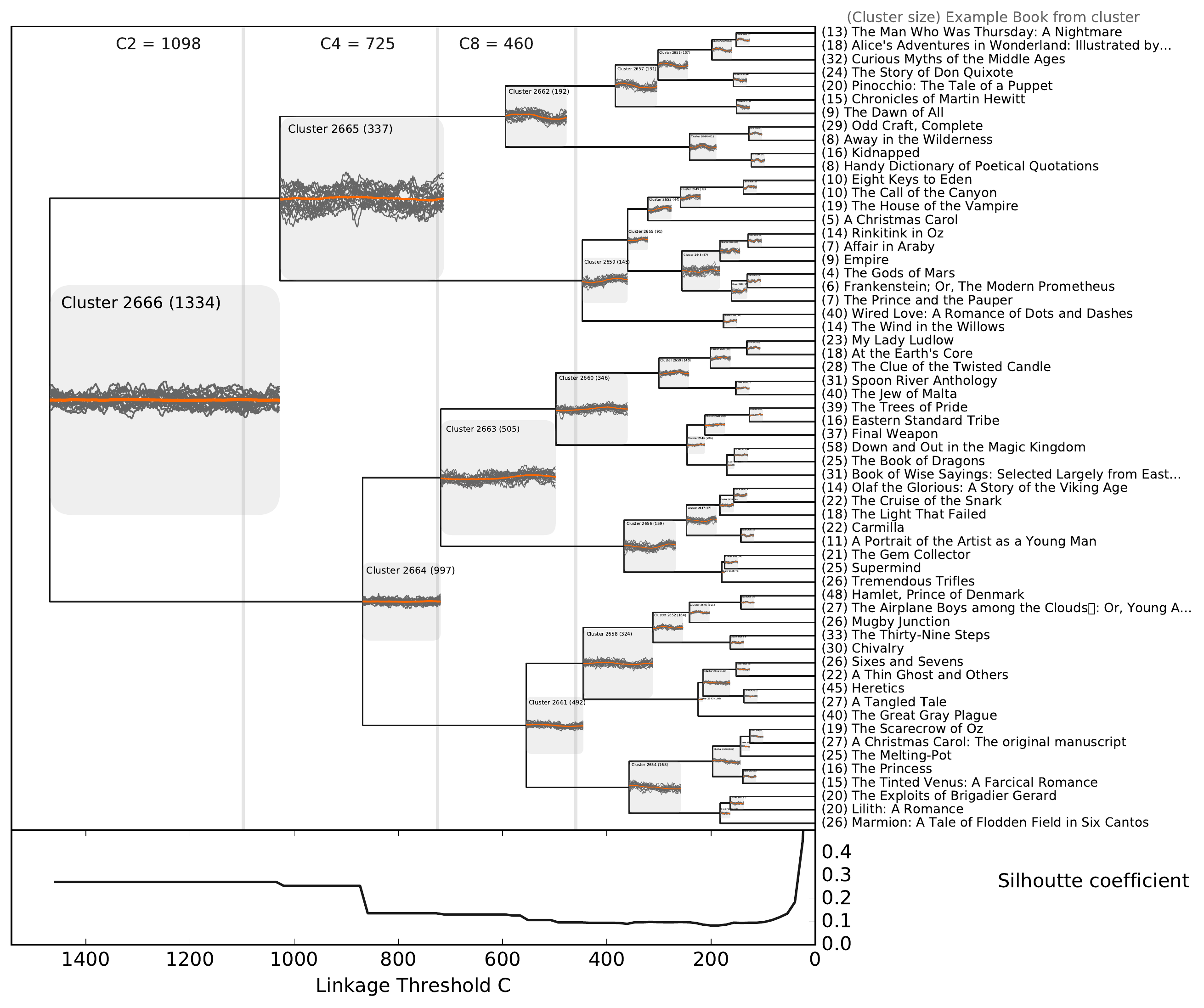}
  \caption[]{
    Dendrogram from the agglomerative clustering procedure using Ward's minimum variance method on nonsense books.
    For each cluster, a selection of the 20 most central books to a fully-connected network of books are shown along with the average of the emotional arc for all books in the cluster, along with the cluster ID and number of books in each cluster (shown in parenthesis).
    At the bottom, we show the average Silhouette value for all books, with higher value representing a more appropriate number of clusters.
    For each of the 60 leaf nodes (right side) we show the number of books within the cluster and the most central book to that cluster's book network.
  }
  \label{fig:ward-small-salad}
\end{figure*}

\begin{figure*}[!htb]
  \centering
\includegraphics[width=0.98\textwidth]{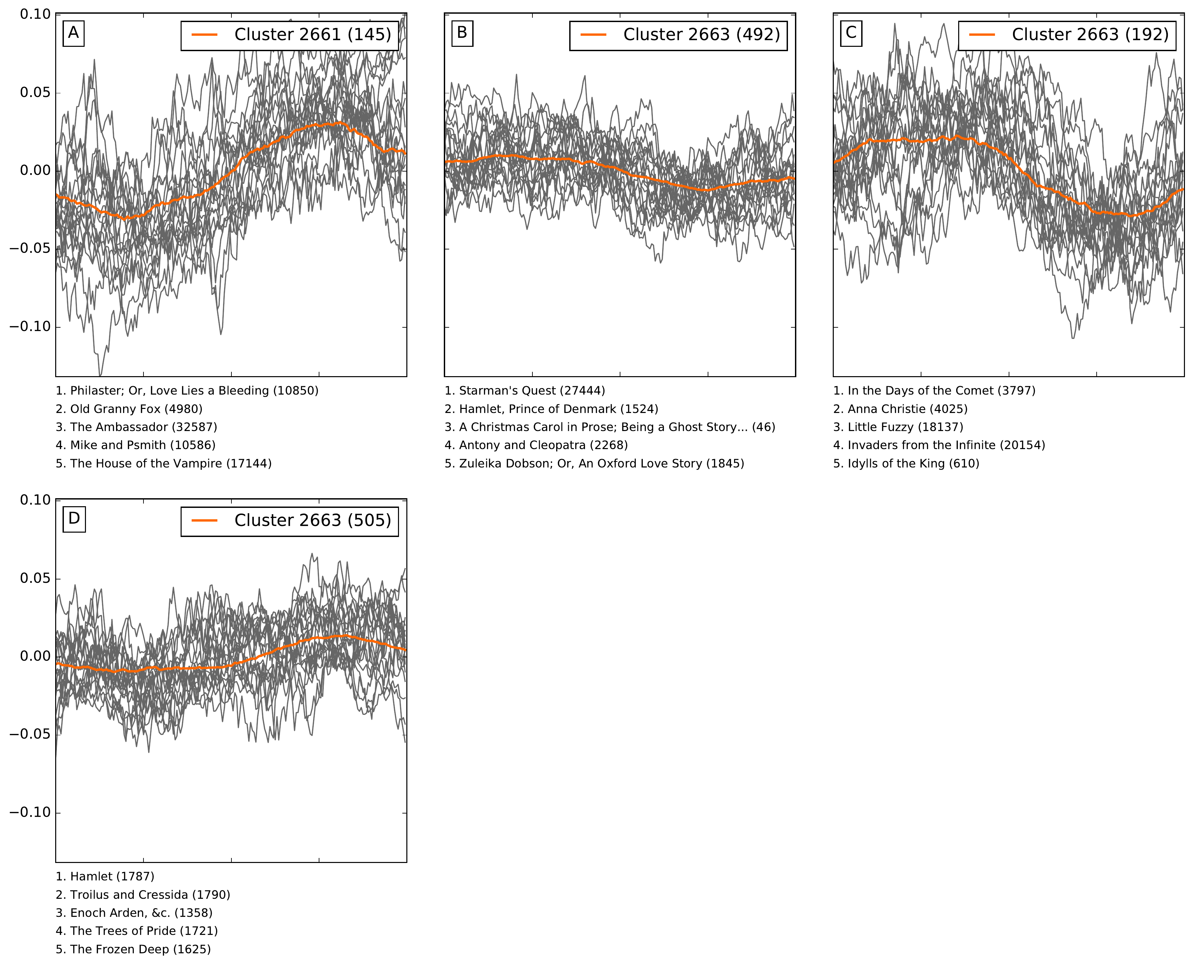}
  \caption[]{
    Four clusters (linkage threshold 850) from the hierarchical clustering of word salad books.
    We observe that the cluster mean emotional arc and the most central emotional arcs have high variance, without a visible signal.}
  \label{fig:ward-4-salad}
\end{figure*}

\clearpage
\pagebreak
\subsection{Null Self Organizing Map (SOM)}

\begin{figure*}[!htb]
  \centering
\includegraphics[width=0.98\textwidth]{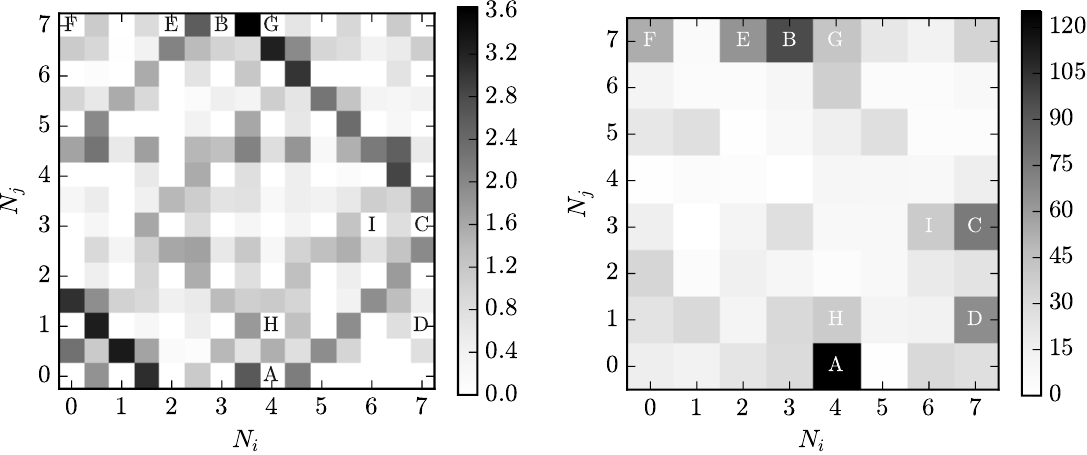}
  \caption[]{
    Results of the SOM applied to nonsense versions of Project Gutenberg books.
    Left panel: Nodes on the 2D SOM grid are shaded by the number of stories for which they are the winner.
    Right panel: The B-Matrix shows that there are clear clusters of stories in the 2D space imposed by the SOM network.
  }
  \label{fig:SOM-matrices-salad}
\end{figure*}

\begin{figure}[!htb]
  \centering
\includegraphics[width=0.78\textwidth]{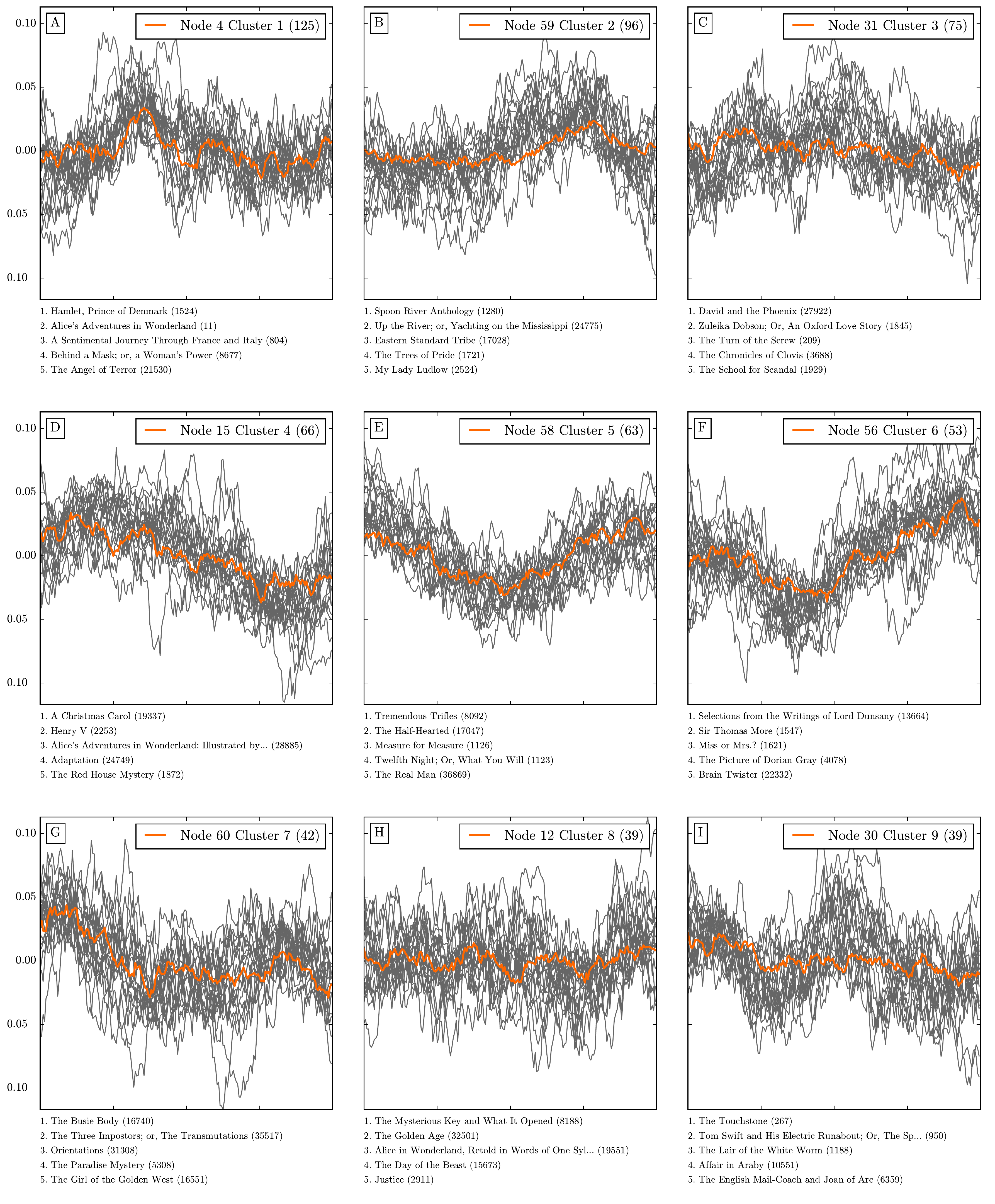}
  \caption[]{
    The vector for each of the top 9 SOM nodes for null emotional arcs, accompanied with those sentiment time series which are closest to that node.
    Panels D and E show what appear to be similar arcs to the six we identified in real books, but overall see that the emotional arcs from null arcs show little coherent structure, especially considering the y-range here being 0.1 compared to the 0.4 of the real books (had we used the same y-range, very little of the variation would be visible at all).
  }
  \label{fig:SOM-stories-salad}
\end{figure}

\clearpage
\pagebreak

\chapter{labMTsimple: A Python Library for Sentiment Analysis}

\section{Getting Started}
\label{getting-started:welcome-to-labmt-simple-s-documentation}\label{getting-started:getting-started}\label{getting-started::doc}

In this chapter, we provide details for a Python package called \verb|labMTsimple|.
The package exposes a simple, but quickly growing, labMT usage library.

\subsection{Usage}
\label{getting-started:usage}
This package uses the language assessment by Mechanical Turk (labMT) word list to score the happiness of a corpus. The labMT word list was created by combining the 5000 words most frequently appearing in four sources: Twitter, the New York Times, Google Books, and music lyrics, and then scoring the words for sentiment on Amazon's Mechanical Turk. The list is described in detail in the publication Dodds' et al. 2011, PLOS ONE, ``Temporal Patterns of Happiness and Information in a Global-Scale Social Network: Hedonometrics and Twitter.''

Given two corpora, the script ``storylab.py'' creates a word-shift graph illustrating the words most responsible for the difference in happiness between the two corpora. The corpora should be large (e.g. at least 10,000 words) in order for the difference to be meaningful, as this is a bag-of-words approach. As an example, a random collection of English tweets from both Saturday January 18 2014 and Tuesday January 21 2014 are included in the ``example'' directory. They can be compared by moving to the test directory, using the command
\begin{lstlisting}
python example.py example-shift.html
\end{lstlisting}

and opening the file \lstinline{example-shift.html} in a web browser. For an explanation of the resulting plot, please visit

\href{http://www.hedonometer.org/shifts.html}{http://www.hedonometer.org/shifts.html}

\subsection{Installation}
\label{getting-started:installation}
Cloning the github directly is recommended, and then installing locally:
\begin{lstlisting}
git clone https://github.com/andyreagan/labMT-simple.git
cd labMT-simple
python setup.py install
\end{lstlisting}

This repository can also be installed using pip
\begin{lstlisting}
pip install labMTsimple
\end{lstlisting}

in which case you can download the tests from github and run them, if desired.

\subsection{Running tests}
\label{getting-started:running-tests}
Tests are based on nose2, \lstinline{pip install nose2}, and can be run inside the by executing
\begin{lstlisting}
nose2
\end{lstlisting}

in the root directory of this repository.

This will compare the two days in test/data and print test.html which shifts them, allowing for a changable lens.

\subsection{Developing with labMT-simple locally}
\label{getting-started:developing-with-labmt-simple-locally}
It is often useful to reload the library when testing it interactively:
\begin{lstlisting}
try:
    reload
except NameError:
    \# Python 3
    from importlib import reload
\end{lstlisting}

\subsection{Building these docs}
\label{getting-started:building-these-docs}
Go into the docs directory (activate local \lstinline{virtualenv} first), and do the following:
\begin{lstlisting}
\rm -rf _build/*
make html
make latexpdf
git add -f *
git commit -am ``new docs, probably should just add a pre-commit hook''
\end{lstlisting}

Note that these docs will build locally in python 2 because the dependencies exist.
With python 3 available, these dependencies will be mocked (and this is set for the online readthedocs site).

(\lstinline{sphinx-apidoc -o . ../labMTsimple} was run once.)

\section{Detailed Examples}
\label{detailed-example::doc}\label{detailed-example:detailed-examples}

\subsection{Preparing texts}
\label{detailed-example:preparing-texts}
This is simple really: just load the text to be scored into python.
This is using a subset of a couple days of public tweets to text, and they have already put the tweet text into \lstinline{.txt} files that are loaded into strings:
\begin{lstlisting}
f = codecs.open(``data/18.01.14.txt'',''r'',''utf8'')
saturday = f.read()
f.close()

f = codecs.open(``data/21.01.14.txt'',''r'',''utf8'')
tuesday = f.read()
f.close()
\end{lstlisting}

\subsection{Loading dictionaries}
\label{detailed-example:loading-dictionaries}
Again this is really simple, just use the \lstinline{emotionFileReader} function:
\begin{lstlisting}
lang = `english'
labMT,labMTvector,labMTwordList = emotionFileReader(stopval=0.0,lang=lang,returnVector=True)
\end{lstlisting}

Then we can score the text and get the word vector at the same time:
\begin{lstlisting}
saturdayValence,saturdayFvec = emotion(saturday,labMT,shift=True,happsList=labMTvector)
tuesdayValence,tuesdayFvec = emotion(tuesday,labMT,shift=True,happsList=labMTvector)
\end{lstlisting}

But we don't want to use these happiness scores yet, because they included all words (including neutral words).
So, set all of the neutral words to 0, and generate the scores:
\begin{lstlisting}
tuesdayStoppedVec = stopper(tuesdayFvec,labMTvector,labMTwordList,stopVal=1.0)
saturdayStoppedVec = stopper(saturdayFvec,labMTvector,labMTwordList,stopVal=1.0)

saturdayValence = emotionV(saturdayStoppedVec,labMTvector)
tuesdayValence = emotionV(tuesdayStoppedVec,labMTvector)
\end{lstlisting}

\section{Making Wordshifts}
\label{wordshifts::doc}\label{wordshifts:making-wordshifts}
With merged updates to the d3 wordshift plotting in labMTsimple, and combined with phantom crowbar (see previous post), it's easier than ever to use the labMT data set to compare texts.

To make an html page with the shift, you'll just need to have labMT-simple installed.
To automate the process into generating svg files, you'll need the phantom crowbar, which depends on phantomjs.
To go all the way to pdf, you'll also need inkscape for making vectorized pdfs, or rsvg for making better formatted, but rasterized, versions.

Let's get set up to make shifts automatically.
Since they're aren't many dependencies all the way down, start by getting phantomjs installed, then the phantom-crowbar.

\subsection{Installing phantom-crowbar}
\label{wordshifts:installing-phantom-crowbar}
For the phantomjs, use homebrew:
\begin{lstlisting}
brew update
brew upgrade
brew install phantomjs
\end{lstlisting}

Then to get the crowbar, clone the git repository.
\begin{lstlisting}
cd \textasciitilde{}
git clone https://github.com/andyreagan/phantom-crowbar
\end{lstlisting}

To use it system-wide, use the bash alias:
\begin{lstlisting}
alias phantom-crowbar=''/usr/local/bin/phantomjs \textasciitilde{}/phantom-crowbar/phantom-crowbar.js''
\end{lstlisting}

Without too much detail, add this to your \lstinline{~/.bash_profile} so that it's loaded every time you start a terminal session.

\subsection{Installing inkscape}
\label{wordshifts:installing-inkscape}
You only need inkscape if you want to go from svg to pdf (and there are other ways too), but this one is easy with, again, homebrew.
\begin{lstlisting}
brew install inkscape
\end{lstlisting}

\subsection{Installing rsvg}
\label{wordshifts:installing-rsvg}
You only need inkscape if you want to go from svg to pdf (and there are other ways too), but this one is easy with, again, homebrew.
\begin{lstlisting}
brew install librsvg
\end{lstlisting}

\subsection{Installing labMTsimple}
\label{wordshifts:installing-labmtsimple}
There are two ways to get it: using pip of cloning the git repo.
If you're not sure, use pip.
Pip makes it easier to keep it up to date, etc.
\begin{lstlisting}
pip install labMTsimple
\end{lstlisting}

\subsection{Making your first shift}
\label{wordshifts:making-your-first-shift}
If you cloned the git repository, install the thing and then you can check out the example in \lstinline{examples/example.py}.
If you went with pip, see that file on \href{https://github.com/andyreagan/labMT-simple/blob/master/examples/example.py}{github}.

Go ahead and run that script!
\begin{lstlisting}
python example-002.py
\end{lstlisting}

You can open the html file to see the shift in any browser, with your choice of local webserver.
Python's SimpleHTTPServer works fine, and generally the node based http-server is a bit more stable.

To take out the svg, go ahead and use the \lstinline{phantom-crowbar.js} file copied to the \lstinline{example/static} directory.
Running it looks like this, for me:
\begin{lstlisting}
/usr/local/bin/phantomjs js/shift-crowbar.js example-002.html shiftsvg wordshift.svg
\end{lstlisting}

Using inkscape or librsvg on my computer look like this:
\begin{lstlisting}
  /Applications/Inkscape.app/Contents/Resources/bin/inkscape \
  -f \$(pwd)/wordshift.svg \
  -A \$(pwd)/wordshift-inkscape.pdf

rsvg-convert --format=eps worshift.svg \textgreater{} wordshift-rsvg.eps
epstopdf wordshift-rsvg.eps
\end{lstlisting}

And again, feel free to tweet suggestions at \href{https://twitter.com/andyreagan}{@andyreagan}, and submit pull requests to the \href{https://github.com/andyreagan/labMT-simple}{source code}!

\subsection{Full Automation}
\label{wordshifts:full-automation}
This procedure wraps up what is potentially the most backwards way to generate figure imaginable.
SThe \lstinline{shiftPDF()} function operates the same way as the \lstinline{shiftHTML()}, but uses the headless web server to render the d3 graphic, then exectues a piece of injected JS to save a local SVG, and uses command line image manipulation libraries to massage it into a PDF.

On my macbook, this works, but your mileage will most certainly vary.

\section{Advanced Usage}
\label{advanced::doc}\label{advanced:advanced-usage}

\subsection{About Tries}
\label{advanced:about-tries}
For dictionary lookup of scores from phrases, the fastest benchmarks available and that were reasonable stable were from the libraries \lstinline{datrie} and \lstinline{marisatrie} which both have python bindings.

They're used in the \lstinline{speedy} module in an attempt to both speed things up, and match against word stems.

\subsection{Advanced Parsing}
\label{advanced:advanced-parsing}
Some dictionaries use word stems to cover the multiple uses of a single word, with a single score.
We can very quickly match these word stems using a prefix match on a trie.
This is much better than using many compiled RE matches, which in my testing took a very long time.

\clearpage
\pagebreak

\chapter{Code for VACC Twitter Database Keyword Searches}

In this Appendix we describe a strategy for utilizing the computational resources available at the University of Vermont's supercomputing center for searching through Twitter data.
A schematic of the general approach is provided below in Figure~\ref{fig:keyword-search}.
We provide scripts for a minimum working example of this approach online at \href{https://github.com/andyreagan/VACC-keyword-search}{https://github.com/andyreagan/VACC-keyword-search}.

The basic approach is to use the cron scheduler to make sure that around 100--150 jobs are running all the time.
Each job is short, processing at most an hour of Tweets, so that each job takes less time to run and can utilize the shortq, which has a limit of 200 jobs that run immediately under most circumstances.

Cron calls the shell script \verb|cron.sh| directly, and that shell script invokes the Python script \verb|qsub.py| to handle the more complex logic of dates and job submission (which is just easier in Python).
The work of looking through Tweets happens in \verb|processTweets.py|.
By utilizing pipes and unzipping from disk directly into Python, no unzipped files are written to disk (this would be prohibitively slow and use too much storage).
Instead, just the minimum necessary output from the search is written to disk, by Python.

To make this run for new keywords, do the following:
\begin{enumerate}
\item Edit the keywords in \verb|processTweets.py|.
\item Set the start date in \verb|currdate.txt|.
\item Create folders for the keywords (manually, or using \verb|makeFolders| function from the \verb|processTweets.py| script). Including the base \verb|raw-tweets| folder.
\item Instantiate a virtual environment. It is called set it up in the job script submitted by \verb|qsub.py|, and in the \verb|cron.sh| to call \verb|qsub.py|. (Or don't use one, at your own peril!).
\item Edit the folders in \verb|cron.sh| and \verb|qsub.py| to where this is running.
\item Add the script \verb|cron.sh| to your crontab.
\item Profit.
\end{enumerate}

Using the commands \verb|mmlsquota| and \verb|showq|, you can see your file system usage and track the individual jobs (look at \verb|currdate.txt| for submitted progress).

\begin{figure}[ht!]
  \centering
  \includegraphics[width=0.95\textwidth]{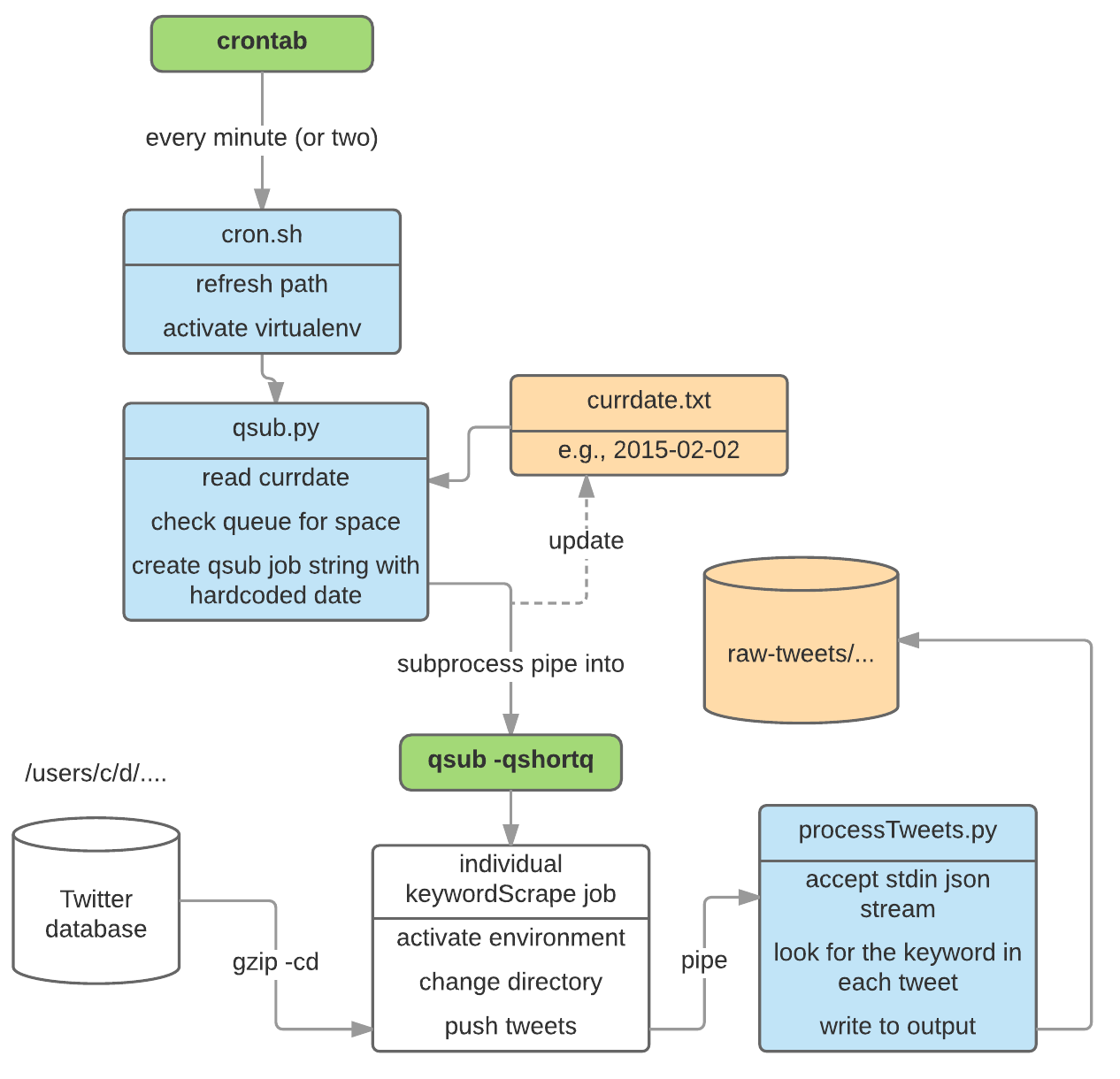}
  \caption[]{
    Schematic of the keyword search framework on the Vermont Advanced Compute Cluster.
    The three core Python files are invoked by the cron scheduler, and the computation is handed off to compute nodes through the PBS schedulers qsub command.
    An example code base is provided at \href{https://github.com/andyreagan/VACC-keyword-search}{https://github.com/andyreagan/VACC-keyword-search}.
  }
  \label{fig:keyword-search}
\end{figure}

\chapter{Infrastructure of Hedonometer.org}

\begin{figure}[ht!]
  \centering
  \includegraphics[width=0.95\textwidth]{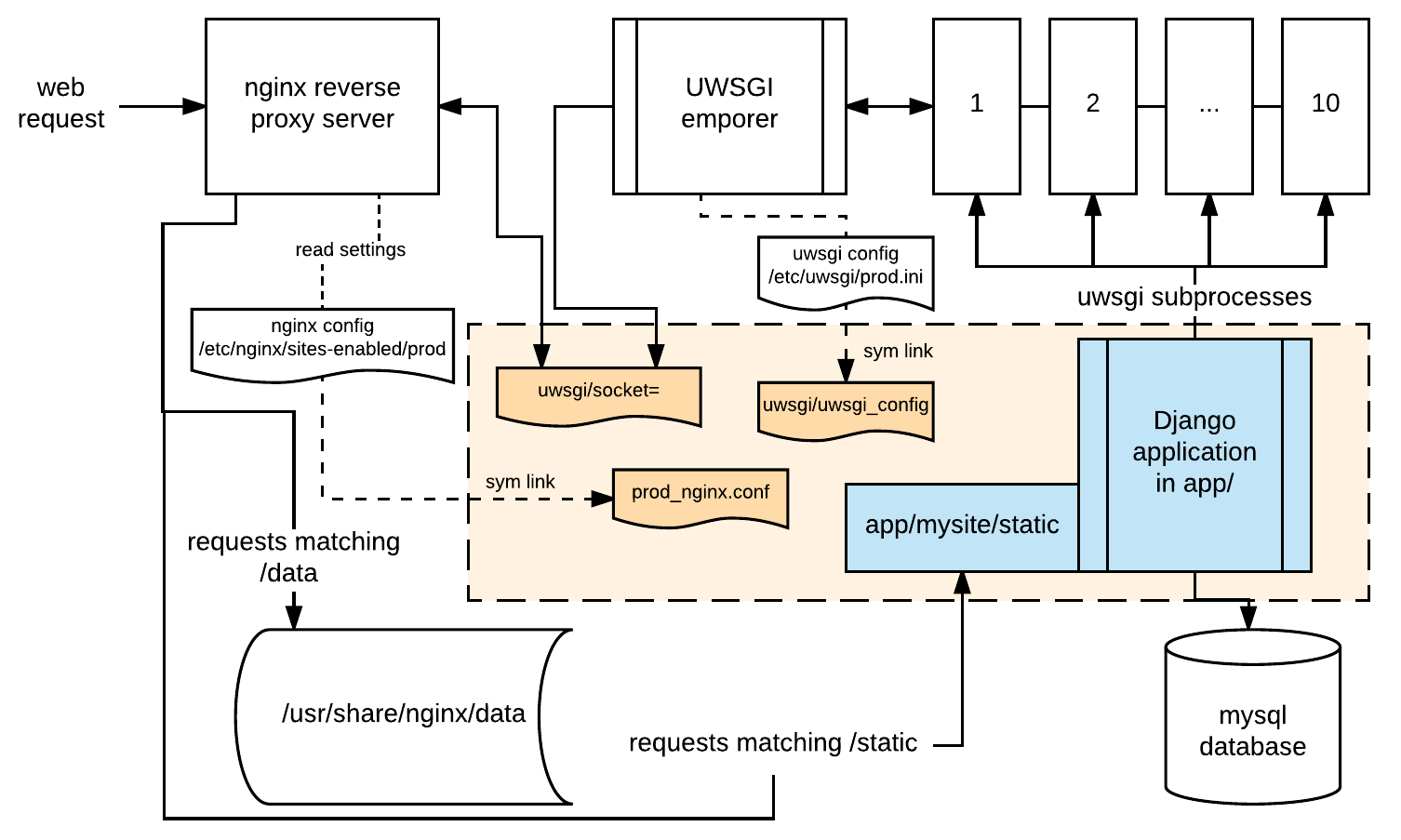}
  \caption[]{
    Schematic of the Hedonometer server architecture.
    The section in orange is contained in the prod user account, and includes the code stored on GitHub at \href{https://github.com/andyreagan/hedonometer}{https://github.com/andyreagan/hedonometer}.
    The settings files for UWSGI and Nginx are written by an Ansible playbook based on the user account under which the code is distributed.
  }
  \label{fig:hedonometer-org}
\end{figure}

The Hedonometer website at \href{http://hedonometer.org}{http://hedonometer.org} is comprised of three main parts: (1) the web server processing including the base code in Python Django, (2) the data processing on the server, and (3) data processing on the VACC.
The deployment of the webserver is done using templates and the Ansible tool.
Settings and detailed instructions for deploying development and production servers are at \href{https://github.com/andyreagan/hedonometer-vagrant-ansible-deployment}{https://github.com/andyreagan/hedonometer-vagrant-ansible-deployment}.
In Figure~\ref{fig:hedonometer-org} we diagram the web server side of the server, included the deployment settings mentioned above and the Django server linked in the caption.

The data side of the server is run separately from the web server side.
Nginx serves all files in the \verb|/data| URL ending at Hedonometer, and the files can be browsed at \href{http://hedonometer.org/data/}{http://hedonometer.org/data/}.
The files here are used in the front end visualizations across the site, and represent files that loaded for the details-on-demand, as well as the overview files.
The structure is optimized for front end performance.
The code base is on GitHub at \href{https://github.com/andyreagan/hedonometer-data-munging}{https://github.com/andyreagan/hedonometer-data-munging}.
As seen in the overall schematic of the server, these files are all inside of \verb|/usr/share/nginx| and they are managed by the \verb|root| user.

Every hour on the hour, these files are updated by a cascade of processes through the cron scheduler.
The process is simple enough to do without a diagram: \verb|cron| calls \verb|cronregions.sh| every hour, which simply calls \verb|regions.py| with Python.
The \verb|regions.py| loops over dates, looks for files in the \verb|word-vectors| folders for each region, and uses \verb|rsync| to copy over the missing files.
Once whole days are downloaded in the 15-minute pieces, it creates the daily summary files (e.g., \verb|word-vectors/vacc/2017-04-07-sum.csv|) and updates the overall summary at \href{http://hedonometer.org/data/word-vectors/vacc/sumhapps.csv}{http://hedonometer.org/data/word-vectors/vacc/sumhapps.csv} (being wary of duplicates, and keeping the most recent).

The copy of files from the VACC uses \verb|rsync|, which operates of \verb|ssh| and relies on the public key of the server being present on the VACC for seamless access.
The files on the VACC are created by a separate process, which is managed in much the same way as the keyword searches in the previous appendix.
The cron scheduler runs every hour on the hour, and submits jobs to the PBS queue that turn 15-minute zipped JSON Tweet files into length 10,222 word vectors.
The full code for this process is available at \href{https://github.com/andyreagan/hedonometer-VACC-processing}{https://github.com/andyreagan/hedonometer-VACC-processing}.

\chapter{Online Code Repositories}

In this Appendix we collect the repositories that store the code used across all of the projects in this dissertation.

From Chapters 2 and 3, we make the code for the papers publicly available in following two repositories :
(1)
\href{https://github.com/andyreagan/sentiment-analysis-comparison}{https://github.com/andyreagan/sentiment-analysis-comparison}
and (2)
\href{https://github.com/andyreagan/core-stories}{https://github.com/andyreagan/core-stories}.
The code for the online Appendix for the sentiment comparison paper is available at \href{https://github.com/andyreagan/sentiment-analysis-comparison-online-appendix}{https://github.com/andyreagan/sentiment-analysis-comparison-online-appendix}, and for the emotional arcs paper at \href{https://github.com/andyreagan/core-stories-online-appendices}{https://github.com/andyreagan/core-stories-online-appendices}.
In addition, the code for the Hedonometer website that hosts the interactive emotional arc visualizations is at \href{https://github.com/andyreagan/hedonometer}{https://github.com/andyreagan/hedonometer}.

From Chapter 4, we provide a link to the repository for each project, in the respective order.

\begin{itemize}
\item Collective Philanthropy: Describing and Modeling the Ecology of Giving --- code at \href{https://github.com/andyreagan/philanthropy-distributions-code}{https://github.com/andyreagan/philanthropy-distributions-code} and online appendices at \href{https://github.com/andyreagan/philanthropy-distributions-online-appendices}{https://github.com/andyreagan/philanthropy-distributions-online-appendices}.
\item Shadow networks: Discovering hidden nodes with models of information flow --- \href{https://github.com/andyreagan/twitter-reply-networks}{https://github.com/andyreagan/twitter-reply-networks}.
\item Human language reveals a universal positivity bias --- \href{https://github.com/andyreagan/many-happy-languages-appendix}{https://github.com/andyreagan/many-happy-languages-appendix}.
\item Climate change sentiment on Twitter: An unsolicited public opinion poll --- code for generating figures is at \href{https://github.com/andyreagan/climate-change-twitter}{https://github.com/andyreagan/climate-change-twitter} and for performing the keyword search is at \href{https://github.com/andyreagan/climate-change-twitter-keyword-search}{https://github.com/andyreagan/climate-change-twitter-keyword-search}.
\item Reply to Garcia et al.: Common mistakes in measuring frequency dependent word characteristics --- the code for this project is contained in the repository for a previous project, at \href{https://github.com/andyreagan/sentiment-analysis-comparison}{https://github.com/andyreagan/sentiment-analysis-comparison}.
\item The game story space of professional sports: Australian Rules Football --- \href{https://github.com/andyreagan/game-stories-code}{https://github.com/andyreagan/game-stories-code}.
\item The Lexicocalorimeter: Gauging public health through caloric input and output on social media --- the code for the online Appendix for this paper is at \href{https://github.com/andyreagan/lexicocalorimeter-appendix}{https://github.com/andyreagan/lexicocalorimeter-appendix}, code for creating wordshift graphs from the website is at \href{https://github.com/andyreagan/lexicocalorimeter-shifts}{https://github.com/andyreagan/lexicocalorimeter-shifts}, and Panometer website is at \href{https://github.com/andyreagan/panometer.org}{https://github.com/andyreagan/panometer.org}.
\item Tracking the Teletherms: The spatiotemporal dynamics of the hottest and coldest days of the year --- the website code is at \href{https://github.com/andyreagan/teletherm.org}{https://github.com/andyreagan/teletherm.org} and the online appendix code is at \href{https://github.com/andyreagan/teletherms-online-appendices}{https://github.com/andyreagan/teletherms-online-appendices}.
\item Divergent Discourse Between Protests and Counter-Protests: \#BlackLivesMatter and \#AllLivesMatter --- \href{https://github.com/andyreagan/livesmatter-keyword-search}{https://github.com/andyreagan/livesmatter-keyword-search}.
\end{itemize}

\end{document}